\documentclass[a4paper,11pt]{article}
\usepackage{ulem}
\usepackage[english]{babel}
\usepackage{csquotes}
\MakeOuterQuote{"}
\usepackage[utf8x]{inputenc}
\usepackage{amsmath}
\usepackage{graphicx}
\usepackage{mathrsfs}
\usepackage{subfig}
\usepackage{bm}
\usepackage{marvosym}
\usepackage{indentfirst}
\usepackage{wasysym}
\usepackage{stackengine}
\usepackage{soul}
\usepackage{stackrel} 
\usepackage[colorinlistoftodos]{todonotes}
\usepackage{amssymb}
\usepackage{nicefrac} 
\usepackage{clrscode3e}
\usepackage{amsthm}
\usepackage{graphics}
\usepackage[font={footnotesize}]{caption}
\usepackage{dsfont}
\usepackage{mathtools}
\usepackage{relsize}
\usepackage[]{algorithm}
\usepackage{color}
\usepackage{enumitem}
\usepackage{xr-hyper}
\usepackage[hidelinks]{hyperref}
\usepackage{cleveref}
\usepackage{upgreek}
\usepackage{centernot}
\usepackage[makeroom]{cancel}
\usepackage{pgfplots}
\usepackage{euscript}
\usepackage{setspace}
\usepackage[noend]{algpseudocode}
\usepackage{xspace}
\usepackage{mathtools}
\usepackage[english]{babel}
\usepackage{titlesec}
\usepackage[skins,theorems]{tcolorbox}
\tcbset{highlight math style={enhanced,
  colframe=red,colback=white,arc=0pt,boxrule=1pt}}
  
\makeatletter
\newcommand*{\addFileDependency}[1]{
  \typeout{(#1)}
  \@addtofilelist{#1}
  \IfFileExists{#1}{}{\typeout{No file #1.}}
}
\makeatother

\usepackage{geometry}
\usepackage{marginnote}
\usepackage[parfill]{parskip}

\DeclareSymbolFont{bbold}{U}{bbold}{m}{n}
\DeclareSymbolFontAlphabet{\mathbbold}{bbold}
\DeclareMathOperator*{\argmin}{argmin\text{ }} 
\DeclareMathOperator*{\argmax}{argmax\text{ } }

\DeclarePairedDelimiterX{\dotprod}[2]{\langle}{\rangle}{#1, #2}
\DeclarePairedDelimiterX{\bigdotprod}[2]{\big\langle}{\big\rangle}{#1, #2}
\DeclarePairedDelimiterX{\Bigdotprod}[2]{\Big\langle}{\Big\rangle}{#1, #2}
\DeclarePairedDelimiterX{\biggdotprod}[2]{\bigg\langle}{\bigg\rangle}{#1, #2}
\DeclarePairedDelimiterX{\Biggdotprod}[2]{\Bigg\langle}{\Bigg\rangle}{#1, #2}

\makeatletter
\newcommand{\vast}{\bBigg@{4}}
\newcommand{\Vast}{\bBigg@{5}}
\makeatother

\addto{\captionsenglish}{}

\newcommand{\numth}[1]{{#1}^{\text{th}}}
\newcommand{\abs}[1]{ | #1 | }
\newcommand{\absbig}[1]{ \big| #1 \big| }
\newcommand{\absBig}[1]{ \Big| #1 \Big| }
\newcommand{\absbigg}[1]{ \bigg| #1 \bigg| }
\newcommand{\absBigg}[1]{ \Bigg| #1 \Bigg| }
\newcommand{\norm}[2]{ || #1 ||_{#2} }
\newcommand{\bignorm}[2]{ \big|\big| #1 \big|\big|_{#2} }
\newcommand{\Bignorm}[2]{ \Big|\Big| #1 \Big|\Big|_{#2} }
\newcommand{\biggnorm}[2]{ \bigg|\bigg| #1 \bigg|\bigg|_{#2} }
\newcommand{\Biggnorm}[2]{ \Bigg|\Bigg| #1 \Bigg|\Bigg|_{#2} }
\newcommand{\partialDerivative}[1]{\frac{\delta}{\delta{#1}}}
\newcommand{\partialSecondDerivative}[1]{\frac{\delta^2}{\delta{ {#1}^2 }}}

\newcommand{\simiid}{\overset{\text{iid}}{\sim}}

\newcommand{\Q}{\mathcal{Q}}
\newcommand{\Z}{\mathcal{Z}}
\newcommand{\G}{\mathcal{G}}
\newcommand{\D}{\EuScript{D}}
\newcommand{\Jcal}{\mathcal{J}}

\renewcommand{\L}{\EuScript{L}}
\newcommand{\F}{\EuScript{F}}

\newcommand{\X}{\mathcal{X}}
\newcommand{\Y}{\mathcal{Y}}
\newcommand{\M}{\mathcal{M}}
\newcommand{\K}{\mathcal{K}}
\newcommand{\R}{\mathbb{R}}

\newcommand{\goesto}[1]{ \xrightarrow{\enskip #1 \enskip} }

\newcommand{\Var}{\textup{\text{Var}}}
\newcommand{\Cov}{\textup{\text{Cov}}}
\newcommand{\Cor}{\textup{\text{Cor}}}

\newcommand{\I}[1]{\mathbbold{1}{ \{ #1 \} } }
\newcommand{\Ibig}[1]{\mathbbold{1}{ \big\{ #1 \big\} } }
\newcommand{\IBig}[1]{\mathbbold{1}{ \Big\{ #1 \Big\} } }
\newcommand{\Ibigg}[1]{\mathbbold{1}{ \bigg\{ #1 \bigg\} } }
\newcommand{\IBigg}[1]{\mathbbold{1}{ \Bigg\{ #1 \Bigg\} } }

\newcommand{\maxSing}[1]{\sigma_{\textup{max}}\big(#1\big)}
\newcommand{\minSing}[1]{\sigma_{\textup{min}}\big(#1\big)}

\newcommand{\maxSingBig}[1]{\sigma_{\textup{max}}\Big\{#1\Big\}}
\newcommand{\minSingBig}[1]{\sigma_{\textup{min}}\Big\{#1\Big\}}

\newcommand{\maxSingBigg}[1]{\sigma_{\textup{max}}\Bigg\{#1\Bigg\}}

\newcommand{\maxEval}[1]{\lambda_{\textup{max}}\big(#1\big)}
\newcommand{\minEval}[1]{\lambda_{\textup{min}}\big(#1\big)}

\newcommand{\maxEvalBig}[1]{\lambda_{\textup{max}}\Big\{#1\Big\}}
\newcommand{\minEvalBig}[1]{\lambda_{\textup{min}}\Big\{#1\Big\}}

\newcommand{\maxEvalBigg}[1]{\lambda_{\textup{max}}\Bigg\{#1\Bigg\}}
\newcommand{\minEvalBigg}[1]{\lambda_{\textup{min}}\Bigg\{#1\Bigg\}}

  \newcommand{\icol}[1]{
  \left(\begin{smallmatrix}#1\end{smallmatrix}\right)%
}

\newcommand{\E}{\mathbb{E}}
\renewcommand{\P}{\mathbb{P}}

\newcommand{\curlyE}{\mathcal{E}}

\newcommand{\T}{\mathcal{T}}

\newcommand{\N}{\mathcal{N}}
\renewcommand{\H}{\mathcal{H}}
\renewcommand{\S}{\mathcal{S}}
\renewcommand{\D}{\mathcal{D}}
\renewcommand{\R}{\mathcal{R}}
\newcommand{\U}{\mathcal{U}}

\newcommand{\CRLB}[1]{\textup{CRLB}\big(#1\big) }

\newcommand{\FisherInfo}[1]{\mathcal{I}(#1)}

\newcommand{\test}{\textup{\text{te}}}
\newcommand{\train}{\textup{\text{tr}}}

\newcommand{\testI}{\textup{\text{te,I}}}

\newcommand{\trainI}{\textup{\text{tr,I}}}
\newcommand{\trainII}{\textup{\text{tr,II}}}

\newcommand{\allD}{\mathcal{D}}
\newcommand{\allDI}{\mathcal{D}^\textup{\text{I}}}
\newcommand{\allDII}{\mathcal{D}^\textup{\text{II}}}

\newcommand{\Dtest}{\mathcal{D}^\textup{\text{te}}}
\newcommand{\Dtrain}{\mathcal{D}^\textup{\text{tr}}}
\newcommand{\DtestI}{\mathcal{D}^\textup{\text{te,I}}}
\newcommand{\DtestII}{\mathcal{D}^\textup{\text{te,II}}}
\newcommand{\DtrainI}{\mathcal{D}^\textup{\text{tr,I}}}
\newcommand{\DtrainII}{\mathcal{D}^\textup{\text{tr,II}}}

\newcommand{\barallDI}{\bar{\mathcal{D}}^\textup{\text{I}}}
\newcommand{\barallDII}{\bar{\mathcal{D}}^\textup{\text{II}}}

\newcommand{\barDtestI}{\bar{\mathcal{D}}^\textup{\text{te,I}}}
\newcommand{\barDtestII}{\bar{\mathcal{D}}^\textup{\text{te,II}}}
\newcommand{\barDtrainI}{\bar{\mathcal{D}}^\textup{\text{tr,I}}}
\newcommand{\barDtrainII}{\bar{\mathcal{D}}^\textup{\text{tr,II}}}

\newcommand{\eff}{\textup{eff}}

\newcommand{\Tr}{\text{Tr}}

\newcommand{\source}{\S}
\newcommand{\target}{\T}

\newtheorem{theorem}{Theorem}[section]
\newtheorem{corollary}{Corollary}[theorem]

\newtheorem{lemma}[theorem]{Lemma}

\definecolor{linkColor}{rgb}{0.,0.11,0.22}
\definecolor{YaleBlue}{rgb}{0.0,0.22,0.444}

\renewcommand{\arraystretch}{1}

\allowdisplaybreaks

\pgfplotsset{compat=1.15}

\usepackage{natbib}
\setcitestyle{authoryear,open={[},close={]}}

\title{A Semiparametric Efficient Approach To Label Shift Estimation and Quantification}
\author{Brandon Chow}

\setlist[itemize]{leftmargin=\parindent + 30pt}
\begin{document}

%
%
\begin{titlepage}
  
\begin{center}
      Abstract\par\bigskip
      { A Semiparametric Efficient Approach To Label Shift Estimation and Quantification}\par\bigskip
      Brandon Chow \\
      2022 \\
\end{center}
\vspace{0.2in}
Transfer Learning is an area of statistics and machine learning research that seeks answers to the following question: how do we build successful learning algorithms when the data available for training our model is qualitatively \textit{different} from the data we hope the model will perform well on? In this thesis, we focus on a specific area of Transfer Learning called \textit{label shift}, also known as \textit{quantification}. In quantification, the aforementioned discrepancy is isolated to a shift in the distribution of the response variable. In such a setting, accurately inferring the response variable's new distribution is both an important estimation task in its own right and a crucial step for ensuring that the learning algorithm can adapt to the new data. We make two contributions to this field. First, we present a new procedure called SELSE which estimates the shift in the response variable's distribution. Second, we prove that SELSE is semiparametric efficient among a large family of quantification algorithms, i.e., SELSE's normalized error has the smallest possible asymptotic variance matrix compared to any other algorithm in that family. This family includes nearly all existing algorithms, including ACC/PACC quantifiers and maximum likelihood based quantifiers such as EMQ and MLLS. Empirical experiments reveal that SELSE is competitive with, and in many cases outperforms, existing state-of-the-art quantification methods, and that this improvement is especially large when the number of test samples is far greater than the number of train samples.

\end{titlepage}
\newpage

%
%
\thispagestyle{empty}
\textcolor{white}{.}
\vspace{1in}
\begin{center}
A Semiparametric Efficient Approach To Label Shift Estimation and Quantification
\end{center}
\vspace{1.7in}

\begin{center}
	A Dissertation
\end{center}

\begin{center}
	Presented to the Faculty of the Graduate School
\end{center}

\begin{center}
	Of
\end{center}

\begin{center}
	Yale University
\end{center}

\begin{center}
	In Candidacy for the Degree of 
\end{center}

\begin{center}
	Doctor of Philosophy
\end{center}
\vspace{1.7in}

\begin{center}
	By
\end{center}
\begin{center}
	Brandon Chow
\end{center}
\vspace{0.3in}

\begin{center}
	Dissertation Director: Harrison Zhou
\end{center}
\vspace{0.3in}

\begin{center}
	December 2022
\end{center}

\newpage

%
%
\thispagestyle{empty}
\textcolor{white}{.}
\vspace{3in}

\begin{center}
Copyright $\textcopyright$ 2022 by Brandon Chow

All rights reserved. 
\end{center}

\newpage

%
%
\pagenumbering{roman} 
\setcounter{page}{3}
\textcolor{white}{.}
\vspace{3in}
\begin{center}
\textit{To my partner Christina and my parents Shih-Fen and Matthew, whose love and understanding made this work possible. Thank you for always believing in me.}
\end{center}
\newpage

%
%
\begin{center}
\textbf{Acknowledgements}
\end{center}
\vspace{0.2in}

To my PhD buddies Colleen and Soham, thank you for sticking with me. You guys made my time at Yale so much fun, and it's because of you two that I think of New Haven as a place I can call home.\\

To Fredrik, thank you for your support early on in my PhD. You always calmed my nerves with sound and empathetic advice throughout my search for an advisor. To Harry, thank you for being the best  advisor I could have asked for over these past four years.  You invested your time, patience and energy into me, and are always looking out for my best interests. You are the real world's Uncle Iroh.\\

To my parents Shih-Fen and Matthew, thank you for always believing in me, and supporting me through my anxiety and stress to see me to where I am today. Your combined unconditional love is what made me able to do this. \\

To Chappy 4, I love you, and wish you were still here. I'm sorry I never took you on a walk. I wish I could have been a better friend. I hope you have found peace, and are chasing deer wherever you are.\\

Finally, to Christina, the cutest meow-mii who understands me the best. So much of my growth since we met in college is because of you. Thank you for making me a better person, through and through $\heartsuit$. We did it!

\newpage
\tableofcontents
\newpage

\setcounter{page}{1}
\pagenumbering{arabic} 

{\centering \section{Introduction}} \label{sec: Introduction}










A critical assumption in supervised machine learning is that the data available for training a classifier follows the same distribution as the future data points we want to accurately classify. When this assumption holds, classifiers will often generalize well to those future data points. However, when this assumption is violated, the classifier's performance on future data degrades, and can be far inferior to what it was during training. \textbf{Transfer Learning}  is an area of statistics and machine learning research that studies precisely these situations. In Transfer Learning, the data available for training the model is called the  \textbf{source} data, and the data we hope the model will perform well on is called the \textbf{target} data. The discrepancy between source and target data can take a variety of forms, and the overarching goal of Transfer Learning is to find ways to overcome the performance drop caused by these discrepancies.


In this thesis, we study a particular form of discrepancy known as \textit{label shift}, also known as \textit{target shift} or \textit{prior probability shift}. Label shift refers to settings where the marginal distribution of the response potentially differs across domains, but the conditional distribution of the covariates given the response is preserved. In other words, the source and target data have different label distributions, but are otherwise identically distributed. This often occurs in public health applications. For example, suppose that the label variable was a binary indicator for whether an individual had a specific disease, and that the covariates were their physical symptoms, e.g., temperature and heart rate. If we measured the covariates and labels of individuals from two geographically separate populations A and B, we may find that the disease prevalence differs between the two populations. However, the typical body temperatures and heart rates experienced by sick individuals in population A should be similar to those experienced by sick individuals in population B-- after all, geography does not alter the impact of disease on the human body, and we are measuring the same disease in both populations. Likewise, the typical temperature and heart rates of healthy individuals should not change based on population. This situation is an example of label shift, because the label distribution (i.e, disease prevalence) might differ across the two populations, but the conditional distribution of the covariate given the label (i.e., body temperature and heart rate of a sick vs. healthy individual) is the same for both populations.

In the label shift setting, one of the most important tasks is to learn the target label distribution-- the basic idea is that if one can infer the source-target discrepancy, then this information can be used to adapt the classifier so it preforms well in the target domain. Unfortunately, this task is also non-trivial, since while the source data is usually labeled, the target data is usually unlabeled. This lack of labels for the target data makes this problem interesting and, at first glance, quite puzzling to solve: indeed, how can one estimate the distribution of a random variable (i.e., the target labels) without actually observing any samples of that random variable?

Our objective will be to develop a method that overcomes this challenge and provides accurate estimates of the target label distribution. The main highlight of our method is the optimality properties of its estimation error, which we derive by drawing on theoretical tools from semiparametric statistics. With the ultimate goal of sharing those results, we have organized the thesis as follows.  In the Introduction section, we present a high level overview of Transfer Learning and its various subfields. Next, we narrow our focus to the label shift (aka, quantification) literature specifically, and review the existing methodologies contained therein. The second section is the Methodology section. In this section, in order to motivate our own algorithm, we provide the mathematical details behind several of the methods described in the Introduction. We then provide a concise description of our own algorithm, SELSE, and present intuition for why it should outperform existing methodologies. The third section is the  Theoretical Results section. Here, we formalize the intuition developed in the Methodology section and present a precise characterization of SELSE's first and second order error terms. We also introduce crucial ideas and tools from the field of semiparametric statistics, which we use to prove that SELSE is semiparametric efficient. Finally, the fourth section is the Numerical Experiments section, which we use to compare the empirical performance of SELSE against various competitors.




\subsection{Transfer Learning: Homogeneous vs. Heterogeneous}

Roughly speaking, Transfer Learning methods can be broken down into two different categories that depend on the type of source-target discrepancy at hand. The simplest form of discrepancy occurs when the distribution of the source data is different from the distribution of the target data, and methods that deal with this type of discrepancy are known as \textbf{Homogeneous Transfer Learning} methods. The exact manner in which the source and target distributions differ (e.g., which conditionals or marginals change or stay the same) varies by application, but generally, it is assumed that the two distributions are supported on the same covariate and label space. As a motivating example, consider the task of building a spam email detector that is tailored to your personal email account. Ideally, you would use a labeled data set of the past emails you have received. In reality, such a dataset is tedious to obtain, as it would require you to manually annotate a large amount of emails from your inbox. While a viable alternative is to use the abundance of publicly available email datasets that exist online to train your model, the emails contained in such datasets are very likely to be different from your own received emails, both in terms of the distribution of features (e.g., topics and word count) and the fraction of emails that are spam. Thus, the distribution of the target data (emails you receive) and the distribution of the source data (emails in public datasets) are different, and even though you would like your spam classifier to perform well on future samples from the target distribution, you only have samples from the source distribution to train on. 

A more complicated discrepancy occurs when the source and target data have qualitatively \textit{different} feature variables.  These are handled by \textbf{Heterogeneous Transfer Learning} methods. For example, consider the task of building a sentiment classifier that determines whether a given online review of the new iPhone $42$ is positive or negative. Natural Language Processing offers a variety of methods to tackle this problem, and a common theme among those methods is that text samples are treated as bags-of-words, where each unique word constitutes a feature. Now, ideally, the data we train our classifier on should contain all the important words (features) that will be present in the reviews for the iPhone 42. This makes the past reviews for iPhones 1-41 a far better training set than reviews for an unrelated product (e.g., a fishing rod). However, even if we train on the reviews of past iPhones, a very realistic problem is that the iPhone 42 will have new characteristics which its predecessors lack, and these new characteristics will be the central focus of many online reviews. For example, reviews of the iPhone 42  will contain words and phrases such as ``seamless 5G streaming'' and ``amazing 3D hologram'' that are relevant to the sentiment we want to predict, but do not appear in the reviews of past iPhones. Thus, the target data (reviews of iPhone 42) contains features (words and phrases) that are highly relevant to the response variable, and while the source data (reviews of iPhones 1-41) we train on does not contain some of the target features, we'd still like our classifier to be able to productively make use of said features when it encounters them in the target data.  

In both homogeneous and heterogeneous settings, the qualitative discrepancy between source and target data requires us to consider how source knowledge can be \textit{transferred} to a target task. This is a significant departure from the traditional supervised learning perspective, where no source-target discrepancy exists. Indeed, from the traditional perspective, it is not even clear how such a transfer would be carried out. In practice, however, we know that it must be possible to transfer knowledge, due to our observations of human learners. For example, a person who has learned the piano can learn the violin faster than others. Human learners also provide us with insight about the important elements of the knowledge transfer problem; for example, human learners show us that the effectiveness of the transfer depends on how and to what degree the source and target are related-- a person who has learned the viola will learn the violin faster than a person who has only learned the piano. Transfer Learning is a field that draws both inspiration and insight from the human capacity to transfer knowledge across different domains, and is a brave attempt to develop statistical models that are capable of the same feat.

\subsection{Homogenous Transfer Learning}  \label{sec: Transfer Learning}

In homogenous transfer learning, the source and target data share the same feature and response spaces, but may  have different distributions on those spaces. In this section, we discuss methods that handle a specific version of this setting, namely, those methods that assume there is exactly \textit{one} source dataset, and that the source data is labeled but the target data is unlabeled. This problem setting is difficult: our goal is build a classifier that preforms well on data from the target distribution, yet the target samples we have are all unlabeled!

This scope is also a strategic choice that reflects this thesis' intended focus. The rationale for this focus is as follows. First, while the multi-source setting is much more complicated than the single source setting (as we must also consider how the different sources are each individually related to the target task) those methods typically draw inspiration from the single-source setting. As such, research advances in the single-source setting tend to contribute to research advances in the multi-source setting. Second, methods designed for the unlabeled target setting can usually be extended to settings where either a small or even a large portion  of the target data is labeled. This should not be surprising, since having target labels makes the learning task far \textit{easier} than not having any target labels at all.


Homogenous transfer learning methods that assume a single source dataset and can handle unlabeled target data are known as single-source \textbf{Unsupervised Domain Adaptation (UDA)} methods. According to the categorization scheme provided by \cite{Kouw}, methods in single-source UDA fall under three general categories: \textbf{importance weighting}, \textbf{feature-based}, and \textbf{adaptive inference methods}.


\subsubsection{Notation} \label{sec: Transfer Learning Notations}

Before diving into each of the three categories of approaches for single-source UDA, we will need to arm ourselves with some notation.  Let $\X$ denote the covariate space and $\Y$ the response (or label) space. For simplicity, we assume that $\Y$ is discrete. Let $p_{x,y}^\source$ and $p_{x,y}^\target$ denote the densities for the source and target distributions on $\X \times \Y$, respectively. In UDA, we assume we have a labeled source dataset $\{(X_i^\source ,Y_i^\source) \}_{i=1}^{n^\source}$ of $n^\source$ samples drawn IID from $p_{x,y}^\source$, and an unlabeled target dataset  $\{X_i^\target \}_{i=1}^{n^\target}$ of $n^\target$ samples drawn IID from $p_{x}^\target$, i.e., the target density's covariate marginal. A technical, sometimes overlooked, assumption made in  the UDA literature is that $\{(X_i^\source ,Y_i^\source) \}_{i=1}^{n^\source}$ and $\{X_i^\target \}_{i=1}^{n^\target}$ are taken as mutually independent. Also, let $L$ denote some loss function of interest (e.g., $0-1$ loss). Our goal is to construct a function $\widehat{f}$ that minimizes the target risk, $R^\target(f):=  \E_{p_{x,y}^\target}[ L(X,Y, f)]$, where $\widehat{f}$ is taken from a hypothesis space $\F$.

\subsubsection{Importance Weighting Methods} \label{sec: Importance Weighting Methods}


Since $p_{x,y}^\source \neq p_{x,y}^\target$, naively training on the labeled source data may not produce a classifier that performs well on the target data. The central idea behind importance weighting is to determine  \textit{which} source examples contribute positively to classifying the target data, and which contribute negatively. When the support of $p_{x,y}^\target$ is a subset of the support of $p_{x,y}^\source$ (i.e., when the source data is ``richer'' than the target data), this notion of ``important'' source data points  is naturally captured by the ratio
\[
	w(x,y) := \frac{p^\target(x,y)}{p^\source(x,y)}
\]
which is typically called the \textit{importance weighting function}. Note that, due to our assumption about the supports of the source and target densities, the denominator in the display above is always nonzero.  The importance weighting function appears naturally by rewriting the population target risk, in the following simple way:
\begin{align*}
	R^\target(f) &= \E_{p_{x,y}^\target}\Big[ L(X,Y, f) \Big]\\
	 &=  \E_{p_{x,y}^\source}\Bigg[ \frac{p^\target(x,y)}{p^\source(x,y)}   L(X,Y, f) \Bigg] \\
	&=  \E_{p_{x,y}^\source}\Big[  w(X,Y)  L(X,Y, f) \Big] .
\end{align*}

In light of the display above, the strategy taken by importance weighting methods is as follows. First, create an estimate $\widehat{w}(x,y)$ of the importance weight function. Then, perform \textbf{weighted empirical risk minimization} (wERM); i.e., find the $\widehat{f} \in\F$ to minimize 
\begin{equation}\label{garlic}
	\frac{1}{n^\source}\sum_{i=1}^{n^\source} \widehat{w}(X_i^\source,Y_i^\source) L(X_i^\source,Y_i^\source,f),
\end{equation}
and declare $\widehat{f}$ to be the classifier for the target data. There are pros and cons to this approach. The upshot is that if we approximate the true importance weights well, i.e. if $\widehat{w}(x,y) \approx w(x,y)$, then it is easily seen that \eqref{garlic} will be an approximately  \textit{unbiased} estimate of $R^\target(f)$. The downside, however, is that if $\widehat{w}(x,y)$ is very large, then the variance of \eqref{garlic} can be huge. This is problematic, because  if \eqref{garlic} is a poor proxy for $R^\target(f) $, then the excess risk $\R^\target(\widehat{f}) -\inf_{f\in\F}\R^\target(f)$ could be very large. This was first studied formally by \citep{Cortes}, who suggested regularizing $\widehat{w}$ to navigate the bias-variance tradeoff. 

While choosing $\widehat{w}$ to balance the above bias-variance tradeoff is important, there is an even more fundamental issue to be addressed: how can we form an estimate of $w(x,y) \equiv p^\target(x,y) / p^\source(x,y) $ when we only have samples from $p_{x}^\target$ and $p_{x,y}^\source$? Indeed, without any prior knowledge, $w(x,y)$ is clearly unidentifiable: the data we observe contains no information about $p_{y|x}^\target$! For this reason, it is necessary to make assumptions about the relationship between $p_{x,y}^\source$ to $p_{x,y}^\target$. Typically, a practitioner will make one of two different assumptions: covariate shift or label shift.  

The \textbf{covariate shift} assumption is that $p_{y|x}^\target = p_{y|x}^\source$. That is, the source and target distributions agree about the relationship between $X$ and $Y$, but disagree on the marginal distribution of the covariates. Note that assuming equality of $p_{y|x}^\target$ and $p_{y|x}^\source$ directly resolves the foregoing unidentifiability of the importance weights, which now have the following form:
\[
	w(x,y) = \frac{p^\target(x,y)}{p^\source(x,y)} = \frac{p^\target(x) p^\target(y|x)}{p^\source(x) p^\source(y|x)} =  \frac{p^\target(x)}{p^\source(x)}.
\]

As the display above shows, to estimate the importance weights, we need only compare the source and target covariate marginals; no training labels are needed. There are many approaches to estimating these weights. Since we observe data from both $p_{x}^\source$ and $p_{x}^\target$, the naive approach would be to directly estimate both densities, and then set $\widehat{w}$ to be the ratio of the density estimates. However, without making strong parametric assumptions about the form of  $p_{x}^\source$ and $p_{x}^\target$, such an estimator would perform poorly in high dimensions. For this reason, several different approaches to estimating $p_{x}^\target /p_{x}^\source$ have been developed.  The Kernel Mean Matching (KMM) method \citep{Huang} directly estimates the importance function at the source data points by re-weighting them so that the mean of the source and target points in a Reproducing Kernel Hilbert Space (RKHS) are close. When the associated kernel is universal, then on the population level, the weights that are the solution to this alignment problem is exactly the importance weights $p^\target(x) / p^\source(x)$. While KMM has the attractive property of being completely nonparametric while avoiding density estimation entirely, the performance of KMM depends heavily upon hyper parameters for regularization, and crucially, the parameters for the universal kernel being used (e.g., if a gaussian kernel is being used, then the width of the kernel must somehow be chosen). So far, the question of how to choose these tuning parameters remains open. 

Another approach, known as Least Squares Importance Fitting (LSIF) \citep{Kanamori}, takes a different approach by assuming that the true density ratio is a linear combination of basis functions. They then form their estimate of the density ratio by minimizing the integrated squared error. While LSIF differs from KMM in that it is no longer completely non-parametric, it is only parametric in its choice of basis functions, not the actual source or target densities; moreover, LSIF's tuning parameters can all be straightforwardly chosen via cross-validation. \cite{Sugiyama} proposed a similar approach, known as Kullback-Leibler Importance Estimation Procedure (KLEIP). In KLEIP, the density ratio is also modeled as a linear combination of basis functions, but instead of minimizing integrated squared error, KLEIP chooses $\widehat{w}$ to minimize the KL divergence between $p^\target(x)$ and $\widehat{w}(x)p^\source(x)$. While KLEIP is interesting theoretically, \cite{Kanamori} demonstrated that its use of log loss instead of squared loss causes its optimization routine to converge much slower than LSIF's. 

In contrast to the covariate shift assumption that  $p_{y|x}^\target = p_{y|x}^\source$, the \textbf{label shift} assumption is that $p_{x|y}^\target = p_{x|y}^\source$. That is, the source and target distributions agree on the class conditional densities, but disagree on the probability of classes. Under these assumptions, the importance weights reduce to 
\begin{equation} \label{importance weight equation for label shift}
	w(x,y) = \frac{p^\target(x,y)}{p^\source(x,y)} = \frac{p^\target(y) p^\target(x|y)}{p^\source(y) p^\source(x|y)} = \frac{p^\target(y)}{p^\source(y)}, 
\end{equation}
a function of only $y$. Since the main focus of this thesis is in the label shift setting, we save our overview of label shift for section \eqref{sec: Label Shift / Quantification}.

\subsubsection{Feature-Based Methods}

Importance weighting methods require the target covariate distribution's support to  be a subset of the source covariate distribution's support. However, in many problems, this assumption may not hold. Feature-based methods are an area of Transfer Learning that does not make assumptions about the source and target supports. These methods appear most commonly in Computer Vision applications. For such methods, one assumes that there exists a transformation that acts as a map between source and target data, i.e., that there exists a way to transform  the source and target data so that their resulting distributions are completely aligned. The main idea behind these methods is that, once we find the aforementioned transformation and align the source and target distributions, we have come back to the classical setting of supervised learning where the source-target discrepancy is nil. As such, a classifier trained on the transformed labeled source data can then be expected to generalize well to the transformed unlabeled target data. Feature-based approaches can be roughly broken down into four different categories: Subspace Mappings, Optimal Transport, and Deep Domain Adaptation methods.

In \textbf{Subspace Mapping} methods, intermediate subspaces are used to model the shift between the source and target distributions. A simple but important example of this was proposed by \cite{Fernando}. In their method, PCA is applied to the source and target covariates separately, and then a linear map between the two PCA-induced subspaces is learned. Once the two PCA-induced subspaces and a linear map between them is learned, the following protocol is used to align the source and target covariates: first, both source and target data get projected onto their respective PCA-induced subspaces; second, the learned linear map is applied to the source data's projections, mapping the projections into the target's subspace. A classifier is the trained on the transformed  labeled source data. Finally, target predictions are obtained by feeding the learned classifier the  transformed, unlabeled target data. The idea proposed by \cite{Fernando} appears consistently in more complicated subspace mapping approaches. For example, instead of using only two subspaces to connect the source and target spaces, \cite{Gong} proposed using a geodesic flow kernel to model a continuous path of intermediate subspaces. Their modeling approach is interesting, because it implies that each point along that path corresponds to a different level of dataset shift. 

Regardless of the particular method, however, all subspace mapping approaches  tend to rely on the assumption that aligning source and target covariate distributions is \textit{equivalent} to aligning the source and target joint (covariate-response) distributions, which is only true if one believes that $p_{y|x}^\target = p_{y|x}^\source$. As such, one can think of subspace mapping methods as ``fancier'' methods for performing covariate shift adaptation when the target data's support is not contained in the source data's supports. Another important theme that must be wrestled with when developing subspace mapping methods (and, indeed, feature-based methods in general) is that the transformation must be chosen so that, when the transformed source data is used to build a classifier, that classifier still has good performance on said transformed source data. To see why this might be important, consider the extreme example whereby the transformation chosen is just a constant function. In this case, the transformed source and transformed target covariate distributions are perfectly aligned, yet the transformed source covariates are now useless for discriminating between the classes. Thus, in addition to choosing the transformation function that aligns the source and target covariate distributions, one must \textit{also} ensure that the transformation is such that the information contained in the source covariates about the response is not lost. Overall, these themes were well captured by \cite{Ben-David}, who developed generalization bounds showing that minimizing the target risk under the covariate shift assumption required choosing a transformation function that would balance a tradeoff between (a) aligning the source and target covariate distributions while (b) preserving the  information contained in the source covariates about the response. 

\textbf{Optimal Transport} methods assume that the shift between the source and target distributions is induced by a function $\phi(x)$. That is, if $(X,Y) \sim p_{x,y}^\source$, then $(\phi(X),Y) \sim p_{x,y}^\target$; i.e., a sample from the joint source distribution whose covariate has been transformed by $\phi$ has the same joint distribution as the target data. Consequentially, $p_{y|x}^\target \neq p_{y|x}^\source$ and $p_{x}^\target \neq p_{x}^\source$. This is far more flexible than the covariate shift regime that Subspace Mapping methods are confined to. However, even though the Optimal Transport regime implies distributional shifts in both $X$ and $Y|X$, this shift is not completely flexible, as we still have $p_{y}^\target = p_{y}^\source$. Furthermore, even though the functional relationship between $Y$ and $X$ have changed between source and target (i.e., $p_{y|x}^\target \neq p_{y|x}^\source$), this shift is homogenous across different classes $y\in\Y$, as $\phi$ is a function of only $x$.  In addition, Optimal Transport methods also  impose a very strong restriction on the search space of possible transformation functions: namely, $\phi$ is taken to be the deterministic coupling (with marginals $X^\source \sim p_{x}^\source$ and $X^\target \sim p_{x}^\target$) that minimizes Wasserstein Distance. Intuitively, this means that $\phi$ is the coupling of $X^\source$ and $X^\target$ such that a distance metric $d(X^\source,X^\target)$ is small on average. While this area of transfer learning has lots of encouraging practical results (particularly with Computer Vision datasets), Optimal Transport methods have far fewer theoretical guarantees than their Subspace Mapping and Importance Weighting competitors. One exception to this general trend is a paper by \cite{Courty}. Utilizing Euclidean distance as their distance metric, \cite{Courty} proved the recovery of $\phi$ when $\phi(x) = Ax + b$, i.e., when $\phi$ was a linear model. In their optimization routine, the authors also propose penalizing those candidate $\phi$ functions that strongly couple source examples from different classes together with the same target samples. While they did not provide any theoretical proof for this approach, their empirical results seem to, at least in part, be due to this novel practical insight. 

Finally, \textbf{Deep Domain Adaptation Methods} comprise the final category of Feature based approaches. In the literature, there are two types of networks that are often considered: Domain-Adversarial Networks and Autoencoders. Domain-Adversarial Neural Networks (DANN) are used to find a function $\phi$ such that (a) source and target examples cannot be distinguished, and (b) the classification error on the transformed source data is minimized \citep{GaninAdversarial}. Note that is exactly the same tradeoff that Subspace Mapping methods must balance, and thus the theoretical justification for DANN's is heavily grounded in the generalization error bounds developed by \cite{Ben-David}. To achieve both goals (a) and (b), DANN's employ two loss layers: one to classify source samples based on their labels, and another layer to classify samples as belonging to either the source or target dataset. By using this architecture, DANN's are thus trained to simultaneously find a representation of the source and target data that makes the two indistinguishable, while also building a classifier that is able to perform well when using the foregoing representation of the source data.  A practical limitation to DANNs are that the  gradients for the two loss layers often point in different directions \citep{KouwReview}, which makes sense in light of our previous discussion of how goals (a) and (b) are at odds. This has been ameliorated to some extent by subjecting the hypothesis space to constraints such as, for example, the \textit{cluster assumption}, which requires decision boundaries not to cross high-density data regions \citep{ShuDirtT}. Finally, autoencoders are another popular type of network used for domain adaptation. Typically, a \textit{single} autoencoder is trained to be able to reconstruct \textit{both} source and target data examples, and the function $\phi$ is simply the encoding layer of the network. However, unlike ordinary autoencoders, the distribution of the encoded source instances and encoded target instances are encouraged to be similar to each other, usually through a penalty such as KL divergence. Much like with DANNs, an additional penalty for the classification loss obtained by a classifier trained on the encoded source instances is also included \citep{ZhuangAutoencoder}.

\subsubsection{Adaptive Inference Methods}

Both importance weighting and feature-based methods take a similar, two-step approach to UDA: first, the discrepancy between source and target data is resolved, and second, a classifier is trained based on either a re-weighting (in the case of importance weighting) or transformation (in the case of feature-based methods) of the labeled source examples. While in the second step this allows users to utilize off-the-shelf supervised learning methods that may work well for their particular datasets, the two-step procedure also has extra overhead, since it is the actual final classifier, \textit{not} the true importance weights or true transformation function, that is of ultimate interest. This overhead can also become particularly wasteful, since some UDA methods (e.g., BBSC \citep{LiptonBBSE}) require dividing the source data into two parts: one for resolving the source-target discrepancy, and another for training the actual classifier. 

Adaptive Inference methods differ from the above two-step approach by explicitly incorporating the adaption step into the inference procedure. In this sense, they entirely drop the extra overhead.  One type of adaptive inference approach that is particularly interesting is the \textbf{minimax} based approach. Since we have labeled source samples and unlabeled target samples, the only distribution we lack information on is $p_{y|x}^\target$. This makes it very natural to choose a hypothesis $f \in \F$ such that the worst possible risk (where each ``possible''  risk corresponds to a different possibility of what $p_{y|x}^\target$ is) is minimized.  A straightforward example of this is the Robust Bias-Aware classifier \citep{LiuMinimax}. In their approach, the set of plausible target posteriors $\{ \widetilde{p}_{y|x}^\target \}$ are those that satisfy a moment constraint:  $\E_{p_{X,Y}^\source}[u(X,Y)] = \E_{p_{X}^\target \widetilde{p}_{Y|X}^\target }[u(X,Y)]$ for some function $u(x,y)$,. The moment constraint allows them to obtain a closed form solution for their minimax classifier. Another minimax-based approach is the Target Contrastive Robust Risk Estimator \citep{Kouw}. Taking a slightly different angle, they focus on choosing a classifier such that, when compared to the naive source classifier, the worst possible \textit{increase} in test risk is as small as possible. In other words, if $\hat{f}^\source \in \F$ is the classifier obtained by simply training on the source data without regard for any potential source-target discrepancy, then their classifier is given by
\[
	\hat{f}^\target := \argmin_{f\in\F}\max_{\widetilde{p}_{y|x}^\target} \Bigg[ R^\target\Big(f ,\widetilde{p}_{y|x}^\target\Big) - R^\target\Big( \hat{f}^\source ,\widetilde{p}_{y|x}^\target\Big)  \Bigg].
\]
This approach is interesting, in that it attempts to choose classifiers that would not do \textit{worse} than if we had not bothered with Transfer Learning at all. However, as \cite{Kouw} point out, the downside of their approach is that if the source-target discrepancy is large enough, then $\hat{f}^\source$ would have been a poor classifier for the target data anyway, and so if $\hat{f}^\target$ is chosen as above, then $\hat{f}^\target$ may also be a poor classifier as well (albeit marginally better than  $\hat{f}^\source$).

\subsection{Label Shift / Quantification} \label{sec: Label Shift / Quantification}

Having provided an overview of UDA, we now turn to the specific subfield of UDA that will be the focus of this thesis: \textit{label shift} estimation, also known as \textit{quantification}. We also choose to focus on the special case where the labels are discrete, which happens to also be the situation that is studied the most. As mentioned in section \eqref{sec: Importance Weighting Methods}, label shift refers to settings where the marginal distribution of the response potentially differs across domains, but the conditional distribution of the covariates given the response is preserved. Like all other settings in UDA, unless extra measures are taken, this source-target discrepancy can negatively impact classifiers that were trained on source data: in particular, if the discrepancy is large enough, then their performance on target data will be especially poor. As an extreme example in binary classification, suppose that the source data had 99\% positive samples and 1\% negative samples, and that the target data had the reverse, i.e., 1\% positive samples and 99\% negative samples. Clearly, traditional classifiers fit on the source data would perform poorly on the target data, simply because they would almost always make positive predictions. 

As was also argued in section \eqref{sec: Importance Weighting Methods}, one way to rectify this problem is to instead fit classifiers via wERM, which weights the source samples in a fashion that adjusts for the source-target discrepancy: samples that are rare in the source domain but common in the target domain are up-weighted, and samples that are common in the source domain but rare in the target domain are down-weighted. In this fashion, classifiers fit via wERM on labeled source data are "adapted" to the target domain, resulting in improved target predictions.   


\subsubsection{Fragmented Literature}

The fact that wERM can adapt classifiers in the face of the source-target discrepancy makes learning the correct weights for the source samples extremely important. Fortunately, as shown in display \eqref{importance weight equation for label shift}, it also happens that these weights take on a particularly simple form for the label shift setting: namely, the weights are the ratio between the target and source response densities. 

Now, one would think that the next natural step in our discussion would be to review the methods for estimating the aforementioned weights. However, to do so effectively, we first must take a small detour, and point out that this area of research is actually quite fragmented: at the time of writing this thesis,  there exists two different "camps" of research that, despite tackling nearly the same problem, appear to be largely unaware of each other! 

So far, we have presented ideas in the same framework utilized by the first camp: this camp refers to our problem of interest as "label shift", and views it as a subcategory of Transfer Learning. However, in the second camp, our problem takes on an entirely different name: \textbf{quantification}. The main difference between the "label shift" camp and the "quantification" camp is that, while the former is interested in building classifiers that perform well in the target domain, the latter is only interested in quantifying the shift between the source and target distributions. Thus, while the label shift camp is focused on learning the ratio between the target and source response densities for the sake of performing wERM, the quantification camp is only concerned with estimating the distribution of the response in the target domain. That is, in mathematical terms, the label shift camp is interested in learning $w(x,y) = \frac{p^\target(y)}{p^\source(y)}$, but the quantification camp is only interested in learning $p^\target(y)$. 

Of course, these two ventures are related: after one estimates $p^\target(y)$, one can then use the source data's labels to form a straightforward estimate of $p^\source(y)$, and then take the ratio of the two to form a plug-in estimator of $\frac{p^\target(y)}{p^\source(y)}$. Conversely, if one first estimated $\frac{p^\target(y)}{p^\source(y)}$  but was actually interested in $p^\target(y)$, an estimate of $p^\target(y)$ can be easily obtained by multiplying the estimate of $\frac{p^\target(y)}{p^\source(y)}$  by an estimate of $p^\source(y)$. Despite the foregoing similarities, however, these two camps have nonetheless been disconnected, a fact recently pointed out by \cite{QuantificationReview2017}.

Given these striking similarities, from here on out, we make the strategic choice to frame all ideas from the quantification camp's perspective. Up until now, the main benefit of introducing our problem from the label shift camp was that it allowed us to situate our problem within the larger context of transfer learning. Since we have already done this, there is no active reason to continue using the label shift framework. However, there \textit{is} a proactive reason to use the quantification framework: namely, the quantification literature is significantly more developed than its label shift counterpart. Indeed, while the first formal definition of quantification  was stated by \cite{Forman2005}, the problem was being discussed long before by \cite{Saerens2002} and by \cite{gart1966comparison}. In contrast, the label shift literature is far less developed, and recently, \cite{LiptonBBSE} even noted that the label shift problem has been "curiously under-explored". Thus, overall, the main benefit of adopting the quantification camp's framework is that it will facilitate our discussion of existing methodologies, since most of those methods had been developed with quantification in mind.

\subsubsection{Existing Quantifiers} \label{sec: Overview of Existing Quantifiers}

We now provide a high level overview of existing quantification methods, also known as \textbf{quantifiers}  \citep{Forman2005}. To be consistent with the literature from which these quantifiers come from, for the remainder of this thesis, we refer to data from the \textit{source} domain as the \textbf{train set} and data from the \textit{target} domain as the \textbf{test set}. All other necessary notational and terminological changes will be saved for section \eqref{sec: Quantification Notation}.  

In the quantification literature, the majority of quantifiers can be categorized as improvements to the \textbf{Classify \& Count (CC)} approach proposed by \cite{Forman2008}. The approach is as follows: fit a hard\footnote{A "hard" classifier is one whose outputs are class labels; a "soft" classifier is one whose outputs are probabilities.} classifier on the train set, and use it to make predictions on the test set. Then, the estimated distribution of class labels is simply the distribution of test predictions. The exact learning algorithm used to fit this classifier could be anything (e.g., SVM, gradient boosted trees or a deep neural network), but is usually selected based on the kind of data at hand. This approach is intuitive because it expresses the idea that if the test labels were available, then learning their distribution would be straightforward, and so because the test labels are in reality \textit{un}available, we should first predict them and then proceed as if those predictions were correct. 

However, this reliance on the accuracy of a classifier's test predictions causes the CC approach to suffer from a Catch-22 dilemma. As highlighted in the example at the beginning of section \eqref{sec: Label Shift / Quantification}, obtaining a classifier that makes accurate test predictions requires one to account for the shift in the label distribution. However, to learn about the shift in the label distribution, the CC approach requires a classifier that can make accurate test predictions! From this perspective, the CC approach is an ineffective way to tackle quantification, because the very bias that it is trying to overcome is also what causes it to fail. Of course, a notable exception to this occurs when the problem at hand involves classes with perfectly separated covariate distributions. In this special situation, if the number of train samples is large enough, then it may be possible to learn this separation boundary via the train set; then, since this separation boundary also holds for the test set, perfect test predictions would be achieved. However, in virtually all real-world applications, the classes are never perfectly separated, so this insight offers little redemption for the CC approach. Other methods nearly identical to the CC approach also exist, but they too are subject to the same biases. For example, the Probability Average or \textbf{Probabilistic Classify \& Count (PCC)} approach proposed by \cite{BellaProbabilistic2010} differs only in that it uses a soft classifier in lieu of a hard one. Specifically, while for the CC approach the estimated test probability for class $y$ is simply the fraction of times the hard classifier predicted test samples to belong to class $y$, for the PCC approach, the estimated probability for class $y$ is the predicted \textit{probability} that a sample belongs to class $y$, averaged over each of the samples in the test set.  

While this limitation may be damning for the CC approach, the CC approach was still impactful, because it inspired several variants that escape its Catch-22 dilemma. The most prominent example of this is \textbf{Adjusted Classify \& Count (ACC)} quantifier proposed by \cite{Forman2005, Forman2008}. The ACC approach is a modification to the CC approach: after computing the CC quantifier, one adjusts the estimate by accounting for the misclassifications that the underlying classifier had made on the training set. For example, in the binary class case, the ACC quantifier is a simple function of the original CC quantifier, and the classifier's true and false positive rates on the train set. Compared to the CC quantifier, the ACC quantifier has the benefit of being a consistent estimator of the test label distribution. While the exact mathematical logic for this will be presented in section \eqref{sec: More Details about Existing Quantifiers}, the intuition is straightforward, and can be discussed here. Put simply, the CC quantifier suffers from misclassifications on the test set, but to adjust its estimate to match the truth, one does not need to know exactly which test samples were misclassified. Rather, only the misclassification \textit{rates} for each class (i.e., the class-normalized confusion matrix) are necessary. Fortunately, under the label shift assumption, these rates are identical within each class across training and testing, at least on the population level \citep{Fawcett2005}. Thus, even though the test labels are unobserved,  the misclassification rates on the test set can nonetheless be estimated for each class via the train data, enabling adjustments to the CC quantifier that yield an accurate estimate of the test label distribution. Other spins on ACC also exist, and inherit the same consistency properties enjoyed by ACC.  For instance, just as PCC is the variant of CC that uses a soft classifier, the \textbf{Probabilistic Adjusted Classify \& Count (PACC)}\footnote{Actually, \cite{BellaProbabilistic2010} refer to their quantifier as Scaled Probability Average (SPA), but we refer to it as PACC to highlight the connection to ACC, and mirror the relationship between PCC and CC.} quantifier proposed by \cite{BellaProbabilistic2010} is the soft-classifier variant of ACC. 

Due to the disconnect between the label shift and quantification camps, it is important to note that there has been some overlapping progress on the ACC / PACC line of thought. In particular, in the label shift camp, \cite{LiptonBBSE} recently proposed an approach called \textbf{Black Box Shift Estimation (BBSE)}. This approach is essentially identical to the ACC and PACC approach: the train data is used to fit a classifier (either hard or soft) and estimate the classifier's confusion matrix, and then the confusion matrix is used to adjust the distribution of the classifier's test predictions, resulting in the final estimate. However, as \cite{Azizzadenesheli} pointed out, BBSE suffers when classes that are rare in the train data become more frequent in the test data, i.e., when the label shift is very large. For this reason, \cite{Azizzadenesheli} expanded upon BBSC by proposing to regularize the estimate of the test label distribution. Specifically, they develop a high probability upper bound, and then minimize that upper bound to choose an appropriate regularization parameter.

Since BBSE can be thought of as a simple "rebranding" of ACC and PACC, the critique put forward by \cite{Azizzadenesheli} applies to ACC and PACC as well. The increased estimation error also makes sense: as mentioned earlier, ACC and PACC overcome the biases that plague CC and PCC via estimating the class-wise misclassification rate for each class. In the aforementioned situation, estimates for these rates can have high variance, simply because there are so few examples from certain classes (e.g., estimating the false positive rate of a hard classifier when there are very few negative training samples is difficult). If those classes suddenly become very prevalent in the test set, then those misclassification rates will play a huge role in updating the original CC/PCC quantifier; thus, the ACC/PACC adjustment will be sensitive to estimation errors for said rates \citep{QuantificationReview2017}. While the promise of consistency still holds in these extreme settings, the larger error should not be ignored.

This concern was noted long before by the quantification community \citep{Forman2006}. However, instead of using regularization like \cite{Azizzadenesheli} had done, some quantification researchers have taken a different route: they modify the classifier that ACC and PACC use. For example, as pointed out by \cite{QuantificationReview2017} in the binary class setting, the magnitude of the adjustment to CC and PCC is inversely related to the difference between the classifier's true and false positive rates: according to them, the larger this difference, the lower the ACC or PACC error. For this reason, several attempts have been made to choose classifiers for ACC so that the foregoing difference is large. For instance, \citep{Forman2006, Forman2008} propose using a SVM for the ACC quantifiers, and then selecting the SVM's threshold according to different criteria. One criteria they propose directly addresses the foregoing concern by choosing the threshold to maximize the difference between the classifier's true and false positive rates on the train set. Another approach they consider is to first compute the ACC quantifier for a grid of different SVM thresholds, and then take the median of those quantifiers as their final estimate. In a separate vein of research, \cite{quantNearestNeighbors} propose using a weighted  $k$-nearest neighbors ($k$-NN) as the underlying classifier. In their approach, the $k$-NN model uses a suite of  weighting policies on nearby points when generating predictions. The goal of these policies is to offset the classifier's bias against the rare classes in the training data, in anticipation that those classes may become more prevalent in the test data. 

While ACC and PACC are the most widely used CC variants, there are several other approaches that do not fall under the CC tradition, yet are popular enough to deserve mentioning. A simple example of this is the Hellinger Distance  \textbf{HDy} approach proposed by \cite{HellingerDistanceQuantification}. The idea behind the HDy approach is to view quantification as a statistical \textbf{mixture} problem: since under the label shift assumption the conditional distribution of the covariate given the label is the \textit{same} for both train and test populations, the goal of quantification can be seen as replacing the train label marginal with a new label distribution, so as to induce a new train covariate marginal which \textit{resembles} the observed test covariate marginal. In the terminology of statistical mixtures, the discrete "mixing distribution" is the new label distribution, and the "mixture" is the induced train covariate marginal. The label distribution that achieves the aforementioned resemblance is then a good estimate of the test label marginal. Of course, in practice, aligning two covariate distributions is difficult, particularly in high-dimensional settings. For this reason, \cite{HellingerDistanceQuantification} instead performs distributional matching on a low-dimensional function of the covariates, namely the output of a trained classifier. The distributional matching is then carried out by choosing the label distribution which minimizes the Hellinger distance between the induced train covariate marginal and the observed test covariate marginal. 

Another set of approaches exist which are based on the principle of maximum likelihood. The first of these approaches is the \textbf{Expectation-Maximization Quantifier (EMQ)}, proposed by \cite{Saerens}. Originally, \cite{Saerens} had designed EMQ for the purpose of adapting a classifier fit on the train data to the test data, and thus, EMQ should be seen as having originated from the label shift camp. However, as noted by \citep{QuantificationReview2017}, EMQ has become quite popular within the quantification community, where its focus has shifted to estimating the distribution of test labels. With this in mind and in light of our objectives, we choose to focus on this latter interpretation of EMQ. 

One way to understand EMQ is as follows. First, note that if the conditional distribution of the covariate given the label was known for each label, then we would find ourselves back in a classical statistical mixture problem: our goal would be to find a convex combination of those conditional densities, so that the resulting mixture density is the same as what had generated our test data. The classic way to solve this would be to use the well known Expectation-Maximization (EM) algorithm \citep{OriginalPaperEMalgorithm}, which would return the \textbf{maximum likelihood estimate (MLE)}. However, the problem with this approach is that, in quantification tasks, the conditional density of the the covariate given the label is unknown for each class. The idea behind EMQ is to address this problem by first fitting a soft classifier to the training data, yielding an estimate of the conditional probability of the label given the covariate. This estimate can then be inverted via Bayes rule, yielding an estimate of the conditional density of the covariate given the label. EMQ then runs the EM algorithm assuming that these estimated conditional densities are, in fact, the true ones. The output of EM is then returned as the output of EMQ.

The second maximum likelihood approach comes from the label shift camp and is called \textbf{Maximum Likelihood Label Shift (MLLS)}, and it is a variation of EMQ. Originally, the method was proposed by \cite{alexandari2020maximum}, but it was later rebranded as MLLS by \cite{gargMLLS2020}, where its theoretical properties (including consistency) were established. There are several differences between MLLS and EMQ. First, as pointed out by many authors \citep{Plessis2012SemiSupervisedLO, alexandari2020maximum, gargMLLS2020}, the likelihood function being maximized by EMQ is concave, and so vanilla optimization routes can be used to compute the MLE in lieu of EM. Second, and most importantly, MLLS replaces the soft classifier used in EMQ with a calibrated version of it. The calibration used is called \textbf{Bias-Corrected Temperature Scaling (BCTS)} \citep{alexandari2020maximum}, and is done on held-out train data after the soft classifier is fit.

The reasoning behind the calibration is interesting, as it addresses a natural concern with EMQ: namely, what happens if the soft classifier used in EMQ is a poor estimate of the probability of the label given the covariate? This is likely to occur in modern machine learning, since many popular soft classifiers such as neural networks are poorly calibrated \citep{Guo2017OnCO}. When this happens, EMQ cannot be said to be performing any approximate version of maximum likelihood estimation. However, as shall be shown in section   \eqref{sec: More Details about Existing Quantifiers}, applying post-hoc calibration to the soft classifier partially fixes this issue. Specifically, it can be shown that MLLS is computing the MLE for a lower dimensional \textit{function} of the observed covariates; that is, the MLE that one would obtain if, instead of receiving the test samples, one only had access to the outputs of some function of the test samples. While using a transformation of the data can only reduce the amount of information that the data contains about the estimand, and thus the MLE corresponding to the original data should have lower asymptotic variance than the MLE for the transformed data, this sacrifice is worth it from a finite sample perspective, because while EMQ might struggle to even approximate the original data's MLE, MLLS is able to approximate the transformed data's MLE relatively well.

\subsubsection{Our Contribution} \label{sec: Our Contribution}


We make several contributions to the quantification and label shift fields. Our first contribution is a new quantifier that falls in line with the ACC/PACC research vein and is called \textbf{Semiparametric Efficient Label Shift Estimation (SELSE)}. Crucially, we prove that SELSE has the smallest asymptotic variance matrix among all possible ACC/PACC quantifiers. The key reason we are able to accomplish this is because we abandon the notion that ACC/PACC quantifiers need to be based on classifiers; instead, we replace the underlying classifier with a special function, which is described in the Methodology Section. We also provide intuition and insight as to why this function minimizes the asymptotic variance.

Second, to universalize the asymptotic optimality of SELSE from solely CC variants to quantifiers in general, we establish a lower bound on the asymptotic variance matrix of quantifiers belonging to a large family which includes SELSE. To do this, we frame the task of quantification as a semiparametric estimation problem, wherein the finite-dimensional parameter of interest is the test distribution of classes, and the infinite-dimensional nuisance parameters are each class's covariate distribution. We also prove that the asymptotic variance matrix of SELSE is equal to the foregoing lower bound. This establishes that SELSE has the smallest possible asymptotic variance matrix in the family (i.e., SELSE is semiparametric efficient).


{\centering \section{Methodology}}  \label{sec: Methodology Section}

The purpose of this section is to present our methodology. The section is organized as follows. First, to better align ourselves with the quantification literature, we update several of the notations presented in section \eqref{sec: Transfer Learning Notations}. Second, with the goal of motivating our general approach, we take a deeper dive into the methods introduced in section \eqref{sec: Overview of Existing Quantifiers} which bear the most relevance to our own work. Third, we present a concise description of our method. Fourth, we provide both intuition and theoretical justifications for the exact steps our algorithm takes, and make several comparisons of our method to maximum likelihood, EMQ and MLLS. Finally, we present any auxiliary lemmas and proofs that had been referenced by the previous subsections.\\

\subsection{Notation} \label{sec: Quantification Notation}

Let $\X$ denote the covariate space and $\Y := \{0,1,\dots,m\}$ the label space, where $m\geq 1$ is an integer. Let $[m]:= \Y/\{0\}$. The training set consists of $n^\train$ covariate-label pairs $(X_{i}^\train,Y_{i}^\train)$ and the test set consists of $n^\test$ covariate points $X_i^\test$. Each test point has a hidden label $Y_i^\test$, which is not included in the test set. Furthermore, each $(X_i^\train,Y_i^\train)$ is an IID copy of a random variable pair $(X^\train,Y^\train) \sim p_{X,Y}^\train$, and each $(X_i^\test, Y_i^\test)$ is an IID copy of a random variable pair $(X^\test,Y^\test) \sim p_{X,Y}^\test$, where $p_{X,Y}^\train$ and $p_{X,Y}^\test$ are densities supported on $\X\times\Y$.  For simplicity, we will write $\pi_y^\train \equiv \P[Y^\train = y]$ and $\pi_y^\ast \equiv \P[Y^\test = y]$ for each $y\in\Y$, and let $\pi^\train = (\pi_1^\train ,\dots,\pi_m^\train)$ and $\pi^\ast = (\pi_1^\ast, \dots, \pi_m^\ast)$. Since probabilities sum to $1$, knowing  $\pi^\ast$ is equivalent to knowing $p_{Y}^\test$ (i.e., knowing $\pi^\ast$ implies knowing $\pi_0^\ast$), and so the goal of quantification can be said to accurately estimate $\pi^\ast$. 

Furthermore, under the label shift assumption, the conditional distributions of $X^\train \mid Y^\train=y$ and $X^\test \mid Y^\test =y$ are the same for each $y\in\Y$. As such, for each class $y$, we will use $p_y:\X \mapsto\mathbb{R}$  to refer to the single density function shared by both distributions. Implicitly, we assume that each $p_y$ is dominated by the same measure $\mu$ (e.g., counting or Lebesgue measure). In addition, for each class $y\in\Y$ and function $f$ with domain $\X$, the conditional expectation and variance of $f(X)$ when $X \sim p_y$ will be denoted as $\E_y[f]$ and $\Var_y[f]$, respectively. We also define $\mathbf{p} := (p_0,\dots, p_m)$ to be the vector of the component densities.

Moreover, we will refer to the set 
\begin{equation}\label{definition of m plus 1 dimensional probability simplex}
	\bigg\{ \beta \ \bigg\lvert\  \beta=(\beta_1,\dots,\beta_m),  \quad \sum_{y=1}^m \beta_y \leq 1, \quad   \beta_y \geq 0 \ \forall y\in[m] \bigg\}
\end{equation}
as the $m+1$ \textit{dimensional probability simplex}, and for each $\beta$ in this set, we adopt the shorthand $\beta_0:=1-\sum_{y=1}^m \beta_y$. We refer to $p_{\beta} := \sum_{y=0}^m \beta_y p_y$  as the $\beta$-\textit{mixture} of the component densities $p_y$, and denote the marginal expectation and variance of $f(X)$ when $X \sim p_{\beta}$ by $\E_\beta[f]$ and $\Var_\beta[f]$, respectively.



\subsection{Related Quantifiers: Further Details} \label{sec: More Details about Existing Quantifiers}

Of the quantifiers presented in subsection \eqref{sec: Overview of Existing Quantifiers}, there are two particular kinds which are relevant to our own work: CC variants and maximum likelihood approaches (i.e., EMQ and MLLS).

\subsubsection{CC Quantifiers and Variants} 

We begin by presenting a mathematical description of CC quantifiers and the variants introduced in section \eqref{sec: Overview of Existing Quantifiers}. Let $f$ denote a classifier. For simplicity, we start with the binary class case of $m=1$, so that $f: \X \mapsto \mathbb{R}$; we also pretend that $f$ is a fixed function, even though in practice it is always selected via a held-out subset of the train data.

If $f$ is a hard classifier, then the empirical counterpart to $\E_{\pi^\ast}[f]$ computed on the test data is the CC quantifier; if $f$ is a soft classifier, then the empirical counterpart to $\E_{\pi^\ast}[f]$ computed on the test data is the PCC quantifier. Recall that $\E_{\pi^\ast}[f] = \sum_{y=0}^m \pi_y^\ast \E_y[f]$, as  defined in subsection \eqref{sec: Quantification Notation}. Given these CC and PCC quantifiers, obtaining their respective ACC and PACC counterparts is also relatively straightforward. To derive them, note that on the population level, it holds that
\[
	\E_{\pi^\ast}[f] = (1-\pi^\ast)\E_0[f] + \pi^\ast \E_1[f],
\]
and so, after a bit of algebra, we can find the following closed-form solution for $\pi^\ast$
\begin{equation} \label{AC basic formula}
	\pi^\ast = \frac{\E_{\pi^\ast}[f] - \E_0[f] }{ \E_1[f] - \E_0[f] },
\end{equation}
assuming of course that $\E_1[f] - \E_0[f] \neq 0$. This assumption is reasonable, because if $f$ is any non-trivial classifier, then the outputs of $f$ when $X \sim p_1$ should tend to be different from the outputs when $X \sim p_0$. When $f$ is a hard classifier, the ACC quantifier is simply the empirical counterpart to the display above: $\E_{\pi^\ast}[f]$ is replaced by  the CC quantifier, and $\E_0[f]$ and $\E_1[f]$ are approximated via sample averages computed on the train data. The PACC quantifier is computed analogously when $f$ is a soft classifier. For both ACC and PACC, the modifications made to the CC and PCC quantifiers represents the adjustment mentioned in section \eqref{sec: Overview of Existing Quantifiers}, with $\E_1[f]$ and $\E_0[f]$ equaling the population-level true and false positive rates, respectively. The display above also highlights the point we made in section \eqref{sec: Overview of Existing Quantifiers} that ACC and PACC can be sensitive to estimation errors in $\E_0[f]$ and $\E_1[f]$ when $\E_1[f] - \E_0[f]$ is small, since those errors would cause the empirical counterpart to the right hand side of display \eqref{AC basic formula} to vary widely.


The multi-class ($m\geq1$) versions of CC variants are direct generalizations of their binary class counterparts. We start with CC and PCC. If $f$ is a hard classifier, then without loss of generality, we may assume that $f$ maps $\X$ to a set which contains two types of $m$-dimensional vectors: the zero vector, and binary vectors that have a single entry equal to $1$ and the remaining entries equal to $0$. In this setup, if $f(x)$ is the zero vector, then this means that $f$ predicts $x$ to belong to class $0$; alternatively, if $f(x)$ is equal to a vector that contains $1$ for its $\numth{y}$ entry for some $y\in[m]$, then this means that $f$ predicts $x$ to belong to class $y$. Then, the multi-class CC quantifier is the empirical counterpart to $\E_{\pi^\ast}[f]$ computed on the test data. A similar set up can be used to describe the multi-class PCC quantifier. If $f$ is a soft-classifier, then without loss of generality, we may assume that $f$ maps $\X$ to the $m+1$ dimensional probability simplex in display \eqref{definition of m plus 1 dimensional probability simplex}. Then, once again, the multi-class PCC quantifier is the empirical counterpart to $\E_{\pi^\ast}[f]$ computed on the test data.

Next, we describe ACC and PACC in the multi-class case. These were originally proposed by \cite{KingVerbalAutopsy2008} and \cite{Hopkins2010AMO} for ACC quantifiers, but we can generalize them to PACC quantifiers as well. Our presentation also slightly modifies their original proposals, to account for the fact that we are focused on estimating $\pi^\ast$ (i.e., we are ignoring $\pi_0^\ast$ since probabilities sum to $1$). The main idea is as follows. If $f$ is either a hard classifier with outputs described in the previous paragraph, or a soft classifier with outputs in the $m+1$ dimensional probability simplex,  then it can be shown on the population level that
\begin{equation}\label{multivariate ACC equation}
	\E_{\pi^\ast}[f] - \E_0[f] =\overbrace{
		\begin{bmatrix}
			 \vrule & & \vrule \\
			\E_1[f] - \E_0[f] & \ldots & \E_m[f] - \E_0[f] \\
			 \vrule & & \vrule 
		\end{bmatrix}}^{=:\text{ } A_f \in \mathbb{R}^{m\times m}}
		\begin{bmatrix}
			\pi_1^\ast\\
			\vdots \\
			\pi_m^\ast
		\end{bmatrix}.
\end{equation}

The multi-class version of the assumption that $\E_1[f] - \E_0[f] \neq 0$ is that $A_f$ is positive definite. When this is the case, we can obtain the following closed form expression for $\pi^\ast$
\begin{equation}\label{multiclass ACC and PACC formula for pi ast}
	\pi^\ast = A_f^{-1}(\E_{\pi^\ast}[f] - \E_0[f]),
\end{equation}
which is the multi-class analogue of display \eqref{AC basic formula}. If $f$ is a hard classifier, then the multi-class ACC quantifier is constructed by replacing the right hand side of display \eqref{multiclass ACC and PACC formula for pi ast} with their sample-based counterparts: both $\E_0[f]$ and the columns of $A_f$ can be estimated via the train data, and $\E_{\pi^\ast}[f]$ can be estimated via the test data. The same procedure holds for obtaining the PACC quantifier when $f$ is a soft classifier. 


\subsubsection{Maximum Likelihood Approaches: EMQ and MLLS} \label{sec: Maximum Likelihood Approaches: EMQ and MLLS}

We now cover the basic mathematics behind EMQ and MLLS, starting with EMQ. For each $\beta$ in the $m+1$ dimensional probability simplex, define the average log likelihood function
\begin{equation} \label{avg log likelihood, for classical mixture problem}
	\frac{1}{n^\test} \log \L(\beta) := \frac{1}{n^\test} \sum_{i=1}^{n^\test} \log p_\beta(X_i^\test).
\end{equation}
EMQ seeks to maximize $\frac{1}{n^\test} \log \L(\beta)$ over $\beta$ in the $m+1$ dimensional probability simplex. However, since learning each of the component densities $p_y$ used in $p_{\beta} = \sum_{y=0}^m \beta_y p_y$ can be difficult in high dimensional settings, EMQ employs the following workaround. Using Bayes' Rule, one can show that
\begin{align*}
	\frac{1}{n^\test} \log \L(\beta) &\propto  \frac{1}{n^\test} \sum_{i=1}^{n^\test} \log \sum_{y=0}^m \beta_y \frac{\P_{\pi^\train}[Y=y \mid X = X_i^\test]}{\pi_y^\train},
\end{align*}
up to an additive constant that does not depend on $\beta$. Consequentially, choosing $\beta$ to maximize $\frac{1}{n^\test} \log \L(\beta)$ is equivalent to choosing $\beta$ to maximize the right hand side of the display above. Of course, $\P_{\pi^\train}[Y=y \mid X = x]$ is unknown, so EMQ replaces it with the estimates from a soft classifier $f$, which had been fit on the train data. Notationally, the objective function of EMQ is therefore:
\[
	\mathbf{F}^{\textup{EMQ}} (\beta) := \frac{1}{n^\test} \sum_{i=1}^{n^\test} \log \sum_{y=0}^m \beta_y \frac{f_y(X_i^\test)}{\pi_y^\train}.
\]
EMQ then outputs $\widehat{\pi}^{\textup{EMQ}}$, which is the value of $\beta$ in the $m+1$ dimensional probability simplex which maximizes the above objective. Now, if $f_y(x) \approx \P_{\pi^\train}[Y = y \mid X = x]$ for each $x\in \X$ and $y\in\Y$, then it is reasonable to expect that $\widehat{\pi}^{\textup{EMQ}}$ approximates the true MLE. However,  as mentioned in subsection \eqref{sec: Overview of Existing Quantifiers}, it is often the case that $f_y(x)$ is a poor estimate of $\P_{\pi^\train}[Y = y \mid X = x]$, in which case $\widehat{\pi}^{\textup{EMQ}}$ may be quite far from the true MLE. 

This is where MLLS comes in. As mentioned in subsection \eqref{sec: Overview of Existing Quantifiers}, MLLS differs from EMQ in that it replaces the soft-classifier $f$ with a calibrated version of itself, $f^\textup{cal}$, resulting in the following objective function for MLLS:
\begin{align*}
	\mathbf{F}^{\textup{MLLS}} (\beta) := \frac{1}{n^\test} \sum_{i=1}^{n^\test} \log \sum_{y=0}^m \beta_y \frac{f_y^\textup{cal}(X_i^\test)}{\pi_y^\train}
\end{align*}
It is important to note that this formulation is slightly different from what was actually discussed in the MLLS analysis done by  \cite{gargMLLS2020}; coming from the label shift camp, their goal was to learn the ratio between $\frac{\pi_y^\ast}{\pi_y^\train}$ for each $y\in\Y$, and so we have modified their formulation to fit our needs. However, the main idea is still the same. In their analysis, $f^\textup{cal}$ is called a calibrated version of $f$ because it satisfies $\P_{\pi^\train}[Y=y \mid f^\textup{cal} (X) = v] = v_y$ for each $y\in\Y$. Of course, in reality, the equality only holds approximately, depending on several factors such as the original $f$, whether BCTS is a reasonable approach for calibration, and the amount of held-out train samples used to calibrate $f$ into $f^\textup{cal}$. However, as an ideal to strive for, perfect calibration is a reasonable property to have: as an example from the binary class case, the calibration property says that if we look at the subset of $\X$ for which $f^\textup{cal}$ predicts there to be a $0.42$ chance of samples being from  class $1$, then indeed, the probability of a sample from that subset belonging to class $1$ is exactly $0.42$. Importantly, one can show that if  the calibration property holds perfectly for $f^\textup{cal}$, then  $\mathbf{F}^{\textup{MLLS}}$ can be rewritten in the following special way:
\begin{align*}
	\mathbf{F}^{\textup{MLLS}} (\beta) 
	&\propto  \frac{1}{n^\test} \sum_{i=1}^{n^\test} \log \sum_{y=0}^m \beta_y  p_y^\textup{f-cal}\big(  f^\textup{cal}(X_i^\test) \big),
\end{align*}
where $p_y^\textup{f-cal}$ denotes the density function of the random vector $f^\textup{cal}(X)$ when $X \sim p_y$, and the symbol $\propto$ means we have dropped a term that does not depend on $\beta$. Crucially, the right hand side of the display above is the average log likelihood (c.f. the definition of $\frac{1}{n^\test} \log \L(\beta)$) that we would be maximizing had someone replaced each $X_i^\test$ in our test set with $f^\textup{cal}(X_i^\test)$. Thus, if $f^\text{cal}$ is at least approximately calibrated, then MLLS is approximately equal to the MLE obtained on this transformed dataset. 





\subsection{A Concise Description of SELSE} \label{sec: A Concise Description of Our Quantifier}

We now describe the method for computing our ACC/PACC quantifier. Since the way our method handles the binary ($m=1$) and multiclass ($m\geq1$) cases are conceptually very similar, we first describe our algorithm for the former and then generalize it to the latter.

The first step in our algorithm is to form two disjoint datasets, $\allDI$ and $\allDII$, each of which contains both training and testing points. These datasets serve different purposes in our algorithm: $\allDII$ will be used to construct a function $f$, and $\allDI$ will use  $f$ to form the ACC/PACC quantifier $\widehat{\pi}$. The main purpose for splitting the data in this way is to facilitate our theoretical analysis of $\widehat{\pi}$ later on. It is also quite simple to construct $\allDI$ and $\allDII$. To start, randomly split the train data into two disjoint parts, $\DtrainI$ and $\DtrainII$, such that $\DtrainI$ and $\DtrainII$ have the same amount of samples for each class. Also randomly split the test data into two equally sized parts, $\DtestI$ and $\DtestII$. Then, define  $\allDI := (\DtrainI,\DtestI)$ and $\allDII := (\DtrainII,\DtestII)$.

After splitting the data, we use $\allDII$ to choose the function $f$ for our ACC/PACC quantifier. In short, our function $f$ will be an estimate of 
\begin{align*}
	s_{\gamma^\ast}(x) = \partialDerivative{\beta} \big\{ \log p_\beta(x) \big\}\Big\lvert_{\beta = \gamma^\ast} = \frac{p_1(x) - p_0(x)}{p_{\gamma^\ast}(x)},
\end{align*}
where $\gamma^\ast$ is given by: 
\[
	\gamma^\ast := \frac{\frac{\pi^\ast}{n^\test} + \frac{1}{n^\train} \frac{(\pi^\ast)^2}{\pi^\train} }{ \frac{1}{n^\test} + \frac{1}{n^\train} \big(\frac{(1-\pi^\ast)^2}{1-\pi^\train} + \frac{(\pi^\ast)^2}{\pi^\train}\big) }.
\]

The main reason for choosing this function is that $s_{\gamma^\ast}$ minimizes, over all choices of $f$, the asymptotic variance of the quantifier in display \eqref{quantifier for binary class case}, which is the empirical counterpart to the formula for $\pi^\ast$ in display \eqref{AC basic formula}. This will be discussed in more depth in subsequent subsections. 

Note that $s_{\gamma^\ast}$ is generally unknown (and thus, must be approximated) because both $\gamma^\ast$ (which depends on $\pi^\ast$) and the class densities $p_0,p_1$ are unknown. To build the approximation, we assume access to a preliminary quantifier for $\pi^\ast$. This quantifier can be poor (e.g., it may have large asymptotic variance), but it is acceptable as long as it converges in probability to $\pi^\ast$. The second step of our algorithm consists of computing this preliminary quantifier using the train and test data in $\allDII$, and then using the result to form a plug-in estimator $\widehat{\gamma}$ of $\gamma^\ast$.

In the third step, we use $\widehat{\gamma}$ and $\allDII$ to construct an estimator $\widehat{s}_{\widehat{\gamma}}$ of $s_{\gamma^\ast}$. Our approach for doing so is motivated by the following fact: among all functions $h$ such that $\Var_{\gamma^\ast}[h] =  \E_1[h] - \E_0[h]$, the function which maximizes $\E_1[h] - \E_0[h]$ is $h = s_{\gamma^\ast}$. This fact is an indirect consequence of Lemma \eqref{Optimal Matching Function}, and it forms the basis of our strategy for constructing $\widehat{s}_{\widehat{\gamma}}$: given a function class $\H$ that we believe contains $s_{\gamma^\ast}$, we choose $\widehat{s}_{\widehat{\gamma}}$ to be the $h\in \H$ that maximizes $\widehat{\E}_1^\trainII[h] - \widehat{\E}_0^\trainII[h]$ subject to the constraint that $\widehat{\Var}_{\widehat{\gamma}}[h] = \widehat{\E}_1^\trainII[h] - \widehat{\E}_0^\trainII[h]$. The operators $\widehat{\E}_0^\trainII, \widehat{\E}_1^\trainII$ and $\widehat{\Var}_{\widehat{\gamma}}^\trainII$ are all constructed using the training data in $\allDII$, and they are given by:
\[
	\widehat{\E}_0^\trainII[h] := \frac{1}{n_0^\train /2 } \sum_{(X_i^\train,Y_i^\train) \in \DtrainII} (1-Y_i^\train) h(X_i^\train) \qquad  \widehat{\E}_1^\trainII[h] := \frac{1}{n_1^\train /2 } \sum_{(X_i^\train,Y_i^\train) \in \DtrainII} Y_i^\train h(X_i^\train) 
\]
\[
	\widehat{\Var}_{\widehat{\gamma}}^\trainII[h] := \widehat{\E}_{\widehat{\gamma}}^\trainII[h^2] - \big(\widehat{\E}_{\widehat{\gamma}}^\trainII[h]\big)^2,
\]
where $n_0^\train$ and $n_1^\train$ are the total number of training samples from class $0$ and $1$, respectively, and 
\[
	\widehat{\E}_{\widehat{\gamma}}^\trainII[h] := (1-\widehat{\gamma})\widehat{\E}_0^\trainII[h] + \widehat{\gamma}\widehat{\E}_1^\trainII[h]. 
\]

We also add a regularization term to the objective function, so that we are maximizing $(\widehat{\E}_1^\trainII[h] - \widehat{\E}_0^\trainII[h]) - \lambda\Omega(h)^2$ for some norm $\Omega$ and parameter $\lambda > 0$. This allows us to choose $\H$ to be an extremely large, expressive class of functions, all the while avoiding overfitting. The exact quality of $\widehat{s}_{\widehat{\gamma}}$ as an estimator of $s_{\gamma^\ast}$ can be found in Lemma \eqref{Rate for Learning the Score Function, Unknown Gamma Star}, which provides details on the rate at which  $\E_{\allDII}\big[ \Var_{\gamma^\ast}[\widehat{s}_{\widehat{\gamma}} - s_{\gamma^\ast} ]  \big] \to 0$ in a setting that allows the complexity of $\H$ to grow as the number of train samples tends to infinity. Here, the $\allDII$ subscript in the operator $\E_{\allDII}$ is used to indicate that we are taking the expectation of $\Var_{\gamma^\ast}[\widehat{s}_{\widehat{\gamma}} - s_{\gamma^\ast}]$ with respect to the distribution of the dataset $\allDII$ (recall that $\widehat{s}_{\widehat{\gamma}}$ is constructed from $\allDII$). 

While this theoretical result holds for general $\H$, in practice we must limit ourselves to those $\H$ for which it is computationally feasible to solve the regularized optimization problem. For example, in the numerical studies section, we assume that $\H$ is a \textbf{Reproducing Kernel Hilbert Space (RKHS)}. For this choice of $\H$, solving the optimization problem is straightforward. Indeed, due to the Representer's theorem, the solution to the objective function is a linear combination of kernels, so the regularized optimization problem reduces to a quadratic program with a single quadratic equality constraint. While this problem is clearly non-convex, \cite{QuadraticOptWithOneQuadraticEqualityConstraint} presents a simple algorithm that ensures that the global optimum can nonetheless be obtained. We provide details of this algorithm in the context of our application in  subsection \eqref{Optimization Algorithm}. 

Having computed $\widehat{s}_{\widehat{\gamma}}$ using $\allDII$, the fourth step in our algorithm is to use $\widehat{s}_{\widehat{\gamma}}$ and $\allDI$ to build our quantifier. This is done by making two modifications to equation \eqref{AC basic formula}: we substitute $f$ with $\widehat{s}_{\widehat{\gamma}}$, and then using $\allDI$, we replace the population-level moments with their sample-level counterparts. The result is a quantifier $\widehat{\pi}^{\text{(a)}}$ which has the following form:
\begin{equation}\label{quantifier for binary class case}
	\widehat{\pi}^{\text{(a)}} := \frac{ \widehat{\E}_{\pi^\ast}^\testI[ \widehat{s}_{\widehat{\gamma}} ] - \widehat{\E}_0^\trainI[\widehat{s}_{\widehat{\gamma}}] }{ \widehat{\E}_1^\trainI[\widehat{s}_{\widehat{\gamma}}] - \widehat{\E}_0^\trainI[\widehat{s}_{\widehat{\gamma}}]  },
\end{equation}
where the operators $\widehat{\E}_0^\trainI, \widehat{\E}_1^\trainI$ are computed in the same fashion as $\widehat{\E}_0^\trainII, \widehat{\E}_1^\trainII$ but on $\allDI$ instead of $\allDII$, and $ \widehat{\E}_{\pi^\ast}^\testI[ \widehat{s}_{\widehat{\gamma}} ]$ is the average of $ \widehat{s}_{\widehat{\gamma}}$ evaluated on the test data in $\allDI$, i.e., 
\[
	\widehat{\E}_{\pi^\ast}^\testI[ \widehat{s}_{\widehat{\gamma}} ] = \frac{1}{n^\test/2} \sum_{X_i^\test \in \DtestI} \widehat{s}_{\widehat{\gamma}}(X_i^\test).
\]

Finally, steps (2)-(4) are repeated, but with the roles of $\allDI$ and $\allDII$ reversed. This results in another quantifier $\widehat{\pi}^{(b)}$. The final quantifier is then simply the average $\widehat{\pi} := \frac{ \widehat{\pi}^{(a)} + \widehat{\pi}^{(b)}  }{2}$. 

Having described the binary ($m=1$) class case, we now turn to the multiclass ($m\geq1$) case. The extension is conceptually identical, but there are three main differences. First, $\gamma^\ast$ is now a $m$-dimensional vector. Namely, $\gamma^\ast = (\gamma_1^\ast, \dots, \gamma_m^\ast)$ where $\gamma_y^\ast$ is given by:
\begin{equation}\label{paper definition of multiclass gamma star}
	\gamma_y^\ast := \frac{ \frac{\pi_y^\ast}{n^\test}  +   \frac{1}{n^\train} \frac{(\pi_y^\ast)^2}{\pi_y^\train} }{\frac{1}{n^\test}+ \frac{1}{n^\train} \sum_{k=0}^{m}\frac{(\pi_k^\ast)^2}{\pi_k^\train} }.
\end{equation}

To estimate $\gamma^\ast$ in step 2, we take the same approach as before: we use $\allDII$ to choose a preliminary quantifier for $\pi^\ast$, which we then use to replace $\pi^\ast$ in display \eqref{paper definition of multiclass gamma star}, resulting in a plug-in estimator $\widehat{\gamma}$ of $\gamma^\ast$.\\

Second, $s_{\gamma^\ast}$ is now a $\mathbb{R}^m$-valued gradient. For each $x\in\X$, 
\begin{align*}
	s_{\gamma^\ast}(x) & := \partialDerivative{\beta} \Big\{ \log p_\beta(x) \Big\}\Big\lvert_{\beta = \gamma^\ast} \\
	&= \frac{\big( p_1(x) - p_0(x), \dots, p_m(x) - p_0(x) \big) }{p_\beta(x)} \\ 
	&\equiv  \big( s_{\gamma^\ast,1}(x), \dots , s_{\gamma^\ast, m}(x) \big).  
\end{align*}

Each of the $m$ different component functions must be estimated. In our algorithm, we estimate each component function separately, and use an approach similar to what was used in the binary class case. Similar to before, one can show that, for each $y\in[m]$, among all functions $h_y:\X \mapsto \mathbb{R}$ such that $\Var_{\gamma^\ast}[h_y] = \E_y[h_y] - \E_0[h_y]$, the function $h_y$ that maximizes   $\E_y[h_y] - \E_0[h_y]$ is $h_y = s_{\gamma^\ast,y}$. This suggests constructing an estimate $\widehat{s}_{ \widehat{\gamma},y}$ of $s_{\gamma^\ast,y}$ in the following manner: given a function class $\H_y$ that we believe contains $s_{\gamma^\ast,y}$, choose $\widehat{s}_{ \widehat{\gamma},y}$ to be the member of $\H_y$ that maximizes $\widehat{\E}_y^\trainII[h_y] - \widehat{\E}_0^\trainII[h_y] - \lambda_y\Omega(h_y)^2$, subject to the constraint that $\Var_{ \widehat{\gamma} }[h_y] = \widehat{\E}_y^\trainII[h_y] - \widehat{\E}_0^\trainII[h_y]$. Analogous to before, $\lambda_y > 0$ is a regularization parameter and $\Omega$ is some norm on $\H_y$. When $\H_y$ is taken to be a union of RKHS's, the optimization problem can be solved using the same algorithm as before by \cite{QuadraticOptWithOneQuadraticEqualityConstraint}. After $\widehat{s}_{\widehat{\gamma},y}$ has been estimated for each $y\in[m]$, the final score function estimate is $\widehat{s}_{\widehat{\gamma}} := (\widehat{s}_{\widehat{\gamma},1}, \dots, \widehat{s}_{\widehat{\gamma},m}  )$.



The third difference lies in how $\widehat{s}_{\widehat{\gamma}}$ is used to construct an ACC quantifier. Similar to the method-of-moments approach used in the $m=1$ case, we make two changes to display \eqref{multivariate ACC equation}: we replace $f$ with $\widehat{s}_{\widehat{\gamma}}$, and we substitute all population moments with their sample-based counterparts.  As a result, the multi-class analogue to equation \eqref{quantifier for binary class case} is then the vector $\widehat{\pi}^{(a)}$ which solves 
\begin{equation} \label{sample multivariate ACC equation}
\widehat{\E}_{\pi^\ast}^\testI[\widehat{s}_{\widehat{\gamma}}] - \widehat{\E}_0^\trainI[\widehat{s}_{\widehat{\gamma}}] = 
		\overbrace{
		\begin{bmatrix}
			 \vrule & & \vrule \\
			\widehat{\E}_1^\trainI[\widehat{s}_{\widehat{\gamma}}] - \widehat{\E}_0^\trainI[\widehat{s}_{\widehat{\gamma}}] & \ldots & \widehat{\E}_m^\trainI[\widehat{s}_{\widehat{\gamma}}] - \widehat{\E}_0^\trainI[\widehat{s}_{\widehat{\gamma}}] \\
			 \vrule & & \vrule 
		\end{bmatrix}
		}^{=: \text{ } \widehat{A} \in \mathbb{R}^{m\times m} }
		\overbrace{\begin{bmatrix}
			\widehat{\pi}_1^{(a)}\\
			\vdots \\
			\widehat{\pi}_m^{(a)}
		\end{bmatrix}}^{ =: \text{ } \widehat{\pi}^{(a)} \in \mathbb{R}^{m}}.
\end{equation}

The final algorithm for the multi-class case is presented below.

\noindent\fbox{%
    \parbox{\textwidth}{%

\textbf{\uline{Semiparametric Efficient Label Shift Estimation (SELSE) Procedure}:} \\ 

\begin{enumerate}[wide, labelwidth=!, labelindent=0pt, label = \textbf{\uline{Step \arabic*.}}]
	\item If there is a class $y$ with only one sample in $\Dtrain$, or if $\Dtest$ only has one sample, then output $\widehat{\pi}  = \frac{1}{m+1}\vec{1}_m$. Else, randomly split $\Dtrain$ into two equally sized parts, $\DtrainI$ and $\DtrainII$, such that $\DtrainI$ and $\DtrainII$ have equal class proportions. Randomly split  $\Dtest$ into two equally sized parts, $\DtestI$ and $\DtestII$. Define $\allDI := (\DtrainI,\DtestI)$ and $\allDII := (\DtrainII,\DtestII)$.  \\
	
	\item Use $\allDII$ to construct an estimate $\widehat{\gamma}$ of $\gamma^\ast$.\\
	
	\item Use $\allDII$ and $\widehat{\gamma}$ to construct an estimate $\widehat{s}_{\widehat{\gamma}}: \X \mapsto\mathbb{R}^{m}$ of the score function $s_{\gamma^\ast}$. This is done by setting $\widehat{s}_{\widehat{\gamma}} := (\widehat{s}_{\widehat{\gamma},1},\dots,\widehat{s}_{\widehat{\gamma},m})$, and for each $y\in[m]$, choosing $\widehat{s}_{\widehat{\gamma},y} $ such that
      	\[
       		\widehat{s}_{\widehat{\gamma},y} \in \argmax_{h_y \in \widehat{C}_{\widehat{\gamma},y} } (\widehat{\E}_y^{\trainII}[h_y] -  \widehat{\E}_0^{\trainII}[h_y] ) - \lambda_y \Omega(h_y)^2,
        \]
        where  $\widehat{C}_{\widehat{\gamma},y}  :=  \Big\{ h_y \in \H_y \text{ } : \text{ }  \widehat{\Var}_{\widehat{\gamma}}^{\trainII} [h_y] =  \widehat{\E}_{y}^{\trainII} [h_y] - \widehat{\E}_{0}^{\trainII} [h_y]   \Big\}$.\\

	\item Use $\allDI$ and $\widehat{s}_{\widehat{\gamma}}$ to compute $\widehat{\pi}^{(a)}$, which is the solution to:
	\[\arraycolsep=1.4pt\def\arraystretch{1}
		\widehat{\E}_{\pi^\ast}^\testI[\widehat{s}_{\widehat{\gamma}}] - \widehat{\E}_0^\trainI[\widehat{s}_{\widehat{\gamma}}] = 
		\overbrace{
		\begin{bmatrix}
			 \vrule & & \vrule \\
			\widehat{\E}_1^\trainI[\widehat{s}_{\widehat{\gamma}}] - \widehat{\E}_0^\trainI[\widehat{s}_{\widehat{\gamma}}] & \ldots & \widehat{\E}_m^\trainI[\widehat{s}_{\widehat{\gamma}}] - \widehat{\E}_0^\trainI[\widehat{s}_{\widehat{\gamma}}] \\
			 \vrule & & \vrule 
		\end{bmatrix}
		}^{=: \text{ } \widehat{A} \in \mathbb{R}^{m\times m} }
		\overbrace{\begin{bmatrix}
			\widehat{\pi}_1^{(a)}\\
			\vdots \\
			\widehat{\pi}_m^{(a)}
		\end{bmatrix}}^{ =: \text{ } \widehat{\pi}^{(a)} \in \mathbb{R}^{m}}.
	\]
	If the matrix $\widehat{A}$ is not invertible, set $\widehat{\pi}^{(a)}  = \frac{1}{m+1}\vec{1}_m$.\\
	
	\item Repeat Steps $(2)-(4)$, but now with the roles of $\allDI$ and $\allDII$ reversed, and let $\widehat{\pi}^{(b)}$ denote the new result of step (4). Output $\widehat{\pi} := \frac{\widehat{\pi}^{(a)} +\widehat{\pi}^{(b)}  }{2}$.
\end{enumerate}
    
    }%
} 
\newpage

\subsection{Optimization Algorithm} \label{Optimization Algorithm}

In this subsection, we present our approach for solving the optimization problem in step 3 of our SELSE algorithm. To make the problem simpler, we assume for each $y\in[m]$ that the solution can be written as a linear combination of functions. That is, we assume that the solution lies in the span of $\phi_{y,1},\dots,\phi_{y,d}$, for some  $d\geq1$ and known functions $\phi_{y,i}: \X\mapsto\mathbb{R}$. This assumption is true when, for example, $\H_y$ is a RKHS or has a known function basis.

Given this linearity assumption, our approach is to rewrite the optimization problem as a quadratic program, with a single quadratic equality constraint. This then enables us to use the algorithm presented by \cite{QuadraticOptWithOneQuadraticEqualityConstraint}, which provably finds the global optimum. Towards that end, we start with rewriting our objective function. Let $h_y = \dotprod{w_y}{ \phi_{y}}$, where $w_y \in \mathbb{R}^d$ and $\phi_y:= (\phi_{y,1},\dots,\phi_{y,d})$. Also, assume that there exists a matrix $P_y \in \mathbb{R}^{d\times d}$ such that  $\lambda_y\Omega(h_y)^2  = w_y^T P_y w_y$. For example, this is true if $\Omega$ is a RKHS norm or if $\Omega(h_y) = \norm{w_y}{2}$; in the former case, $P_y$ is the Gram matrix scaled by $\lambda_y$ and in the latter case, $P_y = \lambda_y I_{d \times d}$. Then, one can show that the objective function in step 3 of SELSE can be rewritten as:
\[
	-2\bigdotprod{w_y}{ \tfrac{1}{2} \widehat{\E}_0^\trainII[\phi_y]  - \tfrac{1}{2}\widehat{\E}_y^{\trainII}[\phi_y]  } - w_y^T P_y w_y. 
\]

Turning this into a minimization problem, the objective function for $w_y$ then becomes:
\[
	w_y^T P_y w_y + 2 w_y^T p_y,
\]
where $p_y := \tfrac{1}{2} \widehat{\E}_0^\trainII[\phi_y]  - \tfrac{1}{2}\widehat{\E}_y^{\trainII}[\phi_y]$. As for the constraint in step 3 of SELSE, one can show that it is equivalent to
\[
	w_y^T Q_yw_y  + 2 w_y^T p_y = 0,
\]
where $Q_y := \widehat{\Var}_{\widehat{\gamma}}^\trainII[\phi_y]  \in \mathbb{R}^{d\times d}$. Thus, the final optimization problem to be solved is:

\noindent\fbox{%
    \parbox{\textwidth}{%
\[
	\text{Find } w_y \in\mathbb{R}^d \text{ to minimize: }  w_y^T P_y w_y + 2 w_y^T p_y
\]
\[
	\text{Subject to: } w_y^T Q_yw_y  + 2 w_y^T p_y = 0.
\]
    }%
}\vspace{0.2in}

Let $\widehat{w}_y$ denote the solution to the above problem. Then, the $\numth{y}$ score function estimate is given by $\widehat{s}_{\widehat{\gamma},y} = \widehat{w}_y^T \phi_y$. Note that the optimization routine has to be run for each $y\in[m]$; after that has been done, the final score function estimate in Step 3 of SELSE is $\widehat{s}_{\widehat{\gamma}} = (\widehat{s}_{\widehat{\gamma},1},\dots,\widehat{s}_{\widehat{\gamma},m})$.

\subsection{Verifying the Asymptotic Optimality of $s_{\gamma^\ast}$ for ACC/PACC Quantifiers} \label{sec: Motivating Methodology Lemmas and Intuition}



Having presented a concise overview of our algorithm, we now focus on substantiating the claim that, among all functions $f: \X \mapsto\mathbb{R}^m$, choosing $f = s_{\gamma^\ast}$ minimizes the asymptotic variance of the normalized error of ACC/PACC quantifiers. 



\subsubsection{Ideal Characteristics of $f$} \label{sec: Ideal Characteristics of f}
Towards that end, it will be helpful to first answer an important question: what even are the \textit{characteristics} of functions $f$ that cause this variance to be small? To identify those characteristics, we consider the following thought experiment. Suppose that we replaced $\widehat{s}_{\widehat{\gamma}}$ in the system of equations in display  \eqref{sample multivariate ACC equation} with some other function $f$, and then defined $\widehat{\pi}_{f}^{(a)} \in\mathbb{R}^m $ to be the quantifier that solves this new system of equations; i.e., $\widehat{\pi}_{f}^{(a)}$ solves 
\begin{equation}\label{damn i really love christina}
\widehat{\E}_{\pi^\ast}^\testI[f] - \widehat{\E}_0^\trainI[f] = 
		\overbrace{
		\begin{bmatrix}
			 \vrule & & \vrule \\
			\widehat{\E}_1^\trainI[f] - \widehat{\E}_0^\trainI[f] & \ldots & \widehat{\E}_m^\trainI[f] - \widehat{\E}_0^\trainI[f] \\
			 \vrule & & \vrule 
		\end{bmatrix}
		}^{=: \text{ } \widehat{A}_f \in \mathbb{R}^{m\times m} } \widehat{\pi}_{f}^{(a)}.
\end{equation}

For the sake of simplicity in this thought experiment, we will assume that $f$ is fixed--  i.e., unlike $\widehat{s}_{\widehat{\gamma}}$, the function $f$ does not depend on $\allDII$. We also assume that  both $\widehat{A}_f$ and $A_f$ (the population level counterpart to $\widehat{A}_f$) are positive definite. Then $\widehat{\pi}_f^{(a)} = \widehat{A}_f^{-1}(\widehat{\E}_{\pi^\ast}^\testI[f]-   \widehat{\E}_0^\trainI[f] )$, and one can show that the error satisfies
\begin{align*}
	\sqrt{\frac{n^\test n^\train}{n^\test + n^\train}}(\widehat{\pi}_f^{(a)} - \pi^\ast) &=  \overbrace{\sqrt{\frac{n^\test n^\train}{n^\test + n^\train}} A_f^{-1}( \widehat{\E}_{\pi^\ast}^\testI[f]  - \widehat{\E}_{\pi^\ast}^\trainI[f])}^{O_{\P}(1) } + o_{\P}(1),
\end{align*}
where $\widehat{\E}_{\pi^\ast}^\trainI[f] := \sum_{y=0}^m \pi_y^\ast \widehat{\E}_y^\trainI[f]$. The normalization factor of $\sqrt{\frac{n^\test n^\train}{n^\test + n^\train}}$ plays the same role as the usual "$\sqrt{n}$" in parametric statistics; the reason for the difference is that, in the quantification setting, there are two types of samples (train and test), and we cannot create consistent estimators for $\pi^\ast$ unless \textit{both} $n^\train,n^\test \to \infty$. The normalization above reflects this, since $\frac{n^\test n^\train}{n^\test + n^\train} \leq \min(n^\train,n^\test)$. The correctness of this scaling  will also be seen in the Theoretical Results section.  

Overall, the display above implies that the asymptotic variance of the normalized error of $\widehat{\pi}_f^{(a)}$ is equal to the variance of $\sqrt{\frac{n^\test n^\train}{n^\test + n^\train}} A_f^{-1}( \widehat{\E}_{\pi^\ast}^\testI[f]  - \widehat{\E}_{\pi^\ast}^\trainI[f])$.  To see how the choice of $f$ impacts this variance, we make use of the following Lemma.\\

\begin{lemma}[\uline{Variance of First Order Term using $f$}] \label{First Order Constant for Any Matching Function}
For any integer $q\geq1$ and function $f: \mathcal{X} \mapsto \mathbb{R}^q$, we have that
\begin{align*}
	\frac{1}{2}\Var_{\allDI}\Big[\widehat{\E}_{\pi^\ast}^\testI[f] -  \widehat{\E}_{\pi^\ast}^\trainI[f] \Big] &=\Bigg[\frac{1}{n^\test}+ \frac{1}{n^\train} \sum_{y=0}^{m}\frac{(\pi_y^\ast)^2}{\pi_y^\train}\Bigg] \Big( \Var_{\gamma^\ast}[f] - A_f \FisherInfo{\gamma^\ast;\textup{Cat}}^{-1} A_f^{T}\Big)\\
	&\qquad+ \frac{1}{n^\test}A_f \FisherInfo{\pi^\ast;\textup{Cat}}^{-1} A_f^{T}, 
\end{align*}
and, if $q=m$ and $A_f$ is positive definite, then
\begin{align*}
	\frac{1}{2}\Var_{\allDI}\Big[A_f^{-1}(\widehat{\E}_{\pi^\ast}^\testI[f] -  \widehat{\E}_{\pi^\ast}^\trainI[f]) \Big] &= \Bigg[\frac{1}{n^\test}+ \frac{1}{n^\train} \sum_{y=0}^{m}\frac{(\pi_y^\ast)^2}{\pi_y^\train}\Bigg] \Big( A_f^{-1}\Var_{\gamma^\ast}[f]A_f^{-T} - \FisherInfo{\gamma^\ast;\textup{Cat}}^{-1}  \Big)\\
	&\qquad  + \frac{1}{n^\test} \FisherInfo{\pi^\ast;\textup{Cat}}^{-1}.
\end{align*}
\end{lemma}
\vspace{0.2in}

To clarify notation, for each $\beta$ in the $m+1$ dimensional probability simplex, $\FisherInfo{\beta;\textup{Cat}}$ denotes the Fisher Information Matrix for $\beta$ when $Y\sim \textup{Cat}(\beta)$, where $\text{Cat}(\beta)$ denotes the categorical distribution supported on $\Y$. For us, $\textup{Cat}(\beta)$ is a generalization of the Bernoulli distribution to multiple outcomes; specifically,  $\P_{\textup{Cat}(\beta)}[Y = 0] = 1-\sum_{k=1}^m \beta_k$ and $\P_{\textup{Cat}(\beta) } [Y=y] = \beta_y$ for each $y\in[m]$. An exact expression for $\FisherInfo{\beta;\textup{Cat}}^{-1}$ can be found in Lemma \eqref{Covariance Matrix of a Finite Mixture}. \\

By Lemma \eqref{First Order Constant for Any Matching Function},  the variance matrix of  $\sqrt{\frac{n^\test n^\train}{n^\test + n^\train}} A_f^{-1}( \widehat{\E}_{\pi^\ast}^\testI[f]  - \widehat{\E}_{\pi^\ast}^\trainI[f])$ is equal to 
\[
	2\Bigg[\frac{n^\train}{n^\test + n^\train}+ \frac{n^\test}{n^\test + n^\train} \sum_{y=0}^{m}\frac{(\pi_y^\ast)^2}{\pi_y^\train}\Bigg] \Big( A_f^{-1}\Var_{\gamma^\ast}[f]A_f^{-T} - \FisherInfo{\gamma^\ast;\textup{Cat}}^{-1}  \Big)  +  2\frac{n^\train}{n^\test + n^\train} \FisherInfo{\pi^\ast;\textup{Cat}}^{-1},
\]
where the factor of $2$ is because $\allDI$ only contains half of the total train and test data, and is eliminated in Step 5 of SELSE when we take the average of $\widehat{\pi}^{(a)}$ and $\widehat{\pi}^{(b)}$. Crucially, the expression above indicates that the impact of $f$ on the asymptotic variance of $\widehat{\pi}_f^{(a)}$'s normalized error is entirely captured by $A_f^{-1}\Var_{\gamma^\ast}[f]A_f^{-T}$. This fact provides us valuable insights about the exact characteristics of $f$ that make the aforementioned asymptotic variance "small". Interestingly, these traits are conceptually identical for both the binary and multi-class cases, so to facilitate the discussion, we focus on the simpler $m=1$ setting (with all formal proofs allowing for $m \geq 1$). In this case, $f$ is scalar-valued and 
\[
	A_f^{-1}\Var_{\gamma^\ast}[f]A_f^{-T}  = \frac{\Var_{\gamma^\ast}[f]}{ (\E_1[f] - \E_0[f])^2},
\]
indicating that the asymptotic variance is low whenever $\Var_{\gamma^\ast}[f]$ is small and $(\E_1[f] - \E_0[f])^2$ is large. This makes sense in light of the first order term of $\widehat{\pi}_f^{(a)} - \pi^\ast$, which when $m=1$, is equal to 
\begin{equation} \label{binary class first order error}
	\frac{\widehat{\E}_{\pi^\ast}^\testI[f] -\widehat{\E}_{\pi^\ast}^\trainI[f] }{ \E_1[f] - \E_0[f] }.
\end{equation}

Roughly, if $\Var_{\gamma^\ast}[f]$ is small then with high probability so is the numerator, and if  $(\E_1[f] - \E_0[f])^2$ is large then so is the denominator. Thus, it is desirable to choose $f$ so that $\Var_{\gamma^\ast}[f]$ is small and $(\E_1[f] - \E_0[f])^2$ is large, as this would make the expression displayed above very small. Choosing $f$ so that both of these criteria are met is also nontrivial, since the criteria are at odds with one another: for example, if $f$ is any constant function, then $\Var_{\gamma^\ast}[f] $ is zero, but so is $(\E_1[f] - \E_0[f])^2$.

At first glance, these two characteristics may appear a bit arbitrary-- indeed, aside from the fact that $\E_1[f] - \E_0[f]$ appears in the denominator of display  \eqref{binary class first order error}, what exactly is the intuition for why ACC/PACC quantifiers benefit from large $(\E_1[f] - \E_0[f])^2$? Furthermore, why is $\Var_{\gamma^\ast}[f]$ even the right metric for the stochastic magnitude of the numerator in display \eqref{binary class first order error}? We now attempt to address both of these questions.

The answer to the first question is simple, and stems from the fact that ACC/PACC quantifiers are, at their core, using the method of moments to solve a system of linear equations.  Specifically, when $m=1$, instead of choosing $\widehat{\pi}_f^{(a)}$ to solve $\E_{\pi^\ast}[f] = (1-\widehat{\pi}_f^{(a)})\E_0[f] + \widehat{\pi}_f^{(a)} \E_1[f]$, ACC quantifiers choose $\widehat{\pi}_f^{(a)}$ to solve  $\widehat{\E}_{\pi^\ast}^\testI [f] = (1-\widehat{\pi}_f^{(a)}) \widehat{\E}_0^{\trainI}[f] + \widehat{\pi}_f^{(a)} \widehat{\E}_1^\trainI [f]$. This substitution of population-level moments for empirical moments is what makes it important to choose $f$ so that $\E_0[f]$ and $\E_1[f]$ are sufficiently different. If they are too similar, then even if $\widehat{\E}_0^\trainI[f]$ and $\widehat{\E}_1^\trainI[f]$ are close to $\E_0[f]$ and $\E_1[f]$ respectively, the solution $\widehat{\pi}_f^{(a)}$ will be hypersensitive to small errors in those approximations. The larger $(\E_1[f] - \E_0[f])^2$  is, the less susceptible $\widehat{\pi}_f^{(a)}$ will be to errors in moment estimation.

The answer to the second question is a bit longer, but nonetheless still straightforward: $\Var_{\gamma^\ast}[f]$ is an excellent proxy for $\Var_{\allDI}\big[ \widehat{\E}_{\pi^\ast}^\testI[f] -\widehat{\E}_{\pi^\ast}^\trainI[f]\big]$, which captures the stochastic size of $\widehat{\E}_{\pi^\ast}^\testI[f] -\widehat{\E}_{\pi^\ast}^\trainI[f]$ because  $\E_{\allDI}\big[ \widehat{\E}_{\pi^\ast}^\testI[f] -\widehat{\E}_{\pi^\ast}^\trainI[f]\big] = 0$. To help see why $\Var_{\gamma^\ast}[f]$ is a good proxy for $\Var_{\allDI}\big[ \widehat{\E}_{\pi^\ast}^\testI[f] -\widehat{\E}_{\pi^\ast}^\trainI[f]\big]$, it may help to first consider a simpler metric, such as
\begin{equation} \label{naive sum of variances for binary class case}
	\Var_{\pi^\test}[f] + \Var_{0}[f] + \Var_1[f].
\end{equation}

This metric seems reasonable if one considers the following expansion of $\widehat{\E}_{\pi^\ast}^\testI[f] -\widehat{\E}_{\pi^\ast}^\trainI[f]$:
\begin{align}
	\widehat{\E}_{\pi^\ast}^\testI[f] -\widehat{\E}_{\pi^\ast}^\trainI[f] &= (\widehat{\E}_{\pi^\ast}^\testI[f] - \E_{\pi^\ast}[f]) + (\E_{\pi^\ast}[f]-\widehat{\E}_{\pi^\ast}^\trainI[f])  \nonumber \\
	&= (\widehat{\E}_{\pi^\ast}^\testI[f] - \E_{\pi^\ast}[f]) + (1-\pi^\ast)(\E_{0}[f]-\widehat{\E}_{0}^\trainI[f]) + \pi^\ast (\E_{1}[f]-\widehat{\E}_{1}^\trainI[f]). \label{numerator decomposition into test and train for binary setting}
\end{align}

Indeed, since each of the three summands in line \eqref{numerator decomposition into test and train for binary setting} has mean zero, this expansion of $\widehat{\E}_{\pi^\ast}^\testI[f] -\widehat{\E}_{\pi^\ast}^\trainI[f]$ suggests that in order for $\Var_{\allDI}\big[ \widehat{\E}_{\pi^\ast}^\testI[f] -\widehat{\E}_{\pi^\ast}^\trainI[f] \big]$ to be small, it would be reasonable to simply choose $f$ so that $\Var_{\pi^\ast}[f], \Var_0[f]$ and $\Var_1[f]$ are all small. However, simply minimizing \eqref{naive sum of variances for binary class case} would be a gross oversimplification: depending on the relationship between  $n^\test$, $n^\train$, $\pi^\ast$ and $\pi^\train$,  the relative contributions of  $\Var_{\pi^\ast}[f], \Var_0[f]$ and $\Var_1[f]$ to $\Var_{\allDI}\big[ \widehat{\E}_{\pi^\ast}^\testI[f] -\widehat{\E}_{\pi^\ast}^\trainI[f] \big]$ may be far from equal. In fact, in certain circumstances, $\Var_{\allDI}\big[ \widehat{\E}_{\pi^\ast}^\testI[f] -\widehat{\E}_{\pi^\ast}^\trainI[f] \big]$ may be nearly equal to only one of those three variances. This is precisely where $\Var_{\gamma^\ast}[f]$ excels as a proxy for $\Var_{\allDI}\big[ \widehat{\E}_{\pi^\ast}^\testI[f] -\widehat{\E}_{\pi^\ast}^\trainI[f]\big]$: $\Var_{\gamma^\ast}[f]$ accurately accounts for each variance's contribution, regardless of the relationship between  $n^\test$, $n^\train$, $\pi^\ast$ and $\pi^\train$. This is highlighted by the behavior of $\Var_{\gamma^\ast}[f]$ when the aforementioned parameters take on extreme values relative to one other, such as in the following settings:\\

\begin{itemize}
	\item If $n^\test  << n^\train$, then $\Var_{\gamma^\ast}[f] \approx \Var_{\pi^\ast}[f]$ because $\gamma^\ast \approx \pi^\ast$, reflecting the fact that the relative scarcity of test data makes $\Var_{\pi^\ast}[f]$ the most important of the three variances to minimize, if we are to ensure that $\Var_{\allDI}\big[ \widehat{\E}_{\pi^\ast}^\testI[f] -\widehat{\E}_{\pi^\ast}^\trainI[f] \big]$ is small.\\
	
	\item If $n^\test >> n^\train$, $\pi^\ast \approx \tfrac{1}{2}$ and $\pi^\train \approx y$ for some $y\in\{0,1\}$, then $\Var_{\gamma^\ast}[f] \approx \Var_{1-y}[f]$ because $\gamma^\ast \approx 1-y$, reflecting the fact that the relative scarcity of training data from class $1-y$ makes $\Var_{1-y}[f]$ the most important of the three variances to minimize. 	\\
	
	\item If $n^\test >> n^\train$, $\pi^\train \approx \tfrac{1}{2}$ and $\pi^\ast \approx y$ for some $y \in \{0,1\}$, then $\Var_{\gamma^\ast}[f] \approx \Var_{y}[f]$ because $\gamma^\ast \approx y$, reflecting the fact that in line \eqref{numerator decomposition into test and train for binary setting}, $\widehat{\E}_{\pi^\ast}^\testI[f] - \E_{\pi^\ast}[f] \approx 0$ and $\pi_{1-y}^\ast (\E_{1-y}[f] - \widehat{\E}_{1-y}^\trainI[f]) \approx 0$, where the former is true because of the abundance of test data and the latter is true because $\pi_{1-y}^\ast \approx 0$. This makes $\Var_{y}[f]$ the most important of the three variances to minimize.
\end{itemize}

\subsubsection{Using $f = s_{\gamma^\ast}$ Maximizes the Ideal Characteristics} \label{sec: score function maximizes ideal characteristics}

In subsection \eqref{sec: Ideal Characteristics of f}, we established that minimizing the asymptotic variance of $\widehat{\pi}_f^{(a)}$'s normalized error amounts to choosing $f$ to minimize $A_f^{-1}\Var_{\gamma^\ast}[f]A_f^{-T}$, which when $m=1$, amounts to minimizing $\frac{\Var_{\gamma^\ast}[f]}{ (\E_1[f] - \E_0[f])^2}$. We also provided intuition for why small values of this metric are desirable. In this section, we now demonstrate that $\frac{\Var_{\gamma^\ast}[f]}{ (\E_1[f] - \E_0[f])^2}$ is minimized by $f = s_{\gamma^\ast}$, for the case of $m=1$. We then extend this result to the general $m \geq 1$ setting, and prove that using a $f$ which is linear in $s_{\gamma^\ast}$ is both necessary and sufficient for minimizing the asymptotic variance of $\widehat{\pi}_f^{(a)}$'s normalized error. Finally, we provide two sources of intuition justifying this result.

To prove that $f = s_{\gamma^\ast}$ minimizes $\frac{\Var_{\gamma^\ast}[f]}{ (\E_1[f] - \E_0[f])^2}$ is straightforward, and relies on the relationship between $A_f$ and $\Cov_{\gamma^\ast}[f,s_{\gamma^\ast}]$, as depicted in the following Lemma.\\

\begin{lemma}[\uline{Moment Identity}] \label{Moment Identity}
	For any integer $q \geq 1$ and function  $f: \X \mapsto \mathbb{R}^q$, we have that 
	\[
		A_f = \E_{\beta}\big[f s_{\beta}^T  \big] = \Cov_{\beta}\big[f,s_{\beta}\big],
	\]
	where  $\beta$ is any vector in the $m+1$ dimensional probability simplex, and $\Cov_{\beta}\big[f,s_{\beta}\big]$ is the cross-covariance matrix between $f$ and $s_{\beta}$.
\end{lemma}
\vspace{0.2in}

In the $m=1$ case, Lemma \eqref{Moment Identity} tells us that $\E_1[f] - \E_0[f] = \Cov_{\gamma^\ast}[f, s_{\gamma^\ast}]$. Using this identity, one can show that: 
\begin{align*}
	\frac{\Var_{\gamma^\ast}[f]}{ (\E_1[f] - \E_0[f])^2} 
	&= \frac{1}{\FisherInfo{\gamma^\ast}} \frac{1}{\Cor_{\gamma^\ast}[f, s_{\gamma^\ast} ]^2 },
\end{align*}
where $\Cor_{\gamma^\ast}[f, s_{\gamma^\ast} ]$ denotes the correlation between $f(X)$ and $s_{\gamma^\ast}(X)$ when $X \sim p_{\gamma^\ast}$. Since $\Cor_{\gamma^\ast}[f, s_{\gamma^\ast} ]^2$ achieves its maximum value of $1$ whenever $f = b s_{\gamma^\ast} + c$ with $b,c \in\mathbb{R}$ and $b\neq 0$, it follows that $\frac{\Var_{\gamma^\ast}[f]}{ (\E_1[f] - \E_0[f])^2}$ can be minimized by choosing $f$ to be linear in $s_{\gamma^\ast}$. Thus, for the case of $m=1$, this establishes that the asymptotic variance of $\widehat{\pi}_f^{(a)}$'s normalized error is minimized by linear functions of $s_{\gamma^\ast}$. 
 
However, several questions remain. For instance,  what version of this result holds in the multi-class case? In addition, is there any other function $f$, not necessarily linear in $s_{\gamma^\ast}$, which also minimizes the asymptotic variance? The latter is an especially important question to answer, because if such a $f$ did exist, then whether we should use that $f$ in place of $s_{\gamma^\ast}$ comes down to which function is easier to learn. To answer both of these questions, we turn to the following Lemma. \\

 \begin{lemma}[\uline{Optimal Function}] \label{Optimal Matching Function}
Let $\beta$ be a vector in the $m+1$ dimensional probability simplex with all entries bounded away from $0$ and $1$, and let $f: \X \mapsto \mathbb{R}^m$ be a function for which  $A_f$ is positive definite. If $\FisherInfo{\beta}$ is positive definite, then so is $A_{s_{\beta}}$, and
\[
	\FisherInfo{\beta}^{-1} = A_{s_{\beta}}^{-1}\Var_{\beta}[s_{\beta}]A_{s_{\beta}}^{-T} \preceq A_f^{-1}\Var_{\beta}[f]A_f^{-T}.
\]
Further, if $A_f^{-1}\Var_{\beta}[f]A_f^{-T}$ and $\FisherInfo{\beta}^{-1}$ have the same eigenvalues, then  $f = Bs_{\beta} + c$ a.e. $[p_\beta]$ for some constant vector $c \in \mathbb{R}^m$ and constant invertible matrix $B \in\mathbb{R}^{m\times m}$.
\end{lemma}
\vspace{0.2in}

To clarify notation in the Lemma above, $\FisherInfo{\beta}  = \Var_\beta[s_\beta] = \E_{\beta}[s_\beta s_\beta^T]$ is the \textbf{Fisher Information Matrix} for the parameter $\beta$, when $X \sim p_{\beta}$ and $\beta$ is a member of the $m+1$ dimensional probability simplex. Furthermore, the relation $\preceq$ denotes the Loewner order, which we use to generalize our method of comparing variances when $m=1$ to when $m \geq 1$. Namely, for any two square matrices $F_1, F_2$ of the same dimension, $F_1 \preceq F_2$ means that the $\numth{i}$ eigenvalue of $F_1$ is no bigger than the $\numth{i}$ eigenvalue of $F_2$, for all $i$.

When we set $\beta = \gamma^\ast$, Lemma \eqref{Optimal Matching Function} directly addresses our two questions from earlier. First, it generalizes the binary class result that $s_{\gamma^\ast}$ minimizes the asymptotic variance to the multi-class case: namely, since $A_{s_{\gamma^\ast}}^{-1}\Var_{\gamma^\ast}[s_{\gamma^\ast}](A_{s_{\gamma^\ast}}^{T})^{-1} \preceq A_f^{-1}\Var_{\gamma^\ast}[f](A_f^{T})^{-1}$, it follows that each eigenvalue of the asymptotic variance matrix of $\widehat{\pi}_{f}^{(a)}$'s normalized error is the smallest when $f = s_{\gamma^\ast}$. Of course, this comparison only applies to those $f$ for which $A_f$ is positive definite, but that is not a restriction for ACC/PACC quantifiers-- this is because $A_f$ \textit{has} to be positive definite in order for the population-level equation \eqref{multiclass ACC and PACC formula for pi ast} to be well-defined, as ACC/PACC quantifiers approximate this equation via method-of-moments. Indeed, as long as $A_f$ is positive definite, $f$ could be any hard, soft or calibrated classifier, ranging from a deep neural network to a simple $k$-NN model-- asymptotically,  $s_{\gamma^\ast}$ is still the better function to use\footnote{Moreover, we can rest assured that $\FisherInfo{\gamma^\ast}$ is positive definite, in all cases for which $\pi^\ast$ is statistically identifiable. The exact logic for this will be reviewed in the Theoretical Results section, but the main idea is that if $\FisherInfo{\gamma^\ast}$ was \textit{not} positive definite, then there would be at least two component densities $p_y$ and $p_{y'}$ that are identical, meaning that the test data carries zero information about $\pi_{y}^\ast$ and $\pi_{y'}^\ast$.}. It is also worthwhile to note that functions which are linear in $s_{\gamma^\ast}$ also share this optimality property, since if $f = B s_{\gamma^\ast} + c$ for some invertible matrix $B$ and vector $c$, then $A_f^{-1}\Var_{\beta}[f]A_f^{-T} = A_{s_{\beta}}^{-1}\Var_{\beta}[s_{\beta}]A_{s_{\beta}}^{-T}$. In summary, this means that linearity in $s_{\gamma^\ast}$ is a \textit{sufficient} condition for minimizing the asymptotic variance of $\widehat{\pi}_f^{(a)}$'s normalized error.


Second, Lemma \eqref{Optimal Matching Function} tells us that the foregoing linearity in $s_{\gamma^\ast}$ is also a \textit{necessary} condition. That is, to obtain the smallest possible asymptotic variance for  $\widehat{\pi}_f^{(a)}$'s normalized error, one \textit{has} to use a linear function of $s_{\gamma^\ast}$. This result can also be strengthened, in light of the following observation. Suppose that $f = Bs_{\gamma^\ast} + c$, for any positive definite $B$. Then, using equation \eqref{damn i really love christina}, one can show that $\widehat{\pi}_{f}^{(a)} = \widehat{\pi}_{s_{\gamma^\ast}}^{(a)}$, meaning that a quantifier built off of a linear function of $s_{\gamma^\ast}$ is \textit{point-wise equal} to the quantifier built off of just $s_{\gamma^\ast}$-- i.e., they are the same quantifier. Combined with the previous result, this invariance to linear transformations implies that $\widehat{\pi}_{s_{\gamma^\ast}}$ is effectively the only ACC/PACC quantifier with minimal asymptotic variance. 

Having established the unique asymptotic optimality of $s_{\gamma^\ast}$ for ACC/PACC quantifiers in the general $m \geq 1$ setting, we now provide some of our intuition behind this result. We do so in two ways. In the first way, we rationalize why $s_{\gamma^\ast}$ is the choice of $f$ which minimizes $A_f^{-1}\Var_{\gamma^\ast}[f]A_f^{-T}$. This compliments our analysis in subsection \eqref{sec: Ideal Characteristics of f}, wherein we interpreted why small values of $A_f^{-1}\Var_{\gamma^\ast}[f]A_f^{-T}$ are ideal for quantification. Since that analysis was done in the $m=1$ case, here we restrict ourselves to that setting as well (thus, $A_f^{-1}\Var_{\gamma^\ast}[f]A_f^{-T} =  \Var_{\gamma^\ast}[f] /(\E_1[f] - \E_0[f])^2$), although the same ideas also hold for $m\geq1$. The second way compares quantification to a classical statistical mixture problem, and uses the optimal procedure in the latter to justify the optimality of $s_{\gamma^\ast}$ in the former.

As mentioned, one way to understand the asymptotic optimality of $s_{\gamma^\ast}$ is to consider why it minimizes $\frac{\Var_{\gamma^\ast}[f] }{(\E_1[f] - \E_0[f])^2}$ among all $f$, or equivalently, why it maximizes $\frac{(\E_1[f] - \E_0[f])^2}{\Var_{\gamma^\ast}[f]}$ among all $f$.  Note that we can also restrict ourselves to those $f$ for which $\E_{\gamma^\ast}[f] = 0$, since $\frac{(\E_1[f] - \E_0[f])^2}{\Var_{\gamma^\ast}[f]}$ is invariant to linear transformations of $f$. As such, the metric in question is:
\[
	\frac{(\E_1[f] - \E_0[f])^2}{\Var_{\gamma^\ast}[f]} = \frac{(\E_1[f] - \E_0[f])^2}{\E_{\gamma^\ast}[f^2]}.
\]

Now, intuitively, for this metric to be large, $f(X)$ should take on very different values when $X \sim p_1$ compared to when $X\sim p_0$. This would make the numerator $(\E_1[f] -\E_0[f])^2$ large.  However, in that venture, the values that $f$ does take on should not be too extreme, as this would make the denominator $\E_{\gamma^\ast}[f^2]$ very big, undoing the benefits of large $(\E_1[f] -\E_0[f])^2$. Choosing $f = s_{\gamma^\ast}$ balances these two competing concerns. To see how, let us consider the general behavior of $s_{\gamma^\ast} = \frac{p_1 - p_0}{p_{\gamma^\ast}}$ over a neighborhood $\R \subset \X$ of the covariate space, chosen to be sufficiently small so as to ensure that all functions in our discussion are approximately constant over $\R$. There are $2$ cases to consider.
\begin{itemize}
	\item Case 1: For each $x\in\R$,  $p_1(x) \approx p_0(x)$.  
	\item Case 2: For each $x\in\R$, $\abs{p_1(x) - p_0(x)} >> 0$. 
\end{itemize}
In Case $1$, the fact that $p_1\approx p_0$ over $\R$ means that, for any given $f$, the region $\R$ will contribute very little to making the difference $(\E_1[f] - \E_0[f])^2$ large, regardless of the magnitude of $f$ over $\R$. However, the magnitude of $f$ on $\R$ \textit{will} contribute to making $\E_{\gamma^\ast}[f^2]$ larger. As such, $f$ having large magnitude on $\R$ can only decrease our metric. In light of this, it is best to choose $f$ to be equal to $0$ on $\R$, which $s_{\gamma^\ast}$ clearly accomplishes by virtue of its numerator being $p_1 - p_0$. In Case $2$, one of the two class densities is significantly higher than the other in the region $\R$. Thus, choosing $f$ to have large magnitude on $\R$ seems like a good idea, since it would amplify the magnitude of exactly one of $\E_0[f]$ and $\E_1[f]$ far more than the other, causing $(\E_1[f] - \E_0[f])^2$ to be large. The score function accomplishes this via its numerator $p_1-p_0$. However, at the same time, the magnitude of the chosen function cannot be too large, as that would also increase our metric's denominator. In light of this tension, $s_{\gamma^\ast}$ standardizes the difference $p_1 - p_0$ via dividing it by $p_{\gamma^\ast}$, which as a convex combination of $p_0$ and $p_1$, has a magnitude on the same scale as those densities. This standardization has the effect of making the magnitude of $s_{\gamma^\ast}$ positively "correlated\footnote{in the colloquial, not statistical, sense}" with the difference in $p_1-p_0$, whilst being agnostic to their actual scales. Essentially, this is what enables $\E_0[s_{\gamma^\ast}]$ and $\E_1[s_{\gamma^\ast}]$ to be very different, without also making $\E_{\gamma^\ast}[s_{\gamma^\ast}^2]$ too large in the process. 






We now turn to our second source of intuition for the optimality of $s_{\gamma^\ast}$: the classical statistical mixture problem. As mentioned earlier, in a statistical mixture problem, the goal is to learn the mixing coefficient vector $\pi^\ast$ under the assumption that each component density $p_y$ is known. Standard parametric theory tells us that, at least asymptotically, the best approach to accomplish this goal is the MLE, since its normalized error distribution has the smallest possible asymptotic variance. This problem is clearly similar to quantification, in that both problems share the same goal of learning $\pi^\ast$. However, quantification differs from the classical mixture problem in that the component densities are not assumed to be known. In lieu of perfect information about these densities, we instead have the imperfect "picture" of them painted for us by the train data. 

This information loss means that the task of quantification is usually more difficult than its classical mixture counterpart. However, an exception occurs when $n^\train$ grows far faster than $n^\test$. Under this asymptotic regime, the relative abundance of train data means that the component densities are essentially known. Thus, when $n^\train >> n^\test$, the two problems are equivalent, and so the optimal procedure for learning $\pi^\ast$ in a quantification problem should be identical to the optimal procedure for learning $\pi^\ast$ in a classical mixture problem. The idea that similar problems mean similar optimal procedures is our second source of intuition, and indeed, one can show that using $s_{\gamma^\ast}$ is what ensures this equivalence. This is easiest to see in the binary class case, when we set $n^\train = \infty$. In this case, the first order Taylor approximation of the classical mixture MLE is
\begin{equation} \label{first order taylor approx of classical MLE}
	\pi^\ast + \frac{\widehat{\E}_{\pi^\ast}^\testI[s_{\pi^\ast}] }{\FisherInfo{\pi^\ast}}.
\end{equation}
It can also be shown that the first order Taylor approximation of $\widehat{\pi}_{s_{\gamma^\ast}}^{(a)}$ is given by 
\begin{align*}
	\pi^\ast + \frac{ \widehat{\E}_{\pi^\ast}^\testI [s_{\gamma^\ast}] }{\E_1[s_{\gamma^\ast}] - \E_0[s_{\gamma^\ast}] }.
\end{align*}
However, since $\gamma^\ast = \pi^\ast$ when $n^\train = \infty$, it follows that the display above is equal to 
\begin{align*}
	\pi^\ast + \frac{ \widehat{\E}_{\pi^\ast}^\testI [s_{\pi^\ast}] }{\E_1[s_{\pi^\ast}] - \E_0[s_{\pi^\ast}] },
\end{align*}
which is identical to display \eqref{first order taylor approx of classical MLE} because $\E_1[s_{\pi^\ast}] - \E_0[s_{\pi^\ast}]  = \FisherInfo{\pi^\ast}$ by virtue of Lemma \eqref{Moment Identity}. Therefore, when $n^\train=\infty$, the quantifier $\widehat{\pi}_{s_{\gamma^\ast}}^{(a)}$ mimics the mixture MLE typically employed when the component densities are known. Thus, overall, the optimality of $s_{\gamma^\ast}$ makes sense in this special setting, since it agrees with what is already known about efficient estimation in parametric statistics. 

\subsection{SELSE vs. EMQ \& MLLS}

Having provided intuition for using $f = s_{\gamma^\ast}$, we briefly compare $\widehat{\pi}_{s_{\gamma^\ast}}^{(a)}$ against EMQ and MLLS. We motivate this by confronting an important question: what is the need for a new quantifier, in the first place? On the one hand, if we focus only on existing ACC / PACC quantifiers, then we have already discussed the answer to this in detail: $\widehat{\pi}_{s_{\gamma^\ast}}^{(a)}$ has the smallest possible asymptotic variance matrix among all ACC / PACC quantifiers. However, as subsection \eqref{sec: Overview of Existing Quantifiers} thoroughly points out, there are many other quantifiers besides CC variants in use. Since our optimality result only holds for the ACC / PACC family, one should therefore wonder: how do those other quantifiers compare against $\widehat{\pi}_{s_{\gamma^\ast}}^{(a)}$?




This question will be answered rigorously in the Theoretical Results section for a broad class of quantifiers, so for now, we restrict ourselves to a high-level comparison against specifically EMQ and MLLS, as we have already introduced these approaches in detail. As discussed in subsection \eqref{sec: Maximum Likelihood Approaches: EMQ and MLLS}, EMQ and MLLS are both based on the maximum likelihood principle: under ideal conditions, EMQ approximates the standard MLE for the classical mixture problem described at the end of subsection \eqref{sec: score function maximizes ideal characteristics}, and MLLS approximates the MLE for the version of the classical mixture problem that arises when the original test set has been transformed, i.e.,  $X_i^\test$ in the test set has been replaced by $f(X_i^\test)$ for some calibrated, soft classifier $f$. 

The fact that both EMQ and MLLS are based on the maximum likelihood principle imply that EMQ and MLLS are optimal when the assumptions of the classical mixture setting are met, which includes the assumption that the component densities $p_y$ are known. However, as discussed at the end of subsection \eqref{sec: score function maximizes ideal characteristics}, unless $n^\train >> n^\test$, this assumption is not valid for quantification tasks. This should cause one to be skeptical about the optimality of EMQ and MLLS for quantification, because intuitively, those methods are based on maximizing \textit{approximations} of likelihood functions formed via the train data, yet their final quantifiers are not adjusted to account for the error in those very approximations. This can be problematic when $n^\train \asymp n^\test$ or $n^\train << n^\test$, because in those situations, the relative uncertainty in the approximated likelihoods are higher. In contrast, $\widehat{\pi}_{s_{\gamma^\ast}}^{(a)}$ accounts for the uncertainty in the component densities $p_y$, as seen from our earlier analysis in subsection \eqref{sec: Ideal Characteristics of f}. To recap, this is because $f = s_{\gamma^\ast}$ minimizes $A_f^{-1}\Var_{\gamma^\ast}[f]A_f^{-T}$, and the dependence of $\gamma^\ast$ on $n^\test, n^\train$ and $\pi^\train$ enables $\Var_{\gamma^\ast}[f]$ to properly balance the varying levels of uncertainty among the different sources of data. For this reason, we would expect $\widehat{\pi}_{s_{\gamma^\ast}}$ to perform better than EMQ and MLLS, especially when $n^\train \asymp n^\test$ or $n^\train << n^\test$.

\subsection{Auxiliary Lemmas \& Proofs} \label{sec: Auxiliary Lemmas and Proofs for Methodology Section}

\begin{lemma}[\uline{Cauchy-Schwarz Equality Condition}] \label{Cauchy-Schwarz Equality Condition}
	Suppose that $\phi$ and $\zeta$ are $\mathbb{R}^m$-valued random vectors, and that $\Upsilon$ is the density function for their joint distribution. If $\E_\Upsilon[\zeta \zeta']$ is invertible and 
	\[
		\maxEval{\E_\Upsilon[ \phi \phi' ] - \E_\Upsilon[\phi \zeta'] \E_\Upsilon[\zeta\zeta']^{-1}\E_\Upsilon[\zeta \phi']} = 0,
	\]
	then
	\[
		\phi = \E_\Upsilon[\phi \zeta']\E_\Upsilon[\zeta \zeta']^{-1}\zeta \quad \textup{ a.e. } [\Upsilon].
	\]
\end{lemma}
\vspace{0.2in}

\begin{proof}[\uline{Proof of Lemma \eqref{Cauchy-Schwarz Equality Condition}}]
The proof of this Lemma is partially modeled after the proof of Lemma 1.1 in \cite{TRIPATHI19991}. First, observe that, for each $\alpha \in \mathbb{R}^m$:
\small
\begin{align*}
	 \alpha' \E_\Upsilon\Big[ \big\{  \phi   -   \E_\Upsilon[\phi \zeta']  \E_\Upsilon[\zeta\zeta']^{-1} \zeta   \big\}\big\{  \phi   -   \E_\Upsilon[\phi \zeta']  \E_\Upsilon[\zeta\zeta']^{-1} \zeta   \big\}'\Big]  \alpha  &= \E_\Upsilon\Big[ \big\{ \alpha' \phi   -  \alpha' \E_\Upsilon[\phi \zeta']  \E_\Upsilon[\zeta\zeta']^{-1} \zeta   \big\}^2\Big] \\
	 &\geq 0.
\end{align*}
\normalsize
At the same time, since $\maxEval{\E_\Upsilon[ \phi \phi' ] - \E_\Upsilon[\phi \zeta'] \E_\Upsilon[\zeta\zeta']^{-1}\E_\Upsilon[\zeta \phi']}  = 0$,  we also have that:
\small
\begin{align*}
	& \alpha' \E_\Upsilon\Big[ \big\{  \phi   -   \E_\Upsilon[\phi \zeta']  \E_\Upsilon[\zeta\zeta']^{-1} \zeta   \big\}\big\{  \phi   -   \E_\Upsilon[\phi \zeta']  \E_\Upsilon[\zeta\zeta']^{-1} \zeta   \big\}'\Big]  \alpha\\
	 &= \alpha' \E_\Upsilon\Big[ \big\{  \phi   -   \E_\Upsilon[\phi \zeta']  \E_\Upsilon[\zeta\zeta']^{-1} \zeta   \big\}\big\{  \phi'   -   \zeta'  \E_\Upsilon[\zeta\zeta']^{-1} \E_\Upsilon[\zeta \phi']     \big\} \Big]  \alpha \\ 
	 &= \alpha' \big\{  \E_\Upsilon[\phi\phi']   -   \E_\Upsilon[\phi \zeta']  \E_\Upsilon[\zeta\zeta']^{-1} \E_\Upsilon[\zeta \phi']   \big\}  \alpha \\ 
	 &\leq \norm{\alpha}{2}^2 \maxEval{\E_\Upsilon[ \phi \phi' ] - \E_\Upsilon[\phi \zeta'] \E_\Upsilon[\zeta\zeta']^{-1}\E_\Upsilon[\zeta \phi']} \\
	 &= 0,
\end{align*}
\normalsize
for each $\alpha \in \mathbb{R}^m$. Thus, whenever $\maxEval{\E_\Upsilon[ \phi \phi' ] - \E_\Upsilon[\phi \zeta'] \E_\Upsilon[\zeta\zeta']^{-1}\E_\Upsilon[\zeta \phi']}  = 0$, it follows that:
\small
\begin{align*}
	&\implies \alpha' \E_\Upsilon\Big[ \big\{  \phi   -   \E_\Upsilon[\phi \zeta']  \E_\Upsilon[\zeta\zeta']^{-1} \zeta   \big\}\big\{  \phi   -   \E_\Upsilon[\phi \zeta']  \E_\Upsilon[\zeta\zeta']^{-1} \zeta   \big\}'\Big]  \alpha  = 0 \qquad  \forall \ \alpha\in\mathbb{R}^m \\ 
	&\implies \E_\Upsilon\Big[ \big\{  \phi   -   \E_\Upsilon[\phi \zeta']  \E_\Upsilon[\zeta\zeta']^{-1} \zeta   \big\}\big\{  \phi   -   \E_\Upsilon[\phi \zeta']  \E_\Upsilon[\zeta\zeta']^{-1} \zeta   \big\}'\Big] = 0_{m\times m} \\ 
	&\implies \Tr\Big\{\E_\Upsilon\Big[ \big\{  \phi   -   \E_\Upsilon[\phi \zeta']  \E_\Upsilon[\zeta\zeta']^{-1} \zeta   \big\}\big\{  \phi   -   \E_\Upsilon[\phi \zeta']  \E_\Upsilon[\zeta\zeta']^{-1} \zeta   \big\}'\Big]\Big\} = 0 \\ 
	&\implies \E_\Upsilon\Big[ \bignorm{\phi   -   \E_\Upsilon[\phi \zeta']  \E_\Upsilon[\zeta\zeta']^{-1} \zeta  }{2}^2 \Big]  = 0 \\ 
	&\implies \phi  = \E_\Upsilon[\phi \zeta']  \E_\Upsilon[\zeta\zeta']^{-1} \zeta \quad \text{a.e.} \quad [\Upsilon],
\end{align*}
\normalsize
as desired.
\end{proof}
\vspace{0.2in}

\begin{proof}[\uline{Proof of Lemma \eqref{Moment Identity}}]
Let any $i \in [q]$ and any $y\in[m]$ be given. The $\numth{(i,y)}$ entry of $A_f$ can be rewritten as:
\begin{align*}
	\big[A_f\big]_{i,y} &= \E_y[f_i] - \E_0[f_i] \\ 
	&= \int_{\X} (p_y(x) - p_0(x)) f_i(x) d\mu(x) \\ 
	&= \int_{\X} p_{\beta}(x) \frac{p_y(x) - p_0(x)}{p_{\beta}(x)} f_i(x) d\mu(x) \\ 
	&= \int_{\X} p_{\beta}(x) s_{\beta,y}(x) f_i(x) d\mu(x) \\ 
	&= \E_{\beta}\big[f_i s_{\beta,y}  \big].
\end{align*}
This implies that $A_f = \E_{\beta}\big[f s_{\beta}^T  \big]$. Also note that $\E_{\beta}[s_{\beta}] = 0$, so $\E_{\beta}\big[f s_{\beta}^T  \big] = \E_{\beta}\big[f s_{\beta}^T  \big] - \E_{\beta}[f]\E_{\beta}[s_{\beta}]^T = \Cov_{\beta}[f, s_{\beta}]$, and so 
\[
	A_f  = \E_{\beta}\big[f s_{\beta}^T  \big] = \Cov_{\beta}[f, s_{\beta}],
\]
as claimed.
\end{proof}
\vspace{0.2in}

\begin{lemma}[\uline{Covariance Matrix of a Finite Mixture}] \label{Covariance Matrix of a Finite Mixture}
	Let $\beta$ be a vector in the $m+1$ dimensional probability simplex. 
	Then, for any function $f: \mathcal{X} \mapsto \mathbb{R}^q$, $q \geq 1$, we have that:
	\[
		\Var_{\beta}[f] =  \sum_{y=0}^{m} \beta_y \Var_y[f] + A_f \FisherInfo{\beta;\textup{Cat}}^{-1} A_f^{T},
	\]
	where 
	\[ \arraycolsep=1.4pt\def\arraystretch{1}
	\FisherInfo{\beta;\textup{Cat}}^{-1} := 
	\begin{bmatrix}
		\beta_1(1-\beta_1) & -\beta_1\beta_2 & \ldots & -\beta_1\beta_m \\ 
		-\beta_2\beta_1  & \beta_2(1-\beta_2) & \ldots &  - \beta_2\beta_m \\ 
		\vdots & \vdots & \ddots& \vdots \\
		-\beta_m\beta_1 & -\beta_m\beta_2 & \ldots & \beta_m(1-\beta_m)
	\end{bmatrix} \in \mathbb{R}^{m\times m}.
	\]  
\end{lemma}
\vspace{0.2in}

\begin{proof}[\uline{Proof of Lemma \eqref{Covariance Matrix of a Finite Mixture}}]
Observe that:
\begin{align*}
	\Var_{\beta}[f] &= \E_{\beta}[f f^{T}] - \E_{\beta}[f]\E_{\beta}[f]^{T} \\ 
	&= \sum_{y=0}^{m} \beta_y \E_{y}[f f^{T}] - \E_{\beta}[f]\E_{\beta}[f]^{T}.
\end{align*}
And that:
\begin{align*}
	\E_{\beta}[f]\E_{\beta}[f]^{T} &= \Bigg( \sum_{y=0}^{m} \beta_y \E_{y}[f] \Bigg)  \Bigg( \sum_{y=0}^{m} \beta_y \E_{y}[f] \Bigg)^{T}  \\
	&= \sum_{y=0}^{m}\sum_{z=0}^{m} \beta_y\beta_z \E_y[f]\E_z[f]^T \\ 
	&= \sum_{y=0}^{m} (\beta_y)^2 \E_y[f]\E_y[f]^T  + \sum_{y\neq z}\beta_y\beta_z \E_y[f]\E_z[f]^T.
\end{align*}
Thus:
\begin{align*}
	\implies \Var_{\beta}[f] &= \sum_{y=0}^{m} \beta_y \E_{y}[f f^{T}] - \sum_{y=0}^{m} (\beta_y)^2 \E_y[f]\E_y[f]^T  - \sum_{y\neq z}\beta_y\beta_z \E_y[f]\E_z[f]^T \\ 
	&= \sum_{y=0}^{m} \beta_y \Big[ \E_{y}[f f^{T}] -  \beta_y \E_y[f]\E_y[f]^T \Big]  - \sum_{y\neq z}\beta_y\beta_z \E_y[f]\E_z[f]^T \\ 
	&= \sum_{y=0}^{m} \beta_y \Big[ \big(\E_{y}[f f^{T}] -\E_y[f]\E_y[f]^T\big)  + \big(\E_y[f]\E_y[f]^T  -  \beta_y \E_y[f]\E_y[f]^T\big) \Big] \\
	&\qquad\qquad - \sum_{y\neq z}\beta_y\beta_z \E_y[f]\E_z[f]^T \\
	&= \sum_{y=0}^{m} \beta_y \Big[ \Var_y[f]  + \big(1 -  \beta_y \big)\E_y[f]\E_y[f]^T \Big]  - \sum_{y\neq z}\beta_y\beta_z \E_y[f]\E_z[f]^T \\ 
	&= \sum_{y=0}^{m} \beta_y \Var_y[f]  + \sum_{y=0}^{m} \beta_y\big(1 -  \beta_y \big)\E_y[f]\E_y[f]^T   - \sum_{y\neq z}\beta_y\beta_z \E_y[f]\E_z[f]^T.
\end{align*}
Define:
\[
	c_{y,z} = 
	\begin{cases} 
      		\beta_y\big(1 -  \beta_y \big) & y=z \\
     		-\beta_y\beta_z  & y\neq z.
  	\end{cases}
\]
Then we have that:
\begin{align*}
	\Var_{\beta}[f] &= \sum_{y=0}^{m} \beta_y \Var_y[f]  + \sum_{y=0}^m\sum_{z=0}^m c_{y,z} \E_y[f]\E_z[f]^T.
\end{align*}

Now, the above equality holds for any function $f: \mathcal{X} \mapsto \mathbb{R}^m$, so it also holds for the function given by $f(x) - \E_{0}[f]$. Thus, we have that:
\begin{align*}
	\implies \Var_{\beta}\Big[f - \E_{0}[f]\Big] &= \sum_{y=0}^{m} \beta_y \Var_y\Big[f - \E_{0}[f]\Big] + \sum_{y=0}^m\sum_{z=0}^m c_{y,z} \E_y\Big[f - \E_{0}[f]\Big]\E_z\Big[f - \E_{0}[f]\Big]^T.  
\end{align*}
\begin{align*}
	\implies \Var_{\beta}[f] &= \sum_{y=0}^{m} \beta_y \Var_y[f] + \sum_{y=0}^m\sum_{z=0}^m c_{y,z} \big(\E_y[f] - \E_{0}[f]\big) \big(\E_z[f] - \E_{0}[f]\big)^T\\
	&= \sum_{y=0}^{m} \beta_y \Var_y[f] + \sum_{y=1}^m\sum_{z=1}^m c_{y,z} \big(\E_y[f] - \E_{0}[f]\big) \big(\E_z[f] - \E_{0}[f]\big)^T\\
	&= \sum_{y=0}^{m} \beta_y \Var_y[f] + A_f \FisherInfo{\beta;\text{Cat}}^{-1} A_f^{T}.
\end{align*}

\end{proof}
\vspace{0.2in}

\begin{proof}[\uline{Proof of Lemma \eqref{First Order Constant for Any Matching Function}}]
By the assumed independence properties of our data, we have that:
\begin{align*}
	\Var_{\allDI}\Big[\widehat{\E}_{\pi^\ast}^\testI[f] -  \widehat{\E}_{\pi^\ast}^\trainI[f] \Big] &= \Var_{\DtestI}\Big[\widehat{\E}_{\pi^\ast}^\testI[f] \Big] + \Var_{\DtrainI}\Big[ \widehat{\E}_{\pi^\ast}^\trainI[f] \Big]\\
	 &= \Var_{\DtestI}\Big[\widehat{\E}_{\pi^\ast}^\testI[f] \Big] + \sum_{y=0}^{m} (\pi_y^\ast)^2 \Var_{\DtrainI}\Big[\widehat{\E}_{y}^\trainI[f] \Big] \\ 
	 &= \frac{1}{\tfrac{1}{2}n^\test}\Var_{\pi^{\ast}}[f] + \frac{1}{\tfrac{1}{2}n^\train} \sum_{y=0}^{m} \frac{(\pi_y^\ast)^2}{\pi_y^\train} \Var_{y}[f] \\ 
	 &= \frac{1}{\tfrac{1}{2}n^\test}\Bigg( \sum_{y=0}^{m} \pi_y^\ast \Var_y[f] + A_f \FisherInfo{\pi^\ast;\textup{Cat}}^{-1} A_f^{T}\Bigg) \\
	 &\qquad\qquad+ \frac{1}{\tfrac{1}{2}n^\train} \sum_{y=0}^{m} \frac{(\pi_y^\ast)^2}{\pi_y^\train} \Var_{y}[f],
\end{align*}
where the last equality follows from Lemma \eqref{Covariance Matrix of a Finite Mixture}. Thus,
\begin{align*}
	\frac{1}{2}\Var_{\allDI}\Big[\widehat{\E}_{\pi^\ast}^\testI[f] -  \widehat{\E}_{\pi^\ast}^\trainI[f] \Big]  &= \frac{1}{n^\test}\Bigg( \sum_{y=0}^{m} \pi_y^\ast \Var_y[f] + A_f \FisherInfo{\pi^\ast;\textup{Cat}}^{-1} A_f^{T}\Bigg)\\
	&\qquad\qquad+ \frac{1}{n^\train} \sum_{y=0}^{m} \frac{(\pi_y^\ast)^2}{\pi_y^\train} \Var_{y}[f]  \\ 
	&= \sum_{y=0}^{m}\Bigg(\frac{\pi_y^\ast}{n^\test}  +   \frac{1}{n^\train} \frac{(\pi_y^\ast)^2}{\pi_y^\train} \Bigg)\Var_y[f] + \frac{1}{n^\test}A_f \FisherInfo{\pi^\ast;\textup{Cat}}^{-1} A_f^{T}\\
	&= \Bigg[\frac{1}{n^\test}+ \frac{1}{n^\train} \sum_{y=0}^{m}\frac{(\pi_y^\ast)^2}{\pi_y^\train}\Bigg] \sum_{y=0}^{m}\gamma_y^\ast\Var_y[f] + \frac{1}{n^\test}A_f \FisherInfo{\pi^\ast;\textup{Cat}}^{-1} A_f^{T}  \\
	&= \Bigg[\frac{1}{n^\test}+ \frac{1}{n^\train} \sum_{y=0}^{m}\frac{(\pi_y^\ast)^2}{\pi_y^\train}\Bigg] \Big( \Var_{\gamma^\ast}[f] - A_f \FisherInfo{\gamma^\ast;\text{Cat}}^{-1} A_f^{T}\Big) \\
	&\qquad + \frac{1}{n^\test}A_f \FisherInfo{\pi^\ast;\textup{Cat}}^{-1} A_f^{T},
\end{align*}
where the last equality again follows from Lemma \eqref{Covariance Matrix of a Finite Mixture}. Thus, if $q=m$ and $A_f$ is positive definite, it follows that:
\begin{align}
	 \frac{1}{2}\Var_{\allDI}\Big[A_f^{-1}(\widehat{\E}_{\pi^\ast}^\testI[f] -  \widehat{\E}_{\pi^\ast}^\trainI[f]) \Big] &= \frac{1}{2}A_f^{-1}\Var_{\allDI}\Big[ \widehat{\E}_{\pi^\ast}^\testI[f] -  \widehat{\E}_{\pi^\ast}^\trainI[f] \Big]A_f^{-T} \nonumber \\ \nonumber \\
	&= \Bigg[\frac{1}{n^\test}+ \frac{1}{n^\train} \sum_{y=0}^{m}\frac{(\pi_y^\ast)^2}{\pi_y^\train}\Bigg] \Big( A_f^{-1}\Var_{\gamma^\ast}[f]A_f^{-T} - \FisherInfo{\gamma^\ast;\text{Cat}}^{-1}  \Big) \nonumber \\
	&\qquad + \frac{1}{n^\test} \FisherInfo{\pi^\ast;\textup{Cat}}^{-1} \label{somewhat free tetris},
\end{align}
as desired. 
\end{proof}
\vspace{0.2in}

\begin{proof}[\uline{Proof of Lemma \eqref{Optimal Matching Function}}]
First, note that $A_{s_\beta} =  \Var_{\beta}[s_{\beta}] \equiv \FisherInfo{\beta}$  by Lemma \eqref{Moment Identity}, so $A_{s_\beta}$ is positive definite as claimed because $\FisherInfo{\beta}$ is assumed to be positive definite. Thus, the inverses of both matrices are well-defined. This also means that 
\begin{align*}
	 A_{s_{\beta}}^{-1}\Var_{\beta}[s_{\beta}](A_{s_{\beta}}^{T})^{-1} &=  \FisherInfo{\beta}^{-1}\FisherInfo{\beta}\FisherInfo{\beta}^{-1} \\ 
	 &= \FisherInfo{\beta}^{-1},
\end{align*} 
as was also claimed. Next, we need to show that $\FisherInfo{\beta}^{-1} \preceq  A_{f}^{-1}\Var_{\beta}[f](A_{f}^{T})^{-1}$ for all functions $f:\X\mapsto\mathbb{R}^m$ for which $A_f$ is positive definite. However, before we do that, it will aid us greatly in our endeavor if we first take a small digression, and prove the following two facts:
\begin{itemize}
	\item $\Var_\beta[f]$ is positive definite.
	\item $\Var_{\beta}[g] \preceq \FisherInfo{\beta}$ for any $g:\X \mapsto \mathbb{R}^m$ satisfying $\Var_\beta[g] = A_g$ and $\E_{\beta}[g] = 0$. 
\end{itemize}

So, let us start by showing that  $\Var_\beta[f]$ is positive definite. Note that the positive definiteness of $A_f$ means that its rows are linearly independent, and so:
\begin{align*}
	A_f \text{ is positive definite} &\implies \nexists \alpha \in \mathbb{R}^m/0_m \ \text{ s.t. } \  \alpha'A_f = 0_m' \\ 
	&\implies  \nexists \alpha \in \mathbb{R}^m/0_m \ \text{ s.t. } \ \alpha'(\E_j[f] - \E_0[f]) = 0 \ \forall j\in[m] \\ 
	&\implies  \nexists \alpha \in \mathbb{R}^m/0_m \ \text{ s.t. }  \ \E_j[\alpha'f] = \E_0[\alpha'f] \ \forall j\in[m] \\  
	&\implies  \forall \alpha \in \mathbb{R}^m/0_m, \ \exists j\in[m]  \ \text{ s.t. } \ \E_j[\alpha'f] \neq \E_0[\alpha'f].
\end{align*}

Next, we will prove that
\begin{equation} \label{thing that contrapositive proves, for Var definiteness thing}
	\forall \alpha \in \mathbb{R}^m/0_m, \ \exists j\in[m]  \ \text{ s.t. } \ \E_j[\alpha'f] \neq \E_0[\alpha'f] \implies \forall \alpha \in \mathbb{R}^m/0_m, \ \nexists \eta \in \mathbb{R} \ \text{ s.t. } \   \P_\beta[\alpha'f = \eta] = 1
\end{equation}
by proving the contrapositive, i.e., that:
\[
	\exists \alpha \in \mathbb{R}^m/0_m, \eta \in \mathbb{R} \ \text{ s.t. } \   \P_\beta[\alpha'f = \eta] = 1 \implies  \exists \alpha \in \mathbb{R}^m/0_m \ \text{ s.t. } \ \E_j[\alpha'f] = \E_0[\alpha'f] \ \forall j\in[m]. 
\]

Towards that end, note that since $\P_\beta[\alpha'f = \eta] = \sum_{y=0}^m\beta_y \P_y[\alpha'f = \eta]$ and  $\beta_y > 0$ for all $y\in\Y$, if $\P_\beta[\alpha'f = \eta] = 1$ then it must be that $\P_y[\alpha'f = \eta] = 1$ for all $y\in\Y$ as well. Ergo:
\begin{align*}
	&\exists \alpha \in \mathbb{R}^m/0_m, \eta \in \mathbb{R} \ \text{ s.t. } \   \P_\beta[\alpha'f = \eta] = 1 \\
	&\implies \exists \alpha \in \mathbb{R}^m/0_m, \eta \in \mathbb{R} \ \text{ s.t. } \   \P_y[\alpha'f = \eta] = 1 \ \forall y\in\Y \\ 
	&\implies  \exists \alpha \in \mathbb{R}^m/0_m, \eta \in \mathbb{R} \ \text{ s.t. } \   \E_y[\alpha'f] = \eta \ \forall y\in\Y \\ 
	&\implies  \exists \alpha \in \mathbb{R}^m/0_m \ \text{ s.t. } \   \E_j[\alpha'f] = \E_0[\alpha'f] \ \forall j\in[m].
\end{align*}

Thus, the implication in display \eqref{thing that contrapositive proves, for Var definiteness thing} is indeed true, so it follows that:
\begin{align*}
	A_f \text{ is positive definite} &\implies \forall \alpha \in \mathbb{R}^m/0_m, \ \nexists \eta \in \mathbb{R} \ \text{ s.t. } \   \P_\beta[\alpha'f = \eta] = 1 \\ 
	&\implies \forall \alpha \in \mathbb{R}^m/0_m, \ \Var_\beta[\alpha'f] > 0 \\ 
	&\implies \forall \alpha \in \mathbb{R}^m/0_m, \ \alpha'\Var_\beta[f]\alpha > 0 \\ 
	&\implies \Var_\beta[f] \text{ is positive definite}.
\end{align*}

Next, we establish that $\Var_{\beta}[g] \preceq \FisherInfo{\beta}$ for any $g:\X \mapsto \mathbb{R}^m$ satisfying $\Var_\beta[g] = A_g$ and $\E_{\beta}[g] = 0$. Towards that end, observe that by Lemma \eqref{Moment Identity}, we have that $A_g = \E_{\beta}[g s_\beta^T]$, meaning that $\Var_\beta[g]  = \E_{\beta}[g s_\beta^T]$. So, for any vector $w\in\mathbb{R}^m$:
\begin{align*}
	w^T \Var_{\beta}[g] w &= w^T \E_{\beta}[g s_\beta^T] w \\
	&= \E_{\beta}\big[ (w^Tg ) (w^T s_\beta)  \big] \\ 
	&\leq \sqrt{\E_{\beta}\big[ (w^Tg )^2 \big]\E_{\beta}\big[ (w^Ts_{\beta} )^2 \big] }  \\ 
	&= \sqrt{\Var_\beta[w^Tg] \Var_\beta[w^Ts_{\beta} ]} \\ 
	&= \sqrt{ w^T\Var_\beta[g]w  \cdot w^T \FisherInfo{\beta}w },
\end{align*}
where the second to last line is because $\E_\beta[g] = 0 = \E_\beta[s_\beta]$. Thus, we have that $w^T \Var_{\beta}[g] w \leq w^T \FisherInfo{\beta}w$, meaning that $\Var_{\beta}[g] \preceq \FisherInfo{\beta}$. \\

Having completed our digression,  we now turn to proving that $\FisherInfo{\beta}^{-1} \preceq  A_{f}^{-1}\Var_{\beta}[f](A_{f}^{T})^{-1}$ for all functions $f:\X\mapsto\mathbb{R}^m$ for which  $A_f$ is positive definite. Note that this is equivalent to proving that $A_{f}^{T}\Var_{\beta}[f]^{-1}A_{f} \preceq \FisherInfo{\beta}$. Now, given any function $f$ such that $A_f$ is positive definite, it follows from our digression that $\Var_\beta[f]$ is also positive definite, so its inverse exists and we can define a new function $g_f: \X \mapsto \mathbb{R}^m$ as follows:
\[
	g_f := A_f^T\Var_\beta[f]^{-1}(f - \E_\beta[f]).
\]

Observe that:
\begin{align*}
	&A_{g_f} \Var_\beta[g_f]^{-1} A_{g_f}^T \\
	&= \Big( A_f^T\Var_\beta[f]^{-1}A_f   \Big)     \Big(A_f^T\Var_\beta[f]^{-1}  \Var_\beta[f]\Var_\beta[f]^{-1}A_f    \Big)^{-1}    \Big(   A_f^T\Var_\beta[f]^{-1}A_f \Big)^{T} \\ 
	&= \Big( A_f^T\Var_\beta[f]^{-1}A_f   \Big)     \Big(A_f^T\Var_\beta[f]^{-1}A_f    \Big)^{-1}    \Big(   A_f^T\Var_\beta[f]^{-1}A_f \Big) \\ 
	&=  A_f^T\Var_\beta[f]^{-1}A_f.
\end{align*}

Thus, given any $f$  for which  $A_f$ is positive definite, in order to show that $A_{f}^{T}\Var_{\beta}[f]^{-1}A_{f} \preceq \FisherInfo{\beta}$, it suffices to instead show that $A_{g_f} \Var_\beta[g_f]^{-1} A_{g_f}^T  \preceq \FisherInfo{\beta}$. Next, note that 
\begin{align*}
	\Var_\beta[g_f] &= A_f^T\Var_\beta[f]^{-1}  \Var_\beta[f]\Var_\beta[f]^{-1}A_f  \\ 
	&= A_f^T\Var_\beta[f]^{-1}A_f \\ 
	&= A_{A_f^T\Var_\beta[f]^{-1}f} \\ 
	&= A_{g_f},
\end{align*}
so $\Var_\beta[g_f] = A_{g_f}$.  Thus, $A_{g_f} \Var_\beta[g_f]^{-1} A_{g_f}^T  = \Var_\beta[g_f]$, so it suffices to instead show that $\Var_\beta[g_f] \preceq \FisherInfo{\beta}$.  By our digression from earlier, this is indeed true, because  $\Var_\beta[g_f] = A_{g_f}$ and $\E_\beta[g_f] = 0$. Therefore, we indeed have that $\FisherInfo{\beta}^{-1} \preceq  A_{f}^{-1}\Var_{\beta}[f](A_{f}^{T})^{-1}$ for all $f$ for which  $A_f$ is positive definite. \\ 



Finally, we will prove that if the eigenvalues of $A_{f}^{-1}\Var_{\beta}[f](A_{f}^{T})^{-1}$ and $\FisherInfo{\beta}^{-1}$ are identical, then there exists a constant vector $c \in \mathbb{R}^m$ and a constant invertible matrix $B \in \mathbb{R}^{m\times m}$ such that $ f = Bs_{\beta} + c$ a.e. $[p_\beta]$. Towards that end, note that if the eigenvalues of $A_{f}^{-1}\Var_{\beta}[f](A_{f}^{T})^{-1}$ and $\FisherInfo{\beta}^{-1}$ are identical, then the eigenvalues of $A_{f}^{T} \Var_{\beta}[f]^{-1}A_{f}$ and $\FisherInfo{\beta}$ are as well. So, $\Tr\big\{A_{f}^{T} \Var_{\beta}[f]^{-1}A_{f}\big\} = \Tr\big\{\FisherInfo{\beta}\big\}$, i.e., $\Tr\big\{\FisherInfo{\beta}  - A_{f}^{T} \Var_{\beta}[f]^{-1}A_{f} \big\} = 0$. But, we just showed that $\FisherInfo{\beta}^{-1} \preceq  A_{f}^{-1}\Var_{\beta}[f](A_{f}^{T})^{-1}$, so $\FisherInfo{\beta}  - A_{f}^{T} \Var_{\beta}[f]^{-1}A_{f}$ is positive semidefinite, i.e., all the eigenvalues of $\FisherInfo{\beta}  - A_{f}^{T} \Var_{\beta}[f]^{-1}A_{f}$ must be non-negative. Thus, the fact that  $\Tr\big\{\FisherInfo{\beta}  - A_{f}^{T} \Var_{\beta}[f]^{-1}A_{f} \big\} = 0$ means that the eigenvalues of $\FisherInfo{\beta}  - A_{f}^{T} \Var_{\beta}[f]^{-1}A_{f} $ must actually all be equal to zero. In particular, we have that $\maxEval{\FisherInfo{\beta}  - A_{f}^{T} \Var_{\beta}[f]^{-1}A_{f} } = 0$. \\

In addition, note that $\FisherInfo{\beta}  - A_{f}^{T} \Var_{\beta}[f]^{-1}A_{f} $ can be rewritten as:
\begin{align*}
	\E_\beta[s_\beta s_\beta^T] - \E_\beta[s_\beta (f-\E_\beta[f])^T] \E_{\beta}[(f-\E_\beta[f])(f-\E_\beta[f])^T]^{-1} \E_\beta[(f-\E_\beta[f]) s_\beta^T].
\end{align*}
So, since $\maxEval{\FisherInfo{\beta}  - A_{f}^{T} \Var_{\beta}[f]^{-1}A_{f} } = 0$ and $\Var_{\beta}[f]$ is positive definite, it follows from Lemma  \eqref{Cauchy-Schwarz Equality Condition} that 
\begin{align*}
	s_\beta &= \E_\beta[s_\beta (f-\E_\beta[f])^T]\E_\beta[(f-\E_\beta[f]) (f-\E_\beta[f])^T]^{-1}(f-\E_\beta[f]) \quad \textup{ a.e. } [p_\beta] \\ 
	&= \E_\beta[s_\beta f^T]   \Var_\beta[ f ]^{-1}   (f-\E_\beta[f]) \quad \textup{ a.e. } [p_\beta] \\ 
	&= A_f^T   \Var_\beta[ f ]^{-1}   (f-\E_\beta[f]) \quad \textup{ a.e. } [p_\beta].
\end{align*}

Since $A_f$ and $\Var_\beta[ f ]^{-1}$ are both invertible, it follows that $A_f^T   \Var_\beta[ f ]^{-1}$ is also invertible. Thus, the display above implies that:
\begin{align*}
	f &= (A_f^T   \Var_\beta[ f ]^{-1})^{-1} s_\beta  +  \E_\beta[f] \quad \textup{ a.e. } [p_\beta] \\ 
	&= \Var_\beta[ f ] A_f^{-T} s_\beta  +  \E_\beta[f] \quad \textup{ a.e. } [p_\beta].
\end{align*}

Set $c = \E_\beta[f]$ and $B = \Var_\beta[ f ] A_f^{-T}$, where $B$ is invertible because $\Var_\beta[ f ]$ and $A_f^{-T}$ are both invertible. Thus, we have found a constant invertible matrix $B \in \mathbb{R}^{m\times m}$ and constant vector $c \in \mathbb{R}^m$ such that $f = B s_\beta  + c \textup{ a.e. } [p_\beta]$, which proves the desiderata. 
\end{proof}
\vspace{0.2in}

\section{Theoretical Results} \label{sec: Theoretical Results Section}






In this section, we dive into some of SELSE's theoretical properties, with the ultimate goal of showing that SELSE is semiparametric efficient. Towards that end, the section is organized as follows. First, we make the Methodology section's discussion of SELSE's asymptotic behavior rigorous by providing a mathematical foundation for the intuitive arguments used in that section.  This includes a more precise analysis of SELSE's first order error term, and a characterization of the rate of its second order error term. All of this is done in the traditional quantification setting discussed in the Methodology section. 

Second, having established SELSE's asymptotic behavior in the traditional quantification setting, we then set out to prove that SELSE is semiparametric efficient. To that end, we motivate why a semiparametric framework is appealing for studying quantification in the first place, and in particular, the performance of quantifiers. We then argue why adjusting the generative process for our data is necessary if we want to utilize existing results from the semiparametrics literature in our derivation of the bound.

Third, we describe two new data generating regimes that are readily handled by the semiparametric literature and retain all key aspects of the traditional quantification problem, as was originally described in the Methodology section. We explain how these two regimes compare to the original, and provide a high-level game plan of how we will use the new regimes and our previous analysis of SELSE's asymptotic behavior to prove that SELSE is semiparametric efficient. 

Fourth, with this game plan in mind, we then dive into the details of the two new regimes. This includes reviewing their data generating mechanisms, as well as stating the assumptions we make about their parameter spaces. We make sure to present the intuition behind these assumptions, and provide justifications for why they are either necessary or reasonable to assume. We also introduce a large family of quantifiers $\curlyE$ that we plan to develop the semiparametric efficiency bound for. 

Fifth, having described the two new regimes in precise mathematical detail, we execute our aforementioned game plan, touring the intermediate lemmas and theorems in the two new regimes that strategically position us to prove our main result for SELSE. We also provide the intuition behind these intermediate results, and in doing so, we take a deeper dive into the concept of semiparametric efficiency. 

Finally, with this groundwork in place, we present the semiparametric efficiency bound, and show that SELSE is semiparametric efficient. We then conclude the section by analyzing the  bound, and we discuss how it reveals deep connections between the general problem of quantification and two other statistical estimation tasks.

\subsection{Asymptotics of SELSE in Fixed Sequence Regime}\label{Formalizing the Asymptotic Behavior of SELSE}

In order to rigorously show that SELSE has the smallest possible asymptotic variance, we first needed to formalize the intuitive arguments that were used in subsection \eqref{sec: Motivating Methodology Lemmas and Intuition} to identify that matrix, and more broadly, characterize SELSE's overall asymptotic behavior. In particular, while those arguments were able to correctly deduce the form of SELSE's asymptotic variance matrix, this conclusion could only be reached by assuming that the true score function $s_{\gamma^\ast}$ was known. However, in reality, SELSE only has access to an estimate $\widehat{s}_{\widehat{\gamma}}$. Thus, we begin this subsection by describing the lemmas that enabled us to show that $\widehat{s}_{\widehat{\gamma}}$ converges to $s_{\gamma^\ast}$, and determine the rate at which this occurs. We then show how this result allowed us to formally characterize the first and second order terms in SELSE's normalized error. 

The backbone of our proof for showing the convergence of $\widehat{s}_{\widehat{\gamma}}$ to $s_{\gamma^\ast}$ is a moment condition that characterizes $s_{\gamma^\ast}$. As mentioned in subsection \eqref{sec: score function maximizes ideal characteristics}, the components of $s_{\gamma^\ast}$ possess a special property: for each $y\in[m]$, $s_{\gamma^\ast,y}$ is a member of 
\begin{equation} \label{moment condition set}
	C_{\gamma^\ast,y}  := \Big\{ h_y \ \Big\lvert\  h_y:\X\mapsto\mathbb{R}, \ \Var_{\gamma^\ast}[h_y] = \E_y[h_y] - \E_0[h_y]  \Big\},
\end{equation}
and among all functions $h_y$ in that set, $\E_y[h_y] - \E_0[h_y]$ is \textit{uniquely} maximized by choosing $h_y = s_{\gamma^\ast,y}$. This property follows directly from Lemma \eqref{Optimal Matching Function}, and was our motivation for how we form the estimate $\widehat{s}_{\widehat{\gamma},y}$ in Step 3 of our algorithm. Indeed, the constraint set 
\[ 
	\widehat{C}_{\widehat{\gamma},y}  =  \Big\{ h_y \in \H_y \text{ } : \text{ }  \widehat{\Var}_{\widehat{\gamma}}^{\trainII} [h_y] =  \widehat{\E}_{y}^{\trainII} [h_y] - \widehat{\E}_{0}^{\trainII} [h_y]   \Big\}
\] 
in Step 3 is the empirical counterpart to $C_{\gamma^\ast,y}$ in display  \eqref{moment condition set}, and in that step, we are choosing $\widehat{s}_{\widehat{\gamma},y}$  to be the $h_y\in\widehat{C}_{\widehat{\gamma},y}$ that maximizes the difference $\widehat{\E}_y^{\trainII}[h_y]  -  \widehat{\E}_0^{\trainII}[h_y]$, which is the empirical counterpart to $\E_y[h_y] - \E_0[h_y]$. This correspondence suggests a strategy for proving the convergence of $\widehat{s}_{\widehat{\gamma}}$: since it seems reasonable to believe that $\widehat{s}_{\widehat{\gamma}}$ will be inside of $C_{\gamma^\ast,y}$ for large sample sizes (with high probability), and the unique maximization property of $s_{\gamma^\ast}$ implies that $\widehat{s}_{\widehat{\gamma}} = s_{\gamma^\ast}$ whenever it is the case that $\widehat{s}_{\widehat{\gamma}}\in C_{\gamma^\ast,y}$ and $\E_y[\widehat{s}_{\widehat{\gamma},y}]-\E_0[\widehat{s}_{\widehat{\gamma},y}] = \E_y[s_{\gamma^\ast,y}] - \E_0[s_{\gamma^\ast,y}]$, then in order to argue that $\widehat{s}_{\widehat{\gamma},y} \approx s_{\gamma^\ast,y}$, it may suffice to show that $\E_y[\widehat{s}_{\gamma^\ast,y}]-\E_0[\widehat{s}_{\gamma^\ast,y}] \approx \E_y[s_{\gamma^\ast,y}] - \E_0[s_{\gamma^\ast,y}]$. That is, it is reasonable to believe that bounding the magnitude of the difference
\begin{equation} \label{difference in yth diagonal entry of fisher info vs estimate when gamma star is known}
	\big(\E_y[s_{\gamma^\ast,y}] - \E_0[s_{\gamma^\ast,y}]\big)  -  \big(\E_y[\widehat{s}_{\gamma^\ast,y}]-\E_0[\widehat{s}_{\gamma^\ast,y}]\big)
\end{equation}
will lead to a bound on $\widehat{s}_{\widehat{\gamma},y} - s_{\gamma^\ast,y}$. To bound the difference above, we consider its positive and negative parts separately. The following lemma addresses its positive part. \\


\begin{lemma}[\uline{Rate for Learning Diagonal of $\FisherInfo{\gamma^\ast}$, When $\gamma^\ast$ is Known.}] \label{Rate for Learning Diagonal of Fisher Information Matrix}
Under Assumptions  \eqref{members of function class}, \eqref{Omega Properties}, \eqref{lambda and uniform convergence properties}, \eqref{bounded pis}, and \eqref{min Fisher Info} in the Appendix, we have for each $y\in[m]$ that
	\[
		\big(\E_y[s_{\gamma^\ast,y}] - \E_0[s_{\gamma^\ast,y}]\big)  -  \big(\E_y[\widehat{s}_{\gamma^\ast,y}]-\E_0[\widehat{s}_{\gamma^\ast,y}]\big) \leq T_y,
	\]
	where $T_y > 0$ is a random variable that satisfies
	\[
		\E_{\DtrainII}[T_y] = O\Bigg( \frac{1}{\sqrt{n^\train}} + \lambda_y \Omega^2(s_{\gamma^\ast,y}) + \E_{\DtrainII}[\U_y(\lambda_y)]  +  \frac{1}{\sqrt{\lambda_y}}e^{-C_1n^\train} \Bigg),
	\]
	for some constant $C_1 > 0$ independent of $n^\train$ or $n^\test$. Further, there exists constants $C_2,C_3 > 0$ independent of $n^\train$ and $n^\test$ such that, for all $a > 0$,
	\begin{align*}
		\P_{\DtrainII}[T_y \geq a] &\leq  16(2m+1) \Big(\exp\Big\{ -C_2 n^\train  \tau(a) \Big\} + \exp\Big\{ -C_3 n^\train  \sqrt{\tau(a)} \Big\}\Big) \\
		&\qquad+  \IBig{ \lambda_y \Omega^2(s_{\gamma^\ast,y})  \geq \frac{1}{12}a } \\
	 &\qquad + \P_{\DtrainII}\Bigg[\sup_{\Omega(h_y) \leq \frac{1}{L \sqrt{\lambda_y}}} \absBig{\widehat{\E}_y^{\trainII}[h_y] - \E_y[h_y]}  \geq \frac{1}{6}a\Bigg] \\
	 &\qquad+  \P_{\DtrainII}\Bigg[\sup_{\Omega(h_y) \leq \frac{1}{L \sqrt{\lambda_y}}} \absBig{\widehat{\E}_0^{\trainII}[h_y] - \E_0[h_y]} \geq \frac{1}{6}a\Bigg],
	\end{align*}
	where $ \tau(a) :=    \min\big\{1, a^2, \frac{a}{  \lambda_y \Omega^2(s_{\gamma^\ast,y}) } \big\}$.
\end{lemma}
\vspace{0.2in}

The proof for Lemma \eqref{Rate for Learning Diagonal of Fisher Information Matrix} can be found in the Appendix. Assumptions  \eqref{members of function class}-\eqref{gamma hat properties} are actually analogous to the assumptions presented and thoroughly investigated in the upcoming subsections \eqref{Assumptions on class proportions}-\eqref{subsection: Assumptions about gamma hat}, and are phrased in terms of the semiparametric framework soon to be introduced in subsection \eqref{semiparametric model specification}. The main reason for this redundancy is that it was far more convenient to prove the lemmas and theorems in the present subsection using Assumptions  \eqref{members of function class}-\eqref{gamma hat properties} than their semiparametric counterparts. However, to avoid unnecessary confusion and in anticipation of subsections \eqref{Assumptions on class proportions}-\eqref{subsection: Assumptions about gamma hat}, we hold off on describing Assumptions \eqref{members of function class}-\eqref{gamma hat properties} here, though the curious reader can read about them in the Appendix. The quantities $L$ and $\U_y(\lambda_y)$ in the statement of Lemma \eqref{Rate for Learning Diagonal of Fisher Information Matrix} will also be formally defined in subsections \eqref{Assumptions on class proportions}-\eqref{subsection: Assumptions about gamma hat}, but essentially, $L \in \mathbb{R}$ is a constant that can be shown to uniformly lower bound the components of $\gamma^\ast$ (proven in Lemma \eqref{Bounded Gamma Star} in the Appendix), and $\U_y(\lambda_y)$ denotes a sum of uniform deviation terms that reflect the tension between the (potentially growing) complexity of the space $\H_y$ and $n^\train$, and should satisfy $\E_{\DtrainII}[\U_y(\lambda_y)] \to 0$ as $n^\train \to \infty$.

In a nutshell, Lemma \eqref{Rate for Learning Diagonal of Fisher Information Matrix} shows that the positive part of the difference 
\[
	\big(\E_y[s_{\gamma^\ast,y}] - \E_0[s_{\gamma^\ast,y}]\big)  -  \big(\E_y[\widehat{s}_{\gamma^\ast,y}]-\E_0[\widehat{s}_{\gamma^\ast,y}]\big) 
\]
 approaches $0$ at a rate which depends on the number of train samples used to approximate $s_{\gamma^\ast}$, the complexity of $s_{\gamma^\ast}$ and the function space it lives in, as well as the amount of regularization. Provided that $n^\train \to \infty$  sufficiently quickly and $\lambda_y \to 0$ sufficiently slowly, we will have that $T_y = o_{\P}(1)$. Of course, what is considered "sufficiently quick" or "sufficiently slow" will depend on how the complexity of $\H_y$ grows; to provide the reader a sense of what these rates look like, we consider an example where $\H_y$ is assumed to be a union of different RKHSs in Lemma \eqref{RKHS Application} in the Appendix. Also note that Lemma \eqref{Rate for Learning Diagonal of Fisher Information Matrix} is phrased in terms of $\widehat{s}_{\gamma^\ast}$, but we are interested in $\widehat{s}_{\widehat{\gamma}}$. To be precise, $\widehat{s}_{\gamma^\ast}$ is the version of $\widehat{s}_{\widehat{\gamma}}$ that would have been obtained if $\widehat{\gamma} = \gamma^\ast$ in Step 2 of our algorithm. The analogue of Lemma \eqref{Rate for Learning Diagonal of Fisher Information Matrix} for $\widehat{s}_{\widehat{\gamma}}$ is Lemma \eqref{Rate for Learning Diagonal of Fisher Information Matrix, Unknown Gamma Star} in the Appendix-- we merely opted to describe the result for $\widehat{s}_{\gamma^\ast}$, since it is conceptually identical to the result for $\widehat{s}_{\widehat{\gamma}}$ but the actual statement for the latter is much more complicated, as it also depends on the rate at which $\widehat{\gamma}$ estimates $\gamma^\ast$. 
  
Bounding the magnitude negative part of display \eqref{difference in yth diagonal entry of fisher info vs estimate when gamma star is known} is far easier to do than its positive counterpart, and also depends on $T_y$. The main idea is that if $\widehat{C}_{\widehat{\gamma},y} = C_{\gamma^\ast,y}$, then the maximization property of $s_{\gamma^\ast,y}$ implies that \eqref{difference in yth diagonal entry of fisher info vs estimate when gamma star is known} should be non-negative, so if $\widehat{C}_{\widehat{\gamma},y} \approx C_{\gamma^\ast,y}$, then we have reason to believe it should typically not be too negative. The degree to which $\widehat{C}_{\widehat{\gamma},y} \approx C_{\gamma^\ast,y}$ depends on the extent to which empirical moments can be used to approximate population moments, and this is captured by $\U_y(\lambda_y)$ and hence by $T_y$. Combining this intuition for the magnitude of the negative part of display \eqref{difference in yth diagonal entry of fisher info vs estimate when gamma star is known} with Lemma \eqref{Rate for Learning Diagonal of Fisher Information Matrix}'s bound on the positive part, we were able to prove the following result about the  error of our score function estimate. \\

\begin{lemma}[\uline{Rate for Learning the Score Function, When $\gamma^\ast$ is Known.}] \label{Rate for Learning the Score Function}
	Under Assumptions  \eqref{members of function class}, \eqref{Omega Properties}, \eqref{lambda and uniform convergence properties}, \eqref{bounded pis} and  \eqref{min Fisher Info} in the Appendix, we have for each $y\in[m]$ that
	\[
		\E_{\DtrainII}\Big[ \Var_{\gamma^\ast}[\widehat{s}_{\gamma^\ast,y} - s_{\gamma^\ast,y} ] \Big] = O\Big( \E_{\DtrainII}[T_y] \Big).
	\]
	Further, for all $a > 0$,
	\begin{align*}
		\P_{\DtrainII}\Big[ \Var_{\gamma^\ast}[  \widehat{s}_{\gamma^\ast,y} - s_{\gamma^\ast,y}]  \geq a \Big] &\leq \P_{\DtrainII}\bigg[ T_y \geq \frac{1}{4}a \bigg] \\
		&\qquad+  \P_{\DtrainII}\Bigg[\sup_{\Omega(h_y) \leq  \frac{1}{L\sqrt{\lambda_y}} } \absBig{ \widehat{\Var}_{\gamma^\ast}^{\trainII}[h_y]  -  \Var_{\gamma^\ast}[h_y]} \geq \frac{1}{4}a \Bigg] \\
		&\qquad +  \P_{\DtrainII}\Bigg[  \sup_{\Omega(h_y)\leq  \frac{1}{L\sqrt{\lambda_y}} } \absBig{ \widehat{\E}_y^{\trainII}[h_y]  -   \E_y[h_y]} \geq \frac{1}{4}a \Bigg] \\ 
		&\qquad +  \P_{\DtrainII}\Bigg[  \sup_{\Omega(h_y)\leq  \frac{1}{L\sqrt{\lambda_y}} } \absBig{ \widehat{\E}_0^{\trainII}[h_y]  -   \E_0[h_y]} \geq \frac{1}{4}a \Bigg].
	\end{align*}
\end{lemma}
\vspace{0.2in}

Note that Lemma \eqref{Rate for Learning the Score Function} only applies to $\widehat{s}_{\gamma^\ast,y}$, but we are interested in $\widehat{s}_{\widehat{\gamma},y}$. Our reason for presenting Lemma \eqref{Rate for Learning the Score Function} is the same as our reason for presenting Lemma \eqref{Rate for Learning Diagonal of Fisher Information Matrix} instead of Lemma \eqref{Rate for Learning Diagonal of Fisher Information Matrix, Unknown Gamma Star} from earlier: namely, the result for the estimation error $\widehat{s}_{\widehat{\gamma},y} - s_{\gamma^\ast,y}$ is nearly identical to the result for the estimation error $\widehat{s}_{\gamma^\ast,y} - s_{\gamma^\ast,y}$, but the statement for the latter is far more complicated than the former, and so it is saved for the Appendix in Lemma \eqref{Rate for Learning the Score Function, Unknown Gamma Star}. 

Overall, Lemma \eqref{Rate for Learning the Score Function} shows that the quality of $\widehat{s}_{\widehat{\gamma},y}$ as an estimate of the score function depends on the same factors as $T_y$: the amount of training samples and the complexity of the space $\H_y$. Provided that these quantities grow at appropriate rates, we can expect that the true score function can be learned. Most importantly,  this convergence result opens the floodgates for the results previously described in the Methodology section to be used, which had assumed that SELSE had access to the true score function. In particular, we can now rigorously show that the first order term of SELSE's total error is  $\FisherInfo{\gamma^\ast}^{-1}\big(\widehat{\E}_{\pi^\ast}^{\testI}[s_{\gamma^\ast}] - \widehat{\E}_{\pi^\ast}^{\trainI}[s_{\gamma^\ast}] \big)$, which is the error incurred when $\widehat{s}_{\widehat{\gamma}} = s_{\gamma^\ast}$ and is due to using $\allDI$ to approximate the population moments in equation \eqref{multivariate ACC equation}. In fact, we were able to fully characterize this first order term up to its asymptotic distribution, as seen in the following lemma.\\

\begin{lemma}[\uline{First Order Term}] \label{First Order Term}
	Under Assumptions \eqref{FIM Difference is Bounded}, \eqref{bounded pis}, and \eqref{min Fisher Info} in the Appendix, we have that
	\[
		\Bigg(  \Var_{\allDI}\Big[\FisherInfo{\gamma^\ast}^{-1}\big(\widehat{\E}_{\pi^\ast}^{\testI}[s_{\gamma^\ast}] - \widehat{\E}_{\pi^\ast}^{\trainI}[s_{\gamma^\ast}] \big) \Big] \Bigg)^{-\frac{1}{2}} \FisherInfo{\gamma^\ast}^{-1}\big(\widehat{\E}_{\pi^\ast}^{\testI}[s_{\gamma^\ast}] - \widehat{\E}_{\pi^\ast}^{\trainI}[s_{\gamma^\ast}] \big)  \goesto{d} \N(0,I_{m}).
	\]
\end{lemma}
\vspace{0.2in}

The proof for Lemma \eqref{First Order Term} is in  the Appendix and makes use of standard triangular array central limit theorems. The following theorem expands on Lemma \eqref{First Order Term}, providing both a closed form for the variance of SELSE's first order error term and a rate-exact characterization of its second order error term.\\


\begin{theorem}[\uline{Normalized Error of $\widehat{\pi}$}] \label{Standardized Error Rate of Full Estimator}
Under Assumptions  \eqref{members of function class},  \eqref{Omega Properties}, \eqref{lambda and uniform convergence properties}, \eqref{FIM Difference is Bounded}, \eqref{bounded pis}, \eqref{min Fisher Info} and \eqref{gamma hat properties} in the Appendix, we have that 
\begin{align*}
	\Bigg( \Bigg[\frac{1}{n^\test}+ \frac{1}{n^\train} \sum_{y=0}^{m}\frac{(\pi_y^\ast)^2}{\pi_y^\train}\Bigg] \Big( \FisherInfo{\gamma^\ast}^{-1} - \FisherInfo{\gamma^\ast;\textup{Cat}}^{-1}  \Big)  + \frac{1}{n^\test} \FisherInfo{\pi^\ast;\textup{Cat}}^{-1} \Bigg)^{-\frac{1}{2}} (\widehat{\pi} - \pi^\ast) &= Z + \epsilon,
\end{align*}
where 
\begin{align*}
 	Z &= \Bigg( \Bigg[\frac{1}{n^\test}+ \frac{1}{n^\train} \sum_{y=0}^{m}\frac{(\pi_y^\ast)^2}{\pi_y^\train}\Bigg] \Big( \FisherInfo{\gamma^\ast}^{-1} - \FisherInfo{\gamma^\ast;\textup{Cat}}^{-1}  \Big)  + \frac{1}{n^\test} \FisherInfo{\pi^\ast;\textup{Cat}}^{-1} \Bigg)^{-\frac{1}{2}}\frac{1}{n} \sum_{i= 1}^{n} \psi_i^\eff  \\
	&\goesto{d} \N(0,I_m)
\end{align*}
where $n=n^\train+n^\test$,  $\psi_i^\eff \in \mathbb{R}^m$ is a function of the $\numth{i}$ data point and is defined in Lemma \eqref{Alternative Expression for First Order Error} in the Appendix, and $\epsilon \in \mathbb{R}^m$ is a random vector satisfying   
\begin{align*}
	\norm{\epsilon}{2} &=   O_{\P}\Bigg(\sqrt{ \frac{1}{\sqrt{n^\train}} + \max_y \lambda_y \Omega(s_{\gamma^\ast,y})^2 +  \frac{1}{\sqrt{\min_y\lambda_y}}e^{-Cn^\train}  }\Bigg) \\ 
	&\qquad+   O_{\P}\Bigg(\sqrt{ \max_y \E_{\DtrainII}\Big[ \U_y(\lambda_y)\Big] + \sqrt{\E_{\allDII}\Big[ \norm{\widehat{\gamma} - \gamma^\ast}{2}^2 \Big]}  }\Bigg) \\ 
	&\qquad + O_{\P}\Bigg( \textcolor{black}{ \frac{1}{\sqrt{ 1/n^\test + 1/n^\train} } }\bigg\{\textcolor{black}{   e^{-C (\min_y\lambda_y)^2 n^\train}   }   + \P_{\allDII}\Big[\bignorm{ \widehat{\gamma} - \gamma^\ast }{2} \geq C \Big] \bigg\} \Bigg),
\end{align*}
\normalsize
for some global constant $C > 0$. 
\end{theorem}
\vspace{0.2in}

Whereas the first order error was due to the method of moments approximation of equation \eqref{multivariate ACC equation}, Theorem \eqref{Standardized Error Rate of Full Estimator} makes clear that the second order error $\epsilon$ depends on how fast $\Var_{\gamma^\ast}[\widehat{s}_{\gamma^\ast,y} - s_{\gamma^\ast,y}]$ goes to $0$, as seen by how the rates for both $\epsilon$ and $\Var_{\gamma^\ast}[\widehat{s}_{\gamma^\ast,y} - s_{\gamma^\ast,y}]$ depend on the same factors used to bound $T_y$, i.e., $n^\train$ and the complexity of $\H_y$. Provided that $n^\train$ grows sufficiently fast relative to this complexity, Theorem \eqref{Standardized Error Rate of Full Estimator} implies that $\epsilon = o_{\P}(1)$, in which case asymptotically, the normalized error is fully characterized by the aforementioned first order term. Furthermore, it is important to note that, unlike the other lemmas previously presented in this subsection, Theorem \eqref{Standardized Error Rate of Full Estimator} applies to the \textit{complete} SELSE algorithm,  in that it accounts for the fact that $\gamma^\ast$ must be estimated via $\widehat{\gamma}$. Theorem \eqref{Standardized Error Rate of Full Estimator} is also able to remove the factor of $2$ that appeared when we first stated SELSE's asymptotic variance in subsection \eqref{sec: Motivating Methodology Lemmas and Intuition} as a result of Step 1's sample splitting. This was achievable because of Step 5, which takes $\widehat{\pi}$ to be the average of $\widehat{\pi}^{(a)}$ and $\widehat{\pi}^{(b)}$. Ultimately, Theorem  \eqref{Standardized Error Rate of Full Estimator} paints a comprehensive picture of SELSE's asymptotic behavior, and in doing so, it provides a much needed theoretical backbone to support the arguments  made in the Methodology section.

\subsection{Motivation for Semiparametric Statistics} \label{sec: semiparametric statistics}

Having established SELSE's asymptotic behavior in the traditional quantification setting, our next goal is to show that SELSE is semiparametric efficient. However, before we can do that, we first must answer two fundamental questions: what even \textit{is} semiparametric statistics, and why are we using it in the first place? This subsection serves to address precisely those questions, and in doing so, will provide motivation for the work we do later on. 

In virtually all data analysis settings, it is often desirable to make the weakest possible assumptions  about the data at hand-- that is, rather than "speaking" for the data, we would like the dataset to speak for itself. From a statistician's point of view, this often translates to restricting oneself to a \textit{nonparametric} model for the data generating mechanism. However, there are also times when one has prior knowledge about certain parts of the underlying mechanism, and incorporating that knowledge would help to improve both the quality and accuracy of the statistical models we fit. 




A simple way of incorporating such knowledge is to allow certain components of the model to be \textit{parametric}, rather than nonparametric. Typically, the nonparametric components reflect those aspects of the data generating mechanism that we cannot confidently make assumptions about, and so we must model them flexibly; in contrast, the parametric components reflect those aspects which we feel comfortable making strong assumptions about, and so we can model them rigidly. This is directly applicable to quantification tasks. In quantification tasks, the problem definition assumes that the hidden label of a test sample follows the $\textup{Cat}(\pi^\ast)$ distribution for some $\pi^\ast$ in the $m+1$ dimensional probability simplex, which is a parametric model. However, we do not know anything about the conditional density of the covariate given the label, except of course that they are unchanged between train and test by virtue of the label shift assumption; as such, those component densities are modeled nonparametrically.

\textbf{Semiparametric statistics} studies precisely these situations. In the language of that field, $\pi^\ast$ is the finite-dimensional \textbf{parameter of interest}, and $\mathbf{p}$ is the infinite-dimensional \textbf{nuisance parameter}. The goal is to learn the parameter of interest in an effective 
fashion. Since the nuisance parameter is generally hard to estimate because it is infinite-dimensional, this goal typically translates to learning the parameter of interest in a fashion that requires us to learn the \textit{least} about the nuisance parameter. In other words: how can we learn $\pi^\ast$, given that we don't care about learning $\mathbf{p}$?

The question stated above articulates both the attitude and objective of the quantification field, and it is for this reason that we believe using a semiparametric framework is appropriate. Moreover, for our own goals, this framework is also actively desirable: as discussed previously, we would like to prove that SELSE has the smallest possible asymptotic variance. So far, we have only been able to establish SELSE's optimality among ACC/PACC quantifiers, but ideally, we would like to universalize this result to quantifiers in general. Fortunately, the semiparametric literature is equipped with the tools to do exactly that. For an excellent and mathematically rigorous presentation of those tools, we refer the interested reader to the seminal book by \cite{bickel1993efficient}; for a simplified and more user-friendly version of those same tools, we point the reader to the paper by \cite{Newey1990}. 

\subsection{Derivation Roadmap} \label{sec: derivation roadmap}

Despite the aforementioned appeal of the semiparametric framework, one limitation of that literature is that its most important results tend to be isolated to regimes where the data is generated IID from a single population. This deviates from the traditional quantification setting which we had presented in the Methodology section, where due to label shift, the train and test sets are generated from \textit{two} different populations. As we developed our research, this put us in an especially awkward position: we saw that analyzing the asymptotic behavior of SELSE was conceptually easier to do in the traditional quantification setting, yet to show  that our variance was the smallest among a large class of quantifiers, it seemed like the best approach was to use the single population setting that was ubiquitous in the semiparametric literature.

Faced with this conflict, we decided to make the best of both worlds. On a high level, our approach was as follows. First, we established the asymptotic properties of SELSE in the traditional quantification setting, which included identifying SELSE's asymptotic variance matrix. In the Appendix, this setting is referred to as the "\textbf{Fixed Sequence Regime}". The results of those efforts have already been presented in subsection \eqref{Formalizing the Asymptotic Behavior of SELSE}. Second, we also separately considered a modified version of the traditional quantification setting, wherein all samples are generated IID from a single density, and a random indicator $D\in\{0,1\}$ determines whether a sample belongs to test or train; i.e., $D_i$ decides whether the $\numth{i}$ sample is drawn from $p_{\pi^\ast}$ or $p_{X,Y}^\train$. This setting is referred to as the "\textbf{Fixed $\tau$-IID Regime}" in the Appendix, because it is assumes that $D \sim \text{Bern}(\tau)$ for some fixed $\tau \in(0,1)$. Crucially, notice that the Fixed $\tau$-IID Regime preserves the essence of the quantification problem: we receive labeled train and unlabeled test samples, with the sole distributional discrepancy still being due to label shift. The only difference is that now the \textit{number} of train and test samples is also random. In this regime, our goal was to derive a preliminary version of the semiparametric efficiency bound-- the bound was "preliminary" because it could only apply to settings where $\tau$ was \textit{fixed}. This was inadequate for us, because we also wanted it to cover settings where the train set size grows far faster than the test set size (or vice versa), which requires $\tau \to 0$ (or $\tau\to 1$) as $n\to\infty$. However, we made the strategic choice to obtain the semiparametric efficiency bound for this simplified setting first, as it made achieving the desired, more flexible bound far easier.  


This is where the "\textbf{$\tau_n$-IID Regime}" comes in. This third regime is nearly identical to the Fixed $\tau$-Regime, differing only in that $\tau$ is replaced with $\tau_n$, which may change with $n$. In this regime, we had three goals. First, using the results from the Fixed Sequence Regime as our stepping stone, we translated the asymptotic properties of SELSE to the $\tau_n$-IID Regime. Second, we generalized the semiparametric efficiency bound in the Fixed $\tau$-Regime to accomodate the fact that $\tau_n$ changes with $n$. Third, combining the results from our first and second goals, we showed that SELSE's asymptotic variance matrix was equal to the aforementioned semiparametric efficiency bound. Ultimately, by using the three regimes in this way, we were able to show that SELSE is semiparametric efficient.

\subsection{Data Generation} \label{sec: Data Generation}

We now formally introduce the mechanism used in the $\tau_n$-IID Regime. As alluded to in subsection \eqref{sec: derivation roadmap}, the modification is conceptually straightforward: rather than observing $n^\test$ IID test samples from $p_{\pi^\ast}$ and $n^\train$ IID train samples from $p_{X,Y}^\train$, we now obtain $n \geq 1$ IID copies of a random vector $Z = (X,Y,D)$, which is generated as follows:

\vspace{0.2in}
\begin{itemize}
	\item $D \sim \text{Bern}(\tau_n)$
	\item If $D = 0$, then $(X,Y) \sim p_{X,Y}^\train$
	\item If $D = 1$, then $X \sim p_{\pi^\ast}$ and $Y = -1$.
\end{itemize}
\vspace{0.2in}

If $D = 0$, then $(X,Y)$ is a labeled training sample, and if $D = 1$, then $X$ is an unlabeled test sample (hence the dummy value of $Y = -1$). The probability of generating a test sample from this process is $\tau_n \in (0,1)$, which may depend on $n$. For simplicity, we assume that both $\tau_n$ and $\pi^\train$ are known. Based on the process displayed above, the density function for the distribution of $Z$ is given by
\begin{align} 
	\mathcal{J}^{\pi^\ast,\mathbf{p},\tau_n}(z) &:= \big( \tau_n p_{\pi^\ast}(x) \big)^d \big((1-\tau_n) p_{X,Y}^\train(x,y) \big)^{1-d} \ \Ibig{x\in\X, \text{ } y \in \Y, \text{ } d \in \{0,1\}  }  \nonumber \\ 
	 &= \big( \tau_n p_{\pi^\ast}(x) \big)^d \Bigg((1-\tau_n) \prod_{j=0}^{m} \big(\pi_j^\train p_j(x) \big)^{\I{y=j}} \Bigg)^{1-d} \Ibig{x\in\X, \text{ } y \in \Y, \text{ } d \in \{0,1\}  }. \label{The Real and True Jcal}
\end{align}

The final dataset is then $Z_1,\dots, Z_n \simiid \Jcal^{\pi^\ast, \mathbf{p}, \tau_n}$, with the number of test and train samples given by $N^\test := \sum_{i=1}^n D_i$ and  $N^\train := n - N^\test$, respectively. Although $N^\test$ and $N^\train$ are now random, recall that the benefit of working with this data generation mechanism is that all data points are from a \textit{single} population (i.e., $n$ IID samples from a single density), which is studied by the semiparametric literature far more often than settings where there are multiple sets of samples from different populations (i.e., $n^\test$ test samples and $n^\train$ train samples). This regime therefore enables us to make the most out of existing results in the semiparametric literature when proving our efficiency bound. 

\subsection{Model Specification}\label{semiparametric model specification}

Having described the new data generating mechanism, we now present several modeling assumptions. We assume that $\pi^\ast, \pi^\train \in \Delta$, where $\Delta$ is a subset of the $m+1$ dimensional probability simplex. We also assume that $\mathbf{p} \in \Q$, where $\Q := \Q_0\times\dots\times\Q_m$ and each $\Q_y$ is a set of densities. The full parameter space is $\Theta^\text{semi} = \Delta \times \Q$, and our semiparametric model is
\[
	\textbf{M}^{\text{semi},\tau_n} := \big\{ \Jcal^{\pi,\mathbf{q},\tau_n} \ \big\lvert \ (\pi,\mathbf{q}) \in \Theta^\text{semi} \big\},
\]
where for each parameter $(\pi,\mathbf{q}) \in \Theta^\text{semi}$, $\Jcal^{\pi,\mathbf{q},\tau_n}$ is a density function for the random vector $Z$ described in the previous subsection, except with $\mathbf{q}$ replacing $\mathbf{p}$ and $\pi$ replacing $\pi^\ast$ in equation \eqref{The Real and True Jcal}.


\subsubsection{Assumptions on $\pi^\ast$ and $\pi^\train$}\label{Assumptions on class proportions}

We make several restrictions on both $\Delta$ and $\Q$. For $\Delta$, we assume that there exists a constant $\xi \in (0,\tfrac{1}{2})$ such that 
\[
	\Delta := \Bigg\{(\pi_1,\dots,\pi_m)\in\mathbb{R}^m  \  \Bigg\lvert \ \pi_0 = 1- \sum_{y=1}^{m}\pi_y, \quad \xi < \pi _y< 1-\xi \quad \forall y\in\Y  \Bigg\}.
\]

Since $\pi^\ast, \pi^\train \in \Delta$, this form for $\Delta$ ensures that each class is sufficiently represented in both the train and test sets. For the test set, this is not strictly necessary, but it is reasonable if one chooses $\xi$ to be extremely small, and the assumption is convenient for our proofs.  However, this type of assumption \textit{is} necessary for the training set: clearly, learning about the mixing parameter used in $p_{\pi^\ast}$ is only possible if we have some knowledge about each of the component densities involved in the mixture. Requiring that $\pi_y^\train \in \Delta$ guarantees this, because it ensures that the average amount of training samples from each component density is greater than $n\xi$.

\subsubsection{Assumptions on $\mathbf{p}$} \label{subsection: Assumptions for p}

Next, we state assumptions about $\Q$. Towards that end, define:
\[
	L := \frac{1}{1+m\big( \frac{1-\xi}{\xi}\big)^3} \qquad U := \max\Bigg( 1-\xi, \ \frac{1}{1+m\big( \frac{\xi}{1-\xi}\big)^3}   \Bigg)
\]
\[
	\Gamma := \Bigg\{ (\gamma_1,\dots,\gamma_m) \in \mathbb{R}^m  \ \Bigg\lvert \    \gamma_0 = 1- \sum_{y=1}^{m}\gamma_y, \quad L \leq \gamma _y \leq U \quad \forall y\in\Y  \Bigg\}.
\] 

We assume that there exists constants $ \nu,\Lambda, B > 0$ such that the following conditions hold for all  $\mathbf{q}\in\Q$ and $\gamma \in \Gamma$:
\vspace{0.2in}
\begin{enumerate}[label={B\arabic*.}, ref=B\arabic*]
	\item $\minSing{ \FisherInfo{\gamma} } > \sqrt{\Lambda}$. \label{Fixed Tau Regime: Mixture FIM not too small}
	\item $\minSing{\FisherInfo{\gamma;\textup{Cat}} - \FisherInfo{\gamma}} > \nu$.  \label{Fixed Tau Regime: Cat minus Mixture FIM}
	\item $(q_0e^{\rho' \zeta_0s_{\gamma} - c_0(\rho) },\dots, q_me^{ \rho' \zeta_m s_{\gamma}  - c_m(\rho)} ) \in \Q$ for all $\rho \in \mathbb{R}^m$ satisfying $\norm{\rho}{2} < B$ and scalars $\zeta_j$ satisfying $0\leq \zeta_j < \frac{1-\xi}{\xi}$. \label{Fixed Tau Regime: expo family}
\end{enumerate}
\vspace{0.2in}

Assumptions  \eqref{Fixed Tau Regime: Mixture FIM not too small} and  \eqref{Fixed Tau Regime: Cat minus Mixture FIM} place bounds on the information that a test sample can contain about $\pi^\ast$. Here's how. On the one hand, Assumption  \eqref{Fixed Tau Regime: Mixture FIM not too small} ensures that no two component densities are the same: if it were the case that $\minSing{ \FisherInfo{\gamma} } = 0$ for any $\gamma\in\Gamma$, then it is easy to show that there would exist at least two component densities which are equal to each other with probability 1, and $\pi^\ast$ would be unidentifiable. A simple example of this unidentifiability comes from the binary class case: if $p_0 = p_1$, then $p_{\beta}$ would be the same for all $\beta \in (0,1)$, in which case it is impossible to learn $\pi^\ast$ from the test data. That is,  even though  $X\sim p_{\pi^\ast}$, $X$ contains zero information about $\pi^\ast$.

On the other hand, Assumption \eqref{Fixed Tau Regime: Cat minus Mixture FIM} ensures that the component densities' supports are not all disjoint. This insight comes from the following lemma:\\

\begin{lemma}[\uline{Relationship between $\FisherInfo{\beta}$ and $\FisherInfo{\beta;\textup{Cat}}$}] \label{Harder Problem}
Let $\beta$ be a vector in the $m+1$ dimensional probability simplex, and suppose that the entries in $\beta$ are bounded away from $0$ and $1$. Then, we have that
\[
	\FisherInfo{\beta} \preceq \FisherInfo{\beta;\textup{Cat}},
\]
where equality is achieved if and only if the supports of the class distributions are all disjoint, i.e., 
\[
	\FisherInfo{\beta;\textup{Cat}} =  \FisherInfo{\beta} \iff \P_{y}\Bigg[ \sum_{j=0\neq y}^{m}p_j(X) = 0 \Bigg] = 1 \qquad \forall y \in \Y. 
\]
\end{lemma}
\vspace{0.2in}


Lemma \eqref{Harder Problem} implies that equality between $\FisherInfo{\gamma;\textup{Cat}}$ and $\FisherInfo{\gamma}$ happens if and only if all the component densities' supports are disjoint; since Assumption \eqref{Fixed Tau Regime: Cat minus Mixture FIM} prevents this equality, it follows that \eqref{Fixed Tau Regime: Cat minus Mixture FIM} ensures that the component densities have at least \textit{some} overlap.

Actually, overlapping component densities is not a necessary requirement for establishing the performance of SELSE; indeed, the more separated the densities are, the better quantifiers tend to perform. This can be easily seen for SELSE by examining its asymptotic variance, as presented in Theorem \eqref{Semiparametric Efficiency in tau n IID Regime}. However, it \textit{is} necessary if we want the efficiency bound to be achievable. To see why, suppose that the supports were all disjoint. Then, every $x\in\X$ corresponds to a single class $y\in\Y$, i.e., $\varrho(x) = y$ for some deterministic function $\varrho$. It is easy to show that $\varrho$ is a sufficient statistic for $\pi^\ast$, implying that the information that each test sample $X$ contains about $\pi^\ast$ equals the information that the missing test label $Y$ contains about $\pi^\ast$ (hence the equality of Fisher Informations  in  Lemma \eqref{Harder Problem}). The fact that these information contents are equal \textit{despite} us having only imperfect knowledge about $\varrho$ via the training data is problematic: it implies that, in order for a quantifier to be "efficient", its first-order error must be \textit{just as small as} what is possible when $\varrho$ is known perfectly! Of course, this requirement can be satisfied if the number of training samples grows far faster than the number of test samples (i.e., $\tau_n \to 0$ sufficiently quickly), because then the error incurred for having to learn $\varrho$ will not be part of the first order error. However,  rather than place restrictions on the relationship between the train and test sample sizes, we prefer to simply assume \eqref{Fixed Tau Regime: Cat minus Mixture FIM}, since in most interesting problems, the supports of the component densities are not all disjoint.

Finally, to understand Assumption \eqref{Fixed Tau Regime: expo family}, it is helpful to first clarify ideas on what is meant by an "efficiency bound" in semiparametric statistics. Towards that end, suppose that $\M^{\text{sub},\tau_n}$ was such that $\Jcal^{\pi^\ast,\mathbf{p},\tau_n} \in\M^{\text{sub},\tau_n}\subseteq \M^{\text{semi},\tau_n}$. Intuitively, the task of learning $\pi^\ast$ (without necessarily learning $\mathbf{p}$) when the assumed model  is $\M^{\text{sub},\tau_n}$ cannot be harder than when the assumed model is $\M^{\text{semi},\tau_n}$, and this holds true for the special case of when the density vector $\mathbf{q}$ involved in defining each $\Jcal^{\pi,\mathbf{q},\tau_n} \in \M^{\text{sub},\tau_n}$ belongs to a smooth parametric family. Moreover, for any such smooth parametric submodel, this estimation difficulty can be measured by the \textbf{Cramer Rao Lower Bound (CRLB)}. Thus, it is intuitive to consider the supremum of all such CRLBs (one CRLB for each possible smooth parametric submodel of $\M^{\text{semi},\tau_n}$) as the semiparametric efficiency bound for learning $\pi^\ast$ when the assumed model is $\M^{\text{semi},\tau_n}$. 

Now ideally, we would want $\Q$ to be as large as possible, because this corresponds to making fewer assumptions about $\mathbf{p}$. However, making $\Q$ larger can only increase the semiparametric efficiency bound, i.e., make it more difficult to learn $\pi^\ast$, because it would expand the set of smooth parametric submodels that the aforementioned supremum indexes over. This is precisely where Assumption \eqref{Fixed Tau Regime: expo family} comes in. Under this assumption, it is possible to construct a parametric submodel with a CRLB that is greater than or equal to the CRLB of \textit{any other possible} parametric submodel. In the semiparametric literature, this is often called the "least favorable" submodel. This result is presented in Lemma \eqref{Largest CRLB}, with the specific submodel described in Lemma \eqref{A Particular Smooth Parametric Submodel}, both located in the Appendix section. This result is significant, because it means that if Assumption \eqref{Fixed Tau Regime: expo family}  holds true, then $\Q$ can be assumed to be arbitrarily large, without the extra expense of making it harder to learn $\pi^\ast$!

\subsubsection{Assumptions on $s_{\gamma^\ast}$} \label{subsection: Assumptions for score function}

In addition to Assumptions \eqref{Fixed Tau Regime: Mixture FIM not too small}, \eqref{Fixed Tau Regime: Cat minus Mixture FIM} and \eqref{Fixed Tau Regime: expo family}, we also place constraints on the score function formed from $\mathbf{p}$. Such assumptions are necessary because SELSE assumes that it is possible to learn $s_{\gamma^\ast}$.  However, before diving into those assumptions, it will be helpful to first establish some new notation.

Let $(\bar{n}^\test, \bar{n}_0^\train,\dots,\bar{n}_m^\train)$ denote any sequence depending on $n$ such that $\bar{n}^\test + \sum_{y=0}^m n_y^\train = n$, with each component approaching infinity. For each $(\pi,\mathbf{q}) \in \Theta^\text{semi}$, let $\barDtrainI$ and $\barDtrainII$ denote datasets each containing $\tfrac{1}{2} \bar{n}_y^\train$ IID samples from $q_y$ for every $y\in\Y$, and let $\barDtestI$ and $\barDtestII$ denote datasets each containing $\tfrac{1}{2}\bar{n}^\test$ IID samples from $q_{\pi}$. Define $\barallDI := (\barDtrainI,\barDtestI)$ and $\barallDII := (\barDtrainII,\barDtestII)$. These datasets are analogous to what was described in the Methodology Section. Also define $\bar{\gamma} := (\bar{\gamma}_1,\dots,\bar{\gamma}_m)$ where
\begin{equation}\label{definition of gamma bar}
	\bar{\gamma}_y := \frac{\frac{\pi_y}{\bar{n}^\test} +  \frac{(\pi_y)^2}{\bar{n}_y^\train }  }{ \frac{1}{\bar{n}^\test} +  \sum_{k=0}^m \frac{( \pi_k)^2}{ \bar{n}_k^\train  } } \qquad 	\forall \ y\in\Y,
\end{equation}
and define
\[
	\H^\text{pre} := \Bigg\{ s_{\gamma}     \ \Bigg\lvert \  s_{\gamma} := \frac{ (q_1 - q_0,\dots,q_m - q_0)' }{\sum_{j=0}^m \gamma_j q_j}, \quad (q_0,\dots,q_m)\in \Q, \quad \gamma \in \Gamma    \Bigg\}.
\]
Finally, let $\H$ be any set for which $\H^\text{pre} \subseteq \H$. We assume that:

\vspace{0.2in}
\begin{enumerate}[resume*]
	\item $\H$ is closed under scalar multiplication. \label{Fixed Tau Regime: H closed under scalar multiplication} 
	\item For each $j\in[m]$, $\Omega:\H_j \mapsto [0,\infty)$ satisfies the following properties for each $h_j\in\H_j$:  \label{Fixed Tau Regime: Omega Properties} 
	\begin{itemize}
		\item  $\Omega(ch_j) = \abs{c}\Omega(h_j)$ for all $c\in\mathbb{R}$
		\item  $\Omega(h_j) \geq D \norm{h_j}{\infty}$ where $D > 0$ is a global constant.
	\end{itemize}	
\end{enumerate}
\vspace{0.2in}
We also make the following assumptions about the complexity of $\H$:

\vspace{0.2in}
\begin{enumerate}[resume*]
	\item For each $j \in [m]$, $(\pi,\mathbf{q}) \in \Theta^\text{semi}$ and sequence  $(\bar{n}^\test, \bar{n}_0^\train,\dots,\bar{n}_m^\train)$, we assume that there exists a corresponding sequence $\lambda_j \in (0,\infty)$ which satisfies the following properties: \label{Tau n Regime: lambda and uniform convergence properties}
	\begin{itemize}
		\item $\lambda_j\sup_{\gamma\in\Gamma} \Omega^2\Big( \frac{q_j - q_0}{q_{\gamma}} \Big)  \to 0$
		\item $\lambda_j^2 \sum_{k=0}^m \bar{n}_k^\train  \to \infty$
		\item $\E_{\barDtrainII}\Big[ \bar{\U}_j(\lambda_j)\Big] \to 0$, where
		\begin{align*}
			\bar{\U}_j(\lambda_j) &:=  \sup\limits_{ \Omega(h_j) \leq \frac{1}{L\sqrt{\lambda_j} } } \absBig{\widehat{\E}_{q_0}^\trainII[h_j] - \E_{q_0}[h_j] }  + \sup\limits_{ \Omega(h_j) \leq \frac{1}{L\sqrt{\lambda_j} } } \absBig{\widehat{\E}_{q_j}^\trainII[h_j] - \E_{q_j}[h_j] } \\
			&\qquad\qquad+ \sup\limits_{ \Omega(h_j) \leq \frac{1}{L\sqrt{\lambda_j} } } \absBig{\widehat{\Var}_{    q_{\bar{\gamma}}  }^\trainII[h_j] - \Var_{  q_{\bar{\gamma}}  } [h_j]  }. 
		\end{align*}
	\end{itemize}
\end{enumerate}
\vspace{0.2in}

$\H^\text{pre}$ is the set of all possible score functions that can be created from density vectors $\mathbf{q} \in \Q$ (hence $s_{\gamma^\ast} \in \H^\text{pre}$), so ideally, we would choose $\widehat{s}_{\widehat{\gamma}}$ in Step 3 of SELSE  by running the optimization routine on $\H^\text{pre}$. However, since there is no reason one should know $\H^\text{pre}$ ahead of time, we instead run the optimization on a more conveniently known function space $\H$ that  is a superset of $\H^\text{pre}$. Fortunately, many of the common choices for spaces $\H$ and penalty terms $\Omega$ satisfy Assumptions  \eqref{Fixed Tau Regime: H closed under scalar multiplication} and \eqref{Fixed Tau Regime: Omega Properties}. For example, if $\H$ is a Sobolev Space or a RKHS, and $\Omega$ the appropriate norm, then both multiplication properties  in  \eqref{Fixed Tau Regime: H closed under scalar multiplication} and \eqref{Fixed Tau Regime: Omega Properties} hold true. The condition involving the $L_\infty$ norm is also satisfied in both cases; see Theorem 8.8 in \cite{SobolevSpaceNormInequality} for when $\H$ is a Sobolev space and see page 124 in \cite{KernelReproducingProperty} for when $\H$ is a RKHS with a bounded reproducing kernel.


Even though $\H^\text{pre}$ is unknown, the hope is that this superset property holds if $\H$ is chosen to be a sufficiently rich and complex function class which is allowed to grow with $n$. Indeed, the larger $\H$ is, the more likely the superset property is satisfied. However, $\H$ also cannot be allowed to grow "too" fast: based on how $\widehat{s}_{\widehat{\gamma},j}$ is constructed for each $j\in[m]$, the quality of  $\widehat{s}_{\widehat{\gamma},j}$ as an  estimator of $s_{\gamma^\ast,j}$ hinges on the degree to which $\E_{q_0}[h_j]$, $\E_{q_j}[h_j]$ and $\Var_{q_{\bar{\gamma}}}[h_j]$ can be well approximated by their empirical counterparts for each $h_j \in \H_j$. Indeed, if $\H_j$ is too large relative to the sample size, then this approximation may be poor for some $h_j$. This is an interesting form of tension, and has been studied for decades within statistical learning theory in the context of excess risk bounds for classification and regression tasks \citep{vapnik1999nature}. In line with that tradition, our algorithm addresses the foregoing concern via regularization in Step 3, which effectively constrains the optimization over each $\H_j$ to a ball in which $\Omega(h_j)$ is bounded. Assumption  \eqref{Tau n Regime: lambda and uniform convergence properties} requires that this ball's radius tends to infinity, but at a rate that is sufficiently slow, so as to ensure that the first and second moments of those $h_j$ inside the ball can be accurately learned.


Finally, even if $\H$ grows extremely slowly, the particular form of the score functions in $\H^\text{pre}$ may actually be enough to ensure that the superset property holds for small $n$. This is because each member of $\H^\text{pre}$ only depends on the \textit{ratio} between the component densities, not the actual densities themselves. For example, when $m=1$, it is easy to show that
\[
	\frac{q_1(x) - q_0(x)}{q_{\gamma}(x)} = \frac{1}{\gamma + \frac{1}{ q_1(x)/q_0(x) -1}}\qquad \forall x\in\X,
\]
meaning that the score is an elementary transformation of the density ratio $q_1(x)/q_0(x)$, with a similar result holding for the more general $m\geq1$ setting. Essentially, this means that even if SELSE is applied to situations where the component densities are complicated, as long as their \textit{density ratios} are not too complex, then the corresponding score functions will not be too complex either. Hence, in these situations, even if $\H$ is not an expressive function class, it may be reasonable to assume that $\H^\text{pre} \subset \H$. An example of one of these situations is when the component densities belong to the same exponential family with an elementary sufficient statistic function $T$, but a complicated reference density. In this case, the reference density causes $q_0,q_1$ to be complicated functions, yet for some constant $c$, we have that $q_1(x) / q_0(x) \propto e^{\dotprod{c}{T(x)}}$, which is relatively simple as it only depends on $T$.

\subsubsection{Assumptions on $\widehat{\gamma}$} \label{subsection: Assumptions about gamma hat}

We also require that there exists an estimator $\widehat{\gamma}$ with the following properties:

\vspace{0.2in}
\begin{enumerate}[resume*]
	\item For each $\mathbf{q}\in\Q$, $\pi\in\Delta$ and sequence $(\bar{n}^\test, \bar{n}_0^\train,\dots,\bar{n}_m^\train)$, there exists an estimator $\widehat{\gamma}$ of $\bar{\gamma}$ such that: \label{tau n regime, estimator of gamma bar}
	\begin{itemize}
		\item $\widehat{\gamma} \in \Gamma$ for each $n$
		\item $\E_{ \barallDII }\Big[ \norm{\widehat{\gamma} - \bar{\gamma}}{2}^2 \Big] \to 0$
		\item For each $a > 0$, $\P_{ \barallDII}\big[ \norm{\widehat{\gamma} - \bar{\gamma}}{2}\geq a\big] = o\Big(\sqrt{\frac{1}{\bar{n}^\test} + \frac{1}{\sum_{k=0}^m \bar{n}_k^\train }}\Big) $
	\end{itemize}
\end{enumerate}
\vspace{0.2in}

These assumptions are easily satisfied when we have access to a preliminary quantifier for $\pi^\ast$ which can be used with equation \eqref{definition of gamma bar} to form a plug-in estimate $\widehat{\gamma}$ of $\bar{\gamma}$, much like in our original algorithm.

\subsubsection{Assumptions on $\tau_n$}

We also make the following assumptions about $\tau_n$:

\vspace{0.2in}
\begin{enumerate}[resume*]
	\item $\tau_n \in (0,1)$ for each $n$. \label{tau n is inside open interval 0 and 1}
	\item $n\tau_n(1-\tau_n)\to\infty$. \label{tau n convergence not too fast}
\end{enumerate}
\vspace{0.2in}

Assumption \eqref{tau n is inside open interval 0 and 1} ensures that the dataset contains both test and train samples with high probability. Assumption \eqref{tau n convergence not too fast} ensures that the expected number of test and train samples goes to infinity, even if the expected proportion of the data that is test or train goes to zero in the limit; that is, both $n\tau_n,n(1-\tau_n)\to\infty$ even if $\tau_n \to 1$ or $\tau_n \to 0$. These conditions enable us to consider many different sequences of $\tau_n$, creating a flexible model for the relationship between the two sample sizes. Examples of such sequences include:

\vspace{0.2in}
\begin{itemize}
	\item $\tau_n = \frac{1}{n^\alpha}$ for some $\alpha \in (0,1)$
	\item $\tau_n = 1 - \frac{1}{n^\alpha}$ for some $\alpha \in (0,1)$
	\item $\tau_n = \frac{1}{2}$.
\end{itemize}
\vspace{0.2in}

The first two choices for $\tau_n$ describe a relationship between the train and test sample sizes, wherein one grows faster than the other. On the one hand, when $\tau_n = \frac{1}{n^\alpha}$,  it can be shown that $ \E\big[ N^\train \big] \asymp \E\big[ N^\test\big]^{\frac{1}{1-\alpha}} $ where $\frac{1}{1-\alpha} \in (1,\infty)$, meaning that the average number of train samples grows polynomially in the average number of test samples. As such, $\E[N^\train] >> \E[N^\test]$. On the other hand, when $\tau_n = 1 - \frac{1}{n^\alpha}$, then $\E\big[ N^\test\big] \asymp \E\big[ N^\train \big]^{\frac{1}{1-\alpha}}$, i.e., the average number of test samples grows polynomially in the average number of train samples. As such, $\E[N^\test] >> \E[N^\train]$. Later in this section, we will interpret the semiparametric efficiency bound in  situations where $\tau_n \to 0$ and $\tau_n \to 1$, and show how in each setting, the task of quantification reduces to other statistical estimation problems. The simple case of when $\tau_n$ equals or converges to a constant in $(0,1)$ can then be thought of as an intermediate "blend" of those two extremes.

\subsection{Family of Quantifiers} \label{sec: Family of Quantifiers}

Next, we describe a large family of quantifiers $\curlyE$ whose asymptotic variance matrix is lower bounded by our semiparametric efficiency bound, which will be presented in the subsequent subsection. This family is extremely large, containing not only SELSE and ACC/PACC quantifiers, but also other popular competitors such as HDy, EMQ and MLLS.   We say that a quantifier $\widetilde{\pi}_n$ is contained in $\curlyE$ if $\widetilde{\pi}_n$ satisfies the following criteria:

\vspace{0.2in}
\begin{enumerate}[label={C\arabic*.}, ref=C\arabic*]
	\item For each fixed $\tau \in (0,1)$, $\widetilde{\pi}_n$ is regular with respect to $\M^{\text{semi},\tau}$ and asymptotically linear with influence function $\psi_{\tau}$ and second order error $\delta_{\tau}$. \label{Definition of E estimator: it should be tau RAL for each tau}
	\item If $\tau_n$ is a sequence satisfying Assumptions \eqref{tau n is inside open interval 0 and 1} and \eqref{tau n convergence not too fast}, then: 
	\begin{enumerate}
		\item $\P_{Z_{1:n}\simiid \Jcal^{\pi^\ast,\mathbf{p}, \tau_n } }\Big[ \absbig{\delta_{\tau_n }(Z_{1:n})} \geq c \Big] \to 0$ for all $c > 0$. \label{Definition of E estimator: second order error to zero with changing tau n}
		\item $\frac{(1-\tau_n)^2}{\tau_n n}\E_{\pi^\ast} \Big[  \norm{\tau_n \psi_{\tau_n}(X,-1,1)}{2}^4  \Big] + \frac{\tau_n^2}{(1-\tau_n)n} \sum\limits_{y=0}^m \pi_y^\train \E_y \Big[  \norm{(1-\tau_n) \psi_{\tau_n}(X,y,0)}{2}^4  \Big] \to 0$. \label{Definition of E estimator: moment condition on influence function with changing tau n}
	\end{enumerate}
\end{enumerate}
\vspace{0.2in}

Regularity and asymptotic linearity are important concepts in semiparametric statistics. Intuitively, a \textbf{regular} estimator is one that is not systematically biased towards certain values of the estimand. For example, quantifiers that are set equal to or shrunk towards a specific estimand value, or are superefficient for particular values (cf. Hodge's estimator) are all non-regular. The idea is that, by having an a priori preference towards certain estimand values, these quantifiers are making use of information contained \textit{outside} of the semiparametric model, and thus should be excluded from consideration \citep{Newey1990}. Further, a quantifier that is \textit{asymptotically linear} with influence function $\psi_{\tau}$ and second order error $\delta_{\tau}$ is one whose normalized estimation error can be written as
\[
	\sqrt{n}(\widetilde{\pi}_n - \pi^\ast) = \frac{1}{\sqrt{n}} \sum_{i=1}^n \psi_\tau(Z_i)  + \delta_\tau(Z_{1:n}),
\]
where $\psi_\tau,\delta_\tau$ are $\mathbb{R}^m$-valued functions,  $\psi_\tau(Z_i)$ is mean zero with a finite and nonsingular covariance matrix, and $\delta_\tau(Z_{1:n}) \goesto{p} 0$. Quantifiers of this form are unbiased to the first order, and more specifically, have an asymptotic error distribution of $\N\big(0, \Var_{ \Jcal^{\pi^\ast,\mathbf{p}, \tau } }[\psi_\tau] \big)$. This leads to a natural way of comparing asymptotically linear quantifiers:  if $\psi_{\tau,1}$ and $\psi_{\tau,2}$ are influence functions for two different quantifiers and $\Var_{ \Jcal^{\pi^\ast,\mathbf{p}, \tau } }[\psi_{\tau,1}] \preceq  \Var_{ \Jcal^{\pi^\ast,\mathbf{p}, \tau } }[\psi_{\tau, 2}]$, then the quantifier with $\psi_{\tau,1}$ as its influence function is more desirable, since its first order error is stochastically smaller. Traditionally, the influence function with the smallest variance matrix is called the \textbf{efficient influence function}. The variance matrix of the efficient influence function is called the \textbf{semiparametric efficiency bound}, and  a quantifier which is asymptotically linear in the efficient influence function is called \textbf{semiparametric efficient}, because its first order variance achieves the semiparametric efficiency bound. For a rigorous treatment of regularity and asymptotic linearity, we refer the reader to \cite{Newey1990} and \cite{bickel1993efficient}.

While condition \eqref{Definition of E estimator: it should be tau RAL for each tau} places requirements on the asymptotic behavior of quantifiers for \textit{fixed} $\tau \in (0,1)$, we would also like to analyze this behavior in settings where one of the train or test sample sizes grows far faster than the other, i.e., when $\tau_n$ approaches either $0$ or $1$. Since the influence function and the second order error may vary for different $\tau$, analyzing the foregoing behavior requires placing restrictions on $\{(\psi_\tau,\delta_\tau) \mid \tau \in (0,1) \}$. This is where conditions \eqref{Definition of E estimator: second order error to zero with changing tau n} and \eqref{Definition of E estimator: moment condition on influence function with changing tau n} come in. By requiring that  $ \absbig{\delta_{\tau_n }(Z_{1:n})}\goesto{p} 0 $ even as $\tau_n$ changes, condition \eqref{Definition of E estimator: second order error to zero with changing tau n} ensures that the asymptotic variance of $\widehat{\pi}_n$ is completely determined by $\psi_{\tau_n}$. On top of that, condition \eqref{Definition of E estimator: moment condition on influence function with changing tau n} is a simple moment condition that controls the size of the influence function. It is also easy to satisfy: for example, the condition holds whenever the norm of the influence function is uniformly bounded.

\subsection{Intermediate Results} \label{sec: intermediate results}

Having fully described our semiparametric model and the assumptions we make about its parameter space, we now walk through some of the intermediate results that were key to establishing SELSE's semiparametric efficiency. Towards that end, we start by taking a deeper dive into the concept of a semiparametric efficiency bound, which we had introduced in subsection \eqref{subsection: Assumptions for p}. This will be done first for the Fixed $\tau$-IID Regime, and generalized later on to the $\tau_n$-IID Regime.

\subsubsection{Fixed $\tau$-IID Regime}\label{sec: fixed tau iid regime intermediate results}

As mentioned previously, an intuitive way to think about a semiparametric efficiency bound is that it is the largest possible CRLB for a smooth parametric submodel, i.e., it is the CRLB for the least favorable parametric submodel. However, for any given smooth parametric submodel, what \textit{exactly} is the form of the CRLB? Answering this question will be quite helpful for our purposes, so we turn to that now.

Recall that a parametric submodel differs from a semiparametric model in that the infinite dimensional nuisance parameter in the semiparametric model is now parameterized by a finite dimensional vector. That is, in the context of the Fixed $\tau$-IID Regime, a parametric submodel $\M^{\text{sub},\tau} \subset \M^{\text{semi},\tau}$ assumes that $\mathbf{p}$ belongs to a parametric family. Accordingly, each density vector $\mathbf{q}$ in this family can be parameterized by some $\rho \in \R$, where $\R$ is some open set of finite dimensional vectors. The full parameter space for general $(\pi, \rho)$ is then $\Delta \times \R$, and the submodel is given by $\M^{\text{sub},\tau} = \{ \Jcal^{\pi,\rho,\tau} \mid (\pi, \rho) \in \Delta \times \R\}$. For example, if the covariates were all one dimensional and we were confident that the component densities were gaussian pdfs with unit variance, then $\R = \mathbb{R}^{m+1}$ and each $\rho\in\R$ corresponds to a different vector of $m+1$ means, with one mean for each of the $m+1$ classes. 

Now, the fact that the parameter space is finite dimensional for $\M^{\text{sub},\tau}$ means that the score function for the joint vector $(\pi,\rho)$ is easy to compute: it is simply $S_{\pi,\rho}(z) := (S_\pi(z), S_\rho(z))$, where 
\begin{align}
	S_\pi(z) &:= \partialDerivative{\pi}  \log \Jcal^{\pi, \rho^\ast, \tau}(z)  \nonumber \\
	&= ds_{\pi}(x) \label{really I dont know what to name things anymore}
\end{align}
and 
\[
	S_{\rho}(z) := \partialDerivative{\rho}  \log \Jcal^{\pi^\ast, \rho, \tau}(z),
\]
where $\rho^\ast$ is the parameter vector that the parametric submodel identifies with $\mathbf{p}$. Note that our use of the word "score" here is in line with the semiparametric literature, but it should not be confused with the score function $s_{\pi}(x)$ that we have typically referred to throughout this thesis, although they are related by line \eqref{really I dont know what to name things anymore}. $S_\pi$ is the score function for the parameter of interest, and $S_{\rho}$ is the score function for the nuisance parameter (aka the \textbf{nuisance score function}). Intuitively, $S_\pi$ captures the sensitivity of the density $\Jcal^{\pi, \rho^\ast, \tau}$ to changes in $\pi$, or equivalently, the degree of dependence of $\Jcal^{\pi, \rho^\ast, \tau}$ on $\pi$, when the component densities are fixed at the truth $\mathbf{p}$. Likewise, $S_\rho$ captures the degree of dependence of $\Jcal^{\pi^\ast, \rho, \tau}$ on $\rho$, when the parameter of interest is fixed at the truth $\pi^\ast$. This perspective encourages an interesting interpretation of the magnitude for both types of score functions. On the one hand,  one can think of the magnitude of $S_{\pi^\ast}(Z)$ as a measure of the amount of information contained in a random variable $Z \sim \Jcal^{\pi^\ast,\rho^\ast,\tau}$ about $\pi^\ast$ when $\rho^\ast$ is already known. On the other hand, one can think of the magnitude of $S_{\rho^\ast}(Z)$ as a measure of the amount of information contained in $Z \sim \Jcal^{\pi^\ast,\rho^\ast,\tau}$ about $\rho^\ast$ when $\pi^\ast$ is already known. 

These information-centric interpretations of $S_{\pi^\ast}$ and $S_{\rho^\ast}$ are vital for understanding the CRLB in a parametric submodel.  As argued in \cite{Newey1990}, one way to interpret the CRLB for learning $\pi^\ast$ (without necessarily learning $\mathbf{p}$) is that it is \textit{the inverse of the variance matrix of the residual we obtain from projecting $S_{\pi^\ast}(Z)$ onto $S_{\rho^\ast}(Z)$}, where in this case, the projection is in terms of mean-square distance, which is generally defined as $\E \norm{V_1 - V_2}{2}^2$ for any two random vectors $V_1,V_2$. Notationally, this means that for some conformable matrix $B$, the aforementioned CRLB is equal to 
\begin{equation} \label{interpreting CRLB as inverse variance of a projection}
	\bigg( \E_{\Jcal^{\pi^\ast,\mathbf{p},\tau}}\Big[ \big( S_{\pi^\ast} - B S_{\rho^\ast} \big) \big( S_{\pi^\ast} - B S_{\rho^\ast}\big)^T \Big] \bigg)^{-1}
\end{equation}
In other words, from a semiparametric perspective, the average information content contained in a single sample $Z$ about $\pi^\ast$ absent a priori knowledge about $\mathbf{p}$ is equal to the variance (i.e., average magnitude) of what is left over from $S_{\pi}(Z)$ \textit{after we remove} the part of $S_{\pi}(Z)$  that lies in the same direction as $S_{\rho^\ast}(Z)$. Based on our previous interpretations of $S_\pi$ and $S_\rho$, this "removed" part of $S_{\pi}(Z)$ can be thought of in terms of $\log\Jcal^{\pi^\ast,\rho,\tau}$'s sensitivity to changes in $\pi^\ast$: that is, the "removed" part is the part of that sensitivity which had been due to freezing $\rho$ at $\rho^\ast$, as opposed to any other member of $\R$. 


This interpretation of a CRLB is useful for understanding how we were able to identify the semiparametric efficiency bound in the Fixed $\tau$-IID Regime. As mentioned earlier, the semiparametric bound can be thought of as the supremum of CRLBs for estimating $\pi^\ast$ in the presence of unknown $\mathbf{p}$, where the supremum indexes over all possible smooth parametric submodels. While intuitive as a definition, this does not lend itself to easy evaluation-- indeed, how would one even find that supremum? The trick lies in the CRLB interpretation we established in the previous paragraph. First, note that $S_{\pi^\ast}$ is the same for all smooth parametric submodels in question; this is because in the definition of $S_{\pi^\ast}$, the finite dimensional nuisance parameter $\rho$ is always frozen at $\rho = \rho^\ast$, and the underlying truth $\mathbf{p}$ is the same for all submodels. Thus, to make the aforementioned residual as small as possible, we need to choose a parametric family for $\mathbf{p}$ for which the distance between $S_{\pi^\ast}(Z)$ and the resulting linear span of the nuisance score is as small as possible. However, rather than looking over all possible parametric families separately and computing the residual each time so as to find the one that is the smallest stochastically, one should note that the minimum of those individual residuals is the same as the residual obtained from the projection of $S_{\pi^\ast}(Z)$ onto the \textit{union} of all the aforementioned nuisance scores. Indeed, this is a simple consequence of the fact that a projection is always based on the shortest distance between $S_{\pi^\ast}(Z)$ and the set in question. 

Of course, in semiparametric statistics, this "union" is replaced by a more formal notion of taking the mean square closure of all $m$-dimensional linear combinations of nuisance score functions, but the fundamental idea is the same. This mean square closure is often called the \textbf{nuisance tangent set}, and once one is able to identify the residual obtained from projecting $S_{\pi^\ast}$ onto this set, finding the semiparametric efficiency bound becomes a routine calculation that mimics what is done in display \eqref{interpreting CRLB as inverse variance of a projection} for the CRLB of smooth parametric submodels. Indeed, as \cite{Newey1990} states in their Theorem 3.2, the semiparametric  bound is precisely the inverse of the variance matrix of the aforementioned residual!

At last, having motivated the steps necessary to identify the semiparametric efficiency bound for the fixed $\tau$-IID Regime, we can finally walk through those exact steps as it applies to quantification. As mentioned in the previous paragraph, there are two steps. The first step consists of identifying the residual of the  projection of $S_{\pi^\ast}(Z)$ onto the nuisance tangent set. This is concisely summarized via the following lemma. \\

\begin{lemma}[\uline{Projection of $S_{\pi^\ast}$ onto $\T$}] \label{Projection of Ordinary Score onto T}
Let $\T$ denote the semiparametric nuisance tangent set defined in the Appendix. Under Assumptions \eqref{Fixed Tau Regime: Mixture FIM not too small}, \eqref{Fixed Tau Regime: Cat minus Mixture FIM} and \eqref{Fixed Tau Regime: expo family},  we have that 
\[
	d s_{\pi^\ast} - \mathbf{V}^\textup{eff}(\tau)^{-1} \psi_\tau^\textup{eff} \in \T
\]
and 
\[
	\E_{\Jcal^{\pi^\ast,\mathbf{p}, \tau}} \Big[ t' \big(S_{\pi^\ast} - \big( Ds_{\pi^\ast} -  \mathbf{V}^\textup{eff}(\tau)^{-1}\psi_\tau^\textup{eff} \big) \big)\Big] = 0 \qquad \forall \ t\in\T,
\]
where
\begin{align*}
	\psi_\tau^\textup{eff}(z) &:= \frac{d}{\tau} \FisherInfo{\gamma^\ast(\tau)}^{-1}\big(s_{\gamma^\ast(\tau)}(x)  -  \E_{\pi^\ast}[s_{\gamma^\ast(\tau)}]\big) \\
	&\qquad - \frac{1-d}{1-\tau} \FisherInfo{\gamma^\ast(\tau)}^{-1} \sum_{j=0}^m \I{y=j} \frac{\pi_j^\ast}{\pi_j^\train}\big(s_{\gamma^\ast(\tau)}(x)  -  \E_{j}[s_{\gamma^\ast(\tau)}]\big)
\end{align*}
and 
\begin{align*}
	\mathbf{V}^\textup{eff}(\tau) := \bigg[ \frac{1}{\tau} + \frac{1}{1-\tau} \sum_{k=0}^{m} \frac{(\pi_k^\ast)^2}{\pi_k^\train} \bigg]\Big(\FisherInfo{ \gamma^\ast(\tau) }^{-1} - \FisherInfo{\gamma^\ast(\tau);\textup{Cat}}^{-1} \Big) + \frac{1}{\tau}\FisherInfo{\pi^\ast;\textup{Cat}}^{-1}.
\end{align*}
\end{lemma} 
\vspace{0.2in}

Lemma \eqref{Projection of Ordinary Score onto T} tells us that the projection of $S_{\pi^\ast}$ onto the nuisance tangent set $\T$ is given by $D s_{\pi^\ast}(X) - \mathbf{V}^\textup{eff}(\tau)^{-1} \psi_\tau^\textup{eff}(Z)$, and that the residual from this projection is equal to $\mathbf{V}^\textup{eff}(\tau)^{-1} \psi_\tau^\textup{eff}(Z)$. The proof for this lemma can be found in the Appendix. Next, having identified the residual, the second step is to compute the inverse of the variance of $\mathbf{V}^\textup{eff}(\tau)^{-1} \psi_\tau^\textup{eff}(Z)$ when $Z \sim \Jcal^{\pi^\ast,\mathbf{p},\tau}$, as the resulting matrix will be equal to the semiparametric efficiency bound. This is a simple but algebraically intense calculation, the results of which are summarized in the following theorem. \\

\begin{theorem}[\uline{Semiparametric Efficiency Bound in Fixed $\tau$-IID Regime}] \label{Efficiency Bound}
Under Assumptions \eqref{Fixed Tau Regime: Mixture FIM not too small},  \eqref{Fixed Tau Regime: Cat minus Mixture FIM} and \eqref{Fixed Tau Regime: expo family}, the efficient influence function is $\psi^\eff_\tau$,  the semiparametric efficiency bound is $\mathbf{V}^\textup{eff}(\tau)$, and $\mathbf{V}^\textup{eff}(\tau) = \Var_{\Jcal^{\pi^\ast, \mathbf{p},\tau}}[\psi^\eff_\tau]$.
\end{theorem} 
\vspace{0.2in}

Theorem \eqref{Efficiency Bound} identifies $\mathbf{V}^\textup{eff}(\tau)$ as the semiparametric efficiency bound for quantifiers that are regular and asymptotically linear within the Fixed $\tau$-IID Regime. This result is extremely encouraging, because upon inspection, $\mathbf{V}^\textup{eff}(\tau)$ is the Fixed $\tau$-IID Regime's analogue of the variance matrix identified for SELSE in the Fixed Sequence Regime within subsection \eqref{Formalizing the Asymptotic Behavior of SELSE} in Theorem \eqref{Standardized Error Rate of Full Estimator}. Furthermore, in establishing this result, we were also able to identify $\psi_\tau^\eff$ as the efficient influence function, which we had first mentioned in subsection \eqref{sec: Family of Quantifiers}. Also, recall from that same subsection that the variance of the efficient influence function should be equal to the semiparametric efficiency bound. As a sanity check, Theorem \eqref{Efficiency Bound} verifies that this is indeed the case.

\subsubsection{$\tau_n$-IID Regime} \label{sec: tau n iid regime intermediate results}

The parallel between the asymptotic variance matrix identified for SELSE in the Fixed Sequence Regime and the semiparametric efficiency bound identified in the Fixed $\tau$-IID Regime was exciting, because it suggested that the two variance matrices would align upon translating our results into the final $\tau_n$-IID Regime. 

Encouraged by this, we first focused on the translation for the asymptotic results of SELSE. While algebraically tedious, the strategy was quite straightforward. Essentially, the main insight comes from recognizing that the stochastic behavior of SELSE in the Fixed Sequence Regime can be thought of as its behavior in the $\tau_n$-IID Regime \textit{when we condition} on the event that $N^\train = (1-\tau_n)n$ and $N^\test =  \tau_n n$. Since $N^\train/n$ and $N^\test/n$ will concentrate around $1-\tau_n$ and $\tau_n$ respectively, it follows that the behavior of SELSE in the Fixed Sequence Regime based on setting $n^\train = (1-\tau_n)n$ and $n^\test =  \tau_n n$ is representative of its behavior in the $\tau_n$-IID Regime. Hence, combining this intuition with Theorem \eqref{Standardized Error Rate of Full Estimator} from the Fixed Sequence Regime, we were well positioned to prove the following lemma.\\

\begin{lemma}[\uline{First and Second Order Error of SELSE in $\tau_n$-IID Regime}] \label{Behavior of First and Second Order Error Terms in tau n IID regime}
Under Assumptions \eqref{Fixed Tau Regime: Mixture FIM not too small}, \eqref{Fixed Tau Regime: Cat minus Mixture FIM},  \eqref{Fixed Tau Regime: H closed under scalar multiplication}, \eqref{Fixed Tau Regime: Omega Properties},  \eqref{Tau n Regime: lambda and uniform convergence properties}, \eqref{tau n regime, estimator of gamma bar}, \eqref{tau n is inside open interval 0 and 1} and \eqref{tau n convergence not too fast}, the SELSE quantifier $\widehat{\pi}_n$ satisfies
\[
	\sqrt{n}(\widehat{\pi}_n - \pi^\ast) = \frac{1}{\sqrt{n}} \sum_{i=1}^n \psi_{\tau_n}^\eff(Z_i) + \delta_{\tau_n}
\] 
where $\psi_{\tau_n}^\eff:\Z\mapsto\mathbb{R}^m$ is as defined in Lemma \eqref{Projection of Ordinary Score onto T} and satisfies
\[
	\frac{(1-{\tau}_n)^2}{{\tau}_n n}\E_{\pi^\ast} \Big[  \norm{{\tau}_n \psi^\eff_{{\tau}_n}(X,-1,1)}{2}^4  \Big] + \frac{{\tau}_n^2}{(1-{\tau}_n)n} \sum_{y=0}^m \pi_y^\train \E_y \Big[  \norm{(1-{\tau}_n) \psi^\eff_{{\tau}_n}(X,y,0)}{2}^4  \Big] \to 0,
\]
and $\delta_{\tau_n}\in\mathbb{R}^m$ satisfies 
\begin{align*}
	\norm{\delta_{\tau_n}}{2} &= O_{\P}\Bigg( \frac{1}{\sqrt{n\tau_n(1-\tau_n)} }   +   \sqrt{  \frac{1}{\sqrt{\bar{n}^\train}} + \max_y \lambda_y \Omega(s_{\bar{\gamma},y})^2 +  \frac{1}{\sqrt{\min_y\lambda_y}}e^{-C\bar{n}^\train}    } \Bigg) \\
	&\qquad+ O_{\P}\Bigg( \sqrt{  \max_y \E_{\barDtrainII}\Big[ \bar{\U}_y(\lambda_y)\Big] + \sqrt{\E_{\barallDII}\Big[ \norm{\widehat{\gamma} - \bar{\gamma} }{2}^2 \Big]}    } \Bigg) \\
	&\qquad + O_{\P}\Bigg(\sqrt{\tau_n(1-\tau_n)n} e^{-C (\min_y\lambda_y)^2 \bar{n}^\train}  + \sqrt{\tau_n(1-\tau_n)n} \P_{\barallDII}\Big[\bignorm{ \widehat{\gamma} - \bar{\gamma} }{2} \geq C \Big]  \Bigg),
\end{align*}

where  $\barDtrainII$, $\barallDII$, $\bar{\gamma}$ and $\bar{\U}_y(\lambda_y)$ are as defined in subsection \eqref{subsection: Assumptions for score function} and correspond to some sequence $(\bar{n}^\test, \bar{n}_0^\train,\dots, \bar{n}_m^\train)$ that satisfies $\absbig{\bar{n}^\test / n - \tau_n}, \absbig{\bar{n}_y^\train/ n^\train - (1-\tau_n)\pi_y^\train} \to 0$ where $\bar{n}^\train := n - \bar{n}^\test$.
\end{lemma}
\vspace{0.2in}

The proof for Lemma \eqref{Behavior of First and Second Order Error Terms in tau n IID regime} is in the Appendix. Lemma \eqref{Behavior of First and Second Order Error Terms in tau n IID regime} establishes several facts about SELSE. First, it shows that SELSE is asymptotically linear in the efficient influence function under the $\tau_n$-IID Regime. Since the Fixed $\tau$-IID Regime is a special case of the $\tau_n$-IID Regime, this means that SELSE satisfies property \eqref{Definition of E estimator: it should be tau RAL for each tau}, modulo the regularity condition, which we prove SELSE satisfies in Lemma \eqref{pi hat is RAL} in the Appendix. Second, Lemma \eqref{Behavior of First and Second Order Error Terms in tau n IID regime} establishes that the second order $\delta_{\tau_n}$ satisfies property \eqref{Definition of E estimator: second order error to zero with changing tau n} and that its efficient influence function $\psi_{\tau_n}^\eff$ satisfies property \eqref{Definition of E estimator: moment condition on influence function with changing tau n}. Thus, overall, Lemma \eqref{Behavior of First and Second Order Error Terms in tau n IID regime} verifies all the conditions needed to show that SELSE is a member of $\curlyE$. The fact that the efficient influence function appears again in SELSE's first order error but now in the context of the $\tau_n-IID$ Regime is also exciting, as it means that we have only a single step remains for establishing SELSE's semiparametric efficiency: we must now translate the semiparametric efficiency bound from the Fixed $\tau$-IID Regime to the $\tau_n$-IID Regime. The culmination of that endeavor, along with all of our results so far, is presented in Theorem \eqref{Semiparametric Efficiency in tau n IID Regime} in the next subsection.

\subsection{Main Result} \label{sec: main result}

We now present our main theoretical result on SELSE's performance and state the semiparametric efficiency bound for quantification tasks, both in the context of the $\tau_n$-IID Regime. \\



\begin{theorem}[\uline{Semiparametric Efficiency in $\tau_n$-IID Regime}] \label{Semiparametric Efficiency in tau n IID Regime}

Let $\tau_n$ denote any sequence satisfying Assumptions \eqref{tau n is inside open interval 0 and 1} and \eqref{tau n convergence not too fast}, and let $\widetilde{\pi}_n $ denote any quantifier and $\psi_{\tau_n}$ its influence function. Under the Assumptions in subsections \eqref{Assumptions on class proportions}, \eqref{subsection: Assumptions for p}, \eqref{subsection: Assumptions for score function} and \eqref{subsection: Assumptions about gamma hat}, if $\widetilde{\pi}_n \in \curlyE$, then there exists a corresponding random vector $G_n$ such that $G_n \goesto{d} \N(0,I_{m\times m})$ and
\[
	\sqrt{\tau_n(1-\tau_n)n}(\widetilde{\pi}_n - \pi^\ast) = \Big( \tau_n(1-\tau_n) \Var_{\Jcal^{\pi^\ast,\mathbf{p},\tau_n}}[ \psi_{\tau_n} ]  \Big)^{\frac{1}{2}} G_n  + o_{\P}(1),
\]
where $\Var_{\Jcal^{\pi^\ast,\mathbf{p},\tau_n}}[ \psi_{\tau_n} ] \succeq \mathbf{V}^\eff(\tau_n)$ for each $n$,  and the matrix $\mathbf{V}^\eff(\tau_n)$ is given by
\[
	\mathbf{V}^\textup{eff}(\tau_n) := \bigg[ \frac{1}{\tau_n} + \frac{1}{1-\tau_n} \sum_{k=0}^{m} \frac{(\pi_k^\ast)^2}{\pi_k^\train} \bigg]\Big(\FisherInfo{ \gamma^\ast(\tau_n) }^{-1} - \FisherInfo{\gamma^\ast(\tau_n);\textup{Cat}}^{-1} \Big) + \frac{1}{\tau_n}\FisherInfo{\pi^\ast;\textup{Cat}}^{-1},
\]
where  $\gamma^\ast(\tau_n) := (\gamma_1^\ast(\tau_n), \dots, \gamma_m^\ast(\tau_n))$ and
\[
	\gamma_y^\ast(\tau_n) := \frac{ \frac{\pi_y^\ast}{\tau_n}  +   \frac{1}{1-\tau_n} \frac{(\pi_y^\ast)^2}{\pi_y^\train} }{\frac{1}{\tau_n}+ \frac{1}{1-\tau_n} \sum_{k=0}^{m}\frac{(\pi_k^\ast)^2}{\pi_k^\train} }\qquad \forall\ y\in\Y.
\]
Furthermore, our SELSE quantifier $\widehat{\pi}_n$ has an influence function $\psi_{\tau_n}^\eff$ which satisfies 
\[
	\Var_{\Jcal^{\pi^\ast,\mathbf{p},\tau_n}}[ \psi_{\tau_n}^\eff ] = \mathbf{V}^\eff(\tau_n)
\]
for each $n$, and $\widehat{\pi}_n\in\curlyE$.

\end{theorem}
\vspace{0.2in}

In a nutshell, Theorem \eqref{Semiparametric Efficiency in tau n IID Regime} states  that SELSE is a member of $\curlyE$, and that among all members of $\curlyE$, SELSE has the smallest possible asymptotic variance matrix for its normalized error.  This variance matrix is equal to $\tau_n(1-\tau_n)\mathbf{V}^\text{eff}(\tau_n)$, and it is the acclaimed semiparametric efficiency bound for the $\tau_n$-IID Regime. The main logic underlying this result is that the asymptotic variance of a quantifier's error is simply the variance (scaled by $\tau_n(1-\tau_n)$) of its own influence function, and the variance of our quantifier's influence function can be shown to be always less than or equal to the variance of any other member's. This result is largely due to SELSE's usage of $\widehat{s}_{\widehat{\gamma}}$ as a proxy for $s_{\gamma^\ast}$ in its ACC/PACC approach: as indicated by the dependence of $\mathbf{V}^\text{eff}(\tau_n)$ on $\tau_n$, the "optimal" performance for estimating $\pi^\ast$ varies greatly with $\tau_n$, and the dependence of $\gamma^\ast$ on the number of train and test samples allows our quantifier to account for this fact when estimating $\pi^\ast$.



The fact that SELSE has the smallest possible eigenvalues for its asymptotic variance matrix is attractive from two different points of view. First, it follows that our quantifier is optimal for learning $\pi^\ast$ under $L_2$ loss: for any other quantifier $\widetilde{\pi} \in \curlyE$, the asymptotic variance of $\sqrt{\tau_n(1-\tau_n)n} \norm{\widehat{\pi}_n - \pi^\ast}{2}$ will always be less than or equal to that of $\sqrt{\tau_n(1-\tau_n)n} \norm{\widetilde{\pi}_n - \pi^\ast}{2}$. Second, our quantifier is optimal for estimating smooth functions of $\pi^\ast$. That is, for any integer $l \geq 1$ and smooth function $\ell: \mathbb{R}^m \mapsto \mathbb{R}^l$, the asymptotic variance matrix of $\sqrt{\tau_n(1-\tau_n)n}(\ell(\widehat{\pi}_n) - \ell(\pi^\ast) )$ will always be less than or equal to that of $\sqrt{\tau_n(1-\tau_n)n}(\ell(\widetilde{\pi}_n) - \ell(\pi^\ast) )$. This can be easily deduced via a first order Taylor expansion of both normalized errors. From a practical point of view, this means that our quantifier is optimal for learning at least three different, interesting scalar quantities, each of which correspond to setting $l = 1$ and $\ell(\beta) = c'\beta + d$ for an appropriate choice of $c\in\mathbb{R}^m$ and $d\in\mathbb{R}$: the probability of $Y^\test$ belonging to a specific class (or group of classes),  the difference in probability of belonging to one class (or group) versus another class (or group), and the expected value of a function of $Y^\test$.

Lastly, in addition to verifying the optimality of our quantifier, the semiparametric efficiency bound also provides us a deeper understanding of the general problem of quantification. In particular, it reveals that quantification problems are actually a "blend" of two other statistical estimation tasks:

\vspace{0.2in}
\begin{enumerate}[label=(\alph*), ref = \alph*]
	\item estimating $\pi^\ast$ when $\mathbf{p}$ is known but $p_{\pi^\ast}$ is unknown, and we have IID samples from $p_{\pi^\ast}$, and  \label{case: lots of train samples}
	\item estimating $\pi^\ast$ when $p_{\pi^\ast}$ is known but $\mathbf{p}$ is unknown, and we have IID samples from each density in $\mathbf{p}$.  \label{case: lots of test samples}
\end{enumerate}
\vspace{0.2in}

The degree to which quantification is similar to problem \eqref{case: lots of train samples} versus \eqref{case: lots of test samples} depends on $\tau_n$. If $\tau_n \to 0$, then quantification becomes identical to problem \eqref{case: lots of train samples}, and if $\tau_n \to 1$, then quantification becomes identical to problem \eqref{case: lots of test samples}. To see why, start by considering the former situation when $\tau_n \to 0$. When this happens, there are far more train samples than test samples, and so for large $n$, we essentially have perfect knowledge of $\mathbf{p}$ and must estimate $\pi^\ast$ using samples from the unknown $p_{\pi^\ast}$. This is exactly the definition of problem \eqref{case: lots of train samples}. Furthermore,  the equivalence between quantification and problem (a) when $\tau_n\to 0$ also agrees with the limiting behavior of the semiparametric efficiency bound. Indeed, since $\gamma^\ast(0) = \pi^\ast$, we have that
\begin{align*}
	\tau_n(1-\tau_n)\mathbf{V}^\textup{eff}(\tau_n) &= \bigg[ (1-\tau_n) + \tau_n \sum_{k=0}^{m} \frac{(\pi_k^\ast)^2}{\pi_k^\train} \bigg]\Big(\FisherInfo{ \gamma^\ast(\tau_n) }^{-1} - \FisherInfo{\gamma^\ast(\tau_n);\textup{Cat}}^{-1} \Big) \\
	&\qquad+ (1-\tau_n)\FisherInfo{\pi^\ast;\textup{Cat}}^{-1} \\ 
	&\goesto{\tau_n\to0} \Big(\FisherInfo{ \pi^\ast }^{-1} - \FisherInfo{ \pi^\ast ;\textup{Cat}}^{-1} \Big) + \FisherInfo{ \pi^\ast ;\textup{Cat}}^{-1}\\
	& = \FisherInfo{ \pi^\ast }^{-1}.
\end{align*}
$\FisherInfo{\pi^\ast}^{-1}$ is the CRLB for estimating $\pi^\ast$ in problem \eqref{case: lots of train samples}, and so the fact that $\tau_n(1-\tau_n)\mathbf{V}^\textup{eff}(\tau_n) \to \FisherInfo{\pi^\ast}^{-1}$ as $\tau_n\to 0$ means that the difficulty of learning $\pi^\ast$ in quantification becomes equal to the difficulty in problem \eqref{case: lots of train samples}, further verifying the equivalence between the two problems.

The opposite happens when $\tau_n \to 1$: in this situation, there are far more test samples than train samples, and so for large $n$, we essentially have perfect knowledge of $p_{\pi^\ast}$ and must estimate $\pi^\ast$ using the samples obtained from each density in $\mathbf{p}$. This is exactly the definition of problem \eqref{case: lots of test samples}. Now, similar to before, one can also show that the semiparametric efficiency bound for quantification converges to the bound for problem \eqref{case: lots of test samples} when $\tau_n \to 1$, but proving the bound for problem \eqref{case: lots of test samples} is tedious since the problem is still semiparametric. Therefore, as a compromise, we instead point out the equivalence for a special \textit{parametric} case of problem \eqref{case: lots of test samples}: namely, when $m=1$ and both $p_{\pi^\ast}$ and $p_0$ are known but $p_1$ is unknown, and we obtain IID samples from $p_{1}$. This special case of problem \eqref{case: lots of test samples} corresponds to a quantification task where $\tau_n \to 1$ and $\pi^\train \to 0$, because when those limits hold, we will have very few train samples from class $1$, but an abundance of train samples from class $0$, as well as many test samples. As such, when $n$ is large, we will have near perfect knowledge of $p_{\pi^\ast}$ and $p_0$, and must estimate $\pi^\ast$ using the samples from $p_1$. This equivalence is corroborated by the limiting behavior of the semiparametric efficiency bound, since
\begin{align*}
	\pi^\train(1-\pi^\train) \tau_n(1-\tau_n)\mathbf{V}^\eff(\tau_n) & \goesto{ (\tau_n,\pi^\train) \to (1,0) } (\pi^\ast)^2 \FisherInfo{1}^{-1},
\end{align*}
where $(\pi^\ast)^2 \FisherInfo{1}^{-1}$ is the CRLB for estimating $\pi^\ast$ in the special parametric version of problem $(b)$. \\


\subsection{Proof of Main Result} \label{sec: Proof of Main Theorem}

\begin{proof}[\uline{Proof of Theorem \eqref{Semiparametric Efficiency in tau n IID Regime}.}]
First, we verify that $\widehat{\pi}_n \in  \curlyE$. Note that, under Assumptions  \eqref{Fixed Tau Regime: Mixture FIM not too small}, \eqref{Fixed Tau Regime: Cat minus Mixture FIM}, \eqref{Fixed Tau Regime: expo family},  \eqref{Fixed Tau Regime: H closed under scalar multiplication}, \eqref{Fixed Tau Regime: Omega Properties}, \eqref{Tau n Regime: lambda and uniform convergence properties} and \eqref{tau n regime, estimator of gamma bar}, we have by Lemma \eqref{pi hat is RAL} that, for each fixed $\tau\in(0,1)$, the estimator $\widehat{\pi}_n$ is $\tau$-RAL with influence function $\psi_\tau^\eff$ and second order error $\delta_\tau$. Thus, property  \eqref{Definition of E estimator: it should be tau RAL for each tau} holds for $\widehat{\pi}_n$. Furthermore, under Assumptions  \eqref{Fixed Tau Regime: Mixture FIM not too small}, \eqref{Fixed Tau Regime: Cat minus Mixture FIM}, \eqref{Fixed Tau Regime: H closed under scalar multiplication}, \eqref{Fixed Tau Regime: Omega Properties}, \eqref{Tau n Regime: lambda and uniform convergence properties} and \eqref{tau n regime, estimator of gamma bar}, for any sequence $\tau_n$ that satisfies Assumptions \eqref{tau n is inside open interval 0 and 1} and \eqref{tau n convergence not too fast}, it follows from Lemma \eqref{Behavior of First and Second Order Error Terms in tau n IID regime} that properties \eqref{Definition of E estimator: second order error to zero with changing tau n} and \eqref{Definition of E estimator: moment condition on influence function with changing tau n} also hold. Thus, $\widehat{\pi}_n \in  \curlyE$. 

Second, we verify the theorem's asymptotic statements concerning an arbitrary quantifier $\widetilde{\pi}_n \in \curlyE$. Since $\widetilde{\pi}_n$ is $\tau$-RAL for all $\tau \in (0,1)$ by virtue of property  \eqref{Definition of E estimator: it should be tau RAL for each tau}, it follows that for any $n$, the equation
\[
	 \sqrt{\tau_n(1-\tau_n)n}(\widetilde{\pi}_n - \pi^\ast) = \sqrt{\frac{\tau_n(1-\tau_n)}{n}} \sum_{i=1}^n \psi_{\tau_n}(Z_i) + \sqrt{\tau_n(1-\tau_n)} \delta_{\tau_n}(Z_{1:n})
\]
holds for every dataset $Z_{1:n}$. Further, due to Assumptions \eqref{tau n is inside open interval 0 and 1} and \eqref{tau n convergence not too fast}, we have by property \eqref{Definition of E estimator: second order error to zero with changing tau n} that $\delta_{\tau_n}(Z_{1:n}) \goesto{p} 0$. So, because $\sqrt{\tau_n(1-\tau_n)}$ is bounded, it follows that
\[
	\implies  \sqrt{\tau_n(1-\tau_n)n}(\widetilde{\pi}_n - \pi^\ast) = \sqrt{\frac{\tau_n(1-\tau_n)}{n}} \sum_{i=1}^n \psi_{\tau_n}(Z_i) + o_{\P}(1).
\]

Additionally, since $\widetilde{\pi}_n$ is $\tau$-RAL for all $\tau\in(0,1)$, the matrix  $\Var_{\Jcal^{\pi^\ast,\mathbf{p},\tau_n}}\big[ \psi_{\tau_n}(Z) \big]$ is nonsingular for all $n$. Thus, the first order term above may be written as:
\begin{align*}
	&\sqrt{\frac{\tau_n(1-\tau_n)}{n}} \sum_{i=1}^n \psi_{\tau_n}(Z_i)  \\
	&= \Big(\tau_n(1-\tau_n)  \Var_{\Jcal^{\pi^\ast,\mathbf{p},\tau_n} }\big[\psi_{\tau_n}(Z)\big] \Big)^{\tfrac{1}{2}}    \underbrace{ \sum_{i=1}^n \overbrace{ n^{-\frac{1}{2}} \Var_{\Jcal^{\pi^\ast,\mathbf{p},\tau_n} }\big[\psi_{\tau_n}(Z)\big]^{-\tfrac{1}{2}} \psi_{\tau_n}(Z_i)}^{=:W_{n,i}} }_{=: G_n}.
\end{align*}

We will now prove that $G_n \equiv \sum_{i=1}^n W_{n,i} \goesto{p} \N(0,I_m)$. Towards that end, observe that $\E_{\Jcal^{\pi^\ast,\mathbf{p},\tau_n} }[W_{n,i}] = 0$ and $\sum_{i=1}^n \E_{\Jcal^{\pi^\ast,\mathbf{p},\tau_n} }\big[ W_{n,i}W_{n,i}'\big ] = I_{m}$. Also, for any $c > 0$, we have that:
\small
\begin{align*}
	&\sum_{i=1}^n  \E_{\Jcal^{\pi^\ast,\mathbf{p},\tau_n} }\Big[ \norm{W_{n,i}}{2}^2 \Ibig{\norm{W_{n,i}}{2} > c} \Big] \\
	&=  \sum_{i=1}^n  \E_{\Jcal^{\pi^\ast,\mathbf{p},\tau_n} }\Big[ \norm{W_{n,i}}{2}^2 \Ibig{\norm{W_{n,i}}{2}^2 > c^2} \Big]\\
	&= \frac{1}{n}\sum_{i=1}^n  \E_{\Jcal^{\pi^\ast,\mathbf{p},\tau_n} }\Big[  \psi_{\tau_n}(Z_i)' \Var_{\Jcal^{\pi^\ast,\mathbf{p},\tau_n} }\big[\psi_{\tau_n}(Z)\big]^{-1} \psi_{\tau_n}(Z_i)  \\
	&\qquad\qquad\qquad\qquad\qquad\times\Ibig{  \psi_{\tau_n}(Z_i)' \Var_{\Jcal^{\pi^\ast,\mathbf{p},\tau_n} }\big[\psi_{\tau_n}(Z)\big]^{-1} \psi_{\tau_n}(Z_i) > nc^2} \Big] \\
	&= \E_{\Jcal^{\pi^\ast,\mathbf{p},\tau_n} }\Big[  \psi_{\tau_n}(Z)' \Var_{\Jcal^{\pi^\ast,\mathbf{p},\tau_n} }\big[\psi_{\tau_n}(Z)\big]^{-1} \psi_{\tau_n}(Z)  \Ibig{  \psi_{\tau_n}(Z)' \Var_{\Jcal^{\pi^\ast,\mathbf{p},\tau_n} }\big[\psi_{\tau_n}(Z)\big]^{-1} \psi_{\tau_n}(Z) > nc^2} \Big].
\end{align*}
\normalsize
Also note that:
\begin{align}
	 \psi_{\tau_n}(Z)' \Var_{\Jcal^{\pi^\ast,\mathbf{p},\tau_n} }\big[\psi_{\tau_n}(Z)\big]^{-1} \psi_{\tau_n}(Z) &\leq \norm{\psi_{\tau_n}(Z)}{2}^2 \maxEval{ \Var_{\Jcal^{\pi^\ast,\mathbf{p},\tau_n} }\big[\psi_{\tau_n}(Z)\big]^{-1}  } \nonumber \\ 
	 &= \frac{\norm{\psi_{\tau_n}(Z)}{2}^2}{\minEval{ \Var_{\Jcal^{\pi^\ast,\mathbf{p},\tau_n} }\big[\psi_{\tau_n}(Z)\big] } } \nonumber \\
	 &\leq \frac{\norm{\psi_{\tau_n}(Z)}{2}^2}{\minEval{ \mathbf{V}^\eff(\tau_n) }} \nonumber \\
	 &\leq \frac{\tau_n(1-\tau_n)\norm{\psi_{\tau_n}(Z)}{2}^2}{   (\xi)^2 \nu   \big(\frac{L}{m+1}\big)^2   }. \nonumber
\end{align}

The third line is due to the fact that $\widetilde{\pi}_n$ is $\tau$-RAL for each fixed $\tau \in (0,1)$, which under Assumptions \eqref{Fixed Tau Regime: Mixture FIM not too small}, \eqref{Fixed Tau Regime: Cat minus Mixture FIM} and \eqref{Fixed Tau Regime: expo family}, implies that $\Var_{\Jcal^{\pi^\ast,\mathbf{p},\tau_n} }\big[  \psi_{\tau_n}^\eff \big] = \mathbf{V}^\eff(\tau_n) \preceq \Var_{\Jcal^{\pi^\ast,\mathbf{p},\tau_n} }\big[\psi_{\tau_n}(Z)\big]$ for each $n$ by virtue of Theorem \eqref{Efficiency Bound}.  The fourth line is due to Lemma \eqref{Projection of Ordinary Score onto G} and Assumption \eqref{Fixed Tau Regime: Mixture FIM not too small}. For notational convenience, temporarily define $b:=  (\xi)^2 \nu   \big(\frac{L}{m+1}\big)^2$. Then, it follows that:
\small
\begin{align*}
	&\sum_{i=1}^n  \E_{\Jcal^{\pi^\ast,\mathbf{p},\tau_n} }\Big[ \norm{W_{n,i}}{2}^2 \Ibig{\norm{W_{n,i}}{2} > c} \Big]  \\
	&\leq \frac{\tau_n(1-\tau_n)}{b} \E_{\Jcal^{\pi^\ast,\mathbf{p},\tau_n} }\Big[   \norm{\psi_{\tau_n}(Z)}{2}^2   \Ibig{  \norm{\psi_{\tau_n}(Z)}{2}^2> nc^2  \tfrac{b}{\tau_n(1-\tau_n)} } \Big] \\
	 &= \frac{\tau_n(1-\tau_n)}{b} \E_{\Jcal^{\pi^\ast,\mathbf{p},\tau_n} }\Bigg[   \frac{\norm{\psi_{\tau_n}(Z)}{2}^4}{\norm{\psi_{\tau_n}(Z)}{2}^2}   \Ibig{  \norm{\psi_{\tau_n}(Z)}{2}^2> nc^2  \tfrac{b}{\tau_n(1-\tau_n)} } \Bigg] \\ 
	 &\leq \frac{\tau_n^2(1-\tau_n)^2}{nc^2 b^2}  \E_{\Jcal^{\pi^\ast,\mathbf{p},\tau_n} } \Big[  \norm{\psi_{\tau_n}(Z)}{2}^4  \Big] \\ 
	 &= \frac{\tau_n^2(1-\tau_n)^2}{nc^2 b^2}  \Bigg\{ \tau_n\E_{\pi^\ast} \Big[  \norm{\psi_{\tau_n}(X,-1,1)}{2}^4  \Big] + (1-\tau_n) \sum_{y=0}^m \pi_y^\train \E_y \Big[  \norm{\psi_{\tau_n}(X,y,0)}{2}^4  \Big] \Bigg\} \\ 
	 &=  \frac{1}{c^2 b^2}  \Bigg\{ \frac{\tau_n^3(1-\tau_n)^2}{n}\E_{\pi^\ast} \Big[  \norm{\psi_{\tau_n}(X,-1,1)}{2}^4  \Big] + \frac{\tau_n^2(1-\tau_n)^3}{n} \sum_{y=0}^m \pi_y^\train \E_y \Big[  \norm{\psi_{\tau_n}(X,y,0)}{2}^4  \Big] \Bigg\} \\ 
	 &=  \frac{1}{c^2 b^2}  \Bigg\{ \frac{(1-\tau_n)^2}{\tau_n n}\E_{\pi^\ast} \Big[  \norm{\tau_n \psi_{\tau_n}(X,-1,1)}{2}^4  \Big] + \frac{\tau_n^2}{(1-\tau_n)n} \sum_{y=0}^m \pi_y^\train \E_y \Big[  \norm{(1-\tau_n) \psi_{\tau_n}(X,y,0)}{2}^4  \Big] \Bigg\},
\end{align*}
\normalsize
where the last line goes to zero as $n\to\infty$ by virtue of property \eqref{Definition of E estimator: moment condition on influence function with changing tau n}.  Therefore, by Theorem 11.1.6 in \cite{Athreya2006}, we may conclude that $G_n \goesto{d} \N(0,I_m)$.
\end{proof}
\vspace{0.4in}

 \section{Numerical Experiments} \label{sec: Numerical Experiments Section}

We now study SELSE's empirical performance. This is an important supplement to our theoretical analysis, because while we have established the optimality of  SELSE's first order error term, we have not discussed the optimality of its higher order ones, and in practice, a quantifier's total error always depends on both. The experiments in this section therefore aim to provide a more complete, practical picture of SELSE and its competitors.




\subsection{Experimental Set-Up}

We performed experiments on synthetic and real world data. The synthetic data involved two classes, with the covariates from each class sampled from different 2D gaussian distributions, each with identity covariance. For class $0$, the associated mean vector was $(-1,-1)^T$, and for class $1$, the associated mean vector was $(1,1)^T$. In terms of real world data, we considered a blood transfusion dataset. This dataset was used as a benchmark in \cite{moreo2021quapy}'s  systematic comparison of different quantification methods, which includes all the quantifiers considered here except for SELSE. 

The experimental set-up for the synthetic and real world datasets were similar. For a given combination of $n, \tau, \pi^\train$ and $\pi^\ast$, we generated $n^\test = \tau n$ test samples and $n^\train = (1-\tau)n$ train samples. For the real world data, the train data for each class $y$ was generated by sampling with replacement from the points in the original dataset that belonged to class $y$. The test data was generated in a similar fashion, except for each test sample, the class that we took the sample from was  determined randomly according to $\text{Cat}(\pi^\ast)$. Then, once the train and test data were formed, we computed six different quantifiers\footnote{The performance of SELSE and MLLS was measured in the R programming language via our own in-house implementations of those quantifiers, whereas the measurements for ACC, PACC, HDy and EMQ were done via their Python implementations in the QuaPy package \citep{moreo2021quapy}.}: ACC, PACC, HDy, EMQ, MLLS and SELSE, all of which were described in the Methodology section. For each quantifier $\widetilde{\pi}$, the normalized error $\sqrt{\tau_n(1-\tau_n)n} \norm{\widetilde{\pi} - \pi^\ast}{2}$ was recorded. This experiment was repeated 100 times, and for each quantifier, the average of its normalized error across those experiments was calculated. 

This average served as our main metric for the performance of a quantifier under a given combination of  $n, \tau, \pi^\train$ and $\pi^\ast$. The combinations we considered were as follows. For both datasets, we set $n = 500$ and $\pi^\train = \frac{1}{2}$, and computed our metric over all combinations of $\tau \in \{0.2, 0.5, 0.8\}$ and $\pi^\ast \in \{0.05, 0.2, 0.35, 0.5, 0.65, 0.80, 0.95\}$. We froze $\pi^\train$ at $\frac{1}{2}$ to limit the amount of simulations that needed to be done.  

\subsection{Quantifier Configuration Details}
The configuration details of the six quantifiers are as follows. Recall that all the quantifiers in question make use of some form of classifier. For ACC, this is a hard classifier, and for PACC, HDy, EMQ and MLLS, this is a soft classifier. SELSE also uses a soft classifier indirectly, since in Step 2, a preliminary quantifier for $\pi^\ast$ is utilized to form the plug-in estimator $\widehat{\gamma}$, and in our experiments, MLLS served as that preliminary quantifier. For all six quantifiers and both datasets, the classifier in question was a SVM with either a radial or polynomial kernel and $\ell_2$ penalty (the exact kernel and penalization parameters were selected via 3-fold cross validation) Finally, SELSE has several other unique parameters in Step 3 that relate to how the score function is estimated, which need to be chosen. Specifically, in our experiments, we chose each space $\H_y$ to be a RKHS, with a gaussian reproducing kernel. The kernel width and penalty parameters  used to construct $f = \widehat{s}_{\widehat{\gamma}}$ were chosen to minimize the maximum eigenvalue of the empirical counterpart to $A_f^{-1} \Var_{\widehat{\gamma}}[f]A_f^{-T}$, which was computed on a separate validation set.  


\begin{figure*}[hbt!] \label{gmm experiment results} 
\begin{tabular}{c}
\subfloat[$\tau = 0.2$]{\includegraphics[width = .5\textwidth,height = 200pt]{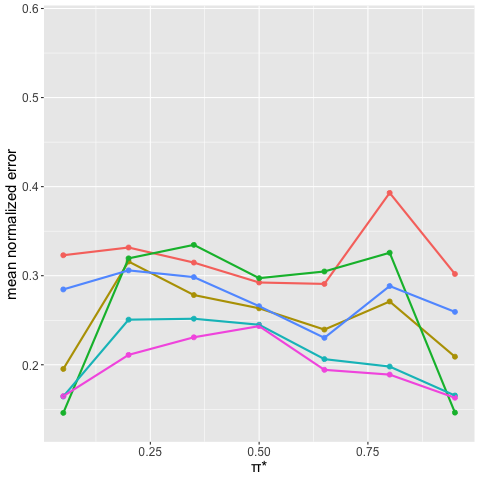}} 
\subfloat[$\tau = 0.8$]{\includegraphics[width = .5\textwidth,height = 200pt]{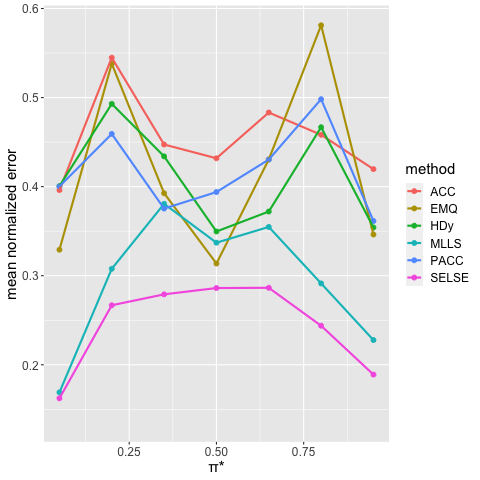}} \\
\subfloat[$\tau = 0.5$]{\includegraphics[width = .5\textwidth,height = 200pt]{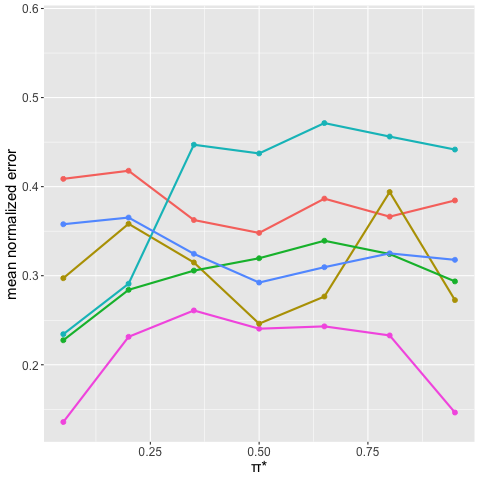}}
\subfloat[SELSE: True Score Correlation]{\includegraphics[width = .5\textwidth,height = 200pt]{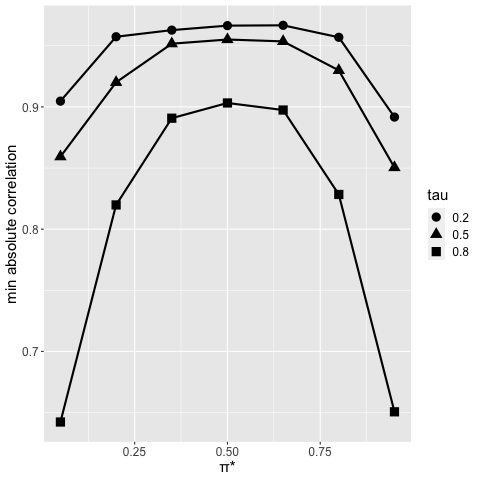}}\label{gmm correlation plot} \\
\end{tabular}
\caption{Simulation Results on Synthetic Gaussian Mixture Data.  }
\end{figure*} 

\begin{figure*}[hbt!]
\begin{tabular}{ccc}
\subfloat[$\tau = 0.2$]{\includegraphics[width = .35\textwidth,height = 180pt]{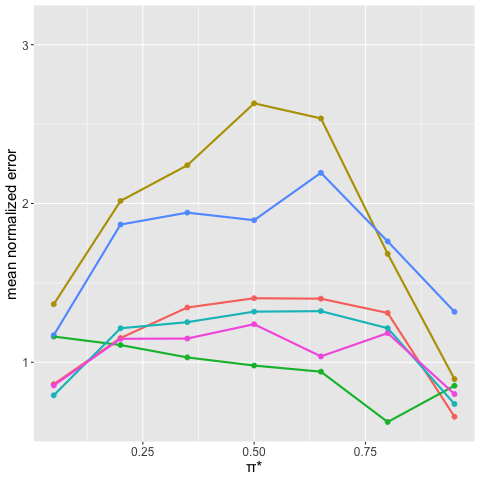}} %
\subfloat[$\tau = 0.5$]{\includegraphics[width = .35\textwidth,height = 180pt]{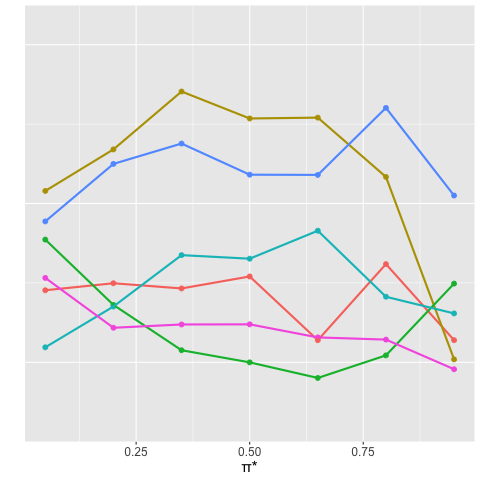}}%
\subfloat[$\tau = 0.8$]{\includegraphics[width = .35\textwidth,height = 180pt]{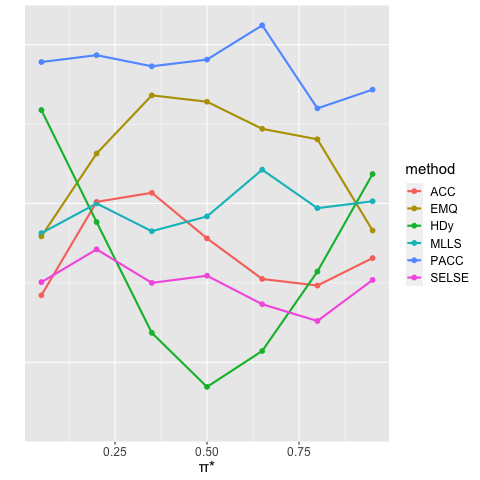}} \\
\end{tabular}
\caption{Simulation Results on Transfusion Data.}
\end{figure*}

\subsection{Results \& Discussion}

We now turn to the results of our experiments. We begin by making two general observations. First, in both experiments, all quantifiers tended to preform worse for larger values of $\tau$ than for smaller values of $\tau$. This means that, at least in these two types of data, a scarcity of training samples is harder to handle than a scarcity of test samples.  Second, when $\tau$ is small, estimating $\pi^\ast$ when $\pi^\ast$ is close to $0$ or $1$ is far easier than when $\pi^\ast$ is close to $\frac{1}{2}$. This makes sense in light of our discussion at the end of the Theoretical Results section, and holds for all datasets (not just the ones considered here). This is because when $\tau\to 0$, the semiparametric efficiency bound approaches the CRLB for the classic parametric mixture problem. The parametric problem is easier when $\pi^\ast$ takes on extreme values, much in the same way that estimating $p$ using independent flips of a $p$-biased coin is easier when $p \approx 0$ or $p\approx1$ than when $p \approx \frac{1}{2}$. However, this pattern tends to weaken as $\tau$ increases. 

Next, we comment on the performance of SELSE specifically. In our synthetic data experiment, SELSE outperforms the other quantifiers for nearly all combinations of $\tau$ and $\pi^\ast$. This makes sense in light of Figure (1d), which shows that SELSE's score function estimates tend to be highly correlated with $s_{\gamma^\ast}$. This correlation is desirable when one estimates $\pi^\ast$ because, as discussed in subsections \eqref{sec: Ideal Characteristics of f} and \eqref{sec: score function maximizes ideal characteristics}, using $s_{\gamma^\ast}$ allows one to strike the ideal balance between one's uncertainty about the mixture distribution (caused by the finite amount of test data) and the component densities (caused by the finite amount of train data). Correspondingly, the fact that the performance gap between SELSE and the other quantifiers grows as $\tau\to 1$ suggests that the other quantifiers are unable to effectively orchestrate this balance. This especially makes sense for EMQ and MLLS, since both of those methods are based on the maximum likelihood principle. As mentioned in subsection \eqref{sec: Maximum Likelihood Approaches: EMQ and MLLS}, maximum likelihood assumes that either the $Y\mid X$ or $Y \mid f(X)$ conditional can be learned, which is a difficult requirement to satisfy when the training set is extremely small. Finally, the same observations made about SELSE for our synthetic data experiment also hold true for the transfusion data experiment. The only exception to this is HDy, which tends to outperform SELSE when $\tau$ is small. However, the performance of HDy also degrades for extreme values of $\pi^\ast$ as $\tau\to1$, making HDy unreliable in those situations. In contrast, SELSE is able to avoid this pitfall: with the exception of HDy, SELSE's error is almost uniformly smaller than the error of any other quantifier, regardless of the value of $\tau$, making it a favorable, general-purpose quantifier to turn to when one does not want to worry about the effect of $\tau$.

{\centering \section{Conclusion}} \label{sec: Conclusion}

In summary, we proposed the SELSE quantifier as a method for inferring the response distribution in the test set under the label shift assumption. We proved that SELSE is semiparametric efficient for a broad class of quantifiers, implying that SELSE's first order error term is optimally small.  We also grounded our theoretical results of SELSE's optimality in two empirical studies, and numerically showed the distinct advantages of SELSE over many of the existing approaches to quantification.


\newpage
{\centering \section{Appendix}} \label{sec: Appendix}

\subsection{Three Regimes}
Here we provide a more in depth description of the three regimes initially presented in subsection \eqref{sec: derivation roadmap}, as this will aid us in proving the supporting lemmas and theorems in the Appendix. Recall that the regimes differ from each other in two key regards: the data generation mechanism, and how the true parameters behave as $n\to \infty$.\\

\subsubsection{Fixed Sequence Regime}

In the Fixed Sequence Regime, there exists a fixed sequence 
\[
	\big\{(n^\test, n^\train, \pi^\ast, \pi^\train, p_{0},p_{1},\dots,p_{m}) \big\}_{n=1}^\infty
\]
where the sample sizes $0\leq n^\test, n^\train \to \infty$ satisfy $n^\test + n^\train = n$, and the class probabilities $\pi^\ast,\pi^\train$ and class densities $p_{0},\dots,p_{m}$ are subject to various restrictions that shall be described later. The class densities are supported on some set $\X$, and are dominated w.r.t some measure $\mu$. The class probabilities and class densities may all depend on $n$.  We assume $\pi^\train$ is known, and that $\pi^\ast$ and $p_{0},\dots,p_{n}$ are unknown.

In terms of data, for each $n$, we obtain $n^\test$ test samples from the mixture $\sum_{y=0}^m \pi_{y}^\ast p_{y}$, as well as $\pi_{y}^\train n^\train$ train samples from $p_{y}$ for each class $y\in\Y:=\{0,1,\dots, m\}$. The $n$ samples are mutually independent. For convenience, we let $\Dtest$ and $\Dtrain$ denote the collection of test and train samples, respectively. Our goal is to use $\Dtest$ and $\Dtrain$ to create an estimator $\widehat{\pi}$ that ``reacts''  to the evolution in the sequence of parameters/sample sizes as $n\to \infty$ so as to ensure $\widehat{\pi} - \pi^\ast$ becomes small. \\

\subsubsection{Fixed $\tau$-IID Regime}

Let any $\tau \in (0,1)$ be given.  In the Fixed $\tau$-IID Regime, we consider an explicit semiparameteric model, $\textbf{M}^{\text{semi},\tau}$, with parameter space $\Theta^\text{semi}$. We define $\Theta^\text{semi}$ in the following manner. Let $\Delta$ denote a subset of the $m+1$ dimensional probability simplex. Let $\Q_0,\Q_1,\dots,\Q_m$ denote sets of densities, and define $\Q := \Q_0\times\Q_1\times\dots\times\Q_m$. The parameter space for the parameter of interest is $\Delta$, and the nuisance parameter space is $\Q$; both $\Delta$ and $\Q$ are subject to various restrictions that shall be described later. Then, $\Theta^\text{semi} := \Delta \times \Q$.

The semiparametric model is then given by $\textbf{M}^{\text{semi},\tau} := \big\{ \Jcal^{\pi,\mathbf{q},\tau} \ \big\lvert \ (\pi,\mathbf{q}) \in \Theta^\text{semi} \big\}$, where for each $(\pi,\mathbf{q}) \in \Theta^\text{semi}$, $\Jcal^{\pi,\mathbf{q},\tau}$ is the density function for the joint distribution of a random vector $Z \equiv (X,Y,D)$ that is generated in the following manner:
\begin{itemize}
	\item $D \sim \text{Bern}(\tau)$
	\item If $D = 1$: $Y = -1$, $X \sim q_{\pi}$
	\item If $D = 0$:  $Y \sim \text{Cat}(\pi^\train)$, $X\mid Y = y \sim q_{y}$,
\end{itemize}
where $\pi^\train \in \Delta$ is some fixed vector. One can show that $\Jcal^{\pi,\mathbf{q},\tau}$ has the following explicit form:
\[
	\mathcal{J}^{\pi,\mathbf{q},\tau}(z) := \big( \tau q_{\pi}(x) \big)^d \Bigg((1-\tau) \prod_{j=0}^{m} \big(\pi_j^\train q_j(x) \big)^{\I{y=j}} \Bigg)^{1-d} \Ibig{x\in\X, \text{ } y \in \Y, \text{ } d \in \{0,1\}  }.
\]

In terms of data, we receive $Z_1, \dots, Z_n \simiid \Jcal^{\pi^\ast,\mathbf{p},\tau}$, where $(\pi^\ast,\mathbf{p}) \in \Theta^\text{semi}$ denotes the true parameters. Here, $\pi^\train,\pi^\ast,\mathbf{p}$ and $\Theta^\text{semi}$ do \textit{not} depend on $n$. We assume $\pi^\train,\tau$ are known, and that $\pi^\ast,\mathbf{p}$ are unknown. \\

\subsubsection{$\tau_n$-IID Regime}

Let $\tau_n \in (0,1)$ be a sequence such that  $\tau_n n, (1-\tau_n)n \to \infty$ and  $\tau_n\to \tau_\infty$ for some $\tau_\infty \in [0,1]$. In the $\tau_n$-IID Regime, we receive $Z_1,\dots,Z_n \simiid \mathcal{J}^{\pi^\ast,\mathbf{p},\tau_n}$ for each $n$, where $\mathcal{J}^{\pi^\ast,\mathbf{p},\tau_n} \in \textbf{M}^{\text{semi},\tau_n}$ and 
\[
	\textbf{M}^{\text{semi},\tau_n} := \big\{ \Jcal^{\pi,\mathbf{q},\tau_n} \ \big\lvert \ (\pi,\mathbf{q}) \in \Theta^\text{semi} \big\}.
\]

Like in the Fixed $\tau$-IID Regime, $(\pi^\ast,\mathbf{p}) \in \Theta^\text{semi}$ and $\pi^\train$ do not depend on $n$; also, we assume $\pi^\train,\tau_n$ are known, and $\pi^\ast,\mathbf{p}$ are unknown. 

\newpage

\subsection{Fixed Sequence Regime}

\subsubsection{SELSE for Other Regimes}
The SELSE procedure at the end of subsection \eqref{sec: A Concise Description of Our Quantifier} is written using the notation of the Fixed Sequence Regime, but it is well defined for all three regimes considered in this Appendix. To obtain the equivalent procedure for the Fixed $\tau$-IID Regime and $\tau_n$-IID Regime, make the following notational adjustments:
\begin{itemize}
	\item $\Dtrain$ is those samples $(X_i,Y_i,D_i)$ for which $D_i = 0$; $\Dtest$ is those samples for which $D_i =1$.
	\item Replace $n^\test$ and $n^\train$ with $N^\test := \sum_{i=1}^n D_i$  and $N^\train := n - N^\test$, respectively. 
	\item For each $y\in\Y$, replace $\pi_y^\train$ with $\widehat{\pi}_y^\train := \frac{N_y^\train}{N^\train}$, where  $N_y^\train :=\sum_{i=1}^n (1-D_i) \I{Y_i = y}$. 
	\item Replace Assumption \eqref{gamma hat properties} with Assumption \eqref{tau n regime, estimator of gamma bar}.\\
\end{itemize}

\subsubsection{Assumptions}

We make the following assumptions about the space that $s_{\gamma^\ast}$ lives in:
\begin{enumerate}[label={A\arabic*.}, ref=A\arabic*]
	\item There exists a set of $\mathbb{R}^m$-valued functions $\H$ (which may depend on $n$)  s.t. $\H$ is closed under scalar multiplication and $s_{\gamma^\ast} \in \H$. \label{members of function class}
	\item For each $n$ and $y\in[m]$, let $\H_y := \{ h_y \mid h\in\H \}$ where $h_y$ denotes the $\numth{y}$ component of a function $h$. Then, the function $\Omega$ satisfies the following properties:
	\begin{itemize}
		\item $\Omega: \H_y \mapsto [0,\infty)$
		\item $\Omega$ is zero at the identically zero function
		\item $\Omega(ch_y) = \abs{c}\Omega(h_y)$ for all $h_y \in \H_y$ and $c\in\mathbb{R}$
		\item $\Omega(h_y) \geq D \norm{h_y}{\infty}$ for all $h_y \in \H_y$ and some fixed global constant $D> 0$.  \label{Omega Properties} 
	\end{itemize}
	\item For each $y\in [m]$, the sequence $\lambda_y$ satisfies the following properties: 
		\begin{itemize}
			\item $0 < \lambda_y < \infty$
			\item $ \lambda_y = o(1)$
			\item $\lambda_y \Omega^2(s_{\gamma^\ast,y}) = o(1)$
			\item $\lambda_y^2 n^\train = \omega(1)$
			\item $\E_{\DtrainII}[ \U_y(\lambda_y)] = o(1)$ where
			\begin{align*}
		\U_y(\lambda_y) & := \sup\limits_{h_y \in \H_y, \text{ } \Omega(h_y) \leq \frac{1}{L\sqrt{\lambda_y}}}  \absBig{\widehat{\E}_0^{\trainII}[h_y] - \E_0[h_y]} \\
		&\qquad+  \sup\limits_{h_y \in \H_y, \text{ } \Omega(h_y) \leq \frac{1}{L\sqrt{\lambda_y}}} \absBig{\widehat{\E}_y^{\trainII}[h_y] - \E_y[h_y]} \\
		&\qquad+  \sup\limits_{h_y \in \H_y , \text{ } \Omega(h_y) \leq \frac{1}{L\sqrt{\lambda_y}}} \absBig{ \widehat{\Var}_{\gamma^\ast}^{\trainII}[h_y] - \Var_{\gamma^\ast}[h_y]} .
	\end{align*}
		\end{itemize} \label{lambda and uniform convergence properties}
\end{enumerate}

We also make the following assumptions about our sequence of "true" parameters:
\begin{enumerate}[label={A\arabic*.}, ref=A\arabic*, resume]
	\item $\exists\ \nu > 0$ such that $\minSing{ \FisherInfo{\gamma^\ast;\text{Cat}} -  \FisherInfo{\gamma^\ast}  }  \geq \nu$, for all $n$.   \label{FIM Difference is Bounded}
	\item $\exists\ \xi \in (0,\tfrac{1}{2})$ such that $\pi_y^\train, \pi_y^\ast \in (\xi,1-\xi)$, for all $y\in\Y$ and $n$. \label{bounded pis}
	\item $\exists\ \Lambda > 0$ such that $\minSing{\FisherInfo{\gamma^\ast}} \geq \sqrt{\Lambda}$, for all $n$. \label{min Fisher Info}\\
\end{enumerate}

 Lastly, we assume several properties about the estimator $\widehat{\gamma}$: 
\begin{enumerate}[label={A\arabic*.}, ref=A\arabic*, resume]
	\item $\widehat{\gamma} \in [L,U]$, $\E_{\allDII}\big[ \norm{\widehat{\gamma} - \gamma^\ast}{2}^2\big] = o(1)$, and for each $a > 0$,  $\P_{\allDII}\big[  \norm{\widehat{\gamma} - \gamma^\ast}{2} \geq a\big] = o\Big(\sqrt{\frac{1}{n^\test} + \frac{1}{n^\train}}\Big)$. \label{gamma hat properties}\\
\end{enumerate}

\subsubsection{Lemmas}

It is important to note that all Corollaries, Lemmas, and Theorems before Corollary \eqref{Error Rate of Full Estimator} in the Fixed Sequence Regime concern themselves with \textit{only} the first run through of the SELSE procedure. That is, in those Corollaries, Lemmas, and Theorems, $\widehat{\gamma}$ is computed on $\allDII$, $\widehat{s}_{\widehat{\gamma}}$ is computed using $\allDII$ and $\widehat{\gamma}$, and $\widehat{\pi}^{(a)}$ is computed using $\allDI$ and $\widehat{s}_{\widehat{\gamma}}$.  In addition, several references are made to an estimator $\widehat{s}_{\gamma^\ast}$. This refers to a "hypothetical" estimator from the first run through of the SELSE procedure, obtained if we somehow knew $\gamma^\ast$ and so could replace $\widehat{\gamma}$ with $\gamma^\ast$ in the constraints of the optimization problem. As $\gamma^\ast$ is unknown (it depends on the unknown $\pi^\ast$), $\widehat{s}_{\gamma^\ast}$ cannot be computed, but for our theoretical analysis, analyzing its statistical properties help to elucidate those of $\widehat{s}_{\widehat{\gamma}}$ and therefore of $\widehat{\pi}^{(a)}$, and so ultimately, of $\widehat{\pi}$.
\vspace{0.3in}


\begin{lemma}[\uline{Bounds on $\gamma^\ast$}] \label{Bounded Gamma Star}
	Define
	\[
		L := \frac{1}{1+m\big(\frac{1-\xi}{\xi}\big)^3}, \qquad  U := \max\Bigg\{ 1-\xi , \text{ }  \frac{1}{1+m\big(\frac{\xi}{1-\xi}\big)^3 }\Bigg\}.
	\] 
	Then,  we have that 
	\[
		L \leq \frac{a \alpha_y+ b \frac{(\alpha_y)^2}{\beta_y} }{a + b \sum\limits_{k=0}^{m}\frac{(\alpha_k)^2}{\beta_k} }  \leq U \qquad \forall \ y\in\Y,
	\]
	whenever $a, b > 0$ and $\alpha,\beta$ belong to the $m+1$ dimensional probability simplex and satisfy $\xi\leq \alpha_y,\beta_y\leq 1-\xi$ for each $y\in\Y$.  Furthermore, as a special case, if we set $a = \frac{1}{n^\test}$,  $b = \frac{1}{n^\train}$, $\alpha = \pi^\ast$ and $\beta = \pi^\train$, then we have that $L \leq \gamma_y^\ast \leq U$ for each $y\in \Y$ under Assumption \eqref{bounded pis}.

\end{lemma}
\vspace{0.4in}

\begin{lemma}[\uline{Upper Bound on $\Omega(\widehat{s}_{\gamma^\ast,y})$}] \label{Bound on Norm of Score Estimate}
	Under Assumptions \eqref{members of function class},  \eqref{Omega Properties} and \eqref{bounded pis}, we have that 
	\[
		\Omega(\widehat{s}_{\gamma^\ast,y}) \leq \frac{1}{L\sqrt{\lambda_y}}
	\] 
	for all $y\in[m]$.
\end{lemma}
\vspace{0.4in}

\begin{lemma}[\uline{Bounds on Eigenvalues of $\FisherInfo{\beta;\textup{Cat}}$}] \label{Bounds on Eigenvalues of Categorical Fisher Info}
	Let $\beta$ be a vector in the $m+1$ dimensional probability simplex, and suppose that the entries in $\beta$ are bounded away from $0$ and $1$. Then, we have that
	\[
		2 \leq \minEval{\FisherInfo{\beta;\textup{Cat}}} ,\qquad \maxEval{\FisherInfo{\beta;\textup{Cat}}} \leq \frac{1}{\min_y \beta_y} + \frac{m}{\beta_0}.
	\]
\end{lemma}
\vspace{0.4in}

\begin{lemma}[\uline{Bounds on Eigenvalues of $\FisherInfo{\alpha;\text{Cat}}\FisherInfo{\beta;\text{Cat}}^{-1}$}] \label{Bounds on Eigenvalues of Inverse Cat FIM times Cat FIM, alpha vs. beta}
Let $\alpha,\beta$ be vectors in the $m+1$ dimensional probability simplex, and suppose that the entries in both vectors are bounded away from $0$ and $1$. Then, we have that
\[
	 1- (m+1) \frac{ \norm{\alpha - \beta}{1}  }{\min_y \alpha_y} \leq \minEval{\FisherInfo{\alpha;\textup{Cat}}\FisherInfo{\beta;\textup{Cat}}^{-1}}
\]
and
\[ 
	\maxEval{\FisherInfo{\alpha;\textup{Cat}}\FisherInfo{\beta;\textup{Cat}}^{-1}} \leq    1 +  (m+1)\frac{ \norm{\alpha - \beta}{1}  }{\min_y \alpha_y}.
\]
\end{lemma}
\vspace{0.4in}

\begin{lemma}[\uline{Bounds on Eigenvalues of $ \Var_{\pi^\ast}[s_{\gamma^\ast}] $}] \label{Bounds on Test Mixture Eigenvalues}
	Under Assumptions \eqref{bounded pis} and  \eqref{min Fisher Info}, we have that
	\[
		 \frac{\Lambda \xi}{m+1} \leq \minEval{ \Var_{\pi^\ast}[s_{\gamma^\ast}] }, \qquad \maxEval{\Var_{\pi^\ast}[s_{\gamma^\ast}]  } \leq \frac{m^2}{L^2}.
	\]
\end{lemma}
\vspace{0.4in}

\begin{lemma}[\uline{Bounds on $s_{\beta,j} - s_{\alpha,j}$}] \label{Bounds On Score Function Differences at Different Parameters}
Let $\alpha,\beta$ be vectors in the $m+1$ dimensional probability simplex, and suppose that the entries in both vectors are bounded away from $0$ and $1$. Then, we have that
\begin{align*}
	 \abs{s_{\beta,j} - s_{\alpha,j}} &\leq \frac{2}{\Big(\min\limits_{0\leq k \leq m}\alpha_k\Big)\Big(\min\limits_{0\leq k \leq m}\beta_k\Big)}  \sum_{k=0}^m \abs{\alpha_k - \beta_k},
\end{align*}
for each $j\in[m]$.
\end{lemma}
\vspace{0.4in}

\begin{lemma}[\uline{Lower Bound on $\minSing{\FisherInfo{\gamma^\ast}^{-1} - \FisherInfo{\gamma^\ast;\textup{Cat}}^{-1} }$}] \label{Inverse FIM Difference is Bounded}
	Under Assumptions \eqref{FIM Difference is Bounded} and \eqref{bounded pis}, we have that
	\[
		\minSing{\FisherInfo{\gamma^\ast}^{-1} - \FisherInfo{\gamma^\ast;\textup{Cat}}^{-1} } \geq \nu \bigg( \frac{L}{m+1}\bigg)^2.
	\]
\end{lemma}
\vspace{0.4in}

\begin{corollary}[\uline{Relationship between Variances of Different Mixtures}] \label{Relationship between Variances of Different Mixtures}
Let $\alpha,\beta$ be vectors in the $m+1$ dimensional probability simplex, and suppose that the entries in both vectors are bounded away from $0$ and $1$.  Then, for any function $f: \mathcal{X} \mapsto \mathbb{R} $, we have that:
\[
	\Var_{\alpha}[f] \leq \bigg(\frac{1}{\min_y\beta_y} + \frac{m}{\beta_0}  \bigg)\Var_{\beta}[f].
\]
\end{corollary}
\vspace{0.4in}

\begin{corollary}[\uline{First Order Constant for $f = s_{\gamma^\ast}$}] \label{First Order Constant When Matching Function is Score Function}
The matrix $\FisherInfo{\gamma^\ast}$ is positive definite under Assumption  \eqref{min Fisher Info} and satisfies $\FisherInfo{\gamma^\ast} = A_{s_{\gamma^\ast}}$. Further,  
\begin{align*}
	\Var_{\allDI}\Big[\FisherInfo{\gamma^\ast}^{-1}(\widehat{\E}_{\pi^\ast}^\testI[s_{\gamma^\ast}] -  \widehat{\E}_{\pi^\ast}^\trainI[s_{\gamma^\ast}]) \Big] &= \Bigg[\frac{2}{n^\test}+ \frac{2}{n^\train} \sum_{y=0}^{m}\frac{(\pi_y^\ast)^2}{\pi_y^\train}\Bigg] \Big( \FisherInfo{\gamma^\ast}^{-1} - \FisherInfo{\gamma^\ast;\textup{Cat}}^{-1}  \Big)  \\
	&\qquad\qquad+ \frac{2}{n^\test} \FisherInfo{\pi^\ast;\textup{Cat}}^{-1},
\end{align*}
which is positive definite under Assumption \eqref{bounded pis}.
\end{corollary}
\vspace{0.4in}

\begin{lemma}[\uline{Alternative Expression for First Order Error}] \label{Alternative Expression for First Order Error}
Consider an arbitrary ordering $i=1,\dots,n$ of the data points in $\allD$, where $i \in [1,n/2]$ if the $\numth{i}$ data point is from $\allDI$ and $i\in[n/2 + 1, n]$ if the $\numth{i}$ data point is from $\allDII$. Let $Z_i := (X_i,Y_i,D_i)$ be defined as follows. If the $\numth{i}$ data point is from the test set, then $(X_i,Y_i,D_i) = (X_i,-1,1)$. If the $\numth{i}$ data point is from the training set, then $(X_i,Y_i,D_i) = (X_i,Y_i,0)$. Then, we have that
\begin{align*}
	\FisherInfo{\gamma^\ast}^{-1}\big(\widehat{\E}_{\pi^\ast}^{\testI}[s_{\gamma^\ast}] - \widehat{\E}_{\pi^\ast}^{\trainI}[s_{\gamma^\ast}] \big) &= \frac{1}{n/2} \sum_{i=1}^{n/2} \psi^\eff(Z_i),
\end{align*}
where $\Z := \X \times \Y \times \{0,1\}$ and, for each $z = (x,y,d)\in \Z$, the function $\psi^\eff: \Z \mapsto \mathbb{R}^m$ is given by
\begin{align*}
	\psi^\eff(z) &:= \frac{d}{n^\test / n} \FisherInfo{\gamma^\ast}^{-1}\big( s_{\gamma^\ast}(x) - \E_{\pi^\ast}[s_{\gamma^\ast}] \big) \\
	&\qquad\qquad - \frac{1-d}{n^\train/n} \sum_{j=0}^m \I{y=j} \frac{\pi_j^\ast}{ n_j^\train /n^\train } \FisherInfo{\gamma^\ast}^{-1}\big( s_{\gamma^\ast}(x) - \E_{j}[s_{\gamma^\ast}] \big) 
\end{align*}
where $n_j^\train := \frac{1}{n/2} \sum_{k=1}^{n/2}\I{Y_k = j}$ is the fraction of the training points in $\DtrainI$ that belong to class $j$. Analogously, we also have that
\begin{align*}
	\FisherInfo{\gamma^\ast}^{-1}\big(\widehat{\E}_{\pi^\ast}^{\test}[s_{\gamma^\ast}] - \widehat{\E}_{\pi^\ast}^{\train}[s_{\gamma^\ast}] \big) &= \frac{1}{n} \sum_{i=1}^{n} \psi^\eff(Z_i).
\end{align*}

\end{lemma}
\vspace{0.4in}


\begin{lemma}[\uline{Uniform Deviation Tail Bounds}]\label{Uniform Deviation Tail Bounds}
Under Assumptions \eqref{Omega Properties} and \eqref{bounded pis}, we have for each $y\in [m]$, $j \in 0\cup[m]$ and all $a > 0$ that
\begin{align*}
	&\P_{\DtrainII}\Bigg[ \sup_{\Omega(h_y) \leq \frac{1}{L\sqrt{\lambda_y}}}\absBig{\widehat{\E}_j^{\trainII}[h_y] - \E_j[h_y] } \geq a\Bigg] \\
	&\leq \exp\Bigg\{ -\frac{D^2 L^2 \xi }{16} \lambda_y n^\train a^2 \Bigg\} \\
	&\qquad +  \IBigg{   \E_{\DtrainII}\bigg( \sup_{\Omega(h_y) \leq \frac{1}{L \sqrt{\lambda_y}}} \absBig{\widehat{\E}_j^{\trainII}[h_y] - \E_j[h_y]} \bigg) \geq  \frac{1}{2}a  }
\end{align*}
and
\begin{align*}
	&\P_{\DtrainII}\Bigg[\sup_{\Omega(h_y) \leq  \frac{1}{L\sqrt{\lambda_y}} } \absBig{ \widehat{\Var}_{\gamma^\ast}^{\trainII}[h_y]  -  \Var_{\gamma^\ast}[h_y]}   \geq a \Bigg] \\
	&\leq \exp\Bigg\{- \frac{D^4L^4}{1024}\lambda_y^2 n^\train a^2 \Bigg\} \\ 
	&\qquad + \IBigg{\E_{\DtrainII}\Bigg(\sup_{\Omega(h_y) \leq  \frac{1}{L\sqrt{\lambda_y}} } \absBig{ \widehat{\Var}_{\gamma^\ast}^{\trainII}[h_y]  -  \Var_{\gamma^\ast}[h_y]}\Bigg) \geq \frac{1}{2}a }.
\end{align*}
\end{lemma}
\vspace{0.4in}

\begin{lemma}[\uline{Rate for Learning Diagonal of $\FisherInfo{\gamma^\ast}$, with Unknown $\gamma^\ast$}] \label{Rate for Learning Diagonal of Fisher Information Matrix, Unknown Gamma Star}
Under Assumptions  \eqref{members of function class},  \eqref{Omega Properties}, \eqref{lambda and uniform convergence properties}, \eqref{bounded pis}, \eqref{min Fisher Info} and \eqref{gamma hat properties},  we have for each $y\in[m]$ that
\[
	\big(\E_y[s_{\gamma^\ast,y}] - \E_0[s_{\gamma^\ast,y}]\big)  -  \big(\E_y[\widehat{s}_{\widehat{\gamma},y}]-\E_0[\widehat{s}_{\widehat{\gamma},y}]\big) \leq \widetilde{T}_y,
\]
where $\widetilde{T}_y > 0$ is a random variable that satisfies
\[
	\E_{\allDII}[\widetilde{T}_y] = O\Bigg(\E_{\DtrainII}[T_y] + \E_{\DtrainII}[\U_y(\lambda_y)]  + \sqrt{\E_{\allDII}\Big[\norm{\widehat{\gamma} - \gamma^\ast}{2}^2\Big]} +  \E_{\DtrainII}\Big[\Var_{\gamma^\ast}[\widehat{s}_{\gamma^\ast,y} - s_{\gamma^\ast,y}]\Big] \Bigg).
\]
Further, for all $a > 0$,
\begin{align*}
	\P_{\allDII}[\widetilde{T}_y \geq a] &\leq \P_{\DtrainII}[T_y \geq a/7] + \P_{\DtrainII}\Bigg[ \sup_{\Omega(h_y) \leq \frac{1}{L\sqrt{\lambda_y}}} \absbig{\widehat{\E}_y^{\trainII}[h_y] - \E_y[h_y]} \geq \frac{1}{14}a \Bigg] \\
	&\qquad + \P_{\DtrainII}\Bigg[ \sup_{\Omega(h_y) \leq \frac{1}{L\sqrt{\lambda_y}}} \absbig{\widehat{\E}_0^{\trainII}[h_y] - \E_0[h_y]} \geq \frac{1}{14}a \Bigg] \\ 
	&\qquad + \P_{\DtrainII}\Bigg[ \sup_{\Omega(h_y) \leq \frac{1}{L\sqrt{\lambda_y}} } \absbig{ \widehat{\Var}_{\gamma^\ast}^\trainII[h_y] - \Var_{\gamma^\ast}[h_y] } \geq \frac{1}{14}\frac{L}{m+1}a \Bigg] \\
	&\qquad + 2\P_{\allDII}\Bigg[  \norm{\widehat{\gamma} - \gamma^\ast}{2} \geq \widetilde{\tau}_1(a) \Bigg] \\ 
	&\qquad +  2\P_{\DtrainII}\Bigg[  \Var_{\gamma^\ast}[\widehat{s}_{\gamma^\ast,y} - s_{\gamma^\ast,y}]  \geq \frac{1}{14}\frac{L^2}{m+1}a \Bigg],
\end{align*}
where 
\[
	\widetilde{\tau}_1(a) := \min\Bigg\{\frac{L^4}{21\sqrt{m}(m+1) +  14\sqrt{m}(m+1)^2} a, \sqrt{\frac{1}{14}\frac{L^2}{m(m+1)}a}\Bigg\}.
\]
\end{lemma}
\vspace{0.4in}

\begin{lemma}[\uline{Rate for Learning the Score Function, with Unknown $\gamma^\ast$}] \label{Rate for Learning the Score Function, Unknown Gamma Star}
Under Assumptions  \eqref{members of function class}, \eqref{Omega Properties}, \eqref{lambda and uniform convergence properties}, \eqref{bounded pis}, \eqref{min Fisher Info}  and \eqref{gamma hat properties}, we have for each $y\in[m]$ that 
\begin{align*}
	\E_{\allDII}\Big[ \Var_{\gamma^\ast}[\widehat{s}_{\widehat{\gamma},y} - s_{\gamma^\ast,y}] \Big] &= O\Bigg(\E_{\allDII}[\widetilde{T}_y] + \E_{\DtrainII}\Big[ \U_y(\lambda_y)\Big] + \sqrt{\E_{\allDII}\Big[ \norm{\widehat{\gamma} - \gamma^\ast}{2}^2 \Big]} \\
	&\qquad\qquad+  \E_{\DtrainII}\Big[ \Var_{\gamma^\ast}[\widehat{s}_{\gamma^\ast,y} - s_{\gamma^\ast,y}] \Big]  \Bigg).
\end{align*}

Further, for all $a > 0$, we have that:
\begin{align*}
	&\P_{\allDII}\Big[  \Var_{\gamma^\ast}[\widehat{s}_{\widehat{\gamma},y} - s_{\gamma^\ast,y}]  \geq  a \Big] \\
	&\leq \P_{\allDII}[\widetilde{T}_y \geq a/7]   +   \P_{\DtrainII}\Bigg[ \sup_{\Omega(h_y) \leq \frac{1}{L\sqrt{\lambda_y}} } \absBig{\widehat{\E}_y^{\trainII}[h_y] - \E_y[h_y]} \geq a/7 \Bigg] \\ 
	&\qquad + \P_{\DtrainII}\Bigg[ \sup_{\Omega(h_y) \leq \frac{1}{L\sqrt{\lambda_y}} } \absBig{\widehat{\E}_0^{\trainII}[h_y] - \E_0[h_y]} \geq a/7 \Bigg] \\
	&\qquad+ \P_{\DtrainII}\Bigg[\sup_{\Omega(h_y) \leq \frac{1}{L\sqrt{\lambda_y}}} \absBig{\widehat{\Var}_{\gamma^\ast}^\trainII[h_y]  -   \Var_{\gamma^\ast}[h_y] } \geq \frac{L^3}{77(m+2)^3}a \Bigg]\\
	&\qquad + 2\P_{\allDII}\Bigg[ \norm{\widehat{\gamma} - \gamma^\ast}{2}\geq \widetilde{\tau}_2(a)\Bigg] \\ 
	&\qquad+ 2\P_{\DtrainII}\Bigg[   \Var_{\gamma^\ast}[\widehat{s}_{\gamma^\ast,y} - s_{\gamma^\ast,y} ]  \geq  \frac{L^4}{70(m+2)^3}a  \Bigg],
\end{align*}
where 
\[
	\widetilde{\tau}_2(a) := \min\Bigg\{\sqrt{ \frac{L^4}{70m(m+2)^3}a } , \frac{L^5}{66\sqrt{m}(m+2)^3}a  \Bigg\}.
\]

\end{lemma}
\vspace{0.4in}

\begin{lemma}[\uline{Rate for Learning Columns of $\FisherInfo{\gamma^\ast}$}] \label{Rate for Learning Columns of Fisher Information Matrix}
	Under Assumptions  \eqref{members of function class}, \eqref{Omega Properties}, \eqref{lambda and uniform convergence properties}, \eqref{bounded pis}, \eqref{min Fisher Info} and \eqref{gamma hat properties}, we have for each $y\in[m]$ that 
	\[
		\E_{\allD}\bigg[ \Bignorm{ \textup{col } y \textup{ of } \widehat{A} - \FisherInfo{\gamma^\ast} }{2}^2 \bigg] = O\Bigg(  \frac{1}{n^\train}  +    \sum_{j=1}^{m}  \E_{\allDII}\Big[\Var_{\gamma^\ast}[\widehat{s}_{\widehat{\gamma},j}-s_{\gamma^\ast,j}]\Big]  \Bigg).
	\]
	Further, for all $a > 0$,	
	\begin{align*}
		\P_{\allD}\Big[\norm{ \textup{col } y \textup{ of } \widehat{A} - \FisherInfo{\gamma^\ast} }{2}^2 \geq a \Big] &\leq 4\sum_{j=1}^{m}\exp\bigg\{-\frac{  \xi  L^2 D^2 \lambda_j n^\train }{64m   }a   \bigg\} \\
		&\qquad + \sum_{j=1}^{m}  \P_{\allDII}\bigg[ \Var_{\gamma^\ast}[\widehat{s}_{\widehat{\gamma},j}-s_{\gamma^\ast,j}]  \geq \frac{L^2}{4m}a \bigg].
	\end{align*}
\end{lemma}
\vspace{0.4in}

\begin{lemma}[\uline{Learning the Inverse of $\FisherInfo{\gamma^\ast}^{-1}$}] \label{Learning the Inverse FIM}
	Under Assumptions  \eqref{members of function class}, \eqref{Omega Properties}, \eqref{lambda and uniform convergence properties}, \eqref{bounded pis}, \eqref{min Fisher Info} and \eqref{gamma hat properties}, we have that
	\[
		\I{ \minSing{\widehat{A}}=0} =  O_{\P}\Bigg(   \sum_{i=1}^{m} \P_{\Dtrain}\Bigg[   \Bignorm{\textup{col } i \textup{ of } \widehat{A} - \FisherInfo{\gamma^\ast}}{2}^2  \geq \frac{1}{m^2}\min\bigg\{ \frac{1}{2}\Lambda  ,\text{ } \frac{1}{16}\frac{L^4}{m^{3}} \Lambda^2\bigg\} \Bigg]  \Bigg)
	\]
	and that
	\[
		\maxSing{\widehat{A}^{-1} - \FisherInfo{\gamma^\ast}^{-1}}\I{ \minSing{\widehat{A}}>0} = O_{\P}\Bigg( \frac{1}{\sqrt{n^\train}}  +     \sqrt{\sum_{k=1}^{m}  \E_{\allDII}\Big[\Var_{\gamma^\ast}[\widehat{s}_{\widehat{\gamma},k}-s_{\gamma^\ast,k}]\Big] }  \Bigg).
	\]
\end{lemma}
\vspace{0.4in}

\begin{lemma}[\uline{Rate of Second Order Term}] \label{Rate of Second Order Term}
	Under Assumptions  \eqref{members of function class}, \eqref{Omega Properties}, \eqref{lambda and uniform convergence properties}, \eqref{bounded pis}, \eqref{min Fisher Info} and \eqref{gamma hat properties}, we have that
	\[
		\E_{\allD}\bigg[\Bignorm{ \widehat{\E}_{\pi^\ast}^\testI[\widehat{s}_{\widehat{\gamma}}-s_{\gamma^\ast}] -  \widehat{\E}_{\pi^\ast}^\trainI[\widehat{s}_{\widehat{\gamma}}-s_{\gamma^\ast}]     }{2}^2 \bigg] =  O\Bigg( \bigg( \frac{1}{n^\test} + \frac{1}{n^\train} \bigg)  \sum_{j=1}^{m}\E_{\allDII}\Big[\Var_{\gamma^\ast}[\widehat{s}_{\widehat{\gamma},j} - s_{\gamma^\ast,j}]  \Big] \Bigg).
	\]
\end{lemma}
\vspace{0.4in}

\begin{theorem}[\uline{Error Rate of $\widehat{\pi}^{(a)}$}] \label{Error Rate of First Step Estimator}
Under Assumptions  \eqref{members of function class}, \eqref{Omega Properties}, \eqref{lambda and uniform convergence properties}, \eqref{FIM Difference is Bounded}, \eqref{bounded pis}, \eqref{min Fisher Info} and \eqref{gamma hat properties}, we have that
\begin{align*}
	\widehat{\pi}^{(a)} - \pi^\ast &= \frac{1}{n/2} \sum_{i=1}^{n/2} \psi^\eff(Z_i) +  \epsilon^{(a)} \\ 
	&= \sqrt{2}\Bigg( \Bigg[\frac{1}{n^\test}+ \frac{1}{n^\train} \sum_{y=0}^{m}\frac{(\pi_y^\ast)^2}{\pi_y^\train}\Bigg] \Big( \FisherInfo{\gamma^\ast}^{-1} - \FisherInfo{\gamma^\ast;\textup{Cat}}^{-1}  \Big)  + \frac{1}{n^\test} \FisherInfo{\pi^\ast;\textup{Cat}}^{-1} \Bigg)^{\frac{1}{2}} Z^{(a)} \\
	&\qquad+ \epsilon^{(a)},
\end{align*}
where $Z^{(a)} \to \N(0, I_m)$, and $\epsilon^{(a)} \in \mathbb{R}^m$ is a random vector satisfying   
\small
\begin{align*}
	\norm{\epsilon^{(a)}}{2} &=   O_{\P}\Bigg(\sqrt{\bigg( \frac{1}{n^\test} + \frac{1}{n^\train} \bigg) \bigg\{ \frac{1}{\sqrt{n^\train}} + \max_y \lambda_y \Omega(s_{\gamma^\ast,y})^2 +  \frac{e^{-Cn^\train}}{\sqrt{\min_y\lambda_y}}   \bigg\}   }\Bigg) \\ 
	&\qquad  + O_{\P}\Bigg(\sqrt{\bigg( \frac{1}{n^\test} + \frac{1}{n^\train} \bigg) \bigg\{ \max_y \E_{\DtrainII}\Big[ \U_y(\lambda_y)\Big] + \sqrt{\E_{\allDII}\Big[ \norm{\widehat{\gamma} - \gamma^\ast}{2}^2 \Big]}  \bigg\}   }\Bigg) \\ 
	&\qquad + O_{\P}\Bigg( e^{-C (\min_y\lambda_y)^2 n^\train}  + \P_{\allDII}\Big[\bignorm{ \widehat{\gamma} - \gamma^\ast }{2} \geq C \Big]  \Bigg),
\end{align*}
\normalsize
for some global constant $C > 0$.
\end{theorem}
\vspace{0.4in}

\begin{corollary}[\uline{Error Rate of $\widehat{\pi}$}] \label{Error Rate of Full Estimator}
Under Assumptions  \eqref{members of function class}, \eqref{Omega Properties}, \eqref{lambda and uniform convergence properties}, \eqref{FIM Difference is Bounded}, \eqref{bounded pis}, \eqref{min Fisher Info} and \eqref{gamma hat properties}, we have that 
\begin{align*}
	\widehat{\pi} - \pi^\ast &= \frac{1}{n} \sum_{i= 1}^{n} \psi^\eff(Z_i) + \epsilon \\
	&= \Bigg( \Bigg[\frac{1}{n^\test}+ \frac{1}{n^\train} \sum_{y=0}^{m}\frac{(\pi_y^\ast)^2}{\pi_y^\train}\Bigg] \Big( \FisherInfo{\gamma^\ast}^{-1} - \FisherInfo{\gamma^\ast;\textup{Cat}}^{-1}  \Big)  + \frac{1}{n^\test} \FisherInfo{\pi^\ast;\textup{Cat}}^{-1} \Bigg)^{\frac{1}{2}} Z + \epsilon,
\end{align*}
where $Z \goesto{d} \N(0,I_m)$, and $\epsilon \in \mathbb{R}^m$ is a random vector satisfying   
\small
\begin{align*}
	\norm{\epsilon}{2} &=   O_{\P}\Bigg(\sqrt{\bigg( \frac{1}{n^\test} + \frac{1}{n^\train} \bigg) \bigg\{ \frac{1}{\sqrt{n^\train}} + \max_y \lambda_y \Omega(s_{\gamma^\ast,y})^2 +  \frac{e^{-Cn^\train}}{\sqrt{\min_y\lambda_y}}  }\Bigg) \\ 
	&\qquad+   O_{\P}\Bigg(\sqrt{\bigg( \frac{1}{n^\test} + \frac{1}{n^\train} \bigg) \bigg\{   \max_y \E_{\DtrainII}\Big[ \U_y(\lambda_y)\Big] + \sqrt{\E_{\allDII}\Big[ \norm{\widehat{\gamma} - \gamma^\ast}{2}^2 \Big]}  \bigg\}   }\Bigg) \\ 
	&\qquad + O_{\P}\Bigg( e^{-C (\min_y\lambda_y)^2 n^\train}  + \P_{\allDII}\Big[\bignorm{ \widehat{\gamma} - \gamma^\ast }{2} \geq C \Big]  \Bigg),
\end{align*}
\normalsize
for some global constant $C > 0$. 
\end{corollary}
\vspace{0.4in}

\begin{corollary}[\uline{RKHS Application}] \label{RKHS Application}
Let $\K$ denote a set of reproducing kernels, where $\K$ may depend on $n$. For each kernel $k \in \K$, let $H_k$ denote the corresponding RKHS and  $\norm{\cdot}{k}$ the RKHS norm. Define $H_\K$ and $\Omega_{\K}: H_{\K}\to [0,\infty)$ by 
\[
	H_{\K} := \cup_{k\in\K} H_k
\]
\[
	\Omega_{\K}(\ell) := \inf\Big\{  \norm{\ell}{k} \ \Big\lvert \ k\in\K, \ \ell \in H_k \Big\} \qquad \forall \ \ell \in H_\K. 
\]
Assume that:
\begin{itemize}
	\item For each $n$, $\sup\limits_{k\in\K}\sup\limits_{x_1,x_2\in\X} \sqrt{k(x_1,x_2)} \leq \kappa$, where $\kappa > 0$ is a finite, global constant. 
	\item For each $n$, the psuedo-dimension $d_{\K}$ of $\K$ (see Definition 3 in \citep{RadChaosComplexityBound}) is finite.
\end{itemize}

If $s_{\gamma^\ast} \in H_{\K}^m$, $\H = H_{\K}^m$ and $\Omega = \Omega_{\K}$, then Assumptions \eqref{members of function class} and \eqref{Omega Properties} are satisfied with $D = \frac{1}{\kappa}$. Further, for each class $y\in [m]$:
\[
	\E_{\DtrainII}\big[ \U_y(\lambda_y)\big] = O\Bigg( \frac{1}{\lambda_y}\sqrt{\frac{1+d_{\K}}{n^\train}} \Bigg).
\]
\end{corollary}

\vspace{0.4in}

\begin{corollary}[\uline{Standardized Error Rate of $\widehat{\pi}$, Application to RKHS}] \label{Standardized Error Rate of Full Estimator, Application to RKHS}
In the setting of Corollary \eqref{RKHS Application}, if  Assumptions  \eqref{FIM Difference is Bounded}, \eqref{bounded pis}, \eqref{min Fisher Info} and \eqref{gamma hat properties} hold, and if additionally we have that
\begin{itemize}
	\item $ \frac{1+d_{\K}}{n^\train} = o(1)$
	\item $\max_{j} \Omega(s_{\gamma^\ast,j}) = o\bigg(\frac{  (n^\train)^{\frac{1}{4}} }{\max\big\{ \sqrt{\log(n^\train)}, (1+d_{\K})^{1/4} \big\} } \bigg)$,
\end{itemize}
then, by choosing
\[
	\lambda_y \asymp \frac{1}{\max_{j} \Omega(s_{\gamma^\ast,j})} \bigg( \frac{1+d_{\K}}{n^\train} \bigg)^{\frac{1}{4}} \qquad \forall y\in[m],
\]
we obtain that
\begin{align*}
	\Bigg( \Bigg[\frac{1}{n^\test}+ \frac{1}{n^\train} \sum_{y=0}^{m}\frac{(\pi_y^\ast)^2}{\pi_y^\train}\Bigg] \Big( \FisherInfo{\gamma^\ast}^{-1} - \FisherInfo{\gamma^\ast;\textup{Cat}}^{-1}  \Big)  + \frac{1}{n^\test} \FisherInfo{\pi^\ast;\textup{Cat}}^{-1} \Bigg)^{-\frac{1}{2}} (\widehat{\pi} - \pi^\ast) = Z + \epsilon,
\end{align*}
where 
\begin{align*}
 	Z &= \Bigg( \Bigg[\frac{1}{n^\test}+ \frac{1}{n^\train} \sum_{y=0}^{m}\frac{(\pi_y^\ast)^2}{\pi_y^\train}\Bigg] \Big( \FisherInfo{\gamma^\ast}^{-1} - \FisherInfo{\gamma^\ast;\textup{Cat}}^{-1}  \Big)  + \frac{1}{n^\test} \FisherInfo{\pi^\ast;\textup{Cat}}^{-1} \Bigg)^{-\frac{1}{2}}\bigg(\frac{1}{n} \sum_{i= 1}^{n} \psi^\eff(Z_i)\bigg) \\
	&\goesto{d} \N(0,I_m),
\end{align*}
and $\epsilon \in \mathbb{R}^m$ is a random vector satisfying
\begin{align*}
	\norm{\epsilon}{2} &= O_{\P}\Bigg(   \bigg(\frac{1+d_\K}{n^\train}\bigg)^{\tfrac{1}{8}}\sqrt{\max_{y} \Omega(s_{\gamma^\ast,y})} + \Big(\E_{\allDII}\Big[ \norm{\widehat{\gamma} - \gamma^\ast}{2}^2 \Big]\Big)^{\frac{1}{4}} + \frac{\P_{\allDII}\big[\norm{ \widehat{\gamma} - \gamma^\ast }{2} \geq C \big]}{\sqrt{\tfrac{1}{n^\test} + \tfrac{1}{n^\train}} }\Bigg) \\ 
	&= o_{\P}(1),
\end{align*}
and $C > 0$ is a global constant. \\
\end{corollary}

\subsubsection{Proofs}

\begin{proof}[\uline{Proof of Lemma \eqref{Bounded Gamma Star}}]

Let any $y \in \Y$ be given, and for convenience, define
\[
	q_y :=\frac{a \alpha_y+ b \frac{(\alpha_y)^2}{\beta_y} }{a + b \sum\limits_{k=0}^{m}\frac{(\alpha_k)^2}{\beta_k} }  
\]
We will prove the desired inequality for $q_y$; the desired result for $\gamma_y^\ast$ then follows as a special case of this more general inequality, namely by setting $a = \frac{1}{n^\test}$, $b = \frac{1}{n^\train}$, $\alpha = \pi^\ast$ and $\beta = \pi^\train$. 

To prove the desired inequality for $q_y$, begin by applying the mediant inequality to the two fractions $\alpha_y  = \frac{a \alpha_y}{a}$ and $\frac{\frac{(\alpha_y)^2}{\beta_y} }{ \sum\limits_{k=0}^{m}\frac{(\alpha_k)^2}{\beta_k} } = \frac{ b\frac{(\alpha_y)^2}{\beta_y} }{b \sum\limits_{k=0}^{m}\frac{(\alpha_k)^2}{\beta_k} }$; then, because $q_y$ is their mediant, we find ourselves with the following two cases:\\
\begin{itemize}
	\item If $\alpha_y\leq \frac{\frac{(\alpha_y)^2}{\beta_y} }{ \sum\limits_{k=0}^{m}\frac{(\alpha_k)^2}{\beta_k} }$, then $\alpha_y \leq q_y\leq \frac{\frac{(\alpha_y)^2}{\beta_y} }{ \sum\limits_{k=0}^{m}\frac{(\alpha_k)^2}{\beta_k} }$.
	\item If $\frac{\frac{(\alpha_y)^2}{\beta_y} }{ \sum\limits_{k=0}^{m}\frac{(\alpha_k)^2}{\beta_k} } < \alpha_y$, then $  \frac{\frac{(\alpha_y)^2}{\beta_y} }{ \sum\limits_{k=0}^{m}\frac{(\alpha_k)^2}{\beta_k} }  \leq q_y \leq   \alpha_y$.\\
\end{itemize}

Since $ \xi \leq \alpha_y \leq 1-\xi$, so the display above implies that:
\begin{itemize}
	\item If $\alpha_y\leq \frac{\frac{(\alpha_y)^2}{\beta_y} }{ \sum\limits_{k=0}^{m}\frac{(\alpha_k)^2}{\beta_k} }$, then $\xi \leq q_y\leq \frac{\frac{(\alpha_y)^2}{\beta_y} }{ \sum\limits_{k=0}^{m}\frac{(\alpha_k)^2}{\beta_k} }$.
	\item If $\frac{\frac{(\alpha_y)^2}{\beta_y} }{ \sum\limits_{k=0}^{m}\frac{(\alpha_k)^2}{\beta_k} } < \alpha_y$, then $  \frac{\frac{(\alpha_y)^2}{\beta_y} }{ \sum\limits_{k=0}^{m}\frac{(\alpha_k)^2}{\beta_k} }  \leq q_y \leq   1-\xi$.\\
\end{itemize}

Next, let's obtain upper and lower bounds on $\frac{\frac{(\alpha_y)^2}{\beta_y} }{ \sum\limits_{k=0}^{m}\frac{(\alpha_k)^2}{\beta_k} } $ in terms of $\xi$. Note that:
\[
	\frac{\frac{(\alpha_y)^2}{\beta_y} }{ \sum\limits_{k=0}^{m}\frac{(\alpha_k)^2}{\beta_k} }  = \frac{\frac{(\alpha_y)^2}{\beta_y} }{\frac{(\alpha_y)^2}{\beta_y} +  \sum\limits_{k=0, k\neq y}^{m}\frac{(\alpha_k)^2}{\beta_k} } = \frac{1}{1 +  \frac{\beta_y}{ (\alpha_y)^2 }  \sum\limits_{k=0, k\neq y}^{m}\frac{(\alpha_k)^2}{\beta_k} }
\]

Now, for each $k \in \Y$, the fact that $\xi \leq \alpha_k, \beta_k \leq 1-\xi$ implies that $ \frac{\xi^2}{1-\xi} \leq \frac{(\alpha_k)^2}{\beta_k}  \leq  \frac{(1-\xi)^2}{\xi}$ and $  \frac{\xi}{(1-\xi)^2} \leq \frac{\beta_y}{ (\alpha_y)^2 }   \leq \frac{1-\xi}{\xi^2}$, so
\[
	m \frac{\xi^2}{1-\xi} \leq \sum\limits_{k=0, k\neq y}^{m}\frac{(\alpha_k)^2}{\beta_k} \leq m \frac{(1-\xi)^2}{\xi}
\]
and so 
\[
	\implies m \frac{\xi^3}{(1-\xi)^3} \leq  \frac{\beta_y}{ (\alpha_y)^2 }  \sum\limits_{k=0, k\neq y}^{m}\frac{(\alpha_k)^2}{\beta_k}  \leq m \frac{(1-\xi)^3}{\xi^3}
\]
\[
	\implies 1 + m \bigg(\frac{\xi}{1-\xi}\bigg)^3 \leq  1 + \frac{\beta_y}{ (\alpha_y)^2 }  \sum\limits_{k=0, k\neq y}^{m}\frac{(\alpha_k)^2}{\beta_k} \leq 1 + m  \bigg(\frac{1-\xi}{\xi}\bigg)^3
\]
\[
	\implies  \frac{1}{1 + m  \big(\frac{1-\xi}{\xi}\big)^3}  \leq  \frac{1}{1 + \frac{\beta_y}{ (\alpha_y)^2 }  \sum\limits_{k=0, k\neq y}^{m}\frac{(\alpha_k)^2}{\beta_k} } \leq   \frac{1}{1 + m \big(\frac{\xi}{1-\xi}\big)^3}    
\]
\[
	\implies  \frac{1}{1 + m  \big(\frac{1-\xi}{\xi}\big)^3}  \leq  \frac{\frac{(\alpha_y)^2}{\beta_y} }{ \sum\limits_{k=0}^{m}\frac{(\alpha_k)^2}{\beta_k} }  \leq   \frac{1}{1 + m \big(\frac{\xi}{1-\xi}\big)^3}    
\]

Thus, based on our earlier observation from the mediant inequality, our two cases now become:
\begin{itemize}
	\item If $\alpha_y\leq \frac{\frac{(\alpha_y)^2}{\beta_y} }{ \sum\limits_{k=0}^{m}\frac{(\alpha_k)^2}{\beta_k} }$, then $ \xi \leq q_y \leq \frac{1}{1 + m \big(\frac{\xi}{1-\xi}\big)^3}    $.
	\item If $\frac{\frac{(\alpha_y)^2}{\beta_y} }{ \sum\limits_{k=0}^{m}\frac{(\alpha_k)^2}{\beta_k} } < \alpha_y$, then $  \frac{1}{1 + m  \big(\frac{1-\xi}{\xi}\big)^3}  \leq q_y \leq  1-\xi $.\\
\end{itemize}

Now, let $L = \min\Bigg\{\xi ,\text{ }   \frac{1}{1 + m  \big(\frac{1-\xi}{\xi}\big)^3}  \Bigg\}$ and $U = \max\Bigg\{1-\xi , \text{ } \frac{1}{1 + m \big(\frac{\xi}{1-\xi}\big)^3}\Bigg\}$. Then, regardless of which of the above two cases we are in, we always have that:
\[
	\implies L \leq q_y \leq U.
\]

Finally, the desired result comes from noticing that:
\[
	\min\Bigg\{\xi ,\text{ }   \frac{1}{1 + m  \big(\frac{1-\xi}{\xi}\big)^3 }  \Bigg\} = \frac{1}{1 + m  \big(\frac{1-\xi}{\xi}\big)^3}.
\]

\end{proof}
\vspace{0.4in}

\begin{proof}[\uline{Proof of Lemma \eqref{Bound on Norm of Score Estimate}}]

For each $k \in \Y$, let $\D_k^{\trainII}$ denote the subset of $\DtrainII$ that belongs to class $k$. Let $\widehat{p}_k(x) := \frac{1}{\abs{\D_k^{\trainII}}} \Ibig{x \in \D_k^{\trainII}}$ be the empirical probability mass function associated with $\D_k^{\trainII}$. In addition, define  $\widehat{p}_{\gamma^\ast}(x) := \sum\limits_{k=0}^{m} \gamma_k^\ast  \widehat{p}_k(x)$.

Now, let any $y\in [m]$ be given. By the constraint in our procedure, we have that:
\begin{align*}
	\widehat{\Var}_{\gamma^\ast}^{\trainII}[\widehat{s}_{\gamma^\ast,y}] &= \widehat{\E}_{y}^{\trainII}[\widehat{s}_{\gamma^\ast,y}] - \widehat{\E}_{0}^{\trainII}[\widehat{s}_{\gamma^\ast,y}] \\
	&= \sum_{x\in\DtrainII} \widehat{p}_y(x)\widehat{s}_{\gamma^\ast,y}(x) -  \sum_{x\in\DtrainII} \widehat{p}_0(x)\widehat{s}_{\gamma^\ast,y}(x) \\ 
	&= \sum_{x\in\DtrainII} \big(\widehat{p}_y(x)-\widehat{p}_0(x)\big) \widehat{s}_{\gamma^\ast,y}(x)  \\ 
	&= \sum_{x\in\DtrainII} \widehat{p}_{\gamma^\ast}(x) \frac{\widehat{p}_y(x)-\widehat{p}_0(x)}{\widehat{p}_{\gamma^\ast}(x)} \widehat{s}_{\gamma^\ast,y}(x) \\ 
	&= \widehat{\Cov}_{\gamma^\ast}^{\trainII}\bigg[ \widehat{s}_{\gamma^\ast,y} ,\text{ } \frac{\widehat{p}_y-\widehat{p}_0}{\widehat{p}_{\gamma^\ast}} \bigg] \\
	&\leq \sqrt{\widehat{\Var}_{\gamma^\ast}^{\trainII}[\widehat{s}_{\gamma^\ast,y}] } \sqrt{\widehat{\Var}_{\gamma^\ast}^{\trainII}\bigg[ \frac{\widehat{p}_y-\widehat{p}_0}{\widehat{p}_{\gamma^\ast}} \bigg] }.
\end{align*}
Ergo:
\[
	\implies \widehat{\Var}_{\gamma^\ast}^{\trainII}[\widehat{s}_{\gamma^\ast,y}]  \leq \widehat{\Var}_{\gamma^\ast}^{\trainII}\bigg[ \frac{\widehat{p}_y-\widehat{p}_0}{\widehat{p}_{\gamma^\ast}} \bigg].
\]

Next, let us obtain upper and lower bounds on $\frac{\widehat{p}_y(x)-\widehat{p}_0(x)}{\widehat{p}_{\gamma^\ast}(x)}$, for $x \in \DtrainII$. Let's start with upper bounds. For any $x\in \DtrainII$, we have that
\[
	\frac{\widehat{p}_y(x)-\widehat{p}_0(x)}{\widehat{p}_{\gamma^\ast}(x)} \leq \frac{\widehat{p}_y(x)}{\widehat{p}_{\gamma^\ast}(x)},
\]
where $\widehat{p}_{\gamma^\ast}(x) > 0 $ since $x \in \DtrainII$. Now, if $x \notin \D_y^{\trainII}$, then $\frac{\widehat{p}_y(x)}{\widehat{p}_{\gamma^\ast}(x)} = 0$, whereas if  $x\in\D_y^{\trainII}$, then 
$\frac{\widehat{p}_y(x)}{\widehat{p}_{\gamma^\ast}(x)} \leq \frac{\widehat{p}_y(x)}{ \gamma_y^\ast \widehat{p}_y(x)} = \frac{1}{\gamma_y^\ast}$. Therefore, in general, it follows that 
\[
	\frac{\widehat{p}_y(x)-\widehat{p}_0(x)}{\widehat{p}_{\gamma^\ast}(x)} \leq \frac{1}{\gamma_y^\ast}  \quad \forall x \in \DtrainII.
\] 
Next, let's address the lower bound. For any $x\in \DtrainII$, we have that
\[
	\frac{\widehat{p}_y(x)-\widehat{p}_0(x)}{\widehat{p}_{\gamma^\ast}(x)} \geq -\frac{\widehat{p}_0(x)}{\widehat{p}_{\gamma^\ast}(x)}.
\]
If $x \notin \D_0^{\trainII}$, then $\frac{\widehat{p}_0(x)}{\widehat{p}_{\gamma^\ast}(x)} = 0$, whereas if $x \in \D_0^{\trainII}$, then $-\frac{\widehat{p}_0(x)}{\widehat{p}_{\gamma^\ast}(x)} \geq -\frac{\widehat{p}_0(x)}{\gamma_0^\ast \widehat{p}_0(x)} = - \frac{1}{\gamma_0^\ast}$. Therefore, in general, it follows that:
\[
	\frac{\widehat{p}_y(x)-\widehat{p}_0(x)}{\widehat{p}_{\gamma^\ast}(x)} \geq - \frac{1}{\gamma_0^\ast}.
\]
So, overall, we've shown that
\[
	- \frac{1}{\gamma_0^\ast} \leq \frac{\widehat{p}_y(x)-\widehat{p}_0(x)}{\widehat{p}_{\gamma^\ast}(x)} \leq  \frac{1}{\gamma_y^\ast},
\]
for all $x\in\DtrainII$. So,
\begin{align*}
	 \widehat{\Var}_{\gamma^\ast}^{\trainII}\bigg[ \frac{\widehat{p}_y-\widehat{p}_0}{\widehat{p}_{\gamma^\ast}} \bigg] & \leq \frac{1}{4}\bigg(\frac{1}{\gamma_0^\ast} + \frac{1}{\gamma_y^\ast} \bigg)^2 \\
	 &\leq \frac{1}{4}\bigg( \frac{2}{L}  \bigg)^2 \\ 
	 &=  \frac{1}{L^2}. 
\end{align*}
where the first line is due to Popoviciu's variance inequality, and the second line is due to Assumption \eqref{bounded pis} and Lemma \eqref{Bounded Gamma Star}. By our work earlier in this proof, it follows that 
\[
	\implies \widehat{\Var}_{\gamma^\ast}^{\trainII}[\widehat{s}_{\gamma^\ast,y}]  \leq \frac{1}{L^2}.
\]

Next, consider again the procedure for constructing $\widehat{s}_{\gamma^\ast,y}$. By Assumption \eqref{members of function class}, the identically zero function is included in $\H_y$. Also note that the identically zero function trivially satisfies the moment constraint in the procedure, and that $\Omega(\cdot)$ equals zero when applied to the identically zero function, by Assumption \eqref{Omega Properties}.  Altogether, these observations imply that:
\[
	0 \leq \big(\widehat{\E}_y^{\trainII}[\widehat{s}_{\gamma^\ast,y}] - \widehat{\E}_0^{\trainII}[\widehat{s}_{\gamma^\ast,y}] \big) - \lambda_y \Omega(\widehat{s}_{\gamma^\ast,y})^2
\]
\begin{align*}
	\implies  \Omega(\widehat{s}_{\gamma^\ast,y}) &\leq \sqrt{ \frac{\widehat{\E}_y^{\trainII}[\widehat{s}_{\gamma^\ast,y}] - \widehat{\E}_0^{\trainII}[\widehat{s}_{\gamma^\ast,y}] }{\lambda_y} } \\
	&= \sqrt{ \frac{   \widehat{\Var}_{\gamma^\ast}^{\trainII}[\widehat{s}_{\gamma^\ast,y}]  }{\lambda_y} } \\ 
	&\leq \sqrt{\frac{1}{\lambda_y L^2}} \\ 
	&= \frac{1}{L\sqrt{\lambda_y}},
\end{align*}
where the second line is due to the procedure's moment constraints.

\end{proof}
\vspace{0.4in}

\begin{proof}[\uline{Proof of Lemma \eqref{Bounds on Eigenvalues of Categorical Fisher Info}}] 

It is easy to show that, for $i,j \in [m]$:
\[
	\FisherInfo{\beta;\textup{Cat}}_{i,j} = 
	\begin{cases} 
      		\frac{1}{\beta_i} + \frac{1}{\beta_0} & i=j \\
      		\frac{1}{\beta_0} & i\neq j.
  	 \end{cases}	
\]

Therefore:
\begin{align*}
	\maxEval{\FisherInfo{\beta;\textup{Cat}}} &\leq \max_{i} \sum_{j=1}^{m} \absbig{\FisherInfo{\beta;\textup{Cat}}_{i,j}} \\ 
	&= \max_{i} \Bigg\{   \frac{1}{\beta_i} + \frac{1}{\beta_0}    +    \frac{m-1}{\beta_0} \Bigg\} \\ 
	&= \max_{i} \Bigg\{   \frac{1}{\beta_i} +  \frac{m}{\beta_0} \Bigg\} \\ 
	&=   \frac{1}{\min_i \beta_i} +  \frac{m}{\beta_0},
\end{align*}
as claimed. As for the upper bound, it is easy to show that $\FisherInfo{\beta;\text{Cat}}^{-1}$ is equal to the expression given in Lemma \eqref{Covariance Matrix of a Finite Mixture}. It follows that:
\begin{align*}
	\frac{1}{\minEval{ \FisherInfo{\beta;\text{Cat}} } } &= \maxEval{\FisherInfo{\beta;\text{Cat}}^{-1}} \\ 
	&\leq  \max_{i} \sum_{j=1}^{m} \absbig{\FisherInfo{\beta;\text{Cat}}^{-1}_{i,j}} \\ 
	&=  \max_{i} \beta_i \Bigg\{(1-\beta_i) +  \sum_{j=1\neq i}^{m} \beta_j \Bigg\} \\
	&\leq  2\max_{i} \beta_i (1-\beta_i)  \\
	&\leq  \frac{1}{2},
\end{align*}
meaning that $\minEval{ \FisherInfo{\beta;\text{Cat}} } \geq 2$, as claimed. 
\end{proof}
\vspace{0.4in}

\begin{proof}[\uline{Proof of Lemma \eqref{Bounds on Eigenvalues of Inverse Cat FIM times Cat FIM, alpha vs. beta}}]
One can verify that, for any $i,j \in [m]$:
\[
\big[\FisherInfo{\alpha;\text{Cat}}\FisherInfo{\beta;\text{Cat}}^{-1}\big]_{i,j} = 
 \begin{cases} 
      \beta_j \Big( \frac{\beta_0}{\alpha_0} + \frac{1-\beta_j}{\alpha_j} \Big) & i=j \\ \\ 
      \beta_j \Big( \frac{\beta_0}{\alpha_0} - \frac{\beta_i}{\alpha_i} \Big) & i\neq j.
   \end{cases}
\]

Now, let $(\lambda,u)$ denote an eigenvalue and unit eigenvector pair for the matrix $\FisherInfo{\alpha;\text{Cat}}\FisherInfo{\beta;\text{Cat}}^{-1}$. Then, for each $i\in[m]$, $(\lambda,u)$ satisfies
\begin{align*}
	\lambda u_i &= \sum_{j=1}^{m} \big[\FisherInfo{\alpha;\text{Cat}}\FisherInfo{\beta;\text{Cat}}^{-1}\big]_{i,j} u_j  \\ 
	&= \sum_{j=1\neq i }^{m} \big[\FisherInfo{\alpha;\text{Cat}}\FisherInfo{\beta;\text{Cat}}^{-1}\big]_{i,j} u_j  + \big[\FisherInfo{\alpha;\text{Cat}}\FisherInfo{\beta;\text{Cat}}^{-1}\big]_{i,i} u_i \\ 
	&= \sum_{j=1\neq i }^{m} \beta_j \bigg( \frac{\beta_0}{\alpha_0} - \frac{\beta_i}{\alpha_i} \bigg) u_j  + \beta_i \Big( \frac{\beta_0}{\alpha_0} + \frac{1-\beta_i}{\alpha_i} \Big) u_i \\ 
	&= \sum_{j=1\neq i }^{m} \beta_j \bigg( \frac{\beta_0}{\alpha_0} - \frac{\beta_i}{\alpha_i} \bigg) u_j  + \beta_i \Big( \frac{\beta_0}{\alpha_0} - \frac{\beta_i}{\alpha_i} \Big) u_i + \frac{\beta_i}{\alpha_i} u_i \\
	&= \bigg( \frac{\beta_0}{\alpha_0} - \frac{\beta_i}{\alpha_i} \bigg) \sum_{j=1}^{m} \beta_j u_j  + \frac{\beta_i}{\alpha_i} u_i,
\end{align*}
meaning that:
\[
	\implies \bigg(\lambda  - \frac{\beta_i}{\alpha_i}\bigg) u_i = \bigg( \frac{\beta_0}{\alpha_0} - \frac{\beta_i}{\alpha_i} \bigg) \sum_{j=1}^{m} \beta_j u_j
\]
\begin{align*}
	\implies \absbigg{\lambda  - \frac{\beta_i}{\alpha_i}} \abs{u_i} &\leq \absbigg{ \frac{\beta_0}{\alpha_0} - \frac{\beta_i}{\alpha_i} }  \norm{u}{1} \\ 
	&\leq \sqrt{m}\absbigg{ \frac{\beta_0}{\alpha_0} - \frac{\beta_i}{\alpha_i} },
\end{align*}
where the second inequality is because $\norm{u}{2} = 1$. Now, the inequality above holds for all $i\in[m]$, so let us now pick a specific $\tilde{i}$. In particular, let $\tilde{i}$ denote an index for which $\frac{1}{\sqrt{m}} \leq \abs{u_{\tilde{i}}}$. To see that such an index $\tilde{i}$ always exists, suppose towards a contradiction that $\abs{u_i} < \frac{1}{\sqrt{m}}$ for all $i\in[m]$. Then, $u_i^2 < \frac{1}{m}$ for all $i\in[m]$, and so $\sum_{i=1}^{m}u_i^2 < 1$. This implies that $\norm{u}{2} < 1$, contradicting the fact that $\norm{u}{2} = 1$. Ergo, there must exist at least one $\tilde{i} \in [m]$ such that  $\abs{u_{\tilde{i}}} \geq \frac{1}{\sqrt{m}}$, as claimed. Applying this fact to the display above, it follows that:
\begin{align*}
	\implies  \absbigg{\lambda  - \frac{\beta_{\tilde{i}} }{\alpha_{\tilde{i}}}  } &\leq m \absbigg{ \frac{\beta_0}{\alpha_0} - \frac{\beta_{\tilde{i}} }{\alpha_{\tilde{i}}} } \\ 
	&\leq m \absbigg{ \frac{\beta_0}{\alpha_0} - 1} + m\absbigg{ 1- \frac{\beta_{\tilde{i}} }{\alpha_{\tilde{i}}} } \\ 
	&\leq m \frac{ \absbig{\beta_0-\alpha_0} + \absbig{\alpha_{\tilde{i}} - \beta_{\tilde{i}} } }{\min_k \alpha_k} \\  
	&\leq m \frac{ \norm{\alpha - \beta}{1}  }{\min_k \alpha_k}.
\end{align*}
Thus, we have that:
\[
	\implies - m \frac{ \norm{\alpha - \beta}{1}  }{\min_k \alpha_k}       \leq      \lambda  - \frac{\beta_{\tilde{i}} }{\alpha_{\tilde{i}}}     \leq      m \frac{ \norm{\alpha - \beta}{1}  }{\min_k \alpha_k}   
\]
\[
	\implies  \frac{\beta_{\tilde{i}} }{\alpha_{\tilde{i}}}  - m \frac{ \norm{\alpha - \beta}{1}  }{\min_k \alpha_k}       \leq      \lambda     \leq      \frac{\beta_{\tilde{i}} }{\alpha_{\tilde{i}}}  +  m \frac{ \norm{\alpha - \beta}{1}  }{\min_k \alpha_k}   
\]
\[
	\implies  1+ \bigg(\frac{\beta_{\tilde{i}} }{\alpha_{\tilde{i}}} - 1\bigg)  - m \frac{ \norm{\alpha - \beta}{1}  }{\min_k \alpha_k}       \leq      \lambda     \leq     1+ \bigg( \frac{\beta_{\tilde{i}} }{\alpha_{\tilde{i}}}-1\bigg)  +  m \frac{ \norm{\alpha - \beta}{1}  }{\min_k \alpha_k}   
\]
\[
	\implies  1- \absbigg{\frac{\beta_{\tilde{i}} -\alpha_{\tilde{i}}}{\alpha_{\tilde{i}}} }  - m \frac{ \norm{\alpha - \beta}{1}  }{\min_k \alpha_k}       \leq      \lambda     \leq     1+  \absbigg{ \frac{\beta_{\tilde{i}} -\alpha_{\tilde{i}}}{\alpha_{\tilde{i}}} }  +  m \frac{ \norm{\alpha - \beta}{1}  }{\min_k \alpha_k}   
\]
\[
	\implies  1- \frac{\norm{\alpha-\beta}{1}}{\min_k\alpha_k}   - m \frac{ \norm{\alpha - \beta}{1}  }{\min_k \alpha_k}       \leq      \lambda     \leq     1+   \frac{\norm{\alpha-\beta}{1}}{\min_k\alpha_k}   +  m \frac{ \norm{\alpha - \beta}{1}  }{\min_k \alpha_k}   
\]
\[
	\implies  1- (m+1) \frac{ \norm{\alpha - \beta}{1}  }{\min_k \alpha_k}       \leq      \lambda     \leq     1 +  (m+1)\frac{ \norm{\alpha - \beta}{1}  }{\min_k \alpha_k}.
\]
Since the display above holds for all eigenvalues, we have proven the desideratum.
\end{proof}
\vspace{0.4in}

\begin{proof}[\uline{Proof of Lemma \eqref{Bounds on Test Mixture Eigenvalues}}]

Under Assumption \eqref{min Fisher Info}, we have that:
\begin{align*}
	\sqrt{\Lambda} &\leq \minEval{\FisherInfo{\gamma^\ast} } \\ 
	&= \min_{\norm{c}{2} = 1} \sum_{i=1}^{m}\sum_{j=1}^{m} c_i c_j \E_{\gamma^\ast}[ s_{\gamma^\ast,i}s_{\gamma^\ast,j}] \\ 
	&= \min_{\norm{c}{2} = 1} \sum_{i=1}^{m}\sum_{j=1}^{m} c_i c_j \int p_{\gamma^\ast}(x) \frac{p_i(x) - p_0(x)}{p_{\gamma^\ast}(x)} s_{\gamma^\ast,j}(x)dx  \\ 
	&= \min_{\norm{c}{2} = 1} \sum_{i=1}^{m}\sum_{j=1}^{m} c_i c_j \int p_{\pi^\ast}(x) \frac{p_i(x) - p_0(x)}{p_{\pi^\ast}(x)} s_{\gamma^\ast,j}(x)dx  \\ 
	&= \min_{\norm{c}{2} = 1} \sum_{i=1}^{m}\sum_{j=1}^{m} c_i c_j \Cov_{\pi^\ast}\big[s_{\pi^\ast,i}, s_{\gamma^\ast,j}\big]  \\ 
	&= \min_{\norm{c}{2} = 1} \Cov_{\pi^\ast}\big[c^T s_{\pi^\ast}, c^Ts_{\gamma^\ast}\big]  \\ 
	&\leq \sqrt{\min_{\norm{c}{2} = 1} \Var_{\pi^\ast}\big[c^T s_{\pi^\ast}\big]  \Var_{\pi^\ast}\big[c^Ts_{\gamma^\ast}\big]}  \\ 
	&\leq \sqrt{\min_{\norm{c}{2} = 1} \maxEval{\FisherInfo{\pi^\ast}}  \Var_{\pi^\ast}\big[c^Ts_{\gamma^\ast}\big]}  \\ 
	&=  \sqrt{\maxEval{\FisherInfo{\pi^\ast}}} \sqrt{\minEval{ \Var_{\pi^\ast}[s_{\gamma^\ast}] }} \\ 
	&\leq  \sqrt{\maxEval{\FisherInfo{\pi^\ast;\textup{Cat} }} } \sqrt{\minEval{ \Var_{\pi^\ast}[s_{\gamma^\ast}] }} \\ 
	&\leq  \sqrt{ \frac{1}{\min_y \pi_y^\ast} +  \frac{m}{\pi_0^\ast}   } \sqrt{\minEval{ \Var_{\pi^\ast}[s_{\gamma^\ast}] }} \\ 
	&\leq  \sqrt{ \frac{m+1}{\xi}   } \sqrt{\minEval{ \Var_{\pi^\ast}[s_{\gamma^\ast}] }},
\end{align*}

where the third to last line is because $\FisherInfo{\pi^\ast} \preceq \FisherInfo{\pi^\ast;\text{Cat}}$,  the second to last line is by Lemma \eqref{Bounds on Eigenvalues of Categorical Fisher Info}, and the last line follows from Assumption \eqref{bounded pis}. Thus, it follows that:
\[
	\implies \frac{\Lambda \xi}{m+1} \leq \minEval{ \Var_{\pi^\ast}[s_{\gamma^\ast}] },
\]
as claimed. As for the upper bound on $\maxEval{ \Var_{\pi^\ast}[s_{\gamma^\ast}] }$, observe that:

\begin{align*}
	\maxEval{ \Var_{\pi^\ast}[s_{\gamma^\ast}] } &\leq \sum_{i=1}^{m}\sum_{j=1}^{m}  \absBig{\Cov_{\pi^\ast}[s_{\gamma^\ast,i}\text{ }, \text{ }s_{\gamma^\ast,j}]} \\ 
	&\leq \sum_{i=1}^{m}\sum_{j=1}^{m}  \sqrt{\Var_{\pi^\ast}[s_{\gamma^\ast,i}] \Var_{\pi^\ast}[s_{\gamma^\ast,j} ]}\\ 
	&\leq  \frac{m^2}{L^2},
\end{align*}
where the third inequality uses Popoviciu's variance inequality.

\end{proof}
\vspace{0.4in}

\begin{proof}[\uline{Proof of Lemma \eqref{Bounds On Score Function Differences at Different Parameters}}]
Observe that, for each $j\in[m]$:
\begin{align*}
	s_{\beta,j} - s_{\alpha,j} &=  s_{\alpha,j}\frac{p_\alpha}{p_\beta} -  s_{\alpha,j} \\
	&=  s_{\alpha,j}\bigg( \frac{p_\alpha}{p_\beta} - 1 \bigg)\\
	&=  s_{\alpha,j}\bigg( \frac{p_\alpha-p_\beta}{p_\beta} \bigg)\\
	&=  s_{\alpha,j} \sum_{k=0}^m  (\alpha_k - \beta_k)\frac{p_k }{p_\beta} .
\end{align*}
Thus, we have that
\begin{align*}
	\abs{s_{\beta,j} - s_{\alpha,j}} &\leq \frac{2}{\min\limits_{0\leq k \leq m}\alpha_k }  \sum_{k=0}^m  \abs{\alpha_k - \beta_k}\frac{p_k }{p_\beta}\\
	&\leq \frac{2}{\min\limits_{0\leq k \leq m}\alpha_k}  \sum_{k=0}^m  \abs{\alpha_k - \beta_k}\frac{p_k}{\beta_k p_k}\\
	&\leq \frac{2}{\Big(\min\limits_{0\leq k \leq m}\alpha_k\Big)\Big(\min\limits_{0\leq k \leq m}\beta_k\Big)}  \sum_{k=0}^m \abs{\alpha_k - \beta_k},
\end{align*}
as claimed.

\end{proof}

\begin{proof}[\uline{Proof of Lemma \eqref{Inverse FIM Difference is Bounded}}]
Observe that:
\begin{align*}
	\minSing{\FisherInfo{\gamma^\ast}^{-1} - \FisherInfo{\gamma^\ast;\textup{Cat}}^{-1} } &=  \minSing{  \FisherInfo{\gamma^\ast}^{-1}[\FisherInfo{\gamma^\ast;\textup{Cat}}- \FisherInfo{\gamma^\ast}]\FisherInfo{\gamma^\ast;\textup{Cat}}^{-1} } \\
	&\geq  \minSing{  \FisherInfo{\gamma^\ast}^{-1}} \minSing{\FisherInfo{\gamma^\ast;\textup{Cat}}- \FisherInfo{\gamma^\ast}}  \minSing{\FisherInfo{\gamma^\ast;\textup{Cat}}^{-1}} \\
	&=  \frac{\minSing{\FisherInfo{\gamma^\ast;\textup{Cat}}- \FisherInfo{\gamma^\ast}}}{\maxEval{\FisherInfo{\gamma^\ast}}  \maxEval{\FisherInfo{\gamma^\ast;\textup{Cat}}}  }  \\
	&\geq  \frac{\nu}{\maxEval{\FisherInfo{\gamma^\ast}}  \maxEval{\FisherInfo{\gamma^\ast;\textup{Cat}}}  }  \\
	&\geq \frac{\nu}{ \maxEval{\FisherInfo{\gamma^\ast;\textup{Cat}}}^2  }  \\
	&\geq  \frac{\nu}{ \Big(\frac{1}{\min_y \gamma_y^\ast} + \frac{m}{\gamma_0^\ast}\Big)^2 }  \\
	&\geq  \frac{\nu}{ \Big( \frac{m+1}{L}\Big)^2 }  \\
	&= \nu \bigg( \frac{L}{m+1}\bigg)^2.
\end{align*}
where the fourth line follows from Assumption \eqref{FIM Difference is Bounded},  the fifth line uses Lemma \eqref{Harder Problem}, the sixth line uses Lemma \eqref{Bounds on Eigenvalues of Categorical Fisher Info}, and the seventh line uses Lemma \eqref{Bounded Gamma Star}.
\end{proof}
\vspace{0.4in}

\begin{proof}[\uline{Proof of Corollary \eqref{Relationship between Variances of Different Mixtures}}]

Let $a = \Big[ \E_1[f] - \E_0[f] ,\dots ,  \E_m[f] - \E_0[f]  \Big]$.  By Lemma \eqref{Covariance Matrix of a Finite Mixture}, we have that:
\begin{align*}
	\Var_{\beta}[f] &= \sum_{y=0}^{m} \beta_y \Var_y[f] + a \FisherInfo{\beta;\text{Cat}}^{-1} a^{T} \\ 
	&\geq (\min_y\beta_y) \sum_{y=0}^{m}  \alpha_y \Var_y[f] + \frac{a \FisherInfo{\beta;\text{Cat}}^{-1} a^{T} }{a \FisherInfo{\alpha;\text{Cat}}^{-1} a^{T} } a \FisherInfo{\alpha;\text{Cat}}^{-1} a^{T}  \\ 
	&\geq (\min_y\beta_y) \sum_{y=0}^{m}  \alpha_y \Var_y[f] + \frac{\minEval{\FisherInfo{\beta;\text{Cat}}^{-1}} }{ \maxEval{\FisherInfo{\alpha;\text{Cat}}^{-1}} } a \FisherInfo{\alpha;\text{Cat}}^{-1} a^{T}  \\
	&= (\min_y\beta_y) \sum_{y=0}^{m}  \alpha_y \Var_y[f] + \frac{\minEval{\FisherInfo{\alpha;\text{Cat}}} }{ \maxEval{\FisherInfo{\beta;\text{Cat}}} } a \FisherInfo{\alpha;\text{Cat}}^{-1} a^{T}  \\ 
	&\geq (\min_y\beta_y) \sum_{y=0}^{m}  \alpha_y \Var_y[f] + \frac{2}{\frac{1}{\min_y \beta_y} + \frac{m}{\beta_0}}  a \FisherInfo{\alpha;\text{Cat}}^{-1} a^{T}  \\ 
	&\geq \frac{1}{\frac{1}{\min_y\beta_y} + \frac{m}{\beta_0} } \sum_{y=0}^{m}  \alpha_y \Var_y[f] + \frac{1}{\frac{1}{\min_y \beta_y} + \frac{m}{\beta_0}}  a \FisherInfo{\alpha;\text{Cat}}^{-1} a^{T}  \\ 
	&= \frac{1}{\frac{1}{\min_y\beta_y} + \frac{m}{\beta_0} } \Var_{\alpha}[f],
\end{align*}
where the third inequality is by Lemma \eqref{Bounds on Eigenvalues of Categorical Fisher Info}, and the last line again uses Lemma \eqref{Covariance Matrix of a Finite Mixture}. Thus, it follows that
\[
	\bigg(\frac{1}{\min_y\beta_y} + \frac{m}{\beta_0}  \bigg)\Var_{\beta}[f] \geq \Var_{\alpha}[f],
\]
as claimed

\end{proof}
\vspace{0.4in}

\begin{proof}[\uline{Proof of Lemma \eqref{Harder Problem}}]

Let $(X,Y)$ be random variables, with marginal distribution given by $Y \sim \text{Cat}(\beta)$ and conditional distribution given by $X|Y= y \sim p_{y}$. Let $p_{\text{Cat}(\beta)}(y) := \prod_{i = 1}^{m} \beta_i^{ \I{y=i} }\cdot \Big(1 - \sum_{i=1}^{m} \beta_i \Big)^{\I{y=0}}  $  denote the probability mass function for the $Y$ marginal. Observe that:
\small
\begin{align*}
	  \nabla_{\beta} \log\big[p_{\text{Cat}(\beta)}(y)p_{y}(x)\big] &=  \nabla_{\beta} \log\big[p_{\text{Cat}(\beta)}(y)\big] + \nabla_{\beta} \log\big[p_{y}(x)\big] \\ 
	  &=  \nabla_{\beta} \log\big[p_{\text{Cat}(\beta)}(y)\big],
\end{align*}
\normalsize
so
\small
\begin{align*}
	&\E_{X,Y}\bigg[  \Big(\nabla_{\beta} \log\big[p_{\text{Cat}(\beta)}(Y)p_{Y}(X)\big]\Big) \Big(\nabla_{\beta} \log\big[p_{\text{Cat}(\beta)}(Y)p_{Y}(X)\big]\Big)^T  \bigg] \\
	&= \E_{X,Y}\bigg[ \Big(\nabla_{\beta} \log\big[p_{\text{Cat}(\beta)}(Y)\big]  \Big) \Big(\nabla_{\beta} \log\big[p_{\text{Cat}(\beta)}(Y)\big]  \Big)^T\bigg] \\ 
	&= \E_{Y}\bigg[ \Big(\nabla_{\beta} \log\big[p_{\text{Cat}(\beta)}(Y)\big]  \Big) \Big(\nabla_{\beta} \log\big[p_{\text{Cat}(\beta)}(Y)\big]  \Big)^T   \bigg] \\ 
	&\equiv \FisherInfo{\beta;\text{Cat}}.
\end{align*}
\normalsize
Now, define $p_{\beta}(y|x) := \frac{p_{\text{Cat}(\beta)}(y) p_y(x)}{p_{\beta}(x)}$. By Lemma 1 in \citep{FisherInfoInequality}, we have that 
\small
\begin{align*}
	&\E_{X,Y}\bigg[  \Big(\nabla_{\beta} \log\big[p_{\text{Cat}(\beta)}(Y)p_{Y}(X)\big]\Big) \Big(\nabla_{\beta} \log\big[p_{\text{Cat}(\beta)}(Y)p_{Y}(X)\big]\Big)^T  \bigg] \\
	&= \E_{X}\bigg[  \Big(\nabla_{\beta} \log\big[p_{\beta}(X)\big]\Big) \Big(\nabla_{\beta} \log\big[p_{\beta}(X)\big]\Big)^T  \bigg] \\
	&\qquad + \E_{X,Y}\bigg[ \Big(\nabla_{\beta}\log p_{\beta}(Y|X) \Big)\Big(\nabla_{\beta}\log p_{\beta}(Y|X) \Big)^{T} \bigg] \\ \\
	&\equiv \FisherInfo{\beta} + \E_{X,Y}\bigg[ \Big(\nabla_{\beta}\log p_{\beta}(Y|X) \Big)\Big(\nabla_{\beta}\log p_{\beta}(Y|X) \Big)^{T} \bigg].
\end{align*}
\normalsize
As a quick aside, we note that the proof in Lemma 1 in \citep{FisherInfoInequality} assumes that a certain derivative and integral operator can be exchanged, without rigorous proof. However, for our case, this exchange is justified, because the integral in question amounts to a finite summation over $m+1$ classes. Now, bringing our work in this proof together, it follows that:
\begin{align*}
	\implies \FisherInfo{\beta;\text{Cat}} -  \FisherInfo{\beta} &=  \E_{X,Y}\bigg[ \Big(\nabla_{\beta}\log p_{\beta}(Y|X) \Big)\Big(\nabla_{\beta}\log p_{\beta}(Y|X) \Big)^{T} \bigg].
\end{align*}
Note that the matrix in the RHS above can be rewritten as an expected value of a covariance matrix, 
\[
	\E_{X} \E_{Y|X} \bigg[ \Big(\nabla_{\beta}\log p_{\beta}(Y|X) \Big)\Big(\nabla_{\beta}\log p_{\beta}(Y|X) \Big)^{T} \bigg],
\]
so the aforementioned matrix is positive semidefinite, and thus, $\FisherInfo{\beta;\text{Cat}} \succeq \FisherInfo{\beta}$, as  claimed. Next, it remains for us to demonstrate that $\FisherInfo{\beta;\text{Cat}} =  \FisherInfo{\beta}$ if and only if the supports of the class densities are all disjoint. Towards that end, observe that:
\begin{align*}
	\FisherInfo{\beta;\text{Cat}} =  \FisherInfo{\beta} &\iff \E_{X,Y}\bigg[ \Big(\nabla_{\beta}\log p_{\beta}(Y|X) \Big)\Big(\nabla_{\beta}\log p_{\beta}(Y|X) \Big)^{T} \bigg] = 0 \\ 
	&\iff \nabla_{\beta}\log p_{\beta}(Y|X) = \vec{0} \quad \text{ a.e. } \P_{Y\sim \text{Cat}(\beta)}\otimes\P_{X|Y}  \\ 
	&\iff \nabla_{\beta}\log\Big\{ p_{\text{Cat}(\beta)}(Y) p_Y(X) \Big\} =  \nabla_{\beta}\log\Big\{ p_{\beta}(X) \Big\}  \quad \text{ a.e. } \P_{Y\sim \text{Cat}(\beta)}\otimes\P_{X|Y}  \\ 
	&\iff \nabla_{\beta}\log\Big\{ p_{\text{Cat}(\beta)}(Y)  \Big\} =  s_{\beta}(X)  \quad \text{ a.e. } \P_{Y\sim \text{Cat}(\beta)}\otimes\P_{X|Y}.
\end{align*}
Now, since $\beta_0,\beta_1,\dots,\beta_m > 0$, it follows that: 
\[
	\FisherInfo{\beta;\text{Cat}} =  \FisherInfo{\beta}  \iff  \nabla_{\beta}\log\Big\{ p_{\text{Cat}(\beta)}(y)  \Big\} =  s_{\beta}(X)  \qquad \text{ a.e. } \P_{X|y}, \quad \forall y \in 0\cup[m].
\]
Next, note that for each $j\in[m]$:
\begin{align*}
	\partialDerivative{\beta_j} \log\Big\{ p_{\text{Cat}(\beta)}(y)  \Big\} &= \partialDerivative{\beta_j} \Bigg\{\sum_{i=1}^{m} \I{y=i} \log\beta_i + \I{y=0}\log\bigg(1 - \sum_{i=1}^{m} \beta_i \bigg) \Bigg\} \\
	&= \I{y=j} \partialDerivative{\beta_j} \log\beta_j   +   \I{y=0}\partialDerivative{\beta_j}  \log\bigg(1 - \sum_{i=1}^{m} \beta_i \bigg) \\ 
	&= \frac{\I{y=j}}{\beta_j}  -   \frac{\I{y=0}}{1 - \sum_{i=1}^{m} \beta_i }.
\end{align*}

Thus, we have that, for each $y\in\Y$:
\begin{align*}
	\FisherInfo{\beta;\text{Cat}} =  \FisherInfo{\beta} &\iff  \frac{\I{y=j}}{\beta_j}  -   \frac{\I{y=0}}{1 - \sum_{i=1}^{m} \beta_i } = \frac{p_j(X) - p_0(X)}{p_{\beta}(X)} \quad \forall j\in[m]  \text{ a.e. } \P_{X|y}.
\end{align*}

Now, consider the RHS above, when $y \in[m]$. In this case, we have that:
\[
	\frac{\I{y=j}}{\beta_j}  -   \frac{\I{y=0}}{1 - \sum_{i=1}^{m} \beta_i } = \frac{p_j(X) - p_0(X)}{p_{\beta}(X)} \quad \forall j\in[m] \quad \text{ a.e. } \P_{X|y} 
\]
\[
	\iff \frac{\I{y=j}}{\beta_j}   = \frac{p_j(X) - p_0(X)}{p_{\beta}(X)} \quad \forall j\in[m] \quad \text{ a.e. } \P_{X|y} 
\]
\[
	\iff \I{y=j} p_{\beta}(X)  = \beta_j\big(p_j(X) - p_0(X)\big)  \quad \forall j\in[m] \quad \text{ a.e. } \P_{X|y} 
\]
\[ \iff 
   \begin{cases} 
      p_j(X) = p_0(X) & j\neq y\\
      p_{\beta}(X) = \beta_j\big(p_j(X) - p_0(X)\big)  & j=y
   \end{cases}\qquad \forall j\in[m] \quad  \text{ a.e. } \P_{X|y} 
\]
\[ \iff 
   \begin{cases} 
      p_j(X) = p_0(X) & j\neq y\\
     p_0(X) = -\sum\limits_{k=1\neq j}^{m}\beta_k\big(p_{k}(X) - p_0(X)\big) & j=y
   \end{cases}\qquad \forall j\in[m] \quad  \text{ a.e. } \P_{X|y}.
\]
\[ \iff 
   \begin{cases} 
      p_j(X) = p_0(X) & j\in [m]/y \\
     p_0(X) = -\sum\limits_{k=1\neq y}^{m}\beta_k\big(p_{k}(X) - p_0(X)\big) &
   \end{cases} \qquad  \text{ a.e. } \P_{X|y}.
\]
If the $m$ equalities above  hold simultaneously $\text{ a.e. } \P_{X|y}$, then it follows that $p_0(x) = 0 =  p_j(x)$, for all $ j\in [m]/y$. Conversely, if  $p_0(x) = 0 =  p_j(x)$ for all $ j\in [m]/y$, then the $m$ equalities  displayed above are trivially true. Ergo, the previous display occurs iff
\[
	\iff p_0(x) = 0 = p_j(x) \quad \forall  j\in [m]/y,  \quad  \text{ a.e. } \P_{X|y}.
\]

Now, what about when $y = 0$? In this case:
\[
	\frac{\I{y=j}}{\beta_j}  -   \frac{\I{y=0}}{1 - \sum_{i=1}^{m} \beta_i } = \frac{p_j(X) - p_0(X)}{p_{\beta}(X)} \quad \forall j\in[m] \quad \text{ a.e. } \P_{X|y} 
\]
\[
	\iff   -   \frac{1}{1 - \sum_{i=1}^{m} \beta_i } = \frac{p_j(X) - p_0(X)}{p_{\beta}(X)} \quad \forall j\in[m] \quad \text{ a.e. } \P_{X|0} 
\]
\[
	\iff     p_{\beta}(X) = - \big(p_j(X) - p_0(X)\big) \bigg(1 - \sum_{i=1}^{m} \beta_i\bigg) \quad \forall j\in[m] \quad \text{ a.e. } \P_{X|0} 
\]
\begin{equation}\label{releaseKracken}
	\iff     \bigg(1-\sum_{i=1}^{m}\beta_i \bigg) p_j(X) = -  \sum_{i=1}^{m}\beta_i p_i(x) \qquad \forall  j\in[m] \quad \text{ a.e. } \P_{X|0} 
\end{equation}

Now, if the display above is true, then it follows that $p_1(X) = \dots = p_m(X)$, a.e. $ \P_{X|0} $. As such, it would follow that 
\[
	\implies \Bigg[   \bigg(1-\sum_{i=1}^{m}\beta_i \bigg) +  \sum_{i=1}^{m}\beta_i   \Bigg]p_1(X) = 0 \quad \text{ a.e. } \P_{X|0} 
\]
\[
	\implies p_1(X) = 0  \quad \text{ a.e. } \P_{X|0} 
\]
\[
	\implies p_1(X) = \dots = p_m(X) = 0  \quad \text{ a.e. } \P_{X|0}.
\]

Conversely, if $p_1(X) = \dots = p_m(X) = 0  \quad \text{ a.e. } \P_{X|0}$, then \eqref{releaseKracken} holds trivially. Ergo, overall, we have shown that 
\begin{align*}
	\FisherInfo{\beta;\text{Cat}} =  \FisherInfo{\beta}   &\iff \frac{\I{y=j}}{\beta_j}  -   \frac{\I{y=0}}{1 - \sum_{i=1}^{m} \beta_i } = \frac{p_j(X) - p_0(X)}{p_{\beta}(X)} \quad \forall j\in[m] \quad \text{ a.e. } \P_{X|y},  \forall y \in\Y \\ \\
	&\iff   \begin{cases} 
			p_j(x) = 0 \quad \forall  j\in 0\cup[m] / y &: y\in[m] \\ 
		 	p_j(x) = 0  \quad \forall j \in[m] &: y = 0
	\end{cases}\qquad   \text{ a.e. } \P_{X|y}, \quad \forall y \in \Y,
\end{align*}
which is equivalent to the Lemma's statement.


\end{proof}
\vspace{0.2in}

\begin{proof}[\uline{Proof of Corollary \eqref{First Order Constant When Matching Function is Score Function}}]

First, note that $s_{\gamma^\ast}$ is a $\mathbb{R}^m$-valued function, so $A_{s_{\gamma^\ast}}$ is square. Also note that, by Lemma \eqref{Moment Identity},
\begin{align*}
	A_{s_{\gamma^\ast}} &= \Cov_{\gamma^\ast}[s_{\gamma^\ast},s_{\gamma^\ast}] \\ 
	&= \Var_{\gamma^\ast}[s_{\gamma^\ast}] \\ 
	&\equiv \FisherInfo{\gamma^\ast},
\end{align*}
so $A_{s_{\gamma^\ast}} =  \FisherInfo{\gamma^\ast}$, as claimed. Furthermore, under Assumption \eqref{min Fisher Info}, $\minSing{\FisherInfo{\gamma^\ast}} > \sqrt{\Lambda} > 0$, so $A_{s_{\gamma^\ast}}$ is positive definite. Thus, by applying Corollary \eqref{First Order Constant for Any Matching Function} to $f = s_{\gamma^\ast}$, it follows that
\begin{align*}
	 &\frac{1}{2}\Var_{\allDI}\Big[\FisherInfo{\gamma^\ast}^{-1}(\widehat{\E}_{\pi^\ast}^\testI[s_{\gamma^\ast}] -  \widehat{\E}_{\pi^\ast}^\trainI[s_{\gamma^\ast}]) \Big] \\
	 &= \Bigg[\frac{1}{n^\test}+ \frac{1}{n^\train} \sum_{y=0}^{m}\frac{(\pi_y^\ast)^2}{\pi_y^\train}\Bigg] \Big( \FisherInfo{\gamma^\ast}^{-1}\FisherInfo{\gamma^\ast}(\FisherInfo{\gamma^\ast}^{T})^{-1} - \FisherInfo{\gamma^\ast;\text{Cat}}^{-1}  \Big) + \frac{1}{n^\test} \FisherInfo{\pi^\ast;\textup{Cat}}^{-1} \\ 
	  &= \Bigg[\frac{1}{n^\test}+ \frac{1}{n^\train} \sum_{y=0}^{m}\frac{(\pi_y^\ast)^2}{\pi_y^\train}\Bigg] \Big( \FisherInfo{\gamma^\ast}^{-1} - \FisherInfo{\gamma^\ast;\text{Cat}}^{-1}  \Big) + \frac{1}{n^\test} \FisherInfo{\pi^\ast;\textup{Cat}}^{-1},
\end{align*}
as was claimed! Lastly, we need to show that $\frac{1}{2}\Var_{\allDI}\Big[\FisherInfo{\gamma^\ast}^{-1}(\widehat{\E}_{\pi^\ast}^\testI[s_{\gamma^\ast}] -  \widehat{\E}_{\pi^\ast}^\trainI[s_{\gamma^\ast}]) \Big]$ is positive definite. It suffices to show that the smallest eigenvalue is strictly positive. Towards that end, observe that under Assumption \eqref{bounded pis}, we have that $\gamma_y^\ast > 0$ for each $y\in\Y$ due to Lemma \eqref{Bounded Gamma Star}. Thus, $ \FisherInfo{\gamma^\ast} \preceq \FisherInfo{\gamma^\ast;\text{Cat}}$ by Lemma \eqref{Harder Problem}. So, $ \FisherInfo{\gamma^\ast}^{-1} \succeq \FisherInfo{\gamma^\ast;\text{Cat}}^{-1}$, meaning that the minimum eigenvalue of $\FisherInfo{\gamma^\ast}^{-1} - \FisherInfo{\gamma^\ast;\text{Cat}}^{-1}$ is non-negative. Ergo:
\begin{align*}
	 &\minEvalBigg{\frac{1}{2}\Var_{\allDI}\Big[\FisherInfo{\gamma^\ast}^{-1}(\widehat{\E}_{\pi^\ast}^\testI[s_{\gamma^\ast}] -  \widehat{\E}_{\pi^\ast}^\trainI[s_{\gamma^\ast}]) \Big]} \\
	 &=  \frac{1}{2}\minEvalBigg{\Var_{\allDI}\Big[\FisherInfo{\gamma^\ast}^{-1}(\widehat{\E}_{\pi^\ast}^\testI[s_{\gamma^\ast}] -  \widehat{\E}_{\pi^\ast}^\trainI[s_{\gamma^\ast}]) \Big]} \\ 
	 &\geq  \frac{1}{2n^\test}\minEval{ \FisherInfo{\pi^\ast;\textup{Cat}}^{-1} } \\
	 &=  \frac{1}{2n^\test} \frac{1}{\maxEval{ \FisherInfo{\pi^\ast;\textup{Cat}}} }.
\end{align*}

Now, by Lemma \eqref{Bounds on Eigenvalues of Categorical Fisher Info} and Assumption \eqref{bounded pis}, we have $\maxEval{\FisherInfo{\pi^\ast;\textup{Cat}}} \leq \frac{1}{\min_y \pi_y^\ast} + \frac{m}{\pi_0^\ast} \leq \frac{1}{\xi} + \frac{m}{\xi} = \frac{m+1}{\xi}$.
Thus, 
\begin{align*}
	 \minEvalBigg{\frac{1}{2}\Var_{\allDI}\Big[\FisherInfo{\gamma^\ast}^{-1}(\widehat{\E}_{\pi^\ast}^\testI[s_{\gamma^\ast}] -  \widehat{\E}_{\pi^\ast}^\trainI[s_{\gamma^\ast}]) \Big]} &\geq \frac{1}{2n^\test} \frac{\xi}{m+1},
\end{align*}
where the RHS is strictly greater than $0$ because $\xi$ and $m+1$ are both fixed, positive constants.

\end{proof}
\vspace{0.4in}

\begin{proof}[\uline{Proof of Lemma \eqref{Alternative Expression for First Order Error}}]
Observe that:
\begin{align*}
	&\widehat{\E}_{\pi^\ast}^{\testI}[s_{\gamma^\ast}] - \widehat{\E}_{\pi^\ast}^{\trainI}[s_{\gamma^\ast}] \\
	&=\Big( \widehat{\E}_{\pi^\ast}^{\testI}[s_{\gamma^\ast}] -\E_{\pi^\ast}[s_{\gamma^\ast}] \Big) - \Big( \widehat{\E}_{\pi^\ast}^{\trainI}[s_{\gamma^\ast}] -  \E_{\pi^\ast}[s_{\gamma^\ast}]  \Big) \\
	&=\Big( \widehat{\E}_{\pi^\ast}^{\testI}[s_{\gamma^\ast}] -\E_{\pi^\ast}[s_{\gamma^\ast}] \Big) -  \sum_{j=0}^m \pi_j^\ast \Big(  \widehat{\E}_{j}^{\trainI}[s_{\gamma^\ast}] - \E_{j}[s_{\gamma^\ast}]  \Big) \\
	&= \widehat{\E}_{\pi^\ast}^{\testI}\big[s_{\gamma^\ast} -\E_{\pi^\ast}[s_{\gamma^\ast}] \big]   -  \sum_{j=0}^m \pi_j^\ast   \widehat{\E}_{j}^{\trainI}\big[s_{\gamma^\ast} - \E_{j}[s_{\gamma^\ast}] \big]    \\
	&= \frac{1}{n^\test/2}\sum_{i=1}^{n/2}D_i\big( s_{\gamma^\ast}(X_i) -\E_{\pi^\ast}[s_{\gamma^\ast}]  \big) - \sum_{j=0}^m \pi_j^\ast \Bigg(      \frac{1}{n_j^\train/2} \sum_{i=1}^{n/2}(1-D_i)\I{Y_i=j}\big( s_{\gamma^\ast}(X_i) -\E_{j}[s_{\gamma^\ast}]  \big)     \Bigg) \\ 
	&= 2\sum_{i=1}^{n/2}\frac{D_i}{n^\test} \big( s_{\gamma^\ast}(X_i) -\E_{\pi^\ast}[s_{\gamma^\ast}]  \big) - 2 \sum_{i=1}^{n/2}\sum_{j=0}^m       \frac{\pi_j^\ast}{n_j^\train} (1-D_i)\I{Y_i=j}\big( s_{\gamma^\ast}(X_i) -\E_{j}[s_{\gamma^\ast}]  \big)     \\ 
	&= \frac{2}{n}\sum_{i=1}^{n/2}\frac{D_i}{n^\test/n}\big( s_{\gamma^\ast}(X_i) -\E_{\pi^\ast}[s_{\gamma^\ast}]  \big) - \frac{2}{n} \sum_{i=1}^{n/2}\sum_{j=0}^m       \frac{1-D_i}{n^\train/n} \frac{\pi_j^\ast}{n_j^\train/n^\train} \I{Y_i=j}\big( s_{\gamma^\ast}(X_i) -\E_{j}[s_{\gamma^\ast}]  \big)     \\ 
	&= \frac{2}{n}\sum_{i=1}^{n/2}\Bigg\{\frac{D_i}{n^\test/n}\big( s_{\gamma^\ast}(X_i) -\E_{\pi^\ast}[s_{\gamma^\ast}]  \big) - \sum_{j=0}^m       \frac{1-D_i}{n^\train/n} \frac{\pi_j^\ast}{n_j^\train/n^\train} \I{Y_i=j}\big( s_{\gamma^\ast}(X_i) -\E_{j}[s_{\gamma^\ast}]  \big)\Bigg\}     \\ 
	&= \frac{1}{n/2}\sum_{i=1}^{n/2}\Bigg\{\frac{D_i}{n^\test/n}\big( s_{\gamma^\ast}(X_i) -\E_{\pi^\ast}[s_{\gamma^\ast}]  \big) - \frac{1-D_i}{n^\train/n} \sum_{j=0}^m  \I{Y_i=j} \frac{\pi_j^\ast}{n_j^\train/n^\train} \big( s_{\gamma^\ast}(X_i) -\E_{j}[s_{\gamma^\ast}]  \big)\Bigg\}.
\end{align*}
This implies that
\begin{align*}
	\implies \FisherInfo{\gamma^\ast}^{-1}\big(\widehat{\E}_{\pi^\ast}^{\testI}[s_{\gamma^\ast}] - \widehat{\E}_{\pi^\ast}^{\trainI}[s_{\gamma^\ast}]\big) &= \frac{1}{n/2}\sum_{i=1}^{n/2}\psi^\eff(Z_i),
\end{align*}
as claimed. The derivation for showing the analogous identity involving $ \FisherInfo{\gamma^\ast}^{-1}\big(\widehat{\E}_{\pi^\ast}^{\test}[s_{\gamma^\ast}] - \widehat{\E}_{\pi^\ast}^{\train}[s_{\gamma^\ast}]\big) $ is identical.

\end{proof}
\vspace{0.4in}

\begin{proof}[\uline{Proof of Lemma \eqref{Uniform Deviation Tail Bounds}}]

First, we will address the uniform deviation tail bounds for the difference between the empirical class expectations, and their population level counterparts. Fix $y\in[m]$. Let any $a > 0$ and class $j \in \Y$ be given. By standard arguments using McDiarmid's inequality, we have that:
\small
\begin{align*}
	&\P_{\DtrainII}\Bigg[\sup_{\Omega(h_y) \leq \frac{1}{L \sqrt{\lambda_y}}} \absBig{\widehat{\E}_j^{\trainII}[h_y] - \E_j[h_y]} \geq a\Bigg]  \\
	&\leq  \P_{\DtrainII}\Bigg[\sup_{\Omega(h_y) \leq \frac{1}{L \sqrt{\lambda_y}}} \absBig{\widehat{\E}_j^{\trainII}[h_y] - \E_j[h_y]} -  \E_{\DtrainII}\bigg( \sup_{\Omega(h_y) \leq \frac{1}{L \sqrt{\lambda_y}}} \absBig{\widehat{\E}_j^{\trainII}[h_y] - \E_j[h_y]}\bigg) \geq  \frac{1}{2}a \Bigg] \\ 
	&\qquad +  \IBigg{   \E_{\DtrainII}\bigg( \sup_{\Omega(h_y) \leq \frac{1}{L \sqrt{\lambda_y}}} \absBig{\widehat{\E}_j^{\trainII}[h_y] - \E_j[h_y]} \bigg) \geq  \frac{1}{2}a  } \\ \\ 
	&\leq  \exp\Bigg\{ -2 \frac{a^2/4}{\tfrac{1}{2}\pi_j^\train n^\train\big( \tfrac{2}{\tfrac{1}{2}\pi_j^\train n^\train DL \sqrt{\lambda_y}}  \big)^2 } \Bigg\} +  \IBigg{   \E_{\DtrainII}\bigg( \sup_{\Omega(h_y) \leq \frac{1}{L \sqrt{\lambda_y}}} \absBig{\widehat{\E}_j^{\trainII}[h_y] - \E_j[h_y]} \bigg) \geq  \frac{1}{2}a  } \\ 
	&\leq \exp\Bigg\{ -\frac{D^2 L^2 \xi }{16} \lambda_y n^\train a^2 \Bigg\} +  \IBigg{   \E_{\DtrainII}\bigg( \sup_{\Omega(h_y) \leq \frac{1}{L \sqrt{\lambda_y}}} \absBig{\widehat{\E}_j^{\trainII}[h_y] - \E_j[h_y]} \bigg) \geq  \frac{1}{2}a  },
\end{align*}
\normalsize
where the second inequality is because, under Assumption \eqref{Omega Properties}, each $h_y$ in the supremum satisfies $\norm{h_y}{\infty} \leq \frac{1}{DL \sqrt{\lambda_y}}$, so it can be shown that each of the $\frac{1}{2}\pi_j^\train n^\train$ bounded differences coefficients (for McDiarmid's inequality) equal $\frac{1}{\tfrac{1}{2}\pi_j^\train n^\train}\cdot \frac{2}{DL\sqrt{\lambda_y}}$.  

Next, using a similar argument, we will obtain a tail bound on the uniform deviation between the empirical variance, and its population level counterpart. Since this type of bound is not very common for variances, we will first go through the details of deriving the bounded differences coefficients. Towards that end,  for any real valued function $f$, let $\widetilde{\Var}_{\gamma^\ast}^{\trainII}[f]$ denote the variance estimate that is based on the same data used to compute $\widehat{\Var}_{\gamma^\ast}^{\trainII}[f]$, except for one data point, which has been arbitrarily altered. Further suppose that this data point belonged to class $k$, for some $k \in \Y$. Then:

\begin{align*}
	\Delta & :=  \sup_{\Omega(h_y) \leq  \frac{1}{L\sqrt{\lambda_y}} } \absBig{ \widehat{\Var}_{\gamma^\ast}^{\trainII}[h_y]  -  \Var_{\gamma^\ast}[h_y]} - \sup_{\Omega(h_y) \leq  \frac{1}{L\sqrt{\lambda_y}} } \absBig{ \widetilde{\Var}_{\gamma^\ast}^{\trainII}[h_y]  -  \Var_{\gamma^\ast}[h_y]} \\ 
	&= \sup_{\Omega(h_y) \leq  \frac{1}{L\sqrt{\lambda_y}} } \Bigg\{ \absBig{ \widehat{\Var}_{\gamma^\ast}^{\trainII}[h_y]  -  \Var_{\gamma^\ast}[h_y]}   - \sup_{\Omega(h_y') \leq  \frac{1}{L\sqrt{\lambda_y}} } \absBig{ \widetilde{\Var}_{\gamma^\ast}^{\trainII}[h_y']  -  \Var_{\gamma^\ast}[h_y']}  \Bigg\}  \\
	&\leq \sup_{\Omega(h_y) \leq  \frac{1}{L\sqrt{\lambda_y}} } \Bigg\{ \absBig{ \widehat{\Var}_{\gamma^\ast}^{\trainII}[h_y]  -  \Var_{\gamma^\ast}[h_y]}   -  \absBig{ \widetilde{\Var}_{\gamma^\ast}^{\trainII}[h_y]  -  \Var_{\gamma^\ast}[h_y]}  \Bigg\}  \\
	&\leq \sup_{\Omega(h_y) \leq  \frac{1}{L\sqrt{\lambda_y}} }  \absBig{ \widehat{\Var}_{\gamma^\ast}^{\trainII}[h_y]  -  \widetilde{\Var}_{\gamma^\ast}^{\trainII}[h_y]}     \\
	&= \sup_{g_y = h_y - \widetilde{\E}_{\gamma^\ast}^{\trainII}[h_y], \Omega(h_y) \leq  \frac{1}{L\sqrt{\lambda_y}} }  \absBig{ \widehat{\Var}_{\gamma^\ast}^{\trainII}[g_y]  -  \widetilde{\Var}_{\gamma^\ast}^{\trainII}[g_y]}     \\
	&= \sup_{g_y = h_y - \widetilde{\E}_{\gamma^\ast}^{\trainII}[h_y], \Omega(h_y) \leq  \frac{1}{L\sqrt{\lambda_y}} }  \absBig{ \widehat{\E}_{\gamma^\ast}^{\trainII}[g_y^2] -  \big( \widehat{\E}_{\gamma^\ast}^{\trainII}[g_y] \big)^2    -  \widetilde{\E}_{\gamma^\ast}^{\trainII}[g_y^2]}     \\
	&\leq \sup_{g_y = h_y - \widetilde{\E}_{\gamma^\ast}^{\trainII}[h_y], \Omega(h_y) \leq  \frac{1}{L\sqrt{\lambda_y}} }  \absBig{ \widehat{\E}_{\gamma^\ast}^{\trainII}[g_y^2]  -  \widetilde{\E}_{\gamma^\ast}^{\trainII}[g_y^2]}   +   \Bigg( \sup_{g_y = h_y - \widetilde{\E}_{\gamma^\ast}^{\trainII}[h_y], \Omega(h_y) \leq  \frac{1}{L\sqrt{\lambda_y}}} \absBig{  \widehat{\E}_{\gamma^\ast}^{\trainII}[g_y] }   \Bigg)^2 \\
	&= \sup_{g_y = h_y - \widetilde{\E}_{\gamma^\ast}^{\trainII}[h_y], \Omega(h_y) \leq  \frac{1}{L\sqrt{\lambda_y}} }  \absBig{ \widehat{\E}_{\gamma^\ast}^{\trainII}[g_y^2]  -  \widetilde{\E}_{\gamma^\ast}^{\trainII}[g_y^2]}   +   \Bigg( \sup_{\Omega(h_y) \leq  \frac{1}{L\sqrt{\lambda_y}}} \absBig{  \widehat{\E}_{\gamma^\ast}^{\trainII}[h_y] - \widetilde{\E}_{\gamma^\ast}^{\trainII}[h_y] }   \Bigg)^2 \\
	&= \pi_k^\train \sup_{g_y = h_y - \widetilde{\E}_{\gamma^\ast}^{\trainII}[h_y], \Omega(h_y) \leq  \frac{1}{L\sqrt{\lambda_y}} }  \absBig{ \widehat{\E}_{k}^{\trainII}[g_y^2]  -  \widetilde{\E}_{k}^{\trainII}[g_y^2]}   +   \Bigg( \pi_k^\train \sup_{\Omega(h_y) \leq  \frac{1}{L\sqrt{\lambda_y}}} \absBig{  \widehat{\E}_{k}^{\trainII}[h_y] - \widetilde{\E}_{k}^{\trainII}[h_y] }   \Bigg)^2.
\end{align*}

Now, if $g_y= h_y - \widetilde{\E}_{\gamma^\ast}[h_y]$ where   $h_y$ satisfies $\Omega(h_y) \leq  \frac{1}{L\sqrt{\lambda_y}}$, then by Assumption \eqref{Omega Properties}, it follows that
\[
	\implies \norm{h_y}{\infty} \leq \frac{1}{DL\sqrt{\lambda_y}} \implies \norm{g_y}{\infty} \leq  \frac{2}{DL\sqrt{\lambda_y}} \implies  \norm{g_y^2}{\infty} \leq  \frac{4}{D^2L^2\lambda_y}.
\]
Next, notice that the operators $\widehat{\E}_{k}^{\trainII}$ and $\widetilde{\E}_{k}^{\trainII}$ are computed on datasets that only differ by one data point. So, it follows that:
\begin{align*}
	\implies \Delta &\leq \frac{\pi_k^\train}{\frac{1}{2}\pi_k^\train n^\train} \frac{8}{D^2L^2\lambda_y}   +   \Bigg( \frac{\pi_k^\train}{\frac{1}{2}\pi_k^\train n^\train} \frac{2}{DL \sqrt{\lambda_y}}   \Bigg)^2 \\ 
	&= \frac{1}{\frac{1}{2} n^\train} \frac{8}{D^2L^2\lambda_y}   +    \frac{1}{\big(\frac{1}{2} n^\train\big)^2} \frac{4}{D^2 L^2 \lambda_y}    \\
	&\leq \frac{1}{\frac{1}{2} n^\train} \frac{8}{D^2L^2\lambda_y}   +    \frac{1}{\frac{1}{2} n^\train} \frac{8}{D^2 L^2 \lambda_y}    \\ 
	&= \frac{1}{\frac{1}{2} n^\train} \frac{16}{D^2L^2\lambda_y} .
\end{align*}

Thus, by McDiarmid's inequality, it follows that:
\small
\begin{align*}
	&\P_{\DtrainII}\Bigg[\sup_{\Omega(h_y) \leq  \frac{1}{L\sqrt{\lambda_y}} } \absBig{ \widehat{\Var}_{\gamma^\ast}^{\trainII}[h_y]  -  \Var_{\gamma^\ast}[h_y]} - \E_{\DtrainII}\Bigg(\sup_{\Omega(h_y) \leq  \frac{1}{L\sqrt{\lambda_y}} } \absBig{ \widehat{\Var}_{\gamma^\ast}^{\trainII}[h_y]  -  \Var_{\gamma^\ast}[h_y]}\Bigg)  \geq \frac{1}{2}a \Bigg]\\
	 &\leq \exp\Bigg\{-2 \frac{ a^2/4 }{ \frac{1}{2}n^\train \Delta^2 } \Bigg\} \\ 
	&= \exp\Bigg\{-2 \frac{ a^2/4 }{ \frac{1}{2}n^\train \Big( \frac{1}{\frac{1}{2} n^\train} \frac{16}{D^2L^2\lambda_y} \Big)^2 } \Bigg\} \\ 
	&= \exp\Bigg\{- \frac{D^4L^4}{1024}\lambda_y^2 n^\train a^2 \Bigg\}.
\end{align*}
\normalsize

Ergo:
\small
\begin{align*}
	 &\P_{\DtrainII}\Bigg[\sup_{\Omega(h_y) \leq  \frac{1}{L\sqrt{\lambda_y}} } \absBig{ \widehat{\Var}_{\gamma^\ast}^{\trainII}[h_y]  -  \Var_{\gamma^\ast}[h_y]}   \geq a \Bigg] \\
	 &\leq \exp\Bigg\{- \frac{D^4L^4}{1024}\lambda_y^2 n^\train a^2 \Bigg\} \\
	&\qquad+ \IBigg{\E_{\DtrainII}\Bigg(\sup_{\Omega(h_y) \leq  \frac{1}{L\sqrt{\lambda_y}} } \absBig{ \widehat{\Var}_{\gamma^\ast}^{\trainII}[h_y]  -  \Var_{\gamma^\ast}[h_y]}\Bigg) \geq \frac{1}{2}a }.
\end{align*}
\normalsize

\end{proof}
\vspace{0.4in}

\begin{proof}[\uline{Proof of Lemma \eqref{Rate for Learning Diagonal of Fisher Information Matrix}}]

We'll begin by making use of the following decomposition:
\[
	\big(\E_y[s_{\gamma^\ast,y}] - \E_0[s_{\gamma^\ast,y}]\big)  -  \big(\E_y[\widehat{s}_{\gamma^\ast,y}]-\E_0[\widehat{s}_{\gamma^\ast,y}]\big) = A + B,
\]
where 
\[
	A = \big(\E_y[s_{\gamma^\ast,y}] - \E_0[s_{\gamma^\ast,y}]\big)     -    \Big[  \big(\widehat{\E}_y^{\trainII}[\widehat{s}_{\gamma^\ast,y}] - \widehat{\E}_0^{\trainII}[\widehat{s}_{\gamma^\ast,y}] \big) - \lambda_y \Omega^2(\widehat{s}_{\gamma^\ast,y}) \Big]
\]
\[
	B = \Big[  \big(\widehat{\E}_y^{\trainII}[\widehat{s}_{\gamma^\ast,y}] - \widehat{\E}_0^{\trainII}[\widehat{s}_{\gamma^\ast,y}] \big) - \lambda_y \Omega^2(\widehat{s}_{\gamma^\ast,y}) \Big] -  \big(\E_y[\widehat{s}_{\gamma^\ast,y}]-\E_0[\widehat{s}_{\gamma^\ast,y}]\big).
\]
\vspace{0.1in}

Next, we may rewrite and bound each of (A) and (B), individually. For term (A), note that:
\small
\begin{align}
	A &= \big(\E_y[s_{\gamma^\ast,y}] - \E_0[s_{\gamma^\ast,y}]\big)   -    \Big[  \big(\widehat{\E}_y^{\trainII}[\widehat{s}_{\gamma^\ast,y}] - \widehat{\E}_0^{\trainII}[\widehat{s}_{\gamma^\ast,y}] \big) - \lambda_y \Omega^2(\widehat{s}_{\gamma^\ast,y}) \Big] \nonumber \\ \nonumber \\
	&\leq  \big(\E_y[s_{\gamma^\ast,y}] - \E_0[s_{\gamma^\ast,y}]\big) \nonumber \\
	&\qquad -    \Bigg[   \frac{\widehat{\E}_y^{\trainII}[s_{\gamma^\ast,y}] - \widehat{\E}_0^{\trainII}[s_{\gamma^\ast,y}]  }{\widehat{\Var}_{\gamma^\ast}^{\trainII}[s_{\gamma^\ast,y}] } \big(\widehat{\E}_y^{\trainII}[s_{\gamma^\ast,y}] - \widehat{\E}_0^{\trainII}[s_{\gamma^\ast,y}] \big) - \lambda_y \Omega^2\bigg( \frac{\widehat{\E}_y^{\trainII}[s_{\gamma^\ast,y}] - \widehat{\E}_0^{\trainII}[s_{\gamma^\ast,y}]  }{\widehat{\Var}_{\gamma^\ast}^{\trainII}[s_{\gamma^\ast,y}] } s_{\gamma^\ast,y}\bigg) \Bigg]\nonumber \\ \nonumber \\
	&=  \big(\E_y[s_{\gamma^\ast,y}] - \E_0[s_{\gamma^\ast,y}]\big)\nonumber \\
	&\qquad -       \frac{\widehat{\E}_y^{\trainII}[s_{\gamma^\ast,y}] - \widehat{\E}_0^{\trainII}[s_{\gamma^\ast,y}]  }{\widehat{\Var}_{\gamma^\ast}^{\trainII}[s_{\gamma^\ast,y}] } \big(\widehat{\E}_y^{\trainII}[s_{\gamma^\ast,y}] - \widehat{\E}_0^{\trainII}[s_{\gamma^\ast,y}] \big)  + \lambda_y \Bigg(\frac{\widehat{\E}_y^{\trainII}[s_{\gamma^\ast,y}] - \widehat{\E}_0^{\trainII}[s_{\gamma^\ast,y}]  }{\widehat{\Var}_{\gamma^\ast}^{\trainII}[s_{\gamma^\ast,y}] } \Bigg)^2\Omega^2(s_{\gamma^\ast,y}) \nonumber \\ \nonumber \\ 
	&=  \big(\E_y[s_{\gamma^\ast,y}] - \E_0[s_{\gamma^\ast,y}]\big) - \big(\widehat{\E}_y^{\trainII}[s_{\gamma^\ast,y}] - \widehat{\E}_0^{\trainII}[s_{\gamma^\ast,y}] \big)  \nonumber \\
	&\qquad + \Bigg(1    -     \frac{\widehat{\E}_y^{\trainII}[s_{\gamma^\ast,y}] - \widehat{\E}_0^{\trainII}[s_{\gamma^\ast,y}]  }{\widehat{\Var}_{\gamma^\ast}^{\trainII}[s_{\gamma^\ast,y}] }\Bigg) \big(\widehat{\E}_y^{\trainII}[s_{\gamma^\ast,y}] - \widehat{\E}_0^{\trainII}[s_{\gamma^\ast,y}] \big) \nonumber  \\
	&\qquad   + \lambda_y \Bigg(\frac{\widehat{\E}_y^{\trainII}[s_{\gamma^\ast,y}] - \widehat{\E}_0^{\trainII}[s_{\gamma^\ast,y}]  }{\widehat{\Var}_{\gamma^\ast}^{\trainII}[s_{\gamma^\ast,y}] } \Bigg)^2\Omega^2(s_{\gamma^\ast,y})  \nonumber \\ \nonumber \\ 
	&\leq   \big(\E_y[s_{\gamma^\ast,y}] - \E_0[s_{\gamma^\ast,y}]\big) - \big(\widehat{\E}_y^{\trainII}[s_{\gamma^\ast,y}] - \widehat{\E}_0^{\trainII}[s_{\gamma^\ast,y}] \big) \nonumber \\
	&+ \absBigg{ \Bigg(1    -     \frac{\widehat{\E}_y^{\trainII}[s_{\gamma^\ast,y}] - \widehat{\E}_0^{\trainII}[s_{\gamma^\ast,y}]  }{\widehat{\Var}_{\gamma^\ast}^{\trainII}[s_{\gamma^\ast,y}] }\Bigg) \big(\widehat{\E}_y^{\trainII}[s_{\gamma^\ast,y}] - \widehat{\E}_0^{\trainII}[s_{\gamma^\ast,y}] \big)  } \nonumber \\
	&\qquad   + 2\lambda_y \Omega^2(s_{\gamma^\ast,y})   + 2\Bigg(\frac{\widehat{\E}_y^{\trainII}[s_{\gamma^\ast,y}] - \widehat{\E}_0^{\trainII}[s_{\gamma^\ast,y}]  }{\widehat{\Var}_{\gamma^\ast}^{\trainII}[s_{\gamma^\ast,y}] } - 1\Bigg)^2 \lambda_y \Omega^2(s_{\gamma^\ast,y}) \nonumber \\ \nonumber \\ 
	&\leq \absBig{ \big(\E_y[s_{\gamma^\ast,y}] - \E_0[s_{\gamma^\ast,y}]\big) - \big(\widehat{\E}_y^{\trainII}[s_{\gamma^\ast,y}] - \widehat{\E}_0^{\trainII}[s_{\gamma^\ast,y}] \big) }  + \frac{4}{L} \absBigg{ \frac{\widehat{\E}_y^{\trainII}[s_{\gamma^\ast,y}] - \widehat{\E}_0^{\trainII}[s_{\gamma^\ast,y}]  }{\widehat{\Var}_{\gamma^\ast}^{\trainII}[s_{\gamma^\ast,y}] } - 1}  \nonumber \\
	&\qquad   + 2\lambda_y \Omega^2(s_{\gamma^\ast,y})   + 2\Bigg(\frac{\widehat{\E}_y^{\trainII}[s_{\gamma^\ast,y}] - \widehat{\E}_0^{\trainII}[s_{\gamma^\ast,y}]  }{\widehat{\Var}_{\gamma^\ast}^{\trainII}[s_{\gamma^\ast,y}] } - 1\Bigg)^2 \lambda_y \Omega^2(s_{\gamma^\ast,y}), \nonumber 
\end{align}
\normalsize
where the second line is due to Assumption \eqref{members of function class} and the fact that $\frac{\widehat{\E}_y^{\trainII}[s_{\gamma^\ast,y}] - \widehat{\E}_0^{\trainII}[s_{\gamma^\ast,y}]  }{\widehat{\Var}_{\gamma^\ast}^{\trainII}[s_{\gamma^\ast,y}] } s_{\gamma^\ast,y}$ is inside the feasible set of the optimization problem,  the third line is due to $\Omega(\cdot)$'s assumed properties, and the sixth line uses Lemma \eqref{Bounded Gamma Star} and  the fact that $\abs{s_{\gamma^\ast,y}} \leq \frac{1}{\gamma_0^\ast} + \frac{1}{\gamma_y^\ast} \leq  \frac{2}{L}$. As for term (B), note that:
\begin{align}
	B &=  \Big[  \big(\widehat{\E}_y^{\trainII}[\widehat{s}_{\gamma^\ast,y}] - \widehat{\E}_0^{\trainII}[\widehat{s}_{\gamma^\ast,y}] \big) - \lambda_y \Omega^2(\widehat{s}_{\gamma^\ast,y}) \Big] -  \big(\E_y[\widehat{s}_{\gamma^\ast,y}]-\E_0[\widehat{s}_{\gamma^\ast,y}]\big) \nonumber \\ 
	 &\leq \sup_{\Omega(h_y) \leq \frac{1}{L \sqrt{\lambda_y}}} \absBig{\widehat{\E}_y^{\trainII}[h_y] - \E_y[h_y]}    +      \sup_{\Omega(h_y) \leq \frac{1}{L \sqrt{\lambda_y}}} \absBig{\widehat{\E}_0^{\trainII}[h_y] - \E_0[h_y]} - \lambda_y \Omega^2(\widehat{s}_{\gamma^\ast,y}), \nonumber 
\end{align}
where the second line is because $\Omega(\widehat{s}_{\gamma^\ast,y}) \leq \frac{1}{L \sqrt{\lambda_y}}$ due to Lemma \eqref{Bound on Norm of Score Estimate}. Therefore:
\begin{align*}
	\big(\E_y[s_{\gamma^\ast,y}] - \E_0[s_{\gamma^\ast,y}]\big)  -  \big(\E_y[\widehat{s}_{\gamma^\ast,y}]-\E_0[\widehat{s}_{\gamma^\ast,y}]\big)  &= A + B \\ 
	&\leq R_y - \lambda_y \Omega^2(\widehat{s}_{\gamma^\ast,y}),
\end{align*}
where $R_y$ is given by 
\begin{align}
	R_y &:= \absBig{\big(\E_y[s_{\gamma^\ast,y}] - \E_0[s_{\gamma^\ast,y}]\big) - \big(\widehat{\E}_y^{\trainII}[s_{\gamma^\ast,y}] - \widehat{\E}_0^{\trainII}[s_{\gamma^\ast,y}] \big) }  + \frac{4}{L} \absBigg{ \frac{\widehat{\E}_y^{\trainII}[s_{\gamma^\ast,y}] - \widehat{\E}_0^{\trainII}[s_{\gamma^\ast,y}]  }{\widehat{\Var}_{\gamma^\ast}^{\trainII}[s_{\gamma^\ast,y}] } - 1} \nonumber \\ 
	&\qquad   + 2\lambda_y \Omega^2(s_{\gamma^\ast,y})   + 2\Bigg( \frac{\widehat{\E}_y^{\trainII}[s_{\gamma^\ast,y}] - \widehat{\E}_0^{\trainII}[s_{\gamma^\ast,y}]  }{\widehat{\Var}_{\gamma^\ast}^{\trainII}[s_{\gamma^\ast,y}] } - 1\Bigg)^2 \lambda_y \Omega^2(s_{\gamma^\ast,y}) \nonumber \\
	&\qquad + \sup_{\Omega(h_y) \leq \frac{1}{L \sqrt{\lambda_y}}} \absBig{\widehat{\E}_y^{\trainII}[h_y] - \E_y[h_y]}    +      \sup_{\Omega(h_y) \leq \frac{1}{L \sqrt{\lambda_y}}} \absBig{\widehat{\E}_0^{\trainII}[h_y] - \E_0[h_y]}. \label{Definition of Ry}  
\end{align}
Now, it is important to note that there is another upper bound on $\big(\E_y[s_{\gamma^\ast,y}] - \E_0[s_{\gamma^\ast,y}]\big)  -  \big(\E_y[\widehat{s}_{\gamma^\ast,y}]-\E_0[\widehat{s}_{\gamma^\ast,y}]\big) $, albeit a trivial one. By Assumption  \eqref{Omega Properties}, we have that $\norm{\widehat{s}_{\gamma^\ast,y}}{\infty} \leq \frac{1}{D} \Omega(\widehat{s}_{\gamma^\ast,y})$, and by Lemma \eqref{Bound on Norm of Score Estimate}, we have that $\Omega(\widehat{s}_{\gamma^\ast,y}) \leq \frac{1}{L\sqrt{\lambda_y}}$. This implies that  $\norm{\widehat{s}_{\gamma^\ast,y}}{\infty}\leq  \frac{1}{DL\sqrt{\lambda_y}}$. We also have that $\norm{s_{\gamma^\ast,y}}{\infty} \leq \frac{2}{L}$. Thus:
\begin{align*}
	 \big(\E_y[s_{\gamma^\ast,y}] - \E_0[s_{\gamma^\ast,y}]\big)  -  \big(\E_y[\widehat{s}_{\gamma^\ast,y}]-\E_0[\widehat{s}_{\gamma^\ast,y}]\big) &\leq \frac{4}{L} + \frac{2}{DL\sqrt{\lambda_y}}.
\end{align*}
So, combining our two upper bounds together, it follows that:
\begin{align}
	&\big(\E_y[s_{\gamma^\ast,y}] - \E_0[s_{\gamma^\ast,y}]\big)  -  \big(\E_y[\widehat{s}_{\gamma^\ast,y}]-\E_0[\widehat{s}_{\gamma^\ast,y}]\big)\\
	 &\leq \min\Bigg\{ R_y - \lambda_y \Omega^2(\widehat{s}_{\gamma^\ast,y})   \text{ },\text{ }   \frac{4}{L} + \frac{2}{DL\sqrt{\lambda_y}}   \Bigg\} \nonumber \\ 
	&= \min\Bigg\{ R_y    \text{ },\text{ }   \frac{4}{L} + \frac{2}{DL\sqrt{\lambda_y}}  + \lambda_y \Omega^2(\widehat{s}_{\gamma^\ast,y})   \Bigg\}  - \lambda_y \Omega^2(\widehat{s}_{\gamma^\ast,y}) \nonumber \\ 
	&\leq \min\Bigg\{ R_y    \text{ },\text{ }   \frac{4}{L} + \frac{2}{DL\sqrt{\lambda_y}}  + \frac{1}{L^2}  \Bigg\}  - \lambda_y \Omega^2(\widehat{s}_{\gamma^\ast,y}) \nonumber \\ 
	&\leq \min\Bigg\{ R_y    \text{ },\text{ }   \frac{5}{L^2} + \frac{2}{DL\sqrt{\lambda_y}}   \Bigg\}  - \lambda_y \Omega^2(\widehat{s}_{\gamma^\ast,y}) \nonumber \\ 
	&= T_y - \lambda_y \Omega^2(\widehat{s}_{\gamma^\ast,y}) \label{Ty - lambda times norm} \\ 
	&\leq T_y, \nonumber
\end{align}
where the third line is by Lemma \eqref{Bound on Norm of Score Estimate}, the fourth line is because $0 < L < 1$, and 
\begin{equation} \label{Definition of Ty}
	T_y := \min\Bigg\{ R_y    \text{ },\text{ }   \frac{5}{L^2} + \frac{2}{DL\sqrt{\lambda_y}}   \Bigg\}.
\end{equation}

\textbf{\uline{Next, let's analyze the asymptotic behavior of $\E_{\DtrainII}[T_y]$}}. Observe that:
\begin{align*}
	&\E_{\DtrainII}[T_y] \\
	&= \int_{0}^{\infty}\P_{\DtrainII}[ T_y \geq \epsilon] d\epsilon \\ 
	 &= \int_{0}^{\frac{5}{L^2} + \frac{2}{DL\sqrt{\lambda_y}}}\P_{\DtrainII}[ T_y \geq \epsilon] d\epsilon \\
	 &\leq \int_{0}^{\frac{5}{L^2} + \frac{2}{DL\sqrt{\lambda_y}}}\P_{\DtrainII}[ R_y \geq \epsilon] d\epsilon \\ \\ 
	 &\leq \int_{0}^{\frac{5}{L^2} + \frac{2}{DL\sqrt{\lambda_y}}}\P_{\DtrainII}\bigg[ \absBig{\big(\E_y[s_{\gamma^\ast,y}] - \E_0[s_{\gamma^\ast,y}]\big) - \big(\widehat{\E}_y^{\trainII}[s_{\gamma^\ast,y}] - \widehat{\E}_0^{\trainII}[s_{\gamma^\ast,y}] \big) } \geq \frac{1}{6}\epsilon\bigg] d\epsilon \\
	 &\qquad +  \int_{0}^{\frac{5}{L^2} + \frac{2}{DL\sqrt{\lambda_y}}}\P_{\DtrainII}\Bigg[ \absBigg{ \frac{\widehat{\E}_y^{\trainII}[s_{\gamma^\ast,y}] - \widehat{\E}_0^{\trainII}[s_{\gamma^\ast,y}]  }{\widehat{\Var}_{\gamma^\ast}^{\trainII}[s_{\gamma^\ast,y}] } - 1}  \geq \frac{L}{24}\epsilon\Bigg] d\epsilon \\
	 &\qquad +  \int_{0}^{\frac{5}{L^2} + \frac{2}{DL\sqrt{\lambda_y}}}\P_{\DtrainII}\Big[ 12\lambda_y \Omega^2(s_{\gamma^\ast,y})  \geq \epsilon\Big] d\epsilon \\
	 &\qquad +  \int_{0}^{\frac{5}{L^2} + \frac{2}{DL\sqrt{\lambda_y}}}\P_{\DtrainII}\Bigg[  \Bigg( \frac{\widehat{\E}_y^{\trainII}[s_{\gamma^\ast,y}] - \widehat{\E}_0^{\trainII}[s_{\gamma^\ast,y}]  }{\widehat{\Var}_{\gamma^\ast}^{\trainII}[s_{\gamma^\ast,y}] } - 1\Bigg)^2 \geq \frac{1}{12  \lambda_y \Omega^2(s_{\gamma^\ast,y})}\epsilon\Bigg] d\epsilon \\
	 &\qquad +  \int_{0}^{\frac{5}{L^2} + \frac{2}{DL\sqrt{\lambda_y}}}\P_{\DtrainII}\Bigg[6\sup_{\Omega(h_y) \leq \frac{1}{L \sqrt{\lambda_y}}} \absBig{\widehat{\E}_y^{\trainII}[h_y] - \E_y[h_y]}  \geq \epsilon\Bigg] d\epsilon \\
	 &\qquad +  \int_{0}^{\frac{5}{L^2} + \frac{2}{DL\sqrt{\lambda_y}}}\P_{\DtrainII}\Bigg[6\sup_{\Omega(h_y) \leq \frac{1}{L \sqrt{\lambda_y}}} \absBig{\widehat{\E}_0^{\trainII}[h_y] - \E_0[h_y]} \geq \epsilon\Bigg] d\epsilon \\ \\ 
	 &\leq \int_{0}^{\infty}\P_{\DtrainII}\bigg[ 6\absBig{\big(\E_y[s_{\gamma^\ast,y}] - \E_0[s_{\gamma^\ast,y}]\big) - \big(\widehat{\E}_y^{\trainII}[s_{\gamma^\ast,y}] - \widehat{\E}_0^{\trainII}[s_{\gamma^\ast,y}] \big) } \geq \epsilon\bigg] d\epsilon \\
	 &\qquad +  \int_{0}^{\frac{5}{L^2} + \frac{2}{DL\sqrt{\lambda_y}}}\P_{\DtrainII}\Bigg[ \absBigg{ \frac{\widehat{\E}_y^{\trainII}[s_{\gamma^\ast,y}] - \widehat{\E}_0^{\trainII}[s_{\gamma^\ast,y}]  }{\widehat{\Var}_{\gamma^\ast}^{\trainII}[s_{\gamma^\ast,y}] } - 1}  \geq \frac{L}{24}\epsilon\Bigg] d\epsilon \\
	 &\qquad +  12\lambda_y \Omega^2(s_{\gamma^\ast,y}) \\
	 &\qquad +  \int_{0}^{\frac{5}{L^2} + \frac{2}{DL\sqrt{\lambda_y}}}\P_{\DtrainII}\Bigg[  \absBigg{\frac{\widehat{\E}_y^{\trainII}[s_{\gamma^\ast,y}] - \widehat{\E}_0^{\trainII}[s_{\gamma^\ast,y}]  }{\widehat{\Var}_{\gamma^\ast}^{\trainII}[s_{\gamma^\ast,y}] } - 1} \geq \frac{1}{\sqrt{12  \lambda_y} \Omega(s_{\gamma^\ast,y})}\sqrt{\epsilon}\Bigg] d\epsilon \\
	 &\qquad +  \int_{0}^{\infty}\P_{\DtrainII}\Bigg[6\sup_{\Omega(h_y) \leq \frac{1}{L \sqrt{\lambda_y}}} \absBig{\widehat{\E}_y^{\trainII}[h_y] - \E_y[h_y]}  \geq \epsilon\Bigg] d\epsilon \\
	 &\qquad +  \int_{0}^{\infty}\P_{\DtrainII}\Bigg[6\sup_{\Omega(h_y) \leq \frac{1}{L \sqrt{\lambda_y}}} \absBig{\widehat{\E}_0^{\trainII}[h_y] - \E_0[h_y]} \geq \epsilon\Bigg] d\epsilon \\ \\ 
	 &\leq  6 \sqrt{\Var_{\DtrainII}\Big[ \widehat{\E}_y^{\trainII}[s_{\gamma^\ast,y}] - \widehat{\E}_0^{\trainII}[s_{\gamma^\ast,y}] \Big]} \\
	 &\qquad +  \int_{0}^{\frac{5}{L^2} + \frac{2}{DL\sqrt{\lambda_y}}}\P_{\DtrainII}\Bigg[ \absBigg{ \frac{\widehat{\E}_y^{\trainII}[s_{\gamma^\ast,y}] - \widehat{\E}_0^{\trainII}[s_{\gamma^\ast,y}]  }{\widehat{\Var}_{\gamma^\ast}^{\trainII}[s_{\gamma^\ast,y}] } - 1}  \geq \frac{L}{24}\epsilon\Bigg] d\epsilon \\
	 &\qquad +  12\lambda_y \Omega^2(s_{\gamma^\ast,y}) \\
	 &\qquad +  \int_{0}^{\frac{5}{L^2} + \frac{2}{DL\sqrt{\lambda_y}}}\P_{\DtrainII}\Bigg[  \absBigg{\frac{\widehat{\E}_y^{\trainII}[s_{\gamma^\ast,y}] - \widehat{\E}_0^{\trainII}[s_{\gamma^\ast,y}]  }{\widehat{\Var}_{\gamma^\ast}^{\trainII}[s_{\gamma^\ast,y}] } - 1} \geq \frac{1}{\sqrt{12  \lambda_y} \Omega(s_{\gamma^\ast,y})}\sqrt{\epsilon}\Bigg] d\epsilon \\
	 &\qquad + 6\E_{\DtrainII}\Bigg[\sup_{\Omega(h_y) \leq \frac{1}{L \sqrt{\lambda_y}}} \absBig{\widehat{\E}_y^{\trainII}[h_y] - \E_y[h_y]} \Bigg]  +  6\E_{\DtrainII}\Bigg[\sup_{\Omega(h_y) \leq \frac{1}{L \sqrt{\lambda_y}}} \absBig{\widehat{\E}_0^{\trainII}[h_y] - \E_0[h_y]} \Bigg].
\end{align*}

Now, note that:
\smaller{}
\begin{align*}
	&\absBigg{\frac{\widehat{\E}_y^{\trainII}[s_{\gamma^\ast,y}] - \widehat{\E}_0^{\trainII}[s_{\gamma^\ast,y}]   }{\widehat{\Var}_{\gamma^\ast}^{\trainII}[s_{\gamma^\ast,y}] } - 1 } \\
	&=    \frac{ \absBig{\widehat{\E}_y^{\trainII}[s_{\gamma^\ast,y}] - \widehat{\E}_0^{\trainII}[s_{\gamma^\ast,y}]   - \widehat{\Var}_{\gamma^\ast}^{\trainII}[s_{\gamma^\ast,y}] } }{\widehat{\Var}_{\gamma^\ast}^{\trainII}[s_{\gamma^\ast,y}] }  \\ 
	&=   \frac{ \absBig{\Big\{ \big(\widehat{\E}_y^{\trainII}[s_{\gamma^\ast,y}] - \widehat{\E}_0^{\trainII}[s_{\gamma^\ast,y}]\big) - \big(\E_y[s_{\gamma^\ast,y}] - \E_0[s_{\gamma^\ast,y}]\big) \Big\} + \Big\{\E_{\gamma^\ast}[s_{\gamma^\ast,y}^2] - \widehat{\Var}_{\gamma^\ast}^{\trainII}[s_{\gamma^\ast,y}] \Big\} } }{\widehat{\Var}_{\gamma^\ast}^{\trainII}[s_{\gamma^\ast,y}] } \\ 
	&\leq  \frac{ \absBig{ \big(\widehat{\E}_y^{\trainII}[s_{\gamma^\ast,y}] - \widehat{\E}_0^{\trainII}[s_{\gamma^\ast,y}]\big) - \big(\E_y[s_{\gamma^\ast,y}] - \E_0[s_{\gamma^\ast,y}]\big) } +  \absBig{ \E_{\gamma^\ast}[s_{\gamma^\ast,y}^2] - \widehat{\E}_{\gamma^\ast}^{\trainII}[s_{\gamma^\ast,y}^2]  }   + \big(\widehat{\E}_{\gamma^\ast}^{\trainII}[s_{\gamma^\ast,y}]\big)^2 }{\widehat{\Var}_{\gamma^\ast}^{\trainII}[s_{\gamma^\ast,y}] }.
\end{align*}
\normalsize{}

So, for any $\delta > 0$: 
\smaller{}
\[
	\absBigg{\frac{\widehat{\E}_y^{\trainII}[s_{\gamma^\ast,y}] - \widehat{\E}_0^{\trainII}[s_{\gamma^\ast,y}]   }{\widehat{\Var}_{\gamma^\ast}^{\trainII}[s_{\gamma^\ast,y}] } - 1 }  \geq \delta 	
\]
\[
	\implies   \absBig{ \big(\widehat{\E}_y^{\trainII}[s_{\gamma^\ast,y}] - \widehat{\E}_0^{\trainII}[s_{\gamma^\ast,y}]\big) - \big(\E_y[s_{\gamma^\ast,y}] - \E_0[s_{\gamma^\ast,y}]\big) } +  \absBig{ \E_{\gamma^\ast}[s_{\gamma^\ast,y}^2] - \widehat{\E}_{\gamma^\ast}^{\trainII}[s_{\gamma^\ast,y}^2]  }   + \big(\widehat{\E}_{\gamma^\ast}^{\trainII}[s_{\gamma^\ast,y}]\big)^2   \geq \delta \widehat{\Var}_{\gamma^\ast}^{\trainII}[s_{\gamma^\ast,y}] 
\]
\[
	\iff   \absBig{ \big(\widehat{\E}_y^{\trainII}[s_{\gamma^\ast,y}] - \widehat{\E}_0^{\trainII}[s_{\gamma^\ast,y}]\big) - \big(\E_y[s_{\gamma^\ast,y}] - \E_0[s_{\gamma^\ast,y}]\big) } +  \absBig{ \E_{\gamma^\ast}[s_{\gamma^\ast,y}^2] - \widehat{\E}_{\gamma^\ast}^{\trainII}[s_{\gamma^\ast,y}^2]  }   + \big(\widehat{\E}_{\gamma^\ast}^{\trainII}[s_{\gamma^\ast,y}]\big)^2 +    \delta\big( \E_{\gamma^\ast}[s_{\gamma^\ast,y}^2] - \widehat{\Var}_{\gamma^\ast}^{\trainII}[s_{\gamma^\ast,y}]\big)  \geq \delta\E_{\gamma^\ast}[s_{\gamma^\ast,y}^2]
\]
\[
	\implies   \absBig{ \big(\widehat{\E}_y^{\trainII}[s_{\gamma^\ast,y}] - \widehat{\E}_0^{\trainII}[s_{\gamma^\ast,y}]\big) - \big(\E_y[s_{\gamma^\ast,y}] - \E_0[s_{\gamma^\ast,y}]\big) } +  \absBig{ \widehat{\E}_{\gamma^\ast}^{\trainII}[s_{\gamma^\ast,y}^2] - \E_{\gamma^\ast}[s_{\gamma^\ast,y}^2]   }   + \big(\widehat{\E}_{\gamma^\ast}^{\trainII}[s_{\gamma^\ast,y}]\big)^2 +    \delta\absBig{ \widehat{\Var}_{\gamma^\ast}^{\trainII}[s_{\gamma^\ast,y}] - \E_{\gamma^\ast}[s_{\gamma^\ast,y}^2] }  \geq \delta \sqrt{\Lambda}
\]
\normalsize{}

Ergo:
\smaller{}
\begin{align*}
	&\P_{\DtrainII} \Bigg[ \absBigg{\frac{\widehat{\E}_y^{\trainII}[s_{\gamma^\ast,y}] - \widehat{\E}_0^{\trainII}[s_{\gamma^\ast,y}]   }{\widehat{\Var}_{\gamma^\ast}^{\trainII}[s_{\gamma^\ast,y}] } - 1 }  \geq \delta \Bigg] \\
	&\leq \P_{\DtrainII} \Bigg[ \frac{4}{\sqrt{\Lambda}}  \absBig{ \big(\widehat{\E}_y^{\trainII}[s_{\gamma^\ast,y}] - \widehat{\E}_0^{\trainII}[s_{\gamma^\ast,y}]\big) - \big(\E_y[s_{\gamma^\ast,y}] - \E_0[s_{\gamma^\ast,y}]\big) }   \geq \delta \Bigg]\\
	&\qquad  +  \P_{\DtrainII} \Bigg[ \frac{4}{\sqrt{\Lambda}} \absBig{ \widehat{\E}_{\gamma^\ast}^{\trainII}[s_{\gamma^\ast,y}^2] - \E_{\gamma^\ast}[s_{\gamma^\ast,y}^2]   }  \geq \delta \Bigg] \\
	&\qquad  +  \P_{\DtrainII} \Bigg[ \frac{4}{\sqrt{\Lambda}} \big(\widehat{\E}_{\gamma^\ast}^{\trainII}[s_{\gamma^\ast,y}]\big)^2  \geq \delta \Bigg] \\
	&\qquad  +  \P_{\DtrainII} \Bigg[ \absBig{  \widehat{\Var}_{\gamma^\ast}^{\trainII}[s_{\gamma^\ast,y}] - \E_{\gamma^\ast}[s_{\gamma^\ast,y}^2] }  \geq \frac{\sqrt{\Lambda}}{ 4} \Bigg].
\end{align*}
\normalsize{}
Therefore:
\smaller{}
\begin{align*}
	 &\int_{0}^{\frac{5}{L^2} + \frac{2}{DL\sqrt{\lambda_y}}}\P_{\DtrainII}\Bigg[ \absBigg{ \frac{\widehat{\E}_y^{\trainII}[s_{\gamma^\ast,y}] - \widehat{\E}_0^{\trainII}[s_{\gamma^\ast,y}]  }{\widehat{\Var}_{\gamma^\ast}^{\trainII}[s_{\gamma^\ast,y}] } - 1}  \geq \frac{L}{24}\epsilon\Bigg] d\epsilon \\
	 &\leq \int_{0}^{\infty}\P_{\DtrainII} \Bigg[ \frac{4}{\sqrt{\Lambda}}  \absBig{ \big(\widehat{\E}_y^{\trainII}[s_{\gamma^\ast,y}] - \widehat{\E}_0^{\trainII}[s_{\gamma^\ast,y}]\big) - \big(\E_y[s_{\gamma^\ast,y}] - \E_0[s_{\gamma^\ast,y}]\big) }   \geq \frac{L}{24}\epsilon \Bigg] d\epsilon\\
	&\qquad  +  \int_{0}^{\infty}\P_{\DtrainII} \Bigg[ \frac{4}{\sqrt{\Lambda}} \absBig{ \widehat{\E}_{\gamma^\ast}^{\trainII}[s_{\gamma^\ast,y}^2] - \E_{\gamma^\ast}[s_{\gamma^\ast,y}^2]   }  \geq \frac{L}{24}\epsilon  \Bigg] d\epsilon \\
	&\qquad  +  \int_{0}^{\infty}\P_{\DtrainII} \Bigg[ \frac{4}{\sqrt{\Lambda}} \big(\widehat{\E}_{\gamma^\ast}^{\trainII}[s_{\gamma^\ast,y}]\big)^2  \geq \frac{L}{24}\epsilon \Bigg] d\epsilon \\
	&\qquad  + \bigg(  \frac{5}{L^2} + \frac{2}{DL\sqrt{\lambda_y}}  \bigg) \P_{\DtrainII} \Bigg[  \absBig{ \widehat{\Var}_{\gamma^\ast}^{\trainII}[s_{\gamma^\ast,y}]  -   \E_{\gamma^\ast}[s_{\gamma^\ast,y}^2] }  \geq \frac{\sqrt{\Lambda}}{ 4} \Bigg]  \\ \\ 
	&\leq \frac{96}{L\sqrt{\Lambda}} \sqrt{\Var_{\DtrainII}\Big[ \widehat{\E}_y^{\trainII}[s_{\gamma^\ast,y}] - \widehat{\E}_0^{\trainII}[s_{\gamma^\ast,y}] \Big]}  + \frac{96}{L\sqrt{\Lambda}} \sqrt{\Var_{\DtrainII}\Big[   \widehat{\E}_{\gamma^\ast}^{\trainII}[s_{\gamma^\ast,y}^2] \Big]} \\
	&\qquad + \frac{96}{L\sqrt{\Lambda}} \Var_{\DtrainII}\Big[   \widehat{\E}_{\gamma^\ast}^{\trainII}[s_{\gamma^\ast,y}] \Big] \\
	&\qquad  +  \bigg(  \frac{5}{L^2} + \frac{2}{DL\sqrt{\lambda_y}}  \bigg) \P_{\DtrainII} \Bigg[ \absBig{\widehat{\Var}_{\gamma^\ast}^{\trainII}[s_{\gamma^\ast,y}] -  \E_{\gamma^\ast}[s_{\gamma^\ast,y}^2] } \geq \frac{\sqrt{\Lambda}}{ 4} \Bigg] \\ \\ 
	&\leq \frac{96}{L\sqrt{\Lambda}} \sqrt{\Var_{\DtrainII}\Big[ \widehat{\E}_y^{\trainII}[s_{\gamma^\ast,y}] - \widehat{\E}_0^{\trainII}[s_{\gamma^\ast,y}] \Big]}  + \frac{96}{L\sqrt{\Lambda}} \sqrt{\Var_{\DtrainII}\Big[   \widehat{\E}_{\gamma^\ast}^{\trainII}[s_{\gamma^\ast,y}^2] \Big]} \\
	&\qquad + \frac{96}{L\sqrt{\Lambda}} \Var_{\DtrainII}\Big[   \widehat{\E}_{\gamma^\ast}^{\trainII}[s_{\gamma^\ast,y}] \Big] \\
	&\qquad  +  \bigg(  \frac{5}{L^2} + \frac{2}{DL\sqrt{\lambda_y}}  \bigg) \P_{\DtrainII} \Bigg[ \absBig{\widehat{\E}_{\gamma^\ast}^{\trainII}[s_{\gamma^\ast,y}^2] -  \E_{\gamma^\ast}[s_{\gamma^\ast,y}^2] } \geq \frac{\sqrt{\Lambda}}{8} \Bigg] \\ 
	&\qquad  +  \bigg(  \frac{5}{L^2} + \frac{2}{DL\sqrt{\lambda_y}}  \bigg) \P_{\DtrainII} \Bigg[ \absBig{ \widehat{\E}_{\gamma^\ast}^{\trainII}[s_{\gamma^\ast,y}] } \geq \sqrt{\frac{\sqrt{\Lambda}}{8}} \Bigg] \\ \\ 
	&\leq \frac{96}{L\sqrt{\Lambda}} \sqrt{\Var_{\DtrainII}\Big[ \widehat{\E}_y^{\trainII}[s_{\gamma^\ast,y}] - \widehat{\E}_0^{\trainII}[s_{\gamma^\ast,y}] \Big]}  + \frac{96}{L\sqrt{\Lambda}} \sqrt{\Var_{\DtrainII}\Big[   \widehat{\E}_{\gamma^\ast}^{\trainII}[s_{\gamma^\ast,y}^2] \Big]} \\
	&\qquad + \frac{96}{L\sqrt{\Lambda}} \Var_{\DtrainII}\Big[   \widehat{\E}_{\gamma^\ast}^{\trainII}[s_{\gamma^\ast,y}] \Big] \\
	&\qquad  +  2\bigg(  \frac{5}{L^2} + \frac{2}{DL\sqrt{\lambda_y}}  \bigg) \Bigg(\exp\bigg\{ - \frac{L^{4} \sqrt{\Lambda}^2}{1024}\xi n^\train \bigg\} + \exp\bigg\{ - \frac{L^2 \sqrt{\Lambda} }{32} \xi n^\train \bigg\}\Bigg),
\end{align*}
\normalsize{}

where the last line is by Hoeffding's inequality.  Analogously:
\smaller{}
\[
	 \int_0^{\frac{5}{L^2} + \frac{2}{DL\sqrt{\lambda_y}}}\P_{\DtrainII}\Bigg[  \absBigg{\frac{\widehat{\E}_y^{\trainII}[s_{\gamma^\ast,y}] - \widehat{\E}_0^{\trainII}[s_{\gamma^\ast,y}]  }{\widehat{\Var}_{\gamma^\ast}^{\trainII}[s_{\gamma^\ast,y}] } - 1} \geq \frac{1}{\sqrt{12  \lambda_y} \Omega(s_{\gamma^\ast,y})}\sqrt{\epsilon}\Bigg] d\epsilon 
\]

\begin{align*}
	 &\leq \int_{0}^{\infty}\P_{\DtrainII} \Bigg[ \frac{4}{\sqrt{\Lambda}}  \absBig{ \big(\widehat{\E}_y^{\trainII}[s_{\gamma^\ast,y}] - \widehat{\E}_0^{\trainII}[s_{\gamma^\ast,y}]\big) - \big(\E_y[s_{\gamma^\ast,y}] - \E_0[s_{\gamma^\ast,y}]\big) }   \geq  \frac{1}{\sqrt{12  \lambda_y} \Omega(s_{\gamma^\ast,y})}\sqrt{\epsilon}  \Bigg] d\epsilon\\
	&\qquad  +  \int_{0}^{\infty}\P_{\DtrainII} \Bigg[ \frac{4}{\sqrt{\Lambda}} \absBig{ \widehat{\E}_{\gamma^\ast}^{\trainII}[s_{\gamma^\ast,y}^2] - \E_{\gamma^\ast}[s_{\gamma^\ast,y}^2]   }  \geq \frac{1}{\sqrt{12  \lambda_y} \Omega(s_{\gamma^\ast,y})}\sqrt{\epsilon}  \Bigg] d\epsilon \\
	&\qquad  +  \int_{0}^{\infty}\P_{\DtrainII} \Bigg[ \frac{4}{\sqrt{\Lambda}} \big(\widehat{\E}_{\gamma^\ast}^{\trainII}[s_{\gamma^\ast,y}]\big)^2  \geq  \frac{1}{\sqrt{12  \lambda_y} \Omega(s_{\gamma^\ast,y})}\sqrt{\epsilon} \Bigg] d\epsilon \\
	&\qquad  + \bigg(  \frac{5}{L^2} + \frac{2}{DL\sqrt{\lambda_y}}  \bigg) \P_{\DtrainII} \Bigg[  \absBig{ \widehat{\Var}_{\gamma^\ast}^{\trainII}[s_{\gamma^\ast,y}]  -   \E_{\gamma^\ast}[s_{\gamma^\ast,y}^2] }  \geq \frac{\sqrt{\Lambda}}{ 4} \Bigg]  \\ \\ 
	 &\leq \int_{0}^{\infty}\P_{\DtrainII} \Bigg[ \frac{16}{\sqrt{\Lambda}^2}  \absBig{ \big(\widehat{\E}_y^{\trainII}[s_{\gamma^\ast,y}] - \widehat{\E}_0^{\trainII}[s_{\gamma^\ast,y}]\big) - \big(\E_y[s_{\gamma^\ast,y}] - \E_0[s_{\gamma^\ast,y}]\big) }^2   \geq   \frac{1}{12  \lambda_y \Omega(s_{\gamma^\ast,y})^2}\epsilon \Bigg] d\epsilon\\
	&\qquad  +  \int_{0}^{\infty}\P_{\DtrainII} \Bigg[ \frac{16}{\sqrt{\Lambda}^2} \absBig{ \widehat{\E}_{\gamma^\ast}^{\trainII}[s_{\gamma^\ast,y}^2] - \E_{\gamma^\ast}[s_{\gamma^\ast,y}^2]   }^2  \geq  \frac{1}{12  \lambda_y \Omega(s_{\gamma^\ast,y})^2}\epsilon  \Bigg] d\epsilon \\
	&\qquad  +  \int_{0}^{\infty}\P_{\DtrainII} \Bigg[ \frac{16}{\sqrt{\Lambda}^2} \big(\widehat{\E}_{\gamma^\ast}^{\trainII}[s_{\gamma^\ast,y}]\big)^4 \geq  \frac{1}{12  \lambda_y \Omega(s_{\gamma^\ast,y})^2}\epsilon  \Bigg] d\epsilon \\
	&\qquad  + \bigg(  \frac{5}{L^2} + \frac{2}{DL\sqrt{\lambda_y}}  \bigg) \P_{\DtrainII} \Bigg[  \absBig{ \widehat{\Var}_{\gamma^\ast}^{\trainII}[s_{\gamma^\ast,y}]  -   \E_{\gamma^\ast}[s_{\gamma^\ast,y}^2] }  \geq \frac{\sqrt{\Lambda}}{ 4} \Bigg]  \\ \\ 
	&= 192 \frac{\lambda_y \Omega(s_{\gamma^\ast,y})^2}{\sqrt{\Lambda}^2} \bigg( \Var_{\DtrainII}\Big[\widehat{\E}_y^{\trainII}[s_{\gamma^\ast,y}] - \widehat{\E}_0^{\trainII}[s_{\gamma^\ast,y}]\Big] + \Var_{\DtrainII}\Big[\widehat{\E}_{\gamma^\ast}^{\trainII}[s_{\gamma^\ast,y}^2]\Big] + \E_{\DtrainII}\Big[ \big(\widehat{\E}_{\gamma^\ast}^{\trainII}[s_{\gamma^\ast,y}]\big)^4  \Big] \bigg) \\ 
	&\qquad  + \bigg(  \frac{5}{L^2} + \frac{2}{DL\sqrt{\lambda_y}}  \bigg) \P_{\DtrainII} \Bigg[  \absBig{ \widehat{\Var}_{\gamma^\ast}^{\trainII}[s_{\gamma^\ast,y}]  -   \E_{\gamma^\ast}[s_{\gamma^\ast,y}^2] }  \geq \frac{\sqrt{\Lambda}}{ 4} \Bigg]  \\ \\ 
	&\leq 192 \frac{\lambda_y \Omega(s_{\gamma^\ast,y})^2}{\sqrt{\Lambda}^2} \bigg( \Var_{\DtrainII}\Big[\widehat{\E}_y^{\trainII}[s_{\gamma^\ast,y}] - \widehat{\E}_0^{\trainII}[s_{\gamma^\ast,y}]\Big] + \Var_{\DtrainII}\Big[\widehat{\E}_{\gamma^\ast}^{\trainII}[s_{\gamma^\ast,y}^2]\Big] + \E_{\DtrainII}\Big[ \big(\widehat{\E}_{\gamma^\ast}^{\trainII}[s_{\gamma^\ast,y}]\big)^4  \Big] \bigg) \\ 
	&\qquad  +  2\bigg(  \frac{5}{L^2} + \frac{2}{DL\sqrt{\lambda_y}}  \bigg) \Bigg(\exp\bigg\{ - \frac{L^{4} \sqrt{\Lambda}^2}{1024}\xi n^\train \bigg\} + \exp\bigg\{ - \frac{L^2 \sqrt{\Lambda} }{32} \xi n^\train \bigg\}\Bigg) \\ \\ 
	&\leq 192 \frac{\lambda_y \Omega(s_{\gamma^\ast,y})^2}{\sqrt{\Lambda}^2} \bigg( \Var_{\DtrainII}\Big[\widehat{\E}_y^{\trainII}[s_{\gamma^\ast,y}] - \widehat{\E}_0^{\trainII}[s_{\gamma^\ast,y}]\Big] + \Var_{\DtrainII}\Big[\widehat{\E}_{\gamma^\ast}^{\trainII}[s_{\gamma^\ast,y}^2]\Big] + \frac{4}{L^2} \Var_{\DtrainII}\Big[ \widehat{\E}_{\gamma^\ast}^{\trainII}[s_{\gamma^\ast,y}] \Big] \bigg) \\ 
	&\qquad  +  2\bigg(  \frac{5}{L^2} + \frac{2}{DL\sqrt{\lambda_y}}  \bigg) \Bigg(\exp\bigg\{ - \frac{L^{4} \sqrt{\Lambda}^2}{1024}\xi n^\train \bigg\} + \exp\bigg\{ - \frac{L^2 \sqrt{\Lambda} }{32} \xi n^\train \bigg\}\Bigg),
\end{align*}
\normalsize{}
where the last line is because $\abs{s_{\gamma^\ast,y}} \leq \frac{2}{L} \implies \big(\widehat{\E}_{\gamma^\ast}^{\trainII}[s_{\gamma^\ast,y}] \big)^2 \leq \frac{4}{L^2}$. Therefore, overall, we have that:
\smaller{}
\begin{align*}
	&\E_{\DtrainII}[T_y] \\
	&\leq  6 \sqrt{\Var_{\DtrainII}\Big[ \widehat{\E}_y^{\trainII}[s_{\gamma^\ast,y}] - \widehat{\E}_0^{\trainII}[s_{\gamma^\ast,y}] \Big]} +  12\lambda_y \Omega^2(s_{\gamma^\ast,y}) \\
	&\qquad + 6\E_{\DtrainII}\Bigg[\sup_{\Omega(h_y) \leq \frac{1}{L \sqrt{\lambda_y}}} \absBig{\widehat{\E}_y^{\trainII}[h_y] - \E_y[h_y]} \Bigg]  +  6\E_{\DtrainII}\Bigg[\sup_{\Omega(h_y) \leq \frac{1}{L \sqrt{\lambda_y}}} \absBig{\widehat{\E}_0^{\trainII}[h_y] - \E_0[h_y]} \Bigg] \\
	 &\qquad +  \int_{0}^{\frac{5}{L^2} + \frac{2}{DL\sqrt{\lambda_y}}}\P_{\DtrainII}\Bigg[ \absBigg{ \frac{\widehat{\E}_y^{\trainII}[s_{\gamma^\ast,y}] - \widehat{\E}_0^{\trainII}[s_{\gamma^\ast,y}]  }{\widehat{\Var}_{\gamma^\ast}^{\trainII}[s_{\gamma^\ast,y}] } - 1}  \geq \frac{L}{24}\epsilon\Bigg] d\epsilon \\
	 &\qquad +  \int_{0}^{\frac{5}{L^2} + \frac{2}{DL\sqrt{\lambda_y}}}\P_{\DtrainII}\Bigg[  \absBigg{\frac{\widehat{\E}_y^{\trainII}[s_{\gamma^\ast,y}] - \widehat{\E}_0^{\trainII}[s_{\gamma^\ast,y}]  }{\widehat{\Var}_{\gamma^\ast}^{\trainII}[s_{\gamma^\ast,y}] } - 1} \geq \frac{1}{\sqrt{12  \lambda_y} \Omega(s_{\gamma^\ast,y})}\sqrt{\epsilon}\Bigg] d\epsilon \\ \\ \\
	 &\leq  6 \sqrt{\Var_{\DtrainII}\Big[ \widehat{\E}_y^{\trainII}[s_{\gamma^\ast,y}] - \widehat{\E}_0^{\trainII}[s_{\gamma^\ast,y}] \Big]} +  12\lambda_y \Omega^2(s_{\gamma^\ast,y}) +  \E_{\DtrainII}[\U_y(\lambda_y)]\\
	 &\qquad +  \frac{96}{L\sqrt{\Lambda}} \bigg(\sqrt{\Var_{\DtrainII}\Big[ \widehat{\E}_y^{\trainII}[s_{\gamma^\ast,y}] - \widehat{\E}_0^{\trainII}[s_{\gamma^\ast,y}] \Big]}  + \sqrt{\Var_{\DtrainII}\Big[   \widehat{\E}_{\gamma^\ast}^{\trainII}[s_{\gamma^\ast,y}^2] \Big]}  + \Var_{\DtrainII}\Big[   \widehat{\E}_{\gamma^\ast}^{\trainII}[s_{\gamma^\ast,y}] \Big]\bigg)  \\
	 &\qquad + 192 \frac{\lambda_y \Omega(s_{\gamma^\ast,y})^2}{\sqrt{\Lambda}^2} \bigg( \Var_{\DtrainII}\Big[\widehat{\E}_y^{\trainII}[s_{\gamma^\ast,y}] - \widehat{\E}_0^{\trainII}[s_{\gamma^\ast,y}]\Big] + \Var_{\DtrainII}\Big[\widehat{\E}_{\gamma^\ast}^{\trainII}[s_{\gamma^\ast,y}^2]\Big] + \frac{4}{L^2}\Var_{\DtrainII}\Big[ \widehat{\E}_{\gamma^\ast}^{\trainII}[s_{\gamma^\ast,y}]  \Big] \bigg) \\ 
	 &\qquad+  4\bigg(  \frac{5}{L^2} + \frac{2}{DL\sqrt{\lambda_y}}  \bigg) \Bigg(\exp\bigg\{ - \frac{L^{4} \sqrt{\Lambda}^2}{1024}\xi n^\train \bigg\} + \exp\bigg\{ - \frac{L^2 \sqrt{\Lambda} }{32} \xi n^\train \bigg\}\Bigg) \\ 
	 \shortintertext{\normalsize{} \newline Now, we know that we have a minimum of $\frac{1}{2}n^{\train}\xi$ samples in $\DtrainII$ for each class. Further, we know that $\abs{s_{\gamma^\ast,y}} \leq \frac{2}{L}$, meaning that the variance of $s_{\gamma^\ast,y}$ (and its powers) can all be bounded above in terms of solely $L$ (via Popoviciu's variance inequality), regardless of the distribution the variance is with respect to. Since $\xi$ and $L$ are constants that are independent of $w$, it therefore follows that all the variance terms displayed above go down like $O(1/n^{\train})$. Ergo, the above is equal to: \newline\smaller{}}
	  &=  O\Bigg( \frac{1}{\sqrt{n^\train}} + \lambda_y \Omega^2(s_{\gamma^\ast,y}) + \E_{\DtrainII}[\U_y(\lambda_y)]  + \bigg(1 + \frac{1}{\sqrt{\lambda_y}}\bigg)e^{-Cn^\train} \Bigg),
\end{align*}
\normalsize{}
for some $C > 0$ that is independent of $w$.

Next, we prove the tail bound on $T_y$. For any $a > 0 $, it follows from our work in bounding $\E_{\DtrainII}[T_y]$ that:
\begin{align*}
	&\P_{\DtrainII}[T_y \geq a]\\
	 &\leq  \P_{\DtrainII}[R_y \geq a] \\ \\
	&\leq  \P_{\DtrainII}\bigg[ \absBig{\big(\E_y[s_{\gamma^\ast,y}] - \E_0[s_{\gamma^\ast,y}]\big) - \big(\widehat{\E}_y^{\trainII}[s_{\gamma^\ast,y}] - \widehat{\E}_0^{\trainII}[s_{\gamma^\ast,y}] \big) } \geq \frac{1}{6}a\bigg]  \\
	 &\qquad +  \P_{\DtrainII}\Bigg[ \absBigg{ \frac{\widehat{\E}_y^{\trainII}[s_{\gamma^\ast,y}] - \widehat{\E}_0^{\trainII}[s_{\gamma^\ast,y}]  }{\widehat{\Var}_{\gamma^\ast}^{\trainII}[s_{\gamma^\ast,y}] } - 1}  \geq \frac{L}{24}a\Bigg]  \\
	 &\qquad +  \P_{\DtrainII}\Big[ \lambda_y \Omega^2(s_{\gamma^\ast,y})  \geq \frac{1}{12}a\Big]  \\
	 &\qquad +  \P_{\DtrainII}\Bigg[  \Bigg( \frac{\widehat{\E}_y^{\trainII}[s_{\gamma^\ast,y}] - \widehat{\E}_0^{\trainII}[s_{\gamma^\ast,y}]  }{\widehat{\Var}_{\gamma^\ast}^{\trainII}[s_{\gamma^\ast,y}] } - 1\Bigg)^2 \geq \frac{1}{12  \lambda_y \Omega^2(s_{\gamma^\ast,y})}a\Bigg]  \\
	 &\qquad +  \P_{\DtrainII}\Bigg[\sup_{\Omega(h_y) \leq \frac{1}{L \sqrt{\lambda_y}}} \absBig{\widehat{\E}_y^{\trainII}[h_y] - \E_y[h_y]}  \geq \frac{1}{6}a\Bigg] +  \P_{\DtrainII}\Bigg[\sup_{\Omega(h_y) \leq \frac{1}{L \sqrt{\lambda_y}}} \absBig{\widehat{\E}_0^{\trainII}[h_y] - \E_0[h_y]} \geq \frac{1}{6}a\Bigg]  \\ \\ 
	 &\leq  \P_{\DtrainII}\bigg[ \absBig{ \widehat{\E}_y^{\trainII}[s_{\gamma^\ast,y}] - \E_y[s_{\gamma^\ast,y}] } \geq \frac{1}{12}a\bigg]  + \P_{\DtrainII}\bigg[ \absBig{ \widehat{\E}_0^{\trainII}[s_{\gamma^\ast,y}] - \E_0[s_{\gamma^\ast,y}]  } \geq \frac{1}{12}a\bigg]  \\
	 &\qquad +  2\P_{\DtrainII}\Bigg[ \absBigg{ \frac{\widehat{\E}_y^{\trainII}[s_{\gamma^\ast,y}] - \widehat{\E}_0^{\trainII}[s_{\gamma^\ast,y}]  }{\widehat{\Var}_{\gamma^\ast}^{\trainII}[s_{\gamma^\ast,y}] } - 1}  \geq \min\bigg\{ \frac{L}{24}a, \frac{1}{\sqrt{12  \lambda_y \Omega^2(s_{\gamma^\ast,y})} }\sqrt{a} \bigg\} \Bigg]  \\
	 &\qquad +  \IBig{ \lambda_y \Omega^2(s_{\gamma^\ast,y})  \geq \frac{1}{12}a } \\
	 &\qquad +  \P_{\DtrainII}\Bigg[\sup_{\Omega(h_y) \leq \frac{1}{L \sqrt{\lambda_y}}} \absBig{\widehat{\E}_y^{\trainII}[h_y] - \E_y[h_y]}  \geq \frac{1}{6}a\Bigg] +  \P_{\DtrainII}\Bigg[\sup_{\Omega(h_y) \leq \frac{1}{L \sqrt{\lambda_y}}} \absBig{\widehat{\E}_0^{\trainII}[h_y] - \E_0[h_y]} \geq \frac{1}{6}a\Bigg]. 
\end{align*}

Let's bound each of the probabilities above. Note that, for any class $j$, we have by Hoeffding's inequality that:
\begin{align*}
	\P_{\DtrainII}\bigg[ \absBig{ \widehat{\E}_j^{\trainII}[s_{\gamma^\ast,y}] - \E_j[s_{\gamma^\ast,y}]  } \geq \frac{1}{12}a\bigg] &\leq 2\exp\bigg\{ -2 \frac{\tfrac{1}{2}\pi_j^\train n^\train \big(\tfrac{a}{12}\big)^2 }{4/L^2}   \bigg\}.
\end{align*}

Also, by our earlier work and Hoeffding's inequality, we have for any $\delta > 0$ that:
\begin{align*}
	&\P_{\DtrainII} \Bigg[ \absBigg{\frac{\widehat{\E}_y^{\trainII}[s_{\gamma^\ast,y}] - \widehat{\E}_0^{\trainII}[s_{\gamma^\ast,y}]   }{\widehat{\Var}_{\gamma^\ast}^{\trainII}[s_{\gamma^\ast,y}] } - 1 }  \geq \delta \Bigg]\\
	 &\leq \P_{\DtrainII} \Bigg[  \absBig{ \big(\widehat{\E}_y^{\trainII}[s_{\gamma^\ast,y}] - \widehat{\E}_0^{\trainII}[s_{\gamma^\ast,y}]\big) - \big(\E_y[s_{\gamma^\ast,y}] - \E_0[s_{\gamma^\ast,y}]\big) }   \geq \frac{\sqrt{\Lambda} }{4} \delta \Bigg]\\
	&\qquad  +  \P_{\DtrainII} \Bigg[ \absBig{ \widehat{\E}_{\gamma^\ast}^{\trainII}[s_{\gamma^\ast,y}^2] - \E_{\gamma^\ast}[s_{\gamma^\ast,y}^2]   }  \geq  \frac{\sqrt{\Lambda} }{4}  \delta \Bigg] \\
	&\qquad  +  \P_{\DtrainII} \Bigg[  \absBig{\widehat{\E}_{\gamma^\ast}^{\trainII}[s_{\gamma^\ast,y}]} \geq \frac{ \Lambda^{1/4} }{2}  \sqrt{\delta} \Bigg] \\
	&\qquad  +  \P_{\DtrainII} \Bigg[ \absBig{  \widehat{\Var}_{\gamma^\ast}^{\trainII}[s_{\gamma^\ast,y}] - \E_{\gamma^\ast}[s_{\gamma^\ast,y}^2] }  \geq \frac{\sqrt{\Lambda}}{ 4} \Bigg].
\end{align*}

As for the first term, we have by Hoeffding's inequality that:
\small{}
\begin{align*}
	&\P_{\DtrainII} \Bigg[  \absBig{ \big(\widehat{\E}_y^{\trainII}[s_{\gamma^\ast,y}] - \widehat{\E}_0^{\trainII}[s_{\gamma^\ast,y}]\big) - \big(\E_y[s_{\gamma^\ast,y}] - \E_0[s_{\gamma^\ast,y}]\big) }   \geq \frac{\sqrt{\Lambda} }{4} \delta \Bigg] \\
	&\leq 2\exp\bigg\{ -2 \frac{\tfrac{1}{2}\pi_0^\train n^\train \big(\frac{\sqrt{\Lambda} }{8} \delta\big)^2 }{4/L^2}   \bigg\} + 2\exp\bigg\{ -2 \frac{\tfrac{1}{2}\pi_y^\train n^\train \big(\frac{\sqrt{\Lambda} }{8} \delta\big)^2 }{4/L^2}   \bigg\}.
\end{align*}
\normalsize{}

For the second term:
\begin{align*}
	  \P_{\DtrainII} \Bigg[ \absBig{ \widehat{\E}_{\gamma^\ast}^{\trainII}[s_{\gamma^\ast,y}^2] - \E_{\gamma^\ast}[s_{\gamma^\ast,y}^2]   }  \geq  \frac{\sqrt{\Lambda} }{4}  \delta \Bigg] &\leq  \sum_{j=0}^{m} \P_{\DtrainII} \Bigg[ \absBig{ \widehat{\E}_{j}^{\trainII}[s_{\gamma^\ast,y}^2] - \E_{j}[s_{\gamma^\ast,y}^2]   }  \geq  \frac{\sqrt{\Lambda} }{4}  \delta \Bigg] \\ 
	   &\leq  2\sum_{j=0}^{m}\exp\bigg\{ -2 \frac{\tfrac{1}{2}\pi_j^\train n^\train \big(\tfrac{\sqrt{\Lambda} }{4}  \delta \big)^2}{16/L^4}\bigg\}.
\end{align*}

For the third term:
\begin{align*}
	\P_{\DtrainII} \Bigg[  \absBig{\widehat{\E}_{\gamma^\ast}^{\trainII}[s_{\gamma^\ast,y}]} \geq \frac{ \Lambda^{1/4} }{2}  \sqrt{\delta} \Bigg] &\leq \sum_{j=0}^{m} \P_{\DtrainII} \Bigg[  \absBig{\widehat{\E}_{j}^{\trainII}[s_{\gamma^\ast,y}]    -    \E_{j}[s_{\gamma^\ast,y}]   } \geq \frac{ \Lambda^{1/4} }{2}  \sqrt{\delta} \Bigg] \\ 
	&\leq 2 \sum_{j=0}^{m} \exp\bigg\{ -2 \frac{\tfrac{1}{2}\pi_j^\train n^\train \big( \frac{ \Lambda^{1/4} }{2} \sqrt{\delta} \big)^2 }{4/L^2}   \bigg\}.
\end{align*}

For the fourth term:
\begin{align*}
	 &\P_{\DtrainII} \Bigg[ \absBig{  \widehat{\Var}_{\gamma^\ast}^{\trainII}[s_{\gamma^\ast,y}] - \E_{\gamma^\ast}[s_{\gamma^\ast,y}^2] }  \geq \frac{\sqrt{\Lambda}}{ 4} \Bigg] \\
	 &\leq  \P_{\DtrainII} \Bigg[ \absBig{  \widehat{\E}_{\gamma^\ast}^{\trainII}[s_{\gamma^\ast,y}^2] - \E_{\gamma^\ast}[s_{\gamma^\ast,y}^2] }  \geq \frac{\sqrt{\Lambda}}{8} \Bigg] + \P_{\DtrainII} \Bigg[ \Big( \widehat{\E}_{\gamma^\ast}^{\trainII}[s_{\gamma^\ast,y}]   \Big)^2  \geq \frac{\sqrt{\Lambda}}{8} \Bigg] \\ 
	 &= \P_{\DtrainII} \Bigg[ \absBig{  \widehat{\E}_{\gamma^\ast}^{\trainII}[s_{\gamma^\ast,y}^2] - \E_{\gamma^\ast}[s_{\gamma^\ast,y}^2] }  \geq \frac{\sqrt{\Lambda}}{8} \Bigg] + \P_{\DtrainII} \Bigg[ \absBig{ \widehat{\E}_{\gamma^\ast}^{\trainII}[s_{\gamma^\ast,y}]   }  \geq \frac{\Lambda^{1/4}}{\sqrt{8}} \Bigg] \\ 
	&\leq  2\sum_{j=0}^{m}\exp\bigg\{ -2 \frac{\tfrac{1}{2}\pi_j^\train n^\train \big(   \tfrac{\sqrt{\Lambda}}{8} \big)^2}{16/L^4}\bigg\} + 2 \sum_{j=0}^{m} \exp\bigg\{ -2 \frac{\tfrac{1}{2}\pi_j^\train n^\train \big( \frac{\Lambda^{1/4}}{\sqrt{8}} \big)^2 }{4/L^2}   \bigg\}.
\end{align*}

Thus, it follows that:
\begin{align*}
	&\P_{\DtrainII} \Bigg[ \absBigg{\frac{\widehat{\E}_y^{\trainII}[s_{\gamma^\ast,y}] - \widehat{\E}_0^{\trainII}[s_{\gamma^\ast,y}]   }{\widehat{\Var}_{\gamma^\ast}^{\trainII}[s_{\gamma^\ast,y}] } - 1 }  \geq \delta \Bigg] \\ 
	&\leq  2\exp\bigg\{ -2 \frac{\tfrac{1}{2}\pi_0^\train n^\train \big(\frac{\sqrt{\Lambda} }{8} \delta\big)^2 }{4/L^2}   \bigg\} + 2\exp\bigg\{ -2 \frac{\tfrac{1}{2}\pi_y^\train n^\train \big(\frac{\sqrt{\Lambda} }{8} \delta\big)^2 }{4/L^2}   \bigg\} \\ 
	&\qquad + 2\sum_{j=0}^{m}\exp\bigg\{ -2 \frac{\tfrac{1}{2}\pi_j^\train n^\train \big(\tfrac{\sqrt{\Lambda} }{4}  \delta \big)^2}{16/L^4}\bigg\} +  2 \sum_{j=0}^{m} \exp\bigg\{ -2 \frac{\tfrac{1}{2}\pi_j^\train n^\train \big( \frac{ \Lambda^{1/4} }{2} \sqrt{\delta} \big)^2 }{4/L^2}   \bigg\} \\ 
	&\qquad + 2\sum_{j=0}^{m}\exp\bigg\{ -2 \frac{\tfrac{1}{2}\pi_j^\train n^\train \big(   \tfrac{\sqrt{\Lambda}}{8} \big)^2}{16/L^4}\bigg\} + 2 \sum_{j=0}^{m} \exp\bigg\{ -2 \frac{\tfrac{1}{2}\pi_j^\train n^\train \big( \frac{\Lambda^{1/4}}{\sqrt{8}} \big)^2 }{4/L^2}   \bigg\} \\ \\
	&\leq  2\exp\bigg\{ -2 \frac{\tfrac{1}{2}\pi_0^\train n^\train \big(\frac{\sqrt{\Lambda} }{8} \delta\big)^2 }{16/L^4}   \bigg\} + 2\exp\bigg\{ -2 \frac{\tfrac{1}{2}\pi_y^\train n^\train \big(\frac{\sqrt{\Lambda} }{8} \delta\big)^2 }{16/L^4}   \bigg\} \\ 
	&\qquad + 4\sum_{j=0}^{m}\exp\bigg\{ -2 \frac{\tfrac{1}{2}\pi_j^\train n^\train \big(\tfrac{\sqrt{\Lambda} }{4}  \delta    \wedge   \frac{ \Lambda^{1/4} }{2} \sqrt{\delta} \big)^2}{16/L^4}\bigg\} \\ 
	&\qquad + 4\sum_{j=0}^{m}\exp\bigg\{ -2 \frac{\tfrac{1}{2}\pi_j^\train n^\train \big(   \tfrac{\sqrt{\Lambda}}{8} \wedge \frac{\Lambda^{1/4}}{\sqrt{8}}  \big)^2}{16/L^4}\bigg\} \\ \\ 
	&\leq  4\exp\bigg\{ -2 \frac{\tfrac{1}{2} \xi n^\train \big(\frac{\sqrt{\Lambda} }{8} \delta\big)^2 }{16/L^4}   \bigg\}  \\ 
	&\qquad + 4m\exp\bigg\{ -2 \frac{\tfrac{1}{2}\xi n^\train \big(\tfrac{\sqrt{\Lambda} }{4}  \delta    \wedge   \frac{ \Lambda^{1/4} }{2} \sqrt{\delta} \big)^2}{16/L^4}\bigg\} \\ 
	&\qquad + 4m\exp\bigg\{ -2 \frac{\tfrac{1}{2}\xi n^\train \big(   \tfrac{\sqrt{\Lambda}}{8} \wedge \frac{\Lambda^{1/4}}{\sqrt{8}}  \big)^2}{16/L^4}\bigg\} \\ \\ 
	&\leq  4m\exp\bigg\{ -2 \frac{\tfrac{1}{2}\xi n^\train \big(\tfrac{\sqrt{\Lambda} }{4}  \delta    \wedge   \frac{ \Lambda^{1/4} }{2} \sqrt{\delta} \big)^2}{16/L^4}\bigg\} \\ 
	&\qquad +( 4m+4)\exp\bigg\{ -2 \frac{\tfrac{1}{2}\xi n^\train \big(   \tfrac{\sqrt{\Lambda}}{8} \wedge \frac{\Lambda^{1/4}}{\sqrt{8}}  \big)^2}{16/L^4}\bigg\} \\ \\ 
	&\leq  (8m+4)\exp\bigg\{ -2 \frac{\tfrac{1}{2}\xi n^\train \big(\tfrac{\sqrt{\Lambda} }{8}  \delta    \wedge   \frac{ \Lambda^{1/4} }{8} \sqrt{\delta}     \wedge     \tfrac{\sqrt{\Lambda}}{8} \wedge \frac{\Lambda^{1/4}}{8}   \big)^2}{16/L^4}\bigg\} \\ 
	&=  4(2m+1)\exp\bigg\{ -\frac{ L^4 \xi  n^\train \big(\sqrt{\Lambda}   \delta    \wedge    \Lambda^{1/4}\sqrt{\delta}     \wedge    \sqrt{\Lambda}  \wedge \Lambda^{1/4}   \big)^2}{ 2^{10}}\bigg\}. 
\end{align*}

Therefore, we have that:
\begin{align*}
	&\P_{\DtrainII}\Bigg[ \absBigg{ \frac{\widehat{\E}_y^{\trainII}[s_{\gamma^\ast,y}] - \widehat{\E}_0^{\trainII}[s_{\gamma^\ast,y}]  }{\widehat{\Var}_{\gamma^\ast}^{\trainII}[s_{\gamma^\ast,y}] } - 1}  \geq \min\bigg\{ \frac{L}{24}a, \frac{1}{\sqrt{12  \lambda_y \Omega^2(s_{\gamma^\ast,y})} }\sqrt{a} \bigg\} \Bigg]\\
	  &\leq 4(2m+1)\exp\bigg\{ -\frac{ L^4 \xi  n^\train \tau(a) }{ 2^{10}}\bigg\},
\end{align*}

where 
\small
\begin{align*}
	\tau(a) &=  \Bigg(\sqrt{\Lambda}   \min\bigg\{ \frac{L}{24}a, \frac{1}{\sqrt{12  \lambda_y \Omega^2(s_{\gamma^\ast,y})} }\sqrt{a} \bigg\}     \wedge    \Lambda^{1/4}\sqrt{\min\bigg\{ \frac{L}{24}a, \frac{1}{\sqrt{12  \lambda_y \Omega^2(s_{\gamma^\ast,y})} }\sqrt{a} \bigg\} }     \wedge    \sqrt{\Lambda}  \wedge \Lambda^{1/4}   \Bigg)^2 \\ 
	&= \Lambda   \min\bigg\{ \frac{L^2}{24^2}a^2, \frac{1}{12  \lambda_y \Omega^2(s_{\gamma^\ast,y}) }a \bigg\}     \wedge    \Lambda^{1/2}\min\bigg\{ \frac{L}{24}a, \frac{1}{\sqrt{12  \lambda_y \Omega^2(s_{\gamma^\ast,y})} }\sqrt{a} \bigg\}     \wedge    \Lambda  \wedge \Lambda^{1/2}.
\end{align*}
\normalsize

Thus, combining all of our work on the tail bound, we have that:
\begin{align*}
	&\P_{\DtrainII}[T_y \geq a] \\
	&\leq  \P_{\DtrainII}\bigg[ \absBig{ \widehat{\E}_y^{\trainII}[s_{\gamma^\ast,y}] - \E_y[s_{\gamma^\ast,y}] } \geq \frac{1}{12}a\bigg]  + \P_{\DtrainII}\bigg[ \absBig{ \widehat{\E}_0^{\trainII}[s_{\gamma^\ast,y}] - \E_0[s_{\gamma^\ast,y}]  } \geq \frac{1}{12}a\bigg]  \\
	 &\qquad +  2\P_{\DtrainII}\Bigg[ \absBigg{ \frac{\widehat{\E}_y^{\trainII}[s_{\gamma^\ast,y}] - \widehat{\E}_0^{\trainII}[s_{\gamma^\ast,y}]  }{\widehat{\Var}_{\gamma^\ast}^{\trainII}[s_{\gamma^\ast,y}] } - 1}  \geq \min\bigg\{ \frac{L}{24}a, \frac{1}{\sqrt{12  \lambda_y \Omega^2(s_{\gamma^\ast,y})} }\sqrt{a} \bigg\} \Bigg]  \\
	 &\qquad +  \IBig{ \lambda_y \Omega^2(s_{\gamma^\ast,y})  \geq \frac{1}{12}a } \\
	 &\qquad +  \P_{\DtrainII}\Bigg[\sup_{\Omega(h_y) \leq \frac{1}{L \sqrt{\lambda_y}}} \absBig{\widehat{\E}_y^{\trainII}[h_y] - \E_y[h_y]}  \geq \frac{1}{6}a\Bigg] +  \P_{\DtrainII}\Bigg[\sup_{\Omega(h_y) \leq \frac{1}{L \sqrt{\lambda_y}}} \absBig{\widehat{\E}_0^{\trainII}[h_y] - \E_0[h_y]} \geq \frac{1}{6}a\Bigg] \\ \\
	 &\leq  4\exp\bigg\{ -2 \frac{\tfrac{1}{2}\xi n^\train \big(\tfrac{a}{12}\big)^2 }{4/L^2}   \bigg\}    +  8(2m+1)\exp\bigg\{ -\frac{ L^4 \xi  n^\train \tau(a) }{ 2^{10}}\bigg\} +  \IBig{ \lambda_y \Omega^2(s_{\gamma^\ast,y})  \geq \frac{1}{12}a } \\
	 &\qquad + \P_{\DtrainII}\Bigg[\sup_{\Omega(h_y) \leq \frac{1}{L \sqrt{\lambda_y}}} \absBig{\widehat{\E}_y^{\trainII}[h_y] - \E_y[h_y]}  \geq \frac{1}{6}a\Bigg] +  \P_{\DtrainII}\Bigg[\sup_{\Omega(h_y) \leq \frac{1}{L \sqrt{\lambda_y}}} \absBig{\widehat{\E}_0^{\trainII}[h_y] - \E_0[h_y]} \geq \frac{1}{6}a\Bigg] \\ \\ 
	 &=  4\exp\bigg\{ - \frac{L^2 \xi n^\train    }{576} a^2  \bigg\}  +  8(2m+1)\exp\bigg\{ -\frac{ L^4 \xi  n^\train  }{ 2^{10}} \tau(a) \bigg\} +  \IBig{ \lambda_y \Omega^2(s_{\gamma^\ast,y})  \geq \frac{1}{12}a } \\
	 &\qquad + \P_{\DtrainII}\Bigg[\sup_{\Omega(h_y) \leq \frac{1}{L \sqrt{\lambda_y}}} \absBig{\widehat{\E}_y^{\trainII}[h_y] - \E_y[h_y]}  \geq \frac{1}{6}a\Bigg] +  \P_{\DtrainII}\Bigg[\sup_{\Omega(h_y) \leq \frac{1}{L \sqrt{\lambda_y}}} \absBig{\widehat{\E}_0^{\trainII}[h_y] - \E_0[h_y]} \geq \frac{1}{6}a\Bigg], 
\end{align*}
as claimed!


\end{proof}
\vspace{0.4in}

\begin{proof}[\uline{Proof of Lemma \eqref{Rate for Learning the Score Function}}]

Consider any $y\in[m]$. Observe that:
\small
\begin{align}
	&\Var_{\gamma^\ast}[  \widehat{s}_{\gamma^\ast,y} - s_{\gamma^\ast,y}] \\
	&= \Var_{\gamma^\ast}[\widehat{s}_{\gamma^\ast,y}] + \Var_{\gamma^\ast}[s_{\gamma^\ast,y}]  -  2\Cov_{\gamma^\ast}[s_{\gamma^\ast,y},\widehat{s}_{\gamma^\ast,y}] \nonumber \\ 
	&= \Var_{\gamma^\ast}[\widehat{s}_{\gamma^\ast,y}] + \big(\E_y[s_{\gamma^\ast,y}] - \E_0[s_{\gamma^\ast,y}]\big)  -  2\big(\E_y[\widehat{s}_{\gamma^\ast,y}]-\E_0[\widehat{s}_{\gamma^\ast,y}]\big) \nonumber \\ \nonumber \\
	&= \Big[\Var_{\gamma^\ast}[\widehat{s}_{\gamma^\ast,y}]-\big(\E_y[\widehat{s}_{\gamma^\ast,y}]-\E_0[\widehat{s}_{\gamma^\ast,y}]\big)\Big] \nonumber \\
	&\qquad+ \Big[\big(\E_y[s_{\gamma^\ast,y}] - \E_0[s_{\gamma^\ast,y}]\big)  -  \big(\E_y[\widehat{s}_{\gamma^\ast,y}]-\E_0[\widehat{s}_{\gamma^\ast,y}]\big) \Big] \nonumber\\ \nonumber \\
	&= \Big[\Var_{\gamma^\ast}[\widehat{s}_{\gamma^\ast,y}] - \widehat{\Var}_{\gamma^\ast}^{\trainII}[\widehat{s}_{\gamma^\ast,y}]\Big] + \Big[\big(\widehat{\E}_y^{\trainII}[\widehat{s}_{\gamma^\ast,y}]-\widehat{\E}_0^{\trainII}[\widehat{s}_{\gamma^\ast,y}]\big) -\big(\E_y[\widehat{s}_{\gamma^\ast,y}]-\E_0[\widehat{s}_{\gamma^\ast,y}]\big)\Big] \nonumber \\
	&\qquad+ \Big[\big(\E_y[s_{\gamma^\ast,y}] - \E_0[s_{\gamma^\ast,y}]\big)  -  \big(\E_y[\widehat{s}_{\gamma^\ast,y}]-\E_0[\widehat{s}_{\gamma^\ast,y}]\big) \Big] \nonumber\\ \nonumber \\
	&\leq \sup_{\Omega(h_y) \leq  \frac{1}{L\sqrt{\lambda_y}} } \absBig{ \widehat{\Var}_{\gamma^\ast}^{\trainII}[h_y]  -  \Var_{\gamma^\ast}[h_y]} + \sup_{\Omega(h_y)\leq  \frac{1}{L\sqrt{\lambda_y}} } \absBig{ \widehat{\E}_y^{\trainII}[h_y]  -   \E_y[h_y]} + \sup_{\Omega(h_y)\leq  \frac{1}{L\sqrt{\lambda_y}} } \absBig{\widehat{\E}_0^{\trainII}[h_y] -\E_0[h_y] }\nonumber \\
	&\qquad+ \Big[\big(\E_y[s_{\gamma^\ast,y}] - \E_0[s_{\gamma^\ast,y}]\big)  -  \big(\E_y[\widehat{s}_{\gamma^\ast,y}]-\E_0[\widehat{s}_{\gamma^\ast,y}]\big) \Big]  \label{ruff}\\\nonumber \\ 
	&= \U_y(\lambda_y)  + \Big[\big(\E_y[s_{\gamma^\ast,y}] - \E_0[s_{\gamma^\ast,y}]\big)  -  \big(\E_y[\widehat{s}_{\gamma^\ast,y}]-\E_0[\widehat{s}_{\gamma^\ast,y}]\big) \Big]\nonumber \\ 
	&\leq \U_y(\lambda_y) + T_y,\nonumber
\end{align}
\normalsize
where the fourth line is due to the constraints of the procedure used to create $\widehat{s}_{\gamma^\ast,y}$, the fifth line uses Lemma \eqref{Bound on Norm of Score Estimate}, and the seventh line is by Lemma \eqref{Rate for Learning Diagonal of Fisher Information Matrix}. Ergo: 
\begin{align*}
	\implies \E_{\DtrainII}\Big[ \Var_{\gamma^\ast}[  \widehat{s}_{\gamma^\ast,y} - s_{\gamma^\ast,y}] \Big] &\leq  \E_{\DtrainII}[\U_y(\lambda_y)] +  \E_{\DtrainII}[T_y] = O\Big(\E_{\DtrainII}[T_y]\Big),
\end{align*}

as claimed. Finally, as for the  tail bound on $\Var_{\gamma^\ast}[  \widehat{s}_{\gamma^\ast,y} - s_{\gamma^\ast,y}]$, it follows from our work above that, for any $a > 0$:
\begin{align*}
	&\P_{\DtrainII}\Big[ \Var_{\gamma^\ast}[  \widehat{s}_{\gamma^\ast,y} - s_{\gamma^\ast,y}]  \geq a \Big]\\
	 &\leq \P_{\DtrainII}\bigg[ T_y \geq \frac{1}{4}a \bigg] +  \P_{\DtrainII}\Bigg[\sup_{\Omega(h_y) \leq  \frac{1}{L\sqrt{\lambda_y}} } \absBig{ \widehat{\Var}_{\gamma^\ast}^{\trainII}[h_y]  -  \Var_{\gamma^\ast}[h_y]} \geq \frac{1}{4}a \Bigg] \\
	&\qquad +  \P_{\DtrainII}\Bigg[  \sup_{\Omega(h_y)\leq  \frac{1}{L\sqrt{\lambda_y}} } \absBig{ \widehat{\E}_y^{\trainII}[h_y]  -   \E_y[h_y]} \geq \frac{1}{4}a \Bigg] \\ 
	&\qquad +  \P_{\DtrainII}\Bigg[  \sup_{\Omega(h_y)\leq  \frac{1}{L\sqrt{\lambda_y}} } \absBig{ \widehat{\E}_0^{\trainII}[h_y]  -   \E_0[h_y]} \geq \frac{1}{4}a \Bigg],
\end{align*}
as claimed!


\end{proof}
\vspace{0.4in}

\begin{proof}[\uline{Proof of Lemma \eqref{Rate for Learning Diagonal of Fisher Information Matrix, Unknown Gamma Star}}]

It follows from line \eqref{Ty - lambda times norm} in the proof of Lemma  \eqref{Rate for Learning Diagonal of Fisher Information Matrix} that:
\begin{align*}
	&\big(\E_y[s_{\gamma^\ast,y}] - \E_0[s_{\gamma^\ast,y}]\big)  -  \big(\E_y[\widehat{s}_{ \widehat{\gamma} ,y}]-\E_0[\widehat{s}_{ \widehat{\gamma} ,y}]\big) \\
	&= \Big[ \big(\E_y[s_{\gamma^\ast,y}] - \E_0[s_{\gamma^\ast,y}]\big)  -  \big(\E_y[\widehat{s}_{\gamma^\ast,y}]-\E_0[\widehat{s}_{\gamma^\ast,y}]\big) \Big] \\
	&\qquad+ \Big[\big(\E_y[\widehat{s}_{\gamma^\ast,y}]-\E_0[\widehat{s}_{\gamma^\ast,y}]\big) -  \big(\E_y[\widehat{s}_{ \widehat{\gamma} ,y}]-\E_0[\widehat{s}_{ \widehat{\gamma} ,y}]\big) \Big] \\ \\ 
	&\leq  T_y -  \lambda_y \Omega^2(\widehat{s}_{\gamma^\ast,y}) \\
	&\qquad+ \Big[\big(\E_y[\widehat{s}_{\gamma^\ast,y}]-\E_0[\widehat{s}_{\gamma^\ast,y}]\big) -  \big(\E_y[\widehat{s}_{ \widehat{\gamma} ,y}]-\E_0[\widehat{s}_{ \widehat{\gamma} ,y}]\big) \Big].
\end{align*}


Next, observe that:
\begin{align*}
	&\big(\E_y[\widehat{s}_{\gamma^\ast,y}]-\E_0[\widehat{s}_{\gamma^\ast,y}]\big) -  \big(\E_y[\widehat{s}_{ \widehat{\gamma} ,y}]-\E_0[\widehat{s}_{ \widehat{\gamma} ,y}]\big) \\
	&=\Big[\big(\E_y[\widehat{s}_{\gamma^\ast,y}]-\E_0[\widehat{s}_{\gamma^\ast,y}]\big) - \big( \widehat{\E}_y^\trainII[\widehat{s}_{\gamma^\ast,y}]-\widehat{\E}_0^\trainII[\widehat{s}_{\gamma^\ast,y}]\big)\Big] \\ 
	&\qquad+ \Big[\big( \widehat{\E}_y^\trainII[\widehat{s}_{\gamma^\ast,y}]-\widehat{\E}_0^\trainII[\widehat{s}_{\gamma^\ast,y}]\big) - \big( \widehat{\E}_y^\trainII[\widehat{s}_{\widehat{\gamma},y}]-\widehat{\E}_0^\trainII[\widehat{s}_{\widehat{\gamma},y}]\big)\Big] \\
	&\qquad+ \Big[  \big( \widehat{\E}_y^\trainII[\widehat{s}_{\widehat{\gamma},y}]-\widehat{\E}_0^\trainII[\widehat{s}_{\widehat{\gamma},y}]\big) - \big(\E_y[\widehat{s}_{ \widehat{\gamma} ,y}]-\E_0[\widehat{s}_{ \widehat{\gamma} ,y}]\big)\Big]  \\ \\ 
	&\leq 2\sup_{\Omega(h_y) \leq \frac{1}{L\sqrt{\lambda_y}}} \absbig{\widehat{\E}_y^{\trainII}[h_y] - \E_y[h_y]} + 2\sup_{\Omega(h_y) \leq \frac{1}{L\sqrt{\lambda_y}}} \absbig{\widehat{\E}_0^{\trainII}[h_y] - \E_y[h_y]} \\ 
	&\qquad+  \Big[\big( \widehat{\E}_y^\trainII[\widehat{s}_{\gamma^\ast,y}]-\widehat{\E}_0^\trainII[\widehat{s}_{\gamma^\ast,y}]\big) - \big( \widehat{\E}_y^\trainII[\widehat{s}_{\widehat{\gamma},y}]-\widehat{\E}_0^\trainII[\widehat{s}_{\widehat{\gamma},y}]\big)\Big],
\end{align*}
where the inequality follows from Lemma \eqref{Bound on Norm of Score Estimate} and the fact that Lemma \eqref{Bound on Norm of Score Estimate} also holds if we replace $\widehat{s}_{\gamma^\ast,y}$ with $\widehat{s}_{\widehat{\gamma},y}$, due to Assumption \eqref{gamma hat properties} (this is evident from a quick inspect of the proof of Lemma \eqref{Bound on Norm of Score Estimate}, wherein the only property of $\gamma^\ast$ that is used is the fact that the components are all lower bounded by $L$). Also observe that:
\small
\begin{align*}
	&\big( \widehat{\E}_y^\trainII[\widehat{s}_{\gamma^\ast,y}]-\widehat{\E}_0^\trainII[\widehat{s}_{\gamma^\ast,y}]\big) - \big( \widehat{\E}_y^\trainII[\widehat{s}_{\widehat{\gamma},y}]-\widehat{\E}_0^\trainII[\widehat{s}_{\widehat{\gamma},y}]\big) \\
	&\leq \big( \widehat{\E}_y^\trainII[\widehat{s}_{\gamma^\ast,y}]-\widehat{\E}_0^\trainII[\widehat{s}_{\gamma^\ast,y}]\big) - \big( \widehat{\E}_y^\trainII[\widehat{s}_{\widehat{\gamma},y}]-\widehat{\E}_0^\trainII[\widehat{s}_{\widehat{\gamma},y}] - \lambda_y\Omega(\widehat{s}_{\widehat{\gamma},y})^2\big) \\ \\ 
	&\leq \big( \widehat{\E}_y^\trainII[\widehat{s}_{\gamma^\ast,y}]-\widehat{\E}_0^\trainII[\widehat{s}_{\gamma^\ast,y}]\big) \\
	&\qquad - \frac{\widehat{\E}_y^\trainII[\widehat{s}_{\gamma^\ast,y}] - \widehat{\E}_0^\trainII[\widehat{s}_{\gamma^\ast,y}] }{\widehat{\Var_{\widehat{\gamma}}}^{\trainII}[\widehat{s}_{\gamma^\ast,y}] } \big( \widehat{\E}_y^\trainII[\widehat{s}_{\gamma^\ast,y}]-\widehat{\E}_0^\trainII[\widehat{s}_{\gamma^\ast,y}]\big)  \\ 
	&\qquad + \Bigg( \frac{\widehat{\E}_y^\trainII[\widehat{s}_{\gamma^\ast,y}] - \widehat{\E}_0^\trainII[\widehat{s}_{\gamma^\ast,y}] }{\widehat{\Var_{\widehat{\gamma}}}^{\trainII}[\widehat{s}_{\gamma^\ast,y}] } \Bigg)^2\lambda_y\Omega(\widehat{s}_{\gamma^\ast,y})^2 \\ \\ 
	&= \big( \widehat{\E}_y^\trainII[\widehat{s}_{\gamma^\ast,y}]-\widehat{\E}_0^\trainII[\widehat{s}_{\gamma^\ast,y}]\big)\Bigg[1-   \frac{\widehat{\E}_y^\trainII[\widehat{s}_{\gamma^\ast,y}] - \widehat{\E}_0^\trainII[\widehat{s}_{\gamma^\ast,y}] }{\widehat{\Var_{\widehat{\gamma}}}^{\trainII}[\widehat{s}_{\gamma^\ast,y}] }\Bigg] \\ 
	&\qquad + \Bigg( \frac{\widehat{\E}_y^\trainII[\widehat{s}_{\gamma^\ast,y}] - \widehat{\E}_0^\trainII[\widehat{s}_{\gamma^\ast,y}] }{\widehat{\Var_{\widehat{\gamma}}}^{\trainII}[\widehat{s}_{\gamma^\ast,y}] } \Bigg)^2\lambda_y\Omega(\widehat{s}_{\gamma^\ast,y})^2 \\ \\ 
	&=\widehat{\Var}_{\gamma^\ast}^\trainII[\widehat{s}_{\gamma^\ast,y}] \Bigg[1-   \frac{\widehat{\Var}_{\gamma^\ast}^\trainII[\widehat{s}_{\gamma^\ast,y}]  }{\widehat{\Var_{\widehat{\gamma}}}^{\trainII}[\widehat{s}_{\gamma^\ast,y}] }\Bigg]   + \Bigg( \frac{\widehat{\Var}_{\gamma^\ast}^\trainII[\widehat{s}_{\gamma^\ast,y}] }{\widehat{\Var_{\widehat{\gamma}}}^{\trainII}[\widehat{s}_{\gamma^\ast,y}] } \Bigg)^2\lambda_y\Omega(\widehat{s}_{\gamma^\ast,y})^2,
\end{align*}
\normalsize
where the second inequality is because of $\Omega(\cdot)$'s assumed properties and the fact that $ \frac{\widehat{\E}_y^\trainII[\widehat{s}_{\gamma^\ast,y}] - \widehat{\E}_0^\trainII[\widehat{s}_{\gamma^\ast,y}] }{\widehat{\Var_{\widehat{\gamma}}}^{\trainII}[\widehat{s}_{\gamma^\ast,y}] } \widehat{s}_{\gamma^\ast,y}$ is inside the feasible set of the optimization problem which produced $\widehat{s}_{\widehat{\gamma},y}$, and the last line uses the moment constraints of the optimization problem which produced $\widehat{s}_{\gamma^\ast,y}$. Combining all of our work in this proof so far, it follows that:
\begin{align}
	&\big(\E_y[s_{\gamma^\ast,y}] - \E_0[s_{\gamma^\ast,y}]\big)  -  \big(\E_y[\widehat{s}_{ \widehat{\gamma} ,y}]-\E_0[\widehat{s}_{ \widehat{\gamma} ,y}]\big)\\
	 &\leq  T_y + 2\sup_{\Omega(h_y) \leq \frac{1}{L\sqrt{\lambda_y}}} \absbig{\widehat{\E}_y^{\trainII}[h_y] - \E_y[h_y]} + 2\sup_{\Omega(h_y) \leq \frac{1}{L\sqrt{\lambda_y}}} \absbig{\widehat{\E}_0^{\trainII}[h_y] - \E_y[h_y]} \nonumber \\ 
	&\qquad+   \widehat{\Var}_{\gamma^\ast}^\trainII[\widehat{s}_{\gamma^\ast,y}]   \absBigg{1-   \frac{ \widehat{\Var}_{\gamma^\ast}^\trainII[\widehat{s}_{\gamma^\ast,y}] }{\widehat{\Var_{\widehat{\gamma}}}^{\trainII}[\widehat{s}_{\gamma^\ast,y}] } } \nonumber \\ 
	&\qquad + \lambda_y \Omega^2(\widehat{s}_{\gamma^\ast,y})  \absBigg{\Bigg( \frac{ \widehat{\Var}_{\gamma^\ast}^\trainII[\widehat{s}_{\gamma^\ast,y}]  }{\widehat{\Var_{\widehat{\gamma}}}^{\trainII}[\widehat{s}_{\gamma^\ast,y}] } \Bigg)^2   -  1 } \nonumber \\ \nonumber \\ 
	&\leq  T_y + 2\sup_{\Omega(h_y) \leq \frac{1}{L\sqrt{\lambda_y}}} \absbig{\widehat{\E}_y^{\trainII}[h_y] - \E_y[h_y]} + 2\sup_{\Omega(h_y) \leq \frac{1}{L\sqrt{\lambda_y}}} \absbig{\widehat{\E}_0^{\trainII}[h_y] - \E_y[h_y]} \nonumber\\ 
	&\qquad +   \widehat{\Var}_{\gamma^\ast}^\trainII[\widehat{s}_{\gamma^\ast,y}]   \absBigg{1-   \frac{ \widehat{\Var}_{\gamma^\ast}^\trainII[\widehat{s}_{\gamma^\ast,y}] }{\widehat{\Var_{\widehat{\gamma}}}^{\trainII}[\widehat{s}_{\gamma^\ast,y}] } }+ \frac{1}{L^2} \absBigg{ 1 -  \Bigg( \frac{ \widehat{\Var}_{\gamma^\ast}^\trainII[\widehat{s}_{\gamma^\ast,y}]  }{\widehat{\Var_{\widehat{\gamma}}}^{\trainII}[\widehat{s}_{\gamma^\ast,y}] } \Bigg)^2    } \nonumber\\ \nonumber \\ 
	&=  T_y + 2\sup_{\Omega(h_y) \leq \frac{1}{L\sqrt{\lambda_y}}} \absbig{\widehat{\E}_y^{\trainII}[h_y] - \E_y[h_y]} + 2\sup_{\Omega(h_y) \leq \frac{1}{L\sqrt{\lambda_y}}} \absbig{\widehat{\E}_0^{\trainII}[h_y] - \E_y[h_y]} \nonumber \\ 
	&\qquad +   \Bigg(\widehat{\Var}_{\gamma^\ast}^\trainII[\widehat{s}_{\gamma^\ast,y}]    + \frac{1}{L^2} \absBigg{ 1   +  \frac{ \widehat{\Var}_{\gamma^\ast}^\trainII[\widehat{s}_{\gamma^\ast,y}]  }{\widehat{\Var_{\widehat{\gamma}}}^{\trainII}[\widehat{s}_{\gamma^\ast,y}] }     } \Bigg)\absBigg{ 1 -   \frac{ \widehat{\Var}_{\gamma^\ast}^\trainII[\widehat{s}_{\gamma^\ast,y}]  }{\widehat{\Var_{\widehat{\gamma}}}^{\trainII}[\widehat{s}_{\gamma^\ast,y}] }     } \label{vitamin g},
\end{align}
where the second inequality follows from Lemma \eqref{Bound on Norm of Score Estimate}. Next, we bound $\widehat{\Var}_{\gamma^\ast}^\trainII[\widehat{s}_{\gamma^\ast,y}]$:
\begin{align}
	&\widehat{\Var}_{\gamma^\ast}^\trainII[\widehat{s}_{\gamma^\ast,y}] \\
	&= \widehat{\Var}_{\gamma^\ast}^\trainII[\widehat{s}_{\gamma^\ast,y}] - \Var_{\gamma^\ast}[\widehat{s}_{\gamma^\ast,y}]  + \Var_{\gamma^\ast}[\widehat{s}_{\gamma^\ast,y}] \nonumber \\ 
	&\leq \sup_{\Omega(h_y) \leq \frac{1}{L\sqrt{\lambda_y}} } \absbig{ \widehat{\Var}_{\gamma^\ast}^\trainII[h_y] - \Var_{\gamma^\ast}[h_y] }  + \Var_{\gamma^\ast}[\widehat{s}_{\gamma^\ast,y}] \nonumber \\ 
	&= \sup_{\Omega(h_y) \leq \frac{1}{L\sqrt{\lambda_y}} } \absbig{ \widehat{\Var}_{\gamma^\ast}^\trainII[h_y] - \Var_{\gamma^\ast}[h_y] }  + \Var_{\gamma^\ast}[s_{\gamma^\ast,y} + (\widehat{s}_{\gamma^\ast,y} - s_{\gamma^\ast,y})] \nonumber \\ 
	&= \sup_{\Omega(h_y) \leq \frac{1}{L\sqrt{\lambda_y}} } \absbig{ \widehat{\Var}_{\gamma^\ast}^\trainII[h_y] - \Var_{\gamma^\ast}[h_y] }  + \Var_{\gamma^\ast}[s_{\gamma^\ast,y}] + \Var_{\gamma^\ast}[\widehat{s}_{\gamma^\ast,y} - s_{\gamma^\ast,y}] + 2\Cov_{\gamma^\ast}[s_{\gamma^\ast,y}, \widehat{s}_{\gamma^\ast,y} - s_{\gamma^\ast,y}]  \nonumber \\ 
	&\leq \sup_{\Omega(h_y) \leq \frac{1}{L\sqrt{\lambda_y}} } \absbig{ \widehat{\Var}_{\gamma^\ast}^\trainII[h_y] - \Var_{\gamma^\ast}[h_y] }  + \frac{1}{L^2} + \Var_{\gamma^\ast}[\widehat{s}_{\gamma^\ast,y} - s_{\gamma^\ast,y}] +  \frac{2}{L} \sqrt{\Var_{\gamma^\ast}[\widehat{s}_{\gamma^\ast,y} - s_{\gamma^\ast,y}]}, \label{vitamin h}
\end{align}
where the last line is due to Popoviciu's variance inequality and the fact that $\abs{s_{\gamma^\ast,y}} \leq \frac{2}{L}$. Next, we will show that $ \frac{ \widehat{\Var}_{\gamma^\ast}^\trainII[\widehat{s}_{\gamma^\ast,y}]  }{\widehat{\Var}_{\widehat{\gamma}}^{\trainII}[\widehat{s}_{\gamma^\ast,y}] }$ cannot be too far from $1$. Towards that end, note that the proof of Lemma \eqref{Covariance Matrix of a Finite Mixture} can easily be extended to the case where the class mixture densities are replaced by their empirical counterparts. That is, let $\beta$ be any vector in the $(m+1)$-dimensional probability simplex,  let $f$ be any real-valued function, and define the row vector $\widehat{a}_f =: \big(\widehat{\E}_1^\trainII[f] - \widehat{\E}_0^\trainII[f], \dots, \widehat{\E}_m^\trainII[f] - \widehat{\E}_0^\trainII[f]  \big)$. Then:
\[
	\widehat{\Var}_\beta^{\trainII}[f] = \sum_{j=0}^m \beta_j \widehat{\Var}_j^\trainII[f] + \widehat{a}_f \FisherInfo{\beta;\text{Cat}}^{-1} \widehat{a}_f^T.
\]
Thus, if $\alpha$ is another vector in the $(m+1)$-dimensional probability simplex, we can write:
\[
	\frac{\widehat{\Var}_\beta^{\trainII}[f] }{\widehat{\Var}_\alpha^{\trainII}[f] } = \frac{\sum_{j=0}^m \beta_j \widehat{\Var}_j^\trainII[f] + \widehat{a}_f \FisherInfo{\beta;\text{Cat}}^{-1} \widehat{a}_f^T}{\sum_{j=0}^m \alpha_j \widehat{\Var}_j^\trainII[f] + \widehat{a}_f \FisherInfo{\alpha;\text{Cat}}^{-1} \widehat{a}_f^T}
\]
Now, a consequence of Lemma \eqref{Bounds on Eigenvalues of Categorical Fisher Info} is that both $\FisherInfo{\gamma^\ast;\text{Cat}}^{-1}$ and $\FisherInfo{\widehat{\gamma};\text{Cat}}^{-1}$ are positive definite. Thus, in the RHS above, both the numerator and denominator are sums of two positive real numbers. Ergo, by the mediant inequality, it follows that:
\begin{align*}
	&\min\Bigg\{  \frac{\sum_{j=0}^m \beta_j \widehat{\Var}_j^\trainII[f] }{\sum_{j=0}^m \alpha_j \widehat{\Var}_j^\trainII[f] }         \text{ },\text{ }       \frac{\widehat{a}_f \FisherInfo{\beta;\text{Cat}}^{-1} \widehat{a}_f^T}{ \widehat{a}_f \FisherInfo{\alpha;\text{Cat}}^{-1} \widehat{a}_f^T}  \Bigg\} \\
	&\leq \frac{\widehat{\Var}_\beta^{\trainII}[f] }{\widehat{\Var}_\alpha^{\trainII}[f] }  \leq \max\Bigg\{  \frac{\sum_{j=0}^m \beta_j \widehat{\Var}_j^\trainII[f] }{\sum_{j=0}^m \alpha_j \widehat{\Var}_j^\trainII[f] }         \text{ },\text{ }       \frac{\widehat{a}_f \FisherInfo{\beta;\text{Cat}}^{-1} \widehat{a}_f^T}{ \widehat{a}_f \FisherInfo{\alpha;\text{Cat}}^{-1} \widehat{a}_f^T}  \Bigg\}
\end{align*}
Further, since $\FisherInfo{\beta;\text{Cat}}^{-1}$ is clearly symmetric and $\FisherInfo{\alpha;\text{Cat}}^{-1}$ is positive definite, we know that:
\[
	\minEval{ \FisherInfo{\alpha;\text{Cat}}\FisherInfo{\beta;\text{Cat}}^{-1}} \leq  \frac{\widehat{a}_f \FisherInfo{\beta;\text{Cat}}^{-1} \widehat{a}_f^T}{ \widehat{a}_f \FisherInfo{\alpha;\text{Cat}}^{-1} \widehat{a}_f^T} \leq \maxEval{ \FisherInfo{\alpha;\text{Cat}}\FisherInfo{\beta;\text{Cat}}^{-1}}.
\]
Applying Lemma \eqref{Bounds on Eigenvalues of Inverse Cat FIM times Cat FIM, alpha vs. beta}, it follows that
\[
	\implies 1 -  (m+1)\frac{ \norm{\alpha - \beta}{1}  }{\min_j \alpha_j} \leq  \frac{\widehat{a}_f \FisherInfo{\beta;\text{Cat}}^{-1} \widehat{a}_f^T}{ \widehat{a}_f \FisherInfo{\alpha;\text{Cat}}^{-1} \widehat{a}_f^T} \leq 1 +  (m+1)\frac{ \norm{\alpha - \beta}{1}  }{\min_j \alpha_j}.
\]
By virtue of the generalized weighted mediant inequality \citep{GeneralizedWeightedMediantInequality}, we can also get the following lower and upper bounds on $\frac{\sum_{j=0}^m \beta_j \widehat{\Var}_j^\trainII[f] }{\sum_{j=0}^m \alpha_j \widehat{\Var}_j^\trainII[f] }$:
\[
	\min_j \frac{\beta_j}{\alpha_j}  \leq   \frac{\sum_{j=0}^m \beta_j \widehat{\Var}_j^\trainII[f] }{\sum_{j=0}^m \alpha_j \widehat{\Var}_j^\trainII[f] }    \leq \max_j \frac{\beta_j}{\alpha_j}
\]
\[
	\implies 1+ \min_j \frac{\beta_j-\alpha_j}{\alpha_j}   \leq   \frac{\sum_{j=0}^m \beta_j \widehat{\Var}_j^\trainII[f] }{\sum_{j=0}^m \alpha_j \widehat{\Var}_j^\trainII[f] }    \leq 1+ \max_j \frac{\beta_j-\alpha_j}{\alpha_j} 
\]
\[
	\implies 1-  \frac{\norm{\alpha - \beta}{1} }{\min_j \alpha_j}   \leq   \frac{\sum_{j=0}^m \beta_j \widehat{\Var}_j^\trainII[f] }{\sum_{j=0}^m \alpha_j \widehat{\Var}_j^\trainII[f] }    \leq 1+  \frac{\norm{\alpha-\beta}{1}}{\min_j \alpha_j}. 
\]
Ergo:
\[
	1 -  (m+1)\frac{ \norm{\alpha - \beta}{1}  }{\min_j \alpha_j}  \leq \frac{\widehat{\Var}_\beta^{\trainII}[f] }{\widehat{\Var}_\alpha^{\trainII}[f] }  \leq 1 +  (m+1)\frac{ \norm{\alpha - \beta}{1}  }{\min_j \alpha_j}.
\]
Thus, we have that
\[
	 \absBigg{1-\frac{\widehat{\Var}_\beta^{\trainII}[f] }{\widehat{\Var}_\alpha^{\trainII}[f] } }  \leq  (m+1)\frac{ \norm{\alpha - \beta}{1}  }{\min_j \alpha_j},
\]
and that
\[
	\absBigg{1 + \frac{\widehat{\Var}_\beta^{\trainII}[f] }{\widehat{\Var}_\alpha^{\trainII}[f] } } \leq 2 +  (m+1)\frac{ \norm{\alpha - \beta}{1}  }{\min_j \alpha_j}.
\]

Since these last two displays are true for all real-valued functions $f$ and probability vectors $\alpha,\beta$, we may apply them to $f = \widehat{s}_{\gamma^\ast}$, $\alpha = \widehat{\gamma}$, and $\beta = \gamma^\ast$. Doing so yields
\[
	 \absBigg{1-\frac{\widehat{\Var}_{\gamma^\ast}^{\trainII}[\widehat{s}_{\gamma^\ast}] }{\widehat{\Var}_{\widehat{\gamma}}^{\trainII}[\widehat{s}_{\gamma^\ast}] } }  \leq  (m+1)\frac{ \norm{\widehat{\gamma} - \gamma^\ast}{1}  }{\min_j \widehat{\gamma}_j}  \leq   \frac{ m+1 }{L} \norm{\widehat{\gamma} - \gamma^\ast}{1},
\]
and
\[
	 \absBigg{1+\frac{\widehat{\Var}_{\gamma^\ast}^{\trainII}[\widehat{s}_{\gamma^\ast}] }{\widehat{\Var}_{\widehat{\gamma}}^{\trainII}[\widehat{s}_{\gamma^\ast}] } } \leq  2 +  \frac{ m+1 }{L} \norm{\widehat{\gamma} - \gamma^\ast}{1}.
\]

Combined these results with the upper bounds in line \eqref{vitamin g} and \eqref{vitamin h}, it follows that:
\small
\[
	\big(\E_y[s_{\gamma^\ast,y}] - \E_0[s_{\gamma^\ast,y}]\big)  -  \big(\E_y[\widehat{s}_{ \widehat{\gamma} ,y}]-\E_0[\widehat{s}_{ \widehat{\gamma} ,y}]\big)
\]

\[
	\leq   T_y + 2\sup_{\Omega(h_y) \leq \frac{1}{L\sqrt{\lambda_y}}} \absbig{\widehat{\E}_y^{\trainII}[h_y] - \E_y[h_y]} + 2\sup_{\Omega(h_y) \leq \frac{1}{L\sqrt{\lambda_y}}} \absbig{\widehat{\E}_0^{\trainII}[h_y] - \E_y[h_y]} 
\]
\[
	 +  \frac{ m+1 }{L} \norm{\widehat{\gamma} - \gamma^\ast}{1} \Bigg( \frac{3}{L^2} + \sup_{\Omega(h_y) \leq \frac{1}{L\sqrt{\lambda_y}} } \absbig{ \widehat{\Var}_{\gamma^\ast}^\trainII[h_y] - \Var_{\gamma^\ast}[h_y] }  +\Var_{\gamma^\ast}[\widehat{s}_{\gamma^\ast,y} - s_{\gamma^\ast,y}] +  \frac{2}{L} \sqrt{\Var_{\gamma^\ast}[\widehat{s}_{\gamma^\ast,y} - s_{\gamma^\ast,y}]}    +  \frac{ m+1 }{L^3} \norm{\widehat{\gamma} - \gamma^\ast}{1}\Bigg)
\]
\vspace{0.1in}

\[
	\leq   T_y + 2\sup_{\Omega(h_y) \leq \frac{1}{L\sqrt{\lambda_y}}} \absbig{\widehat{\E}_y^{\trainII}[h_y] - \E_y[h_y]} + 2\sup_{\Omega(h_y) \leq \frac{1}{L\sqrt{\lambda_y}}} \absbig{\widehat{\E}_0^{\trainII}[h_y] - \E_y[h_y]} 
\]
\[
	 +  \frac{ m+1 }{L} \norm{\widehat{\gamma} - \gamma^\ast}{1} \Bigg( \frac{3 +  2(m+1)}{L^3} + \sup_{\Omega(h_y) \leq \frac{1}{L\sqrt{\lambda_y}} } \absbig{ \widehat{\Var}_{\gamma^\ast}^\trainII[h_y] - \Var_{\gamma^\ast}[h_y] }  +\Var_{\gamma^\ast}[\widehat{s}_{\gamma^\ast,y} - s_{\gamma^\ast,y}] +  \frac{2}{L} \sqrt{\Var_{\gamma^\ast}[\widehat{s}_{\gamma^\ast,y} - s_{\gamma^\ast,y}]}    \Bigg)
\]
\vspace{0.1in}

\[
	\leq   T_y + 2\sup_{\Omega(h_y) \leq \frac{1}{L\sqrt{\lambda_y}}} \absbig{\widehat{\E}_y^{\trainII}[h_y] - \E_y[h_y]} + 2\sup_{\Omega(h_y) \leq \frac{1}{L\sqrt{\lambda_y}}} \absbig{\widehat{\E}_0^{\trainII}[h_y] - \E_y[h_y]}  + 2\frac{ m+1 }{L}\sup_{\Omega(h_y) \leq \frac{1}{L\sqrt{\lambda_y}} } \absbig{ \widehat{\Var}_{\gamma^\ast}^\trainII[h_y] - \Var_{\gamma^\ast}[h_y] } 
\]
\[
	 +  \frac{3(m+1) +  2(m+1)^2}{L^4} \norm{\widehat{\gamma} - \gamma^\ast}{1}    +  \frac{ m+1 }{L} \norm{\widehat{\gamma} - \gamma^\ast}{1}\Bigg( \Var_{\gamma^\ast}[\widehat{s}_{\gamma^\ast,y} - s_{\gamma^\ast,y}] +  \frac{2}{L} \sqrt{\Var_{\gamma^\ast}[\widehat{s}_{\gamma^\ast,y} - s_{\gamma^\ast,y}]}    \Bigg).
\]
\vspace{0.1in}

\[
	\leq   T_y + 2\sup_{\Omega(h_y) \leq \frac{1}{L\sqrt{\lambda_y}}} \absbig{\widehat{\E}_y^{\trainII}[h_y] - \E_y[h_y]} + 2\sup_{\Omega(h_y) \leq \frac{1}{L\sqrt{\lambda_y}}} \absbig{\widehat{\E}_0^{\trainII}[h_y] - \E_y[h_y]}  + 2\frac{ m+1 }{L}\sup_{\Omega(h_y) \leq \frac{1}{L\sqrt{\lambda_y}} } \absbig{ \widehat{\Var}_{\gamma^\ast}^\trainII[h_y] - \Var_{\gamma^\ast}[h_y] } 
\]
\[
	 +  \frac{3(m+1) +  2(m+1)^2}{L^4} \norm{\widehat{\gamma} - \gamma^\ast}{1}    +  2\frac{ m+1 }{L}  \Var_{\gamma^\ast}[\widehat{s}_{\gamma^\ast,y} - s_{\gamma^\ast,y}] +  2\frac{ m+1 }{L^2} \norm{\widehat{\gamma} - \gamma^\ast}{1} \sqrt{\Var_{\gamma^\ast}[\widehat{s}_{\gamma^\ast,y} - s_{\gamma^\ast,y}]}.
\]
\vspace{0.1in}

\[
	=: \widetilde{T}_y.
\]

Note that:
\begin{align*}
	&\E_{\allDII}[\widetilde{T}_y] \\
	&\leq \E_{\DtrainII}[T_y] + 2\frac{m+1}{L}\E_{\DtrainII}[\U_y(\lambda_y)] +  \frac{3(m+1) +  2(m+1)^2}{L^4} \E_{\allDII}\Big[\norm{\widehat{\gamma} - \gamma^\ast}{1}\Big] \\ 
	&\qquad +  2\frac{ m+1 }{L} \E_{\DtrainII}\Big[\Var_{\gamma^\ast}[\widehat{s}_{\gamma^\ast,y} - s_{\gamma^\ast,y}]\Big] +  2\frac{ m+1 }{L^2} \E_{\allDII}\bigg[\norm{\widehat{\gamma} - \gamma^\ast}{1} \sqrt{\Var_{\gamma^\ast}[\widehat{s}_{\gamma^\ast,y} - s_{\gamma^\ast,y}]}\bigg] \\ \\ 
	&\leq \E_{\DtrainII}[T_y] + 2\frac{m+1}{L}\E_{\DtrainII}[\U_y(\lambda_y)] +  \frac{3(m+1) +  2(m+1)^2}{L^4} \sqrt{\E_{\allDII}\Big[\norm{\widehat{\gamma} - \gamma^\ast}{1}^2\Big]} \\ 
	&\qquad +  2\frac{ m+1 }{L} \E_{\DtrainII}\Big[\Var_{\gamma^\ast}[\widehat{s}_{\gamma^\ast,y} - s_{\gamma^\ast,y}]\Big] +  2\frac{ m+1 }{L^2} \sqrt{\E_{\allDII}\Big[\norm{\widehat{\gamma} - \gamma^\ast}{1}^2
	\Big]  \E_{\DtrainII}\Big[\Var_{\gamma^\ast}[\widehat{s}_{\gamma^\ast,y} - s_{\gamma^\ast,y}]\Big] } \\ \\
	&\leq \E_{\DtrainII}[T_y] + 2\frac{m+1}{L}\E_{\DtrainII}[\U_y(\lambda_y)] +  \frac{3(m+1) +  2(m+1)^2}{L^4} \sqrt{m\E_{\allDII}\Big[\norm{\widehat{\gamma} - \gamma^\ast}{2}^2\Big]} \\ 
	&\qquad +  2\frac{ m+1 }{L} \E_{\DtrainII}\Big[\Var_{\gamma^\ast}[\widehat{s}_{\gamma^\ast,y} - s_{\gamma^\ast,y}]\Big] +  2\frac{ m+1 }{L^2} \sqrt{m\E_{\allDII}\Big[\norm{\widehat{\gamma} - \gamma^\ast}{2}^2
	\Big]  \E_{\DtrainII}\Big[\Var_{\gamma^\ast}[\widehat{s}_{\gamma^\ast,y} - s_{\gamma^\ast,y}]\Big] } \\ \\ 
	&= O\Bigg(\E_{\DtrainII}[T_y] + \E_{\DtrainII}[\U_y(\lambda_y)]  + \sqrt{\E_{\allDII}\Big[\norm{\widehat{\gamma} - \gamma^\ast}{2}^2\Big]} +  \E_{\DtrainII}\Big[\Var_{\gamma^\ast}[\widehat{s}_{\gamma^\ast,y} - s_{\gamma^\ast,y}]\Big] \Bigg),
\end{align*}

as desired. Next, we obtain a tail bound on $\widetilde{T}_y$. Using the definition of $\widetilde{T}_y$, it follows for any $a > 0$ that:

\begin{align*}
	&\P_{\allDII}[\widetilde{T}_y \geq a] \\
	&\leq \P_{\DtrainII}[T_y \geq a/7] + \P_{\DtrainII}\Bigg[ \sup_{\Omega(h_y) \leq \frac{1}{L\sqrt{\lambda_y}}} \absbig{\widehat{\E}_y^{\trainII}[h_y] - \E_y[h_y]} \geq \frac{1}{14}a \Bigg] + \P_{\DtrainII}\Bigg[ \sup_{\Omega(h_y) \leq \frac{1}{L\sqrt{\lambda_y}}} \absbig{\widehat{\E}_0^{\trainII}[h_y] - \E_0[h_y]} \geq \frac{1}{14}a \Bigg] \\ 
	&\qquad + \P_{\DtrainII}\Bigg[ \sup_{\Omega(h_y) \leq \frac{1}{L\sqrt{\lambda_y}} } \absbig{ \widehat{\Var}_{\gamma^\ast}^\trainII[h_y] - \Var_{\gamma^\ast}[h_y] } \geq \frac{1}{14}\frac{L}{m+1}a \Bigg] + \P_{\allDII}\Bigg[  \norm{\widehat{\gamma} - \gamma^\ast}{1} \geq  \frac{L^4}{21(m+1) +  14(m+1)^2} a \Bigg] \\ 
	&\qquad +  \P_{\DtrainII}\Bigg[  \Var_{\gamma^\ast}[\widehat{s}_{\gamma^\ast,y} - s_{\gamma^\ast,y}]  \geq \frac{1}{14}\frac{L}{m+1}a \Bigg] + \P_{\DtrainII}\Bigg[\norm{\widehat{\gamma} - \gamma^\ast}{1} \sqrt{\Var_{\gamma^\ast}[\widehat{s}_{\gamma^\ast,y} - s_{\gamma^\ast,y}]} \geq \frac{1}{14}\frac{L^2}{m+1}a \Bigg] \\ \\ 
	&\leq \P_{\DtrainII}[T_y \geq a/7] + \P_{\DtrainII}\Bigg[ \sup_{\Omega(h_y) \leq \frac{1}{L\sqrt{\lambda_y}}} \absbig{\widehat{\E}_y^{\trainII}[h_y] - \E_y[h_y]} \geq \frac{1}{14}a \Bigg] + \P_{\DtrainII}\Bigg[ \sup_{\Omega(h_y) \leq \frac{1}{L\sqrt{\lambda_y}}} \absbig{\widehat{\E}_0^{\trainII}[h_y] - \E_0[h_y]} \geq \frac{1}{14}a \Bigg] \\ 
	&\qquad + \P_{\DtrainII}\Bigg[ \sup_{\Omega(h_y) \leq \frac{1}{L\sqrt{\lambda_y}} } \absbig{ \widehat{\Var}_{\gamma^\ast}^\trainII[h_y] - \Var_{\gamma^\ast}[h_y] } \geq \frac{1}{14}\frac{L}{m+1}a \Bigg] + \P_{\allDII}\Bigg[  \norm{\widehat{\gamma} - \gamma^\ast}{1} \geq  \frac{L^4}{21(m+1) +  14(m+1)^2} a \Bigg] \\ 
	&\qquad +  \P_{\DtrainII}\Bigg[  \Var_{\gamma^\ast}[\widehat{s}_{\gamma^\ast,y} - s_{\gamma^\ast,y}]  \geq \frac{1}{14}\frac{L}{m+1}a \Bigg] + \P_{\DtrainII}\Bigg[ \norm{\widehat{\gamma} - \gamma^\ast}{1}  \geq \sqrt{\frac{1}{14}\frac{L^2}{m+1}a} \Bigg] \\
	&\qquad + \P_{\DtrainII}\Bigg[ \Var_{\gamma^\ast}[\widehat{s}_{\gamma^\ast,y} - s_{\gamma^\ast,y}] \geq \frac{1}{14}\frac{L^2}{m+1}a \Bigg] \\ \\ 
	&\leq \P_{\DtrainII}[T_y \geq a/7] + \P_{\DtrainII}\Bigg[ \sup_{\Omega(h_y) \leq \frac{1}{L\sqrt{\lambda_y}}} \absbig{\widehat{\E}_y^{\trainII}[h_y] - \E_y[h_y]} \geq \frac{1}{14}a \Bigg] + \P_{\DtrainII}\Bigg[ \sup_{\Omega(h_y) \leq \frac{1}{L\sqrt{\lambda_y}}} \absbig{\widehat{\E}_0^{\trainII}[h_y] - \E_0[h_y]} \geq \frac{1}{14}a \Bigg] \\ 
	&\qquad + \P_{\DtrainII}\Bigg[ \sup_{\Omega(h_y) \leq \frac{1}{L\sqrt{\lambda_y}} } \absbig{ \widehat{\Var}_{\gamma^\ast}^\trainII[h_y] - \Var_{\gamma^\ast}[h_y] } \geq \frac{1}{14}\frac{L}{m+1}a \Bigg] + 2\P_{\allDII}\Bigg[  \norm{\widehat{\gamma} - \gamma^\ast}{2} \geq \widetilde{\tau}_1(a) \Bigg] \\ 
	&\qquad +  2\P_{\DtrainII}\Bigg[  \Var_{\gamma^\ast}[\widehat{s}_{\gamma^\ast,y} - s_{\gamma^\ast,y}]  \geq \frac{1}{14}\frac{L^2}{m+1}a \Bigg], 
\end{align*}
\normalsize
as desired!
\end{proof}
\vspace{0.4in}

\begin{proof}[\uline{Proof of Lemma \eqref{Rate for Learning the Score Function, Unknown Gamma Star}}]

Observe that:
\begin{align*}
	&\Var_{\gamma^\ast}[ \widehat{s}_{\widehat{\gamma},y} - s_{\gamma^\ast,y} ] \\
	&= \Var_{\gamma^\ast}[\widehat{s}_{\widehat{\gamma},y}] + \Var_{\gamma^\ast}[s_{\gamma^\ast,y}] - 2\Cov_{\gamma^\ast}[\widehat{s}_{\widehat{\gamma},y}, s_{\gamma^\ast,y}] \\ 
	&= \Var_{\gamma^\ast}[\widehat{s}_{\widehat{\gamma},y}] + \big(\E_{y}[s_{\gamma^\ast,y}] - \E_0[s_{\gamma^\ast,y}] \big) - 2\big(\E_y[\widehat{s}_{\widehat{\gamma},y}] - \E_0[\widehat{s}_{\widehat{\gamma},y}]\big)  \\ 
	&= \Var_{\gamma^\ast}[\widehat{s}_{\widehat{\gamma},y}] - \big(\E_y[\widehat{s}_{\widehat{\gamma},y}] - \E_0[\widehat{s}_{\widehat{\gamma},y}]\big)\\
	&\qquad + \big(\E_{y}[s_{\gamma^\ast,y}] - \E_0[s_{\gamma^\ast,y}] \big) - \big(\E_y[\widehat{s}_{\widehat{\gamma},y}] - \E_0[\widehat{s}_{\widehat{\gamma},y}]\big)  \\ \\
	&= \Big[\Var_{\gamma^\ast}[\widehat{s}_{\widehat{\gamma},y}] - \widehat{\Var}_{\widehat{\gamma}}^\trainII[\widehat{s}_{\widehat{\gamma},y}]\Big] + \Big[ \big(\widehat{\E}_y^{\trainII}[\widehat{s}_{\widehat{\gamma},y}] - \widehat{\E}_0^{\trainII}[\widehat{s}_{\widehat{\gamma},y}]\big) - \big(\E_y[\widehat{s}_{\widehat{\gamma},y}] - \E_0[\widehat{s}_{\widehat{\gamma},y}]\big) \Big] \\
	&\qquad + \big(\E_{y}[s_{\gamma^\ast,y}] - \E_0[s_{\gamma^\ast,y}] \big) - \big(\E_y[\widehat{s}_{\widehat{\gamma},y}] - \E_0[\widehat{s}_{\widehat{\gamma},y}]\big)  \\ \\
	&\leq \Big[\Var_{\gamma^\ast}[\widehat{s}_{\widehat{\gamma},y}] - \widehat{\Var}_{\widehat{\gamma}}^\trainII[\widehat{s}_{\widehat{\gamma},y}]\Big] + \sup_{\Omega(h_y) \leq \frac{1}{L\sqrt{\lambda_y}} } \absBig{\widehat{\E}_y^{\trainII}[h_y] - \E_y[h_y]}  + \sup_{\Omega(h_y) \leq \frac{1}{L\sqrt{\lambda_y}} } \absBig{\widehat{\E}_0^{\trainII}[h_y] - \E_0[h_y] } \\
	&\qquad + \widetilde{T}_y,
\end{align*}
where the last inequality uses Lemma \eqref{Rate for Learning Diagonal of Fisher Information Matrix, Unknown Gamma Star}. Next, let's obtain an upper bound on $\Var_{\gamma^\ast}[\widehat{s}_{\widehat{\gamma},y}] - \widehat{\Var}_{\widehat{\gamma}}^\trainII[\widehat{s}_{\widehat{\gamma},y}]$. Using an argument very similar to the one employed in Lemma \eqref{Rate for Learning Diagonal of Fisher Information Matrix, Unknown Gamma Star}, one can show that
\[
	\frac{\widehat{\Var}_{\widehat{\gamma}}^\trainII[\widehat{s}_{\widehat{\gamma},y}]}{\widehat{\Var}_{\gamma^\ast}^\trainII[\widehat{s}_{\widehat{\gamma},y}]} \geq 1 - \frac{m+1}{L}\norm{\widehat{\gamma} - \gamma^\ast}{1},
\]
which implies that
\begin{align*}
	 & \Var_{\gamma^\ast}[\widehat{s}_{\widehat{\gamma},y}] - \widehat{\Var}_{\widehat{\gamma}}^\trainII[\widehat{s}_{\widehat{\gamma},y}] \\
	 &=  \Var_{\gamma^\ast}[\widehat{s}_{\widehat{\gamma},y}] - \widehat{\Var}_{\gamma^\ast}^\trainII[\widehat{s}_{\widehat{\gamma},y}] \frac{\widehat{\Var}_{\widehat{\gamma}}^\trainII[\widehat{s}_{\widehat{\gamma},y}]}{\widehat{\Var}_{\gamma^\ast}^\trainII[\widehat{s}_{\widehat{\gamma},y}]} \\
	 &\leq   \Var_{\gamma^\ast}[\widehat{s}_{\widehat{\gamma},y}] - \widehat{\Var}_{\gamma^\ast}^\trainII[\widehat{s}_{\widehat{\gamma},y}] + \frac{m+1}{L}\norm{\widehat{\gamma} - \gamma^\ast}{1}\widehat{\Var}_{\gamma^\ast}^\trainII[\widehat{s}_{\widehat{\gamma},y}] \\ 
	 &\leq   \sup_{\Omega(h_y) \leq \frac{1}{L\sqrt{\lambda_y}}} \absBig{\widehat{\Var}_{\gamma^\ast}^\trainII[h_y]  -   \Var_{\gamma^\ast}[h_y] } + \frac{m+1}{L}\norm{\widehat{\gamma} - \gamma^\ast}{1}\widehat{\Var}_{\gamma^\ast}^\trainII[\widehat{s}_{\widehat{\gamma},y}].
\end{align*}

Next, we will get a bound on $\widehat{\Var}_{\gamma^\ast}^\trainII[\widehat{s}_{\widehat{\gamma},y}]$. Observe that:
\small
\begin{align*}
	&\widehat{\Var}_{\gamma^\ast}^\trainII[\widehat{s}_{\widehat{\gamma},y}] \\
	&= \Bigg(\frac{ \widehat{\Var}_{\gamma^\ast}^{\trainII}[ \widehat{s}_{\widehat{\gamma},y}] }{\widehat{\E}_y^\trainII[ \widehat{s}_{\widehat{\gamma},y} ] - \widehat{\E}_0^\trainII[ \widehat{s}_{\widehat{\gamma},y}]  } \Bigg)^2\widehat{\Var}_{\gamma^\ast}^\trainII\Bigg[ \frac{\widehat{\E}_y^\trainII[ \widehat{s}_{\widehat{\gamma},y} ] - \widehat{\E}_0^\trainII[ \widehat{s}_{\widehat{\gamma},y}]  }{ \widehat{\Var}_{\gamma^\ast}^{\trainII}[ \widehat{s}_{\widehat{\gamma},y}] } \widehat{s}_{\widehat{\gamma},y} \Bigg] \\ \\
	 &= \Bigg(\frac{ \widehat{\Var}_{\gamma^\ast}^{\trainII}[ \widehat{s}_{\widehat{\gamma},y}] }{\widehat{\E}_y^\trainII[ \widehat{s}_{\widehat{\gamma},y} ] - \widehat{\E}_0^\trainII[ \widehat{s}_{\widehat{\gamma},y}]  } \Bigg)^2 \Bigg\{ \widehat{\Var}_{\gamma^\ast}^\trainII\Bigg[ \frac{\widehat{\E}_y^\trainII[ \widehat{s}_{\widehat{\gamma},y} ] - \widehat{\E}_0^\trainII[ \widehat{s}_{\widehat{\gamma},y}]  }{ \widehat{\Var}_{\gamma^\ast}^{\trainII}[ \widehat{s}_{\widehat{\gamma},y}] } \widehat{s}_{\widehat{\gamma},y} \Bigg] - \lambda_y\Omega\Bigg( \frac{\widehat{\E}_y^\trainII[ \widehat{s}_{\widehat{\gamma},y} ] - \widehat{\E}_0^\trainII[ \widehat{s}_{\widehat{\gamma},y}]  }{ \widehat{\Var}_{\gamma^\ast}^{\trainII}[ \widehat{s}_{\widehat{\gamma},y}] } \widehat{s}_{\widehat{\gamma},y}\Bigg)^2 \Bigg\} \\ 
	 &\qquad + \lambda_y\Bigg(\frac{ \widehat{\Var}_{\gamma^\ast}^{\trainII}[ \widehat{s}_{\widehat{\gamma},y}] }{\widehat{\E}_y^\trainII[ \widehat{s}_{\widehat{\gamma},y} ] - \widehat{\E}_0^\trainII[ \widehat{s}_{\widehat{\gamma},y}]  } \Bigg)^2 \Omega\Bigg( \frac{\widehat{\E}_y^\trainII[ \widehat{s}_{\widehat{\gamma},y} ] - \widehat{\E}_0^\trainII[ \widehat{s}_{\widehat{\gamma},y}]  }{ \widehat{\Var}_{\gamma^\ast}^{\trainII}[ \widehat{s}_{\widehat{\gamma},y}] } \widehat{s}_{\widehat{\gamma},y}\Bigg)^2 \\ \\ 
	 &\leq \Bigg(\frac{ \widehat{\Var}_{\gamma^\ast}^{\trainII}[ \widehat{s}_{\widehat{\gamma},y}] }{\widehat{\E}_y^\trainII[ \widehat{s}_{\widehat{\gamma},y} ] - \widehat{\E}_0^\trainII[ \widehat{s}_{\widehat{\gamma},y}]  } \Bigg)^2 \Bigg\{ \widehat{\Var}_{\gamma^\ast}^\trainII[\widehat{s}_{\gamma^\ast,y}] - \lambda_y \Omega(\widehat{s}_{\gamma^\ast,y})^2 \Bigg\}  + \lambda_y\Omega(\widehat{s}_{\widehat{\gamma},y})^2 \\ 
	 &\leq \Bigg(\frac{ \widehat{\Var}_{\gamma^\ast}^{\trainII}[ \widehat{s}_{\widehat{\gamma},y}] }{\widehat{\E}_y^\trainII[ \widehat{s}_{\widehat{\gamma},y} ] - \widehat{\E}_0^\trainII[ \widehat{s}_{\widehat{\gamma},y}]  } \Bigg)^2 \widehat{\Var}_{\gamma^\ast}^\trainII[\widehat{s}_{\gamma^\ast,y}]   + \frac{1}{L^2} \\ 
	 &= \Bigg(\frac{ \widehat{\Var}_{\gamma^\ast}^{\trainII}[ \widehat{s}_{\widehat{\gamma},y}] }{ \widehat{\Var}_{\widehat{\gamma}}^\trainII[\widehat{s}_{\widehat{\gamma},y}]  } \Bigg)^2 \widehat{\Var}_{\gamma^\ast}^\trainII[\widehat{s}_{\gamma^\ast,y}]   + \frac{1}{L^2} \\ 
	 &= \Bigg[\Bigg(\frac{ \widehat{\Var}_{\gamma^\ast}^{\trainII}[ \widehat{s}_{\widehat{\gamma},y}] }{ \widehat{\Var}_{\widehat{\gamma}}^\trainII[\widehat{s}_{\widehat{\gamma},y}]  } \Bigg)^2 -1 \Bigg]\widehat{\Var}_{\gamma^\ast}^\trainII[\widehat{s}_{\gamma^\ast,y}] + \widehat{\Var}_{\gamma^\ast}^\trainII[\widehat{s}_{\gamma^\ast,y}]   + \frac{1}{L^2} \\ 
	 &\leq \absBigg{\Bigg(\frac{ \widehat{\Var}_{\gamma^\ast}^{\trainII}[ \widehat{s}_{\widehat{\gamma},y}] }{ \widehat{\Var}_{\widehat{\gamma}}^\trainII[\widehat{s}_{\widehat{\gamma},y}]  } \Bigg) -1} \absBigg{\Bigg(\frac{ \widehat{\Var}_{\gamma^\ast}^{\trainII}[ \widehat{s}_{\widehat{\gamma},y}] }{ \widehat{\Var}_{\widehat{\gamma}}^\trainII[\widehat{s}_{\widehat{\gamma},y}]  } \Bigg) + 1 }\widehat{\Var}_{\gamma^\ast}^\trainII[\widehat{s}_{\gamma^\ast,y}] + \widehat{\Var}_{\gamma^\ast}^\trainII[\widehat{s}_{\gamma^\ast,y}]   + \frac{1}{L^2} \\ 
	 &\leq \frac{ m+1 }{L} \norm{\widehat{\gamma} - \gamma^\ast}{1}\Big(2 + \frac{ m+1 }{L} \norm{\widehat{\gamma} - \gamma^\ast}{1} \Big)\widehat{\Var}_{\gamma^\ast}^\trainII[\widehat{s}_{\gamma^\ast,y}] + \widehat{\Var}_{\gamma^\ast}^\trainII[\widehat{s}_{\gamma^\ast,y}]   + \frac{1}{L^2}  \\ 
	 &\leq 2\frac{ m+1 }{L} \Big(2 + 2\frac{ m+1 }{L}  \Big)\widehat{\Var}_{\gamma^\ast}^\trainII[\widehat{s}_{\gamma^\ast,y}] + \widehat{\Var}_{\gamma^\ast}^\trainII[\widehat{s}_{\gamma^\ast,y}]   + \frac{1}{L^2}  \\ 
	 &= 4\frac{ m+1 }{L} \Big(1 + \frac{ m+1 }{L}  \Big)\widehat{\Var}_{\gamma^\ast}^\trainII[\widehat{s}_{\gamma^\ast,y}] + \widehat{\Var}_{\gamma^\ast}^\trainII[\widehat{s}_{\gamma^\ast,y}]   + \frac{1}{L^2}  \\
	 &= \bigg[4\frac{ m+1 }{L} \Big(1 + \frac{ m+1 }{L}  \Big) +1\bigg] \widehat{\Var}_{\gamma^\ast}^\trainII[\widehat{s}_{\gamma^\ast,y}]   + \frac{1}{L^2}  \\ 
	 &\leq  \frac{5(m+2)^2 }{L^2} \widehat{\Var}_{\gamma^\ast}^\trainII[\widehat{s}_{\gamma^\ast,y}]   + \frac{1}{L^2}  \\  
	 &\leq \frac{5(m+2)^2 }{L^2} \bigg( \sup_{\Omega(h_y) \leq \frac{1}{L\sqrt{\lambda_y}} } \absbig{ \widehat{\Var}_{\gamma^\ast}^\trainII[h_y] - \Var_{\gamma^\ast}[h_y] }  + \frac{1}{L^2} + \Var_{\gamma^\ast}[\widehat{s}_{\gamma^\ast,y} - s_{\gamma^\ast,y}] \\
	 &\qquad\qquad+  \frac{2}{L} \sqrt{\Var_{\gamma^\ast}[\widehat{s}_{\gamma^\ast,y} - s_{\gamma^\ast,y}]} \bigg)   + \frac{1}{L^2},
\end{align*}
\normalsize
where the third line utilizes the fact that the function $ \frac{\widehat{\E}_y^\trainII[ \widehat{s}_{\widehat{\gamma},y} ] - \widehat{\E}_0^\trainII[ \widehat{s}_{\widehat{\gamma},y}]  }{ \widehat{\Var}_{\gamma^\ast}^{\trainII}[ \widehat{s}_{\widehat{\gamma},y}] } \widehat{s}_{\widehat{\gamma},y}$ is inside of the feasible set of the optimization problem solved by $\widehat{s}_{\gamma^\ast,y}$,  the fifth line utilizes the fact $\widehat{s}_{\widehat{\gamma},y}$ satisfies the constraints of the optimization problem it solves, the eight line uses the same argument from Lemma  \eqref{Rate for Learning Diagonal of Fisher Information Matrix, Unknown Gamma Star} to control the ratio of empirical variances, and the last line is by line \eqref{vitamin h}. Thus, it follows that:
\begin{align*}
	&\Var_{\gamma^\ast}[\widehat{s}_{\widehat{\gamma},y}] - \widehat{\Var}_{\widehat{\gamma}}^\trainII[\widehat{s}_{\widehat{\gamma},y}]\\
	 &\leq \sup_{\Omega(h_y) \leq \frac{1}{L\sqrt{\lambda_y}}} \absBig{\widehat{\Var}_{\gamma^\ast}^\trainII[h_y]  -   \Var_{\gamma^\ast}[h_y] } + \frac{m+1}{L}\norm{\widehat{\gamma} - \gamma^\ast}{1}\widehat{\Var}_{\gamma^\ast}^\trainII[\widehat{s}_{\widehat{\gamma},y}] \\ \\ 
	&\leq 11\frac{(m+2)^3}{L^3}\sup_{\Omega(h_y) \leq \frac{1}{L\sqrt{\lambda_y}}} \absBig{\widehat{\Var}_{\gamma^\ast}^\trainII[h_y]  -   \Var_{\gamma^\ast}[h_y] } + \frac{6(m+2)^3}{L^5} \norm{\widehat{\gamma} - \gamma^\ast}{1} \\
	&\qquad + \frac{10(m+2)^3}{L^3} \Var_{\gamma^\ast}[\widehat{s}_{\gamma^\ast,y} - s_{\gamma^\ast,y}] + \frac{10(m+2)^3}{L^4} \norm{\widehat{\gamma} - \gamma^\ast}{1} \sqrt{\Var_{\gamma^\ast}[\widehat{s}_{\gamma^\ast,y} - s_{\gamma^\ast,y} ] }.
\end{align*}

So, bringing all of our work in this proof together, it follows that:
\begin{align*}
	&\Var_{\gamma^\ast}[ \widehat{s}_{\widehat{\gamma},y} - s_{\gamma^\ast,y} ] \\
	&\leq \widetilde{T}_y  + \sup_{\Omega(h_y) \leq \frac{1}{L\sqrt{\lambda_y}} } \absBig{\widehat{\E}_y^{\trainII}[h_y] - \E_y[h_y]}  + \sup_{\Omega(h_y) \leq \frac{1}{L\sqrt{\lambda_y}} } \absBig{\widehat{\E}_0^{\trainII}[h_y] - \E_0[h_y] } + \Big[\Var_{\gamma^\ast}[\widehat{s}_{\widehat{\gamma},y}] - \widehat{\Var}_{\widehat{\gamma}}^\trainII[\widehat{s}_{\widehat{\gamma},y}]\Big] \\ \\ 
	&\leq \widetilde{T}_y  + \sup_{\Omega(h_y) \leq \frac{1}{L\sqrt{\lambda_y}} } \absBig{\widehat{\E}_y^{\trainII}[h_y] - \E_y[h_y]}  + \sup_{\Omega(h_y) \leq \frac{1}{L\sqrt{\lambda_y}} } \absBig{\widehat{\E}_0^{\trainII}[h_y] - \E_0[h_y] } \\
	&\qquad +  11\frac{(m+2)^3}{L^3}\sup_{\Omega(h) \leq \frac{1}{L\sqrt{\lambda_y}}} \absBig{\widehat{\Var}_{\gamma^\ast}^\trainII[h]  -   \Var_{\gamma^\ast}[h] } + \frac{6(m+2)^3}{L^5} \norm{\widehat{\gamma} - \gamma^\ast}{1} \\ 
	&\qquad +  \frac{10(m+2)^3}{L^3} \Var_{\gamma^\ast}[\widehat{s}_{\gamma^\ast,y} - s_{\gamma^\ast,y}] + \frac{10(m+2)^3}{L^4} \norm{\widehat{\gamma} - \gamma^\ast}{1} \sqrt{\Var_{\gamma^\ast}[\widehat{s}_{\gamma^\ast,y} - s_{\gamma^\ast,y} ] }.
\end{align*}
\normalsize

So:
\begin{align*}
	&\E_{\allDII}\Big[ \Var_{\gamma^\ast}[ \widehat{s}_{\widehat{\gamma},y} - s_{\gamma^\ast,y} ]  \Big] \\
	&\leq \E_{\allDII}[\widetilde{T}_y] + 11\frac{(m+2)^3}{L^3}\E_{\DtrainII}\Big[ \U_y(\lambda_y)\Big] +  \frac{6(m+2)^3}{L^5} \sqrt{m\E_{\allDII}\Big[ \norm{\widehat{\gamma} - \gamma^\ast}{2}^2 \Big]} \\ 
	&\qquad + \frac{10(m+2)^3}{L^3} \E_{\allDII}\Big[ \Var_{\gamma^\ast}[\widehat{s}_{\gamma^\ast,y} - s_{\gamma^\ast,y}] \Big]  \\
	&\qquad +  \frac{10(m+2)^3}{L^4}  \sqrt{m\E_{\allDII}\Big[ \norm{\widehat{\gamma} - \gamma^\ast}{2}^2 \Big]  \E_{\DtrainII}\Big[\Var_{\gamma^\ast}[\widehat{s}_{\gamma^\ast,y} - s_{\gamma^\ast,y}]\Big] } \\ \\ 
	&= O\Bigg(\E_{\allDII}[\widetilde{T}_y] + \E_{\DtrainII}\Big[ \U_y(\lambda_y)\Big] + \sqrt{\E_{\allDII}\Big[ \norm{\widehat{\gamma} - \gamma^\ast}{2}^2 \Big]} +  \E_{\allDII}\Big[ \Var_{\gamma^\ast}[\widehat{s}_{\gamma^\ast,y} - s_{\gamma^\ast,y}] \Big]  \Bigg),
\end{align*}
as claimed. Finally, let's obtain a tail bound on $\Var_{\gamma^\ast}[ \widehat{s}_{\widehat{\gamma},y} - s_{\gamma^\ast,y}]$. For any $a > 0$, we have that:
\begin{align*}
	\P_{\allDII}\Big[\Var_{\gamma^\ast}[ \widehat{s}_{\widehat{\gamma},y} - s_{\gamma^\ast,y} ]  \geq a\Big] &\leq \P_{\allDII}[\widetilde{T}_y \geq a/7]   +   \P_{\DtrainII}\Bigg[ \sup_{\Omega(h_y) \leq \frac{1}{L\sqrt{\lambda_y}} } \absBig{\widehat{\E}_y^{\trainII}[h_y] - \E_y[h_y]} \geq a/7 \Bigg] \\ 
	&\qquad + \P_{\DtrainII}\Bigg[ \sup_{\Omega(h_y) \leq \frac{1}{L\sqrt{\lambda_y}} } \absBig{\widehat{\E}_0^{\trainII}[h_y] - \E_0[h_y]} \geq a/7 \Bigg] \\
	&\qquad+ \P_{\DtrainII}\Bigg[\sup_{\Omega(h_y) \leq \frac{1}{L\sqrt{\lambda_y}}} \absBig{\widehat{\Var}_{\gamma^\ast}^\trainII[h_y]  -   \Var_{\gamma^\ast}[h_y] } \geq \frac{L^3}{77(m+2)^3}a \Bigg]\\
	&\qquad + \P_{\allDII}\Bigg[ \norm{\widehat{\gamma} - \gamma^\ast}{2}\geq \frac{L^5}{66\sqrt{m}(m+2)^3}a \Bigg] \\ 
	&\qquad + \P_{\DtrainII}\Bigg[ \Var_{\gamma^\ast}[\widehat{s}_{\gamma^\ast,y} - s_{\gamma^\ast,y}] \geq \frac{L^3}{70(m+2)^3}a \Bigg] \\ 
	&\qquad + \P_{\DtrainII}\Bigg[  \norm{\widehat{\gamma} - \gamma^\ast}{2} \geq \sqrt{ \frac{L^4}{70m(m+2)^3}a }  \Bigg] \\
	&\qquad+ \P_{\DtrainII}\Bigg[   \Var_{\gamma^\ast}[\widehat{s}_{\gamma^\ast,y} - s_{\gamma^\ast,y} ]  \geq  \frac{L^4}{70(m+2)^3}a  \Bigg] \\ \\ 
	&\leq \P_{\allDII}[\widetilde{T}_y \geq a/7]   +   \P_{\DtrainII}\Bigg[ \sup_{\Omega(h_y) \leq \frac{1}{L\sqrt{\lambda_y}} } \absBig{\widehat{\E}_y^{\trainII}[h_y] - \E_y[h_y]} \geq a/7 \Bigg] \\ 
	&\qquad + \P_{\DtrainII}\Bigg[ \sup_{\Omega(h_y) \leq \frac{1}{L\sqrt{\lambda_y}} } \absBig{\widehat{\E}_0^{\trainII}[h_y] - \E_0[h_y]} \geq a/7 \Bigg] \\
	&\qquad+ \P_{\DtrainII}\Bigg[\sup_{\Omega(h_y) \leq \frac{1}{L\sqrt{\lambda_y}}} \absBig{\widehat{\Var}_{\gamma^\ast}^\trainII[h_y]  -   \Var_{\gamma^\ast}[h_y] } \geq \frac{L^3}{77(m+2)^3}a \Bigg]\\
	&\qquad + 2\P_{\allDII}\Bigg[ \norm{\widehat{\gamma} - \gamma^\ast}{2}\geq \widetilde{\tau}_2(a)\Bigg] \\ 
	&\qquad+ 2\P_{\DtrainII}\Bigg[   \Var_{\gamma^\ast}[\widehat{s}_{\gamma^\ast,y} - s_{\gamma^\ast,y} ]  \geq  \frac{L^4}{70(m+2)^3}a  \Bigg],  
\end{align*}
as claimed. 

\end{proof}
\vspace{0.4in}

\begin{proof}[\uline{Proof of Lemma \eqref{Rate for Learning Columns of Fisher Information Matrix}}]

Observe that:
\begin{align*}
	& \norm{ \text{col } y \text{ of } \widehat{A} - \FisherInfo{\gamma^\ast} }{2}^2\\
	 &=  \Bignorm{ (\widehat{\E}_y^{\trainI}[\widehat{s}_{\widehat{\gamma}}] - \widehat{\E}_0^{\trainI}[\widehat{s}_{\widehat{\gamma}}]) - (\E_y[s_{\gamma^\ast}] - \E_0[s_{\gamma^\ast}])}{2}^2 \\ 
	 &\leq  2\Bignorm{ (\widehat{\E}_y^{\trainI}[\widehat{s}_{\widehat{\gamma}}] - \widehat{\E}_0^{\trainI}[\widehat{s}_{\widehat{\gamma}}]) - (\E_y[\widehat{s}_{\widehat{\gamma}}] - \E_0[\widehat{s}_{\widehat{\gamma}}]) }{2}^2 + 2\Bignorm{ (\E_y[\widehat{s}_{\widehat{\gamma}}] - \E_0[\widehat{s}_{\widehat{\gamma}}]) - (\E_y[s_{\gamma^\ast}] - \E_0[s_{\gamma^\ast}])}{2}^2.
\end{align*}

Let's get control of the second term. Note that:
\begin{align*}
	\Bignorm{ (\E_y[\widehat{s}_{\widehat{\gamma}}] - \E_0[\widehat{s}_{\widehat{\gamma}}]) - (\E_y[s_{\gamma^\ast}] - \E_0[s_{\gamma^\ast}])}{2}^2 &=  \sum_{j=1}^{m}\Big((\E_y[\widehat{s}_{\widehat{\gamma},j}] - \E_0[\widehat{s}_{\widehat{\gamma},j}]) - (\E_y[s_{\gamma^\ast,j}] - \E_0[s_{\gamma^\ast,j}]) \Big)^2 \\ 
	&=  \sum_{j=1}^{m}\Big( \Cov_{\gamma^\ast}[s_{\gamma^\ast,y},\widehat{s}_{\widehat{\gamma},j}-s_{\gamma^\ast,j}] \Big)^2 \\ 
	&\leq  \Var_{\gamma^\ast}[s_{\gamma^\ast,y}] \sum_{j=1}^{m}  \Var_{\gamma^\ast}[\widehat{s}_{\widehat{\gamma},j}-s_{\gamma^\ast,j}] \\ 
	&\leq \frac{1}{L^2} \sum_{j=1}^{m}  \Var_{\gamma^\ast}[\widehat{s}_{\widehat{\gamma},j}-s_{\gamma^\ast,j}].
\end{align*}
Ergo:
\begin{align*}
	&\norm{ \text{col } y \text{ of } \widehat{A} - \FisherInfo{\gamma^\ast} }{2}^2 \\
	&\leq 2 \sum_{j=1}^{m} \Big((\widehat{\E}_y^{\trainI}[\widehat{s}_{\widehat{\gamma},j}] - \widehat{\E}_0^{\trainI}[\widehat{s}_{\widehat{\gamma},j}]) - (\E_y[\widehat{s}_{\widehat{\gamma},j}] - \E_0[\widehat{s}_{\widehat{\gamma},j}])\Big)^2 + \frac{2}{L^2} \sum_{j=1}^{m}  \Var_{\gamma^\ast}[\widehat{s}_{\widehat{\gamma},j}-s_{\gamma^\ast,j}].
\end{align*}
Next, let us analyze the behavior of $(\widehat{\E}_y^{\trainI}[\widehat{s}_{\widehat{\gamma},j}] - \widehat{\E}_0^{\trainI}[\widehat{s}_{\widehat{\gamma},j}]) - (\E_y[\widehat{s}_{\widehat{\gamma},j}] - \E_0[\widehat{s}_{\widehat{\gamma},j}])$, for arbitrary $j \in [m]$. Note that the conditional expectation (given $\allDII$) of this random variable is $0$. Ergo:
\small
\begin{align}
&\E_{\allD} \bigg[ \Big( (\widehat{\E}_y^{\trainI}[\widehat{s}_{\widehat{\gamma},j}] - \widehat{\E}_0^{\trainI}[\widehat{s}_{\widehat{\gamma},j}]) - (\E_y[\widehat{s}_{\widehat{\gamma},j}] - \E_0[\widehat{s}_{\widehat{\gamma},j}]) \Big)^2 \bigg]  \\
&= \E_{\allDII}\E_{\DtrainI \mid \allDII} \bigg[ \Big( (\widehat{\E}_y^{\trainI}[\widehat{s}_{\widehat{\gamma},j}] - \widehat{\E}_0^{\trainI}[\widehat{s}_{\widehat{\gamma},j}]) - (\E_y[\widehat{s}_{\widehat{\gamma},j}] - \E_0[\widehat{s}_{\widehat{\gamma},j}]) \Big)^2 \bigg] \nonumber \\ 
	&= \E_{\allDII}\Var_{\DtrainI \mid \allDII} \Big[  \widehat{\E}_y^{\trainI}[\widehat{s}_{\widehat{\gamma},j}] - \widehat{\E}_0^{\trainI}[\widehat{s}_{\widehat{\gamma},j}]   \Big] \nonumber \\ 
	&= \E_{\allDII}\bigg\{ \Var_{\DtrainI \mid \allDII} \Big[  \widehat{\E}_y^{\trainI}[\widehat{s}_{\widehat{\gamma},j}]\Big] + \Var_{\DtrainI \mid \allDII} \Big[ \widehat{\E}_0^{\trainI}[\widehat{s}_{\widehat{\gamma},j}]  \Big]   \bigg\} \nonumber \\ 
	&\leq \frac{2}{\xi n^\train} \E_{\allDII}\bigg\{ \Var_{y} [\widehat{s}_{\widehat{\gamma},j}]  + \Var_{0} [\widehat{s}_{\widehat{\gamma},j}]   \bigg\} \nonumber\\ 
	&\leq \frac{2}{L \xi n^\train} \E_{\allDII}\bigg\{ \gamma_y^\ast \Var_{y} [\widehat{s}_{\widehat{\gamma},j}]  + \gamma_0^\ast \Var_{0} [\widehat{s}_{\widehat{\gamma},j}]   \bigg\} \nonumber \\ 
	&\leq \frac{2}{L \xi n^\train} \E_{\allDII}\bigg\{ \Var_{\gamma^\ast} [\widehat{s}_{\widehat{\gamma},j}]   \bigg\} \nonumber \\ 
	&= \frac{2}{L \xi n^\train} \E_{\allDII}\bigg\{ \Var_{\gamma^\ast} [(\widehat{s}_{\widehat{\gamma},j} - s_{\gamma^\ast,j}) + s_{\gamma^\ast,j}]   \bigg\} \nonumber \\ \nonumber\\
	&= \frac{2}{L \xi n^\train} \E_{\allDII}\bigg\{ \Var_{\gamma^\ast} [\widehat{s}_{\widehat{\gamma},j} - s_{\gamma^\ast,j} ] \bigg\} +   \frac{2}{L \xi n^\train} \E_{\allDII}\bigg\{ \Var_{\gamma^\ast} [s_{\gamma^\ast,j}] \bigg\} \nonumber \\ 
	&\qquad + \frac{4}{L \xi n^\train} \E_{\allDII}\bigg\{\Cov_{\gamma^\ast}[\widehat{s}_{\widehat{\gamma},j} - s_{\gamma^\ast,j},s_{\gamma^\ast,j}]  \bigg\} \nonumber \\ \nonumber \\
	&\leq \frac{2}{L \xi n^\train} \E_{\allDII}\bigg\{ \Var_{\gamma^\ast} [\widehat{s}_{\widehat{\gamma},j} - s_{\gamma^\ast,j} ] \bigg\} +   \frac{2}{L^3 \xi n^\train} \nonumber \\ 
	&\qquad + \frac{4}{L \xi n^\train} \E_{\allDII}\bigg\{\sqrt{ \Var_{\gamma^\ast}[\widehat{s}_{\widehat{\gamma},j} - s_{\gamma^\ast,j}] \Var_{\gamma^\ast}[s_{\gamma^\ast,j}]  } \bigg\} \nonumber \\ \nonumber \\
	&\leq \frac{2}{L \xi n^\train} \E_{\allDII}\Big[ \Var_{\gamma^\ast} [\widehat{s}_{\widehat{\gamma},j} - s_{\gamma^\ast,j} ] \Big] +   \frac{2}{L^3 \xi n^\train} \nonumber \\ 
	&\qquad + \frac{4}{L^2 \xi n^\train} \sqrt{ \E_{\allDII} \Big[ \Var_{\gamma^\ast}[\widehat{s}_{\widehat{\gamma},j} - s_{\gamma^\ast,j}]  \Big] } \nonumber\\ \nonumber\\
	&\leq  \frac{4}{L^3 \xi n^\train}\bigg( 1+ \E_{\allDII}\Big[ \Var_{\gamma^\ast} [\widehat{s}_{\widehat{\gamma},j} - s_{\gamma^\ast,j} ] \Big] + \sqrt{ \E_{\allDII} \Big[ \Var_{\gamma^\ast}[\widehat{s}_{\widehat{\gamma},j} - s_{\gamma^\ast,j}]  \Big] } \bigg) \nonumber \\ 
	&= O\bigg(\frac{1}{n^\train} \bigg), \label{MEOW}
\end{align}
\normalsize
where the sixth line follows from the Law of Total Variance. Thus, it follows that:
\begin{align*}
	\E_{\allD}\Big[ \norm{ \text{col } y \text{ of } \widehat{A} - \FisherInfo{\gamma^\ast} }{2}^2\Big] 
	&= O\Bigg(  \frac{1}{n^\train}  +    \sum_{j=1}^{m}  \E_{\allDII}\Big[\Var_{\gamma^\ast}[\widehat{s}_{\widehat{\gamma},j}-s_{\gamma^\ast,j}]\Big]  \Bigg), 
\end{align*}
as claimed! Next, let's prove our tail bound on $\norm{ \text{col } y \text{ of } \widehat{A} - \FisherInfo{\gamma^\ast} }{2}^2$. By our work earlier in this proof, we have that, for any $a > 0$:
\begin{align*}
	&\P_{\allD}\Big[\norm{ \text{col } y \text{ of } \widehat{A} - \FisherInfo{\gamma^\ast} }{2}^2 \geq a \Big] \\
	&\leq \P_{\allD}\bigg[2 \sum_{j=1}^{m} \Big((\widehat{\E}_y^{\trainI}[\widehat{s}_{\widehat{\gamma},j}] - \widehat{\E}_0^{\trainI}[\widehat{s}_{\widehat{\gamma},j}]) - (\E_y[\widehat{s}_{\widehat{\gamma},j}] - \E_0[\widehat{s}_{\widehat{\gamma},j}])\Big)^2  \geq \frac{1}{2}a \bigg] \\
	&\qquad + \P_{\allDII}\bigg[\frac{2}{L^2} \sum_{j=1}^{m}  \Var_{\gamma^\ast}[\widehat{s}_{\widehat{\gamma},j}-s_{\gamma^\ast,j}]  \geq \frac{1}{2}a \bigg] \\ \\ 
	&\leq \sum_{j=1}^{m} \P_{\allD}\bigg[  \absBig{(\widehat{\E}_y^{\trainI}[\widehat{s}_{\widehat{\gamma},j}] - \widehat{\E}_0^{\trainI}[\widehat{s}_{\widehat{\gamma},j}]) - (\E_y[\widehat{s}_{\widehat{\gamma},j}] - \E_0[\widehat{s}_{\widehat{\gamma},j}])} \geq \frac{1}{2\sqrt{m}}\sqrt{a} \bigg] \\
	&\qquad + \sum_{j=1}^{m}  \P_{\allDII}\bigg[ \Var_{\gamma^\ast}[\widehat{s}_{\widehat{\gamma},j}-s_{\widehat{\gamma},j}]  \geq \frac{L^2}{4m}a \bigg] \\ \\ 
	&\leq \sum_{j=1}^{m}\P_{\allD}\bigg[ \absBig{ \widehat{\E}_0^{\trainI}[\widehat{s}_{\widehat{\gamma},j}] - \E_0[\widehat{s}_{\widehat{\gamma},j}] } \geq \frac{1}{4\sqrt {m}}\sqrt{a}  \bigg] \\ 
	&\qquad + \sum_{j=1}^{m} \P_{\allD}\bigg[  \absBig{ \widehat{\E}_y^{\trainI}[\widehat{s}_{\widehat{\gamma},j}] -\E_y[\widehat{s}_{\widehat{\gamma},j}]    } \geq \frac{1}{4\sqrt{m}}\sqrt{a} \bigg] \\
	&\qquad + \sum_{j=1}^{m}  \P_{\allDII}\bigg[ \Var_{\gamma^\ast}[\widehat{s}_{\widehat{\gamma},j}-s_{\gamma^\ast,j}]  \geq \frac{L^2}{4m}a \bigg] \\ \\ 
	&=  \sum_{j=1}^{m}\E_{\allDII}\bigg\{ \P_{\DtrainI}\bigg[ \absBig{ \widehat{\E}_0^{\trainI}[\widehat{s}_{\widehat{\gamma},j}] - \E_0[\widehat{s}_{\widehat{\gamma},j}] } \geq \frac{1}{4\sqrt {m}}\sqrt{a}  \bigg] \bigg\} \\ 
	&\qquad + \sum_{j=1}^{m} \E_{\allDII}\bigg\{  \P_{\DtrainI}\bigg[  \absBig{ \widehat{\E}_y^{\trainI}[\widehat{s}_{\widehat{\gamma},j}] -\E_y[\widehat{s}_{\widehat{\gamma},j}]    } \geq \frac{1}{4\sqrt{m}}\sqrt{a} \bigg] \bigg\} \\
	&\qquad + \sum_{j=1}^{m}  \P_{\allDII}\bigg[ \Var_{\gamma^\ast}[\widehat{s}_{\widehat{\gamma},j}-s_{\gamma^\ast,j}]  \geq \frac{L^2}{4m}a \bigg].
\end{align*}

Towards bounding the display above, let's get a bound on $\P_{\DtrainI}\bigg[ \absBig{ \widehat{\E}_k^{\trainI}[\widehat{s}_{\widehat{\gamma},j}] - \E_k[\widehat{s}_{\widehat{\gamma},j}] } \geq \frac{1}{4\sqrt {m}}\sqrt{a}  \bigg]$, for each class $k$. As was argued earlier,  $\norm{\widehat{s}_{\widehat{\gamma},j}}{\infty}\leq  \frac{1}{DL\sqrt{\lambda_j}}$, so applying Hoeffding's inequality yields (for fixed $\DtrainII$):
\begin{align*}
	\P_{\DtrainI}\bigg[ \absBig{ \widehat{\E}_k^{\trainI}[\widehat{s}_{\widehat{\gamma},j}] - \E_k[\widehat{s}_{\widehat{\gamma},j}] } \geq \frac{1}{4\sqrt {m}}\sqrt{a}  \bigg] &\leq 2\exp\bigg\{-2\frac{\big(\frac{1}{2}\pi_k^\train n^\train\big)\big(\frac{1}{4\sqrt {m}}\sqrt{a}\big)^2 }{\big( \frac{2}{DL\sqrt{\lambda_j}} \big)^2 } \bigg\} \\ 
	&\leq 2\exp\bigg\{-\frac{ \frac{1}{16m}a }{ \frac{4}{D^2 L^2\lambda_j}  } \xi n^\train \bigg\} \\ 
	&\leq 2\exp\bigg\{-\frac{  \xi  L^2 D^2 \lambda_j n^\train }{64m   }a   \bigg\}.
\end{align*}

Therefore, it follows that:
\begin{align*}
	&\P_{\Dtrain}\Big[\norm{ \text{col } y \text{ of } \widehat{A} - \FisherInfo{\gamma^\ast} }{2}^2 \geq a \Big] \\
	&\leq 4\sum_{j=1}^{m}\exp\bigg\{-\frac{  \xi  L^2 D^2 \lambda_j n^\train }{64m   }a   \bigg\} + \sum_{j=1}^{m}  \P_{\DtrainII}\bigg[ \Var_{\gamma^\ast}[\widehat{s}_{\widehat{\gamma},j}-s_{\gamma^\ast,j}]  \geq \frac{L^2}{4m}a \bigg],
\end{align*}
as claimed!

\end{proof}
\vspace{0.4in}


\begin{proof}[\uline{Proof of Lemma \eqref{Learning the Inverse FIM}}]

Observe that:
\begin{align*}
	\maxSing{\widehat{A}^{-1} -  \FisherInfo{\gamma^\ast}^{-1}}\Ibig{ \minSing{\widehat{A}} > 0}  &= \maxSing{ \widehat{A}^{-1}[ \FisherInfo{\gamma^\ast} - \widehat{A}] \FisherInfo{\gamma^\ast}^{-1} } \Ibig{ \minSing{\widehat{A}} > 0} \\ 
	&\leq \maxSing{ \widehat{A}^{-1}} \maxSing{ \FisherInfo{\gamma^\ast} - \widehat{A}} \maxSing{ \FisherInfo{\gamma^\ast}^{-1} } \Ibig{ \minSing{\widehat{A}} > 0} \\
	&= \frac{\maxSing{\widehat{A}-  \FisherInfo{\gamma^\ast}}}{  \minSing{\widehat{A}}  \minSing{ \FisherInfo{\gamma^\ast}} }  \Ibig{ \minSing{\widehat{A}} > 0} \\ 
	&\leq \frac{\maxSing{\widehat{A}-  \FisherInfo{\gamma^\ast}}}{  \minSing{\widehat{A}}  \sqrt{\Lambda} }  \Ibig{ \minSing{\widehat{A}} > 0} .
\end{align*}

First, let's get control over $\maxSing{\widehat{A} -  \FisherInfo{\gamma^\ast}}$. Note that:
\begin{align*}
	\maxSing{\widehat{A}-  \FisherInfo{\gamma^\ast}} &= \sqrt{\maxEval{  (\widehat{A}-  \FisherInfo{\gamma^\ast})^T (\widehat{A}-  \FisherInfo{\gamma^\ast})  } } \\ 
	&\leq \sqrt{\sum_{i=1}^{m} \sum_{j=1}^{m} \absBig{\big[(\widehat{A}-  \FisherInfo{\gamma^\ast})^T (\widehat{A}-  \FisherInfo{\gamma^\ast})\big]_{i,j}} } \\ 
	&= \sqrt{\sum_{i=1}^{m} \sum_{j=1}^{m} \absBig{  \bigdotprod{\text{col } i \text{ of } \widehat{A}-  \FisherInfo{\gamma^\ast} }{  \text{col } j \text{ of } \widehat{A}-  \FisherInfo{\gamma^\ast}  }    } } \\ 
	&\leq \sqrt{\sum_{i=1}^{m} \sum_{j=1}^{m}  \bignorm{\text{col } i \text{ of } \widehat{A}-  \FisherInfo{\gamma^\ast} }{2}    \bignorm{ \text{col } j \text{ of } \widehat{A}-  \FisherInfo{\gamma^\ast}  }{2}     } \\ 
	&= \sum_{j=1}^{m} \bignorm{   \text{col } j \text{ of } \widehat{A}-  \FisherInfo{\gamma^\ast}  }{2}.
\end{align*}

Thus, so far, we have shown that:
\begin{align*}
	\implies \maxSing{\widehat{A}^{-1} -  \FisherInfo{\gamma^\ast}^{-1}}\Ibig{ \minSing{\widehat{A}} > 0} &\leq \frac{\sum_{j=1}^{m} \bignorm{   \text{col } j \text{ of } \widehat{A}-  \FisherInfo{\gamma^\ast}  }{2}}{ \sqrt{\Lambda} }  \cdot \frac{\Ibig{ \minSing{\widehat{A}} > 0} }{\minSing{\widehat{A}}}.
\end{align*}

Next, let's get control over $\minSing{\widehat{A}}$. Namely, we need to show that  $\minSing{\widehat{A}}$ cannot be ``too'' small.  Note that:
\begin{align}
	\sqrt{\Lambda} &\leq \minSing{ \FisherInfo{\gamma^\ast}}  \nonumber \\ 
	&= \sqrt{\min_{\norm{v}{2} =1}v^T \FisherInfo{\gamma^\ast}^T \FisherInfo{\gamma^\ast}v } \nonumber \\ 
	&= \sqrt{\min_{\norm{v}{2} =1}\Big\{v^T\widehat{A}^T \widehat{A}v + v^T\Big( \FisherInfo{\gamma^\ast}^T \FisherInfo{\gamma^\ast} - \widehat{A}^T \widehat{A}\Big)v } \Big\}  \nonumber \\
	&\leq \sqrt{\min_{\norm{v}{2} =1} v^T\widehat{A}^T \widehat{A}v + \max_{\norm{v}{2} =1} v^T\Big( \FisherInfo{\gamma^\ast}^T \FisherInfo{\gamma^\ast} - \widehat{A}^T \widehat{A}\Big)v }  \nonumber \\ 
	&=\sqrt{\min_{\norm{v}{2} =1} v^T\widehat{A}^T \widehat{A}v + \maxEval{ \FisherInfo{\gamma^\ast}^T  \FisherInfo{\gamma^\ast} - \widehat{A}^T\widehat{A} } }   \nonumber \\ 
	&\leq \sqrt{\min_{\norm{v}{2} =1} v^T\widehat{A}^T \widehat{A}v + \abs{\maxEval{ \FisherInfo{\gamma^\ast}^T  \FisherInfo{\gamma^\ast} - \widehat{A}^T \widehat{A} } }}   \nonumber \\
	&\leq \sqrt{\min_{\norm{v}{2} =1} v^T\widehat{A}^T \widehat{A}v} + \sqrt{\abs{\maxEval{ \FisherInfo{\gamma^\ast}^T \FisherInfo{\gamma^\ast} - \widehat{A}^T \widehat{A} } }}   \nonumber \\ 
	&= \minSing{\widehat{A}} + \sqrt{\abs{\maxEval{  \FisherInfo{\gamma^\ast}^T  \FisherInfo{\gamma^\ast} - \widehat{A}^T \widehat{A} } } }. \label{flying pig feet} 
\end{align}
The next step is to find an upper bound on $\sqrt{\abs{\maxEval{  \FisherInfo{\gamma^\ast}^T  \FisherInfo{\gamma^\ast} - \widehat{A}^T \widehat{A} } } }$. Towards that end, observe that:
\begin{align}
	\absbig{ \maxEval{  \FisherInfo{\gamma^\ast}^T  \FisherInfo{\gamma^\ast} - \widehat{A}^T \widehat{A} } } &\leq \sum_{i=1}^{m} \sum_{j=1}^{m}     \absBig{\big[\widehat{A}^T \widehat{A} - \FisherInfo{\gamma^\ast}^T  \FisherInfo{\gamma^\ast} \big]_{i,j} } \nonumber   \\ 
	&= \sum_{i=1}^{m} \sum_{j=1}^{m}     \absBig{\big[\widehat{A}^T \widehat{A} - \widehat{A}^T\FisherInfo{\gamma^\ast} +  \widehat{A}^T\FisherInfo{\gamma^\ast}  - \FisherInfo{\gamma^\ast}^T  \FisherInfo{\gamma^\ast} \big]_{i,j} }\nonumber \\
	&\leq \sum_{i=1}^{m} \sum_{j=1}^{m}     \absbigg{\Big[\widehat{A}^T\big( \widehat{A} - \FisherInfo{\gamma^\ast}\big) \Big]_{i,j} } +  \sum_{i=1}^{m} \sum_{j=1}^{m}   \absbigg{\Big[ \big(\widehat{A}^T - \FisherInfo{\gamma^\ast}^T \big)  \FisherInfo{\gamma^\ast} \Big]_{i,j} }  \nonumber \\ 
	&= \sum_{i=1}^{m} \sum_{j=1}^{m}     \absbigg{\Big[\widehat{A}^T\big( \widehat{A} - \FisherInfo{\gamma^\ast}\big) \Big]_{i,j} } +  \sum_{i=1}^{m} \sum_{j=1}^{m}   \absbigg{\Big[ \big(\widehat{A} - \FisherInfo{\gamma^\ast} \big)^T  \FisherInfo{\gamma^\ast} \Big]_{i,j} }. \label{flying horse hooves} 
\end{align}
Note that:
\begin{align*}
	\absbigg{\Big[\widehat{A}^T\big( \widehat{A} - \FisherInfo{\gamma^\ast}\big) \Big]_{i,j} } &= \absbigg{ \Bigdotprod{    \widehat{\E}_i^{\trainI}[\widehat{s}_{\widehat{\gamma}}] - \widehat{\E}_0^{\trainI}[\widehat{s}_{\widehat{\gamma}}]     }{ \text{col } j \text{ of }  \widehat{A} - \FisherInfo{\gamma^\ast}} } \\
	&\leq \Bignorm{\widehat{\E}_i^{\trainI}[\widehat{s}_{\widehat{\gamma}}] - \widehat{\E}_0^{\trainI}[\widehat{s}_{\widehat{\gamma}}]   }{2}   \Bignorm{  \text{col } j \text{ of }  \widehat{A} - \FisherInfo{\gamma^\ast}  }{2},
\end{align*}
where we know, by Lemma \eqref{Rate for Learning Columns of Fisher Information Matrix}, that
\[
	 \Bignorm{ \textup{col } j \text{ of } \widehat{A} - \FisherInfo{\gamma^\ast} }{2}  = O_{\P}\Bigg( \frac{1}{\sqrt{n^\train}}  +     \sqrt{\sum_{k=1}^{m}  \E_{\allDII}\Big[\Var_{\gamma^\ast}[\widehat{s}_{\widehat{\gamma},k}-s_{\gamma^\ast,k}]\Big] }  \Bigg).
\]
As for $\Bignorm{\widehat{\E}_i^{\trainI}[\widehat{s}_{\widehat{\gamma}}] - \widehat{\E}_0^{\trainI}[\widehat{s}_{\widehat{\gamma}}]   }{2}$, observe that:
\begin{align*}
	&\Bignorm{\widehat{\E}_i^{\trainI}[\widehat{s}_{\widehat{\gamma}}] - \widehat{\E}_0^{\trainI}[\widehat{s}_{\widehat{\gamma}}]   }{2}  \\
	&\leq \Bignorm{\E_i[s_{\gamma^\ast}] - \E_0[s_{\gamma^\ast}]}{2} + \Bignorm{\E_i[\widehat{s}_{\widehat{\gamma}} - s_{\gamma^\ast} ] - \E_0[\widehat{s}_{\widehat{\gamma}} - s_{\gamma^\ast}]}{2} \\
	&\qquad  +   \Bignorm{\big( \widehat{\E}_{i}^{\trainI}[\widehat{s}_{\widehat{\gamma}}] - \widehat{\E}_{0}^{\trainI}[\widehat{s}_{\widehat{\gamma}}] \big) - \big( \E_{i}[\widehat{s}_{\widehat{\gamma}}] - \E_0[\widehat{s}_{\widehat{\gamma}}] \big)}{2} \\ \\ 
	&= \sqrt{ \sum_{z= 1}^{m} \big(\E_i[s_{\gamma^\ast,z}] - \E_0[s_{\gamma^\ast,z}] \big)^2 }  + \sqrt{ \sum_{z=1}^{m} \big(\E_i[\widehat{s}_{\widehat{\gamma},z} - s_{\gamma^\ast,z} ] - \E_0[\widehat{s}_{\widehat{\gamma},z} - s_{\gamma^\ast,z}]\big)^2  } \\
	&\qquad  +  \Bignorm{\big( \widehat{\E}_{i}^{\trainI}[\widehat{s}_{\widehat{\gamma}}] - \widehat{\E}_{0}^{\trainI}[\widehat{s}_{\widehat{\gamma}}] \big) - \big( \E_{i}[\widehat{s}_{\widehat{\gamma}}] - \E_0[\widehat{s}_{\widehat{\gamma}}] \big)}{2} \\ \\ 
	&\leq \sqrt{ \sum_{z= 1}^{m} \big(\E_i[s_{\gamma^\ast,z}] - \E_0[s_{\gamma^\ast,z}] \big)^2 } + \sqrt{ \Var_{\gamma^\ast}[s_{\gamma^\ast,i}]\sum_{z=1}^{m}  \Var_{\gamma^\ast}[\widehat{s}_{\widehat{\gamma},z} - s_{\gamma^\ast,z}]  } \\
	&\qquad  +  \Bignorm{\big( \widehat{\E}_{i}^{\trainI}[\widehat{s}_{\widehat{\gamma}}] - \widehat{\E}_{0}^{\trainI}[\widehat{s}_{\widehat{\gamma}}] \big) - \big( \E_{i}[\widehat{s}_{\widehat{\gamma}}] - \E_0[\widehat{s}_{\widehat{\gamma}}] \big)}{2} \\ \\ 
	&\leq \sqrt{m\frac{16}{L^2} } + \sqrt{ \frac{1}{L^2} \sum_{z=1}^{m}  \Var_{\gamma^\ast}[\widehat{s}_{\widehat{\gamma},z} - s_{\gamma^\ast,z}]  } \\
	&\qquad  +  \Bignorm{\big( \widehat{\E}_{i}^{\trainI}[\widehat{s}_{\widehat{\gamma}}] - \widehat{\E}_{0}^{\trainI}[\widehat{s}_{\widehat{\gamma}}] \big) - \big( \E_{i}[\widehat{s}_{\widehat{\gamma}}] - \E_0[\widehat{s}_{\widehat{\gamma}}] \big)}{2} \\ \\ 
	&= O_{\P}(1),
\end{align*}
where the last line is a consequence of \eqref{MEOW} from Lemma \eqref{Rate for Learning Columns of Fisher Information Matrix}. 
Ergo, it follows that 
\[
	\absbigg{\Big[\widehat{A}^T\big( \widehat{A} - \FisherInfo{\gamma^\ast}\big) \Big]_{i,j} } =  O_{\P}\Bigg( \frac{1}{\sqrt{n^\train}}  +     \sqrt{\sum_{k=1}^{m}  \E_{\allDII}\Big[\Var_{\gamma^\ast}[\widehat{s}_{\widehat{\gamma},k}-s_{\gamma^\ast,k}]\Big] }  \Bigg).
\]


Next, let's analyze the behavior of $\absbigg{\Big[ \big(\widehat{A} - \FisherInfo{\gamma^\ast} \big)^T  \FisherInfo{\gamma^\ast} \Big]_{i,j} }$. Observe that:
\begin{align*}
	\absbigg{\Big[ \big(\widehat{A} - \FisherInfo{\gamma^\ast} \big)^T  \FisherInfo{\gamma^\ast} \Big]_{i,j} } &= \Bigdotprod{\text{col } i \text{ of } \widehat{A} - \FisherInfo{\gamma^\ast}}{ \E_{j}[s_{\gamma^\ast}] -\E_0[s_{\gamma^\ast}] } \\ 
	&\leq \Bignorm{\text{col } i \text{ of } \widehat{A} - \FisherInfo{\gamma^\ast}}{2} \Bignorm{  \E_{j}[s_{\gamma^\ast}] -\E_0[s_{\gamma^\ast}]    }{2}.
\end{align*}

Note that: 
\begin{align*}
	\Bignorm{  \E_{j}[s_{\gamma^\ast}] -\E_0[s_{\gamma^\ast}]    }{2} &= \sqrt{  \sum_{k=1}^{m} \Big(  \E_j[s_{\gamma^\ast,k}] - \E_0[s_{\gamma^\ast,k}]  \Big)^2  } \\ 
	&= \sqrt{  \sum_{k=1}^{m} \Big(  \Cov_{\gamma^\ast}[s_{\gamma^\ast,j} , s_{\gamma^\ast,k}] \Big)^2  } \\ 
	&\leq \sqrt{  \Var_{\gamma^\ast}[s_{\gamma^\ast,j}] \sum_{k=1}^{m}     \Var_{\gamma^\ast}[s_{\gamma^\ast,k}]   } \\ 
	&\leq \frac{\sqrt{m}}{L^2}.
\end{align*}

Ergo:
\begin{align*}
	\absbigg{\Big[ \big(\widehat{A} - \FisherInfo{\gamma^\ast} \big)^T  \FisherInfo{\gamma^\ast} \Big]_{i,j} } &\leq \Bignorm{\text{col } i \text{ of } \widehat{A} - \FisherInfo{\gamma^\ast}}{2}  \frac{\sqrt{m}}{L^2} \\ 
	&= O_{\P}\Bigg( \frac{1}{\sqrt{n^\train}}  +     \sqrt{\sum_{k=1}^{m}  \E_{\allDII}\Big[\Var_{\gamma^\ast}[\widehat{s}_{\widehat{\gamma},k}-s_{\gamma^\ast,k}]\Big] }  \Bigg).
\end{align*}

Therefore, combining our analysis of $\absbigg{\Big[\widehat{A}^T\big( \widehat{A} - \FisherInfo{\gamma^\ast}\big) \Big]_{i,j} }$ and  $\absbigg{\Big[ \big(\widehat{A} - \FisherInfo{\gamma^\ast} \big)^T  \FisherInfo{\gamma^\ast} \Big]_{i,j} }$ with the inequality in display \eqref{flying horse hooves}, it follows that:

\[
	\implies \absbig{ \maxEval{  \FisherInfo{\gamma^\ast}^T  \FisherInfo{\gamma^\ast} - \widehat{A}^T \widehat{A} } } = O_{\P}\Bigg( \frac{1}{\sqrt{n^\train}}  +     \sqrt{\sum_{k=1}^{m}  \E_{\allDII}\Big[\Var_{\gamma^\ast}[\widehat{s}_{\widehat{\gamma},k}-s_{\gamma^\ast,k}]\Big] }  \Bigg).
\]

Now, since we already showed that $\sqrt{\Lambda} \leq \minSing{\widehat{A}} + \sqrt{\abs{\maxEval{  \FisherInfo{\gamma^\ast}^T  \FisherInfo{\gamma^\ast} - \widehat{A}^T \widehat{A} } } }$ in display \eqref{flying pig feet}, it follows that:

\[
	\implies \sqrt{\Lambda} + O_{\P}\Bigg( \frac{1}{(n^\train)^{\tfrac{1}{4}}} +     \bigg(\sum_{k=1}^{m}  \E_{\allDII}\Big[\Var_{\gamma^\ast}[\widehat{s}_{\widehat{\gamma},k}-s_{\gamma^\ast,k}]\Big]   \bigg)^{\tfrac{1}{4}}    \Bigg) \leq  \minSing{\widehat{A}}
\]
\[
	\implies \frac{\Ibig{ \minSing{\widehat{A}} > 0}  }{\minSing{\widehat{A}}} \leq \frac{1}{\sqrt{\Lambda}} + o_{\P}(1)
\]
\begin{align*}
	\implies \maxSing{\widehat{A}^{-1} -  \FisherInfo{\gamma^\ast}^{-1}}\Ibig{ \minSing{\widehat{A}} > 0} &\leq \bigg( \frac{1}{\sqrt{\Lambda}} + o_{\P}(1) \bigg) \frac{ \sum_{j=1}^{m} \bignorm{   \text{col } j \text{ of } \widehat{A}-  \FisherInfo{\gamma^\ast}  }{2} }{ \sqrt{\Lambda} } \\
	&= O_{\P}\Bigg( \frac{1}{\sqrt{n^\train}}  +     \sqrt{\sum_{k=1}^{m}  \E_{\allDII}\Big[\Var_{\gamma^\ast}[\widehat{s}_{\widehat{\gamma},k}-s_{\gamma^\ast,k}]\Big] }  \Bigg),
\end{align*}

where the last line follows from Lemma \eqref{Rate for Learning Columns of Fisher Information Matrix}. Next, we need to prove that $\Ibig{\minSing{\widehat{A}}=0}$ goes down exponentially fast. It follows from display \eqref{flying pig feet} that $\sqrt{\Lambda}-\sqrt{\abs{\maxEval{  \FisherInfo{\gamma^\ast}^T  \FisherInfo{\gamma^\ast} - \widehat{A}^T \widehat{A} } } } \leq \minSing{\widehat{A}}$, so
\[
	\Ibig{\minSing{\widehat{A}} > 0} \geq \Ibig{\sqrt{\Lambda}-\sqrt{\abs{\maxEval{  \FisherInfo{\gamma^\ast}^T  \FisherInfo{\gamma^\ast} - \widehat{A}^T \widehat{A} } } }> 0},
\]
which implies that:
\begin{align*}
	\P_{\allD}\big[\minSing{\widehat{A}}=0\big] &= 1 - \P_{\allD}\big[\minSing{\widehat{A}} > 0 \big]  \\ 
	&\leq 1 - \P_{\allD}\Big[ \sqrt{\Lambda}-\sqrt{\abs{\maxEval{  \FisherInfo{\gamma^\ast}^T  \FisherInfo{\gamma^\ast} - \widehat{A}^T \widehat{A} } } }> 0 \Big]  \\
	&= 1 - \P_{\allD}\Big[ \Lambda > \abs{\maxEval{  \FisherInfo{\gamma^\ast}^T  \FisherInfo{\gamma^\ast} - \widehat{A}^T \widehat{A} } }  \Big]  \\ 
	&= \P_{\allD}\Big[ \Lambda \leq \abs{\maxEval{  \FisherInfo{\gamma^\ast}^T  \FisherInfo{\gamma^\ast} - \widehat{A}^T \widehat{A} } }  \Big].
\end{align*}

Now, recall from earlier in the present proof that
\[
	\abs{\maxEval{  \FisherInfo{\gamma^\ast}^T  \FisherInfo{\gamma^\ast} - \widehat{A}^T \widehat{A} } } \leq \sum_{i=1}^{m} \sum_{j=1}^{m}     \absbigg{\Big[\widehat{A}^T\big( \widehat{A} - \FisherInfo{\gamma^\ast}\big) \Big]_{i,j} } +  \sum_{i=1}^{m} \sum_{j=1}^{m}   \absbigg{\Big[ \big(\widehat{A} - \FisherInfo{\gamma^\ast} \big)^T  \FisherInfo{\gamma^\ast} \Big]_{i,j} },
\]
and that we also showed that
\begin{align*}
	 \sum_{i=1}^{m} \sum_{j=1}^{m}     \absbigg{\Big[\widehat{A}^T\big( \widehat{A} - \FisherInfo{\gamma^\ast}\big) \Big]_{i,j} }  &\leq  \sum_{i=1}^{m}  \Bignorm{\widehat{\E}_i^{\trainI}[\widehat{s}_{\widehat{\gamma}}] - \widehat{\E}_0^{\trainI}[\widehat{s}_{\widehat{\gamma}}]   }{2}   \sum_{j=1}^{m} \Bignorm{  \text{col } j \text{ of }  \widehat{A} - \FisherInfo{\gamma^\ast}  }{2},
\end{align*}

as well as that
\begin{align*}
	\sum_{i=1}^{m} \sum_{j=1}^{m}   \absbigg{\Big[ \big(\widehat{A} - \FisherInfo{\gamma^\ast} \big)^T  \FisherInfo{\gamma^\ast} \Big]_{i,j} } &\leq   \frac{\sqrt{m}}{L^2} \sum_{i=1}^{m} \sum_{j=1}^{m} \Bignorm{\text{col } i \text{ of } \widehat{A} - \FisherInfo{\gamma^\ast}}{2} \\ 
	&=   \frac{m^{3/2}}{L^2} \sum_{i=1}^{m} \Bignorm{\text{col } i \text{ of } \widehat{A} - \FisherInfo{\gamma^\ast}}{2}.
\end{align*}

So, from this earlier work, it follows that
\begin{align*}
	&\abs{\maxEval{  \FisherInfo{\gamma^\ast}^T  \FisherInfo{\gamma^\ast} - \widehat{A}^T \widehat{A} } } \\
	&\leq \sum_{i=1}^{m}  \Bignorm{\widehat{\E}_i^{\trainI}[\widehat{s}_{\widehat{\gamma}}] - \widehat{\E}_0^{\trainI}[\widehat{s}_{\widehat{\gamma}}]   }{2}   \sum_{j=1}^{m} \Bignorm{  \text{col } j \text{ of }  \widehat{A} - \FisherInfo{\gamma^\ast}  }{2}      +        \frac{m^{3/2}}{L^2} \sum_{i=1}^{m} \Bignorm{\text{col } i \text{ of } \widehat{A} - \FisherInfo{\gamma^\ast}}{2} \\
	&= \Bigg(\sum_{i=1}^{m}  \Bignorm{ \widehat{\E}_i^{\trainI}[\widehat{s}_{\widehat{\gamma}}] - \widehat{\E}_0^{\trainI}[\widehat{s}_{\widehat{\gamma}}]   }{2}        +        \frac{m^{3/2}}{L^2} \Bigg) \sum_{i=1}^{m} \Bignorm{\text{col } i \text{ of } \widehat{A} - \FisherInfo{\gamma^\ast}}{2} \\ 
	&\leq \Bigg(\sum_{i=1}^{m}  \Bignorm{\text{col } i \text{ of } \widehat{A} - \FisherInfo{\gamma^\ast} }{2} + \sum_{i=1}^{m}  \Bignorm{ \text{col } i \text{ of }  \FisherInfo{\gamma^\ast} }{2}        +        \frac{m^{3/2}}{L^2} \Bigg) \sum_{i=1}^{m} \Bignorm{\text{col } i \text{ of } \widehat{A} - \FisherInfo{\gamma^\ast}}{2} \\ 
	&\leq \Bigg(\sum_{i=1}^{m}  \Bignorm{\text{col } i \text{ of } \widehat{A} - \FisherInfo{\gamma^\ast} }{2} +       2\frac{m^{3/2}}{L^2} \Bigg) \sum_{i=1}^{m} \Bignorm{\text{col } i \text{ of } \widehat{A} - \FisherInfo{\gamma^\ast}}{2} \\
	&= \bigg(\sum_{i=1}^{m}  \Bignorm{\text{col } i \text{ of } \widehat{A} - \FisherInfo{\gamma^\ast} }{2}\bigg)^2 +       2\frac{m^{3/2}}{L^2} \sum_{i=1}^{m} \Bignorm{\text{col } i \text{ of } \widehat{A} - \FisherInfo{\gamma^\ast}}{2},
\end{align*}
where the second inequality above uses the fact that $\widehat{\E}_i^{\trainI}[\widehat{s}_{\widehat{\gamma}}] - \widehat{\E}_0^{\trainI}[\widehat{s}_{\widehat{\gamma}}] $ is the $\numth{i}$ column of $\widehat{A}$. Ergo:
\begin{align*}
	&\P_{\Dtrain}\big[\minSing{\widehat{A}}=0\big] \\
	&\leq  \P_{\Dtrain}\Big[ \abs{\maxEval{  \FisherInfo{\gamma^\ast}^T  \FisherInfo{\gamma^\ast} - \widehat{A}^T \widehat{A} } } \geq \Lambda  \Big] \\ 
	&\leq \P_{\Dtrain}\Bigg[  \bigg(\sum_{i=1}^{m}  \Bignorm{\text{col } i \text{ of } \widehat{A} - \FisherInfo{\gamma^\ast} }{2}\bigg)^2 \geq \frac{1}{2}\Lambda \Bigg] +  \P_{\Dtrain}\Bigg[  2\frac{m^{3/2}}{L^2} \sum_{i=1}^{m} \Bignorm{\text{col } i \text{ of } \widehat{A} - \FisherInfo{\gamma^\ast}}{2}  \geq \frac{1}{2}\Lambda \Bigg] \\ 
	&= \P_{\Dtrain}\Bigg[  \sum_{i=1}^{m}  \Bignorm{\text{col } i \text{ of } \widehat{A} - \FisherInfo{\gamma^\ast} }{2} \geq \frac{1}{\sqrt{2}}\sqrt{\Lambda} \Bigg] +  \P_{\Dtrain}\Bigg[   \sum_{i=1}^{m} \Bignorm{\text{col } i \text{ of } \widehat{A} - \FisherInfo{\gamma^\ast}}{2}  \geq \frac{1}{4}\frac{L^2}{m^{3/2}} \Lambda \Bigg] \\ 
	&\leq  2\P_{\Dtrain}\Bigg[   \sum_{i=1}^{m} \Bignorm{\text{col } i \text{ of } \widehat{A} - \FisherInfo{\gamma^\ast}}{2}  \geq \min\bigg\{ \frac{1}{\sqrt{2}}\sqrt{\Lambda}  ,\text{ } \frac{1}{4}\frac{L^2}{m^{3/2}} \Lambda\bigg\} \Bigg] \\
	&\leq  2  \sum_{i=1}^{m} \P_{\Dtrain}\Bigg[   \Bignorm{\text{col } i \text{ of } \widehat{A} - \FisherInfo{\gamma^\ast}}{2}  \geq \frac{1}{m}\min\bigg\{ \frac{1}{\sqrt{2}}\sqrt{\Lambda}  ,\text{ } \frac{1}{4}\frac{L^2}{m^{3/2}} \Lambda\bigg\} \Bigg] \\ 
	&=  2  \sum_{i=1}^{m} \P_{\Dtrain}\Bigg[   \Bignorm{\text{col } i \text{ of } \widehat{A} - \FisherInfo{\gamma^\ast}}{2}^2  \geq \frac{1}{m^2}\min\bigg\{ \frac{1}{2}\Lambda  ,\text{ } \frac{1}{16}\frac{L^4}{m^{3}} \Lambda^2\bigg\} \Bigg]. 
\end{align*}

Since $\Ibig{\minSing{\widehat{A}}=0} = O_{\P}\Big( \P_{\Dtrain}\big[\minSing{\widehat{A}}=0\big] \Big)$, the claim in the Lemma holds.





\end{proof}
\vspace{0.4in}


\begin{proof}[\uline{Proof of Lemma \eqref{Rate of Second Order Term}}]Observe that:
\begin{align*}
	&\E_{\allD}\bigg[\Bignorm{ \widehat{\E}_{\pi^\ast}^\testI[\widehat{s}_{\widehat{\gamma}}-s_{\gamma^\ast}] -  \widehat{\E}_{\pi^\ast}^\trainI[\widehat{s}_{\widehat{\gamma}}-s_{\gamma^\ast}]     }{2}^2 \bigg] \\
	&= \E_{\allDII}\E_{\allDI \mid \allDII} \bigg[\Bignorm{ \widehat{\E}_{\pi^\ast}^\testI[\widehat{s}_{\widehat{\gamma}}-s_{\gamma^\ast}] -  \widehat{\E}_{\pi^\ast}^\trainI[\widehat{s}_{\widehat{\gamma}}-s_{\gamma^\ast}]     }{2}^2 \bigg]  \\ 
	&= \sum_{j=1}^{m}\E_{\allDII}\bigg\{  \Var_{\allDI \mid \allDII}\Big[\widehat{\E}_{\pi^\ast}^\testI[\widehat{s}_{\widehat{\gamma},j}-s_{\gamma^\ast,j}] -  \widehat{\E}_{\pi^\ast}^\trainI[\widehat{s}_{\widehat{\gamma},j}-s_{\gamma^\ast,j}] \Big]    \bigg\}.
\end{align*}

Let's focus on controlling the conditional variance term, for arbitrary $j\in[m]$. Define the random $1\times m$ \textit{row} vector 
\[
	\delta_j  := \Big(\E_1[\widehat{s}_{\widehat{\gamma},j} - s_{\gamma^\ast,j}] - \E_0[\widehat{s}_{\widehat{\gamma},j} - s_{\gamma^\ast,j}], \dots, \E_{m}[\widehat{s}_{\widehat{\gamma},j} - s_{\gamma^\ast,j}] - \E_0[\widehat{s}_{\widehat{\gamma},j} - s_{\gamma^\ast,j}] \Big).
\]

By Corollary \eqref{First Order Constant for Any Matching Function}, we have that:
\[
	\frac{1}{2} \Var_{\allDI \mid \allDII}\Big[\widehat{\E}_{\pi^\ast}^\testI[\widehat{s}_{\widehat{\gamma},j}-s_{\gamma^\ast,j}] -  \widehat{\E}_{\pi^\ast}^\trainI[\widehat{s}_{\widehat{\gamma},j}-s_{\gamma^\ast,j}] \Big]  
\]
\[
	 = \Bigg[\frac{1}{n^\test}+ \frac{1}{n^\train} \sum_{y=0}^{m}\frac{(\pi_y^\ast)^2}{\pi_y^\train}\Bigg] \Bigg( \Var_{\gamma^\ast}[\widehat{s}_{\widehat{\gamma},j} - s_{\gamma^\ast,j}] - \delta_j \FisherInfo{\gamma^\ast;\text{Cat}}^{-1} \delta_j^{T}\Bigg) + \frac{1}{n^\test}\delta_j \FisherInfo{\pi^\ast;\textup{Cat}}^{-1} \delta_j^{T}
\]
\[
	 \leq \Bigg[\frac{1}{n^\test}+ \frac{1}{n^\train} \sum_{y=0}^{m}\frac{(\pi_y^\ast)^2}{\pi_y^\train}\Bigg] \Var_{\gamma^\ast}[\widehat{s}_{\widehat{\gamma},j} - s_{\gamma^\ast,j}]  + \frac{1}{n^\test}\delta_j \FisherInfo{\pi^\ast;\textup{Cat}}^{-1} \delta_j^{T}
\]
\[
	 \leq \Bigg[\frac{1}{n^\test}+ \frac{1}{n^\train} \sum_{y=0}^{m}\frac{(\pi_y^\ast)^2}{\pi_y^\train}\Bigg] \Var_{\gamma^\ast}[\widehat{s}_{\widehat{\gamma},j} - s_{\gamma^\ast,j}]  + \frac{1}{n^\test} \maxEval{\FisherInfo{\pi^\ast;\textup{Cat}}^{-1} } \bignorm{\delta_j}{2}^2
\]
\[
	 \leq \Bigg[\frac{1}{n^\test}+ \frac{1}{n^\train} \sum_{y=0}^{m}\frac{(\pi_y^\ast)^2}{\pi_y^\train}\Bigg] \Var_{\gamma^\ast}[\widehat{s}_{\widehat{\gamma},j} - s_{\gamma^\ast,j}]  + \frac{1}{2n^\test} \bignorm{\delta_j}{2}^2,
\]
where the first and third inequalities are consequences of Lemma \eqref{Bounds on Eigenvalues of Categorical Fisher Info} and the fact that the components of $\widehat{\gamma}$ and $\pi^\ast$ are globally bounded away from $0$ and $1$. Next, bound the $\bignorm{\delta_j}{2}^2$ term:
\begin{align*}
	\bignorm{\delta_j}{2}^2 &= \sum_{k=1}^{m} \big(\E_k[\widehat{s}_{\widehat{\gamma},j} - s_{\gamma^\ast,j}] - \E_0[\widehat{s}_{\widehat{\gamma},j} - s_{\gamma^\ast,j}]\big)^2 \\ 
	&= \sum_{k=1}^{m} \big(\Cov_{\gamma^\ast}[ s_{\gamma^\ast,k}, \widehat{s}_{\widehat{\gamma},j} - s_{\gamma^\ast,j}] \big)^2 \\ 
	&\leq \Var_{\gamma^\ast}[\widehat{s}_{\widehat{\gamma},j} - s_{\gamma^\ast,j}]\sum_{k=1}^{m}  \Var_{\gamma^\ast}[ s_{\gamma^\ast,k}].
\end{align*}

Ergo, it follows that:
\begin{align*}
	 &\frac{1}{2} \Var_{\allDI \mid \allDII}\Big[\widehat{\E}_{\pi^\ast}^\testI[\widehat{s}_{\widehat{\gamma},j}-s_{\gamma^\ast,j}] -  \widehat{\E}_{\pi^\ast}^\trainI[\widehat{s}_{\widehat{\gamma},j}-s_{\gamma^\ast,j}] \Big]  \\
	 &\leq      \Bigg(\Bigg[\frac{1}{n^\test}+ \frac{1}{n^\train} \sum_{y=0}^{m}\frac{(\pi_y^\ast)^2}{\pi_y^\train}\Bigg]  + \frac{1}{2n^\test} \sum_{k=1}^{m}  \Var_{\gamma^\ast}[ s_{\gamma^\ast,k}]\Bigg)\Var_{\gamma^\ast}[\widehat{s}_{\widehat{\gamma},j} - s_{\gamma^\ast,j}]
\end{align*}

Thus:
\small
\begin{align*}
	&\E_{\allD}\bigg[\Bignorm{ \widehat{\E}_{\pi^\ast}^\testI[\widehat{s}_{\widehat{\gamma}}-s_{\gamma^\ast}] -  \widehat{\E}_{\pi^\ast}^\trainI[\widehat{s}_{\widehat{\gamma}}-s_{\gamma^\ast}]     }{2}^2 \bigg] \\
	&= \sum_{j=1}^{m}\E_{\allDII}\bigg\{  \Var_{\allDI \mid \allDII}\Big[\widehat{\E}_{\pi^\ast}^\testI[\widehat{s}_{\widehat{\gamma},j}-s_{\gamma^\ast,j}] -  \widehat{\E}_{\pi^\ast}^\trainI[\widehat{s}_{\widehat{\gamma},j}-s_{\gamma^\ast,j}] \Big]    \bigg\} \\ 
	 &\leq 2\sum_{j=1}^{m}\E_{\allDII}\Bigg\{  \Bigg(\Bigg[\frac{1}{n^\test}+ \frac{1}{n^\train} \sum_{y=0}^{m}\frac{(\pi_y^\ast)^2}{\pi_y^\train}\Bigg]  + \frac{1}{2n^\test} \sum_{k=1}^{m}  \Var_{\gamma^\ast}[ s_{\gamma^\ast,k}]\Bigg)\Var_{\gamma^\ast}[\widehat{s}_{\widehat{\gamma},j} - s_{\gamma^\ast,j}]    \Bigg\} \\ 
	 &= 2\Bigg(\Bigg[\frac{1}{n^\test}+ \frac{1}{n^\train} \sum_{y=0}^{m}\frac{(\pi_y^\ast)^2}{\pi_y^\train}\Bigg]  + \frac{1}{2n^\test} \sum_{k=1}^{m}  \Var_{\gamma^\ast}[ s_{\gamma^\ast,k}]\Bigg) \sum_{j=1}^{m}\E_{\allDII}\Big[\Var_{\gamma^\ast}[\widehat{s}_{\widehat{\gamma},j} - s_{\gamma^\ast,j}]  \Big]\\
	 &\leq 2\Bigg(\Bigg[\frac{1}{n^\test}+ \frac{1}{n^\train} m\frac{(1-\xi)^2}{\xi}\Bigg]  + \frac{1}{2n^\test}  \frac{m}{L^2}\Bigg) \sum_{j=1}^{m}\E_{\allDII}\Big[\Var_{\gamma^\ast}[\widehat{s}_{\widehat{\gamma},j} - s_{\gamma^\ast,j}]  \Big]\\
	 &= O\Bigg( \bigg( \frac{1}{n^\test} + \frac{1}{n^\train} \bigg)  \sum_{j=1}^{m}\E_{\allDII}\Big[\Var_{\gamma^\ast}[\widehat{s}_{\widehat{\gamma},j} - s_{\gamma^\ast,j}]  \Big] \Bigg),
\end{align*}
\normalsize
as desired!

\end{proof}
\vspace{0.4in}

\begin{proof}[\uline{Proof of Lemma \eqref{First Order Term}}]

Define the following:
\begin{itemize}
	\item $\kappa := \Var_{\DtrainI}\Big[\FisherInfo{\gamma^\ast}^{-1} \big( \widehat{\E}_{\pi^\ast}^{\testI}[s_{\gamma^\ast}]  - \widehat{\E}_{\pi^\ast}^{\trainI}[s_{\gamma^\ast}] \big) \Big] $
	\item $\phi(x) := \kappa^{-\frac{1}{2}}\FisherInfo{\gamma^\ast}^{-1}s_{\gamma^\ast}(x)$.\\
\end{itemize}

Our goal is to show that:
\begin{align*}
	&\widehat{\E}_{\pi^\ast}^{\testI}[\phi]  - \widehat{\E}_{\pi^\ast}^{\trainI}[\phi] \\
	&\equiv \bigg(\Var_{\DtrainI}\Big[\FisherInfo{\gamma^\ast}^{-1} \big( \widehat{\E}_{\pi^\ast}^{\testI}[s_{\gamma^\ast}]  - \widehat{\E}_{\pi^\ast}^{\trainI}[s_{\gamma^\ast}] \big) \Big] \bigg)^{-\frac{1}{2}} \FisherInfo{\gamma^\ast}^{-1}  \big( \widehat{\E}_{\pi^\ast}^{\testI}[s_{\gamma^\ast}]  - \widehat{\E}_{\pi^\ast}^{\trainI}[s_{\gamma^\ast}] \big) \goesto{d} \N(0,I_m).
\end{align*}

Towards that end, rewrite $\widehat{\E}_{\pi^\ast}^{\testI}[\phi]  - \widehat{\E}_{\pi^\ast}^{\trainI}[\phi] $ as follows:
\begin{align}
	\widehat{\E}_{\pi^\ast}^{\testI}[\phi]  - \widehat{\E}_{\pi^\ast}^{\trainI}[\phi] &= \big(\widehat{\E}_{\pi^\ast}^{\testI}[\phi] - \E_{\pi^\ast}[\phi]\big) +  \big(\E_{\pi^\ast}[\phi]  - \widehat{\E}_{\pi^\ast}^{\trainI}[\phi]\big) \nonumber \\ 
	&= \big(\widehat{\E}_{\pi^\ast}^{\testI}[\phi] - \E_{\pi^\ast}[\phi]\big) + \sum_{y=0}^{m} \pi_y^\ast\big(\E_{y}[\phi]  - \widehat{\E}_{y}^{\trainI}[\phi]\big) \nonumber \\ 
	&= \sum_{i=1}^{\tfrac{1}{2}n^\test} \frac{1}{\tfrac{1}{2}n^\test}   \big( \phi(X_i^{\test})- \E_{\pi^\ast}[\phi]\big) + \sum_{y=0}^{m}\sum_{i=1}^{\tfrac{1}{2}\pi_y^\train n^\train } \frac{\pi_y^\ast}{\tfrac{1}{2}\pi_y^\train n^\train} \big(\E_{y}[\phi]  - \phi(X_{i,y}^{\train}) \big) \label{YELP}
\end{align}

For convenience, let $r_w := \frac{1}{2}n^\test + \frac{1}{2}n^\train$. Construct the random variables $\{ Q_{w,j} \mid 1\leq j \leq r_w   \} $, as follows:
\begin{itemize}
	\item If $1 \leq j \leq \tfrac{1}{2}n^\test$, then $Q_{w,j} := \frac{1}{\tfrac{1}{2}n^\test}   \big( \phi(X_j^{\test})- \E_{\pi^\ast}[\phi]\big)$.
	\item If $\tfrac{1}{2}n^\test + 1 \leq j \leq \tfrac{1}{2}n^\test  + \tfrac{1}{2} \pi_0^\train n^\train$, then $Q_{w,j} :=   \frac{\pi_0^\ast}{\tfrac{1}{2}\pi_0^\train n^\train} \big(\E_{0}[\phi]  - \phi(X_{j - n^\test/2,0}^{\train}) \big)$.
	\item ....etc.
\end{itemize}

Then, by display \eqref{YELP}, it follows that $\widehat{\E}_{\pi^\ast}^{\testI}[\phi]  - \widehat{\E}_{\pi^\ast}^{\trainI}[\phi] = \sum\limits_{j = 1}^{r_w} Q_{w,j}$, so our goal is to now show that:
\[
	 \sum_{j = 1}^{r_w} Q_{w,j} \goesto{d} \N(0,I_m).
\]
Note that $\E[ Q_{w,j}] = 0$ and $\sum_{j = 1}^{r_w} \Var[Q_{w,j}] = \Var\Big[\sum_{j = 1}^{r_w} Q_{w,j} \Big] =  I_m$, so by Theorem 6.9.2 of \citep{TriangularArrayCLT}, to show the above weak convergence result, it suffices to prove that
\begin{equation}\label{Lindeberg-Feller Condition}
	\lim_{w\to\infty} \sum_{j=1}^{r_w}\E\Big[ \norm{Q_{w,j}}{2}^2 \Ibig{\norm{Q_{w,j}}{2}^2 \geq \epsilon} \Big] = 0, 
\end{equation}
for every $\epsilon \in (0,\infty)$. But how do we show this? Well, start by examining $Q_{w,j}$. First, suppose that $1 \leq j \leq \tfrac{1}{2} n^\test$. Then:
\begin{align*}
	Q_{w,j} &=  \frac{1}{\tfrac{1}{2}n^\test}   \big( \phi(X_j^{\test})- \E_{\pi^\ast}[\phi]\big) \\ 
	&= \frac{1}{\tfrac{1}{2}n^\test}   \kappa^{-\frac{1}{2}}\FisherInfo{\gamma^\ast}^{-1}  \big( s_{\gamma^\ast}(X_j^{\test})- \E_{\pi^\ast}[ s_{\gamma^\ast}]\big).
\end{align*}
\begin{align*}
	 &\norm{Q_{w,j}}{2}^2 \\
	&=  \frac{1}{(\tfrac{1}{2}n^\test)^2} \big( s_{\gamma^\ast}(X_j^{\test})- \E_{\pi^\ast}[ s_{\gamma^\ast}]\big)^{T} \FisherInfo{\gamma^\ast}^{-1} \kappa^{-\frac{1}{2}} \kappa^{-\frac{1}{2}}\FisherInfo{\gamma^\ast}^{-1} \big( s_{\gamma^\ast}(X_j^{\test})- \E_{\pi^\ast}[ s_{\gamma^\ast}]\big) \\
	&=  \frac{1}{(\tfrac{1}{2}n^\test)^2} \big( s_{\gamma^\ast}(X_j^{\test})- \E_{\pi^\ast}[ s_{\gamma^\ast}]\big)^{T} \FisherInfo{\gamma^\ast}^{-1} \kappa^{-1}\FisherInfo{\gamma^\ast}^{-1} \big( s_{\gamma^\ast}(X_j^{\test})- \E_{\pi^\ast}[ s_{\gamma^\ast}]\big) \\
	&=  \frac{1}{(\tfrac{1}{2}n^\test)^2} \big( s_{\gamma^\ast}(X_j^{\test})- \E_{\pi^\ast}[ s_{\gamma^\ast}]\big)^{T} \Big(\FisherInfo{\gamma^\ast} \kappa\FisherInfo{\gamma^\ast} \Big)^{-1} \big( s_{\gamma^\ast}(X_j^{\test})- \E_{\pi^\ast}[ s_{\gamma^\ast}]\big) \\
	&=  \frac{1}{(\tfrac{1}{2}n^\test)^2} \big( s_{\gamma^\ast}(X_j^{\test})- \E_{\pi^\ast}[ s_{\gamma^\ast}]\big)^{T} \bigg(    \Var_{\DtrainI}\Big[ \widehat{\E}_{\pi^\ast}^{\testI}[s_{\gamma^\ast}]  - \widehat{\E}_{\pi^\ast}^{\trainI}[s_{\gamma^\ast}]  \Big]    \bigg)^{-1} \big( s_{\gamma^\ast}(X_j^{\test})- \E_{\pi^\ast}[ s_{\gamma^\ast}]\big).
\end{align*}

Now, note that:
\begin{align*}
	 \Var_{\DtrainI}\Big[ \widehat{\E}_{\pi^\ast}^{\testI}[s_{\gamma^\ast}]  - \widehat{\E}_{\pi^\ast}^{\trainI}[s_{\gamma^\ast}]  \Big] &= \frac{1}{\tfrac{1}{2}n^\test}\Var_{\pi^\ast}[s_{\gamma^\ast}] + \frac{1}{\tfrac{1}{2}n^\train} \sum_{y=0}^{m} \frac{(\pi_y^\ast)^2}{\pi_y^\train}\Var_y[s_{\gamma^\ast}] \\ 
	 &\succeq \frac{1}{\tfrac{1}{2}n^\test}\Var_{\pi^\ast}[s_{\gamma^\ast}].
\end{align*}

By Lemma \eqref{Bounds on Test Mixture Eigenvalues}, we know that $\Var_{\pi^\ast}[s_{\gamma^\ast}] \succ 0$, so its inverse exists. Thus, we have that:
\[
	\implies \bigg( \Var_{\DtrainI}\Big[ \widehat{\E}_{\pi^\ast}^{\testI}[s_{\gamma^\ast}]  - \widehat{\E}_{\pi^\ast}^{\trainI}[s_{\gamma^\ast}]  \Big] \bigg)^{-1} \preceq  \tfrac{1}{2}n^\test \big(\Var_{\pi^\ast}[s_{\gamma^\ast}]\big)^{-1}.
\]
\begin{align*}
	\implies \norm{Q_{w,j}}{2}^2 &\leq \frac{\tfrac{1}{2}n^\test}{(\tfrac{1}{2}n^\test)^2} \big( s_{\gamma^\ast}(X_j^{\test})- \E_{\pi^\ast}[ s_{\gamma^\ast}]\big)^{T}    \big(\Var_{\pi^\ast}[s_{\gamma^\ast}]\big)^{-1} \big( s_{\gamma^\ast}(X_j^{\test})- \E_{\pi^\ast}[ s_{\gamma^\ast}]\big) \\ 
	&= \frac{1}{\tfrac{1}{2}n^\test} \big( s_{\gamma^\ast}(X_j^{\test})- \E_{\pi^\ast}[ s_{\gamma^\ast}]\big)^{T}    \big(\Var_{\pi^\ast}[s_{\gamma^\ast}]\big)^{-1} \big( s_{\gamma^\ast}(X_j^{\test})- \E_{\pi^\ast}[ s_{\gamma^\ast}]\big) \\ 
	&\leq  \frac{1}{\tfrac{1}{2}n^\test} \norm{s_{\gamma^\ast}(X_j^{\test})- \E_{\pi^\ast}[ s_{\gamma^\ast}]}{2}^2  \maxEval{ \Var_{\pi^\ast}[s_{\gamma^\ast}]^{-1} }\\
	&=  \frac{1}{\tfrac{1}{2}n^\test} \norm{s_{\gamma^\ast}(X_j^{\test})- \E_{\pi^\ast}[ s_{\gamma^\ast}]}{2}^2  / \minEval{ \Var_{\pi^\ast}[s_{\gamma^\ast}]}\\ 
	&\leq \frac{1}{\tfrac{1}{2}n^\test} \norm{s_{\gamma^\ast}(X_j^{\test})- \E_{\pi^\ast}[ s_{\gamma^\ast}]}{2}^2  \cdot \frac{m+1}{\Lambda\xi} \\ 
	&\leq \frac{1}{\tfrac{1}{2}n^\test} \cdot \frac{16m(m+1)}{L^2\Lambda\xi},
\end{align*}
where the second to last inequality uses Lemma  \eqref{Bounds on Test Mixture Eigenvalues}. Next, let's handle the case when $j \in [\tfrac{1}{2}n^\test + 1, \tfrac{1}{2}n^\test  + \tfrac{1}{2} \pi_0^\train n^\train]$. Observe that: 
\begin{align*}
	 &\frac{1}{2}\minEvalBigg{\Var_{\DtrainI}\Big[ \widehat{\E}_{\pi^\ast}^{\testI}[s_{\gamma^\ast}]  - \widehat{\E}_{\pi^\ast}^{\trainI}[s_{\gamma^\ast}]  \Big] } \\
	 &\geq \Bigg[\frac{1}{n^\train} \sum_{y=0}^{m}\frac{(\pi_y^\ast)^2}{\pi_y^\train}\Bigg] \minEval{	\FisherInfo{\gamma^\ast} -  \FisherInfo{\gamma^\ast} \FisherInfo{\gamma^\ast;\textup{Cat}}^{-1}  \FisherInfo{\gamma^\ast} } \\ 
	 &\geq \frac{m\xi^2}{n^\train}  \minEval{ \FisherInfo{\gamma^\ast} [\FisherInfo{\gamma^\ast}^{-1} -  \FisherInfo{\gamma^\ast;\textup{Cat}}^{-1}]  \FisherInfo{\gamma^\ast} },
\end{align*}
where the first line is due to  Lemmas \eqref{Bounds on Eigenvalues of Categorical Fisher Info} and \eqref{First Order Constant for Any Matching Function}, and the second line is due to Assumption \eqref{bounded pis}. Now, note that due to Lemma \eqref{Harder Problem}, the difference $\FisherInfo{\gamma^\ast}^{-1} -  \FisherInfo{\gamma^\ast;\textup{Cat}}^{-1}$ is positive semi-definite. Since the matrix is also symmetric, it follows that the eigenvalues and singular values coincide; thus, by Lemma \eqref{Inverse FIM Difference is Bounded}, we know that all the eigenvalues are strictly positive. Hence, $\FisherInfo{\gamma^\ast}^{-1} -  \FisherInfo{\gamma^\ast;\textup{Cat}}^{-1}$ is symmetric positive definite. Since $\FisherInfo{\gamma^\ast}$ is symmetric positive definite by Assumption \eqref{min Fisher Info}, it follows that the product $ \FisherInfo{\gamma^\ast} [\FisherInfo{\gamma^\ast}^{-1} -  \FisherInfo{\gamma^\ast;\textup{Cat}}^{-1}]  \FisherInfo{\gamma^\ast} $ is symmetric positive definite as well. Thus, the eigenvalues and singular values coincide, meaning that:
\begin{align*}
	&\minEval{ \FisherInfo{\gamma^\ast} [\FisherInfo{\gamma^\ast}^{-1} -  \FisherInfo{\gamma^\ast;\textup{Cat}}^{-1}]  \FisherInfo{\gamma^\ast} } \\&=\minSing{ \FisherInfo{\gamma^\ast} [\FisherInfo{\gamma^\ast}^{-1} -  \FisherInfo{\gamma^\ast;\textup{Cat}}^{-1}]  \FisherInfo{\gamma^\ast} } \\ 
	&\geq \minSing{ \FisherInfo{\gamma^\ast}} \minSing{\FisherInfo{\gamma^\ast}^{-1} -  \FisherInfo{\gamma^\ast;\textup{Cat}}^{-1}} \minSing{ \FisherInfo{\gamma^\ast} } \\
	&\geq \Lambda \nu \bigg( \frac{L}{m+1}\bigg)^2,
\end{align*}
where the last inequality is due to Assumption \eqref{min Fisher Info} and Lemma \eqref{Inverse FIM Difference is Bounded}. Ergo, it follows that:
\[
	\implies \minEvalBigg{\Var_{\DtrainI}\Big[ \widehat{\E}_{\pi^\ast}^{\testI}[s_{\gamma^\ast}]  - \widehat{\E}_{\pi^\ast}^{\trainI}[s_{\gamma^\ast}]  \Big] } \geq 2 \frac{m\xi^2}{n^\train}  \Lambda \nu \bigg( \frac{L}{m+1}\bigg)^2.
\]
Thus, $\Var_{\DtrainI}\Big[ \widehat{\E}_{\pi^\ast}^{\testI}[s_{\gamma^\ast}]  - \widehat{\E}_{\pi^\ast}^{\trainI}[s_{\gamma^\ast}]  \Big] $ is positive definite. Now, using our existing work in this proof, we have that:
\begin{align*}
	&\norm{Q_{w,j}}{2}^2\\
	 &= \frac{(\pi_0^\ast)^2}{\big(\tfrac{1}{2}\pi_0^\train n^\train \big)^2} \big( s_{\gamma^\ast}(X_{j - n^\test/2,0}^{\train})  - \E_{0}[ s_{\gamma^\ast}] \big)^{T}  \Bigg(  \Var_{\DtrainI}\Big[ \widehat{\E}_{\pi^\ast}^{\testI}[s_{\gamma^\ast}]  - \widehat{\E}_{\pi^\ast}^{\trainI}[s_{\gamma^\ast}]  \Big]  \Bigg)^{-1}\big( s_{\gamma^\ast}(X_{j - n^\test/2,0}^{\train})  - \E_{0}[ s_{\gamma^\ast}] \big) \\ 
	&\leq \frac{(\pi_0^\ast)^2}{\big(\tfrac{1}{2}\pi_0^\train n^\train \big)^2} \bignorm{ s_{\gamma^\ast}(X_{j - n^\test/2,0}^{\train})  - \E_{0}[ s_{\gamma^\ast}] }{2}^2 \maxEvalBigg{\Var_{\DtrainI}\Big[ \widehat{\E}_{\pi^\ast}^{\testI}[s_{\gamma^\ast}]  - \widehat{\E}_{\pi^\ast}^{\trainI}[s_{\gamma^\ast}]  \Big]^{-1} } \\ 
	&= \frac{(\pi_0^\ast)^2}{\big(\tfrac{1}{2}\pi_0^\train\big)^2 (n^\train)^2 } \bignorm{ s_{\gamma^\ast}(X_{j - n^\test/2,0}^{\train})  - \E_{0}[ s_{\gamma^\ast}] }{2}^2 /  \minEvalBigg{\Var_{\DtrainI}\Big[ \widehat{\E}_{\pi^\ast}^{\testI}[s_{\gamma^\ast}]  - \widehat{\E}_{\pi^\ast}^{\trainI}[s_{\gamma^\ast}]  \Big] } \\ 
	&\leq \frac{(\pi_0^\ast)^2}{\big(\tfrac{1}{2}\pi_0^\train\big)^2 (n^\train)^2 } \bignorm{ s_{\gamma^\ast}(X_{j - n^\test/2,0}^{\train})  - \E_{0}[ s_{\gamma^\ast}] }{2}^2 \frac{1}{2 \frac{m\xi^2}{n^\train}  \Lambda \nu \big( \frac{L}{m+1}\big)^2} \\ 
	&\leq \frac{(\pi_0^\ast)^2}{\big(\tfrac{1}{2}\pi_0^\train\big)^2 n^\train } \cdot m\frac{16}{L^2} \cdot \frac{1}{2 m\xi^2  \Lambda \nu} \bigg( \frac{m+1}{L}\bigg)^2 \\ 
	&\leq \frac{1}{ \tfrac{1}{2}\pi_0^\train n^\train } \cdot   \frac{16}{L^2\xi^3  \Lambda \nu} \bigg( \frac{m+1}{L}\bigg)^2.
\end{align*}

A similar argument can be used to bound $\norm{Q_{w,j}}{2}^2$ for $j >  \tfrac{1}{2}n^\test  + \tfrac{1}{2} \pi_0^\train n^\train$. It follows that, for each $\epsilon > 0$,
\begin{align*}
	 &\sum_{j=1}^{r_w}\E\Big[ \norm{Q_{w,j}}{2}^2 \Ibig{\norm{Q_{w,j}}{2}^2 \geq \epsilon} \Big]  \\
	 &\leq \frac{16m(m+1)}{L^2\Lambda\xi} \Ibigg{\frac{16m(m+1)}{L^2\Lambda\xi}  \geq \tfrac{1}{2}n^\test\epsilon } \\
	&\qquad + \sum_{y=0}^{m} \frac{16}{L^2\xi^3  \Lambda \nu} \bigg( \frac{m+1}{L}\bigg)^2 \IBigg{ \frac{16}{L^2\xi^3  \Lambda \nu} \bigg( \frac{m+1}{L}\bigg)^2 \geq \tfrac{1}{2}\pi_y^\train n^\train \epsilon },
\end{align*}
which clearly goes to zero as $w\to\infty$, meaning we've proven the desired convergence result.


\end{proof}
\vspace{0.4in}

\begin{proof}[\uline{Proof of Theorem \eqref{Error Rate of First Step Estimator}}]

Note that:
\small
\begin{align*}
	\widehat{\pi}^{(a)} - \pi^\ast &= (\widehat{\pi}^{(a)} - \pi^\ast) \Ibig{\minSing{\widehat{A}} > 0 }  +   (\widehat{\pi}^{(a)} - \pi^\ast) \Ibig{\minSing{\widehat{A}} = 0 } \\ 
	&= (\widehat{\pi}^{(a)} - \pi^\ast) \Ibig{\minSing{\widehat{A}} > 0 }  +   \Big(\frac{1}{m+1}\vec{1}_m - \pi^\ast \Big) \Ibig{\minSing{\widehat{A}} = 0 }.
\end{align*}
\normalsize

And also that:
\small
\begin{align*}
	&(\widehat{\pi}^{(a)} - \pi^\ast) \Ibig{\minSing{\widehat{A}} > 0 } \\
	&= \Big[\widehat{A}^{-1} (\widehat{\E}_{\pi^\ast}^\testI[\widehat{s}_{\widehat{\gamma}}] - \widehat{\E}_0^\trainI[\widehat{s}_{\widehat{\gamma}}]) -   \pi^\ast  \Big]  \Ibig{\minSing{\widehat{A}} > 0 } \\
	&= \widehat{A}^{-1}\Big[(\widehat{\E}_{\pi^\ast}^\testI[\widehat{s}_{\widehat{\gamma}}] - \widehat{\E}_0^\trainI[\widehat{s}_{\widehat{\gamma}}]) -  \widehat{A} \pi^\ast  \Big]  \Ibig{\minSing{\widehat{A}} > 0 } \\
	&= \widehat{A}^{-1}\Bigg[(\widehat{\E}_{\pi^\ast}^\testI[\widehat{s}_{\widehat{\gamma}}] - \widehat{\E}_0^\trainI[\widehat{s}_{\widehat{\gamma}}]) -  \sum_{y=1}^{m}\pi_y^{\ast}(\widehat{\E}_y^\trainI[\widehat{s}_{\widehat{\gamma}}] - \widehat{\E}_0^\trainI[\widehat{s}_{\widehat{\gamma}}]) \Bigg]  \Ibig{\minSing{\widehat{A}} > 0 }  \\
	&=  \widehat{A}^{-1}(\widehat{\E}_{\pi^\ast}^\testI[\widehat{s}_{\widehat{\gamma}}] -  \widehat{\E}_{\pi^\ast}^\trainI[\widehat{s}_{\widehat{\gamma}}] ) \Ibig{\minSing{\widehat{A}} > 0 } \\ 
	&=  \widehat{A}^{-1}\bigg[ \big(\widehat{\E}_{\pi^\ast}^\testI[s_{\gamma^\ast}] -  \widehat{\E}_{\pi^\ast}^\trainI[s_{\gamma^\ast}]\big)  +\big(\widehat{\E}_{\pi^\ast}^\testI[\widehat{s}_{\widehat{\gamma}}-s_{\gamma^\ast}] -  \widehat{\E}_{\pi^\ast}^\trainI[\widehat{s}_{\widehat{\gamma}}-s_{\gamma^\ast}] \big)   \bigg] \Ibig{\minSing{\widehat{A}} > 0 } \\ \\ 
	&= \FisherInfo{\gamma^\ast}^{-1}\bigg[ \big(\widehat{\E}_{\pi^\ast}^\testI[s_{\gamma^\ast}] -  \widehat{\E}_{\pi^\ast}^\trainI[s_{\gamma^\ast}]\big)  +\big(\widehat{\E}_{\pi^\ast}^\testI[\widehat{s}_{\widehat{\gamma}}-s_{\gamma^\ast}] -  \widehat{\E}_{\pi^\ast}^\trainI[\widehat{s}_{\widehat{\gamma}}-s_{\gamma^\ast}] \big)   \bigg] \Ibig{\minSing{\widehat{A}} > 0 } \\ 
	&\quad + (\widehat{A}^{-1} - \FisherInfo{\gamma^\ast}^{-1})\bigg[ \big(\widehat{\E}_{\pi^\ast}^\testI[s_{\gamma^\ast}] -  \widehat{\E}_{\pi^\ast}^\trainI[s_{\gamma^\ast}]\big)  +\big(\widehat{\E}_{\pi^\ast}^\testI[\widehat{s}_{\widehat{\gamma}}-s_{\gamma^\ast}] -  \widehat{\E}_{\pi^\ast}^\trainI[\widehat{s}_{\widehat{\gamma}}-s_{\gamma^\ast}] \big)   \bigg] \Ibig{\minSing{\widehat{A}} > 0 } \\ \\ 
	&= \FisherInfo{\gamma^\ast}^{-1} \big(\widehat{\E}_{\pi^\ast}^\testI[s_{\gamma^\ast}] -  \widehat{\E}_{\pi^\ast}^\trainI[s_{\gamma^\ast}]\big) \Ibig{\minSing{\widehat{A}} > 0 }  \\
	&\quad + \FisherInfo{\gamma^\ast}^{-1} \big(\widehat{\E}_{\pi^\ast}^\testI[\widehat{s}_{\widehat{\gamma}}-s_{\gamma^\ast}] -  \widehat{\E}_{\pi^\ast}^\trainI[\widehat{s}_{\widehat{\gamma}}-s_{\gamma^\ast}] \big)   \Ibig{\minSing{\widehat{A}} > 0 } \\ 
	&\quad + (\widehat{A}^{-1} - \FisherInfo{\gamma^\ast}^{-1})\bigg[ \big(\widehat{\E}_{\pi^\ast}^\testI[s_{\gamma^\ast}] -  \widehat{\E}_{\pi^\ast}^\trainI[s_{\gamma^\ast}]\big)  +\big(\widehat{\E}_{\pi^\ast}^\testI[\widehat{s}_{\widehat{\gamma}}-s_{\gamma^\ast}] -  \widehat{\E}_{\pi^\ast}^\trainI[\widehat{s}_{\widehat{\gamma}}-s_{\gamma^\ast}] \big)   \bigg] \Ibig{\minSing{\widehat{A}} > 0 } \\ \\ 
	&= \FisherInfo{\gamma^\ast}^{-1} \big(\widehat{\E}_{\pi^\ast}^\testI[s_{\gamma^\ast}] -  \widehat{\E}_{\pi^\ast}^\trainI[s_{\gamma^\ast}]\big)  - \FisherInfo{\gamma^\ast}^{-1} \big(\widehat{\E}_{\pi^\ast}^\testI[s_{\gamma^\ast}] -  \widehat{\E}_{\pi^\ast}^\trainI[s_{\gamma^\ast}]\big) \Ibig{\minSing{\widehat{A}} = 0 }  \\
	&\quad + \FisherInfo{\gamma^\ast}^{-1} \big(\widehat{\E}_{\pi^\ast}^\testI[\widehat{s}_{\widehat{\gamma}}-s_{\gamma^\ast}] -  \widehat{\E}_{\pi^\ast}^\trainI[\widehat{s}_{\widehat{\gamma}}-s_{\gamma^\ast}] \big)   \Ibig{\minSing{\widehat{A}} > 0 } \\ 
	&\quad + (\widehat{A}^{-1} - \FisherInfo{\gamma^\ast}^{-1})\bigg[ \big(\widehat{\E}_{\pi^\ast}^\testI[s_{\gamma^\ast}] -  \widehat{\E}_{\pi^\ast}^\trainI[s_{\gamma^\ast}]\big)  +\big(\widehat{\E}_{\pi^\ast}^\testI[\widehat{s}_{\widehat{\gamma}}-s_{\gamma^\ast}] -  \widehat{\E}_{\pi^\ast}^\trainI[\widehat{s}_{\widehat{\gamma}}-s_{\gamma^\ast}] \big)   \bigg] \Ibig{\minSing{\widehat{A}} > 0 }.
\end{align*}
\normalsize

Thus, we have that:
\small
\begin{align*}
	\widehat{\pi}^{(a)} - \pi^\ast &=  \FisherInfo{\gamma^\ast}^{-1} \big(\widehat{\E}_{\pi^\ast}^\testI[s_{\gamma^\ast}] -  \widehat{\E}_{\pi^\ast}^\trainI[s_{\gamma^\ast}]\big)  - \FisherInfo{\gamma^\ast}^{-1} \big(\widehat{\E}_{\pi^\ast}^\testI[s_{\gamma^\ast}] -  \widehat{\E}_{\pi^\ast}^\trainI[s_{\gamma^\ast}]\big) \Ibig{\minSing{\widehat{A}} = 0 }  \\
	&\quad + \FisherInfo{\gamma^\ast}^{-1} \big(\widehat{\E}_{\pi^\ast}^\testI[\widehat{s}_{\widehat{\gamma}}-s_{\gamma^\ast}] -  \widehat{\E}_{\pi^\ast}^\trainI[\widehat{s}_{\widehat{\gamma}}-s_{\gamma^\ast}] \big)   \Ibig{\minSing{\widehat{A}} > 0 } \\ 
	&\quad + (\widehat{A}^{-1} - \FisherInfo{\gamma^\ast}^{-1})\bigg[ \big(\widehat{\E}_{\pi^\ast}^\testI[s_{\gamma^\ast}] -  \widehat{\E}_{\pi^\ast}^\trainI[s_{\gamma^\ast}]\big)  +\big(\widehat{\E}_{\pi^\ast}^\testI[\widehat{s}_{\widehat{\gamma}}-s_{\gamma^\ast}] -  \widehat{\E}_{\pi^\ast}^\trainI[\widehat{s}_{\widehat{\gamma}}-s_{\gamma^\ast}] \big)   \bigg] \Ibig{\minSing{\widehat{A}} > 0 } \\
	&\quad + \Big(\frac{1}{m+1}\vec{1}_m - \pi^\ast \Big) \Ibig{\minSing{\widehat{A}} = 0 }.
\end{align*}
\normalsize

Let's handle each of the terms on the RHS above, one at a time. For convenience, employ the shorthand:
\small
\begin{itemize}
	\item $W_1 := \FisherInfo{\gamma^\ast}^{-1} \big(\widehat{\E}_{\pi^\ast}^\testI[s_{\gamma^\ast}] -  \widehat{\E}_{\pi^\ast}^\trainI[s_{\gamma^\ast}]\big)$
	\item $W_2 := -\FisherInfo{\gamma^\ast}^{-1} \big(\widehat{\E}_{\pi^\ast}^\testI[s_{\gamma^\ast}] -  \widehat{\E}_{\pi^\ast}^\trainI[s_{\gamma^\ast}]\big) \Ibig{\minSing{\widehat{A}} = 0 }$
	\item $W_3 := \FisherInfo{\gamma^\ast}^{-1} \big(\widehat{\E}_{\pi^\ast}^\testI[\widehat{s}_{\widehat{\gamma}}-s_{\gamma^\ast}] -  \widehat{\E}_{\pi^\ast}^\trainI[\widehat{s}_{\widehat{\gamma}}-s_{\gamma^\ast}] \big)   \Ibig{\minSing{\widehat{A}} > 0 }$
	\item $W_4 :=  (\widehat{A}^{-1} - \FisherInfo{\gamma^\ast}^{-1})\big(\widehat{\E}_{\pi^\ast}^\testI[s_{\gamma^\ast}] -  \widehat{\E}_{\pi^\ast}^\trainI[s_{\gamma^\ast}]\big)\Ibig{\minSing{\widehat{A}} > 0 }$
	\item $W_5 := (\widehat{A}^{-1} - \FisherInfo{\gamma^\ast}^{-1})\big(\widehat{\E}_{\pi^\ast}^\testI[\widehat{s}_{\widehat{\gamma}}-s_{\gamma^\ast}] -  \widehat{\E}_{\pi^\ast}^\trainI[\widehat{s}_{\widehat{\gamma}}-s_{\gamma^\ast}] \big)\Ibig{\minSing{\widehat{A}} > 0 }$
	\item $W_6 := \Big(\frac{1}{m+1}\vec{1}_m - \pi^\ast \Big) \Ibig{\minSing{\widehat{A}} = 0}$,
\end{itemize}
\normalsize
so that $\widehat{\pi}^{(a)} - \pi^\ast = \sum\limits_{l = 1}^{6} W_l$. \\ \\

\uline{Let's start with $W_1$}. Observe that:
\small
\begin{align*}
	&W_1 \\
	&=  \FisherInfo{\gamma^\ast}^{-1} \big(\widehat{\E}_{\pi^\ast}^\testI[s_{\gamma^\ast}] -  \widehat{\E}_{\pi^\ast}^\trainI[s_{\gamma^\ast}]\big) \\ 
	&= \bigg(  \Var_{\allDI}\Big[\FisherInfo{\gamma^\ast}^{-1}\big(\widehat{\E}_{\pi^\ast}^{\testI}[s_{\gamma^\ast}] - \widehat{\E}_{\pi^\ast}^{\trainI}[s_{\gamma^\ast}] \big) \Big] \bigg)^{\frac{1}{2}}    \bigg(  \Var_{\allDI}\Big[\FisherInfo{\gamma^\ast}^{-1}\big(\widehat{\E}_{\pi^\ast}^{\testI}[s_{\gamma^\ast}] - \widehat{\E}_{\pi^\ast}^{\trainI}[s_{\gamma^\ast}] \big) \Big] \bigg)^{-\frac{1}{2}}\FisherInfo{\gamma^\ast}^{-1} \big(\widehat{\E}_{\pi^\ast}^\testI[s_{\gamma^\ast}] -  \widehat{\E}_{\pi^\ast}^\trainI[s_{\gamma^\ast}]\big) \\ 
	&\equiv \bigg(  \Var_{\allDI}\Big[\FisherInfo{\gamma^\ast}^{-1}\big(\widehat{\E}_{\pi^\ast}^{\testI}[s_{\gamma^\ast}] - \widehat{\E}_{\pi^\ast}^{\trainI}[s_{\gamma^\ast}] \big) \Big] \bigg)^{\frac{1}{2}} Z^{(a)} \\ 
	&= \Bigg( 2\Bigg[\frac{1}{n^\test}+ \frac{1}{n^\train} \sum_{y=0}^{m}\frac{(\pi_y^\ast)^2}{\pi_y^\train}\Bigg] \Big( \FisherInfo{\gamma^\ast}^{-1} - \FisherInfo{\gamma^\ast;\textup{Cat}}^{-1}  \Big)  + 2\frac{1}{n^\test} \FisherInfo{\pi^\ast;\textup{Cat}}^{-1} \Bigg)^{\frac{1}{2}} Z^{(a)} \\
	&= \sqrt{2}\Bigg( \Bigg[\frac{1}{n^\test}+ \frac{1}{n^\train} \sum_{y=0}^{m}\frac{(\pi_y^\ast)^2}{\pi_y^\train}\Bigg] \Big( \FisherInfo{\gamma^\ast}^{-1} - \FisherInfo{\gamma^\ast;\textup{Cat}}^{-1}  \Big)  + \frac{1}{n^\test} \FisherInfo{\pi^\ast;\textup{Cat}}^{-1} \Bigg)^{\frac{1}{2}} Z^{(a)},
\end{align*}
\normalsize
where the second to last line is by Corollary \eqref{First Order Constant When Matching Function is Score Function}, and under Assumptions \eqref{FIM Difference is Bounded}, \eqref{bounded pis}, and \eqref{min Fisher Info}, we have that $Z^{(a)} \goesto{d} \N(0,I_{m})$ by Lemma \eqref{First Order Term}. We also take this opportunity to note that $W_1  = \frac{1}{n/2} \sum_{i=1}^{n/2} \psi^\eff(Z_i)$ by Lemma \eqref{Alternative Expression for First Order Error}, and so the claim that
\[
	\sqrt{2}\Bigg( \Bigg[\frac{1}{n^\test}+ \frac{1}{n^\train} \sum_{y=0}^{m}\frac{(\pi_y^\ast)^2}{\pi_y^\train}\Bigg] \Big( \FisherInfo{\gamma^\ast}^{-1} - \FisherInfo{\gamma^\ast;\textup{Cat}}^{-1}  \Big)  + \frac{1}{n^\test} \FisherInfo{\pi^\ast;\textup{Cat}}^{-1} \Bigg)^{\frac{1}{2}} Z^{(a)} =  \frac{1}{n/2} \sum_{i=1}^{n/2} \psi^\eff(Z_i)
\] 
is indeed true.\\ \\

\uline{Next, let's tackle $W_2$}. For some finite global constant $C > 0$, we have that:
\small
\begin{align*}
	\norm{W_2}{2} &\leq \maxSing{\FisherInfo{\gamma^\ast}^{-1}} \bignorm{\widehat{\E}_{\pi^\ast}^\testI[s_{\gamma^\ast}] -  \widehat{\E}_{\pi^\ast}^\trainI[s_{\gamma^\ast}]}{2}  \Ibig{\minSing{\widehat{A}} = 0 } \\ 
	&\leq \frac{1}{\sqrt{\Lambda}} \bignorm{\widehat{\E}_{\pi^\ast}^\testI[s_{\gamma^\ast}] -  \widehat{\E}_{\pi^\ast}^\trainI[s_{\gamma^\ast}]}{2}   \Ibig{\minSing{\widehat{A}} = 0 }\\
	&= O_{\P}\bigg(\sqrt{\frac{1}{n^\test} + \frac{1}{n^\train}}\bigg)  \Ibig{\minSing{\widehat{A}} = 0 } \\ 
	&= O_{\P}\bigg(\sqrt{\frac{1}{n^\test} + \frac{1}{n^\train}}\bigg) O_{\P}\bigg( e^{-C (\min_y\lambda_y)^2 n^\train}  + \P_{\allDII}\Big[\bignorm{ \widehat{\gamma} - \gamma^\ast }{2} \geq C \Big]  \bigg),
\end{align*}
\normalsize
where the third line uses Popoviciu's variance inequality with the fact that the magnitude of the components of $s_{\gamma^\ast}$ are bounded by a finite, global constant, and the fourth line is a consequence of the tail bounds in Lemmas \eqref{Uniform Deviation Tail Bounds}, \eqref{Rate for Learning Diagonal of Fisher Information Matrix}, \eqref{Rate for Learning the Score Function}, \eqref{Rate for Learning Diagonal of Fisher Information Matrix, Unknown Gamma Star}, \eqref{Rate for Learning the Score Function, Unknown Gamma Star}, \eqref{Rate for Learning Columns of Fisher Information Matrix} and \eqref{Learning the Inverse FIM},   as well as the properties of $\min_y \lambda_y$ in Assumption \eqref{lambda and uniform convergence properties}.  \\ \\

\uline{Next, let's get control of $W_3$}. Observe that:

 \small
\begin{align*}
	&\norm{W_3}{2} \\
	&\leq \maxSing{\FisherInfo{\gamma^\ast}^{-1}}\bignorm{ \widehat{\E}_{\pi^\ast}^\testI[\widehat{s}_{\widehat{\gamma}}-s_{\gamma^\ast}] -  \widehat{\E}_{\pi^\ast}^\trainI[\widehat{s}_{\widehat{\gamma}}-s_{\gamma^\ast}] }{2} \\ 
	&\leq \frac{1}{\sqrt{\Lambda}} \bignorm{ \widehat{\E}_{\pi^\ast}^\testI[\widehat{s}_{\widehat{\gamma}}-s_{\gamma^\ast}] -  \widehat{\E}_{\pi^\ast}^\trainI[\widehat{s}_{\widehat{\gamma}}-s_{\gamma^\ast}] }{2} \\ 
	&= O_{\P}\Bigg(\sqrt{\bigg( \frac{1}{n^\test} + \frac{1}{n^\train} \bigg)  \sum_{y=1}^{m}\E_{\allDII}\Big[\Var_{\gamma^\ast}[\widehat{s}_{\widehat{\gamma},y} - s_{\gamma^\ast,y}]  \Big] }\Bigg) \\
	&= O_{\P}\Bigg(\sqrt{\bigg( \frac{1}{n^\test} + \frac{1}{n^\train} \bigg)  \sum_{y=1}^{m}\bigg\{ \E_{\allDII}[\widetilde{T}_y] + \E_{\DtrainII}\Big[ \U_y(\lambda_y)\Big] + \sqrt{\E_{\allDII}\Big[ \norm{\widehat{\gamma} - \gamma^\ast}{2}^2 \Big]} +  \E_{\DtrainII}\Big[ \Var_{\gamma^\ast}[\widehat{s}_{\gamma^\ast,y} - s_{\gamma^\ast,y}] \Big] \bigg\}   }\Bigg) \\ 
	&= O_{\P}\Bigg(\sqrt{\bigg( \frac{1}{n^\test} + \frac{1}{n^\train} \bigg)  \sum_{y=1}^{m}\bigg\{ \E_{\DtrainII}[T_y] + \E_{\DtrainII}\Big[ \U_y(\lambda_y)\Big] + \sqrt{\E_{\allDII}\Big[ \norm{\widehat{\gamma} - \gamma^\ast}{2}^2 \Big]} +  \E_{\DtrainII}\Big[ \Var_{\gamma^\ast}[\widehat{s}_{\gamma^\ast,y} - s_{\gamma^\ast,y}] \Big] \bigg\}   }\Bigg) \\ 
	&= O_{\P}\Bigg(\sqrt{\bigg( \frac{1}{n^\test} + \frac{1}{n^\train} \bigg)  \sum_{y=1}^{m}\bigg\{ \E_{\DtrainII}[T_y] + \E_{\DtrainII}\Big[ \U_y(\lambda_y)\Big] + \sqrt{\E_{\allDII}\Big[ \norm{\widehat{\gamma} - \gamma^\ast}{2}^2 \Big]}  \bigg\}   }\Bigg) \\ 
	&= O_{\P}\Bigg(\sqrt{\bigg( \frac{1}{n^\test} + \frac{1}{n^\train} \bigg)  \sum_{y=1}^{m}\bigg\{ \frac{1}{\sqrt{n^\train}} + \lambda_y \Omega(s_{\gamma^\ast,y})^2 + \frac{1}{\sqrt{\lambda_y}}e^{-Cn^\train} + \E_{\DtrainII}\Big[ \U_y(\lambda_y)\Big] + \sqrt{\E_{\allDII}\Big[ \norm{\widehat{\gamma} - \gamma^\ast}{2}^2 \Big]}  \bigg\}   }\Bigg),
 \end{align*}
 \normalsize
where the third line follows from Lemma \eqref{Rate of Second Order Term}, the fourth line follows from Lemma \eqref{Rate for Learning the Score Function, Unknown Gamma Star}, the fifth line follows from Lemma \eqref{Rate for Learning Diagonal of Fisher Information Matrix, Unknown Gamma Star}, the sixth line follows from Lemma \eqref{Rate for Learning the Score Function}, and the seventh line follows from Lemma \eqref{Rate for Learning Diagonal of Fisher Information Matrix}.  \\ \\ 

\uline{Next, let's examine $W_4$}. Observe that:
\small
\begin{align*}
	&\norm{W_4}{2} \\
	&\leq \maxSing{\widehat{A}^{-1} - \FisherInfo{\gamma^\ast}^{-1}}\Ibig{\minSing{\widehat{A}} > 0 } \cdot  \bignorm{ \widehat{\E}_{\pi^\ast}^\testI[s_{\gamma^\ast}] -  \widehat{\E}_{\pi^\ast}^\trainI[s_{\gamma^\ast}]}{2} \\ 
	&= O_{\P}\Bigg(   \frac{1}{\sqrt{n^\train}}  +     \sqrt{\sum_{y=1}^{m}  \E_{\allDII}\Big[\Var_{\gamma^\ast}[\widehat{s}_{\widehat{\gamma},y}-s_{\gamma^\ast,y}]\Big] }   \Bigg)  O_{\P}\bigg(\sqrt{\frac{1}{n^\test} + \frac{1}{n^\train}}\bigg) \\ 
	&= O_{\P}\Bigg( \frac{1}{\sqrt{n^\train}} + \sqrt{\sum_{y=1}^m \bigg\{ \frac{1}{\sqrt{n^\train}} + \lambda_y \Omega(s_{\gamma^\ast,y})^2 + \frac{1}{\sqrt{\lambda_y}}e^{-Cn^\train} + \E_{\DtrainII}\Big[ \U_y(\lambda_y)\Big] + \sqrt{\E_{\allDII}\Big[ \norm{\widehat{\gamma} - \gamma^\ast}{2}^2 \Big]}  \bigg\}    }  \Bigg)  O_{\P}\bigg(\sqrt{\frac{1}{n^\test} + \frac{1}{n^\train}}\bigg) \\ 
	&= O_{\P}\Bigg(\sqrt{\sum_{y=1}^m \bigg\{ \frac{1}{\sqrt{n^\train}} + \lambda_y \Omega(s_{\gamma^\ast,y})^2 + \frac{1}{\sqrt{\lambda_y}}e^{-Cn^\train} + \E_{\DtrainII}\Big[ \U_y(\lambda_y)\Big] + \sqrt{\E_{\allDII}\Big[ \norm{\widehat{\gamma} - \gamma^\ast}{2}^2 \Big]}  \bigg\}    }  \Bigg)  O_{\P}\bigg(\sqrt{\frac{1}{n^\test} + \frac{1}{n^\train}}\bigg),
\end{align*}
\normalsize
where the second line is due to Lemma \eqref{Learning the Inverse FIM}, and the third line uses the same logic employed when we bounded $\norm{W_3}{2}$.  \\ \\

\uline{Next, let's examine $W_5$}. Using the same logic used to bound  $\norm{W_3}{2}$ and  $\norm{W_4}{2}$, we have that:
\small
\begin{align*}
	&\norm{W_5}{2} \\
	&\leq \maxSing{\widehat{A}^{-1} - \FisherInfo{\gamma^\ast}^{-1}}\Ibig{\minSing{\widehat{A}} > 0}  \cdot \bignorm{\widehat{\E}_{\pi^\ast}^\testI[\widehat{s}_{\widehat{\gamma}}-s_{\gamma^\ast}] -  \widehat{\E}_{\pi^\ast}^\trainI[\widehat{s}_{\widehat{\gamma}}-s_{\gamma^\ast}]  }{2} \\ \\
	&= O_{\P}\Bigg(\sqrt{\sum_{y=1}^m \bigg\{ \frac{1}{\sqrt{n^\train}} + \lambda_y \Omega(s_{\gamma^\ast,y})^2 + \frac{1}{\sqrt{\lambda_y}}e^{-Cn^\train} + \E_{\DtrainII}\Big[ \U_y(\lambda_y)\Big] + \sqrt{\E_{\allDII}\Big[ \norm{\widehat{\gamma} - \gamma^\ast}{2}^2 \Big]}  \bigg\}    }  \Bigg) \\
	&\qquad \times O_{\P}\Bigg(\sqrt{\bigg( \frac{1}{n^\test} + \frac{1}{n^\train} \bigg)  \sum_{y=1}^{m}\bigg\{ \frac{1}{\sqrt{n^\train}} + \lambda_y \Omega(s_{\gamma^\ast,y})^2 + \frac{1}{\sqrt{\lambda_y}}e^{-Cn^\train} + \E_{\DtrainII}\Big[ \U_y(\lambda_y)\Big] + \sqrt{\E_{\allDII}\Big[ \norm{\widehat{\gamma} - \gamma^\ast}{2}^2 \Big]}  \bigg\}   }\Bigg) \\ \\ 
	&= O_{\P}\Bigg(\sqrt{\bigg( \frac{1}{n^\test} + \frac{1}{n^\train} \bigg) } \sum_{y=1}^{m}\bigg\{ \frac{1}{\sqrt{n^\train}} + \lambda_y \Omega(s_{\gamma^\ast,y})^2 + \frac{1}{\sqrt{\lambda_y}}e^{-Cn^\train} + \E_{\DtrainII}\Big[ \U_y(\lambda_y)\Big] + \sqrt{\E_{\allDII}\Big[ \norm{\widehat{\gamma} - \gamma^\ast}{2}^2 \Big]}  \bigg\}\Bigg).
\end{align*}
\normalsize
\vspace{0.4in}

\uline{Finally, let's examine $W_6$}. Observe that:
\small
\begin{align*}
	\norm{W_6}{2} &\leq  \biggnorm{\frac{1}{m+1}\vec{1}_m - \pi^\ast}{2} \Ibig{\minSing{\widehat{A}} = 0} \\ 
	&= O_{\P}\bigg( e^{-C (\min_y\lambda_y)^2 n^\train}  + \P_{\allDII}\Big[\bignorm{ \widehat{\gamma} - \gamma^\ast }{2} \geq C \Big]  \bigg),
\end{align*}
\normalsize
where the second line makes use of the work done in bounding $\norm{W_2}{2}$. \\ \\

Finally,  having obtained bounds on $\norm{W_2}{2},\dots,\norm{W_6}{2}$, we now make two observations. First, note that the upper bound for the rate of $\norm{W_2}{2}$ goes down to 0 faster than the upper bound for the rate of $\norm{W_6}{2}$. Second, note that, due to Assumption \eqref{lambda and uniform convergence properties},
\small
\begin{align*}
	\frac{e^{-C n^\train}}{\sqrt{\min_y\lambda_y}} &= \frac{1}{\sqrt{\min_y\lambda_y} \cdot e^{C n^\train} } \\ 
	&= o\bigg(  \frac{1}{ ( \min_y\lambda_y)^2 n^\train } \bigg) \\ 
	&= o(1).
\end{align*}
\normalsize
Thus, the upper bound for the rate of $\norm{W_5}{2}$ goes down to 0 faster than the upper bound for the rate of $\norm{W_4}{2}$.  Defining  $\epsilon^{(a)} := \sum_{l=2}^{6} W_l$, it follows that:
\small
\begin{align*}
	&\norm{\epsilon^{(a)}}{2}\\
	 &\leq \sum_{l = 2}^{6} \norm{W_l}{2} \\ \\ 
	&=   O_{\P}\Bigg(\sqrt{\bigg( \frac{1}{n^\test} + \frac{1}{n^\train} \bigg)  \sum_{y=1}^{m}\bigg\{ \frac{1}{\sqrt{n^\train}} + \lambda_y \Omega(s_{\gamma^\ast,y})^2 + \frac{1}{\sqrt{\lambda_y}}e^{-Cn^\train} + \E_{\DtrainII}\Big[ \U_y(\lambda_y)\Big] + \sqrt{\E_{\allDII}\Big[ \norm{\widehat{\gamma} - \gamma^\ast}{2}^2 \Big]}  \bigg\}   }\Bigg) \\ 
	&\qquad + O_{\P}\bigg( e^{-C (\min_y\lambda_y)^2 n^\train}  + \P_{\allDII}\Big[\bignorm{ \widehat{\gamma} - \gamma^\ast }{2} \geq C \Big]  \bigg) \\ \\ 
	&=   O_{\P}\Bigg(\sqrt{\bigg( \frac{1}{n^\test} + \frac{1}{n^\train} \bigg) \bigg\{ \frac{1}{\sqrt{n^\train}} + \max_y \lambda_y \Omega(s_{\gamma^\ast,y})^2 +  \frac{1}{\sqrt{\min_y\lambda_y}}e^{-Cn^\train} +  \max_y \E_{\DtrainII}\Big[ \U_y(\lambda_y)\Big] + \sqrt{\E_{\allDII}\Big[ \norm{\widehat{\gamma} - \gamma^\ast}{2}^2 \Big]}  \bigg\}   }\Bigg) \\ 
	&\qquad + O_{\P}\bigg( e^{-C (\min_y\lambda_y)^2 n^\train}  + \P_{\allDII}\Big[\bignorm{ \widehat{\gamma} - \gamma^\ast }{2} \geq C \Big]  \bigg).
\end{align*}
\normalsize

So, overall, we have that:
\small
\begin{align*}
	\widehat{\pi}^{(a)} - \pi^\ast = \sqrt{2}\Bigg( \Bigg[\frac{1}{n^\test}+ \frac{1}{n^\train} \sum_{y=0}^{m}\frac{(\pi_y^\ast)^2}{\pi_y^\train}\Bigg] \Big( \FisherInfo{\gamma^\ast}^{-1} - \FisherInfo{\gamma^\ast;\textup{Cat}}^{-1}  \Big)  + \frac{1}{n^\test} \FisherInfo{\pi^\ast;\textup{Cat}}^{-1} \Bigg)^{\frac{1}{2}} Z^{(a)}+ \epsilon^{(a)},
\end{align*}
\normalsize
where $Z^{(a)} \goesto{d} \N(0,I_m)$, and $\epsilon^{(a)} \in \mathbb{R}^m$ satisfies 
\small
\begin{align*}
	\norm{\epsilon^{(a)}}{2} &=   O_{\P}\Bigg(\sqrt{\bigg( \frac{1}{n^\test} + \frac{1}{n^\train} \bigg) \bigg\{ \frac{1}{\sqrt{n^\train}} + \max_y \lambda_y \Omega(s_{\gamma^\ast,y})^2 +  \frac{1}{\sqrt{\min_y\lambda_y}}e^{-Cn^\train} +  \max_y \E_{\DtrainII}\Big[ \U_y(\lambda_y)\Big] + \sqrt{\E_{\allDII}\Big[ \norm{\widehat{\gamma} - \gamma^\ast}{2}^2 \Big]}  \bigg\}   }\Bigg) \\ 
	&\qquad + O_{\P}\bigg( e^{-C (\min_y\lambda_y)^2 n^\train}  + \P_{\allDII}\Big[\bignorm{ \widehat{\gamma} - \gamma^\ast }{2} \geq C \Big]  \bigg),
\end{align*}
\normalsize
as claimed!

\end{proof}
\vspace{0.4in}

\begin{proof}[\uline{Proof of Corollary \eqref{Error Rate of Full Estimator}}]

Note that the datasets $\allDI$ and $\allDII$ have identical joint distributions, so the results of Theorem \eqref{Error Rate of First Step Estimator} also apply analogously to $\widehat{\pi}^{(b)}$; that is, there exists random vectors $Z^{(b)},\epsilon^{(b)} \in \mathbb{R}^m$ such that 
\small
\begin{align*}
	\widehat{\pi}^{(b)} - \pi^\ast &= \frac{1}{n/2} \sum_{i= n/2 + 1}^n \psi^\eff(Z_i) + \epsilon^{(b)} \\
	&= \sqrt{2}\Bigg( \Bigg[\frac{1}{n^\test}+ \frac{1}{n^\train} \sum_{y=0}^{m}\frac{(\pi_y^\ast)^2}{\pi_y^\train}\Bigg] \Big( \FisherInfo{\gamma^\ast}^{-1} - \FisherInfo{\gamma^\ast;\textup{Cat}}^{-1}  \Big)  + \frac{1}{n^\test} \FisherInfo{\pi^\ast;\textup{Cat}}^{-1} \Bigg)^{\frac{1}{2}} Z^{(b)}+ \epsilon^{(b)},
\end{align*}
\normalsize
where $Z^{(b)} \goesto{d} \N(0,I_m)$, and $\epsilon^{(b)}$ satisfies 
\small
\begin{align*}
	&\norm{\epsilon^{(b)}}{2} \\
	&=   O_{\P}\Bigg(\sqrt{\bigg( \frac{1}{n^\test} + \frac{1}{n^\train} \bigg) \bigg\{ \frac{1}{\sqrt{n^\train}} + \max_y \lambda_y \Omega(s_{\gamma^\ast,y})^2 +  \frac{1}{\sqrt{\min_y\lambda_y}}e^{-Cn^\train} +  \max_y \E_{\DtrainII}\Big[ \U_y(\lambda_y)\Big] + \sqrt{\E_{\allDII}\Big[ \norm{\widehat{\gamma} - \gamma^\ast}{2}^2 \Big]}  \bigg\}   }\Bigg) \\ 
	&\qquad + O_{\P}\bigg( e^{-C (\min_y\lambda_y)^2 n^\train}  + \P_{\allDII}\Big[\bignorm{ \widehat{\gamma} - \gamma^\ast }{2} \geq C \Big]  \bigg).
\end{align*}
\normalsize

Thus:
\begin{align*}
	&\widehat{\pi} - \pi^\ast \\
	&= \frac{\widehat{\pi}^{(a)} + \widehat{\pi}^{(b)} }{2} - \pi^\ast \\ 
	 &= \frac{(\widehat{\pi}^{(a)}-\pi^\ast) }{2}+ \frac{(\widehat{\pi}^{(b)}-\pi^\ast) }{2}  \\ 
	 &= \Bigg( \Bigg[\frac{1}{n^\test}+ \frac{1}{n^\train} \sum_{y=0}^{m}\frac{(\pi_y^\ast)^2}{\pi_y^\train}\Bigg] \Big( \FisherInfo{\gamma^\ast}^{-1} - \FisherInfo{\gamma^\ast;\textup{Cat}}^{-1}  \Big)  + \frac{1}{n^\test} \FisherInfo{\pi^\ast;\textup{Cat}}^{-1} \Bigg)^{\frac{1}{2}}  \frac{Z^{(a)} + Z^{(b)} }{\sqrt{2}} + \frac{1}{2}\big(\epsilon^{(a)} + \epsilon^{(b)} \big).
\end{align*}

Now, recall from the proof of Theorem \eqref{Error Rate of First Step Estimator} that $Z^{(a)}$ is a function of only $\allDI$ (and not $\allDII$), and analogously, $Z^{(b)}$ is a function of only $\allDII$ (and not $\allDI$). Since $\allDI$ and $\allDII$ are independent, this means that $Z^{(a)}$ and $Z^{(b)}$ are also independent. Thus, it follows from the Continuity Theorem that $Z^{(a)} + Z^{(b)} \goesto{d} \N(0,2I_m)$, meaning that $Z:= \frac{Z^{(a)} + Z^{(b)} }{\sqrt{2}} \goesto{d} \N(0,I_m)$ by way of Slutsky's Theorem. Defining $\epsilon := \frac{1}{2} (\epsilon^{(a)} + \epsilon^{(b)})$, we therefore have that:
\begin{align*}
	\widehat{\pi} - \pi^\ast &=  \Bigg( \Bigg[\frac{1}{n^\test}+ \frac{1}{n^\train} \sum_{y=0}^{m}\frac{(\pi_y^\ast)^2}{\pi_y^\train}\Bigg] \Big( \FisherInfo{\gamma^\ast}^{-1} - \FisherInfo{\gamma^\ast;\textup{Cat}}^{-1}  \Big)  + \frac{1}{n^\test} \FisherInfo{\pi^\ast;\textup{Cat}}^{-1} \Bigg)^{\frac{1}{2}} Z + \epsilon,
\end{align*}
where $Z \goesto{d} \N(0,I_m)$, and $\epsilon$ satisfies 
\small
\begin{align*}
	\norm{\epsilon}{2} &=   O_{\P}\Bigg(\sqrt{\bigg( \frac{1}{n^\test} + \frac{1}{n^\train} \bigg) \bigg\{ \frac{1}{\sqrt{n^\train}} + \max_y \lambda_y \Omega(s_{\gamma^\ast,y})^2 +  \frac{1}{\sqrt{\min_y\lambda_y}}e^{-Cn^\train} +  \max_y \E_{\DtrainII}\Big[ \U_y(\lambda_y)\Big] + \sqrt{\E_{\allDII}\Big[ \norm{\widehat{\gamma} - \gamma^\ast}{2}^2 \Big]}  \bigg\}   }\Bigg) \\ 
	&\qquad + O_{\P}\bigg( e^{-C (\min_y\lambda_y)^2 n^\train}  + \P_{\allDII}\Big[\bignorm{ \widehat{\gamma} - \gamma^\ast }{2} \geq C \Big]  \bigg),
\end{align*}
\normalsize
as claimed. Finally, observe that:
\begin{align*}
	\frac{1}{n} \sum_{i= 1}^{n} \psi^\eff(Z_i) &= \frac{1}{2}\bigg(   \frac{1}{n/2} \sum_{i= 1}^{n/2} \psi^\eff(Z_i) +  \frac{1}{n/2} \sum_{i= n/2 + 1}^n \psi^\eff(Z_i)  \bigg) \\ 
	&=  \Bigg( \Bigg[\frac{1}{n^\test}+ \frac{1}{n^\train} \sum_{y=0}^{m}\frac{(\pi_y^\ast)^2}{\pi_y^\train}\Bigg] \Big( \FisherInfo{\gamma^\ast}^{-1} - \FisherInfo{\gamma^\ast;\textup{Cat}}^{-1}  \Big)  + \frac{1}{n^\test} \FisherInfo{\pi^\ast;\textup{Cat}}^{-1} \Bigg)^{\frac{1}{2}}  \frac{Z^{(a)} + Z^{(b)} }{\sqrt{2}} \\
	&= \Bigg( \Bigg[\frac{1}{n^\test}+ \frac{1}{n^\train} \sum_{y=0}^{m}\frac{(\pi_y^\ast)^2}{\pi_y^\train}\Bigg] \Big( \FisherInfo{\gamma^\ast}^{-1} - \FisherInfo{\gamma^\ast;\textup{Cat}}^{-1}  \Big)  + \frac{1}{n^\test} \FisherInfo{\pi^\ast;\textup{Cat}}^{-1} \Bigg)^{\frac{1}{2}} Z,
\end{align*}
so the desiderata have all been proven.
\end{proof}
\vspace{0.4in}

\begin{proof}[\uline{Proof of Theorem \eqref{Standardized Error Rate of Full Estimator}}]

Define
\[
 	V:=  \Bigg[\frac{1}{n^\test}+ \frac{1}{n^\train} \sum_{y=0}^{m}\frac{(\pi_y^\ast)^2}{\pi_y^\train}\Bigg] \Big( \FisherInfo{\gamma^\ast}^{-1} - \FisherInfo{\gamma^\ast;\textup{Cat}}^{-1}  \Big)  + \frac{1}{n^\test} \FisherInfo{\pi^\ast;\textup{Cat}}^{-1}.
\]
By Corollary \eqref{First Order Constant When Matching Function is Score Function}, $V$ is a symmetric positive definite matrix. So, using the results of Corollary \eqref{Error Rate of Full Estimator}, it follows that there exists random vectors $Z,\delta \in \mathbb{R}^m$ such that 
\[
	V^{-\frac{1}{2}} (\widehat{\pi} - \pi^\ast) = Z + \epsilon,
\]
where 
\[
 	Z = V^{-\frac{1}{2}}\bigg(\frac{1}{n} \sum_{i= 1}^{n} \psi^\eff(Z_i)\bigg) \goesto{d} \N(0,I_m)
\]
and $\epsilon = V^{-\frac{1}{2}} \delta$, and $\delta \in \mathbb{R}^m$ is a random vector which satisfies
\small
\begin{align*}
	&\norm{\delta}{2} \\
	&=   O_{\P}\Bigg(\sqrt{\bigg( \frac{1}{n^\test} + \frac{1}{n^\train} \bigg) \bigg\{ \frac{1}{\sqrt{n^\train}} + \max_y \lambda_y \Omega(s_{\gamma^\ast,y})^2 +  \frac{1}{\sqrt{\min_y\lambda_y}}e^{-Cn^\train} +  \max_y \E_{\DtrainII}\Big[ \U_y(\lambda_y)\Big] + \sqrt{\E_{\allDII}\Big[ \norm{\widehat{\gamma} - \gamma^\ast}{2}^2 \Big]}  \bigg\}   }\Bigg) \\ 
	&\qquad + O_{\P}\bigg( e^{-C (\min_y\lambda_y)^2 n^\train}  + \P_{\allDII}\Big[\bignorm{ \widehat{\gamma} - \gamma^\ast }{2} \geq C \Big]  \bigg).
\end{align*}
\normalsize

Note that 
\begin{align*}
	\minSing{V} &= \minEval{V} \\ 
	&\geq \bigg(\frac{1}{n^\test} + \frac{1}{n^\train} \bigg)\xi^2 \minEval{   \FisherInfo{\gamma^\ast}^{-1} - \FisherInfo{\gamma^\ast;\textup{Cat}}^{-1}   }\\
	&\geq \bigg(\frac{1}{n^\test} + \frac{1}{n^\train} \bigg)\xi^2 \nu \bigg( \frac{L}{m+1}\bigg)^2, 
\end{align*}
where the first line is because $V$ is symmetric positive definite, the second line is due to Assumption \eqref{bounded pis} and Lemma \eqref{Bounds on Eigenvalues of Categorical Fisher Info}, and the third line uses the same logic employed in the proof of Lemma \eqref{First Order Term} to obtain a lower bound on $\minEval{   \FisherInfo{\gamma^\ast}^{-1} - \FisherInfo{\gamma^\ast;\textup{Cat}}^{-1}   }$. Thus, it follows that:
\begin{align*}
	\norm{\epsilon}{2} &= \norm{ V^{-\frac{1}{2}} \delta}{2} \\ 
	&\leq \maxSing{V^{-\frac{1}{2}}} \norm{\delta}{2} \\ 
	&=  \frac{\norm{\delta}{2}}{\sqrt{\minSing{V}}} \\ 
	&\leq \frac{1}{\sqrt{\frac{1}{n^\test} + \frac{1}{n^\train} }}  \frac{\norm{\delta}{2}}{\sqrt{\xi^2 \nu \bigg( \frac{L}{m+1}\bigg)^2 }}.
\end{align*}
Therefore, we have that:
\small
\begin{align*}
	\norm{\epsilon}{2} &= O_{\P}\Bigg(\sqrt{ \frac{1}{\sqrt{n^\train}} + \max_y \lambda_y \Omega(s_{\gamma^\ast,y})^2 +  \frac{1}{\sqrt{\min_y\lambda_y}}e^{-Cn^\train} +  \max_y \E_{\DtrainII}\Big[ \U_y(\lambda_y)\Big] + \sqrt{\E_{\allDII}\Big[ \norm{\widehat{\gamma} - \gamma^\ast}{2}^2 \Big]}  }\Bigg) \\ 
	&\qquad + O_{\P}\Bigg( \frac{1}{\sqrt{\frac{1}{n^\test} + \frac{1}{n^\train}}}\bigg\{e^{-C (\min_y\lambda_y)^2 n^\train}  + \P_{\allDII}\Big[\bignorm{ \widehat{\gamma} - \gamma^\ast }{2} \geq C \Big] \bigg\} \Bigg),
\end{align*}
\normalsize
as claimed!

\end{proof}
\vspace{0.4in}

\begin{proof}[\uline{Proof of Corollary \eqref{RKHS Application}}]
First, we show that Assumption \eqref{members of function class} is satisfied. Since $s_{\gamma^\ast} \in H_{\K}^m$ and $\H = H_{\K}^m$, we need only verify that $\H$ is closed under scalar multiplication. This is true because if $h\in\H$ and $c\in\mathbb{R}$, then since $\H =  H_{\K}^m = \big[ \cup_{k\in\K} H_k \big]^m$, it must be that $h_y\in \cup_{k\in\K} H_k$ for each $y\in[m]$. However, since $H_k$ is a RKHS, this means that $ch_y \in  \cup_{k\in\K} H_k$ for each $y\in[m]$ as well. Thus, it follows that $ch \in \big[ \cup_{k\in\K} H_k \big]^m = \H$. \\ 

Second, we show that Assumption \eqref{Omega Properties} is satisfied. Let any $y\in[m]$ be given. Clearly, $\Omega: \H_y \mapsto [0,\infty)$. In addition, for any $k\in\K$, $\norm{\cdot}{k}$ is a norm so it is equal to $0$ at the identically zero function, meaning that $\Omega$ must also be equal to $0$ at the identically zero function. Also, for any $c \in \mathbb{R}$ and $h_y \in \H_y$:
\begin{align*}
	\Omega(ch_y) &= \inf\Big\{    \norm{ch_y}{k}    \ \Big\lvert \  k\in\K, \ ch_y \in H_k      \Big\} \\
	&= \abs{c} \inf\Big\{    \norm{h_y}{k}    \ \Big\lvert \  k\in\K, \ h_y \in H_k      \Big\} \\
	&= \abs{c} \Omega(h_y).
\end{align*}
Lastly, for any $k\in\K$, if $h_y\in H_k$, then $\norm{h_y}{k} \geq D \norm{h_y}{\infty}$ with $D = \frac{1}{\kappa}$ (see the beginning of the proof of  Lemma 4.23 in \citep{KernelReproducingProperty}). This implies that:
\begin{align*}
	\Omega(h_y) &= \inf\Big\{    \norm{h_y}{k}    \ \Big\lvert \  k\in\K, \ h_y \in H_k      \Big\} \\
	&\geq \inf\Big\{     D \norm{h_y}{\infty}   \ \Big\lvert \  k\in\K, \ h_y \in H_k      \Big\} \\
	&=   D \norm{h_y}{\infty}. 
\end{align*}
Thus, Assumption \eqref{Omega Properties} is satisfied with $D = \frac{1}{\kappa}$. \\ 

Third, it remains for us to obtain the bound on $\E_{\DtrainII}[\U_y(\lambda_y)]$. Towards that end, note that for each $h_y \in \H_y$ which satisfies $\Omega(h_y) \leq  \frac{1}{L\sqrt{\lambda_y}}$, we have that:
\begin{align*}
	\absBig{ \widehat{\Var}_{\gamma^\ast}^{\trainII}[h_y]  -  \Var_{\gamma^\ast}[h_y]} &\leq \absBig{ \widehat{\E}_{\gamma^\ast}^{\trainII}[h_y^2]  -  \E_{\gamma^\ast}[h_y^2]} + \absBig{ \big(\widehat{\E}_{\gamma^\ast}^{\trainII}[h_y]\big)^2  -  \big(\E_{\gamma^\ast}[h_y]\big)^2 } \\ 
	&=   \absBig{ \widehat{\E}_{\gamma^\ast}^{\trainII}[h_y^2]  -  \E_{\gamma^\ast}[h_y^2]} + \absBig{ \widehat{\E}_{\gamma^\ast}^{\trainII}[h_y]  +  \E_{\gamma^\ast}[h_y] }  \absBig{ \widehat{\E}_{\gamma^\ast}^{\trainII}[h_y]  -  \E_{\gamma^\ast}[h_y] } \\ 
	&\leq \absBig{ \widehat{\E}_{\gamma^\ast}^{\trainII}[h_y^2]  -  \E_{\gamma^\ast}[h_y^2]} +   2\norm{h_y}{\infty}\absBig{ \widehat{\E}_{\gamma^\ast}^{\trainII}[h_y]  -  \E_{\gamma^\ast}[h_y] } \\  
	&\leq \absBig{ \widehat{\E}_{\gamma^\ast}^{\trainII}[h_y^2]  -  \E_{\gamma^\ast}[h_y^2]} +   \frac{2\kappa}{L\sqrt{\lambda_y}} \absBig{ \widehat{\E}_{\gamma^\ast}^{\trainII}[h_y]  -  \E_{\gamma^\ast}[h_y] }\\
	&\leq \sum_{j = 0}^{m}  \absBig{ \widehat{\E}_{j}^{\trainII}[h_y^2]  -  \E_{j}[h_y^2]}  + \frac{2\kappa}{L\sqrt{\lambda_y}} \sum_{j = 0}^{m}  \absBig{ \widehat{\E}_{j}^{\trainII}[h_y]  -  \E_{j}[h_y]}, 
\end{align*} 
where the fourth line uses the fact that $D\norm{h_y}{\infty} \leq \Omega(h)$ where $D = \frac{1}{\kappa}$. Ergo:
\begin{align*}
	&\E_{\DtrainII}[\U_y(\lambda_y)]\\
	 &= \E_{\DtrainII}\Bigg( \sup\limits_{\Omega(h_y) \leq \frac{1}{L\sqrt{\lambda_y}}}  \absBig{\widehat{\E}_0^{\trainII}[h_y] - \E_0[h_y]} \Bigg) +  \E_{\DtrainII}\Bigg( \sup\limits_{\Omega(h_y) \leq \frac{1}{L\sqrt{\lambda_y}}} \absBig{\widehat{\E}_y^{\trainII}[h_y] - \E_y[h_y]} \Bigg) \\
	&\qquad +  \E_{\DtrainII}\Bigg( \sup\limits_{ \Omega(h_y) \leq \frac{1}{L\sqrt{\lambda_y}}} \absBig{ \widehat{\Var}_{\gamma^\ast}^{\trainII}[h_y] - \Var_{\gamma^\ast}[h_y]}\Bigg) \\ \\
	&\leq  \E_{\DtrainII}\Bigg(  \sup\limits_{\Omega(h_y) \leq \frac{1}{L\sqrt{\lambda_y}}}  \absBig{\widehat{\E}_0^{\trainII}[h_y] - \E_0[h_y]} \Bigg) +   \E_{\DtrainII}\Bigg( \sup\limits_{\Omega(h_y) \leq \frac{1}{L\sqrt{\lambda_y}}} \absBig{\widehat{\E}_y^{\trainII}[h_y] - \E_y[h_y]}\Bigg) \\
	&\qquad+ \sum_{j = 0}^{m}  \E_{\DtrainII}\Bigg( \sup\limits_{\Omega(h_y) \leq \frac{1}{L\sqrt{\lambda_y}}} \absBig{ \widehat{\E}_{j}^{\trainII}[h_y^2]  -  \E_{j}[h_y^2]}\Bigg) +  \frac{2\kappa}{L\sqrt{\lambda_y}} \sum_{j = 0}^{m}  \E_{\DtrainII}\Bigg( \sup\limits_{\Omega(h_y) \leq \frac{1}{L\sqrt{\lambda_y}}} \absBig{ \widehat{\E}_{j}^{\trainII}[h_y]  -  \E_{j}[h_y]}\Bigg) \\ \\ 
	&= O\Bigg(\sum_{j = 0}^{m}  \E_{\DtrainII}\Bigg[ \sup\limits_{\Omega(h_y) \leq \frac{1}{L\sqrt{\lambda_y}}} \absBig{ \widehat{\E}_{j}^{\trainII}[h_y^2]  -  \E_{j}[h_y^2]}\Bigg] +  \frac{1}{\sqrt{\lambda_y}} \sum_{j = 0}^{m}  \E_{\DtrainII}\Bigg[ \sup\limits_{\Omega(h_y) \leq \frac{1}{L\sqrt{\lambda_y}}} \absBig{ \widehat{\E}_{j}^{\trainII}[h_y]  -  \E_{j}[h_y]}\Bigg] \Bigg).
\end{align*}

Next, define the function   $\Phi(t) := \min\Big\{t^2,\frac{\kappa^2}{L^2 \lambda_y }  \Big\}$, and let $\Big\{\xi_{j,q}: 0\leq j \leq m,  \ 1\leq q \leq \tfrac{1}{2}\pi_j^\train n^\train \Big\}$ be a set of independent Rademacher random variables. For each class $j \in 0\cup [m]$, define the Rademacher Complexities
\[
	\R_j(\lambda_y, \H_y) :=  \E_{\xi,\DtrainII} \Bigg[ \sup\limits_{\Omega(h_y) \leq \frac{1}{L\sqrt{\lambda_y}}} \frac{1}{\tfrac{1}{2}\pi_j^\train n^\train} \sum_{q=1}^{\tfrac{1}{2}\pi_j^\train n^\train} \xi_{j,q} h_y(X_{j,q}^\train ) \Bigg]
\]
and
\begin{align*}
	\R_j(\lambda_y, \Phi(\H_y)) &:= \E_{\xi,\DtrainII} \Bigg[ \sup\limits_{\Omega(h_y) \leq \frac{1}{L\sqrt{\lambda_y}}} \frac{1}{\tfrac{1}{2}\pi_j^\train n^\train} \sum_{q=1}^{\tfrac{1}{2}\pi_j^\train n^\train} \xi_{j,q} \Phi(h_y(X_{j,q}^\train)) \Bigg] \\
	&= \E_{\xi,\DtrainII} \Bigg[ \sup\limits_{\Omega(h_y) \leq \frac{1}{L\sqrt{\lambda_y}}} \frac{1}{\tfrac{1}{2}\pi_j^\train n^\train} \sum_{q=1}^{\tfrac{1}{2}\pi_j^\train n^\train} \xi_{j,q} h_y^2(X_{j,q}^\train)\Bigg],
\end{align*}
where the second line above uses the fact that $\Omega(h_y) \leq \frac{1}{L\sqrt{\lambda_y}} \implies \norm{h_y}{\infty} \leq \frac{\kappa}{L\sqrt{\lambda_y}}$, so $\norm{h_y^2}{\infty} \leq \frac{\kappa^2}{L^2 \lambda_y}$ meaning that $\Phi(h_y(x)) =  \min\Big\{h_y^2(x),\frac{\kappa^2}{L^2 \lambda_y }  \Big\} = h_y^2(x)$. Using standard symmetrization arguments (e.g., see Lemma 2.3.1 in \citep{UniformDeviationVsRademacherComplexity}), it follows that:
\[
	 \E_{\DtrainII}\Bigg[ \sup\limits_{\Omega(h_y) \leq \frac{1}{L\sqrt{\lambda_y}}} \absBig{ \widehat{\E}_{j}^{\trainII}[h_y]  -  \E_{j}[h_y]}\Bigg] \leq 2 \R_j(\lambda_y, \H_y)
\]
and
\[
	\E_{\DtrainII}\Bigg[ \sup\limits_{\Omega(h_y) \leq \frac{1}{L\sqrt{\lambda_y}}} \absBig{ \widehat{\E}_{j}^{\trainII}[h_y^2]  -  \E_{j}[h_y^2]}\Bigg]  \leq  2 \R_j(\lambda_y, \Phi(\H_y)).
\]
Now, the function $\Phi(t)$ is $\frac{2\kappa}{L\sqrt{\lambda_y}}$-Lipschitz, because for any two $t_1,t_2\in\mathbb{R}$:
\begin{align*}
	\abs{\Phi(t_1) - \Phi(t_2)} &= \absbigg{  \min\Big\{\abs{t_1},\frac{\kappa}{L \sqrt{\lambda_y} }  \Big\}^2 - \min\Big\{\abs{t_2},\frac{\kappa}{L \sqrt{\lambda_y} } \Big\}^2 } \\ 
	&=   \absbigg{  \min\Big\{\abs{t_1},\frac{\kappa}{L \sqrt{\lambda_y} }  \Big\} + \min\Big\{\abs{t_2},\frac{\kappa}{L \sqrt{\lambda_y} } \Big\}  }  \absbigg{  \min\Big\{\abs{t_1},\frac{\kappa}{L^2 \sqrt{\lambda_y} }  \Big\} - \min\Big\{\abs{t_2},\frac{\kappa}{L \sqrt{\lambda_y} } \Big\}  } \\ 
	&\leq  \frac{2\kappa}{L \sqrt{\lambda_y} }  \absbigg{  \min\Big\{\abs{t_1},\frac{\kappa}{L^2 \sqrt{\lambda_y} }  \Big\} - \min\Big\{\abs{t_2},\frac{\kappa}{L \sqrt{\lambda_y} } \Big\}  } \\  
	&\leq  \frac{2\kappa}{L \sqrt{\lambda_y} } \absbig{ \abs{t_1} - \abs{t_2}} \\ 
	&\leq  \frac{2\kappa}{L \sqrt{\lambda_y} } \absbig{ t_1 - t_2},
\end{align*}
where the fourth line is a consequence of considering 4 separate cases: (i) $\abs{t_1},\abs{t_2} \leq \frac{\kappa}{L \sqrt{\lambda_y} }$, (ii)  $ \frac{\kappa}{L \sqrt{\lambda_y} } \leq \abs{t_1},\abs{t_2} $,  (iii) $  \abs{t_1} \leq \frac{\kappa}{L \sqrt{\lambda_y} } \leq \abs{t_2} $ and (iv) $  \abs{t_2} \leq \frac{\kappa}{L \sqrt{\lambda_y} } \leq \abs{t_1}$. In case (i), the fourth line's inequality is an equality. In case (ii), the third line equals $0$ and so the fourth line is trivially true. Finally, cases (iii) and (iv) are symmetric, so we need only consider what happens in case (iii). In case (iii), the distance between $\abs{t_1}$ and $\frac{\kappa}{L \sqrt{\lambda_y} } $ is less than or equal to the distance between $\abs{t_1}$ and $\abs{t_2}$, so 
\[
	 \absbigg{  \min\Big\{\abs{t_1},\frac{\kappa}{L^2 \sqrt{\lambda_y} }  \Big\} - \min\Big\{\abs{t_2},\frac{\kappa}{L \sqrt{\lambda_y} } \Big\}  } =  \absBig{ \abs{t_1} - \frac{\kappa}{L \sqrt{\lambda_y} }  } \leq   \absBig{ \abs{t_1} - \abs{t_2}}.
\]
This establishes that the function $\Phi(t)$ is indeed $\frac{2\kappa}{L\sqrt{\lambda_y}}$-Lipschitz. So, invoking Lemma 4.12 in \cite{RademacherContraction}, it follows that $R_j(\lambda_y, \Phi(\H_y)) \leq \frac{2\kappa}{L\sqrt{\lambda_y}}R_j(\lambda_y, \H_y)$. Therefore, we may conclude that:
\begin{align*}
	\E_{\DtrainII}[\U_y(\lambda_y)]  &=   O\Bigg(\sum_{j = 0}^{m} \R_j(\lambda_y, \Phi(\H_y)) +  \frac{1}{\sqrt{\lambda_y}} \sum_{j = 0}^{m} \R_j(\lambda_y, \H_y) \Bigg) \\ 
	&=   O\Bigg(   \frac{1}{\sqrt{\lambda_y}} \sum_{j = 0}^{m} \R_j(\lambda_y, \H_y) \Bigg).
\end{align*}

Next, we'll develop an upper bound on $\R_j(\lambda_y,\H_y)$. Towards that end, consider any $h_y \in \H_y$ such that $\Omega(h_y) \leq \frac{1}{L\sqrt{\lambda_y}}$. Trivially, this means that $\Omega(h_y) < \frac{2}{L\sqrt{\lambda_y}}$, which by definition of $\Omega$, implies the existence of a kernel $k\in \K$ such that $h_y \in H_k$ and $\norm{h_y}{k} < \frac{2}{L \sqrt{\lambda_y}}$. This means that
\begin{align*}
	&\frac{1}{\tfrac{1}{2}\pi_j^\train n^\train} \sum_{q=1}^{\tfrac{1}{2}\pi_j^\train n^\train} \xi_{j,q} h_y(X_{j,q}^\train ) \\
	&=  \frac{1}{\tfrac{1}{2}\pi_j^\train n^\train} \Biggdotprod{h_y}{\sum_{q=1}^{\tfrac{1}{2}\pi_j^\train n^\train} \xi_{j,q}k(\cdot,X_{j,q}^\train)  }_{k} \\
	&\leq  \frac{1}{\tfrac{1}{2}\pi_j^\train n^\train} \bignorm{h_y}{k} \Biggnorm{\sum_{q=1}^{\tfrac{1}{2}\pi_j^\train n^\train} \xi_{j,q}k(\cdot,X_{j,q}^\train)  }{k} \\
	&<  \frac{1}{\tfrac{1}{2}\pi_j^\train n^\train}    \frac{2}{L \sqrt{\lambda_y}}   \sqrt{  \Biggnorm{\sum_{q=1}^{\tfrac{1}{2}\pi_j^\train n^\train} \xi_{j,q}k(\cdot,X_{j,q}^\train)  }{k}^2 } \\
	&=  \frac{1}{\tfrac{1}{2}\pi_j^\train n^\train}    \frac{2}{L \sqrt{\lambda_y}}   \sqrt{   \sum_{q,r}^{\tfrac{1}{2}\pi_j^\train n^\train} \xi_{j,q}\xi_{j,r}k(X_{j,q}^\train,X_{j,r}^\train)    } \\
	&=  \frac{1}{\tfrac{1}{2}\pi_j^\train n^\train}    \frac{2}{L \sqrt{\lambda_y}}   \sqrt{   \sum_{q=1}^{\tfrac{1}{2}\pi_j^\train n^\train}  k(X_{j,q}^\train,X_{j,q}^\train)      +          2\sum_{q < r}^{\tfrac{1}{2}\pi_j^\train n^\train} \xi_{j,q}\xi_{j,r}k(X_{j,q}^\train,X_{j,r}^\train)    } \\
	&\leq  \frac{1}{\tfrac{1}{2}\pi_j^\train n^\train}    \frac{2}{L \sqrt{\lambda_y}}   \sqrt{   \tfrac{1}{2}\pi_j^\train n^\train \kappa^2       +          2\sum_{q < r}^{\tfrac{1}{2}\pi_j^\train n^\train} \xi_{j,q}\xi_{j,r}k(X_{j,q}^\train,X_{j,r}^\train)    },
\end{align*}
where the first line uses the reproducing property (e.g., see Definition 4.18 in \cite{KernelReproducingProperty}), and the fourth line uses Theorem 4.21 in \cite{KernelReproducingProperty}. Thus, it follows that:
\begin{align*}
	&\R_j(\lambda_y,\H_y) \\
	&=  \E_{\xi,\DtrainII} \Bigg[ \sup\limits_{\Omega(h_y) \leq \frac{1}{L\sqrt{\lambda_y}}} \frac{1}{\tfrac{1}{2}\pi_j^\train n^\train} \sum_{k=1}^{\tfrac{1}{2}\pi_j^\train n^\train} \xi_{j,q} h_y(X_{j,q}^\train ) \Bigg] \\ 
	&\leq  \frac{1}{\tfrac{1}{2}\pi_j^\train n^\train}    \frac{2}{L \sqrt{\lambda_y}} \E_{\xi,\DtrainII} \Bigg[ \sup\limits_{k\in\K}     \sqrt{   \tfrac{1}{2}\pi_j^\train n^\train \kappa^2       +          2\sum_{q < r}^{\tfrac{1}{2}\pi_j^\train n^\train} \xi_{j,q}\xi_{j,r}k(X_{j,q}^\train,X_{j,r}^\train)    } \Bigg] \\ 
	&\leq  \frac{1}{\tfrac{1}{2}\pi_j^\train n^\train}    \frac{2}{L \sqrt{\lambda_y}}  \sqrt{ \E_{\xi,\DtrainII} \Bigg[   \tfrac{1}{2}\pi_j^\train n^\train \kappa^2       +          2 \sup\limits_{k\in\K}   \sum_{q < r}^{\tfrac{1}{2}\pi_j^\train n^\train} \xi_{j,q}\xi_{j,r}k(X_{j,q}^\train,X_{j,r}^\train)     \Bigg]} \\ 
	&=  \frac{1}{\tfrac{1}{2}\pi_j^\train n^\train}    \frac{2}{L \sqrt{\lambda_y}}  \sqrt{   \tfrac{1}{2}\pi_j^\train n^\train \kappa^2       +        \pi_j^\train n^\train    \E_{\xi,\DtrainII} \Bigg[  \frac{1}{  \tfrac{1}{2}\pi_j^\train n^\train} \sup\limits_{k\in\K}   \sum_{q < r}^{\tfrac{1}{2}\pi_j^\train n^\train} \xi_{j,q}\xi_{j,r}k(X_{j,q}^\train,X_{j,r}^\train)     \Bigg]}.
\end{align*}

Now, the expectation in the last line in the display above is the population-level Rademacher Chaos complexity of order two (e.g., see section 4.4 of \citep{RademacherContraction} and Definition 2 of   \citep{RadChaosComplexityBound}), over the set of kernels $\K$. By Theorem 3 of  \citep{RadChaosComplexityBound}, this can be upper bounded by the psuedo-dimension $d_{\K}$ of the set of kernels $\K$:
\[
	 \E_{\xi,\DtrainII} \Bigg[ \frac{1}{  \tfrac{1}{2}\pi_j^\train n^\train}  \sup\limits_{k\in\K}   \sum_{q < r}^{\tfrac{1}{2}\pi_j^\train n^\train} \xi_{j,q}\xi_{j,r}k(X_{j,q}^\train,X_{j,r}^\train)     \Bigg] \leq (96e + 1)\kappa^2d_{\K}.
\]
This implies that:
\begin{align*}
	\implies \R_j(\lambda_y,\H_y) &\leq  \frac{1}{\tfrac{1}{2}\pi_j^\train n^\train}    \frac{2}{L \sqrt{\lambda_y}}  \sqrt{   \tfrac{1}{2}\pi_j^\train n^\train \kappa^2       +          \pi_j^\train n^\train   (96e + 1)\kappa^2d_{\K}    }\\
	&=  \frac{1}{\tfrac{1}{2}\pi_j^\train n^\train}    \frac{2}{L \sqrt{\lambda_y}} \sqrt{   \big(\tfrac{\kappa^2}{2} + (96e + 1)\kappa^2 d_{\K}\big) \pi_j^\train n^\train  }  \\
	&=    \frac{4}{L \sqrt{\pi_j^\train n^\train\lambda_y}} \sqrt{   \tfrac{\kappa^2}{2} + (96e + 1)\kappa^2 d_{\K}  }  \\
	&= O\Bigg(\sqrt{\frac{1+d_{\K}}{\lambda_y n^\train}}\Bigg).
\end{align*}

Combining this result with our work earlier in the proof, it follows that:
\begin{align*}
	\implies \E_{\DtrainII}[\U_y(\lambda_y)]  &= O\Bigg( \frac{1}{\sqrt{\lambda_y}} \sum_{j = 0}^{m} \R_j(\lambda_y, \H_y) \Bigg) \\ 
	&= O\Bigg( \frac{1}{\sqrt{\lambda_y}}\sqrt{\frac{1+d_{\K}}{\lambda_y n^\train}} \Bigg) \\ 
	&= O\Bigg( \frac{1}{\lambda_y}\sqrt{\frac{1+d_{\K}}{n^\train}} \Bigg),
\end{align*}
as claimed!

\end{proof}
\vspace{0.4in}

\begin{proof}[\uline{Proof of Corollary \eqref{Standardized Error Rate of Full Estimator, Application to RKHS}}]
We'll begin by demonstrating that Assumption  \eqref{lambda and uniform convergence properties} holds in the setting specified by this Corollary. By Corollary \eqref{RKHS Application}, we know that, for each $y\in[m]$:
\begin{align*}
	\E_{\DtrainII}\big[ \U_y(\lambda_y)\big] &= O\Bigg( \frac{1}{\lambda_y}\sqrt{\frac{1+d_{\K}}{n^\train}} \Bigg) \\ 
	&=  O\Bigg( \bigg( \frac{n^\train}{1+d_{\K}} \bigg)^{\frac{1}{4}}  \sqrt{\frac{1+d_{\K}}{n^\train}}  \max_{j} \Omega(s_{\gamma^\ast,j}) \Bigg) \\ 
	&=  O\Bigg( \bigg( \frac{1+d_{\K}}{n^\train} \bigg)^{\frac{1}{4}}  \max_{j} \Omega(s_{\gamma^\ast,j})\Bigg),
\end{align*}
where the second line is because $\lambda_y \asymp \frac{1}{\max_{j} \Omega(s_{\gamma^\ast,j})} \big( \frac{1+d_{\K}}{n^\train} \big)^{\frac{1}{4}}$ for each $j\in[m]$. Now, it follows from the assumption of
$\max_{j} \Omega(s_{\gamma^\ast,j}) = o\bigg(\frac{  (n^\train)^{\frac{1}{4}} }{\max\big\{ \sqrt{\log(n^\train)}, (1+d_{\K})^{1/4} \big\} } \bigg)$ that $\max_{j} \Omega(s_{\gamma^\ast,j}) = o\bigg(\frac{  (n^\train)^{\frac{1}{4}} }{ (1+d_{\K})^{1/4} } \bigg)$, i.e., $\frac{ (1+d_{\K})^{1/4} }{  (n^\train)^{\frac{1}{4}} } \max_{j} \Omega(s_{\gamma^\ast,j}) = o(1)$. Thus, the above display implies that $\E_{\DtrainII}\big[ \U_y(\lambda_y)\big] = o(1)$. \\

Next, notice that for all $j\in[m]$:
\begin{align*}
	\sqrt{\Lambda} &\leq \Var_{\gamma^\ast}[s_{\gamma^\ast,j}] \\
	&\leq \frac{2\norm{s_{\gamma^\ast,j}}{\infty}}{4} \\
	&\leq \frac{\kappa \Omega(s_{\gamma^\ast,j}) }{2},
\end{align*} 
where the first line follows from Assumption \eqref{min Fisher Info}, and the second line is by Popoviciu's variance inequality. Thus, it follows that:
\begin{align*}
	\lambda_y &= O\Bigg( \frac{1}{\max_{j} \Omega(s_{\gamma^\ast,j})} \bigg( \frac{1+d_{\K}}{n^\train} \bigg)^{\frac{1}{4}}\Bigg) \\ 
	&= O\Bigg(  \bigg( \frac{1+d_{\K}}{n^\train} \bigg)^{\frac{1}{4}}\Bigg),
\end{align*}
which due to the Assumption of  $\frac{1+d_{\K}}{n^\train} = o(1)$, implies that $\lambda_y  = o(1)$. Next, observe that
\begin{align*}
	\max_y \lambda_y \Omega^2(s_{\gamma^\ast,y}) &= O\Bigg( \max_{j} \Omega(s_{\gamma^\ast,j}) \bigg( \frac{1+d_{\K}}{n^\train} \bigg)^{\frac{1}{4}}  \Bigg),
\end{align*}
so by the same argument used to show that $\E_{\DtrainII}\big[ \U_y(\lambda_y)\big] = o(1)$, it follows that $\max_y \lambda_y \Omega^2(s_{\gamma^\ast,y})  = o(1)$ as well. Lastly, observe that:
\begin{align*}
	(\min_y\lambda_y)^2 n^\train &\asymp \frac{1}{\max_{j} \Omega(s_{\gamma^\ast,j})^2} \bigg( \frac{1+d_{\K}}{n^\train} \bigg)^{\frac{1}{2}}n^\train \\ 
	&= \frac{1}{\max_{j} \Omega(s_{\gamma^\ast,j})^2}  \sqrt{(1+d_{\K})n^\train}  \\ 
	&= \bigg(   \frac{ ((1+d_{\K})n^\train)^{\frac{1}{4}} }{\max_{j} \Omega(s_{\gamma^\ast,j})}    \bigg)^{2}\\ 
	&\geq \bigg(   \frac{ (n^\train)^{\frac{1}{4}} }{\max_{j} \Omega(s_{\gamma^\ast,j})}    \bigg)^{2}.
\end{align*}
Since we already showed that $\max_{j} \Omega(s_{\gamma^\ast,j}) \big( \frac{1+d_{\K}}{n^\train} \big)^{\frac{1}{4}} = o(1)$, it must also be true that $\frac{\max_{j} \Omega(s_{\gamma^\ast,j})}{ (n^\train)^{\frac{1}{4}} } = o(1)$, meaning that $\frac{ (n^\train)^{\frac{1}{4}} }{\max_{j} \Omega(s_{\gamma^\ast,j})}\to\infty$. Thus, it follows that $(\min_y\lambda_y)^2 n^\train \to \infty$, and so Assumption \eqref{lambda and uniform convergence properties} holds true! 

Now, this present Corollary already assumes that Assumptions  \eqref{FIM Difference is Bounded}, \eqref{bounded pis}, \eqref{min Fisher Info} and \eqref{gamma hat properties} are true, and Assumptions \eqref{members of function class} and \eqref{Omega Properties} are already true by virtue of Corollary \eqref{RKHS Application}. So, since we just proved that  Assumption \eqref{lambda and uniform convergence properties} also holds, we can now invoke Theorem \eqref{Standardized Error Rate of Full Estimator}. Namely, we have that:
\begin{align*}
	\Bigg( \Bigg[\frac{1}{n^\test}+ \frac{1}{n^\train} \sum_{y=0}^{m}\frac{(\pi_y^\ast)^2}{\pi_y^\train}\Bigg] \Big( \FisherInfo{\gamma^\ast}^{-1} - \FisherInfo{\gamma^\ast;\textup{Cat}}^{-1}  \Big)  + \frac{1}{n^\test} \FisherInfo{\pi^\ast;\textup{Cat}}^{-1} \Bigg)^{-\frac{1}{2}} (\widehat{\pi} - \pi^\ast) = Z + \epsilon,
\end{align*}
where  $Z \goesto{d} \N(0,I_m)$, and $\epsilon \in \mathbb{R}^m$ is a random vector satisfying   
\begin{align}
	&\norm{\epsilon}{2}\nonumber\\
	 &=   O_{\P}\Bigg(\sqrt{ \frac{1}{\sqrt{n^\train}} + \max_y \lambda_y \Omega(s_{\gamma^\ast,y})^2 +  \frac{1}{\sqrt{\min_y\lambda_y}}e^{-Cn^\train} +  \max_y \E_{\DtrainII}\Big[ \U_y(\lambda_y)\Big] + \sqrt{\E_{\allDII}\Big[ \norm{\widehat{\gamma} - \gamma^\ast}{2}^2 \Big]}  }\Bigg) \nonumber \\ 
	&\qquad + O_{\P}\Bigg(  \frac{1}{\sqrt{\frac{1}{n^\test} + \frac{1}{n^\train}}} \bigg\{    e^{-C (\min_y\lambda_y)^2 n^\train}      + \P_{\allDII}\Big[\bignorm{ \widehat{\gamma} - \gamma^\ast }{2} \geq C \Big] \bigg\} \Bigg), \label{I luv hinata}
\end{align}
\normalsize
where  $C > 0$ is a global constant. The identity for $Z$ presented in the statement of the present Corollary follows from Theorem \eqref{Standardized Error Rate of Full Estimator}. Now, it remains for us to show that, in fact,
\begin{align*}
	\norm{\epsilon}{2} &= O_{\P}\Bigg( \frac{1}{(n^\train)^{1/4}}  +  \bigg(\frac{1+d_\K}{n^\train}\bigg)^{\tfrac{1}{8}}\sqrt{\max_{y} \Omega(s_{\gamma^\ast,y})} + \Big(\E_{\allDII}\Big[ \norm{\widehat{\gamma} - \gamma^\ast}{2}^2 \Big]\Big)^{\frac{1}{4}} + \frac{\P_{\allDII}\big[\norm{ \widehat{\gamma} - \gamma^\ast }{2} \geq C \big]}{\sqrt{\tfrac{1}{n^\test} + \tfrac{1}{n^\train}} }\Bigg),
\end{align*}
and furthermore, that the RHS above is $O_{\P}(1)$. The latter task is quite simple: the second term is $o(1)$ by the previous work done in the proof of this Corollary, and the third and fourth terms are $o(1)$ due to Assumption \eqref{gamma hat properties}. \\

Thus, all that remains is for us to show that the display above is true. Towards that end, note that we have already shown that 
\[
	\E_{\DtrainII}\big[ \U_y(\lambda_y)\big], \text{ }\max_y \lambda_y \Omega^2(s_{\gamma^\ast,y}) = O\Bigg(  \bigg( \frac{1+d_{\K}}{n^\train} \bigg)^{\frac{1}{4}}  \max_{j} \Omega(s_{\gamma^\ast,j})  \Bigg),
\]
so it follows from line \eqref{I luv hinata} that 
\begin{align}
	\norm{\epsilon}{2} &=   O_{\P}\Bigg(\sqrt{ \frac{1}{\sqrt{n^\train}}  +  \frac{1}{\sqrt{\min_y\lambda_y}}e^{-Cn^\train} + \bigg( \frac{1+d_{\K}}{n^\train} \bigg)^{\frac{1}{4}}  \max_{j} \Omega(s_{\gamma^\ast,j})   + \sqrt{\E_{\allDII}\Big[ \norm{\widehat{\gamma} - \gamma^\ast}{2}^2 \Big]}  }\Bigg) \nonumber \\ 
	&\qquad + O_{\P}\Bigg(  \frac{1}{\sqrt{\frac{1}{n^\test} + \frac{1}{n^\train}}} \bigg\{    e^{-C (\min_y\lambda_y)^2 n^\train}      + \P_{\allDII}\Big[\bignorm{ \widehat{\gamma} - \gamma^\ast }{2} \geq C \Big] \bigg\} \Bigg). \label{I luv sakura}
\end{align}
Next, we will show that $ \frac{1}{\sqrt{\min_y\lambda_y}}e^{-Cn^\train} = o\Bigg(  \bigg( \frac{1+d_{\K}}{n^\train} \bigg)^{\frac{1}{4}}  \max_{j} \Omega(s_{\gamma^\ast,j})  \Bigg)$. Towards that end, observe that:
\begin{align*} 
	\frac{1}{\sqrt{\min_y\lambda_y}}e^{-Cn^\train} &= O\Bigg( \sqrt{\max_{j} \Omega(s_{\gamma^\ast,j})}  \bigg( \frac{n^\train}{1+d_{\K}} \bigg)^{\frac{1}{8}} e^{-Cn^\train} \Bigg) \\
	&= O\Bigg( \sqrt{\max_{j} \Omega(s_{\gamma^\ast,j})}  \big( n^\train \big)^{\frac{1}{8}} e^{-Cn^\train} \Bigg) \\ 
	&= O\Bigg( \sqrt{\frac{\max_{j} \Omega(s_{\gamma^\ast,j})}{\big(n^\train\big)^{\frac{1}{4}} } \cdot \big( n^\train \big)^{\frac{1}{2}} e^{-2Cn^\train} }   \Bigg) \\ 
	&= O\Bigg( \sqrt{  \bigg(\frac{1+d_{\K}}{n^\train}\bigg)^{\frac{1}{4}} \max_{j} \Omega(s_{\gamma^\ast,j}) \cdot \big( n^\train \big)^{\frac{1}{2}} e^{-2Cn^\train} }   \Bigg) \\ 
	&= o\Bigg(  \bigg(\frac{1+d_{\K}}{n^\train}\bigg)^{\frac{1}{4}} \max_{j} \Omega(s_{\gamma^\ast,j})   \Bigg),
\end{align*}
where the last line is based on the presumption that $\sqrt{n^\train} e^{-2Cn^\train} = o\Big( \big( \frac{1+d_{\K}}{n^\train} \big)^{\frac{1}{4}} \max_j\Omega(s_{\gamma^\ast,j}) \Big)$. We will now verify this presumption. Since $\frac{2}{\kappa}\sqrt{\Lambda} \leq \Omega(s_{\gamma^\ast,j})$ for all $j\in[m]$, we have that:
\begin{align*}
	\sqrt{n^\train} e^{-2Cn^\train} = o\Bigg( \bigg( \frac{1+d_{\K}}{n^\train} \bigg)^{\frac{1}{4}} \max_j\Omega(s_{\gamma^\ast,j}) \Bigg) &\Longleftarrow \sqrt{n^\train} e^{-2Cn^\train} = o\Bigg( \bigg( \frac{1+d_{\K}}{n^\train} \bigg)^{\frac{1}{4}} \Bigg)  \\ 
	 &\iff (n^\train)^{\frac{3}{4}} e^{-2Cn^\train} = o\Big( (1+d_{\K})^{\frac{1}{4}}\Big) \\ 
	 &\iff (n^\train)^{3} e^{-8Cn^\train} = o\Big( 1+d_{\K}\Big) \\
	 &\Longleftarrow (n^\train)^{3} e^{-8Cn^\train} = o( 1),
\end{align*}
and the last line above is clearly true, meaning that it is indeed the case that $\sqrt{n^\train} e^{-2Cn^\train} = o\Big( \big( \frac{1+d_{\K}}{n^\train} \big)^{\frac{1}{4}} \max_j\Omega(s_{\gamma^\ast,j}) \Big)$. Thus, we have proved that:
\[
	\frac{1}{\sqrt{\min_y\lambda_y}}e^{-Cn^\train}  = o\Bigg(  \bigg(\frac{1+d_{\K}}{n^\train}\bigg)^{\frac{1}{4}} \max_{j} \Omega(s_{\gamma^\ast,j})   \Bigg),
\]
so it follows from line \eqref{I luv sakura} that:
\begin{align}
	\norm{\epsilon}{2} &=   O_{\P}\Bigg(\sqrt{ \frac{1}{\sqrt{n^\train}}  + \bigg( \frac{1+d_{\K}}{n^\train} \bigg)^{\frac{1}{4}}  \max_{j} \Omega(s_{\gamma^\ast,j})   + \sqrt{\E_{\allDII}\Big[ \norm{\widehat{\gamma} - \gamma^\ast}{2}^2 \Big]}  }\Bigg) \nonumber \\ 
	&\qquad + O_{\P}\Bigg(  \frac{1}{\sqrt{\frac{1}{n^\test} + \frac{1}{n^\train}}} \bigg\{    e^{-C (\min_y\lambda_y)^2 n^\train}      + \P_{\allDII}\Big[\bignorm{ \widehat{\gamma} - \gamma^\ast }{2} \geq C \Big] \bigg\} \Bigg). \label{I luv ino}
\end{align}
Finally, we will  show that 
\[
	 \frac{e^{-C (\min_y\lambda_y)^2 n^\train}}{\sqrt{\frac{1}{n^\test} + \frac{1}{n^\train}}} = o\Bigg(  \bigg(\frac{1+d_{\K}}{n^\train}\bigg)^{\frac{1}{8}} \sqrt{\max_{j} \Omega(s_{\gamma^\ast,j})} \Bigg).
\]
Towards that end, note that the requirement that   $\lambda_y \asymp \frac{1}{\max_{j} \Omega(s_{\gamma^\ast,j})} \big( \frac{1+d_{\K}}{n^\train} \big)^{\frac{1}{4}}$ implies that there exists a global constant $B > 0$ such that $\lambda_y \geq B \frac{1}{\max_{j} \Omega(s_{\gamma^\ast,j})} \big( \frac{1+d_{\K}}{n^\train} \big)^{\frac{1}{4}}$. Thus, we have that:
\begin{align}
	 \frac{e^{-C (\min_y\lambda_y)^2 n^\train}}{\sqrt{\frac{1}{n^\test} + \frac{1}{n^\train}}} &\leq \sqrt{n^\train}e^{-C (\min_y\lambda_y)^2 n^\train} \nonumber \\ 
	 &\leq \sqrt{n^\train}\exp\Bigg\{-CB^2 \frac{1}{\max_{j} \Omega(s_{\gamma^\ast,j})^2} \bigg( \frac{1+d_{\K}}{n^\train} \bigg)^{\frac{1}{2}}  n^\train \Bigg\} \nonumber \\ 
	 &\leq \sqrt{n^\train}\exp\Bigg\{-CB^2 \frac{1}{\max_{j} \Omega(s_{\gamma^\ast,j})^2}  \sqrt{n^\train} \Bigg\} \nonumber 
\end{align}

Now, observe that
\[
	\sqrt{n^\train}\exp\Bigg\{-CB^2 \frac{1}{\max_{j} \Omega(s_{\gamma^\ast,j})^2}  \sqrt{n^\train} \Bigg\} = o\Bigg(  \bigg(\frac{1+d_{\K}}{n^\train}\bigg)^{\frac{1}{8}} \sqrt{\max_{j} \Omega(s_{\gamma^\ast,j})} \Bigg)
\]
\[
	\Longleftarrow  \sqrt{n^\train}\exp\Bigg\{-CB^2 \frac{1}{\max_{j} \Omega(s_{\gamma^\ast,j})^2}  \sqrt{n^\train} \Bigg\} = o\Bigg(  \bigg(\frac{1}{n^\train}\bigg)^{\frac{1}{8}} \sqrt{\max_{j} \Omega(s_{\gamma^\ast,j})} \Bigg)
\]
\[
	\iff  (n^\train)^{\frac{5}{8}} \exp\Bigg\{-CB^2 \frac{1}{\max_{j} \Omega(s_{\gamma^\ast,j})^2}  \sqrt{n^\train} \Bigg\} = o\Bigg( \sqrt{\max_{j} \Omega(s_{\gamma^\ast,j})} \Bigg). 
\]
\[
	\Longleftarrow (n^\train)^{\frac{5}{8}} \exp\Bigg\{-CB^2 \frac{1}{\max_{j} \Omega(s_{\gamma^\ast,j})^2}  \sqrt{n^\train} \Bigg\} = o( 1 )
\]
\[
	\iff   \exp\Bigg\{-CB^2 \frac{1}{\max_{j} \Omega(s_{\gamma^\ast,j})^2}  \sqrt{n^\train} + \frac{5}{8} \log(n^\train)  \Bigg\} = o( 1 )
\]
\[
	\Longleftarrow	 \frac{1}{\max_{j} \Omega(s_{\gamma^\ast,j})^2}  \sqrt{n^\train} - \log(n^\train)  \to \infty
\]
\[
	\iff \log(n^\train)\Bigg[  \frac{ \sqrt{n^\train}}{\max_{j} \Omega(s_{\gamma^\ast,j})^2} \frac{1}{\log(n^\train)}     -       1\Bigg] \to \infty
\]
\[
	\Longleftarrow \frac{\max_{j} \Omega(s_{\gamma^\ast,j})^2}{\sqrt{n^\train}}  \log(n^\train)    = o(1)
\]
\[
	\iff \max_{j} \Omega(s_{\gamma^\ast,j})^2 = o\bigg(\frac{\sqrt{n^\train}}{\log(n^\train)}\bigg)
\]
\[
	\iff \max_{j} \Omega(s_{\gamma^\ast,j}) = o\bigg(\frac{  (n^\train)^{\frac{1}{4}} }{ \sqrt{\log(n^\train)} }\bigg)
\]
\[
	\iff \max_{j} \Omega(s_{\gamma^\ast,j}) = o\bigg(\frac{  (n^\train)^{\frac{1}{4}} }{ \max\{\sqrt{\log(n^\train)}, (1+d_{\K})^{1/4} } \bigg)
\]
Thus, the assumption that $\max_{j} \Omega(s_{\gamma^\ast,j}) = o\Big(\frac{  (n^\train)^{\frac{1}{4}} }{ \max\{ \sqrt{\log(n^\train)}, (1+d_{\K})^{1/4}  \} }\Big)$ implies that 
\[
	 \frac{e^{-C (\min_y\lambda_y)^2 n^\train}}{\sqrt{\frac{1}{n^\test} + \frac{1}{n^\train}}} = o\Bigg(  \bigg(\frac{1+d_{\K}}{n^\train}\bigg)^{\frac{1}{8}} \sqrt{\max_{j} \Omega(s_{\gamma^\ast,j})} \Bigg).
\]
Thus, it follows from line \eqref{I luv ino} that:
\begin{align*}
	\norm{\epsilon}{2} &=   O_{\P}\Bigg( \frac{1}{(n^\train)^{\frac{1}{4}} }  + \bigg( \frac{1+d_{\K}}{n^\train} \bigg)^{\frac{1}{8}}  \sqrt{ \max_{j} \Omega(s_{\gamma^\ast,j}) }   + \Big(\E_{\allDII}\Big[ \norm{\widehat{\gamma} - \gamma^\ast}{2}^2 \Big]\Big)^{\frac{1}{4}} +  \frac{ \P_{\allDII}\big[\norm{ \widehat{\gamma} - \gamma^\ast }{2} \geq C \big]}{\sqrt{\frac{1}{n^\test} + \frac{1}{n^\train}}}  \Bigg)\\
	 &=   O_{\P}\Bigg(  \bigg( \frac{1+d_{\K}}{n^\train} \bigg)^{\frac{1}{8}}  \sqrt{ \max_{j} \Omega(s_{\gamma^\ast,j}) }   + \Big(\E_{\allDII}\Big[ \norm{\widehat{\gamma} - \gamma^\ast}{2}^2 \Big]\Big)^{\frac{1}{4}} +  \frac{ \P_{\allDII}\big[\norm{ \widehat{\gamma} - \gamma^\ast }{2} \geq C \big]}{\sqrt{\frac{1}{n^\test} + \frac{1}{n^\train}}}  \Bigg),
\end{align*}
where the second line is because $d_{\K} \geq 0$ and $\max_j\Omega(s_{\gamma^\ast,j})$ can be lower bounded by a fixed constant.

\end{proof}
\vspace{0.4in}

\subsection{Fixed $\tau$-IID Regime}

\subsubsection{Assumptions}

These are Assumptions \eqref{Fixed Tau Regime: Mixture FIM not too small}, \eqref{Fixed Tau Regime: Cat minus Mixture FIM}, \eqref{Fixed Tau Regime: expo family}, \eqref{Fixed Tau Regime: H closed under scalar multiplication} and \eqref{Fixed Tau Regime: Omega Properties}, which are presented in the Theoretical Results section of the paper. Note that these Assumptions are stated in terms of the more general $\tau_n$-IID Regime, but can be used for the  Fixed $\tau$-IID Regime by substituting $\tau$ for $\tau_n$.

\subsubsection{Definition of $\tau$-RAL Estimator} \label{subsection: tau RAL estimator definition}

Now, let $\mathbf{M}^{\text{sub},\tau} = \Big\{ \Jcal^{\pi,\rho,\tau} \ \Big\lvert \ (\pi,\rho) \in \Theta^\text{sub} \Big\} \subset \textbf{M}^{\text{semi},\tau}$ denote a parametric submodel of $\textbf{M}^{\text{semi},\tau}$ parameterized by $(\pi,\rho) \in \Theta^\text{sub}$, where $\Theta^\text{sub} = \Delta \times \R$ and $\R$ is an open set  of finite dimensional vectors. Let $(\pi^\ast,\rho^\ast)$ be the true parameter for the submodel, so that $\Jcal^{\pi^\ast,\rho^\ast,\tau} = \Jcal^{\pi^\ast,\mathbf{p},\tau}$. $\mathbf{M}^{\text{sub},\tau}$ is called a \textbf{regular parametric submodel} if $\sqrt{ \Jcal^{\pi,\rho,\tau}}$ satisfies the mean-square differentiability conditions in Appendix A of \cite{Newey1990} for all $(\pi,\rho)  \in \Theta^\text{sub}$. 

Next, let $\{(\pi_n,\rho_n )\} \subset \Theta^\text{sub}$ be any sequence such that $ \sqrt{n} \bignorm{  \icol{ \pi_n \\  \rho_n } - \icol{ \pi^\ast \\ \rho^\ast} }{2}$ is bounded. Let the corresponding local data generating process, $\text{LDGP}(\{(\pi_n,\rho_n,\tau )\})$, be one where $Z_1,\dots,Z_n \simiid \Jcal^{\pi_n, \rho_n, \tau }$ for each value of $n$. An estimator $\widetilde{\pi}_n$ is called "\textbf{regular} w.r.t $\mathbf{M}^{\text{sub},\tau}$ in the fixed $\tau$ IID regime" if the limiting distribution of $ \sqrt{n}(\widetilde{\pi}_n-\pi_n )$ is the same for all sequences $\{(\pi_n,\rho_n)\}$ satisfying the aforementioned property. The estimator $\widetilde{\pi}_n$ is called "regular w.r.t $\mathbf{M}^{\text{semi},\tau}$ in the fixed $\tau$-IID regime" if it is regular w.r.t all regular parametric submodels in the fixed $\tau$-IID regime.

Finally, $\widetilde{\pi}_n$ is \textbf{asymptotically linear} if there exists $\mathbb{R}^m$-valued functions $\psi$ and $\delta$ such that:
\begin{itemize}
	\item $\sqrt{n}(\widetilde{\pi}_n - \pi^\ast) = \frac{1}{\sqrt{n}} \sum_{i=1}^n \psi(Z_i)  + \delta(Z_{1:n})$
	\item $\E_{\Jcal^{\pi^\ast,\mathbf{p},\tau} }[ \psi(Z)] = 0$ and $\E_{\Jcal^{\pi^\ast,\mathbf{p},\tau} }[ \psi(Z)\psi(Z)' ]$ is finite and nonsingular
	\item $ \delta(Z_{1:n}) \goesto{p} 0$.
\end{itemize}

An estimator that is both regular and asymptotically linear w.r.t $\mathbf{M}^{\text{semi},\tau}$ in the fixed-$\tau$ IID is called $\mathbf{\tau}$\textbf{-RAL}.

\subsubsection{Lemmas}

It is important to note that there are several notational discrepancies between the Fixed $\tau$-IID Regime and the previous Fixed Sequence Regime:
\begin{enumerate}
	\item In the Fixed Sequence Regime, $\pi_y^\train$ was the fraction of the training data that belonged to class $y$. In the Fixed $\tau$-IID Regime, $\pi_y^\train$ is the probability of a training data point belonging to class $y$; the actual fraction of the training data that belongs to class $y$ is random, and is denoted by $\widehat{\pi}_y^\train$.
	\item In the Fixed Sequence Regime, $\gamma^\ast$ was a non-random function that depended on the number of test/train samples, the fraction of the train samples that belong to each class, and $\pi^\ast$. In the Fixed $\tau$-IID Regime, all the aforementioned quantities (except for $\pi^\ast$) are now random, and so $\gamma^\ast$ is a random variable (defined in line \eqref{definition of random gamma ast}). The non-random counterpart to $\gamma^\ast$ is $\gamma^\ast(\tau)$ (defined in Lemma \eqref{A Particular Smooth Parametric Submodel}), which has all the random parts of $\gamma^\ast$ replaced by their expected values. Additionally, in the remainder of this manuscript, whenever we work with $\gamma^\ast$, we do not have to worry about any issues arising from division by $0$; this is because the only time that $\gamma^\ast$ appears in our analysis is in the context of the estimation procedure after step $2$, which we only reach when $N^\test$, $N^\train$ and $\widehat{\pi}_y^\train$ are all nonzero (see step 1 of the procedure). 
	\item  In the Fixed Sequence Regime, the score function for the mixture at a probability parameter $\beta$ was denoted by $s_\beta$. In the Fixed $\tau$-IID Regime, we use $S_{\pi,\rho}$ to refer to the score function for a given parametric submodel that parameterizes the class densities by a vector $\rho$. When the score function is for only one of $\pi$ or $\rho$, then we write $S_{\pi}$ or $S_{\rho}$, respectively.
\end{enumerate}
\vspace{0.4in}

\begin{lemma}[\uline{Smooth Score Function for $\pi$}] \label{Smooth Score Function for Pi}
The density $\Jcal^{\pi,\mathbf{p},\tau}$ is smooth with respect to $\pi \in \Delta$ in the sense of \cite{Newey1990}, with score function $S_{\pi}$ for the parameter of interest equal to
\begin{align*}
	S_{\pi}(z) &=  d  s_{\pi}(x) \I{ p_{\pi}(x) > 0},
\end{align*}
which, in turn, is equal to $ \partialDerivative{\pi} \Big\{ \log \Jcal^{\pi,\mathbf{p},\tau}(z) \Big\}$ a.e. $[\Jcal^{\pi,\mathbf{p},\tau}]$.
\end{lemma} 
\vspace{0.4in}

\begin{lemma}[\uline{A Particular Smooth Parametric Submodel}] \label{A Particular Smooth Parametric Submodel}
Let $\Theta^\textup{sub} = \Delta \times \R$ where 
\[
	\R := \Big\{     \rho \in \mathbb{R}^{m}     \ \Big\lvert \   \norm{\rho}{2} < B   \Big\}
\]
and $B$ is the constant given in Assumption \eqref{Fixed Tau Regime: expo family}. For each $j\in \Y$ and $\rho \in \R$, define
\[
	q_j^\rho(x) := p_j(x) e^{\rho' \frac{\pi_j^\ast}{\pi_j^\train} s_{\gamma^\ast(\tau)}(x) - c_j(\rho)}
\] 
where
\begin{align*}
	\gamma_j^\ast(\tau) := \frac{\frac{\pi_j^\ast}{\tau} + \frac{1}{1-\tau} \frac{(\pi_j^\ast)^2}{\pi_j^\train} }{ \frac{1}{\tau}  + \frac{1}{1-\tau} \sum\limits_{k=0}^{m} \frac{(\pi_k^\ast)^2}{\pi_k^\train} },
\end{align*}
and let 
\[
	\mathcal{J}^{\pi,\rho,\tau}(z) = \big( \tau q_{\pi}^{\rho}(x) \big)^d \Bigg((1-\tau) \prod_{j=0}^{m} \big(\pi_j^\train q_j^{\rho}(x) \big)^{\I{y=j}} \Bigg)^{1-d} \Ibig{x\in\X, \text{ } y \in \Y, \text{ } d \in \{0,1\}  }.
\]
Then, under Assumption \eqref{Fixed Tau Regime: expo family}, the set 
\[
	\mathbf{M}^{\textup{sub},\tau} = \Big\{ \Jcal^{\pi,\rho,\tau} \ \Big\lvert \ (\pi,\rho) \in \Theta^\textup{sub} \Big\}
\]
is a smooth (in the sense of \cite{Newey1990}) parametric submodel of $\mathbf{M}^{\textup{semi},\tau}$, with score function $S_{\pi,\rho}^{\textup{sub},\tau}$ for both the parameter of interest and nuisance parameter equal to
\begin{equation*}
	S_{\pi,\rho}^{\textup{sub},\tau}(z) = 
	\arraycolsep=1.4pt\def\arraystretch{1}
	\begin{pmatrix}
		   d s_{\pi}^{\rho}(x)  \Ibig{  q_{\pi}^{\rho}(x) > 0} \\ 
		   \vartheta_{\pi,\rho}^{\textup{sub},\tau,1}(z)  + \vartheta_{\pi,\rho}^{\textup{sub},\tau,2}(z) 
	\end{pmatrix},
\end{equation*}
where
\[
	\vartheta_{\pi,\rho}^{\textup{sub},\tau,1}(z) := d \sum\limits_{j=0}^m \pi_j  \frac{\pi_j^\ast}{\pi_j^\train}  \frac{  q_j^{\rho}(x) }{ q_{\pi}^\rho(x) }  (s_{\gamma^\ast(\tau)}(x) - \E_j^{\rho}[s_{\gamma^\ast(\tau)}])\Ibig{q_{j}^\rho(x) > 0} 
\]
and
\[
	\vartheta_{\pi,\rho}^{\textup{sub},\tau,2}(z): =  (1-d)  \sum\limits_{j=0}^{m} \I{y=j}  \frac{\pi_j^\ast}{\pi_j^\train}  \big(s_{\gamma^\ast(\tau)}(x) - \E_j^{\rho}[s_{\gamma^\ast(\tau)}] \big)\Ibig{q_j^\rho(x) > 0}.
\]

\end{lemma} 
\vspace{0.4in}

\begin{lemma}[\uline{Semiparametric Nuisance Tangent Set is a Subset of $\G$}] \label{Nuisance Tangent Set is a Subset of G}
The semiparametric nuisance tangent set $\T$ defined in \cite{Newey1990} is a subset of $\G$, where $\G$ is given by 
\begin{align*}
	\G &:= \Bigg\{ g \text{ }\bigg|\text{ }\forall j\in\Y \text{ } f_j:\X \mapsto \mathbb{R}^m, \text{ }  \E_{j}[f_j] = 0, \text{ } \E_j[\norm{f_j}{2}^2] < \infty,   \\
	&\qquad\qquad g(z) =  d \sum_{j=0}^{m} \pi_j^\ast \frac{p_{j} (x)}{p_{\pi^\ast}(x) } f_j(x) + (1-d) \sum_{j=0}^{m} \I{y=j} f_j(x) \quad \textup{ a.s. } \big[ \Jcal^{\pi^\ast,\mathbf{p},\tau} \big] \Bigg\}.
\end{align*}
\end{lemma} 
\vspace{0.4in}

\begin{lemma}[\uline{Projection of $S_{\pi^\ast}$ onto $\G$}] \label{Projection of Ordinary Score onto G}

Define:
\begin{align*}
	\mathbf{V}^\textup{eff}(\tau) := \bigg[ \frac{1}{\tau} + \frac{1}{1-\tau} \sum_{k=0}^{m} \frac{(\pi_k^\ast)^2}{\pi_k^\train} \bigg]\Big(\FisherInfo{ \gamma^\ast(\tau) }^{-1} - \FisherInfo{\gamma^\ast(\tau);\textup{Cat}}^{-1} \Big) + \frac{1}{\tau}\FisherInfo{\pi^\ast;\textup{Cat}}^{-1} 
\end{align*}
\begin{align*}
	\psi_\tau^\textup{eff}(z) &:= \frac{d}{\tau} \FisherInfo{\gamma^\ast(\tau)}^{-1}\big(s_{\gamma^\ast(\tau)}(x)  -  \E_{\pi^\ast}[s_{\gamma^\ast(\tau)}]\big) \\
	&\qquad - \frac{1-d}{1-\tau} \FisherInfo{\gamma^\ast(\tau)}^{-1} \sum_{j=0}^m \I{y=j} \frac{\pi_j^\ast}{\pi_j^\train}\big(s_{\gamma^\ast(\tau)}(x)  -  \E_{j}[s_{\gamma^\ast(\tau)}]\big).
\end{align*}
Then, under Assumption \eqref{Fixed Tau Regime: Mixture FIM not too small}, we have that:
\[
	 \frac{ (\xi)^2 \nu }{\tau(1-\tau)}  \bigg(\frac{L}{m+1}\bigg)^2  \leq \minEval{\mathbf{V}^\textup{eff}(\tau)} \leq \maxEval{\mathbf{V}^\textup{eff}(\tau)} \leq \bigg[ \frac{1}{\tau} + \frac{1}{(1-\tau)\xi}  \bigg] \frac{1}{ \sqrt{\Lambda} } + \frac{1}{2\tau}.
\]
Further, under Assumptions   \eqref{Fixed Tau Regime: Mixture FIM not too small}  and \eqref{Fixed Tau Regime: Cat minus Mixture FIM}, we have that 
\[ 
	d s_{\pi^\ast} - \mathbf{V}^\textup{eff}(\tau)^{-1} \psi_\tau^\textup{eff} \in \G
\] 
and 
\[
	\E_{\Jcal^{\pi^\ast,\mathbf{p}, \tau}} \Big[ g' \big(S_{\pi^\ast} - \big( Ds_{\pi^\ast} -  \mathbf{V}^\textup{eff}(\tau)^{-1}\psi_\tau^\textup{eff} \big) \big)\Big] = 0 \qquad \forall \ g\in\G.
\]
\end{lemma} 
\vspace{0.4in}

\begin{lemma}[\uline{Largest CRLB}] \label{Largest CRLB}
Let $\mathbf{M}^{\textup{sub},\tau}$ denote the smooth parametric submodel defined in Lemma \eqref{A Particular Smooth Parametric Submodel}, and let $\mathbf{M}^{\widetilde{\textup{sub}},\tau}$ denote an arbitrary smooth parametric submodel. Then, under Assumptions \eqref{Fixed Tau Regime: Mixture FIM not too small}, \eqref{Fixed Tau Regime: Cat minus Mixture FIM} and \eqref{Fixed Tau Regime: expo family}, we have that 
\[
	\CRLB{\mathbf{M}^{\widetilde{\textup{sub}},\tau}} \preceq \CRLB{\mathbf{M}^{\textup{sub},\tau}} = \mathbf{V}^\textup{eff}(\tau),
\]
where $\CRLB{\mathbf{M}^{\widetilde{\textup{sub}},\tau}}$ and $ \CRLB{\mathbf{M}^{\textup{sub},\tau}}$  denote the Cramer Rao Lower Bounds for learning only $\pi^\ast$ in the submodels $\mathbf{M}^{\widetilde{\textup{sub}},\tau}$ and $\mathbf{M}^{\textup{sub},\tau}$, respectively.
\end{lemma}
\vspace{0.4in}

\subsubsection{Proofs}

\begin{proof}[\uline{Proof of Lemma \eqref{Smooth Score Function for Pi}}]
We will show that $\Jcal^{\pi,\mathbf{p},\tau}$ is smooth with respect to $\pi \in \Delta$ in the sense of Definition A.1 of \cite{Newey1990}. This requires showing three conditions, and doing so will allow us to also prove the identities for $S_{\pi}$ stated in Lemma \eqref{Smooth Score Function for Pi}. 

The first condition is that $\Delta$ must be an open set, which is clearly the case. The second condition is that there exists a measure $\widetilde{\mu}$ dominating $\Jcal^{\pi,\mathbf{p},\tau}(z)$ such that $\Jcal^{\pi,\mathbf{p},\tau}(z)$ is continuous on $\Delta$ a.s. $[\widetilde{\mu}]$. This is trivially true-- while  $\widetilde{\mu}$ can be constructed from counting measures and $\mu$, the continuity of $\Jcal^{\pi,\mathbf{p},\tau}(z)$ on $\Delta$ holds, in fact, for all $z$: when $d = 0$,  $\Jcal^{\pi,\mathbf{p},\tau}(z)$ does not depend on $\pi$, and when $d=1$,  $\Jcal^{\pi,\mathbf{p},\tau}(z)$ is linear in $\pi$. 

The third condition is that $\sqrt{\Jcal^{\pi,\mathbf{p},\tau}(z) }$ is $\widetilde{\mu}$-mean square continuously differentiable with respect to $\pi\in \Delta$ with derivative $D(z,\pi)$, where 
\[
	D(z,\pi) := \frac{d}{2} \sqrt{\tau p_{\pi}(x)} s_{\pi}(x).
\]
This, in turn, requires showing three different properties. For every $\pi \in \Delta$ and sequence $\pi^{(i)} \to \pi$, we must show that:
\begin{enumerate}[label=(\alph*)]
	\item $\int \norm{D(z,\pi)}{2}^2 d\widetilde{\mu}(z) < \infty$.
	\item $\int \norm{D(z,\pi^{(i)}) - D(z,\pi)}{2}^2 d\widetilde{\mu}(z) \to 0$.
	\item $\int \big[ \Jcal^{\pi^{(i)},\mathbf{p},\tau}(z)^{1/2}  - \Jcal^{\pi,\mathbf{p},\tau}(z)^{1/2} - D(z,\pi)' (\pi^{(i)} - \pi) \big]^2  d\widetilde{\mu}(z) \Big/ \norm{\pi^{(i)} - \pi}{2}^2 \to 0 $. \\ 
\end{enumerate}

\uline{Let's start with showing property (a)}. Observe that:
\begin{align*}
	\int \norm{D(z,\pi)}{2}^2 d\widetilde{\mu}(z) &= \frac{\tau}{4} \int d  p_{\pi}(x)   \norm{s_{\pi}(x) }{2}^2 d\widetilde{\mu}(z) \\ 
	&= \frac{\tau}{4} \E_{\pi} \Big[ \norm{s_{\pi}(X) }{2}^2\Big]  \\ 
	&\leq   \frac{\tau m}{\xi^2}  < \infty,
\end{align*}
where the last line is because $\pi \in \Delta \implies \abs{s_{\pi,y}(x)} \leq \frac{2}{\xi} \implies \norm{ s_{\pi}(x) }{2}^2 \leq \frac{4m}{\xi^2}$. \\ 

\uline{Next, let's prove property (b)}. Observe that:
\begin{align*}
	\bigg(  \frac{1}{p_{\pi^{(i)}}(x)^{1/2}}  -  \frac{1}{p_{\pi}(x)^{1/2}} \bigg)^2 &= \bigg( \frac{p_{\pi}(x)^{1/2} - p_{\pi^{(i)}}(x)^{1/2}}{ p_{\pi^{(i)}}(x)^{1/2}  p_{\pi}(x)^{1/2} }   \bigg)^2 \\ 
	&= \Bigg( \frac{p_{\pi}(x) - p_{\pi^{(i)}}(x) }{ p_{\pi^{(i)}}(x)^{1/2}  p_{\pi}(x)^{1/2} \big(p_{\pi}(x)^{1/2} + p_{\pi^{(i)}}(x)^{1/2}  \big) }   \Bigg)^2 \\ 
	&= \Bigg( \frac{p_{\pi}(x) - p_{\pi^{(i)}}(x) }{ p_{\pi^{(i)}}(x)^{1/2}  p_{\pi}(x) + p_{\pi}(x)^{1/2} p_{\pi^{(i)}}(x) }   \Bigg)^2 \\ 
	&\leq  \frac{ \big( p_{\pi}(x) - p_{\pi^{(i)}}(x) \big)^2 }{ p_{\pi^{(i)}}(x)  p_{\pi}(x)^2 }.
\end{align*}
Also observe that:
\begin{align*}
	D(z,\pi^{(i)}) - D(z,\pi) &=  \frac{d\sqrt{\tau}}{2} \Big(  p_{\pi^{(i)}}(x)^{1/2} s_{\pi^{(i)} }(x) -  p_{\pi}(x)^{1/2} s_{\pi}(x) \Big) \\ 
	&=  \frac{d\sqrt{\tau}}{2} [p_1(x) - p_0(x),\dots, p_m(x) - p_0(x) ]' \bigg(  \frac{1}{p_{\pi^{(i)}}(x)^{1/2}}  -  \frac{1}{p_{\pi}(x)^{1/2}} \bigg).
\end{align*}
Thus, it follows that:
\begin{align*}
	 &\norm{D(z,\pi^{(i)}) - D(z,\pi) }{2}^2\\
	 &= \frac{d\tau}{4} \Bignorm{[p_1(x) - p_0(x),\dots, p_m(x) - p_0(x) ]'}{2}^2 \bigg(  \frac{1}{p_{\pi^{(i)}}(x)^{1/2}}  -  \frac{1}{p_{\pi}(x)^{1/2}} \bigg)^2 \\
	&\leq \frac{d\tau}{4} \Bignorm{[p_1(x) - p_0(x),\dots, p_m(x) - p_0(x) ]'}{2}^2  \frac{ \big( p_{\pi}(x) - p_{\pi^{(i)}}(x) \big)^2 }{ p_{\pi^{(i)}}(x)  p_{\pi}(x)^2 } \\
	&= \frac{d\tau}{4} \norm{ s_{\pi}(x)}{2}^2  \frac{ \big( p_{\pi}(x) - p_{\pi^{(i)}}(x) \big)^2 }{ p_{\pi^{(i)}}(x)  } \\
	&= \frac{d\tau}{4} \norm{ s_{\pi}(x)}{2}^2  \frac{ \bigdotprod{(p_1(x) - p_0(x),\dots,p_m(x)-p_0(x)) }{ \pi^{(i)} - \pi }^2 }{ p_{\pi^{(i)}}(x)  } \\
	&\leq \frac{d\tau}{4} \norm{ s_{\pi}(x)}{2}^2  \frac{ \norm{(p_1(x) - p_0(x),\dots,p_m(x)-p_0(x))}{2}^2 }{ p_{\pi^{(i)}}(x)  } \norm{\pi^{(i)} - \pi}{2}^2 \\
	&= \frac{d\tau}{4} p_{\pi^{(i)}}(x) \norm{ s_{\pi}(x)}{2}^2  \frac{ \norm{(p_1(x) - p_0(x),\dots,p_m(x)-p_0(x))}{2}^2 }{ p_{\pi^{(i)}}(x)^2  } \norm{\pi^{(i)} - \pi}{2}^2 \\
	&= \frac{d\tau}{4} p_{\pi^{(i)}}(x) \norm{ s_{\pi}(x)}{2}^2   \norm{ s_{\pi^{(i)}}(x) }{2}^2  \norm{\pi^{(i)} - \pi}{2}^2 \\ 
	&\leq \frac{d\tau}{4} \frac{16m^2}{\xi^4}  p_{\pi^{(i)}}(x)   \norm{\pi^{(i)} - \pi}{2}^2 \\ 
	&\leq \frac{4\tau m^2 d}{\xi^4}  p_{\pi^{(i)}}(x)   \norm{\pi^{(i)} - \pi}{2}^2.
\end{align*}

Thus, it follows that
\begin{align*}
	\int \norm{D(z,\pi^{(i)}) - D(z,\pi)}{2}^2 d\widetilde{\mu}(z) &\leq \frac{4\tau m^2 }{\xi^4}  \norm{\pi^{(i)} - \pi}{2}^2 \int  d p_{\pi^{(i)}}(x)    d\widetilde{\mu}(z) \\
	&=  \frac{4\tau m^2 }{\xi^4}  \norm{\pi^{(i)} - \pi}{2}^2\\
	&\to 0,
\end{align*}
since $\pi^{(i)} \to \pi$. \\ 

\uline{Finally, let's prove property (c)}. Note that when $d= 0$, $\Jcal^{\pi^{(i)},\mathbf{p},\tau}(z) - \Jcal^{\pi,\mathbf{p},\tau}(z) = 0$, so:
\begin{align*}
	\Jcal^{\pi^{(i)},\mathbf{p},\tau}(z) - \Jcal^{\pi,\mathbf{p},\tau}(z) &= d(\Jcal^{\pi^{(i)},\mathbf{p},\tau}(z) - \Jcal^{\pi,\mathbf{p},\tau}(z)) \\ 
	 &= d\tau \big( p^{\pi^{(i)} }(x) - p^{\pi }(x)\big) \\ 
	 &= d\tau \bigdotprod{(p_1(x) - p_0(x),\dots,p_m(x)-p_0(x)) }{ \pi^{(i)} - \pi }.
\end{align*}
So:
\begin{align*}
	\Jcal^{\pi^{(i)},\mathbf{p},\tau}(z)^{1/2} - \Jcal^{\pi,\mathbf{p},\tau}(z)^{1/2} &= \frac{\Jcal^{\pi^{(i)},\mathbf{p},\tau}(z) - \Jcal^{\pi,\mathbf{p}}(z)}{\Jcal^{\pi^{(i)},\mathbf{p},\tau}(z)^{1/2} + \Jcal^{\pi,\mathbf{p},\tau}(z)^{1/2}} \\
	&= \frac{ d\tau \bigdotprod{(p_1(x) - p_0(x),\dots,p_m(x)-p_0(x)) }{ \pi^{(i)} - \pi }   }{\Jcal^{\pi^{(i)},\mathbf{p},\tau}(z)^{1/2} + \Jcal^{\pi,\mathbf{p},\tau}(z)^{1/2}} \\
	&= \frac{ d\tau \bigdotprod{(p_1(x) - p_0(x),\dots,p_m(x)-p_0(x)) }{ \pi^{(i)} - \pi }   }{ \sqrt{\tau} p_{\pi^{(i)}}(x)^{1/2} + \sqrt{\tau} p_{\pi}(x)^{1/2} } \\
	&=  d\sqrt{\tau}  \frac{ \bigdotprod{(p_1(x) - p_0(x),\dots,p_m(x)-p_0(x)) }{ \pi^{(i)} - \pi }   }{  p_{\pi^{(i)}}(x)^{1/2} +  p_{\pi}(x)^{1/2} }.
\end{align*}
Also note that:
\begin{align*}
	 D(z,\pi)' (\pi^{(i)} - \pi) &= \frac{d}{2} \sqrt{\tau p_{\pi}(x)} \dotprod{s_{\pi}(x)}{\pi^{(i)} - \pi} \\
	  &= \frac{d\sqrt{\tau}}{2} \frac{1}{\sqrt{ p_{\pi}(x)}} \bigdotprod{ (p_1(x) - p_0(x),\dots,p_m(x)- p_0(x) )}{\pi^{(i)} - \pi} \\
\end{align*}

Ergo:
\[
	\Jcal^{\pi^{(i)},\mathbf{p},\tau}(z)^{1/2} - \Jcal^{\pi,\mathbf{p},\tau}(z)^{1/2} - D(z,\pi)' (\pi^{(i)} - \pi)  
\]
\[
	= d\sqrt{\tau} \bigg(  \frac{ 1 }{  p_{\pi^{(i)}}(x)^{1/2} +  p_{\pi}(x)^{1/2} } - \frac{1}{2 p_{\pi}(x)^{1/2} } \bigg) \bigdotprod{(p_1(x) - p_0(x),\dots,p_m(x)-p_0(x)) }{ \pi^{(i)} - \pi } 
\]
\[
	= d\sqrt{\tau} p_{\pi}(x) \bigg(  \frac{ 1 }{  p_{\pi^{(i)}}(x)^{1/2} +  p_{\pi}(x)^{1/2} } - \frac{1}{2 p_{\pi}(x)^{1/2} } \bigg) \bigdotprod{s_{\pi}(x) }{ \pi^{(i)} - \pi } 
\]
\[
	= d\sqrt{\tau p_{\pi}(x)} \bigg(  \frac{  p_{\pi}(x)^{1/2} }{  p_{\pi^{(i)}}(x)^{1/2} +  p_{\pi}(x)^{1/2} } - \frac{1  }{2 } \bigg) \bigdotprod{s_{\pi}(x) }{ \pi^{(i)} - \pi } 
\]
\[
	= d\sqrt{\tau p_{\pi}(x)} \bigg(  \frac{  p_{\pi}(x)^{1/2} -  p_{\pi^{(i)}}(x)^{1/2} }{  2(p_{\pi^{(i)}}(x)^{1/2} +  p_{\pi}(x)^{1/2})  }  \bigg) \bigdotprod{s_{\pi}(x) }{ \pi^{(i)} - \pi } 
\]
\[
	= d\sqrt{\tau p_{\pi}(x)} \bigg(  \frac{  p_{\pi}(x) -  p_{\pi^{(i)}}(x) }{  2(p_{\pi^{(i)}}(x)^{1/2} +  p_{\pi}(x)^{1/2})(p_{\pi}(x)^{1/2} +  p_{\pi^{(i)}}(x)^{1/2})   }  \bigg) \bigdotprod{s_{\pi}(x) }{ \pi^{(i)} - \pi }. 
\]
\[
	= \frac{d}{2}\sqrt{\tau p_{\pi}(x)}   \frac{  p_{\pi}(x) -  p_{\pi^{(i)}}(x) }{  (p_{\pi^{(i)}}(x)^{1/2} +  p_{\pi}(x)^{1/2})^2    }   \bigdotprod{s_{\pi}(x) }{ \pi^{(i)} - \pi } 
\]
Thus:
\begin{align*}
	&\big[ \Jcal^{\pi^{(i)},\mathbf{p},\tau}(z)^{1/2} - \Jcal^{\pi,\mathbf{p},\tau}(z)^{1/2} - D(z,\pi)' (\pi^{(i)} - \pi) \big]^2 \\
	&=  \frac{d}{4} \tau p_{\pi}(x)    \frac{  \big( p_{\pi}(x) -  p_{\pi^{(i)}}(x) \big)^2 }{  (p_{\pi^{(i)}}(x)^{1/2} +  p_{\pi}(x)^{1/2})^4    }   \bigdotprod{s_{\pi}(x) }{ \pi^{(i)} - \pi }^2 \\
	&\leq  \frac{d}{4} \tau p_{\pi}(x)    \frac{  \big( p_{\pi}(x) -  p_{\pi^{(i)}}(x) \big)^2 }{  (p_{\pi^{(i)}}(x)^{1/2} +  p_{\pi}(x)^{1/2})^4    }   \norm{s_{\pi}(x) }{2}^2 \norm{ \pi^{(i)} - \pi }{2}^2 \\
	&\leq  \frac{d}{4} \tau p_{\pi}(x)    \frac{  \big( p_{\pi}(x) -  p_{\pi^{(i)}}(x) \big)^2 }{  \big(  p_{\pi}(x)^{1/2} \big)^4    }   \norm{s_{\pi}(x) }{2}^2 \norm{ \pi^{(i)} - \pi }{2}^2 \\
	&=  \frac{d}{4} \tau p_{\pi}(x)  \bigg(  \frac{   p_{\pi}(x) -  p_{\pi^{(i)}}(x) }{   p_{\pi}(x)     } \bigg)^2  \norm{s_{\pi}(x) }{2}^2 \norm{ \pi^{(i)} - \pi }{2}^2 \\
	&=  \frac{d}{4} \tau p_{\pi}(x)   \bigdotprod{ s_{\pi}(x) }{ \pi^{(i)} - \pi }^2  \norm{s_{\pi}(x) }{2}^2 \norm{ \pi^{(i)} - \pi }{2}^2 \\
	&\leq  \frac{d}{4} \tau p_{\pi}(x)    \norm{s_{\pi}(x) }{2}^4 \norm{ \pi^{(i)} - \pi }{2}^4 \\
	&\leq  4\frac{\tau m^2}{\xi^4}  dp_{\pi}(x)    \norm{ \pi^{(i)} - \pi }{2}^4.
\end{align*}
Thus:
\begin{align*}
	&\int \big[ \Jcal^{\pi^{(i)},\mathbf{p},\tau}(z)^{1/2}  - \Jcal^{\pi,\mathbf{p},\tau}(z)^{1/2} - D(z,\pi)' (\pi^{(i)} - \pi) \big]^2  d\widetilde{\mu}(z) \Big/ \norm{\pi^{(i)} - \pi}{2}^2\\
	&\leq  4\frac{\tau m^2}{\xi^4}  \norm{ \pi^{(i)} - \pi }{2}^2 \int dp_{\pi}(x)  d\widetilde{\mu}(z) \\ 
	&=  4\frac{\tau m^2}{\xi^4}  \norm{ \pi^{(i)} - \pi }{2}^2 \\ 
	&\to 0,
\end{align*}
since $\pi^{(i)} \to \pi$. \\

By \citep{Newey1990}, having proved properties (a), (b) and (c), it follows that $\sqrt{\Jcal^{\pi,\mathbf{p},\tau}(z) }$ is $\widetilde{\mu}(z)$-mean square continuously differentiable with respect to $\pi\in \Delta$ with derivative $D(z,\pi) = \frac{d}{2} \sqrt{\tau p_{\pi}(x)} s_{\pi}(x)$. By \citep{Newey1990}, this means that the score for $\pi$ is $S_{\pi}(z) = 2\I{\Jcal^{\pi,\mathbf{p}}(z) > 0} D(z,\pi) / \sqrt{\Jcal^{\pi,\mathbf{p},\tau}(z) }$. That is,
\begin{align*}
	S_{\pi}(z) &= 2\I{\Jcal^{\pi,\mathbf{p},\tau}(z) > 0} D(z,\pi) / \sqrt{\Jcal^{\pi,\mathbf{p},\tau}(z) } \\ 
	&=  d \I{\Jcal^{\pi,\mathbf{p},\tau}(z) > 0}  \sqrt{\tau p_{\pi}(x)} s_{\pi}(x)  / \sqrt{\Jcal^{\pi,\mathbf{p},\tau}(z) } \\ 
	&=  d \I{ \tau p_{\pi}(x) > 0}  \sqrt{\tau p_{\pi}(x)} s_{\pi}(x)  / \sqrt{ \tau p_{\pi}(x) } \\ 
	&=  d  s_{\pi}(x) \I{ p_{\pi}(x) > 0},
\end{align*}
which is equal to $d s_{\pi}(x)$ a.e. $[\Jcal^{\pi,\mathbf{p},\tau}]$. Finally, it remains to show that $\partialDerivative{\pi} \Big\{ \log \Jcal^{\pi,\mathbf{p},\tau}(z) \Big\}$ is equal to  $d s_{\pi}(x)$ a.e. $[\Jcal^{\pi,\mathbf{p},\tau}]$. Observe that, on non-negligble sets, the indicator function in $\Jcal^{\pi,\mathbf{p},\tau}(z)$ is equal to 1. So, on these sets, it follows from elementary calculus that:
\begin{align*}
	\partialDerivative{\pi} \Big\{ \log \Jcal^{\pi,\mathbf{p},\tau}(z) \Big\} &= d \partialDerivative{\pi} \Big\{ \log \big( \tau p_{\pi}(x) \big) \Big\} \\
	&= d \partialDerivative{\pi} \Big\{ \log  p_{\pi}(x)  \Big\} \\
	&= d s_{\pi}(x),
\end{align*}
as desired!

\end{proof}
\vspace{0.4in}

\begin{proof}[\uline{Proof of Lemma \eqref{A Particular Smooth Parametric Submodel}}]

\uline{First, we need to argue that $\mathbf{M}^{\textup{sub},\tau}$ is a parametric submodel of $\mathbf{M}^{\textup{semi},\tau}$}. Towards that end, consider any density $\Jcal^{\pi,\rho,\tau} \in \mathbf{M}^{\textup{sub},\tau}$. Since $\mathbf{p}\in \Q$, $\gamma^\ast(\tau) \in \Gamma$ and $\frac{\pi_j^\ast}{\pi_j^\train} < \frac{1-\xi}{\xi}$ for each $j\in\Y$, it follows from Assumption \eqref{Fixed Tau Regime: expo family} that  
\[
	(p_0e^{\rho' \frac{\pi_0^\ast}{\pi_0^\train}s_{\gamma^\ast(\tau)} - c_0(\rho) },\dots, p_me^{ \rho'  \frac{\pi_m^\ast}{\pi_m^\train} s_{\gamma^\ast(\tau)}  - c_m(\rho)} ) \in \Q
\]
for all $\rho \in \mathbb{R}^m$ satisfying $\norm{\rho}{2} < B$. These exponentially tilted densities are precisely the class densities $(q_0^{\rho},\dots,q_m^{\rho})$ which appear in the definition for $\Jcal^{\pi,\rho,\tau}$, so it must be that $\Jcal^{\pi,\rho,\tau} \in \mathbf{M}^{\textup{semi},\tau}$. Thus, $\mathbf{M}^{\textup{sub},\tau} \subseteq \mathbf{M}^{\textup{semi},\tau}$. In addition, note that $0\in \R$ so $\Jcal^{\pi^\ast,0,\tau} \in \mathbf{M}^{\textup{sub},\tau}$, and that when $\rho = 0$,
\[
	(p_0e^{\rho' \frac{\pi_0^\ast}{\pi_0^\train}s_{\gamma^\ast(\tau)} - c_0(\rho) },\dots, p_me^{ \rho'  \frac{\pi_m^\ast}{\pi_m^\train} s_{\gamma^\ast(\tau)}  - c_m(\rho)} )  = \mathbf{p}.
\]
Thus,  $\Jcal^{\pi^\ast,0,\tau}= \Jcal^{\pi^\ast,\mathbf{p},\tau}$, and so $\Jcal^{\pi^\ast,\mathbf{p},\tau} \in \mathbf{M}^{\textup{sub},\tau}$. Thus, indeed, $\mathbf{M}^{\textup{sub},\tau}$ is a parametric submodel of $\mathbf{M}^{\textup{semi},\tau}$. 
 
\uline{Next, we need to prove that $\mathbf{M}^{\textup{sub},\tau}$  is \textit{smooth}} in the sense of Definition A.1 of \citep{Newey1990}. That is, we need to prove that $\Jcal^{\pi,\rho,\tau}$ is smooth at each $(\pi,\rho) \in \Theta^\text{sub}$. This involves verifying the three conditions in Definition A.1. Since the proofs for verifying conditions (i) and (ii) are the same as what was done in the proof of Lemma \eqref{Smooth Score Function for Pi}, here we focus on only checking (iii).

Condition (iii) requires us to show that $\sqrt{\Jcal^{\theta,\tau}(z)}  \equiv \sqrt{\Jcal^{\pi,\rho,\tau}(z)} $ is $\widetilde{\mu}$-mean square continuously differentiable with respect to $\theta\equiv (\pi,\rho) \in \Theta^\text{sub}$, with derivative $D(z,\theta)$ given by:
\[
	D(z,\theta) := \big[ D(z,\pi;\rho) ,\ D(z,\rho;\pi)  \big]' \in \mathbb{R}^{2m},
\]
where 
\[
	D(z,\pi;\rho) := \frac{d}{2} \sqrt{\tau q_{\pi}^{\rho}(x)} s_{\pi}^{\rho}(x)
\]
and 
\[
	D(z,\rho;\pi) := d\sqrt{\tau} \sum_{j=0}^m \pi_j \frac{\pi_j^\ast}{\pi_j^\train} \frac{ q_j^{\rho}(x) }{ 2\sqrt{q_{\pi}^\rho(x)} }  (s_{\gamma^\ast(\tau)}(x) - \E_j^{\rho}[s_{\gamma^\ast(\tau)}]) + (1-d)\sqrt{(1-\tau) \pi_y^{\train}}\frac{\pi_y^\ast}{\pi_y^\train} \frac{s_{\gamma^\ast(\tau)}(x) - \E_y^{\rho}[s_{\gamma^\ast(\tau)}]}{2}\sqrt{q_y^\rho(x)}. \\ 
\]
Showing this, in turn, requires us to verify three different properties. Namely, for every $\theta \in \Delta \times \R$ and sequence $\theta^{(i)} \to \theta$, we must show that:
\begin{enumerate}[label=(\alph*)]
	\item $\int \norm{D(z,\theta)}{2}^2 d\widetilde{\mu}(z) < \infty$.
	\item $\int \norm{D(z,\theta^{(i)}) - D(z,\theta)}{2}^2 d\widetilde{\mu}(z) \to 0$.
	\item $\int \big[ \Jcal^{\theta^{(i)},\tau}(z)^{1/2}  - \Jcal^{\theta,\tau}(z)^{1/2} - D(z,\theta)' (\theta^{(i)} - \theta) \big]^2  d\widetilde{\mu}(z) \Big/ \norm{\theta^{(i)} - \theta}{2}^2 \to 0 $. \\ 
\end{enumerate}

\uline{Let's begin by proving property (c)}. First, we will rewrite the $\Jcal^{\theta^{(i)},\tau}(z)^{1/2}  - \Jcal^{\theta,\tau}(z)^{1/2}$ term. Observe that:
\begin{align*}
	\Jcal^{\theta^{(i)},\tau}(z)^{1/2}  - \Jcal^{\theta,\tau}(z)^{1/2} &= \frac{\Jcal^{\theta^{(i)},\tau}(z) - \Jcal^{\theta,\tau}(z)}{\Jcal^{\theta^{(i)},\tau}(z)^{1/2}  + \Jcal^{\theta,\tau}(z)^{1/2} } \\ 
	&= d\frac{\Jcal^{\theta^{(i)},\tau}(z) - \Jcal^{\theta,\tau}(z)}{\Jcal^{\theta^{(i)},\tau}(z)^{1/2}  + \Jcal^{\theta,\tau}(z)^{1/2} }  + (1-d)\frac{\Jcal^{\theta^{(i)},\tau}(z) - \Jcal^{\theta,\tau}(z)}{\Jcal^{\theta^{(i)},\tau}(z)^{1/2}  + \Jcal^{\theta,\tau}(z)^{1/2} }  \\ 
	&= d\sqrt{\tau}\frac{    q_{\pi^{(i)}}^{\rho^{(i)}} (x)  - q_{\pi}^{\rho}(x)   }{ \sqrt{q_{\pi^{(i)}}^{\rho^{(i)}} (x)}  + \sqrt{q_{\pi}^{\rho}(x) }}  + (1-d)\sqrt{(1-\tau)\pi_y^\train}\frac{ q_y^{\rho^{(i)}}(x) - q_y^\rho(x) }{ \sqrt{ q_y^{\rho^{(i)}}(x) } + \sqrt{q_y^\rho(x)} }.
\end{align*}


Also observe that:
\begin{align*}
	q_{\pi^{(i)}}^{\rho^{(i)}} (x)  - q_{\pi}^{\rho}(x)  &=  \sum_{j=0}^m \pi_j^{(i)} q_j^{\rho^{(i)}}(x) - \sum_{j=0}^m \pi_j  q_j^{\rho}(x)  \\ 
	&=  \sum_{j=0}^m \pi_j^{(i)}  (q_j^{\rho^{(i)}}(x) - q_j^{\rho}(x) ) + \sum_{j=0}^m ( \pi_j^{(i)} - \pi_j)  q_j^{\rho}(x)  \\ 
\end{align*}
Note that $\pi_0^{(i)}-\pi_0 = \Big( 1- \sum_{j=1}^m \pi_j^{(i)} \Big) - \Big( 1- \sum_{j=1}^m \pi_j \Big) = \sum_{j=1}^m (\pi_j - \pi_j^{(i)}) $, so $(\pi_0^{(i)}-\pi_0)q_0^{\rho }(x) =   \sum_{j=1}^m (\pi_j - \pi_j^{(i)})q_0^{\rho}(x)$ and therefore
\begin{align*}
	\sum_{j=0}^m (\pi_j^{(i)}-\pi_j) q_j^{\rho}(x) &= \sum_{j=1}^m (\pi_j^{(i)}-\pi_j) q_j^{\rho }(x) + (\pi_0^{(i)}-\pi_0)q_0^{\rho }(x) \\
	&= \sum_{j=1}^m (\pi_j^{(i)}-\pi_j) (q_j^{\rho }(x)-q_0^{\rho }(x)) \\ 
	&= q_{\pi}^{\rho }(x) \sum_{j=1}^m (\pi_j^{(i)}-\pi_j) s_{\pi,j}^{\rho }(x) \\ 
	&=  q_{\pi}^{\rho }(x) \dotprod{ s_{\pi}^{\rho }(x) }{ \pi^{(i)}-\pi }.
\end{align*}
Therefore:
\begin{equation}\label{diff in tilted mixture}
	\implies q_{\pi^{(i)}}^{\rho^{(i)}} (x)  - q_{\pi}^{\rho}(x) =  \sum_{j=0}^m \pi_j^{(i)}  (q_j^{\rho^{(i)}}(x) - q_j^{\rho}(x) )   + q_{\pi}^{\rho }(x) \dotprod{ s_{\pi}^{\rho }(x) }{ \pi^{(i)}-\pi }.
\end{equation}
Ergo:
\small
\begin{align*}
	&\Jcal^{\theta^{(i)},\tau}(z)^{1/2}  - \Jcal^{\theta,\tau}(z)^{1/2} \\
	&=  d\sqrt{\tau}\frac{   \sum_{j=0}^m \pi_j^{(i)}  (q_j^{\rho^{(i)}}(x) - q_j^{\rho}(x) )   + q_{\pi}^{\rho }(x) \dotprod{ s_{\pi}^{\rho }(x) }{ \pi^{(i)}-\pi }    }{ \sqrt{q_{\pi^{(i)}}^{\rho^{(i)}} (x)}  + \sqrt{q_{\pi}^{\rho}(x) }}  + (1-d)\sqrt{(1-\tau)\pi_y^\train}\frac{ q_y^{\rho^{(i)}}(x) - q_y^\rho(x) }{ \sqrt{ q_y^{\rho^{(i)}}(x) } + \sqrt{q_y^\rho(x)} }.
\end{align*}
\normalsize

Thus, we have that:
\smaller
\begin{align*}
	&\Jcal^{\theta^{(i)},\tau}(z)^{1/2}  - \Jcal^{\theta,\tau}(z)^{1/2} - D(z,\theta)' (\theta^{(i)} - \theta) \\ &= \Jcal^{\theta^{(i)},\tau}(z)^{1/2}  - \Jcal^{\theta,\tau}(z)^{1/2} - D(z,\pi;\rho)'(\pi^{(i)} - \pi) - D(z,\rho;\pi)(\rho^{(i)} - \rho) \\ \\
	&= d\sqrt{\tau}\frac{   \sum_{j=0}^m \pi_j^{(i)}  (q_j^{\rho^{(i)}}(x) - q_j^{\rho}(x) )    }{ \sqrt{q_{\pi^{(i)}}^{\rho^{(i)}} (x)}  + \sqrt{q_{\pi}^{\rho}(x) }}  + (1-d)\sqrt{(1-\tau)\pi_y^\train}\frac{ q_y^{\rho^{(i)}}(x) - q_y^\rho(x) }{ \sqrt{ q_y^{\rho^{(i)}}(x) } + \sqrt{q_y^\rho(x)} }  \\
	&\qquad+ d\sqrt{\tau}\frac{  q_{\pi}^{\rho }(x) \dotprod{ s_{\pi}^{\rho }(x) }{ \pi^{(i)}-\pi }    }{ \sqrt{q_{\pi^{(i)}}^{\rho^{(i)}} (x)}  + \sqrt{q_{\pi}^{\rho}(x) }}  -  d\sqrt{\tau} \frac{\sqrt{ q_{\pi}^{\rho}(x)}}{2} \dotprod{  s_{\pi}^{\rho}(x) }{\pi^{(i)} - \pi} - D(z,\rho;\pi)(\rho^{(i)} - \rho) \\ \\ 
	&= \Bigg[ d\sqrt{\tau}\frac{   \sum_{j=0}^m \pi_j^{(i)}  (q_j^{\rho^{(i)}}(x) - q_j^{\rho}(x) )    }{ \sqrt{q_{\pi^{(i)}}^{\rho^{(i)}} (x)}  + \sqrt{q_{\pi}^{\rho}(x) }}  \\
	&\qquad\qquad+ (1-d)\sqrt{(1-\tau)\pi_y^\train}\frac{ q_y^{\rho^{(i)}}(x) - q_y^\rho(x) }{ \sqrt{ q_y^{\rho^{(i)}}(x) } + \sqrt{q_y^\rho(x)} }  -  D(z,\rho;\pi)(\rho^{(i)} - \rho) \Bigg] \\ 
	&\qquad+d\sqrt{\tau q_{\pi}^{\rho}(x) }\Bigg[  \frac{  \sqrt{q_{\pi}^{\rho}(x)}   }{ \sqrt{q_{\pi^{(i)}}^{\rho^{(i)}} (x)}  + \sqrt{q_{\pi}^{\rho}(x) }}   - \frac{1}{2} \Bigg]  \dotprod{ s_{\pi}^{\rho }(x) }{ \pi^{(i)}-\pi }  \\ \\ 
	&= d\sqrt{\tau}\overbrace{\Bigg[ \sum_{j=0}^m \pi_j^{(i)}  \frac{    q_j^{\rho^{(i)}}(x) - q_j^{\rho}(x)   }{ \sqrt{q_{\pi^{(i)}}^{\rho^{(i)}} (x)}  + \sqrt{q_{\pi}^{\rho}(x) }}   -   \sum_{j=0}^m \pi_j \frac{\pi_j^\ast}{\pi_j^\train} \frac{ q_j^{\rho}(x) }{ 2\sqrt{q_{\pi}^\rho(x)} } \dotprod{ s_{\gamma^\ast(\tau)}(x) - \E_j^{\rho}[s_{\gamma^\ast(\tau)}]}{\rho^{(i)} - \rho}    \Bigg]}^{=: \ K_1(z,\theta,\theta^{(i)}) } \\ 
	&\qquad+ (1-d)\sqrt{(1-\tau)\pi_y^\train}\overbrace{\Bigg[\frac{ q_y^{\rho^{(i)}}(x) - q_y^\rho(x) }{ \sqrt{ q_y^{\rho^{(i)}}(x) } + \sqrt{q_y^\rho(x)} }  - \frac{\pi_j^\ast}{\pi_j^\train}\frac{ \dotprod{s_{\gamma^\ast(\tau)}(x) - \E_y^{\rho}[s_{\gamma^\ast(\tau)}]}{\rho^{(i)} -\rho} }{2}\sqrt{q_y^\rho(x)}  \Bigg]}^{ =: \ K_2(z,\theta,\theta^{(i)}) }\\
	&\qquad+ d\sqrt{\tau q_{\pi}^{\rho}(x) }\overbrace{\Bigg[  \frac{  \sqrt{q_{\pi}^{\rho}(x)}   }{ \sqrt{q_{\pi^{(i)}}^{\rho^{(i)}} (x)}  + \sqrt{q_{\pi}^{\rho}(x) }}   - \frac{1}{2} \Bigg]  \dotprod{ s_{\pi}^{\rho }(x) }{ \pi^{(i)}-\pi }}^{=: \ K_3(z,\theta,\theta^{(i)}) }.
\end{align*}
\normalsize

Hence,
\small
\begin{align*}
	&\int \frac{\big[ \Jcal^{\theta^{(i)},\tau}(z)^{1/2}  - \Jcal^{\theta,\tau}(z)^{1/2} - D(z,\theta)' (\theta^{(i)} - \theta) \big]^2}{ \norm{\theta^{(i)} - \theta}{2}^2 }  d\widetilde{\mu}(z) \\
	&\leq 3\int \Big[ d\sqrt{\tau} K_1(z,\theta,\theta^{(i)})\Big]^2 d\widetilde{\mu}(z) \Big/ \norm{\theta^{(i)} - \theta}{2}^2   \\
	&\qquad +    3\int \Big[   (1-d)\sqrt{(1-\tau)\pi_y^\train} K_2(z,\theta,\theta^{(i)})  \Big]^2 d\widetilde{\mu}(z) \Big/ \norm{\theta^{(i)} - \theta}{2}^2 \\ 
	&\qquad	+   3\int \Big[  d\sqrt{\tau q_{\pi}^{\rho}(x) } K_3(z,\theta,\theta^{(i)})   \Big]^2 d\widetilde{\mu}(z) \Big/ \norm{\theta^{(i)} - \theta}{2}^2 \\ \\ 
	&\leq 3 \int \Big[ d K_1(z,\theta,\theta^{(i)})\Big]^2 d\widetilde{\mu}(z) \Big/ \norm{\theta^{(i)} - \theta}{2}^2   \\
	&\qquad +    3\int \Big[   (1-d) K_2(z,\theta,\theta^{(i)})  \Big]^2 d\widetilde{\mu}(z) \Big/ \norm{\theta^{(i)} - \theta}{2}^2 \\ 
	&\qquad	+   3 \int \Big[  d\sqrt{ q_{\pi}^{\rho}(x) } K_3(z,\theta,\theta^{(i)})   \Big]^2 d\widetilde{\mu}(z) \Big/ \norm{\theta^{(i)} - \theta}{2}^2.
\end{align*}
\normalsize

Now, for each $j \in \Y$, let $f_j(\rho;x) := e^{\rho'\frac{\pi_j^\ast}{\pi_j^\train}s_{\gamma^\ast(\tau)}(x) - c_j(\rho)}$. Then by the mean value theorem, for some point $\bar{\rho}_{x,j}$ which depends on $x$ and is a convex combination of $\rho^{(i)}$ and $\rho$, we have that:
\begin{align}
	&q_j^{\rho^{(i)}}(x) - q_j^{\rho}(x) \\
	&= p_j(x)\Big[f_j( \rho^{(i)};x) - f_j( \rho;x) \Big] \nonumber \\ 
	&= p_j(x)\Bigg[ \Bigdotprod{\partialDerivative{\rho} f_j(\rho;x)   }{\rho^{(i)} - \rho} + \frac{1}{2}(\rho^{(i)} - \rho)' \partialSecondDerivative{\rho}f_j(\bar{\rho}_{x,j};x)(\rho^{(i)} - \rho) \Bigg] \nonumber \\
	&= p_j(x)\Bigg[ e^{\rho'\frac{\pi_j^\ast}{\pi_j^\train}s_{\gamma^\ast(\tau)}(x) - c_j(\rho)} \frac{\pi_j^\ast}{\pi_j^\train}\Bigdotprod{ s_{\gamma^\ast(\tau)}(x)- \E_{j}^{\rho}[s_{\gamma^\ast(\tau)}]  }{\rho^{(i)} - \rho} + \frac{1}{2}(\rho^{(i)} - \rho)' \partialSecondDerivative{\rho}f_j(\bar{\rho}_{x,j};x)(\rho^{(i)} - \rho) \Bigg] \nonumber \\
	&=q_j^{\rho}(x) \frac{\pi_j^\ast}{\pi_j^\train}\Bigdotprod{ s_{\gamma^\ast(\tau)}(x)- \E_{j}^{\rho}[s_{\gamma^\ast(\tau)}]  }{\rho^{(i)} - \rho} + \frac{p_j(x)}{2}(\rho^{(i)} - \rho)' \partialSecondDerivative{\rho}f_j(\bar{\rho}_{x,j};x)(\rho^{(i)} - \rho). \label{diff in tilted component}
\end{align}

Therefore:
\small
\begin{align*}
	&\abs{K_2(z,\theta,\theta^{(i)})} \\
	 &= \absBigg{\frac{ q_y^{\rho^{(i)}}(x) - q_y^\rho(x) }{ \sqrt{ q_y^{\rho^{(i)}}(x) } + \sqrt{q_y^\rho(x)} }  - \frac{\pi_j^\ast}{\pi_j^\train}\frac{ \dotprod{s_{\gamma^\ast(\tau)}(x) - \E_y^{\rho}[s_{\gamma^\ast(\tau)}]}{\rho^{(i)} -\rho} }{2}\sqrt{q_y^\rho(x)}}  \\ \\
	&= \absBigg{\frac{ q_y^{\rho}(x) \frac{\pi_j^\ast}{\pi_j^\train} \bigdotprod{ s_{\gamma^\ast(\tau)}(x)- \E_{y}^{\rho}[s_{\gamma^\ast(\tau)}]  }{\rho^{(i)} - \rho} + \frac{p_y(x)}{2}(\rho^{(i)} - \rho)' \partialSecondDerivative{\rho}f_y(\bar{\rho}_{x,y};x)(\rho^{(i)} - \rho)   }{ \sqrt{ q_y^{\rho^{(i)}}(x) } + \sqrt{q_y^\rho(x)} } \\
	&\qquad  - \frac{\pi_j^\ast}{\pi_j^\train} \frac{ \dotprod{s_{\gamma^\ast(\tau)}(x) - \E_y^{\rho}[s_{\gamma^\ast(\tau)}]}{\rho^{(i)} -\rho} }{2}\sqrt{q_y^\rho(x)} } \\ \\
	&= \absBigg{ \sqrt{q_y^\rho(x)} \Bigg( \frac{ \sqrt{q_y^{\rho}(x)}  }{ \sqrt{ q_y^{\rho^{(i)}}(x) } + \sqrt{q_y^\rho(x)} }  - \frac{1}{2}\Bigg)  \frac{\pi_j^\ast}{\pi_j^\train} \dotprod{s_{\gamma^\ast(\tau)}(x) - \E_y^{\rho}[s_{\gamma^\ast(\tau)}]}{\rho^{(i)} -\rho}   + \frac{\frac{p_y(x)}{2}(\rho^{(i)} - \rho)' \partialSecondDerivative{\rho}f_y(\bar{\rho}_{x,y};x)(\rho^{(i)} - \rho) }{\sqrt{ q_y^{\rho^{(i)}}(x) } + \sqrt{q_y^\rho(x)}}} \\ 
	&\leq  \sqrt{q_y^\rho(x)} \absBigg{ \frac{ \sqrt{q_y^{\rho}(x)}  }{ \sqrt{ q_y^{\rho^{(i)}}(x) } + \sqrt{q_y^\rho(x)} }  - \frac{1}{2}} \frac{\pi_j^\ast}{\pi_j^\train} \norm{ s_{\gamma^\ast(\tau)}(x) - \E_y^{\rho}[s_{\gamma^\ast(\tau)}]}{2} \norm{\rho^{(i)} -\rho}{2}   + \frac{p_y(x)}{2} \frac{ \maxEval{ \partialSecondDerivative{\rho}f_y(\bar{\rho}_{x,y};x)} \norm{\rho^{(i)} - \rho}{2}^2 }{\sqrt{ q_y^{\rho^{(i)}}(x) } + \sqrt{q_y^\rho(x)}}  \\
	&\leq \frac{4\sqrt{m}(1-\xi)}{L\xi}  \sqrt{q_y^\rho(x)} \absBigg{ \frac{ \sqrt{q_y^{\rho}(x)}  }{ \sqrt{ q_y^{\rho^{(i)}}(x) } + \sqrt{q_y^\rho(x)} }  - \frac{1}{2}} \norm{\rho^{(i)} -\rho}{2}   + \frac{\sqrt{p_y(x)} }{2} \frac{ \maxEval{ \partialSecondDerivative{\rho}f_y(\bar{\rho}_{x,y};x)}  }{\sqrt{ f_y(\rho^{(i)};x )} + \sqrt{f_y(\rho;x)}} \norm{\rho^{(i)} - \rho}{2}^2.
\end{align*}
\normalsize
And so:
\small
\begin{align*}
	  &\int (1-d)\frac{ K_2(z,\theta,\theta^{(i)}) ^2}{ \norm{\theta^{(i)} - \theta}{2}^2} d\widetilde{\mu}(z)  \\
	  &\leq 2\frac{16m(1-\xi)^2}{L^2\xi^2} \frac{\norm{\rho^{(i)} -\rho}{2}^2}{ \norm{\theta^{(i)} - \theta}{2}^2}  \int (1-d) q_y^\rho(x)\absBigg{ \frac{ \sqrt{q_y^{\rho}(x)}  }{ \sqrt{ q_y^{\rho^{(i)}}(x) } + \sqrt{q_y^\rho(x)} }  - \frac{1}{2}}^2  d\widetilde{\mu}(z)  \\ 
	  &\qquad\qquad + 2 \frac{1}{4} \frac{\norm{\rho^{(i)} - \rho}{2}^4}{\norm{\theta^{(i)} - \theta}{2}^2} \int (1-d)p_y(x) \Bigg( \frac{ \maxEval{ \partialSecondDerivative{\rho}f_y(\bar{\rho}_{x,y};x)}  }{\sqrt{ f_y(\rho^{(i)};x )} + \sqrt{f_y(\rho;x)}} \Bigg)^2  d\widetilde{\mu}(z)  \\ \\ 
	  &\leq \frac{32m(1-\xi)^2}{L^2\xi^2}   \E_{y}^{\rho } \Bigg[\bigg( \frac{ \sqrt{q_y^{\rho}(X)}  }{ \sqrt{ q_y^{\rho^{(i)}}(X) } + \sqrt{q_y^\rho(X)} }  - \frac{1}{2} \bigg)^2 \Bigg] \\ 
	  &\qquad\qquad + \frac{1}{2}\norm{\rho^{(i)} - \rho}{2}^2 \E_y \Bigg[ \bigg( \frac{ \maxEval{ \partialSecondDerivative{\rho}f_y(\bar{\rho}_{x,y};x)}  }{\sqrt{ f_y(\rho^{(i)};x )} + \sqrt{f_y(\rho;x)}} \bigg)^2  \Bigg] \\ \\ 
	  &= \frac{32m(1-\xi)^2}{L^2 \xi^2}    \E_{y}^{\rho } \Bigg[\bigg(  \frac{q_y^{\rho}(X) - q_y^{\rho^{(i)}}(X)}{ 2(q_y^{\rho^{(i)}}(X)^{1/2} + q_y^{\rho}(X)^{1/2})^2 } \bigg)^2 \Bigg] + \frac{\norm{\rho^{(i)} - \rho}{2}^2}{2} \E_y \Bigg[ \bigg( \frac{ \maxEval{ \partialSecondDerivative{\rho}f_y(\bar{\rho}_{x,y};x)}  }{\sqrt{ f_y(\rho^{(i)};x )} + \sqrt{f_y(\rho;x)}} \bigg)^2  \Bigg] \\ \\ 
	  &\leq \frac{8m(1-\xi)^2}{L^2\xi^2}   \E_{y}^{\rho } \Bigg[\bigg(  \frac{q_y^{\rho}(X) - q_y^{\rho^{(i)}}(X)}{ q_y^{\rho}(X)  } \bigg)^2 \Bigg]  + \frac{\norm{\rho^{(i)} - \rho}{2}^2}{2} \E_y \Bigg[ \bigg( \frac{ \maxEval{ \partialSecondDerivative{\rho}f_y(\bar{\rho}_{x,y};x)}  }{\sqrt{ f_y(\rho^{(i)};x )} + \sqrt{f_y(\rho;x)}} \bigg)^2  \Bigg].
\end{align*}
\normalsize
Now, by line \eqref{diff in tilted component}, note that:
\begin{align}
	&\bigg(\frac{q_y^{\rho^{(i)}}(x) - q_y^{\rho}(x)}{q_y^{\rho}(x)} \bigg)^2 \\
	&= \bigg( \frac{\pi_y^\ast}{\pi_y^\train} \bigdotprod{ s_{\gamma^\ast(\tau)}(x)- \E_{y}^{\rho}[s_{\gamma^\ast(\tau)}]  }{\rho^{(i)} - \rho} + \frac{1}{2 f_y(\rho; x)}(\rho^{(i)} - \rho)' \partialSecondDerivative{\rho}f_y(\bar{\rho}_{x,y};x)(\rho^{(i)} - \rho) \bigg)^2 \nonumber \\ 
	&\leq 2\frac{(\pi_y^\ast)^2}{(\pi_y^\train)^2}\bigdotprod{ s_{\gamma^\ast(\tau)}(x)- \E_{y}^{\rho}[s_{\gamma^\ast(\tau)}]  }{\rho^{(i)} - \rho}^2 + 2 \bigg(\frac{1}{2 f_y(\rho; x)}(\rho^{(i)} - \rho)' \partialSecondDerivative{\rho}f_y(\bar{\rho}_{x,y};x)(\rho^{(i)} - \rho) \bigg)^2 \nonumber \\ 
	&\leq 2 \frac{16m(1-\xi)^2}{L^2\xi^2}  \norm{\rho^{(i)} - \rho}{2}^2 + 2 \bigg(\frac{1}{2 f_y(\rho; x)}  \maxEvalBigg{\partialSecondDerivative{\rho}f_y(\bar{\rho}_{x,y};x)}\norm{\rho^{(i)} - \rho}{2}^2  \bigg)^2 \nonumber \\
	&= \frac{32m(1-\xi)^2}{L^2 \xi^2}  \norm{\rho^{(i)} - \rho}{2}^2 + \frac{1}{2 f_y(\rho; x)^2 } \maxEvalBigg{\partialSecondDerivative{\rho}f_y(\bar{\rho}_{x,y};x)} ^2 \norm{\rho^{(i)} - \rho}{2}^4. \label{purple castle}
\end{align}
Thus,
\begin{align*}
	 &\int (1-d)\frac{ K_2(z,\theta,\theta^{(i)}) ^2}{ \norm{\theta^{(i)} - \theta}{2}^2} d\widetilde{\mu}(z) \\
	 &\leq     \frac{256m^2(1-\xi)^4}{L^4\xi^4}  \norm{\rho^{(i)} - \rho}{2}^2     +  \frac{4m}{L^2} \E_{y}^{\rho} \Bigg[ \frac{1}{ f_y(\rho; x)^2 } \maxEvalBigg{\partialSecondDerivative{\rho}f_y(\bar{\rho}_{x,y};x)}^2 \Bigg] \norm{\rho^{(i)} - \rho}{2}^4  \\
	 &\qquad\qquad + \frac{\norm{\rho^{(i)} - \rho}{2}^2}{2} \E_y \Bigg[ \bigg( \frac{ \maxEval{ \partialSecondDerivative{\rho}f_y(\bar{\rho}_{x,y};x)}  }{\sqrt{ f_y(\rho^{(i)};x )} + \sqrt{f_y(\rho;x)}} \bigg)^2  \Bigg].
\end{align*}

One can show that $f_y(\rho^{(i)};x)$ is uniformly bounded away from $0$ and $\infty$, and that $\maxEval{\partialSecondDerivative{\rho}f_y(\bar{\rho}_{x,y};x)}$ is uniformly bounded away from $\infty$. This implies that both expectations in the display above are finite. Since $\theta^{(i)} \to \theta$, the integral on the LHS therefore converges to $0$.

Next, let's tackle the integral involving $K_1$. Towards that end, observe that:
\smaller
\begin{align*}
	&\abs{K_1(z,\theta,\theta^{(i)})} \\
	&= \absBigg{\sum_{j=0}^m \pi_j^{(i)}  \frac{    q_j^{\rho^{(i)}}(x) - q_j^{\rho}(x)   }{ \sqrt{q_{\pi^{(i)}}^{\rho^{(i)}} (x)}  + \sqrt{q_{\pi}^{\rho}(x) }}   -   \sum_{j=0}^m \pi_j \frac{ q_j^{\rho}(x) }{ 2\sqrt{q_{\pi}^\rho(x)} }  \frac{\pi_j^\ast}{\pi_j^\train}\dotprod{ s_{\gamma^\ast(\tau)}(x) - \E_j^{\rho}[s_{\gamma^\ast(\tau)}]}{\rho^{(i)} - \rho} } \\ \\ 
	&= \Bigg\lvert \sum_{j=0}^m \pi_j^{(i)}  \frac{  q_j^{\rho}(x)  \frac{\pi_j^\ast}{\pi_j^\train} \bigdotprod{ s_{\gamma^\ast(\tau)}(x)- \E_{j}^{\rho}[s_{\gamma^\ast(\tau)}]  }{\rho^{(i)} - \rho} + \frac{p_j(x)}{2}(\rho^{(i)} - \rho)' \partialSecondDerivative{\rho}f_j(\bar{\rho}_{x,j};x)(\rho^{(i)} - \rho)   }{ \sqrt{q_{\pi^{(i)}}^{\rho^{(i)}} (x)}  + \sqrt{q_{\pi}^{\rho}(x) }}  \\
	&\qquad\qquad\qquad\qquad -   \sum_{j=0}^m \pi_j \frac{ q_j^{\rho}(x) }{ 2\sqrt{q_{\pi}^\rho(x)} }  \frac{\pi_j^\ast}{\pi_j^\train} \dotprod{ s_{\gamma^\ast(\tau)}(x) - \E_j^{\rho}[s_{\gamma^\ast(\tau)}]}{\rho^{(i)} - \rho} \Bigg\lvert  \\ \\ 
	&\leq \absBigg{ \sum_{j=0}^m \pi_j^{(i)}  \frac{  q_j^{\rho}(x)  \frac{\pi_j^\ast}{\pi_j^\train} \bigdotprod{ s_{\gamma^\ast(\tau)}(x)- \E_{j}^{\rho}[s_{\gamma^\ast(\tau)}]  }{\rho^{(i)} - \rho}   }{ \sqrt{q_{\pi^{(i)}}^{\rho^{(i)}} (x)}  + \sqrt{q_{\pi}^{\rho}(x) }}   -   \sum_{j=0}^m \pi_j \frac{ q_j^{\rho}(x) }{ 2\sqrt{q_{\pi}^\rho(x)} }  \frac{\pi_j^\ast}{\pi_j^\train} \dotprod{ s_{\gamma^\ast(\tau)}(x) - \E_j^{\rho}[s_{\gamma^\ast(\tau)}]}{\rho^{(i)} - \rho} } \\
	&\qquad\qquad\qquad\qquad + \absBigg{\sum_{j=0}^m \pi_j^{(i)}  \frac{ \frac{p_j(x)}{2}(\rho^{(i)} - \rho)' \partialSecondDerivative{\rho}f_j(\bar{\rho}_{x,j};x)(\rho^{(i)} - \rho)  }{ \sqrt{q_{\pi^{(i)}}^{\rho^{(i)}} (x)}  + \sqrt{q_{\pi}^{\rho}(x) }} } \\ \\ 
	&=  \absBigg{\sum_{j=0}^m \Bigg( \pi_j^{(i)}  \frac{  q_j^{\rho}(x)  }{ \sqrt{q_{\pi^{(i)}}^{\rho^{(i)}} (x)}  + \sqrt{q_{\pi}^{\rho}(x) }} -  \pi_j \frac{ q_j^{\rho}(x) }{ 2\sqrt{q_{\pi}^\rho(x)} } \Bigg)  \frac{\pi_j^\ast}{\pi_j^\train} \dotprod{ s_{\gamma^\ast(\tau)}(x) - \E_j^{\rho}[s_{\gamma^\ast(\tau)}]}{\rho^{(i)} - \rho}}   \\
	&\qquad\qquad\qquad\qquad+ \absBigg{\sum_{j=0}^m \pi_j^{(i)}  \frac{ \frac{p_j(x)}{2}(\rho^{(i)} - \rho)' \partialSecondDerivative{\rho}f_j(\bar{\rho}_{x,j};x)(\rho^{(i)} - \rho)  }{ \sqrt{q_{\pi^{(i)}}^{\rho^{(i)}} (x)}  + \sqrt{q_{\pi}^{\rho}(x) }} } \\ \\ 
	&\leq \frac{4\sqrt{m}(1-\xi)}{L\xi} \norm{\rho^{(i)} - \rho}{2} \sum_{j=0}^m \absBigg{ \pi_j^{(i)}  \frac{  q_j^{\rho}(x)  }{ \sqrt{q_{\pi^{(i)}}^{\rho^{(i)}} (x)}  + \sqrt{q_{\pi}^{\rho}(x) }} -  \pi_j \frac{ q_j^{\rho}(x) }{ 2\sqrt{q_{\pi}^\rho(x)} } }   + \norm{\rho^{(i)} -\rho}{2}^2 \sum_{j=0}^m \pi_j^{(i)}  \frac{ \frac{1}{2}p_j(x)  \maxEval{\partialSecondDerivative{\rho}f_j(\bar{\rho}_{x,j};x)}  }{ \sqrt{q_{\pi^{(i)}}^{\rho^{(i)}} (x)}  + \sqrt{q_{\pi}^{\rho}(x) }}  \\ 
	&\leq \frac{4\sqrt{m}(1-\xi)}{L\xi} \norm{\rho^{(i)} - \rho}{2} \sum_{j=0}^m \absBigg{ \pi_j^{(i)}  \frac{  q_j^{\rho}(x)  }{ \sqrt{q_{\pi^{(i)}}^{\rho^{(i)}} (x)}  + \sqrt{q_{\pi}^{\rho}(x) }} -  \pi_j \frac{ q_j^{\rho}(x) }{ 2\sqrt{q_{\pi}^\rho(x)} } }   + \norm{\rho^{(i)} -\rho}{2}^2 \sum_{j=0}^m \frac{\pi_j^{(i)}}{\sqrt{\pi_j^{(i)} \wedge \pi_j} }  \frac{ \frac{1}{2}p_j(x)\maxEval{\partialSecondDerivative{\rho}f_j(\bar{\rho}_{x,j};x)}    }{ \sqrt{q_{j}^{\rho^{(i)}} (x)}  + \sqrt{q_{j}^{\rho}(x) }}  \\ 
	&\leq \frac{4\sqrt{m}(1-\xi)}{L\xi} \norm{\rho^{(i)} - \rho}{2} \sum_{j=0}^m \absBigg{ \pi_j^{(i)}  \frac{  q_j^{\rho}(x)  }{ \sqrt{q_{\pi^{(i)}}^{\rho^{(i)}} (x)}  + \sqrt{q_{\pi}^{\rho}(x) }} -  \pi_j \frac{ q_j^{\rho}(x) }{ 2\sqrt{q_{\pi}^\rho(x)} } }   + \norm{\rho^{(i)} -\rho}{2}^2 \frac{1-\xi}{\sqrt{\xi}} \sum_{j=0}^m  \frac{ \frac{1}{2}\sqrt{p_j(x)}\maxEval{\partialSecondDerivative{\rho}f_j(\bar{\rho}_{x,j};x)}    }{ \sqrt{  f_j(\rho^{(i)};x) }  + \sqrt{ f_j(\rho;x) }}.
\end{align*}
\normalsize

And so:
\small
\begin{align*}
	&\int  d \frac{ K_1(z,\theta,\theta^{(i)})^2}{\norm{\theta^{(i)} - \theta}{2}^2  } d\widetilde{\mu}(z) \\
	 &\leq 2 \frac{16m^2(1-\xi)^2}{L^2\xi^2} \frac{ \norm{\rho^{(i)} - \rho}{2}^2}{\norm{\theta^{(i)} - \theta}{2}^2 } \int  d\sum_{j=0}^m \absBigg{ \frac{   \pi_j^{(i)} q_j^{\rho}(x)  }{ \sqrt{q_{\pi^{(i)}}^{\rho^{(i)}} (x)}  + \sqrt{q_{\pi}^{\rho}(x) }} -   \frac{ \pi_j q_j^{\rho}(x) }{ 2\sqrt{q_{\pi}^\rho(x)} } }^2 d\widetilde{\mu}(z) \\
	&\qquad + 2 \frac{(1-\xi)^2}{\xi} \frac{\norm{\rho^{(i)} -\rho}{2}^4}{\norm{\theta^{(i)} - \theta}{2}^2 }m \int d   \sum_{j=0}^m  \Bigg( \frac{ \frac{1}{2}\sqrt{p_j(x)}\maxEval{\partialSecondDerivative{\rho}f_j(\bar{\rho}_{x,j};x)}    }{ \sqrt{  f_j(\rho^{(i)};x) }  + \sqrt{ f_j(\rho;x) }} \Bigg)^2d\widetilde{\mu}(z) \\ \\ 
	&\leq  \frac{32m^2(1-\xi)^2}{L^2\xi^2} \int  d\sum_{j=0}^m q_j^{\rho}(x) \absBigg{ \frac{   \pi_j^{(i)} \sqrt{q_j^{\rho}(x)}  }{ \sqrt{q_{\pi^{(i)}}^{\rho^{(i)}} (x)}  + \sqrt{q_{\pi}^{\rho}(x) }} -   \frac{ \pi_j \sqrt{q_j^{\rho}(x)} }{ 2\sqrt{q_{\pi}^\rho(x)} } }^2 d\widetilde{\mu}(z) \\
	&\qquad + \frac{1}{2}m \frac{(1-\xi)^2}{\xi} \norm{\rho^{(i)} -\rho}{2}^2  \int d   \sum_{j=0}^m p_j(x) \Bigg( \frac{ \maxEval{\partialSecondDerivative{\rho}f_j(\bar{\rho}_{x,j};x)}    }{ \sqrt{  f_j(\rho^{(i)};x) }  + \sqrt{ f_j(\rho;x) }} \Bigg)^2d\widetilde{\mu}(z) \\ \\ 
	&=  \frac{32m^2(1-\xi)^2}{L^2\xi^2} \sum_{j=0}^m \E_j^{\rho} \Bigg[ \Bigg( \frac{   \pi_j^{(i)} \sqrt{q_j^{\rho}(X)}  }{ \sqrt{q_{\pi^{(i)}}^{\rho^{(i)}} (X)}  + \sqrt{q_{\pi}^{\rho}(X) }} -   \frac{ \pi_j \sqrt{q_j^{\rho}(X)} }{ 2\sqrt{q_{\pi}^\rho(X)} } \Bigg)^2 \Bigg] \\
	&\qquad + \frac{1}{2}m \frac{(1-\xi)^2}{\xi} \norm{\rho^{(i)} -\rho}{2}^2   \sum_{j=0}^m \E_j \Bigg[ \Bigg( \frac{ \maxEval{\partialSecondDerivative{\rho}f_j(\bar{\rho}_{X,j};X)}    }{ \sqrt{  f_j(\rho^{(i)};X) }  + \sqrt{ f_j(\rho;X) }} \Bigg)^2 \Bigg] \\ \\ 
	&\leq  \frac{64m^2(1-\xi)^2}{L^2 \xi^2} \sum_{j=0}^m \E_j^{\rho} \Bigg[ \Bigg( \frac{   \pi_j^{(i)} \sqrt{q_j^{\rho}(X)}  }{ \sqrt{q_{\pi^{(i)}}^{\rho^{(i)}} (X)}  + \sqrt{q_{\pi}^{\rho}(X) }} -  \frac{   \pi_j \sqrt{q_j^{\rho}(X)}  }{ \sqrt{q_{\pi^{(i)}}^{\rho^{(i)}} (X)}  + \sqrt{q_{\pi}^{\rho}(X) }}  \Bigg)^2\Bigg] \\ 
	&\qquad +  \frac{64m^2(1-\xi)^2}{L^2\xi^2} \sum_{j=0}^m \E_j^{\rho} \Bigg[ \Bigg(\frac{   \pi_j \sqrt{q_j^{\rho}(X)}  }{ \sqrt{q_{\pi^{(i)}}^{\rho^{(i)}} (X)}  + \sqrt{q_{\pi}^{\rho}(X) }} -   \frac{ \pi_j \sqrt{q_j^{\rho}(X)} }{ 2\sqrt{q_{\pi}^\rho(X)} } \Bigg)^2 \Bigg] \\
	&\qquad + \frac{1}{2}m \frac{(1-\xi)^2}{\xi} \norm{\rho^{(i)} -\rho}{2}^2   \sum_{j=0}^m \E_j \Bigg[ \Bigg( \frac{ \maxEval{\partialSecondDerivative{\rho}f_j(\bar{\rho}_{X,j};X)}    }{ \sqrt{  f_j(\rho^{(i)};X) }  + \sqrt{ f_j(\rho;X) }} \Bigg)^2 \Bigg] \\ \\ 
	&\leq  \frac{64m^2(1-\xi)^2}{L^2\xi^3} \sum_{j=0}^m ( \pi_j^{(i)} - \pi_j)^2  +  \frac{64m^2(1-\xi)^2}{L^2\xi^2} \sum_{j=0}^m \E_j^{\rho} \Bigg[ \frac{q_j^{\rho}(X)}{q_{\pi}^{\rho}(X) } \Bigg(\frac{  \sqrt{q_{\pi}^\rho(X)}  }{ \sqrt{q_{\pi^{(i)}}^{\rho^{(i)}} (X)}  + \sqrt{q_{\pi}^{\rho}(X) }} -   \frac{  1 }{ 2  } \Bigg)^2 \Bigg] \\
	&\qquad + \frac{1}{2}m \frac{(1-\xi)^2}{\xi} \norm{\rho^{(i)} -\rho}{2}^2   \sum_{j=0}^m \E_j \Bigg[ \Bigg( \frac{ \maxEval{\partialSecondDerivative{\rho}f_j(\bar{\rho}_{X,j};X)}    }{ \sqrt{  f_j(\rho^{(i)};X) }  + \sqrt{ f_j(\rho;X) }} \Bigg)^2 \Bigg] \\ \\ 
	&\leq  \frac{64m^2(1-\xi)^2}{L^2\xi^3} \norm{\pi^{(i)} - \pi}{2}^2  +  \frac{64m^2(1-\xi)^2}{L^2\xi^3} \sum_{j=0}^m \E_j^{\rho} \Bigg[ \Bigg(\frac{  \sqrt{q_{\pi}^\rho(X)}  }{ \sqrt{q_{\pi^{(i)}}^{\rho^{(i)}} (X)}  + \sqrt{q_{\pi}^{\rho}(X) }} -   \frac{  1 }{ 2  } \Bigg)^2 \Bigg] \\
	&\qquad + \frac{1}{2}m \frac{(1-\xi)^2}{\xi} \norm{\rho^{(i)} -\rho}{2}^2   \sum_{j=0}^m \E_j \Bigg[ \Bigg( \frac{ \maxEval{\partialSecondDerivative{\rho}f_j(\bar{\rho}_{X,j};X)}    }{ \sqrt{  f_j(\rho^{(i)};X) }  + \sqrt{ f_j(\rho;X) }} \Bigg)^2 \Bigg] \\ \\ 
	&=  \frac{64m^2(1-\xi)^2}{L^2\xi^3} \norm{\pi^{(i)} - \pi}{2}^2  +  \frac{64m^2(1-\xi)^2}{L^2\xi^3} \sum_{j=0}^m \E_j^{\rho} \Bigg[ \Bigg( \frac{q_{\pi}^\rho(X)- q_{\pi^{(i)}}^{\rho^{(i)}}(X) }{2\big(q_{\pi^{(i)}}^{\rho^{(i)} }(X)^{1/2}  + q_{\pi}^\rho(X)^{1/2}  \big)^2} \Bigg)^2 \Bigg] \\
	&\qquad + \frac{1}{2}m \frac{(1-\xi)^2}{\xi} \norm{\rho^{(i)} -\rho}{2}^2   \sum_{j=0}^m \E_j \Bigg[ \Bigg( \frac{ \maxEval{\partialSecondDerivative{\rho}f_j(\bar{\rho}_{X,j};X)}    }{ \sqrt{  f_j(\rho^{(i)};X) }  + \sqrt{ f_j(\rho;X) }} \Bigg)^2 \Bigg].
\end{align*}
\normalsize
Thus:
\small
\begin{align*}
	\int  d \frac{ K_1(z,\theta,\theta^{(i)})^2}{\norm{\theta^{(i)} - \theta}{2}^2  } d\widetilde{\mu}(z) \\
	 &\leq   \frac{64m^2(1-\xi)^2}{L^2\xi^3} \norm{\pi^{(i)} - \pi}{2}^2  +  \frac{16m^2(1-\xi)^2}{L^2\xi^3} \sum_{j=0}^m \E_j^{\rho} \Bigg[ \Bigg( \frac{q_{\pi}^\rho(X)- q_{\pi^{(i)}}^{\rho^{(i)}}(X) }{  q_{\pi}^\rho(X) } \Bigg)^2 \Bigg] \\
	&\qquad + \frac{1}{2}m \frac{(1-\xi)^2}{\xi} \norm{\rho^{(i)} -\rho}{2}^2   \sum_{j=0}^m \E_j \Bigg[ \Bigg( \frac{ \maxEval{\partialSecondDerivative{\rho}f_j(\bar{\rho}_{X,j};X)}    }{ \sqrt{  f_j(\rho^{(i)};X) }  + \sqrt{ f_j(\rho;X) }} \Bigg)^2 \Bigg].
\end{align*}
\normalsize
Thus, to show that the LHS above converges to $0$, it suffices to show that $\E_j^{\rho} \Big[ \Big( \frac{q_{\pi}^\rho(X)- q_{\pi^{(i)}}^{\rho^{(i)}}(X) }{  q_{\pi}^\rho(X) } \Big)^2 \Big] \to 0$ for each $j\in\Y$. Towards that end, note that by equations \eqref{diff in tilted mixture} and \eqref{purple castle}:
\small
\begin{align}
	  &\Bigg( \frac{q_{\pi}^\rho(x)- q_{\pi^{(i)}}^{\rho^{(i)}}(x) }{  q_{\pi}^\rho(x) } \Bigg)^2   \\
	  &=   \Bigg( \frac{\sum_{l=0}^m \pi_l^{(i)}  (q_l^{\rho^{(i)}}(x) - q_l^{\rho}(x) )   + q_{\pi}^{\rho }(x) \dotprod{ s_{\pi}^{\rho }(x) }{ \pi^{(i)}-\pi } }{  q_{\pi}^\rho(x) } \Bigg)^2  \nonumber \\ 
	&\leq 2   \Bigg( \frac{\sum_{l=0}^m \pi_l^{(i)}  (q_l^{\rho^{(i)}}(x) - q_l^{\rho}(x) )  }{  q_{\pi}^\rho(x) }  \Bigg)^2 +  2  \Big[ \dotprod{ s_{\pi}^{\rho }(x) }{ \pi^{(i)}-\pi }^2 \Big] \nonumber \\ 
	&\leq 2   \frac{\sum_{l=0}^m  (q_l^{\rho^{(i)}}(x) - q_l^{\rho}(x) )^2  }{  q_{\pi}^\rho(x)^2 }   +  2 \norm{\pi^{(i)}-\pi}{2}^2  \Big[ \norm{ s_{\pi}^{\rho }(x)}{2}^2 \Big] \nonumber \\ 
	&\leq \frac{2}{\xi^2}   \sum_{l=0}^m  \frac{(q_l^{\rho^{(i)}}(x) - q_l^{\rho}(x) )^2  }{  q_{l}^\rho(x)^2 }   +  2 \norm{\pi^{(i)}-\pi}{2}^2  \frac{4m}{L^2}\nonumber \\ 
	&= \frac{2}{\xi^2} \sum_{l=0}^m     \bigg(\frac{q_l^{\rho^{(i)}}(x) - q_l^{\rho}(x)    }{  q_{l}^\rho(x)} \bigg)^2   +  \frac{8m}{L^2} \norm{\pi^{(i)}-\pi}{2}^2 \nonumber  \\  
	&\leq \frac{2}{\xi^2} \sum_{l=0}^m    \Bigg[  \frac{32m}{L^2}  \norm{\rho^{(i)} - \rho}{2}^2 + \frac{1}{2 f_l(\rho; x)^2 } \maxEvalBigg{\partialSecondDerivative{\rho}f_l(\bar{\rho}_{x,l};x)} ^2 \norm{\rho^{(i)} - \rho}{2}^4   \Bigg] +  \frac{8m}{L^2} \norm{\pi^{(i)}-\pi}{2}^2 \nonumber  \\  
	&=  \frac{64m(m+1)}{L^2\xi^2}  \norm{\rho^{(i)} - \rho}{2}^2 +   \frac{ \norm{\rho^{(i)} - \rho}{2}^4}{\xi^2}\sum_{l=0}^m    \Bigg[ \frac{1}{ f_l(\rho; x)^2 } \maxEvalBigg{\partialSecondDerivative{\rho}f_l(\bar{\rho}_{x,l};x)} ^2   \Bigg] +  \frac{8m}{L^2} \norm{\pi^{(i)}-\pi}{2}^2. \label{high polymer}
\end{align}
\normalsize
Now, $f_l(\rho; x)$ can be shown to be uniformly bounded away from $0$, and $\maxEval{\partialSecondDerivative{\rho}f_l(\bar{\rho}_{x,l};x)}$ can be shown to be uniformly bounded away from $\infty$. Thus, the expected value of the LHS above has an upper bound that converges to $0$ as $\theta^{(i)} \to \theta$. That is, $\E_j^{\rho} \Big[ \Big( \frac{q_{\pi}^\rho(X)- q_{\pi^{(i)}}^{\rho^{(i)}}(X) }{  q_{\pi}^\rho(X) } \Big)^2 \Big] \to 0$, and so 
\[
	\int  d \frac{ K_1(z,\theta,\theta^{(i)})^2}{\norm{\theta^{(i)} - \theta}{2}^2  } d\widetilde{\mu}(z)   \to 0
\]
as well. \\ 

Finally, we'll show that the integral involving $K_3$ also converges to 0. 
\smaller
\begin{align*}
	\int \frac{\Big[  d\sqrt{ q_{\pi}^{\rho}(x) } K_3(z,\theta,\theta^{(i)})   \Big]^2}{ \norm{\theta^{(i)} - \theta}{2}^2  } d\widetilde{\mu}(z) &= \int d q_{\pi}^{\rho}(x) \frac{   K_3(z,\theta,\theta^{(i)})^2}{ \norm{\theta^{(i)} - \theta}{2}^2  } d\widetilde{\mu}(z) \\ 
	&= \E_{\pi}^\rho\Bigg[   \Bigg(  \frac{  \sqrt{q_{\pi}^{\rho}(X)}   }{ \sqrt{q_{\pi^{(i)}}^{\rho^{(i)}} (X)}  + \sqrt{q_{\pi}^{\rho}(X) }}   - \frac{1}{2} \Bigg)^2    \frac{  \dotprod{ s_{\pi}^{\rho }(X) }{ \pi^{(i)}-\pi }^2}{ \norm{\theta^{(i)} - \theta}{2}^2  }   \Bigg] \\ 
	&\leq \E_{\pi}^\rho\Bigg[   \Bigg(  \frac{  \sqrt{q_{\pi}^{\rho}(X)}   }{ \sqrt{q_{\pi^{(i)}}^{\rho^{(i)}}(X)}  + \sqrt{q_{\pi}^{\rho}(X) }}   - \frac{1}{2} \Bigg)^2    \frac{  \norm{s_{\pi}^{\rho }(X) }{2}^2 \norm{\pi^{(i)}-\pi}{2}^2 }{ \norm{\theta^{(i)} - \theta}{2}^2  }   \Bigg] \\ 
	&\leq  \frac{4m}{L^2} \E_{\pi}^\rho\Bigg[   \Bigg(  \frac{  \sqrt{q_{\pi}^{\rho}(X)}   }{ \sqrt{q_{\pi^{(i)}}^{\rho^{(i)}}(X)}  + \sqrt{q_{\pi}^{\rho}(X) }}   - \frac{1}{2} \Bigg)^2     \Bigg] \\ 
	&= \frac{4m}{L^2} \E_{\pi}^\rho\Bigg[  \Bigg( \frac{q_{\pi}^\rho(X)- q_{\pi^{(i)}}^{\rho^{(i)}}(X) }{2\big(q_{\pi^{(i)}}^{\rho^{(i)} }(X)^{1/2}  + q_{\pi}^\rho(X)^{1/2}  \big)^2} \Bigg)^2    \Bigg] \\ 
	&\leq \frac{m}{L^2} \E_{\pi}^\rho\Bigg[  \Bigg( \frac{q_{\pi}^\rho(X)- q_{\pi^{(i)}}^{\rho^{(i)}}(X) }{   q_{\pi}^\rho(X)  } \Bigg)^2    \Bigg] \\
	&\leq \frac{m}{L^2} \sum_{j=0}^m \pi_j \E_{j}^\rho\Bigg[  \Bigg( \frac{q_{\pi}^\rho(X)- q_{\pi^{(i)}}^{\rho^{(i)}}(X) }{   q_{\pi}^\rho(X)  } \Bigg)^2    \Bigg].
\end{align*}
\normalsize
Since we already showed that each summand in the last line displayed above converges to $0$, it follows that the LHS converges to $0$ as well. Thus, we have shown that all three integrals (involving $K_1,K_2$ and $K_3$) converge to $0$, so it follows that
\[
	\int \frac{\big[ \Jcal^{\theta^{(i)},\tau}(z)^{1/2}  - \Jcal^{\theta,\tau}(z)^{1/2} - D(z,\theta)' (\theta^{(i)} - \theta) \big]^2}{ \norm{\theta^{(i)} - \theta}{2}^2 }  d\widetilde{\mu}(z)  \to 0
\] 
as well, and so property (c) has been verified! \\ \\

\uline{Next, let's prove property (b)}. Observe that:
\small
\begin{align*}
	&\norm{D(z,\theta^{(i)}) - D(z,\theta)}{2}^2 \\
	 &= \norm{D(z,\pi^{(i)};\rho^{(i)}) - D(z,\pi;\rho)}{2}^2   +  \norm{D(z,\rho^{(i)};\pi^{(i)}) - D(z,\rho;\pi)}{2}^2 \\ \\
	&= \frac{d\tau}{4} \overbrace{\Bignorm{  q_{\pi^{(i)}}^{\rho^{(i)}}(x)^{1/2} s_{\pi^{(i)}}^{\rho^{(i)}}(x)  - q_{\pi}^{\rho}(x)^{1/2} s_{\pi}^{\rho}(x) }{2}^2}^{=: \ W_1(z,\theta,\theta^{(i)}) } \\ 
	&\qquad + \frac{d\tau}{4} \overbrace{ \Biggnorm{  \sum_{j=0}^m \pi_j^{(i)} \frac{ q_j^{\rho^{(i)}}(x) }{ \sqrt{q_{\pi^{(i)}}^{\rho^{(i)}}(x)} }   \frac{\pi_j^\ast}{\pi_j^\train} (s_{\gamma^\ast(\tau)}(x) - \E_j^{\rho^{(i)}}[s_{\gamma^\ast(\tau)}])   - \pi_j \frac{ q_j^{\rho}(x) }{ \sqrt{q_{\pi}^\rho(x)} }   \frac{\pi_j^\ast}{\pi_j^\train} (s_{\gamma^\ast(\tau)}(x) - \E_j^{\rho}[s_{\gamma^\ast(\tau)}])  }{2}^2}^{=:\ W_2(z,\theta,\theta^{(i)})} \\ 
	&\qquad + \frac{(1-d)(1-\tau)\pi_y^\train}{4}  \frac{(\pi_y^\ast)^2}{(\pi_y^\train)^2} \overbrace{\Bignorm{ \big(s_{\gamma^\ast(\tau)}(x) - \E_y^{\rho^{(i)}}[s_{\gamma^\ast(\tau)}]\big) q_y^{\rho^{(i)}}(x)^{1/2}    -  \big(s_{\gamma^\ast(\tau)}(x) - \E_y^{\rho}[s_{\gamma^\ast(\tau)}]\big) q_y^\rho(x)^{1/2}   }{2}^2}^{=: \ W_3(z,\theta,\theta^{(i)})}.
\end{align*}
\normalsize
Ergo:
\small
\begin{align*}
	&\int \norm{D(z,\theta^{(i)}) - D(z,\theta)}{2}^2 d\widetilde{\mu}(z)\\
	 &\leq  \int  dW_1(z,\theta,\theta^{(i)}) \  d\widetilde{\mu}(z) +  \int  d W_2(z,\theta,\theta^{(i)}) \  d\widetilde{\mu}(z) +    \frac{(1-\xi)^2}{\xi} \int  (1-d) W_3(z,\theta,\theta^{(i)})  \ d\widetilde{\mu}(z).
\end{align*}
\normalsize

Let's start with tackling the $W_1$ term.  Observe that:
\begin{align*}
	&W_1(z,\theta,\theta^{(i)})^{1/2} \\
	&=\Bignorm{  q_{\pi^{(i)}}^{\rho^{(i)}}(x)^{1/2} s_{\pi^{(i)}}^{\rho^{(i)}}(x)  - q_{\pi}^{\rho}(x)^{1/2} s_{\pi}^{\rho}(x) }{2} \\ 
	&\leq \Bignorm{  q_{\pi^{(i)}}^{\rho^{(i)}}(x)^{1/2} s_{\pi^{(i)}}^{\rho^{(i)}}(x)  - q_{\pi}^{\rho}(x)^{1/2}s_{\pi^{(i)}}^{\rho^{(i)}}(x)  }{2}     +      \Bignorm{ q_{\pi}^{\rho}(x)^{1/2}s_{\pi^{(i)}}^{\rho^{(i)}}(x)  - q_{\pi}^{\rho}(x)^{1/2} s_{\pi}^{\rho}(x) }{2}\\ 
	&= \Bignorm{  q_{\pi^{(i)}}^{\rho^{(i)}}(x)^{1/2} s_{\pi^{(i)}}^{\rho^{(i)}}(x)  - q_{\pi}^{\rho}(x)^{1/2}s_{\pi^{(i)}}^{\rho^{(i)}}(x)  }{2}     +     q_{\pi}^{\rho}(x)^{1/2} \Bignorm{  s_{\pi^{(i)}}^{\rho^{(i)}}(x)  - s_{\pi}^{\rho}(x) }{2}.
\end{align*}
Note that:
\begin{align*}
	 \Bignorm{  q_{\pi^{(i)}}^{\rho^{(i)}}(x)^{1/2} s_{\pi^{(i)}}^{\rho^{(i)}}(x)  - q_{\pi}^{\rho}(x)^{1/2}s_{\pi^{(i)}}^{\rho^{(i)}}(x)  }{2} &= \sqrt{\sum_{j=1}^m \Big(  q_{\pi^{(i)}}^{\rho^{(i)}}(x)^{1/2}s_{\pi^{(i)},j}^{\rho^{(i)}}(x)   - q_{\pi}^{\rho}(x)^{1/2}s_{\pi^{(i)},j}^{\rho^{(i)}}(x) \Big)^2}   \\
	&=   \sqrt{\sum_{j=1}^m s_{\pi^{(i)},j}^{\rho^{(i)}}(x)^2 \Big(  q_{\pi^{(i)}}^{\rho^{(i)}}(x)^{1/2}   - q_{\pi}^{\rho}(x)^{1/2}  \Big)^2}    \\ 
	&=   \absBig{ q_{\pi^{(i)}}^{\rho^{(i)}}(x)^{1/2}   - q_{\pi}^{\rho}(x)^{1/2}}\sqrt{\sum_{j=1}^m s_{\pi^{(i)},j}^{\rho^{(i)}}(x)^2}    \\ 
	&=   \absBig{ q_{\pi^{(i)}}^{\rho^{(i)}}(x)^{1/2}   - q_{\pi}^{\rho}(x)^{1/2}} \norm{s_{\pi^{(i)}}^{\rho^{(i)}}(x)}{2}   \\ 
	&\leq  \frac{2\sqrt{m}}{\xi} \absBig{ q_{\pi^{(i)}}^{\rho^{(i)}}(x)^{1/2}   - q_{\pi}^{\rho}(x)^{1/2} }   \\ 
	&=  \frac{2\sqrt{m}}{\xi} \frac{ \absbig{ q_{\pi^{(i)}}^{\rho^{(i)}}(x)   - q_{\pi}^{\rho}(x)} }{ q_{\pi^{(i)}}^{\rho^{(i)}}(x)^{1/2}   + q_{\pi}^{\rho}(x)^{1/2}  }  \\ 
	&\leq  \frac{2\sqrt{m}}{\xi} \frac{ \absbig{ q_{\pi^{(i)}}^{\rho^{(i)}}(x)   - q_{\pi}^{\rho}(x)} }{   q_{\pi}^{\rho}(x)^{1/2}  }.
\end{align*}
Thus, by equations \eqref{diff in tilted mixture} and \eqref{diff in tilted component}, we have that:
\smaller
\begin{align*}
	 &\Bignorm{  q_{\pi^{(i)}}^{\rho^{(i)}}(x)^{1/2} s_{\pi^{(i)}}^{\rho^{(i)}}(x)  - q_{\pi}^{\rho}(x)^{1/2}s_{\pi^{(i)}}^{\rho^{(i)}}(x)  }{2} \\
	 &\leq \frac{2\sqrt{m}}{\xi} \frac{ \absbig{ \sum_{j=0}^m \pi_j^{(i)}  (q_j^{\rho^{(i)}}(x) - q_j^{\rho}(x) )   + q_{\pi}^{\rho }(x) \dotprod{ s_{\pi}^{\rho }(x) }{ \pi^{(i)}-\pi }  } }{   q_{\pi}^{\rho}(x)^{1/2}  } \\ 
	 &\leq \frac{2\sqrt{m}}{\xi} \Bigg\{ \frac{ \sum_{j=0}^m \pi_j^{(i)}   \absbig{ q_j^{\rho^{(i)}}(x) - q_j^{\rho}(x) }   + q_{\pi}^{\rho}(x)  \abs{ \dotprod{ s_{\pi}^{\rho }(x) }{ \pi^{(i)}-\pi } }  }{   q_{\pi}^{\rho}(x)^{1/2}  }\Bigg\} \\ 
	 &= \frac{2\sqrt{m}}{\xi} \Bigg\{ \sum_{j=0}^m \pi_j^{(i)}  \frac{   \absbig{ q_j^{\rho^{(i)}}(x) - q_j^{\rho}(x) }     }{   q_{\pi}^{\rho}(x)^{1/2}  } + q_{\pi}^{\rho}(x)^{1/2} \abs{ \dotprod{ s_{\pi}^{\rho }(x) }{ \pi^{(i)}-\pi } } \Bigg\} \\ 
	 &\leq \frac{2\sqrt{m}}{\xi^{3/2}} \Bigg\{ \sum_{j=0}^m   \frac{   \absbig{ q_j^{\rho^{(i)}}(x) - q_j^{\rho}(x) }     }{    q_{j}^{\rho}(x)^{1/2}  } +  \frac{2\sqrt{m}}{\xi}\norm{ \pi^{(i)}-\pi }{2} q_{\pi}^{\rho}(x)^{1/2}   \Bigg\} \\ 
	 &= \frac{2\sqrt{m}}{\xi^{3/2}} \Bigg\{ \sum_{j=0}^m  q_{j}^{\rho}(x)^{1/2}  \frac{   \absbig{ q_j^{\rho^{(i)}}(x) - q_j^{\rho}(x) }     }{    q_{j}^{\rho}(x)  } +  \frac{2\sqrt{m}}{\xi}\norm{ \pi^{(i)}-\pi }{2} q_{\pi}^{\rho}(x)^{1/2}   \Bigg\} \\  \\ 
	 &= \frac{2\sqrt{m}}{\xi^{3/2}} \Bigg\{ \sum_{j=0}^m  q_{j}^{\rho}(x)^{1/2}  \frac{\pi_j^\ast}{\pi_j^\train} \absBigg{\bigdotprod{ s_{\gamma^\ast(\tau)}(x)- \E_{j}^{\rho}[s_{\gamma^\ast(\tau)}]  }{\rho^{(i)} - \rho} + \frac{ \norm{\rho^{(i)} - \rho}{2}^2  }{2 f_j(\rho; x)}  \maxEvalBigg{\partialSecondDerivative{\rho}f_j(\bar{\rho}_{x,j};x)} } \\
	 &\qquad\qquad\qquad +  \frac{2\sqrt{m}}{\xi} \norm{ \pi^{(i)}-\pi }{2} q_{\pi}^{\rho}(x)^{1/2}   \Bigg\} \\ \\
	 &\leq \frac{2\sqrt{m}}{\xi^{3/2}} \Bigg\{ \sum_{j=0}^m  q_{j}^{\rho}(x)^{1/2} \absBigg{ \frac{4\sqrt{m}(1-\xi)}{L\xi} \norm{\rho^{(i)} - \rho}{2} + \frac{\norm{\rho^{(i)} - \rho}{2}^2}{2 f_j(\rho; x)}  \maxEvalBigg{\partialSecondDerivative{\rho}f_j(\bar{\rho}_{x,j};x)} } \\
	 &\qquad\qquad\qquad +  \frac{2\sqrt{m}}{\xi} \norm{ \pi^{(i)}-\pi }{2} q_{\pi}^{\rho}(x)^{1/2}   \Bigg\} \\ \\
	 &= \frac{2\sqrt{m}}{\xi^{3/2}}  \sum_{j=0}^m  q_{j}^{\rho}(x)^{1/2}  \absBigg{ \frac{4\sqrt{m}(1-\xi)}{L\xi} \norm{\rho^{(i)} - \rho}{2} + \frac{\norm{\rho^{(i)} - \rho}{2}^2 }{2 f_j(\rho; x)} \maxEvalBigg{\partialSecondDerivative{\rho}f_j(\bar{\rho}_{x,j};x)} } \\
	 &\qquad\qquad\qquad + \frac{4 m}{\xi^{5/2}}\norm{ \pi^{(i)}-\pi }{2} q_{\pi}^{\rho}(x)^{1/2}. 
\end{align*}
\normalsize

Next, observe that:
\small
\begin{align*}
	&\Bignorm{ s_{\pi^{(i)}}^{\rho^{(i)}}(x)  -  s_{\pi}^{\rho}(x) }{2}   \\
	&=   \sqrt{\sum_{j=1}^m \Big( s_{\pi^{(i)},j }^{\rho^{(i)}}(x)  -  s_{\pi,j}^{\rho}(x) \Big)^2 } \\ 
	 &=   \sqrt{\sum_{j=1}^m \Bigg( \frac{q_j^{\rho^{(i)}}(x)- q_0^{\rho^{(i)}}(x) }{q_{\pi^{(i)} }^{\rho^{(i)}}(x)}  -  \frac{q_j^{\rho}(x)- q_0^{\rho}(x) }{q_{\pi}^{\rho}(x)} \Bigg)^2 } \\ 
	\shortintertext{which, after some simple algebra, can be shown to be upper bounded by} 
	&\leq  \sqrt{2 \sum_{j=1}^m \Bigg[        \Bigg(\frac{[q_j^{\rho^{(i)}}(x)- q_0^{\rho^{(i)}}(x)]-[q_j^{\rho}(x)- q_0^{\rho}(x)] }{q_{\pi^{(i)} }^{\rho^{(i)}}(x)}\Bigg)^2      + \Bigg(\frac{ [q_j^{\rho}(x)- q_0^{\rho}(x)][q_{\pi^{(i)} }^{\rho^{(i)}}(x) - q_{\pi}^{\rho}(x)] }{      q_{\pi^{(i)} }^{\rho^{(i)}}(x)   q_{\pi}^{\rho}(x)      } \Bigg)^2   \Bigg] } \\ 
	&=  \sqrt{ 2 \sum_{j=1}^m         \Bigg(\frac{ [q_j^{\rho^{(i)}}(x) - q_j^{\rho}(x) ] }{ q_{\pi^{(i)} }^{\rho^{(i)}}(x) } + \frac{  [ q_0^{\rho}(x) - q_0^{\rho^{(i)}}(x)]   }{  q_{\pi^{(i)} }^{\rho^{(i)}}(x)  }\Bigg)^2      + 2  \Bigg(  \frac{ [q_{\pi^{(i)} }^{\rho^{(i)}}(x) - q_{\pi}^{\rho}(x)] }{         q_{\pi}^{\rho}(x)      } \Bigg)^2 \sum_{j=1}^m  \Bigg( \frac{[q_j^{\rho}(x)- q_0^{\rho}(x)]}{q_{\pi^{(i)} }^{\rho^{(i)}}(x)}\Bigg)^2    } \\ 
	&\leq  \sqrt{ 4m \sum_{j=1}^m         \Bigg(\frac{ q_j^{\rho^{(i)}}(x) - q_j^{\rho}(x)  }{ q_{\pi^{(i)} }^{\rho^{(i)}}(x) }\Bigg)^2 + 4m\Bigg(   \frac{   q_0^{\rho}(x) - q_0^{\rho^{(i)}}(x)   }{  q_{\pi^{(i)} }^{\rho^{(i)}}(x)  }    \Bigg)^2       + 2  \Bigg(  \frac{ q_{\pi^{(i)} }^{\rho^{(i)}}(x) - q_{\pi}^{\rho}(x) }{         q_{\pi}^{\rho}(x)      } \Bigg)^2 \sum_{j=1}^m  \Bigg( \frac{q_j^{\rho}(x)}{q_{\pi^{(i)} }^{\rho^{(i)}}(x)} + \frac{q_0^{\rho}(x)}{ q_{\pi^{(i)} }^{\rho^{(i)}}(x) }\Bigg)^2    } \\ 
	&\leq  \sqrt{ 4m \sum_{j=0}^m         \Bigg(\frac{ q_j^{\rho^{(i)}}(x) - q_j^{\rho}(x)  }{ \pi_j^{(i)} q_{j }^{\rho^{(i)}}(x) }\Bigg)^2        + 2  \Bigg(  \frac{ q_{\pi^{(i)} }^{\rho^{(i)}}(x) - q_{\pi}^{\rho}(x) }{         q_{\pi}^{\rho}(x)      } \Bigg)^2 \sum_{j=1}^m  \Bigg( \frac{q_j^{\rho}(x)}{ \pi_j^{(i)} q_{j}^{\rho^{(i)}}(x)} + \frac{q_0^{\rho}(x)}{ \pi_0^{(i)} q_{0}^{\rho^{(i)}}(x) }\Bigg)^2    } \\ 
	&\leq  \sqrt{ \frac{4m}{\xi^2} \sum_{j=0}^m         \Bigg(\frac{ q_j^{\rho^{(i)}}(x) - q_j^{\rho}(x)  }{ q_{j }^{\rho^{(i)}}(x) }\Bigg)^2        + \frac{2}{\xi^2} \Bigg(  \frac{ q_{\pi^{(i)} }^{\rho^{(i)}}(x) - q_{\pi}^{\rho}(x) }{         q_{\pi}^{\rho}(x)      } \Bigg)^2 \sum_{j=1}^m  \Bigg( \frac{q_j^{\rho}(x)}{ q_{j}^{\rho^{(i)}}(x)} + \frac{q_0^{\rho}(x)}{  q_{0}^{\rho^{(i)}}(x) }\Bigg)^2    } \\ 
	&=  \sqrt{   \frac{4m}{\xi^2} \sum_{j=0}^m         \Bigg(\frac{ q_j^{\rho^{(i)}}(x) - q_j^{\rho}(x)  }{ q_{j}^{\rho}(x) } \Bigg)^2 \Bigg( \frac{f_j(\rho;x)}{ f_j(\rho^{(i)};x) }  \Bigg)^2        + \frac{2}{\xi^2} \Bigg(  \frac{ q_{\pi^{(i)} }^{\rho^{(i)}}(x) - q_{\pi}^{\rho}(x) }{         q_{\pi}^{\rho}(x)      } \Bigg)^2 \sum_{j=1}^m  \Bigg(\frac{f_j(\rho;x)}{ f_j(\rho^{(i)};x) } + \frac{f_0(\rho;x)}{ f_0(\rho^{(i)};x) }\Bigg)^2    }.
\end{align*}
\normalsize

Thus, overall, we have that:
\small
\begin{align*}
	& \int  dW_1(z,\theta,\theta^{(i)}) \  d\widetilde{\mu}(z) \\
	&\leq  \int  d \bigg(  \Bignorm{  q_{\pi^{(i)}}^{\rho^{(i)}}(x)^{1/2} s_{\pi^{(i)}}^{\rho^{(i)}}(x)  - q_{\pi}^{\rho}(x)^{1/2}s_{\pi^{(i)}}^{\rho^{(i)}}(x)  }{2}     +     q_{\pi}^{\rho}(x)^{1/2} \Bignorm{  s_{\pi^{(i)}}^{\rho^{(i)}}(x)  - s_{\pi}^{\rho}(x) }{2} \bigg)^2 \  d\widetilde{\mu}(z) \\ 
	 &\leq 2\int  d \Bignorm{  q_{\pi^{(i)}}^{\rho^{(i)}}(x)^{1/2} s_{\pi^{(i)}}^{\rho^{(i)}}(x)  - q_{\pi}^{\rho}(x)^{1/2}s_{\pi^{(i)}}^{\rho^{(i)}}(x)  }{2}^2   \ d\widetilde{\mu}(z)        +          2\int  d q_{\pi}^{\rho}(x) \Bignorm{  s_{\pi^{(i)}}^{\rho^{(i)}}(x)  - s_{\pi}^{\rho}(x) }{2}^2 \ d\widetilde{\mu}(z) \\ 
	 &= 2\int  d \Bignorm{  q_{\pi^{(i)}}^{\rho^{(i)}}(x)^{1/2} s_{\pi^{(i)}}^{\rho^{(i)}}(x)  - q_{\pi}^{\rho}(x)^{1/2}s_{\pi^{(i)}}^{\rho^{(i)}}(x)  }{2}^2   \ d\widetilde{\mu}(z)        +          2 \E_{\pi}^{\rho}\Big[ \Bignorm{  s_{\pi^{(i)}}^{\rho^{(i)}}(X)  - s_{\pi}^{\rho}(X) }{2}^2 \Big],
\end{align*}
\normalsize
where 
\smaller
\begin{align*}
	&\int  d \Bignorm{  q_{\pi^{(i)}}^{\rho^{(i)}}(x)^{1/2} s_{\pi^{(i)}}^{\rho^{(i)}}(x)  - q_{\pi}^{\rho}(x)^{1/2}s_{\pi^{(i)}}^{\rho^{(i)}}(x)  }{2}^2   \ d\widetilde{\mu}(z)\\
	 &\leq  \frac{8m^2}{\xi^3} \sum_{j=0}^m \int  d q_j^{\rho}(x) \Bigg(  \frac{4\sqrt{m}(1-\xi)}{L\xi} \norm{\rho^{(i)} - \rho}{2} + \frac{\norm{\rho^{(i)} - \rho}{2}^2 }{2 f_j(\rho; x)} \maxEvalBigg{\partialSecondDerivative{\rho}f_j(\bar{\rho}_{x,j};x)} \Bigg)^2   \ d\widetilde{\mu}(z) \\
	&\qquad\qquad + \frac{32m^2}{\xi^5} \norm{ \pi^{(i)} - \pi }{2}^2 \int dq_{\pi}^{\rho}(x)  \ d\widetilde{\mu}(z) \\ \\ 
	&=  \frac{8m^2}{\xi^3} \sum_{j=0}^m \E_j^{\rho}\Bigg[ \Bigg(  \frac{4\sqrt{m}(1-\xi)}{L\xi} \norm{\rho^{(i)} - \rho}{2} + \frac{\norm{\rho^{(i)} - \rho}{2}^2 }{2 f_j(\rho; X)} \maxEvalBigg{\partialSecondDerivative{\rho}f_j(\bar{\rho}_{X,j};X)} \Bigg)^2  \Bigg]  \\
	&\qquad\qquad + \frac{32m^2}{\xi^5} \norm{ \pi^{(i)} - \pi }{2}^2.
\end{align*}
\normalsize
Since $\norm{s_{\gamma^\ast(\tau)}}{2}$ and $\norm{\rho}{2}$ are bounded, one may show that $\frac{1}{f_j(\rho; x)} \maxEvalBig{\partialSecondDerivative{\rho}f_j(\bar{\rho}_{x,j};x)}  $ is uniformly bounded for all $x\in \X$, meaning that the last line above tends to $0$ as  $\theta^{(i)} \to \theta$. Also note that 
\smaller
\begin{align*}
	&\E_{\pi}^{\rho}\Big[ \Bignorm{  s_{\pi^{(i)}}^{\rho^{(i)}}(X)  - s_{\pi}^{\rho}(X) }{2}^2 \Big] \\
	&\leq  \E_{\pi}^{\rho}\Bigg[ \frac{4m}{\xi^2} \sum_{j=0}^m         \Bigg(\frac{ q_j^{\rho^{(i)}}(X) - q_j^{\rho}(X)  }{ q_{j}^{\rho}(X) } \Bigg)^2 \Bigg( \frac{f_j(\rho;x)}{ f_j(\rho^{(i)};X) }  \Bigg)^2        + \frac{2}{\xi^2} \Bigg(  \frac{ q_{\pi^{(i)} }^{\rho^{(i)}}(x) - q_{\pi}^{\rho}(X) }{         q_{\pi}^{\rho}(X)      } \Bigg)^2 \sum_{j=1}^m  \Bigg(\frac{f_j(\rho;X)}{ f_j(\rho^{(i)};X) } + \frac{f_0(\rho;X)}{ f_0(\rho^{(i)};X) }\Bigg)^2 \Bigg],
\end{align*}
\normalsize
which by applying equations  \eqref{purple castle} and  \eqref{high polymer}, can be showed to converge to zero via an analogous argument. Thus, we may conclude that $\int  dW_1(z,\theta,\theta^{(i)}) \  d\widetilde{\mu}(z) \to 0$ as $\theta^{(i)} \to \theta$.

Next, we show that $\int  d W_2(z,\theta,\theta^{(i)}) \  d\widetilde{\mu}(z) $ converges to 0. Towards that end, define $\epsilon_j(x,\rho,\rho^{(i)}) := \partialDerivative{\rho}\{ f_j(\rho^{(i)};x) \} -  \partialDerivative{\rho}\{ f_j(\rho;x) \} $, and observe that:
\small
\begin{align*}
	& W_2(z,\theta,\theta^{(i)})^{1/2}\\
	 &= \Biggnorm{  \sum_{j=0}^m \pi_j^{(i)} \frac{\pi_j^\ast}{\pi_j^\train} \frac{ q_j^{\rho^{(i)}}(x) }{ \sqrt{q_{\pi^{(i)}}^{\rho^{(i)}}(x)} }  (s_{\gamma^\ast(\tau)}(x) - \E_j^{\rho^{(i)}}[s_{\gamma^\ast(\tau)}])   - \pi_j  \frac{\pi_j^\ast}{\pi_j^\train} \frac{ q_j^{\rho}(x) }{ \sqrt{q_{\pi}^\rho(x)} }  (s_{\gamma^\ast(\tau)}(x) - \E_j^{\rho}[s_{\gamma^\ast(\tau)}])  }{2} \\ 
	 &\leq    \sum_{j=0}^m  \frac{\pi_j^\ast}{\pi_j^\train} \Biggnorm{ \pi_j^{(i)} \frac{ q_j^{\rho^{(i)}}(x) }{ \sqrt{q_{\pi^{(i)}}^{\rho^{(i)}}(x)} }  (s_{\gamma^\ast(\tau)}(x) - \E_j^{\rho^{(i)}}[s_{\gamma^\ast(\tau)}])   - \pi_j \frac{ q_j^{\rho}(x) }{ \sqrt{q_{\pi}^\rho(x)} }  (s_{\gamma^\ast(\tau)}(x) - \E_j^{\rho}[s_{\gamma^\ast(\tau)}])  }{2} \\
	 &=    \sum_{j=0}^m  \frac{\pi_j^\ast}{\pi_j^\train}  \Biggnorm{ \pi_j^{(i)} \frac{ p_j(x) \partialDerivative{\rho}\{ f_j(\rho^{(i)};x) \}  }{ \sqrt{q_{\pi^{(i)}}^{\rho^{(i)}}(x)} }     - \pi_j \frac{  p_j(x) \partialDerivative{\rho}\{ f_j(\rho;x) \}   }{ \sqrt{q_{\pi}^\rho(x)} }   }{2} \\
	  &=    \sum_{j=0}^m   \frac{\pi_j^\ast}{\pi_j^\train} p_j(x)\Biggnorm{ \pi_j^{(i)} \frac{   \partialDerivative{\rho}\{ f_j(\rho;x) \}   +  \epsilon_j(x,\rho,\rho^{(i)})  }{ \sqrt{q_{\pi^{(i)}}^{\rho^{(i)}}(x)} }     - \pi_j \frac{  \partialDerivative{\rho}\{ f_j(\rho;x) \}   }{ \sqrt{q_{\pi}^\rho(x)} }   }{2} \\
	  &\leq  \frac{1-\xi}{\xi}  \sum_{j=0}^m  p_j(x)\Biggnorm{ \pi_j^{(i)} \frac{   \partialDerivative{\rho}\{ f_j(\rho;x) \}    }{ \sqrt{q_{\pi^{(i)}}^{\rho^{(i)}}(x)} }      - \pi_j \frac{  \partialDerivative{\rho}\{ f_j(\rho;x) \}   }{ \sqrt{q_{\pi}^\rho(x)} }   }{2} + \frac{1-\xi}{\xi} \sum_{j=0}^m p_j(x) \Biggnorm{  \frac{ \epsilon_j(x,\rho,\rho^{(i)})}{ \sqrt{q_{\pi^{(i)}}^{\rho^{(i)}}(x)}  }   }{2} \\
	  &=   \frac{1-\xi}{\xi}  \sum_{j=0}^m  p_j(x) \biggnorm{\partialDerivative{\rho}\{ f_j(\rho;x) \}}{2} \absBigg{  \frac{   \pi_j^{(i)}    }{ q_{\pi^{(i)}}^{\rho^{(i)}}(x)^{1/2} }      -  \frac{  \pi_j  }{ q_{\pi}^\rho(x)^{1/2} } } + \frac{1-\xi}{\xi}  \sum_{j=0}^m p_j(x) \Biggnorm{  \frac{ \epsilon_j(x,\rho,\rho^{(i)})}{ \sqrt{q_{\pi^{(i)}}^{\rho^{(i)}}(x)}  }   }{2} \\
	  &=  \frac{1-\xi}{\xi}  \sum_{j=0}^m  q_j^{\rho}(x) \bignorm{ s_{\gamma^\ast(\tau)}(x) - \E_j^{\rho}[s_{\gamma^\ast(\tau)}] }{2} \absBigg{  \frac{   \pi_j^{(i)}    }{ q_{\pi^{(i)}}^{\rho^{(i)}}(x)^{1/2} }      -  \frac{  \pi_j  }{ q_{\pi}^\rho(x)^{1/2} } } +  \frac{1-\xi}{\xi} \sum_{j=0}^m p_j(x) \Biggnorm{  \frac{ \epsilon_j(x,\rho,\rho^{(i)})}{ \sqrt{q_{\pi^{(i)}}^{\rho^{(i)}}(x)}  }   }{2}\\
	  &\leq   \frac{4\sqrt{m}(1-\xi)}{L\xi} \sum_{j=0}^m  q_j^{\rho}(x)  \absBigg{  \frac{   \pi_j^{(i)}    }{ q_{\pi^{(i)}}^{\rho^{(i)}}(x)^{1/2} }      -  \frac{  \pi_j  }{ q_{\pi}^\rho(x)^{1/2} } } + \frac{1-\xi}{\xi^{3/2}} \sum_{j=0}^m p_j(x) \Biggnorm{  \frac{ \epsilon_j(x,\rho,\rho^{(i)})}{ \sqrt{  q_{j}^{\rho^{(i)}}(x)}  }   }{2} \\ 
	  &=   \frac{4\sqrt{m}(1-\xi)}{L\xi}\sum_{j=0}^m  q_j^{\rho}(x)  \absBigg{  \frac{   \pi_j^{(i)}    }{ q_{\pi^{(i)}}^{\rho^{(i)}}(x)^{1/2} }      -  \frac{  \pi_j  }{ q_{\pi}^\rho(x)^{1/2} } } + \frac{1-\xi}{\xi^{3/2}}  \sum_{j=0}^m \frac{\sqrt{p_j(x)} }{\sqrt{  f_j(\rho^{(i)},x)} } \bignorm{   \epsilon_j(x,\rho,\rho^{(i)})  }{2}.
\end{align*}
\normalsize

Note that:
\begin{align*}
	\absBigg{  \frac{   \pi_j^{(i)}    }{ q_{\pi^{(i)}}^{\rho^{(i)}}(x)^{1/2} }      -  \frac{  \pi_j  }{ q_{\pi}^\rho(x)^{1/2} } } &= \absBigg{        \frac{\pi_j^{(i)} q_{\pi}^\rho(x)^{1/2} -  \pi_jq_{\pi^{(i)}}^{\rho^{(i)}}(x)^{1/2}   }{q_{\pi^{(i)}}^{\rho^{(i)}}(x)^{1/2} q_{\pi}^\rho(x)^{1/2} }         } \\ 
	 &\leq        \frac{ \absBig{    \pi_j^{(i)} q_{\pi}^\rho(x)^{1/2} -  \pi_j^{(i)} q_{\pi^{(i)}}^{\rho^{(i)}}(x)^{1/2} } +  \absBig{\pi_j^{(i)} q_{\pi^{(i)}}^{\rho^{(i)} }(x)^{1/2}  -  \pi_jq_{\pi^{(i)}}^{\rho^{(i)}}(x)^{1/2} }       }{q_{\pi^{(i)}}^{\rho^{(i)}}(x)^{1/2} q_{\pi}^\rho(x)^{1/2} }          \\ 
	 &=        \frac{\pi_j^{(i)}}{  q_{\pi}^\rho(x)^{1/2}  } \frac{ \absBig{     q_{\pi}^\rho(x)^{1/2} -   q_{\pi^{(i)}}^{\rho^{(i)}}(x)^{1/2} }  }{   q_{\pi^{(i)}}^{\rho^{(i)}}(x)^{1/2}     }     +     \frac{ \absbig{\pi_j^{(i)}   -  \pi_j  }      }{     q_{\pi}^\rho(x)^{1/2}   }          \\ 
	 &=       \frac{1}{ q_{\pi}^\rho(x)^{1/2} } \Bigg(\pi_j^{(i)} \frac{ \absbig{     q_{\pi}^\rho(x)^{1/2} -   q_{\pi^{(i)}}^{\rho^{(i)}}(x)^{1/2} }  }{   q_{\pi^{(i)}}^{\rho^{(i)}}(x)^{1/2}     }     +      \absbig{\pi_j^{(i)}   -  \pi_j  }      \Bigg)        \\ 
	 &\leq       \frac{1}{ \sqrt{\xi} q_{j}^\rho(x)^{1/2} } \Bigg(  \frac{ \absbig{     q_{\pi}^\rho(x)^{1/2} -   q_{\pi^{(i)}}^{\rho^{(i)}}(x)^{1/2} }  }{   q_{\pi^{(i)}}^{\rho^{(i)}}(x)^{1/2}     }     +      \absbig{\pi_j^{(i)}   -  \pi_j  }      \Bigg)        \\ 
	 &=       \frac{1}{ \sqrt{\xi} q_{j}^\rho(x)^{1/2} } \Bigg(  \frac{ \absbig{     q_{\pi}^\rho(x)^{1/2} -   q_{\pi^{(i)}}^{\rho^{(i)}}(x)^{1/2} }  }{   q_{\pi  }^{\rho }(x)^{1/2}     }\frac{q_{\pi}^{\rho}(x)^{1/2}   }{q_{\pi^{(i)}}^{\rho^{(i)}}(x)^{1/2}  }     +      \absbig{\pi_j^{(i)}   -  \pi_j  }      \Bigg)        \\ 
	 &=       \frac{1}{ \sqrt{\xi} q_{j}^\rho(x)^{1/2} } \Bigg( \frac{ \absbig{     q_{\pi}^\rho(x) -   q_{\pi^{(i)}}^{\rho^{(i)}}(x) }  }{   q_{\pi  }^{\rho }(x)^{1/2}  \big(q_{\pi}^\rho(x)^{1/2} +   q_{\pi^{(i)}}^{\rho^{(i)}}(x)^{1/2} \big)   }\frac{ q_{\pi}^{\rho}(x)^{1/2}   }{q_{\pi^{(i)}}^{\rho^{(i)}}(x)^{1/2}  }     +      \absbig{\pi_j^{(i)}   -  \pi_j  }      \Bigg)        \\ 
	 &\leq       \frac{1}{ \sqrt{\xi} q_{j}^\rho(x)^{1/2} } \Bigg(  \frac{ \absbig{     q_{\pi}^\rho(x) -   q_{\pi^{(i)}}^{\rho^{(i)}}(x) }  }{   q_{\pi  }^{\rho }(x)  }\frac{q_{\pi}^{\rho}(x)^{1/2}   }{q_{\pi^{(i)}}^{\rho^{(i)}}(x)^{1/2}  }     +      \absbig{\pi_j^{(i)}   -  \pi_j  }      \Bigg).
\end{align*}

Therefore:
\small
\begin{align*}
	 &W_2(z,\theta,\theta^{(i)})^{1/2}   \\
	 &\leq   \frac{4\sqrt{m} }{L\xi^{3/2} }\sum_{j=0}^m  q_{j}^\rho(x)^{1/2}  \Bigg(  \frac{ \absbig{     q_{\pi}^\rho(x) -   q_{\pi^{(i)}}^{\rho^{(i)}}(x) }  }{   q_{\pi  }^{\rho }(x)  }\frac{q_{\pi}^{\rho}(x)^{1/2}   }{q_{\pi^{(i)}}^{\rho^{(i)}}(x)^{1/2}  }     +      \absbig{\pi_j^{(i)}   -  \pi_j  }      \Bigg)   + \frac{1}{\xi^{3/2}}  \sum_{j=0}^m \frac{\sqrt{p_j(x)} }{\sqrt{  f_j(\rho^{(i)},x)} } \bignorm{   \epsilon_j(x,\rho,\rho^{(i)})  }{2}.
\end{align*}
\normalsize

And so:
\small
\begin{align*}
	&W_2(z,\theta,\theta^{(i)})   \\
	&\leq   \frac{32 m^2}{L^2 \xi^3 } \sum_{j=0}^m  q_{j}^\rho(x)  \Bigg(  \frac{ \absbig{     q_{\pi}^\rho(x) -   q_{\pi^{(i)}}^{\rho^{(i)}}(x) }  }{   q_{\pi  }^{\rho }(x)  }\frac{q_{\pi}^{\rho}(x)^{1/2}   }{q_{\pi^{(i)}}^{\rho^{(i)}}(x)^{1/2}  }     +      \absbig{\pi_j^{(i)}   -  \pi_j  }      \Bigg)^2 + \frac{2m}{\xi^3} \sum_{j=0}^m \frac{p_j(x) }{  f_j(\rho^{(i)},x) } \bignorm{   \epsilon_j(x,\rho,\rho^{(i)})  }{2}^2
\end{align*}
\normalsize

Meaning:
\smaller
\begin{align*}
	&  \int  d W_2(z,\theta,\theta^{(i)}) \  d\widetilde{\mu}(z) \\
	  &\leq \frac{32 m^2}{L^2 \xi^3 } \sum_{j=0}^m   \E_{j}^{\rho}\Bigg[  \Bigg(  \frac{ \absbig{     q_{\pi}^\rho(X) -   q_{\pi^{(i)}}^{\rho^{(i)}}(X) }  }{   q_{\pi  }^{\rho }(X)  }\frac{q_{\pi}^{\rho}(X)^{1/2}   }{q_{\pi^{(i)}}^{\rho^{(i)}}(X)^{1/2}  }     +      \absbig{\pi_j^{(i)}   -  \pi_j  }      \Bigg)^2   \Bigg] + \frac{2m}{\xi^3} \sum_{j=0}^m \E_{j}\Bigg[  \frac{\bignorm{   \epsilon_j(X,\rho,\rho^{(i)})  }{2}^2}{ f_j(\rho^{(i)},X)} \Bigg] \\ 
	  &\leq \frac{64 m^2}{L^2 \xi^3 } \sum_{j=0}^m   \E_{j}^{\rho}\Bigg[  \Bigg(  \frac{ \absbig{     q_{\pi}^\rho(X) -   q_{\pi^{(i)}}^{\rho^{(i)}}(X) }  }{   q_{\pi  }^{\rho }(X)  }\frac{q_{\pi}^{\rho}(X)^{1/2}   }{q_{\pi^{(i)}}^{\rho^{(i)}}(X)^{1/2}  }        \Bigg)^2   \Bigg] +  \frac{64 m^2}{L^2 \xi^3 } \sum_{j=0}^m        \absbig{\pi_j^{(i)}   -  \pi_j  }^2  + \frac{2m}{\xi^3} \sum_{j=0}^m \E_{j}\Bigg[  \frac{\bignorm{   \epsilon_j(X,\rho,\rho^{(i)})  }{2}^2}{ f_j(\rho^{(i)},X)} \Bigg] \\ 
	  &= \frac{64 m^2}{L^2 \xi^3 } \sum_{j=0}^m   \E_{j}^{\rho}\Bigg[  \Bigg(  \frac{      q_{\pi}^\rho(X) -   q_{\pi^{(i)}}^{\rho^{(i)}}(X)   }{   q_{\pi  }^{\rho }(X)  }\Bigg)^2 \frac{q_{\pi}^{\rho}(X)    }{q_{\pi^{(i)}}^{\rho^{(i)}}(X)   }         \Bigg] +  \frac{64 m^2}{L^2 \xi^3 } \norm{\pi^{(i)} - \pi}{2}^2     + \frac{2m}{\xi^3} \sum_{j=0}^m \E_{j}\Bigg[  \frac{\bignorm{   \epsilon_j(X,\rho,\rho^{(i)})  }{2}^2}{ f_j(\rho^{(i)},X)} \Bigg] \\ 
	 &\leq \frac{64 m^2}{L^2 \xi^4 } \sum_{j=0}^m   \E_{j}^{\rho}\Bigg[  \Bigg(  \frac{      q_{\pi}^\rho(X) -   q_{\pi^{(i)}}^{\rho^{(i)}}(X)   }{   q_{\pi  }^{\rho }(X)  }\Bigg)^2 \frac{q_{\pi}^{\rho}(X)    }{ q_{ j }^{\rho^{(i)}}(X)   }         \Bigg] +  \frac{64 m^2}{L^2 \xi^3 } \norm{\pi^{(i)} - \pi}{2}^2     + \frac{2m}{\xi^3} \sum_{j=0}^m \E_{j}\Bigg[  \frac{\bignorm{   \epsilon_j(X,\rho,\rho^{(i)})  }{2}^2}{ f_j(\rho^{(i)},X)} \Bigg] \\ 
	 &= \frac{64 m^2}{L^2 \xi^4 } \sum_{j=0}^m   \E_{\pi}^{\rho}\Bigg[  \Bigg(  \frac{      q_{\pi}^\rho(X) -   q_{\pi^{(i)}}^{\rho^{(i)}}(X)   }{   q_{\pi  }^{\rho }(X)  }\Bigg)^2 \frac{ q_{j}^{\rho}(X)    }{ q_{ j }^{\rho^{(i)}}(X)   }         \Bigg] +  \frac{64 m^2}{L^2 \xi^3 } \norm{\pi^{(i)} - \pi}{2}^2     + \frac{2m}{\xi^3} \sum_{j=0}^m \E_{j}\Bigg[  \frac{\bignorm{   \epsilon_j(X,\rho,\rho^{(i)})  }{2}^2}{ f_j(\rho^{(i)},X)} \Bigg] \\
	 &= \frac{64 m^2}{L^2 \xi^4 }   \E_{\pi}^{\rho}\Bigg[  \Bigg(  \frac{      q_{\pi}^\rho(X) -   q_{\pi^{(i)}}^{\rho^{(i)}}(X)   }{   q_{\pi  }^{\rho }(X)  }\Bigg)^2 \sum_{j=0}^m  \frac{ f_j(\rho;X)    }{ f_{ j }(\rho^{(i)};X)   }         \Bigg] +  \frac{64 m^2}{L^2 \xi^3 } \norm{\pi^{(i)} - \pi}{2}^2     + \frac{2m}{\xi^3} \sum_{j=0}^m \E_{j}\Bigg[  \frac{\bignorm{   \epsilon_j(X,\rho,\rho^{(i)})  }{2}^2}{ f_j(\rho^{(i)},X)} \Bigg].
\end{align*}
\normalsize

As stated earlier, the $f_j(\rho;x)$ are uniformly bounded away from $0$ and $\infty$, for all $x \in \X$ and $\rho \in \R$. Combined with the inequality in display \eqref{high polymer}, this implies that the first expectation in the last line displayed above converges to $0$ as $\theta^{(i)} \to \theta$. Consequentially, to show that $  \int  d W_2(z,\theta,\theta^{(i)}) \  d\widetilde{\mu}(z) \to 0$ as $\theta^{(i)} \to \theta$, it suffices to show that $\E_j\Big[ \bignorm{   \epsilon_j(X,\rho,\rho^{(i)})  }{2}^2 \Big] \to 0$ as $\theta^{(i)} \to \theta$. 

With that goal in mind,  note that $\partialDerivative{\rho}\{ f_j(\rho;x) \}: \R  \mapsto \mathbb{R}^m$ is a differentiable, vector-valued function and that $\R \subset \mathbb{R}^m$ is an open convex set. Thus, by the mean value theorem for vector-valued functions proved in \cite{MVTspecial}, we have that:
\begin{align*}
	\bignorm{   \epsilon_j(X,\rho,\rho^{(i)})  }{2} &= \biggnorm{  \partialDerivative{\rho}\{ f_j(\rho^{(i)};x) \} -  \partialDerivative{\rho}\{ f_j(\rho;x) \}   }{2} \\
	&\leq \norm{\rho^{(i)} - \rho }{2} \sup_{0 < t < 1} \maxSingBigg{ \bigg\{ \frac{\delta^2}{ \delta a \delta b}  f_j\big( (1-t)\rho + t\rho^{(i)};x\big) \bigg\}_{ 1\leq a,b\leq m}  }.
\end{align*}
Using the fact that $\norm{ (1-t)\rho + t\rho^{(i)} }{2} < B <\infty$  and $\norm{s_{\gamma^\ast(\tau)} (x)}{2} \leq \frac{2\sqrt{m}}{\xi} < \infty$ for all $x\in\X$, one can show that the singular values of the Hessian matrix of  $ f_j\big( (1-t)\rho + t\rho^{(i)};x\big)$ are bounded, uniformly over all choices of $x\in\X$ and $\rho,\rho^{(i)} \in \R$. Hence, $\E_j\Big[ \bignorm{   \epsilon_j(X,\rho,\rho^{(i)})  }{2}^2 \Big] $ is upper bounded by a constant times $ \norm{\rho^{(i)} - \rho }{2}$, meaning that $\E_j\Big[ \bignorm{   \epsilon_j(X,\rho,\rho^{(i)})  }{2}^2 \Big] \to 0$ as $\theta^{(i)} \to \theta$. Thus, we have shown that  $  \int  d W_2(z,\theta,\theta^{(i)}) \  d\widetilde{\mu}(z) \to 0$ as $\theta^{(i)} \to \theta$.

Finally, we show that $\int  (1-d) W_3(z,\theta,\theta^{(i)})  \ d\widetilde{\mu}(z)$ converges to 0. Towards that end, observe that:
\begin{align*}
	 &W_3(z,\theta,\theta^{(i)})\\
	  &= \Bignorm{ \big(s_{\gamma^\ast(\tau)}(x) - \E_y^{\rho^{(i)}}[s_{\gamma^\ast(\tau)}]\big) q_y^{\rho^{(i)}}(x)^{1/2}    -  \big(s_{\gamma^\ast(\tau)}(x) - \E_y^{\rho}[s_{\gamma^\ast(\tau)}]\big) q_y^\rho(x)^{1/2}   }{2}^2 \\ \\
	 &\leq  2\Bignorm{ \big(s_{\gamma^\ast(\tau)}(x) - \E_y^{\rho^{(i)}}[s_{\gamma^\ast(\tau)}]\big) q_y^{\rho^{(i)}}(x)^{1/2}    -   \big(s_{\gamma^\ast(\tau)}(x) - \E_y^{\rho^{(i)}}[s_{\gamma^\ast(\tau)}]\big) q_y^{\rho}(x)^{1/2}   }{2}^2 \\
	 &\qquad + 2\Bignorm{ \big(s_{\gamma^\ast(\tau)}(x) - \E_y^{\rho^{(i)}}[s_{\gamma^\ast(\tau)}]\big) q_y^{\rho}(x)^{1/2}   -  \big(s_{\gamma^\ast(\tau)}(x) - \E_y^{\rho}[s_{\gamma^\ast(\tau)}]\big) q_y^\rho(x)^{1/2}  }{2}^2 \\ \\ 
	 &=  2\Bignorm{ s_{\gamma^\ast(\tau)}(x) - \E_y^{\rho^{(i)}}[s_{\gamma^\ast(\tau)}]     }{2}^2 \Big(q_y^{\rho^{(i)}}(x)^{1/2} -   q_y^{\rho}(x)^{1/2}\Big)^2 + 2q_y^\rho(x) \Bignorm{  \E_y^{\rho^{(i)}}[s_{\gamma^\ast(\tau)}]  -     \E_y^{\rho}[s_{\gamma^\ast(\tau)}]    }{2}^2 \\ 
	 &\leq  \frac{32 m }{L^2}  \Big(q_y^{\rho^{(i)}}(x)^{1/2} -   q_y^{\rho}(x)^{1/2}\Big)^2 + 2q_y^\rho(x) \Bignorm{  \E_y^{\rho^{(i)}}[s_{\gamma^\ast(\tau)}]  -     \E_y^{\rho}[s_{\gamma^\ast(\tau)}]    }{2}^2 \\ 
	 &=  \frac{32 m }{L^2} q_y^{\rho}(x)  \Bigg( \frac{ q_y^{\rho^{(i)}}(x)^{1/2} -   q_y^{\rho}(x)^{1/2}}{q_y^{\rho}(x)^{1/2}}\Bigg)^2 + 2q_y^\rho(x) \Bignorm{  \E_y^{\rho^{(i)}}[s_{\gamma^\ast(\tau)}]  -     \E_y^{\rho}[s_{\gamma^\ast(\tau)}]    }{2}^2.
\end{align*}
Ergo:
\begin{align*}
	&\int  (1-d) W_3(z,\theta,\theta^{(i)})  \ d\widetilde{\mu}(z) \\
	&\leq  \frac{32 m }{L^2} \sum_{y=0}^m \E_y^{\rho}\Bigg[ \Bigg( \frac{ q_y^{\rho^{(i)}}(x)^{1/2} -   q_y^{\rho}(x)^{1/2}}{q_y^{\rho}(x)^{1/2}}\Bigg)^2 \Bigg]       +        2\sum_{y=0}^m  \Bignorm{  \E_y^{\rho^{(i)}}[s_{\gamma^\ast(\tau)}]  -     \E_y^{\rho}[s_{\gamma^\ast(\tau)}]    }{2}^2 .
\end{align*}

Note that:
\small
\begin{align*}
	&\Bignorm{  \E_y^{\rho^{(i)}}[s_{\gamma^\ast(\tau)}]  -     \E_y^{\rho}[s_{\gamma^\ast(\tau)}]    }{2}^2  \\
	&=  \biggnorm{  \E_y\Big[f_y(\rho^{(i)};X)s_{\gamma^\ast(\tau)}(X)\Big]  -     \E_y\Big[ f_y(\rho;X)  s_{\gamma^\ast(\tau)}(X) \Big]    }{2}^2 \\
	&= \biggnorm{  \E_y\Big[ \big(f_y(\rho^{(i)};X) -  f_y(\rho;X)\big)  s_{\gamma^\ast(\tau)}(X) \Big]    }{2}^2  \\ 
	&\leq   \E_y\Big[  \bignorm{ \big(f_y(\rho^{(i)};X) -  f_y(\rho;X)\big)  s_{\gamma^\ast(\tau)}(X)    }{2}^2\Big]   \\ 
	&=   \E_y\Big[\big(f_y(\rho^{(i)};X) -  f_y(\rho;X)\big)^2  \bignorm{ s_{\gamma^\ast(\tau)}(X)  }{2}^2\Big]   \\ 
	&\leq \frac{4m}{L^2}  \E_y\Big[\big(f_y(\rho^{(i)};X) -  f_y(\rho;X)\big)^2  \Big] \\ 
	&= \frac{4m}{L^2}  \E_y\Bigg[\Bigg(     f_y(\rho;X) \frac{\pi_j^\ast}{\pi_j^\train} \bigdotprod{ s_{\gamma^\ast(\tau)}(X)- \E_{y}^{\rho}[s_{\gamma^\ast(\tau)}]  }{\rho^{(i)} - \rho} + \frac{1}{2}(\rho^{(i)} - \rho)' \partialSecondDerivative{\rho}f_y(\bar{\rho}_{X,y};X)(\rho^{(i)} - \rho)\Bigg)^2  \Bigg]  \\ 
	&\leq \frac{8m}{L^2}  \E_y\Bigg[     f_y(\rho;X)^2 \frac{(\pi_j^\ast)^2}{(\pi_j^\train)^2} \norm{ s_{\gamma^\ast(\tau)}(X)- \E_{y}^{\rho}[s_{\gamma^\ast(\tau)}]  }{2}^2 \norm{\rho^{(i)} - \rho}{2}^2 + \frac{1}{4} \maxEvalBigg{ \partialSecondDerivative{\rho}f_y(\bar{\rho}_{X,y};X) }^2  \norm{\rho^{(i)} - \rho}{2}^4  \Bigg]  \\ 
	&\leq \frac{8m}{L^2}  \E_y\Bigg[  \frac{16m}{L^2\xi^2}    f_y(\rho;X)^2  \norm{\rho^{(i)} - \rho}{2}^2 + \frac{1}{4} \maxEvalBigg{ \partialSecondDerivative{\rho}f_y(\bar{\rho}_{X,y};X) }^2  \norm{\rho^{(i)} - \rho}{2}^4  \Bigg], 
\end{align*}
\normalsize
which, by our earlier arguments, converges to $0$ as $\theta^{(i)} \to \theta$. In addition, note that:
\begin{align*}
	 \E_y^{\rho}\Bigg[ \Bigg( \frac{ q_y^{\rho^{(i)}}(x)^{1/2} -   q_y^{\rho}(x)^{1/2} }{q_y^{\rho}(x)^{1/2}}\Bigg)^2 \Bigg] &=  \E_y^{\rho}\Bigg[ \Bigg( \frac{ q_y^{\rho^{(i)}}(x) -   q_y^{\rho}(x)}{q_y^{\rho}(x)^{1/2} \big(q_y^{\rho^{(i)}}(x)^{1/2} +   q_y^{\rho}(x)^{1/2}\big)}\Bigg)^2 \Bigg] \\ 
	 &\leq  \E_y^{\rho}\Bigg[ \Bigg( \frac{ q_y^{\rho^{(i)}}(x) -   q_y^{\rho}(x)}{q_y^{\rho}(x) }\Bigg)^2 \Bigg].
\end{align*}
By line \eqref{purple castle}, the RHS converges to $0$ as $\theta^{(i)} \to \theta$. Thus, overall, we have that $\int  (1-d) W_3(z,\theta,\theta^{(i)})  \ d\widetilde{\mu}(z)  \to 0$ as $\theta^{(i)} \to\theta$.
And, having shown that all three integrals (involving $W_1,W_2$ and $W_3$) converge to $0$, we may conclude that $\int \norm{D(z,\theta^{(i)}) - D(z,\theta)}{2}^2 d\widetilde{\mu}(z)  \to 0$. Thus, we have verified property (b)!

\uline{Finally, let's prove property (a)}. Observe that:
\begin{align*}
	\int \norm{D(z,\theta)}{2}^2 d\widetilde{\mu}(z) &= \int \norm{D(z,\pi;\rho)}{2}^2 d\widetilde{\mu}(z) + \int \norm{D(z,\rho;\pi)}{2}^2 d\widetilde{\mu}(z) \\ 
	&\leq  \int d q_{\pi}^{\rho}(x)\norm{ s_{\pi}^{\rho}(x)}{2}^2 d\widetilde{\mu}(z)  +  \int \norm{D(z,\rho;\pi)}{2}^2 d\widetilde{\mu}(z)  \\
	&\leq \frac{4m}{\xi^2}  +  \int \norm{D(z,\rho;\pi)}{2}^2 d\widetilde{\mu}(z).
\end{align*}

Now, note that:
\small
\begin{align*}
	&\norm{D(z,\rho;\pi)}{2} \\
	&\leq d\sqrt{\tau} \frac{1}{\xi}\sum_{j=0}^m \pi_j \frac{ q_j^{\rho}(x) }{ 2\sqrt{q_{\pi}^\rho(x)} }  \bignorm{ s_{\gamma^\ast(\tau)}(x) - \E_j^{\rho}[s_{\gamma^\ast(\tau)}]}{2} + (1-d)\sqrt{(1-\tau) \pi_y^{\train}} \frac{1}{2}\sqrt{q_y^\rho(x)}  \frac{1}{\xi} \norm{s_{\gamma^\ast(\tau)}(x) - \E_y^{\rho}[s_{\gamma^\ast(\tau)}]}{2} \\
	&\leq d  \frac{1}{\xi}\sum_{j=0}^m  \frac{ q_j^{\rho}(x) }{ \sqrt{q_{\pi}^\rho(x)} }  \bignorm{ s_{\gamma^\ast(\tau)}(x) - \E_j^{\rho}[s_{\gamma^\ast(\tau)}]}{2} + (1-d)\sqrt{q_y^\rho(x)}  \frac{1}{\xi}\norm{s_{\gamma^\ast(\tau)}(x) - \E_y^{\rho}[s_{\gamma^\ast(\tau)}]}{2} \\ 
	&\leq d  \frac{4\sqrt{m}}{L\xi} \sum_{j=0}^m  \frac{ q_j^{\rho}(x) }{ \sqrt{q_{\pi}^\rho(x)} } + (1-d) \frac{4\sqrt{m}}{L\xi} \sqrt{q_y^\rho(x)} \\ 
	&\leq d  \frac{4\sqrt{m}}{L\xi} \sum_{j=0}^m  \frac{ q_j^{\rho}(x) }{ \sqrt{ \pi_j q_{j}^\rho(x) } } + (1-d) \frac{4\sqrt{m}}{L\xi} \sqrt{q_y^\rho(x)} \\ 
	&= d  \frac{4\sqrt{m}}{L\xi^{3/2}} \sum_{j=0}^m  \sqrt{ q_{j}^\rho(x) } + (1-d) \frac{4\sqrt{m}}{L\xi} \sqrt{q_y^\rho(x)}.
\end{align*}
\normalsize
Thus:
\begin{align*}
	\norm{D(z,\rho;\pi)}{2}^2 &\leq  d  \frac{32 m}{L^2 \xi^3} \bigg(\sum_{j=0}^m  \sqrt{ q_{j}^\rho(x) }\bigg)^2 + (1-d) \frac{32 m}{L^2\xi^2} q_y^\rho(x) \\
	 &\leq  d  \frac{32 m^2}{L^2 \xi^3} \sum_{j=0}^m  q_{j}^\rho(x)  + (1-d) \frac{32 m}{L^2\xi^2} q_y^\rho(x).
\end{align*}
Ergo:
\begin{align*}
	\int \norm{D(z,\rho;\pi)}{2}^2 d\widetilde{\mu}(z) &\leq \int  \bigg[      d  \frac{32 m^2}{L^2 \xi^3} \sum_{j=0}^m  q_{j}^\rho(X)  + (1-d) \frac{32 m}{L^2\xi^2} q_y^\rho(X)      \bigg] d\widetilde{\mu}(z) \\ 
	&\leq        \frac{32 m^3}{L^2 \xi^3}   + \frac{32 m^2}{L^2\xi^2}.
\end{align*}
Thus, overall, 
\begin{align*}
	\int \norm{D(z,\theta)}{2}^2 d\widetilde{\mu}(z) &\leq \frac{4m}{\xi^2}  +  \frac{32 m^3}{L^2 \xi^3}   + \frac{32 m^2}{L^2\xi^2} \\ 
	&< \infty,
\end{align*}
meaning that property (a) is true! 

We have therefore verified properties (a), (b) and (c). Finally, it remains for us to prove the score function identities in the statement of this lemma. By   Definition A.1 of \citep{Newey1990}, the score function $S_{\pi,\rho}^{\tau}$ for the parameters $(\pi,\rho)$ is given by:
\begin{align*}
	S_{\pi,\rho}^{\tau}(z) &= 2 \frac{D(z,\pi,\rho)}{\Jcal^{\pi,\rho,\tau}(z)^{1/2}} \Ibig{\Jcal^{\pi,\rho,\tau}(z) > 0} \\ 
	&= 2 \frac{ \arraycolsep=1.4pt\def\arraystretch{1}  \begin{pmatrix}  D(z,\pi;\rho) \\ D(z,\rho;\pi) \end{pmatrix}    }{\Jcal^{\pi,\rho,\tau}(z)^{1/2}} \Ibig{\Jcal^{\pi,\rho,\tau}(z) > 0}.
\end{align*}
Note that:
\begin{align*}
	2 \frac{  D(z,\pi;\rho) }{\Jcal^{\pi,\rho,\tau}(z)^{1/2}} \Ibig{\Jcal^{\pi,\rho,\tau}(z) > 0} &= 2 \frac{ \frac{d}{2} \sqrt{\tau q_{\pi}^{\rho}(x)} s_{\pi}^{\rho}(x) }{\Jcal^{\pi,\rho,\tau}(z)^{1/2}} \Ibig{\Jcal^{\pi,\rho,\tau}(z) > 0}  \\
	 &=  \frac{ d\sqrt{\tau q_{\pi}^{\rho}(x)} s_{\pi}^{\rho}(x) }{ \sqrt{ \tau q_{\pi}^{\rho}(x) } } \Ibig{ \tau q_{\pi}^{\rho}(x) > 0}  \\
	 &=  d s_{\pi}^{\rho}(x)  \Ibig{  q_{\pi}^{\rho}(x) > 0}.
\end{align*}

Also note that:
\smaller
\begin{align*}
	 &\frac{ 2D(z,\rho;\pi)   }{\Jcal^{\pi,\rho,\tau}(z)^{1/2}} \Ibig{\Jcal^{\pi,\rho,\tau}(z) > 0}\\
	 	&=  \frac{   d\sqrt{\tau} \sum_{j=0}^m \frac{ \pi_j q_j^{\rho}(x) }{ \sqrt{q_{\pi}^\rho(x)} }  \frac{\pi_j^\ast}{\pi_j^\train}(s_{\gamma^\ast(\tau)}(x) - \E_j^{\rho}[s_{\gamma^\ast(\tau)}]) + (1-d)\sqrt{(1-\tau) \pi_y^{\train}q_y^\rho(x)}\frac{\pi_y^\ast}{\pi_y^\train} \big(s_{\gamma^\ast(\tau)}(x) - \E_y^{\rho}[s_{\gamma^\ast(\tau)}]\big)      }{\Jcal^{\pi,\rho,\tau}(z)^{1/2}} \Ibig{\Jcal^{\pi,\rho,\tau}(z) > 0} \\
	 &= d \sum_{j=0}^m \pi_j\frac{\pi_j^\ast}{\pi_j^\train} \frac{  q_j^{\rho}(x) }{ q_{\pi}^\rho(x) }  (s_{\gamma^\ast(\tau)}(x) - \E_j^{\rho}[s_{\gamma^\ast(\tau)}])\Ibig{q_{\pi}^\rho(x) > 0} + (1-d)  \frac{\pi_y^\ast}{\pi_y^\train}\big(s_{\gamma^\ast(\tau)}(x) - \E_y^{\rho}[s_{\gamma^\ast(\tau)}]\big)\Ibig{q_y^\rho(x) > 0}.     
\end{align*}
\normalsize

Ergo:
\smaller
\begin{align*}
	&S_{\pi,\rho}^{\tau}(z) \\
	&= \arraycolsep=1.4pt\def\arraystretch{1.5}\begin{pmatrix}
		   d s_{\pi}^{\rho}(x)  \Ibig{  q_{\pi}^{\rho}(x) > 0} \\ 
		   d \sum\limits_{j=0}^m \pi_j \frac{\pi_j^\ast}{\pi_j^\train} \frac{  q_j^{\rho}(x) }{ q_{\pi}^\rho(x) }  (s_{\gamma^\ast(\tau)}(x) - \E_j^{\rho}[s_{\gamma^\ast(\tau)}])\Ibig{q_{\pi}^\rho(x) > 0} + (1-d)  \sum\limits_{j=0}^{m} \I{y=j}\frac{\pi_j^\ast}{\pi_j^\train}\big(s_{\gamma^\ast(\tau)}(x) - \E_j^{\rho}[s_{\gamma^\ast(\tau)}] \big)\Ibig{q_j^\rho(x) > 0}
	\end{pmatrix},
\end{align*}
\normalsize
as claimed.

\end{proof}
\vspace{0.4in}

\begin{proof}[\uline{Proof of Lemma \eqref{Nuisance Tangent Set is a Subset of G}}]

First, we restate the definition of the semiparametric nuisance tangent set from  \citep{Newey1990}, but in the context of our fixed $\tau$ IID regime. The semiparametric nuisance tangent set  $\T$ is defined as: 
\small
\begin{align}
	\T &:= \bigg\{ t: \mathcal{X}\times \mathcal{Y}\times\{0,1\} \mapsto \mathbb{R}^m \text{ } \bigg| \text{ } \E_{ \Jcal^{\pi^\ast,\mathbf{p},\tau} }\big[ \norm{t(Z)}{2}^2 \big] < \infty, \nonumber\\
	&\qquad\qquad\qquad\text{ } \exists \{ B^{(k)} \},\{ \mathbf{M}^{\textup{sub},\tau,(k)} \} \quad \text{s.t.}\quad  \lim_{k\to\infty} \E_{ \Jcal^{\pi^\ast,\mathbf{p},\tau} }\big[ \norm{ t(Z) - B^{(k)} S_{\rho^\ast}^{\tau,(k)}(Z) }{2}^2 \big] = 0 \bigg\}.  \label{Definition of Nuisance Tangent Set}
\end{align}
\normalsize
In the definition above, $\mathbf{M}^{\textup{sub},\tau,(k)}$ denotes a smooth (in the sense of \citep{Newey1990}) parametric submodel (of $\mathbf{M}^{\text{semi},\tau}$) containing densities $\Jcal^{\pi,\rho,\tau,(k)}$, $\rho$ is a finite dimensional vector that is used by $\mathbf{M}^{\textup{sub},\tau,(k)}$ to parameterize a subset of $\Q$ in a fashion that ensures $\exists \rho^\ast$ such that $\Jcal^{\pi^\ast,\rho^\ast,\tau,(k)} = \Jcal^{\pi^\ast,\mathbf{p},\tau} $, $B^{(k)}$ is a conformable matrix with $m$ rows, and $S_{\rho}^{\tau,(k)}$ is the score function for $\rho$ evaluated at $(\pi^\ast,\rho)$.  Note that 
\[
	\mathcal{J}^{\pi,\rho,\tau,(k)}(z) \equiv \big( \tau q_{\pi}^{\rho,(k)}(x) \big)^d \Bigg((1-\tau) \prod_{j=0}^{m} \big(\pi_j^\train q_j^{\rho,(k)}(x) \big)^{\I{y=j}} \Bigg)^{1-d} \Ibig{x\in\X, \text{ } y \in \Y, \text{ } d \in \{0,1\}  },
\]
where $(q_0^{\rho,(k)},q_1^{\rho,(k)},\dots,q_m^{\rho,(k)}) \in \Q$ depends $\rho$.  Also, for any given smooth parametric submodel $\mathbf{M}^{\textup{sub},\tau,(k)}$,  one can use arguments similar to those employed in the proof of Lemma \eqref{Smooth Score Function for Pi} to show that $S_{\rho}^{\tau,(k)} =  \partialDerivative{\rho} \log \mathcal{J}^{\pi^\ast,\rho,\tau,(k)}(z)$ a.s. $[\mathcal{J}^{\pi^\ast,\rho,\tau,(k)}]$. Now, observe that: 
\begin{align*}
	&\partialDerivative{\rho} \log \Jcal^{\pi^\ast,\rho,\tau,(k)}(z) \\
	&= \partialDerivative{\rho} \log \Bigg\{ \big(\tau q_{\pi^\ast}^{\rho,(k)} (x) \big)^d \Big((1-\tau)\prod_{j=0}^{m}\big(\pi_j^\train q_{j}^{\rho,(k)} (x)\big)^{\I{y=j}}\Big)^{(1-d)} \Bigg\} \\ 
	&= \partialDerivative{\rho} \Bigg\{d \log(\tau) + d\log q_{\pi^\ast}^{\rho,(k)}(x) + (1-d)\log(1-\tau) + (1-d) \sum_{j=0}^{m} \I{y=j}\Big[\log(\pi_j^\train) + \log q_{j}^{\rho,(k)}(x)\Big]\Bigg\} \\
	&= \partialDerivative{\rho} \Bigg\{d\log q_{\pi^\ast}^{\rho,(k)}(x) + (1-d) \sum_{j=0}^{m} \I{y=j} \log q_{j}^{\rho,(k)}(x) \Bigg\} \\
	&=  d \partialDerivative{\rho} \log q_{\pi^\ast}^{\rho,(k)}(x) + (1-d) \sum_{j=0}^{m} \I{y=j}  \partialDerivative{\rho}\log q_{j}^{\rho,(k)}(x) \\ 
	&=  d \frac{\sum_{j=0}^{m}\pi_j^\ast  \partialDerivative{\rho} q_{j}^{\rho,(k)}(x) }{q_{\pi^\ast}^{\rho,(k)}(x) } + (1-d) \sum_{j=0}^{m} \I{y=j}  \partialDerivative{\rho}\log q_{j}^{\rho,(k)}(x) \\
	&=  d \frac{\sum_{j=0}^{m}\pi_j^\ast q_{j}^{\rho,(k)}(x) \partialDerivative{\rho} \log q_{j}^{\rho,(k)}(x)}{q_{\pi^\ast}^{\rho,(k)}(x) } + (1-d) \sum_{j=0}^{m} \I{y=j}  \partialDerivative{\rho}\log q_{j}^{\rho,(k)}(x) \\ 
	&=  d \sum_{j=0}^{m} \pi_j^\ast \frac{q_{j}^{\rho,(k)}(x)}{q_{\pi^\ast}^{\rho,(k)}(x) }\partialDerivative{\rho} \log q_{j}^{\rho,(k)}(x) + (1-d) \sum_{j=0}^{m} \I{y=j}  \partialDerivative{\rho}\log q_{j}^{\rho,(k)}(x).
\end{align*}
Ergo, the following holds a.s. $[\mathcal{J}^{\pi^\ast,\rho^\ast,\tau}]$:
\begin{align*}
	S_{\rho^\ast}(z) &=d \sum_{j=0}^{m} \pi_j^\ast \frac{p_{j}(x)}{p_{\pi^\ast}(x) }\partialDerivative{\rho} \big\{ \log q_{j}^{\rho,(k)}(x) \big\}\big\lvert_{\rho=\rho^\ast}+ (1-d) \sum_{j=0}^{m} \I{y=j}  \partialDerivative{\rho}\big\{\log q_{j}^{\rho,(k)}(x)\big\}\big\lvert_{\rho=\rho^\ast}.
\end{align*}

Now, we want to prove that $\T \subseteq\G$. So, let any $t\in\mathcal{T}$ be given. We will prove that $t \in \G$. Towards that end, let $\{B^{(k)} \}, \{ \mathbf{M}^{\textup{sub},\tau,(k)}  \}$ be the matrices and submodels used in the the definition of $t$, respectively. Observe that:
\begin{align*}
	t(z) &\equiv t(x,y,d) \\ 
	&= d t(x,y,1) + (1-d)t(x,y,0) \\
	&= d t(x,y,1) + (1-d)\sum_{j=0}^m \I{y=j} t(x,j,0).
\end{align*}

So, the difference $t - B^{(k)} S_{\rho^\ast}^{\tau,(k)}$ can be written as follows:

\begin{align*}
	 &t(z) - B^{(k)} S_{\rho^\ast}^{\tau,(k)}(z) \\
	 &\equiv t(x,y,d) - B^{(k)} S_{\rho^\ast}^{\tau,(k)}(x,y,d) \\
	 &= t(x,y,d) - d \sum_{j=0}^{m} \pi_j^\ast \frac{p_{j}(x)}{p_{\pi^\ast}(x) } B^{(k)}  \partialDerivative{\rho} \big\{ \log q_{j}^{\rho,(k)}(x) \big\}\big\lvert_{\rho=\rho^\ast}+ (1-d) \sum_{j=0}^{m} \I{y=j}  B^{(k)} \partialDerivative{\rho}\big\{\log q_{j}^{\rho,(k)}(x)\big\}\big\lvert_{\rho=\rho^\ast} \\ \\
	 &= d\Bigg[t(x,y,1) -   \sum_{j=0}^{m} \pi_j^\ast \frac{p_{j}(x)}{p_{\pi^\ast}(x) } B^{(k)}  \partialDerivative{\rho} \big\{ \log q_{j}^{\rho,(k)}(x) \big\}\big\lvert_{\rho=\rho^\ast} \Bigg] \\ 
	 &\qquad\qquad + (1-d) \sum_{j=0}^m  \I{y=j} \Bigg[ t(x,j,0) - B^{(k)}  \partialDerivative{\rho} \big\{ \log q_{j}^{\rho,(k)}(x) \big\}\big\lvert_{\rho=\rho^\ast} \Bigg].
\end{align*}
Due to the indicators $d$ and $\I{y=j}$, we therefore have that:
\begin{align*}
	&\Bignorm{t(z) - B^{(k)} S_{\rho^\ast}^{\tau,(k)}(z) }{2}^2 \\
	&= d\biggnorm{ t(x,y,1) -   \sum_{j=0}^{m} \pi_j^\ast \frac{p_{j}(x)}{p_{\pi^\ast}(x) } B^{(k)}  \partialDerivative{\rho} \big\{ \log q_{j}^{\rho,(k)}(x) \big\}\big\lvert_{\rho=\rho^\ast} }{2}^2 \\ 
	&\qquad\qquad+ (1-d)  \sum_{j=0}^m  \I{y=j}   \biggnorm{t(x,j,0) - B^{(k)}  \partialDerivative{\rho} \big\{ \log q_{j}^{\rho,(k)}(x) \big\}\big\lvert_{\rho=\rho^\ast}  }{2}^2,
\end{align*}
and so, taking expectations w.r.t. the true distribution $\Jcal^{\pi^\ast,\mathbf{p},\tau}$:
\begin{align}
	&\E_{\Jcal^{\pi^\ast,\mathbf{p},\tau}}\Bigg[\Bignorm{t(Z) - B^{(k)} S_{\rho^\ast}^{\tau,(k)}(Z) }{2}^2\Bigg] \nonumber\\ 
	&= \tau \E_{\Jcal^{\pi^\ast,\mathbf{p},\tau} \mid D = 1}\Bigg[ \biggnorm{ t(X,Y,1) -   \sum_{j=0}^{m} \pi_j^\ast \frac{p_{j}(X)}{p_{\pi^\ast}(X) } B^{(k)}  \partialDerivative{\rho} \big\{ \log q_{j}^{\rho,(k)}(X) \big\}\big\lvert_{\rho=\rho^\ast} }{2}^2 \Bigg] \nonumber \\ 
	&\qquad + (1-\tau)\sum_{j=0}^{m}\pi_j^\train \E_{ \Jcal^{\pi^\ast,\mathbf{p},\tau} \mid D = 0, Y = j}\Bigg[ \biggnorm{t(X,j,0) - B^{(k)}  \partialDerivative{\rho} \big\{ \log q_{j}^{\rho,(k)}(X) \big\}\big\lvert_{\rho=\rho^\ast}  }{2}^2 \Bigg] \nonumber \\ \nonumber \\
	&= \tau \E_{\pi^\ast}\Bigg[ \biggnorm{ t(X,Y,1) -   \sum_{j=0}^{m} \pi_j^\ast \frac{p_{j}(X)}{p_{\pi^\ast}(X) } B^{(k)}  \partialDerivative{\rho} \big\{ \log q_{j}^{\rho,(k)}(X) \big\}\big\lvert_{\rho=\rho^\ast} }{2}^2 \Bigg] \nonumber \\ 
	&\qquad + (1-\tau)\sum_{j=0}^{m}\pi_j^\train \E_{j}\Bigg[ \biggnorm{t(X,j,0) - B^{(k)}  \partialDerivative{\rho} \big\{ \log q_{j}^{\rho,(k)}(X) \big\}\big\lvert_{\rho=\rho^\ast}  }{2}^2 \Bigg]. \label{expected values that will go to zero}
\end{align}
Now, since $t \in \T$, we know that the sequences $\{B^{(k)}\}$ and $ \{ \mathbf{M}^{\textup{sub},\tau,(k)}  \}$ satisfy $\E_{ \Jcal^{\pi^\ast,\mathbf{p},\tau} }\Bigg[\Bignorm{t(Z) - B^{(k)} S_{\rho^\ast}^{\tau,(k)}(Z) }{2}^2\Bigg] \to 0$ as $k\to\infty$. Therefore,  by display \eqref{expected values that will go to zero}, it must be that $\lim_{k\to\infty} \E_{j}\Bigg[ \biggnorm{t(X,j,0) - B^{(k)}  \partialDerivative{\rho} \big\{ \log q_{j}^{\rho,(k)}(X) \big\}\big\lvert_{\rho=\rho^\ast}  }{2}^2 \Bigg] =  0$  for each class $j$, and  $\lim_{k\to\infty}\E_{\pi^\ast}\Bigg[ \biggnorm{ t(X,Y,1) -   \sum_{j=0}^{m} \pi_j^\ast \frac{p_{j}(X)}{p_{\pi^\ast}(X) } B^{(k)}  \partialDerivative{\rho} \big\{ \log q_{j}^{\rho,(k)}(X) \big\}\big\lvert_{\rho=\rho^\ast} }{2}^2 \Bigg]  = 0$ as well. Now, observe that, for all $k$:
\smaller
\[
	\E_{\pi^\ast}\Bigg[ \biggnorm{ t(X,Y,1) -  \sum_{j=0}^{m} \pi_j^\ast \frac{p_{j}(X)}{p_{\pi^\ast}(X) }  t(X,j,0) }{2} \Bigg] 
\]

\[
	\leq \E_{\pi^\ast}\Bigg[ \biggnorm{ t(X,Y,1) -  \sum_{j=0}^{m} \pi_j^\ast \frac{p_{j}(X)}{p_{\pi^\ast}(X) }  B^{(k)} \partialDerivative{\rho} \big\{ \log q_{j}^{\rho,(k)}(X) \big\}\big\lvert_{\rho=\rho^\ast}   }{2} \Bigg] 
\]
\[
	\qquad\qquad  +   \sum_{j=0}^{m} \pi_j^\ast \E_{\pi^\ast}\Bigg[ \frac{p_{j}(X)}{p_{\pi^\ast}(X) } \biggnorm{   B^{(k)} \partialDerivative{\rho} \big\{ \log q_{j}^{\rho,(k)}(X) \big\}\big\lvert_{\rho=\rho^\ast} -   t(X,j,0) }{2} \Bigg] 
\]

\[
		= \E_{\pi^\ast}\Bigg[ \biggnorm{ t(X,Y,1) -  \sum_{j=0}^{m} \pi_j^\ast \frac{p_{j}(X)}{p_{\pi^\ast}(X) }  B^{(k)} \partialDerivative{\rho} \big\{ \log q_{j}^{\rho,(k)}(X) \big\}\big\lvert_{\rho=\rho^\ast}   }{2} \Bigg]  +  \sum_{j=0}^{m} \pi_j^\ast \E_{j}\Bigg[  \biggnorm{   B^{(k)} \partialDerivative{\rho} \big\{ \log q_{j}^{\rho,(k)}(X) \big\}\big\lvert_{\rho=\rho^\ast} -   t(X,j,0) }{2} \Bigg].
\]
\normalsize
By our argument earlier, the two summands in the display above converge to zero as $k\to\infty$, so since the inequality holds for all $k$, it must be that $\E_{\pi^\ast}\Bigg[ \biggnorm{ t(X,Y,1) -  \sum_{j=0}^{m} \pi_j^\ast \frac{p_{j}(X)}{p_{\pi^\ast}(X) }  t(X,j,0) }{2} \Bigg]  = 0$. Thus, we may say that:
\[
	t(x,y,1) =  \sum_{j=0}^{m} \pi_j^\ast \frac{p_{j}(x)}{p_{\pi^\ast}(x) }  t(x,j,0) \quad \text{ a.s. } [p_{\pi^\ast}],
\]
meaning that:
\[
	t(z) = d  \sum_{j=0}^{m} \pi_j^\ast \frac{p_{j}(x)}{p_{\pi^\ast}(x) }  t(x,j,0)  + (1-d)\sum_{j=0}^m \I{y=j} t(x,j,0)  \quad \text{ a.s. } [\Jcal^{\pi^\ast,\mathbf{p},\tau} ].
\]
So, to show that $t \in \G$, it remains to show that $\E_{j}[t(X,j,0)] = 0$ and $\E_{j}\Big[\bignorm{t(X,j,0)}{2}^2\Big] < \infty $  for each class $j$. The latter criteria is clearly true, because the fact that $t\in\T$ means that $\E_{\Jcal^{\pi^\ast,\mathbf{p},\tau}}\big[ \norm{t(Z)}{2}^2 \big] < \infty$, which since $\tau < 1$ and $\xi \leq \pi_j^\train \leq 1-\xi$, can in turn only be true if $\E_{j}\Big[\bignorm{t(X,j,0)}{2}^2\Big] < \infty $ for each $j$. To see that $\E_{j}[t(X,j,0)] = 0$, observe that:
\begin{align*}
	 \Bignorm{  \E_{j}[t(X,j,0) ]   }{2}^2 &= \Bignorm{  \E_{j}[t(X,j,0) ]   - 0 }{2}^2  \\ 
	 &= \Biggnorm{  \E_{j}[t(X,j,0) ]   - \E_{j}\Big[ B^{(k)}  \partialDerivative{\rho} \big\{ \log q_{j}^{\rho,(k)}(X) \big\}\big\lvert_{\rho=\rho^\ast}  \Big] }{2}^2  \\ 
	 &\leq  \E_{j}\Bigg[  \biggnorm{ t(X,j,0) -  B^{(k)}  \partialDerivative{\rho} \big\{ \log q_{j}^{\rho,(k)}(X) \big\}\big\lvert_{\rho=\rho^\ast} }{2}^2 \Bigg].
\end{align*}
Since the inequality above holds for all $k$ and the last line goes to zero as $k\to\infty$, it must be that $ \E_{j}[t(X,j,0) ]  = 0$, as desired. Thus, we have shown that $t \in \G$. Hence, $\T \subseteq \G$, as claimed. \\

\end{proof}
\vspace{0.4in}

\begin{proof}[\uline{Proof of Lemma \eqref{Projection of Ordinary Score onto G}}]
First, we will prove the eigenvalue bounds on $\mathbf{V}^\textup{eff}(\tau)$. Note that  $\pi^\ast,\pi^\train \in \Delta$, which means that $\gamma^\ast(\tau)\in \Gamma$ by virtue of  Lemma \eqref{Bounded Gamma Star} (this can be seen by setting $a = \frac{1}{\tau}$ and $b = \frac{1}{1-\tau}$ in the statement of Lemma \eqref{Bounded Gamma Star}). Thus, by Assumption \eqref{Fixed Tau Regime: Mixture FIM not too small}, the matrix $\FisherInfo{ \gamma^\ast(\tau) }$ is positive definite, and so its inverse in the definition of $\mathbf{V}^\textup{eff}(\tau)$ is well defined. We start with the upper bound:
\begin{align*}
	&\maxEval{\mathbf{V}^\textup{eff}(\tau)}\\
	&\leq  \bigg[ \frac{1}{\tau} + \frac{1}{1-\tau} \sum_{k=0}^{m} \frac{(\pi_k^\ast)^2}{\pi_k^\train} \bigg] \maxEvalBig{  \FisherInfo{ \gamma^\ast(\tau) }^{-1} - \FisherInfo{\gamma^\ast(\tau);\textup{Cat}}^{-1} } + \frac{1}{\tau}\maxEvalBig{\FisherInfo{\pi^\ast;\textup{Cat}}^{-1}  } \\ 
	&\leq  \bigg[ \frac{1}{\tau} + \frac{1}{1-\tau} \sum_{k=0}^{m} \frac{(\pi_k^\ast)^2}{\pi_k^\train} \bigg] \maxEvalBig{  \FisherInfo{ \gamma^\ast(\tau) }^{-1} } + \frac{1}{\tau}\maxEvalBig{\FisherInfo{\pi^\ast;\textup{Cat}}^{-1}  } \\ 
	&=  \bigg[ \frac{1}{\tau} + \frac{1}{1-\tau} \sum_{k=0}^{m} \frac{(\pi_k^\ast)^2}{\pi_k^\train} \bigg] \frac{1}{\minEval{\FisherInfo{ \gamma^\ast(\tau) }}} + \frac{1}{\tau} \frac{1}{ \minEval{ \FisherInfo{\pi^\ast;\textup{Cat}}  } } \\ 
	&\leq \bigg[ \frac{1}{\tau} + \frac{1}{1-\tau} \sum_{k=0}^{m} \frac{(\pi_k^\ast)^2}{\pi_k^\train} \bigg] \frac{1}{ \sqrt{\Lambda} } + \frac{1}{2\tau} \\ 
	&\leq \bigg[ \frac{1}{\tau} + \frac{1}{(1-\tau)\xi}  \bigg] \frac{1}{ \sqrt{\Lambda} } + \frac{1}{2\tau}.
\end{align*}
The second line is because $\gamma^\ast(\tau) \in \Gamma$, so the components of $\gamma^\ast(\tau)$ are all bounded away from $0$ and $1$, and so by Lemma \eqref{Bounds on Eigenvalues of Categorical Fisher Info}, $\FisherInfo{\gamma^\ast(\tau);\textup{Cat}}^{-1}$ is positive definite. The fourth line is by Assumption \eqref{Fixed Tau Regime: Mixture FIM not too small} and Lemma  \eqref{Bounds on Eigenvalues of Categorical Fisher Info}, and the fifth line is because $\pi^\ast,\pi^\train \in \Delta$. As for the lower bound:
\begin{align*}
	&\minEval{\mathbf{V}^\textup{eff}(\tau)}\\
	 &\geq \bigg[ \frac{1}{\tau} + \frac{1}{1-\tau} \sum_{k=0}^{m} \frac{(\pi_k^\ast)^2}{\pi_k^\train} \bigg] \minEvalBig{  \FisherInfo{ \gamma^\ast(\tau) }^{-1} - \FisherInfo{\gamma^\ast(\tau);\textup{Cat}}^{-1} }  + \frac{1}{\tau}\minEvalBig{\FisherInfo{\pi^\ast;\textup{Cat}}^{-1}  } \\ 
	&\geq \bigg[ \frac{1}{\tau} + \frac{1}{1-\tau} \sum_{k=0}^{m} \frac{(\pi_k^\ast)^2}{\pi_k^\train} \bigg] \minEvalBig{  \FisherInfo{ \gamma^\ast(\tau) }^{-1} - \FisherInfo{\gamma^\ast(\tau);\textup{Cat}}^{-1} } \\ 
	&\geq (\xi)^2\bigg[ \frac{1}{\tau} + \frac{1}{1-\tau}   \bigg] \minEvalBig{  \FisherInfo{ \gamma^\ast(\tau) }^{-1} - \FisherInfo{\gamma^\ast(\tau);\textup{Cat}}^{-1} } \\ 
	&=  \frac{ (\xi)^2}{\tau(1-\tau)}  \minSingBig{  \FisherInfo{ \gamma^\ast(\tau) }^{-1} - \FisherInfo{\gamma^\ast(\tau);\textup{Cat}}^{-1} } \\ 
	&=  \frac{ (\xi)^2}{\tau(1-\tau)}  \minSingBig{   \FisherInfo{\gamma^\ast(\tau)}^{-1}   \big[  \FisherInfo{\gamma^\ast(\tau);\textup{Cat}} - \FisherInfo{ \gamma^\ast(\tau) }  \big]   \FisherInfo{\gamma^\ast(\tau);\text{Cat}}^{-1}    } \\
	&\geq  \frac{ (\xi)^2}{\tau(1-\tau)}  \minSing{\FisherInfo{\gamma^\ast(\tau)}^{-1} } \minSing{       \FisherInfo{\gamma^\ast(\tau);\textup{Cat}} - \FisherInfo{ \gamma^\ast(\tau) }       }  \minSing{\FisherInfo{\gamma^\ast(\tau);\text{Cat}}^{-1}  }  \\ 
	&\geq  \frac{ (\xi)^2 \nu}{\tau(1-\tau)}  \minSing{\FisherInfo{\gamma^\ast(\tau)}^{-1} }  \minSing{\FisherInfo{\gamma^\ast(\tau);\text{Cat}}^{-1}  }  \\ 
	&=  \frac{ (\xi)^2 \nu }{\tau(1-\tau)} \frac{1}{\maxSing{\FisherInfo{\gamma^\ast(\tau)} }  \maxSing{\FisherInfo{\gamma^\ast(\tau);\text{Cat}}  }   }   \\ 
	&\geq  \frac{ (\xi)^2 \nu }{\tau(1-\tau)} \frac{1}{   \maxSing{\FisherInfo{\gamma^\ast(\tau);\text{Cat}}  }^2   }   \\ 
	&\geq  \frac{ (\xi)^2 \nu }{\tau(1-\tau)}  \bigg(\frac{L}{m+1}\bigg)^2.
\end{align*}

The second line is by Lemma \eqref{Bounds on Eigenvalues of Categorical Fisher Info} and the fact that $\pi^\ast \in \Delta$. The fourth line is because each component of $\gamma^\ast(\tau)$ is  bounded away from $0$ and $1$, and so by Lemma \eqref{Harder Problem}, the matrix $ \FisherInfo{ \gamma^\ast(\tau) }^{-1} - \FisherInfo{\gamma^\ast(\tau);\textup{Cat}}^{-1}$ is symmetric positive semidefinite. The seventh, ninth and tenth line all use the fact that $\gamma^\ast(\tau) \in \Gamma$; in addition, the seventh, ninth and tenth lines use  Assumption \eqref{Fixed Tau Regime: Cat minus Mixture FIM}, Lemma \eqref{Harder Problem} and Lemma  \eqref{Bounds on Eigenvalues of Categorical Fisher Info}, respectively.



Second, we need to prove that $ds_{\pi^\ast} -  \mathbf{V}^\text{eff}(\tau)^{-1}\psi_\tau^\textup{eff} \in \G$. To do that, we will show that the function
\[
	\bar{g}(z) := d \sum_{j=0}^m \pi_j^\ast \frac{p_j(x)}{p_{\pi^\ast}(x) }\bar{f}_j(x) + (1-d)\sum_{j=0}^{m} \I{y=j} \bar{f}_j(x)
\]
where
\[
	\bar{f}_j(x) := \frac{1}{1-\tau} \frac{\pi_j^\ast}{\pi_j^\train} \mathbf{V}^\text{eff}(\tau)^{-1}\FisherInfo{\gamma^\ast(\tau)}^{-1}\big(s_{\gamma^\ast(\tau)}(x) - \E_j[s_{\gamma^\ast(\tau)}]\big)
\]
satisfies $\bar{g} \in \G$, and furthermore, that  $\bar{g}(z) = ds_{\pi^\ast}(x) -  \mathbf{V}^\text{eff}(\tau)^{-1}\psi_\tau^\textup{eff}(z)$. \\

To see that $\bar{g} \in \G$, first observe that  $\E_j[\bar{f}_j] = 0$. One can also argue that $\E_j\big[  \norm{ \bar{f}_j}{2}^2  \big] < \infty$, in the following manner. By the argument above, both $\mathbf{V}^\text{eff}(\tau)$ and  $\FisherInfo{\gamma^\ast(\tau)}$ are symmetric positive definite. Therefore, it follows that:
\begin{align*}
	\maxEval{ \mathbf{V}^\text{eff}(\tau)^{-1}\FisherInfo{\gamma^\ast(\tau)}^{-1} } &=\maxEvalBig{ \Big(\FisherInfo{\gamma^\ast(\tau)} \mathbf{V}^\text{eff}(\tau)  \Big)^{-1}} \\ 
	&= \frac{1}{\minEval{ \FisherInfo{\gamma^\ast(\tau)} \mathbf{V}^\text{eff}(\tau) }} \\
	&\leq \frac{1}{\minEval{ \FisherInfo{\gamma^\ast(\tau)} } \minEval{\mathbf{V}^\text{eff}(\tau) }}   \\
	&\leq \frac{1 }{ \sqrt{\Lambda}  \frac{ (\xi)^2 \nu }{\tau(1-\tau)}  \big(\frac{L}{m+1}\big)^2 }   \\
	& < \infty,
\end{align*}
where the fourth line is by Assumption \eqref{Fixed Tau Regime: Mixture FIM not too small} and the lower bound on $\minEval{\mathbf{V}^\text{eff}(\tau) }$ that we just established.  Now, since $\norm{s_{\gamma^\ast(\tau)}(x) }{ 2}^2$ is uniformly bounded over all $x\in\X$, the fact that $\maxEval{ \mathbf{V}^\text{eff}(\tau)^{-1}\FisherInfo{\gamma^\ast(\tau)}^{-1} } < \infty$ means that $ \norm{ \bar{f}_j(x)}{2}^2$ is also uniformly bounded, too. Ergo, it must be that $\E_j\big[  \norm{ \bar{f}_j}{2}^2  \big] < \infty$, as claimed. This shows that $\bar{g} \in \G$. 

Having shown that $\bar{g} \in \G$, we now need to verify that $\bar{g}(z)$ equals $ ds_{\pi^\ast}(x) -  \mathbf{V}^\text{eff}(\tau)^{-1}\psi_\tau^\textup{eff}(z)$. Towards that end, note that  
\begin{align*}
	\bar{g}(z)  &\equiv  d \sum_{j=0}^m \pi_j^\ast \frac{p_j(x)}{p_{\pi^\ast}(x) }\bar{f}_j(x) + (1-d)\sum_{j=0}^{m} \I{y=j} \bar{f}_j(x) \\ 
	&= d \sum_{j=0}^m \pi_j^\ast \frac{p_j(x)}{p_{\pi^\ast}(x) }\bar{f}_j(x) + \frac{1-d}{1-\tau}\mathbf{V}^\text{eff}(\tau)^{-1}\FisherInfo{\gamma^\ast(\tau)}^{-1} \sum_{j=0}^{m} \I{y=j}   \frac{\pi_j^\ast}{\pi_j^\train} \big(s_{\gamma^\ast(\tau)}(x) - \E_j[s_{\gamma^\ast(\tau)}]\big).
\end{align*}

Thus:
\tiny
\[
	\bar{g}(z) =  ds_{\pi^\ast}(x) -  \mathbf{V}^\text{eff}(\tau)^{-1}\psi_\tau^\textup{eff}(z)
\]
\[
	\iff \bar{g}(z) =  ds_{\pi^\ast}(x) - \mathbf{V}^\text{eff}(\tau)^{-1}\Bigg(  \frac{d}{\tau} \FisherInfo{\gamma^\ast(\tau)}^{-1}\big(s_{\gamma^\ast(\tau)}(x)  -  \E_{\pi^\ast}[s_{\gamma^\ast(\tau)}]\big) - \frac{1-d}{1-\tau} \FisherInfo{\gamma^\ast(\tau)}^{-1} \sum_{j=0}^m \I{y=j} \frac{\pi_j^\ast}{\pi_j^\train}\big(s_{\gamma^\ast(\tau)}(x)  -  \E_{j}[s_{\gamma^\ast(\tau)}]\big)  \Bigg)
\]
\begin{equation}\label{verify spongebob}
	\iff  d \sum_{j=0}^m \pi_j^\ast \frac{p_j(x)}{p_{\pi^\ast}(x) }\bar{f}_j(x) =  ds_{\pi^\ast}(x) - \mathbf{V}^\text{eff}(\tau)^{-1}\Bigg(  \frac{d}{\tau} \FisherInfo{\gamma^\ast(\tau)}^{-1}\big(s_{\gamma^\ast(\tau)}(x)  -  \E_{\pi^\ast}[s_{\gamma^\ast(\tau)}]\big)  \Bigg).
\end{equation}
\normalsize

So, to prove that $\bar{g}(z) =  ds_{\pi^\ast}(x) -  \mathbf{V}^\text{eff}(\tau)^{-1}\psi_\tau^\textup{eff}(z)$,  we need to verify that display \eqref{verify spongebob} holds. Towards that end, observe that:
\small
\begin{align*}
	&\sum_{j=0}^m \pi_j^\ast \frac{p_j(x)}{p_{\pi^\ast}(x) }\bar{f}_j(x)\\
	 &= \mathbf{V}^\text{eff}(\tau)^{-1}\FisherInfo{\gamma^\ast(\tau)}^{-1} \sum_{j=0}^m \frac{1}{1-\tau} \frac{(\pi_j^\ast)^2}{\pi_j^\train}  \frac{p_j(x)}{p_{\pi^\ast}(x) }   \big(s_{\gamma^\ast(\tau)}(x) - \E_j[s_{\gamma^\ast(\tau)}]\big)  \\
	&=  \mathbf{V}^\text{eff}(\tau)^{-1}\FisherInfo{\gamma^\ast(\tau)}^{-1}s_{\gamma^\ast(\tau)}(x) \sum_{j=0}^m \frac{1}{1-\tau} \frac{(\pi_j^\ast)^2}{\pi_j^\train}   \frac{p_j(x) }{p_{\pi^\ast} (x)} - \mathbf{V}^\text{eff}(\tau)^{-1}\FisherInfo{\gamma^\ast(\tau)}^{-1}\sum_{j=0}^m \frac{1}{1-\tau} \frac{(\pi_j^\ast)^2}{\pi_j^\train}   \frac{p_j(x) }{p_{\pi^\ast} (x)} \E_j[s_{\gamma^\ast(\tau)}].
\end{align*}
\normalsize

Note that:
\begin{align*}
	 \sum_{j=0}^m \frac{1}{1-\tau} \frac{(\pi_j^\ast)^2}{\pi_j^\train}   \frac{p_j(x) }{p_{\pi^\ast} (x)} &=  \sum_{j=0}^m \bigg[ \frac{1}{1-\tau} \frac{(\pi_j^\ast)^2}{\pi_j^\train}  +  \frac{ \pi_j^\ast}{\tau} \bigg] \frac{p_j(x) }{p_{\pi^\ast} (x)} -  \sum_{j=0}^m \bigg[ \frac{ \pi_j^\ast}{\tau}  \bigg]\frac{p_j(x) }{p_{\pi^\ast} (x)} \\
	 &=  \bigg[ \frac{1}{1-\tau} \sum_{j=0}^m \frac{(\pi_j^\ast)^2}{\pi_j^\train}  +  \frac{1}{\tau} \bigg]  \sum_{j=0}^m \gamma_j^\ast(\tau) \frac{p_j(x) }{p_{\pi^\ast} (x)} - \frac{1}{\tau}  \sum_{j=0}^m  \pi_j^\ast  \frac{p_j(x) }{p_{\pi^\ast} (x)} \\
	 &=  \bigg[ \frac{1}{1-\tau} \sum_{j=0}^m \frac{(\pi_j^\ast)^2}{\pi_j^\train}  +  \frac{1}{\tau} \bigg]   \frac{p_{\gamma^\ast(\tau)}(x) }{p_{\pi^\ast} (x)} - \frac{1}{\tau}.
\end{align*}

Thus:
\begin{align}
	\sum_{j=0}^m \pi_j^\ast \frac{p_j(x)}{p_{\pi^\ast}(x) }\bar{f}_j(x) &=  \mathbf{V}^\text{eff}(\tau)^{-1}\FisherInfo{\gamma^\ast(\tau)}^{-1}s_{\gamma^\ast(\tau)}(x) \Bigg(  \bigg[ \frac{1}{1-\tau} \sum_{j=0}^m \frac{(\pi_j^\ast)^2}{\pi_j^\train}  +  \frac{1}{\tau} \bigg]   \frac{p_{\gamma^\ast(\tau)}(x) }{p_{\pi^\ast} (x)} - \frac{1}{\tau} \Bigg)  \nonumber \\
	&\qquad\qquad- \mathbf{V}^\text{eff}(\tau)^{-1}\FisherInfo{\gamma^\ast(\tau)}^{-1}\sum_{j=0}^m \frac{1}{1-\tau} \frac{(\pi_j^\ast)^2}{\pi_j^\train}   \frac{p_j(x) }{p_{\pi^\ast} (x)} \E_j[s_{\gamma^\ast(\tau)}] \nonumber \\ \nonumber \\ 
	&=  \mathbf{V}^\text{eff}(\tau)^{-1}\FisherInfo{\gamma^\ast(\tau)}^{-1} \Bigg(  \bigg[ \frac{1}{1-\tau} \sum_{j=0}^m \frac{(\pi_j^\ast)^2}{\pi_j^\train}  +  \frac{1}{\tau} \bigg]   s_{\pi^\ast}(x) - \frac{s_{\gamma^\ast(\tau)}(x)}{\tau} \Bigg) \nonumber \\
	&\qquad\qquad- \mathbf{V}^\text{eff}(\tau)^{-1}\FisherInfo{\gamma^\ast(\tau)}^{-1}\sum_{j=0}^m \frac{1}{1-\tau} \frac{(\pi_j^\ast)^2}{\pi_j^\train}   \frac{p_j(x) }{p_{\pi^\ast} (x)} \E_j[s_{\gamma^\ast(\tau)}] \label{repair man}
\end{align}

Next, we will focus on the term $\sum_{j=0}^m \frac{1}{1-\tau} \frac{(\pi_j^\ast)^2}{\pi_j^\train}   \frac{p_j(x) }{p_{\pi^\ast} (x)} \E_j[s_{\gamma^\ast(\tau)}]$. Towards that end, note that:
\begin{align*}
	0 &= \bigg[ \frac{1}{1-\tau} \sum_{j=0}^m \frac{(\pi_j^\ast)^2}{\pi_j^\train}  +  \frac{1}{\tau} \bigg] \E_{\gamma^\ast(\tau)}[s_{\gamma^\ast(\tau)}] \\ 
	&=  \bigg[ \frac{1}{1-\tau} \sum_{j=0}^m \frac{(\pi_j^\ast)^2}{\pi_j^\train}  +  \frac{1}{\tau} \bigg] \sum_{j=0}^m \gamma_j^\ast(\tau) \E_{j}[s_{\gamma^\ast(\tau)}] \\
	&= \sum_{j=0}^m\bigg[ \frac{1}{1-\tau}  \frac{(\pi_j^\ast)^2}{\pi_j^\train}  +  \frac{\pi_j^\ast}{\tau} \bigg] \E_{j}[s_{\gamma^\ast(\tau)}] \\ 
	&=  \sum_{j=0}^m  \frac{1}{1-\tau} \frac{(\pi_j^\ast)^2}{\pi_j^\train}  \E_{j}[s_{\gamma^\ast(\tau)}] +  \frac{1}{\tau} \sum_{j=0}^m \pi_j^\ast \E_{j}[s_{\gamma^\ast(\tau)}] \\
	&=  \sum_{j=0}^m   \frac{1}{1-\tau}  \frac{(\pi_j^\ast)^2}{\pi_j^\train}  \E_{j}[s_{\gamma^\ast(\tau)}] +  \frac{1}{\tau}  \E_{\pi^\ast}[s_{\gamma^\ast(\tau)}],
\end{align*}
Thus:
\begin{align}
	&\sum_{j=0}^m \frac{1}{1-\tau} \frac{(\pi_j^\ast)^2}{\pi_j^\train}   \frac{p_j(x) }{p_{\pi^\ast} (x)} \E_j[s_{\gamma^\ast(\tau)}] \\
	&= \sum_{j=0}^m \frac{1}{1-\tau} \frac{(\pi_j^\ast)^2}{\pi_j^\train}   \frac{p_j(x) }{p_{\pi^\ast} (x)} \E_j[s_{\gamma^\ast(\tau)}] - 0 \nonumber \\ 
	 &= \sum_{j=0}^m \frac{1}{1-\tau} \frac{(\pi_j^\ast)^2}{\pi_j^\train}   \frac{p_j(x) }{p_{\pi^\ast} (x)} \E_j[s_{\gamma^\ast(\tau)}] -   \sum_{j=0}^m   \frac{1}{1-\tau}  \frac{(\pi_j^\ast)^2}{\pi_j^\train}  \E_{j}[s_{\gamma^\ast(\tau)}] -  \frac{1}{\tau}  \E_{\pi^\ast}[s_{\gamma^\ast(\tau)}] \nonumber \\ 
	  &= \sum_{j=0}^m \frac{1}{1-\tau} \frac{(\pi_j^\ast)^2}{\pi_j^\train}   \frac{p_j(x) - p_{\pi^\ast} (x)}{p_{\pi^\ast} (x)} \E_j[s_{\gamma^\ast(\tau)}] -  \frac{1}{\tau}  \E_{\pi^\ast}[s_{\gamma^\ast(\tau)}]. \label{Are you the straw to my berry?}
\end{align}

We will now show that
\small
\[
	 \sum_{j=0}^m \frac{1}{1-\tau} \frac{(\pi_j^\ast)^2}{\pi_j^\train}   \frac{p_j(x) - p_{\pi^\ast} (x)}{p_{\pi^\ast} (x)} \E_j[s_{\gamma^\ast(\tau)}] 
\]
\[
	= \Bigg(\bigg[ \frac{1}{\tau} + \frac{1}{1-\tau} \sum_{k=0}^{m} \frac{(\pi_k^\ast)^2}{\pi_k^\train} \bigg] \FisherInfo{\gamma^\ast(\tau)} \FisherInfo{\gamma^\ast(\tau);\textup{Cat}}^{-1} - \frac{1}{\tau} \FisherInfo{\gamma^\ast(\tau)} \FisherInfo{\pi^\ast;\textup{Cat}}^{-1}\Bigg)s_{\pi^\ast}(x).
\]
\normalsize
Towards that end, it will be helpful to first derive the form of $\FisherInfo{\gamma^\ast(\tau)} \FisherInfo{\beta;\textup{Cat}}^{-1}$ for arbitrary $\beta$. The $\numth{(i,j)}$ entry of this matrix product is equal to:
\small
\begin{align*}
	&\Big[ \FisherInfo{\gamma^\ast(\tau)} \FisherInfo{\beta;\textup{Cat}}^{-1} \Big]_{ij} \\
	&= \Bigdotprod{\textup{row } i \textup{ of } \FisherInfo{\gamma^\ast(\tau)} }{\textup{ col } j \textup{ of } \FisherInfo{\beta;\textup{Cat}}^{-1} } \\
	&= \Bigdotprod{\big( \E_1[s_{\gamma^\ast(\tau),i}] - \E_0[s_{\gamma^\ast(\tau),i}] ,\dots,   \E_m[s_{\gamma^\ast(\tau),i}] - \E_0[s_{\gamma^\ast(\tau),i}] \big) }{ \big(-\beta_1\beta_j,\dots,\beta_j(1-\beta_j),\dots,-\beta_m \beta_j\big) } \\ 
	&= -\beta_j \Bigdotprod{\big( \E_1[s_{\gamma^\ast(\tau),i}] - \E_0[s_{\gamma^\ast(\tau),i}] ,\dots,   \E_m[s_{\gamma^\ast(\tau),i}] - \E_0[s_{\gamma^\ast(\tau),i}] \big) }{ \big(\beta_1,\dots,\beta_j-1,\dots,\beta_m \big) } \\ 
	&= -\beta_j \Bigg( \sum_{k=1}^m\beta_k\big( \E_k[s_{\gamma^\ast(\tau),i}] - \E_0[s_{\gamma^\ast(\tau),i}]\big) - \big( \E_j[s_{\gamma^\ast(\tau),i}] - \E_0[s_{\gamma^\ast(\tau),i}]\big)  \Bigg) \\ 
	&= -\beta_j \Bigg( \sum_{k=1}^m\beta_k\E_k[s_{\gamma^\ast(\tau),i}] - \bigg( \sum_{k=1}^m\beta_k\bigg)\E_0[s_{\gamma^\ast(\tau),i}] - \E_j[s_{\gamma^\ast(\tau),i}] + \E_0[s_{\gamma^\ast(\tau),i}]  \Bigg) \\ 
	&= -\beta_j \Bigg( \sum_{k=1}^m\beta_k\E_k[s_{\gamma^\ast(\tau),i}] + \bigg(1- \sum_{k=1}^m\beta_k\bigg)\E_0[s_{\gamma^\ast(\tau),i}] - \E_j[s_{\gamma^\ast(\tau),i}]  \Bigg) \\ 
	&= -\beta_j \big( \E_\beta[s_{\gamma^\ast(\tau),i}] - \E_j[s_{\gamma^\ast(\tau),i}]  \big).
\end{align*}
\normalsize
Hence, the $\numth{j}$ column of $\FisherInfo{\gamma^\ast(\tau)} \FisherInfo{\beta;\textup{Cat}}^{-1}$ is given by:
\begin{align*}
	\implies \Big[\FisherInfo{\gamma^\ast(\tau)} \FisherInfo{\beta;\textup{Cat}}^{-1}\Big]_{\cdot,j} = -\beta_j \big( \E_\beta[s_{\gamma^\ast(\tau)}] - \E_j[s_{\gamma^\ast(\tau)}]  \big).
\end{align*}
Since this holds for general $\beta$, it follows that:
\begin{align*}
	&\Bigg[ \bigg[\frac{1}{\tau} + \frac{1}{1-\tau} \sum_{k=0}^m \frac{(\pi_k^\ast)^2}{\pi_k^\train} \bigg] \FisherInfo{\gamma^\ast(\tau)} \FisherInfo{\gamma^\ast(\tau);\textup{Cat}}^{-1}\Bigg]_{\cdot,j} \\
	&= -\bigg[\frac{1}{\tau} + \frac{1}{1-\tau} \sum_{k=0}^m \frac{(\pi_k^\ast)^2}{\pi_k^\train} \bigg]\gamma_j^\ast(\tau) \big( \E_{\gamma^\ast(\tau)}[s_{\gamma^\ast(\tau)}] - \E_j[s_{\gamma^\ast(\tau)}]  \big) \\ 
	&= \bigg[\frac{1}{\tau} + \frac{1}{1-\tau} \sum_{k=0}^m \frac{(\pi_k^\ast)^2}{\pi_k^\train} \bigg]\gamma_j^\ast(\tau) \E_j[s_{\gamma^\ast(\tau)}]  \\ 
	&= \bigg[\frac{\pi_j^\ast}{\tau} + \frac{1}{1-\tau} \frac{(\pi_j^\ast)^2}{\pi_j^\train} \bigg]\E_j[s_{\gamma^\ast(\tau)}] \\ 
	&= \frac{\pi_j^\ast}{\tau}\E_j[s_{\gamma^\ast(\tau)}] + \frac{1}{1-\tau} \frac{(\pi_j^\ast)^2}{\pi_j^\train} \E_j[s_{\gamma^\ast(\tau)}]. 
\end{align*}
And also that:
\begin{align*}
	\frac{1}{\tau} \Big[ \FisherInfo{\gamma^\ast(\tau)}\FisherInfo{\pi^\ast;\textup{Cat}}^{-1} \Big]_{\cdot,j} &= -\frac{\pi_j^\ast}{\tau} \big( \E_{\pi^\ast}[s_{\gamma^\ast(\tau)}] -\E_j[s_{\gamma^\ast(\tau)}]\big) \\ 
	&=  -\frac{\pi_j^\ast}{\tau}  \E_{\pi^\ast}[s_{\gamma^\ast(\tau)}]  + \frac{\pi_j^\ast}{\tau}\E_j[s_{\gamma^\ast(\tau)}].
\end{align*}
Ergo:
\begin{align*}
	&\Bigg[ \bigg[\frac{1}{\tau} + \frac{1}{1-\tau} \sum_{k=0}^m \frac{(\pi_k^\ast)^2}{\pi_k^\train} \bigg] \FisherInfo{\gamma^\ast(\tau)} \FisherInfo{\gamma^\ast(\tau);\textup{Cat}}^{-1} - \frac{1}{\tau}\FisherInfo{\gamma^\ast(\tau)}\FisherInfo{\pi^\ast;\textup{Cat}}^{-1}  \Bigg]_{\cdot,j} \\
	&= \frac{1}{1-\tau} \frac{(\pi_j^\ast)^2}{\pi_j^\train} \E_j[s_{\gamma^\ast(\tau)}] + \frac{\pi_j^\ast}{\tau}  \E_{\pi^\ast}[s_{\gamma^\ast(\tau)}].
\end{align*}
Now, note that:
\begin{align*}
	p_{\pi^\ast}(x) - p_0(x) &= \bigg(1- \sum_{j=1}^m\pi_j^\ast\bigg) p_0(x) + \sum_{j=1}^m \pi_j^\ast p_j(x) - p_0(x) \\ 
	&=  - \sum_{j=1}^m\pi_j^\ast p_0(x) + \sum_{j=1}^m \pi_j^\ast p_j(x)  \\ 
	&=  \sum_{j=1}^m \pi_j^\ast \big(p_j(x)-p_0(x)\big),
\end{align*}
meaning that $\frac{p_{\pi^\ast}(x) - p_0(x) }{p_{\pi^\ast}(x)} =  \sum_{j=1}^m \pi_j^\ast s_{\pi^\ast,j}(x)$. Combining this with our matrix results, it follows that:
\small
\[
	\Bigg[ \bigg[\frac{1}{\tau} + \frac{1}{1-\tau} \sum_{k=0}^m \frac{(\pi_k^\ast)^2}{\pi_k^\train} \bigg] \FisherInfo{\gamma^\ast(\tau)} \FisherInfo{\gamma^\ast(\tau);\textup{Cat}}^{-1} - \frac{1}{\tau}\FisherInfo{\gamma^\ast(\tau)}\FisherInfo{\pi^\ast;\textup{Cat}}^{-1}  \Bigg] s_{\pi^\ast}(x)
\]
\[
	= \sum_{j=1}^m \bigg(\frac{1}{1-\tau} \frac{(\pi_j^\ast)^2}{\pi_j^\train} \E_j[s_{\gamma^\ast(\tau)}] + \frac{\pi_j^\ast}{\tau}  \E_{\pi^\ast}[s_{\gamma^\ast(\tau)}]\bigg) s_{\pi^\ast,j}(x)
\]
\[
	=  \sum_{j=1}^m \frac{1}{1-\tau} \frac{(\pi_j^\ast)^2}{\pi_j^\train} \E_j[s_{\gamma^\ast(\tau)}]s_{\pi^\ast,j}(x)   +  \frac{\E_{\pi^\ast}[s_{\gamma^\ast(\tau)}] }{\tau}   \sum_{j=1}^m \pi_j^\ast s_{\pi^\ast,j}(x)
\]
\[
	= \sum_{j=1}^m \frac{1}{1-\tau} \frac{(\pi_j^\ast)^2}{\pi_j^\train} \E_j[s_{\gamma^\ast(\tau)}]s_{\pi^\ast,j}(x)   +  \frac{\E_{\pi^\ast}[s_{\gamma^\ast(\tau)}] }{\tau}  \frac{p_{\pi^\ast}(x) - p_0(x)}{p_{\pi^\ast}(x)} 
\]
\[
	= \sum_{j=1}^m \frac{1}{1-\tau} \frac{(\pi_j^\ast)^2}{\pi_j^\train} \E_j[s_{\gamma^\ast(\tau)}] \frac{p_j(x) - p_{\pi^\ast}(x)}{p_{\pi^\ast}(x)} + \sum_{j=1}^m \frac{1}{1-\tau} \frac{(\pi_j^\ast)^2}{\pi_j^\train} \E_j[s_{\gamma^\ast(\tau)}] \frac{p_{\pi^\ast}(x) - p_0(x)}{p_{\pi^\ast}(x)}   +  \frac{\E_{\pi^\ast}[s_{\gamma^\ast(\tau)}] }{\tau}  \frac{p_{\pi^\ast}(x) - p_0(x)}{p_{\pi^\ast}(x)} 
\]
\[
	= \sum_{j=1}^m \frac{1}{1-\tau} \frac{(\pi_j^\ast)^2}{\pi_j^\train} \E_j[s_{\gamma^\ast(\tau)}] \frac{p_j(x) - p_{\pi^\ast}(x)}{p_{\pi^\ast}(x)} + \bigg(\sum_{j=1}^m \frac{1}{1-\tau} \frac{(\pi_j^\ast)^2}{\pi_j^\train} \E_j[s_{\gamma^\ast(\tau)}]   +  \frac{\E_{\pi^\ast}[s_{\gamma^\ast(\tau)}] }{\tau}\bigg) \frac{p_{\pi^\ast}(x) - p_0(x)}{p_{\pi^\ast}(x)}. 
\]
\normalsize
Now, note that:
\begin{align*}
	&\sum_{j=1}^m \frac{1}{1-\tau} \frac{(\pi_j^\ast)^2}{\pi_j^\train} \E_j[s_{\gamma^\ast(\tau)}]   +  \frac{\E_{\pi^\ast}[s_{\gamma^\ast(\tau)}] }{\tau}\\
	 &= \sum_{j=1}^m \frac{1}{1-\tau} \frac{(\pi_j^\ast)^2}{\pi_j^\train} \E_j[s_{\gamma^\ast(\tau)}]   +  \frac{1}{\tau} \sum_{j=0}^m \pi_j^\ast\E_{j}[s_{\gamma^\ast(\tau)}]  \\ 
	&= \sum_{j=0}^m \frac{1}{1-\tau} \frac{(\pi_j^\ast)^2}{\pi_j^\train} \E_j[s_{\gamma^\ast(\tau)}]   +  \frac{1}{\tau} \sum_{j=0}^m \pi_j^\ast\E_{j}[s_{\gamma^\ast(\tau)}]  - \frac{1}{1-\tau} \frac{(\pi_0^\ast)^2}{\pi_0^\train} \E_0[s_{\gamma^\ast(\tau)}] \\ 
	&= \sum_{j=0}^m \bigg(\frac{1}{1-\tau} \frac{(\pi_j^\ast)^2}{\pi_j^\train}   +  \frac{\pi_j^\ast}{\tau}\bigg)  \E_{j}[s_{\gamma^\ast(\tau)}]  - \frac{1}{1-\tau} \frac{(\pi_0^\ast)^2}{\pi_0^\train} \E_0[s_{\gamma^\ast(\tau)}] \\ 
	&= \bigg[\frac{1}{\tau} + \frac{1}{1-\tau} \sum_{k=0}^m \frac{(\pi_k^\ast)^2}{\pi_k^\train} \bigg]  \E_{\gamma^\ast(\tau)}[s_{\gamma^\ast(\tau)}]  - \frac{1}{1-\tau} \frac{(\pi_0^\ast)^2}{\pi_0^\train} \E_0[s_{\gamma^\ast(\tau)}] \\ 
	&= - \frac{1}{1-\tau} \frac{(\pi_0^\ast)^2}{\pi_0^\train} \E_0[s_{\gamma^\ast(\tau)}].
\end{align*}
Therefore, we have that:
\small
\[
	\Bigg[ \bigg[\frac{1}{\tau} + \frac{1}{1-\tau} \sum_{k=0}^m \frac{(\pi_k^\ast)^2}{\pi_k^\train} \bigg] \FisherInfo{\gamma^\ast(\tau)} \FisherInfo{\gamma^\ast(\tau);\textup{Cat}}^{-1} - \frac{1}{\tau}\FisherInfo{\gamma^\ast(\tau)}\FisherInfo{\pi^\ast;\textup{Cat}}^{-1}  \Bigg] s_{\pi^\ast}(x) 
\]
\[
	= \sum_{j=0}^m \frac{1}{1-\tau} \frac{(\pi_j^\ast)^2}{\pi_j^\train} \E_j[s_{\gamma^\ast(\tau)}] \frac{p_j(x) - p_{\pi^\ast}(x)}{p_{\pi^\ast}(x)},
\]
\normalsize
as claimed. Plugging in this identity into line \eqref{Are you the straw to my berry?}, it follows that:
\small
\begin{align*}
	&\sum_{j=0}^m \frac{1}{1-\tau} \frac{(\pi_j^\ast)^2}{\pi_j^\train}   \frac{p_j(x) }{p_{\pi^\ast} (x)} \E_j[s_{\gamma^\ast(\tau)}] \\
	&= \sum_{j=0}^m \frac{1}{1-\tau} \frac{(\pi_j^\ast)^2}{\pi_j^\train}   \frac{p_j(x) - p_{\pi^\ast} (x)}{p_{\pi^\ast} (x)} \E_j[s_{\gamma^\ast(\tau)}] -  \frac{1}{\tau}  \E_{\pi^\ast}[s_{\gamma^\ast(\tau)}] \\ 
	&= \FisherInfo{\gamma^\ast(\tau)} \Bigg[ \bigg[\frac{1}{\tau} + \frac{1}{1-\tau} \sum_{k=0}^m \frac{(\pi_k^\ast)^2}{\pi_k^\train} \bigg] \FisherInfo{\gamma^\ast(\tau);\textup{Cat}}^{-1} - \frac{1}{\tau} \FisherInfo{\pi^\ast;\textup{Cat}}^{-1}  \Bigg] s_{\pi^\ast}(x) - \frac{1}{\tau}  \E_{\pi^\ast}[s_{\gamma^\ast(\tau)}]. 
\end{align*}
\normalsize

Plugging in the above into line \eqref{repair man}, we have that:
\small
\begin{align*}
	 &\sum_{j=0}^m \pi_j^\ast \frac{p_j(x) }{p_{\pi^\ast} (x)} \bar{f}_j(x)  \\
	 &=\mathbf{V}^\text{eff}(\tau)^{-1}  \FisherInfo{\gamma^\ast(\tau)}^{-1} \Bigg(\bigg[ \frac{1}{1-\tau} \sum_{j=0}^m \frac{(\pi_j^\ast)^2}{\pi_j^\train}  +  \frac{1}{\tau} \bigg]s_{\pi^\ast}(x)   - \frac{s_{\gamma^\ast(\tau)}(x)}{\tau}\Bigg) \\
	 &\qquad - \mathbf{V}^\text{eff}(\tau)^{-1}\FisherInfo{\gamma^\ast(\tau)}^{-1}\sum_{j=0}^m \frac{1}{1-\tau} \frac{(\pi_j^\ast)^2}{\pi_j^\train}   \frac{p_j(x) }{p_{\pi^\ast} (x)} \E_j[s_{\gamma^\ast(\tau)}] \\ \\ 
	&=  \mathbf{V}^\text{eff}(\tau)^{-1}\FisherInfo{\gamma^\ast(\tau)}^{-1} \Bigg(\bigg[ \frac{1}{1-\tau} \sum_{j=0}^m \frac{(\pi_j^\ast)^2}{\pi_j^\train}  +  \frac{1}{\tau} \bigg]s_{\pi^\ast}(x)   - \frac{s_{\gamma^\ast(\tau)}(x)}{\tau}\Bigg) \\
	&\qquad - \mathbf{V}^\text{eff}(\tau)^{-1} \Bigg[ \bigg[\frac{1}{\tau} + \frac{1}{1-\tau} \sum_{k=0}^m \frac{(\pi_k^\ast)^2}{\pi_k^\train} \bigg] \FisherInfo{\gamma^\ast(\tau);\textup{Cat}}^{-1} - \frac{1}{\tau}\FisherInfo{\pi^\ast;\textup{Cat}}^{-1}  \Bigg] s_{\pi^\ast}(x) \\ 
	&\qquad + \mathbf{V}^\text{eff}(\tau)^{-1}\frac{\FisherInfo{\gamma^\ast(\tau)}^{-1} }{\tau}  \E_{\pi^\ast}[s_{\gamma^\ast(\tau)}]  \\ \\ 
	&=  \mathbf{V}^\text{eff}(\tau)^{-1}\Bigg[ \bigg[\frac{1}{\tau} + \frac{1}{1-\tau} \sum_{k=0}^m \frac{(\pi_k^\ast)^2}{\pi_k^\train} \bigg] \big(\FisherInfo{\gamma^\ast(\tau)}^{-1} -\FisherInfo{\gamma^\ast(\tau);\textup{Cat}}^{-1}\big) + \frac{1}{\tau}\FisherInfo{\pi^\ast;\textup{Cat}}^{-1}  \Bigg] s_{\pi^\ast}(x) \\
	&\qquad - \mathbf{V}^\text{eff}(\tau)^{-1} \frac{\FisherInfo{\gamma^\ast(\tau)}^{-1}}{\tau}\big(s_{\gamma^\ast(\tau)}(x) -\E_{\pi^\ast}[s_{\gamma^\ast(\tau)}]\big) \\ \\ 
	&=  \mathbf{V}^\text{eff}(\tau)^{-1}  \mathbf{V}^\text{eff}(\tau)   s_{\pi^\ast}(x)  - \mathbf{V}^\text{eff}(\tau)^{-1} \frac{\FisherInfo{\gamma^\ast(\tau)}^{-1}}{\tau}\big(s_{\gamma^\ast(\tau)}(x) -\E_{\pi^\ast}[s_{\gamma^\ast(\tau)}]\big) \\ 
	&=   s_{\pi^\ast}(x)  - \mathbf{V}^\text{eff}(\tau)^{-1} \frac{\FisherInfo{\gamma^\ast(\tau)}^{-1}}{\tau}\big(s_{\gamma^\ast(\tau)}(x) -\E_{\pi^\ast}[s_{\gamma^\ast(\tau)}]\big).
\end{align*}
\normalsize
This verifies the identity stated in display \eqref{verify spongebob}, meaning that, indeed, it is true that $\bar{g}(z) =  ds_{\pi^\ast}(x) -  \mathbf{V}^\text{eff}(\tau)^{-1}\psi_\tau^\textup{eff}(z)$. Consequentially, because we already proved that $\bar{g} \in \G$, it follows that $ ds_{\pi^\ast} -  \mathbf{V}^\text{eff}(\tau)^{-1}\psi_\tau^\textup{eff} \in \G$, as claimed! \\ \\

Finally, it remains for us to prove the following orthogonality condition:
\[
	\E_{\Jcal^{\pi^\ast,\mathbf{p}, \tau}} \Big[ g' \big(S_{\pi^\ast} - \big( Ds_{\pi^\ast} -  \mathbf{V}^\text{eff}(\tau)^{-1}\psi_\tau^\textup{eff} \big) \big)\Big] = 0 \qquad \forall \ g\in\G.
\]
By Lemma \eqref{Smooth Score Function for Pi}, this is equivalent to showing that
\[
	 \E_{\Jcal^{\pi^\ast,\mathbf{p}, \tau}} \Big[ g' \mathbf{V}^\text{eff}(\tau)^{-1}\psi_\tau^\textup{eff} \Big] = 0 \qquad \forall \ g\in\G.
\]
Towards that end, consider any $g \in \G$, and let $f_j$ denote the functions used in the definition of $g$. For $z=(x,y,d)$ that are non-negligible under $\Jcal^{\pi^\ast,\mathbf{p},\tau}$, we have that:
\small
\begin{align*}
	&g(z)' \mathbf{V}^\text{eff}(\tau)^{-1}\psi_\tau^\textup{eff}(z)\\
	 &= \Bigg(  d \sum_{j=0}^{m} \pi_j^\ast \frac{p_{j} (x)}{p_{\pi^\ast}(x) } f_j(x) + (1-d) \sum_{j=0}^{m} \I{y=j} f_j(x)   \Bigg)'  \mathbf{V}^\text{eff}(\tau)^{-1} \FisherInfo{\gamma^\ast(\tau)}^{-1} \\ 
	&\qquad\qquad\times \Bigg(  \frac{d}{\tau} \big(s_{\gamma^\ast(\tau)}(x)  -  \E_{\pi^\ast}[s_{\gamma^\ast(\tau)}]\big) - \frac{1-d}{1-\tau} \sum_{j=0}^m \I{y=j} \frac{\pi_j^\ast}{\pi_j^\train}\big(s_{\gamma^\ast(\tau)}(x)  -  \E_{j}[s_{\gamma^\ast(\tau)}]\big)   \Bigg) \\ \\ 
	&=   \frac{d}{\tau} \Bigg(\sum_{j=0}^{m} \pi_j^\ast \frac{p_{j} (x)}{p_{\pi^\ast}(x) } f_j(x) \Bigg)' \mathbf{V}^\text{eff}(\tau)^{-1} \FisherInfo{\gamma^\ast(\tau)}^{-1} \Bigg( s_{\gamma^\ast(\tau)}(x)  -  \E_{\pi^\ast}[s_{\gamma^\ast(\tau)}] \Bigg) \\ 
	&\qquad\qquad -  \frac{1-d}{1-\tau} \Bigg(  \sum_{j=0}^{m} \I{y=j} f_j(x)  \Bigg)'\mathbf{V}^\text{eff}(\tau)^{-1} \FisherInfo{\gamma^\ast(\tau)}^{-1} \Bigg(  \sum_{j=0}^m \I{y=j} \frac{\pi_j^\ast}{\pi_j^\train}\big(s_{\gamma^\ast(\tau)}(x)  -  \E_{j}[s_{\gamma^\ast(\tau)}]\big) \Bigg) \\ \\ 
	&=   \frac{d}{\tau} \Bigg(\sum_{j=0}^{m} \pi_j^\ast \frac{p_{j} (x)}{p_{\pi^\ast}(x) } f_j(x) \Bigg)' \mathbf{V}^\text{eff}(\tau)^{-1} \FisherInfo{\gamma^\ast(\tau)}^{-1} \Bigg( s_{\gamma^\ast(\tau)}(x)  -  \E_{\pi^\ast}[s_{\gamma^\ast(\tau)}] \Bigg) \\ 
	&\qquad\qquad -  \frac{1-d}{1-\tau}\sum_{j=0}^{m} \I{y=j}   \frac{\pi_j^\ast}{\pi_j^\train} \Bigg(  f_j(x)  \Bigg)'\mathbf{V}^\text{eff}(\tau)^{-1} \FisherInfo{\gamma^\ast(\tau)}^{-1} \Bigg(  s_{\gamma^\ast(\tau)}(x)  -  \E_{j}[s_{\gamma^\ast(\tau)}] \Bigg) \\ \\ 
\end{align*}
\normalsize
Now, since $ \E_j \big[  f_j\big] = 0$, we have that:
\smaller
\begin{align*}
	&\E_{\Jcal^{\pi^\ast,\mathbf{p},\tau}} \Bigg[ \frac{D}{\tau} \Bigg(\sum_{j=0}^{m} \pi_j^\ast \frac{p_{j} (X)}{p_{\pi^\ast}(X) } f_j(X) \Bigg)' \mathbf{V}^\text{eff}(\tau)^{-1} \FisherInfo{\gamma^\ast(\tau)}^{-1}\E_{\pi^\ast}[s_{\gamma^\ast(\tau)}]   \Bigg] \\
	&= \E_{\pi^\ast} \Bigg[  \Bigg(\sum_{j=0}^{m} \pi_j^\ast \frac{p_{j} (X)}{p_{\pi^\ast}(X) } f_j(X) \Bigg)' \mathbf{V}^\text{eff}(\tau)^{-1} \FisherInfo{\gamma^\ast(\tau)}^{-1}\E_{\pi^\ast}[s_{\gamma^\ast(\tau)}]   \Bigg] \\
	 &=  \Bigg(\sum_{j=0}^{m} \pi_j^\ast \E_{\pi^\ast} \Bigg[ \frac{p_{j} (X)}{p_{\pi^\ast}(X) } f_j(X)\Bigg] \Bigg)' \mathbf{V}^\text{eff}(\tau)^{-1} \FisherInfo{\gamma^\ast(\tau)}^{-1}\E_{\pi^\ast}[s_{\gamma^\ast(\tau)}] \\
	  &=  \Bigg(\sum_{j=0}^{m} \pi_j^\ast \E_j \big[  f_j\big] \Bigg)' \mathbf{V}^\text{eff}(\tau)^{-1} \FisherInfo{\gamma^\ast(\tau)}^{-1}\E_{\pi^\ast}[s_{\gamma^\ast(\tau)}] \\
	  &= 0.
\end{align*}
\normalsize
Likewise, note that:
\small
\begin{align*}
	&\E_{\Jcal^{\pi^\ast,\mathbf{p},\tau}} \Bigg[ \frac{1-d}{1-\tau}\sum_{j=0}^{m} \I{y=j}   \frac{\pi_j^\ast}{\pi_j^\train}   f_j(x)' \mathbf{V}^\text{eff}(\tau)^{-1} \FisherInfo{\gamma^\ast(\tau)}^{-1}   \E_{j}[s_{\gamma^\ast(\tau)}]\Bigg] \\
	&= \sum_{j=0}^{m}   \pi_j^\ast   \E_{j} \big[ f_j\big]' \mathbf{V}^\text{eff}(\tau)^{-1} \FisherInfo{\gamma^\ast(\tau)}^{-1}   \E_{j}[s_{\gamma^\ast(\tau)}] \\ 
	&= 0.
\end{align*}
\normalsize

Ergo:
\begin{align*}
	 &\E_{\Jcal^{\pi^\ast,\mathbf{p}, \tau}} \Big[ g' \mathbf{V}^\text{eff}(\tau)^{-1}\psi_\tau^\textup{eff} \Big] \\
	 &=   \E_{\Jcal^{\pi^\ast,\mathbf{p}, \tau}} \Bigg[ \frac{D}{\tau} \Bigg(\sum_{j=0}^{m} \pi_j^\ast \frac{p_{j} (X)}{p_{\pi^\ast}(X) } f_j(X) \Bigg)' \mathbf{V}^\text{eff}(\tau)^{-1} \FisherInfo{\gamma^\ast(\tau)}^{-1}  s_{\gamma^\ast(\tau)}(X)  \Bigg] \\
	 &\qquad  -  \E_{\Jcal^{\pi^\ast,\mathbf{p}, \tau}} \Bigg[  \frac{1-D}{1-\tau}\sum_{j=0}^{m} \I{Y=j}   \frac{\pi_j^\ast}{\pi_j^\train}   f_j(X)' \mathbf{V}^\text{eff}(\tau)^{-1} \FisherInfo{\gamma^\ast(\tau)}^{-1}   s_{\gamma^\ast(\tau)}(X) \Bigg] \\ \\ 
	  &=  \sum_{j=0}^{m} \pi_j^\ast  \E_{\pi^\ast} \Bigg[  \frac{p_{j} (X)}{p_{\pi^\ast}(X) } f_j(X)' \mathbf{V}^\text{eff}(\tau)^{-1} \FisherInfo{\gamma^\ast(\tau)}^{-1}  s_{\gamma^\ast(\tau)}(X)  \Bigg] \\
	 &\qquad  - \sum_{j=0}^{m}\pi_j^\ast   \E_{j} \Bigg[   f_j(X)' \mathbf{V}^\text{eff}(\tau)^{-1} \FisherInfo{\gamma^\ast(\tau)}^{-1}   s_{\gamma^\ast(\tau)}(X) \Bigg] \\ \\ 
	   &=  \sum_{j=0}^{m} \pi_j^\ast  \E_{j} \Bigg[  f_j(X)' \mathbf{V}^\text{eff}(\tau)^{-1} \FisherInfo{\gamma^\ast(\tau)}^{-1}  s_{\gamma^\ast(\tau)}(X)  \Bigg] \\
	 &\qquad  - \sum_{j=0}^{m}\pi_j^\ast   \E_{j} \Bigg[   f_j(X)' \mathbf{V}^\text{eff}(\tau)^{-1} \FisherInfo{\gamma^\ast(\tau)}^{-1}   s_{\gamma^\ast(\tau)}(X) \Bigg] \\ \\ 
	 &= 0,
\end{align*}
meaning that the orthogonality condition holds. 
\end{proof}
\vspace{0.4in}

\begin{proof}[\uline{Proof of Corollary \eqref{Projection of Ordinary Score onto T}}]
By Lemma \eqref{Nuisance Tangent Set is a Subset of G}, we know that $\T \subseteq \G$, and so under Assumptions  \eqref{Fixed Tau Regime: Cat minus Mixture FIM} and \eqref{Fixed Tau Regime: Mixture FIM not too small}, it directly follows from Lemma \eqref{Projection of Ordinary Score onto G} that
\[
	\E_{\Jcal^{\pi^\ast,\mathbf{p}, \tau}} \Big[ t' \big(S_{\pi^\ast} - \big( Ds_{\pi^\ast} -  \mathbf{V}^\text{eff}(\tau)^{-1}\psi_\tau^\textup{eff} \big) \big)\Big] = 0 \qquad \forall \ t\in\T.
\]

It remains for us to show that $d s_{\pi^\ast} - \mathbf{V}^\textup{eff}(\tau)^{-1} \psi_\tau^\textup{eff} \in \T$. With that goal in mind, consider the parametric submodel defined in Lemma \eqref{A Particular Smooth Parametric Submodel}, which is indeed a parametric submodel by virtue of Assumption \eqref{Fixed Tau Regime: expo family}. Let $S_{0}^{\text{sub},\tau}$ denote the submodel's score function for the nuisance parameter (i.e., components $m+1$ to $2m$ of $S_{\pi,\rho}^{\text{sub},\tau}$, as defined in Lemma \eqref{A Particular Smooth Parametric Submodel}), evaluated at $(\pi^\ast,0)$. Then for each $z$, $S_{0}^{\text{sub},\tau}(z)$ is given by
\small
\begin{align*}
	S_{0}^{\text{sub},\tau}(z)&= 	 d \sum\limits_{j=0}^m   \frac{(\pi_j^\ast)^2}{\pi_j^\train}  \frac{  p_j(x) }{ p_{\pi^\ast}(x) }  (s_{\gamma^\ast(\tau)}(x) - \E_j[s_{\gamma^\ast(\tau)}])\Ibig{p_{j}(x) > 0} \\
	&\qquad\qquad\qquad+ (1-d)  \sum\limits_{j=0}^{m} \I{y=j}  \frac{\pi_j^\ast}{\pi_j^\train}  \big(s_{\gamma^\ast(\tau)}(x) - \E_j [s_{\gamma^\ast(\tau)}] \big)\Ibig{p_j(x) > 0}.
\end{align*}
\normalsize

Now, by Lemma \eqref{A Particular Smooth Parametric Submodel}, the parametric submodel in question is smooth (in the sense of Definition A.1 in \citep{Newey1990}), so by definition of $\T$ (see the proof of Lemma \eqref{Nuisance Tangent Set is a Subset of G}), if $C \in \mathbb{R}^{m\times m}$ satisfies $\E_{\Jcal^{\pi^\ast,\mathbf{p},\tau}}\big[ \norm{C S_{0}^{\text{sub},\tau}}{2}^2 \big] < \infty$, then $CS_{0}^{\text{sub},\tau} \in \T$. Note that $\E_{\Jcal^{\pi^\ast,\mathbf{p},\tau}} \big[ \norm{C S_{0}^{\text{sub},\tau} }{2}^2\big] \leq \maxEval{C} \E_{\Jcal^{\pi^\ast,\mathbf{p},\tau}} \big[ \norm{S_{0}^{\text{sub},\tau}}{2}^2\big]$ and that $\E_{\Jcal^{\pi^\ast,\mathbf{p},\tau}} \big[ \norm{S_{0}^{\text{sub},\tau}}{2}^2\big] < \infty$, so to show that $\E_{\Jcal^{\pi^\ast,\mathbf{p},\tau}}\big[ \norm{C S_{0}^{\text{sub},\tau} }{2}^2 \big] < \infty$, it suffices to show that $\maxEval{C} < \infty$.\\ 

Here, we will choose $C = \frac{1}{1-\tau} \mathbf{V}^\textup{eff}(\tau)^{-1} \FisherInfo{\gamma^\ast(\tau)}^{-1}$, and show that $CS_{0}^{\text{sub},\tau} \in \T$ by verifying that $\maxEval{C} < \infty$. Towards that end, observe that:
\begin{align*}
	\maxEval{C} &=  \frac{1}{1-\tau} \maxEvalBigg{  \mathbf{V}^\textup{eff}(\tau)^{-1} \FisherInfo{\gamma^\ast(\tau)}^{-1}  } \\
	&\leq  \frac{1}{1-\tau} \frac{1}{\minEval{\mathbf{V}^\textup{eff}(\tau)} \minEval{ \FisherInfo{\gamma^\ast(\tau)}} } \\ 
	&\leq  \frac{1}{(1-\tau)\sqrt{\Lambda}} \frac{1}{ \minEval{\mathbf{V}^\textup{eff}(\tau)}  } \\
	&\leq  \frac{1}{(1-\tau)\sqrt{\Lambda}} \frac{1}{ \minEval{   \frac{1}{\tau}\FisherInfo{\pi^\ast;\textup{Cat}}^{-1}  }  }  \\
	&=  \frac{\tau  }{(1-\tau)\sqrt{\Lambda}}  \maxEval{\FisherInfo{\pi^\ast;\textup{Cat}}} \\ 
	&\leq  \frac{\tau }{(1-\tau)\sqrt{\Lambda}}   \frac{m+1}{\xi} \\
	&< \infty.
\end{align*}
where the second to last inequality follows from Lemma \eqref{Bounds on Eigenvalues of Categorical Fisher Info}. Thus, by our argument earlier,  we may conclude that $ \frac{1}{1-\tau} \mathbf{V}^\textup{eff}(\tau)^{-1} \FisherInfo{\gamma^\ast(\tau)}^{-1} S_{0}^{\text{sub},\tau} \in \T$.

Next, note that $\frac{1}{1-\tau} \mathbf{V}^\textup{eff}(\tau)^{-1} \FisherInfo{\gamma^\ast(\tau)}^{-1} S_{0}^{\text{sub},\tau}(z) = \bar{g}(z)$ where $\bar{g}(z)$ was defined in the proof of Lemma \eqref{Projection of Ordinary Score onto G}. Also recall that, in the proof of Lemma \eqref{Projection of Ordinary Score onto G}, we proved that $\bar{g}(z)= d s_{\pi^\ast}(x) - \mathbf{V}^\textup{eff}(\tau)^{-1} \psi_\tau^\textup{eff}(z)$. Consequentially, we have that $\frac{1}{1-\tau} \mathbf{V}^\textup{eff}(\tau)^{-1} \FisherInfo{\gamma^\ast(\tau)}^{-1} S_{0}^{\text{sub},\tau}(z) = d s_{\pi^\ast}(x) - \mathbf{V}^\textup{eff}(\tau)^{-1} \psi_\tau^\textup{eff}(z)$, and so $ d s_{\pi^\ast} - \mathbf{V}^\textup{eff}(\tau)^{-1} \psi_\tau^\textup{eff} \in \T$, as claimed!

\end{proof}
\vspace{0.4in}

\begin{proof}[\uline{Proof of Lemma \eqref{Efficiency Bound}}] 

To prove that $\mathbf{V}^{\textup{eff}}(\tau)$ is the semiparametric efficiency bound and $\psi_\tau^{\textup{eff}}$ is the efficient influence function for the fixed $\tau$-IID regime, we will verify the conditions of Theorem 3.2 in \citep{Newey1990}. In our setting, these conditions are:
\begin{enumerate}[label=(\Alph*)]
	\item $\Jcal^{\pi,\mathbf{p},\tau}(z)$ is smooth with score function $S_{\pi}(z) = d  s_{\pi}(x) \I{ p_{\pi}(x) > 0}$. 
	\item The residual of the projection of $S_{\pi^\ast}$ on $\T$ has a nonsingular covariance matrix.
\end{enumerate}

As a quick aside, note that the Theorem 3.2 which appears in \citep{Newey1990} additionally requires that $\T$ be linear. However, as Newey points out (see \cite[p.~106]{Newey1990}), this is actually not a necessary condition-- it is only used to ensure that the projection of $S_{\pi^\ast}$ onto $\T$ exists. Indeed, his proof of Theorem 3.2 does not depend on $\T$ to be linear. For us, since we have already identified the projection of $S_{\pi^\ast}$ onto $\T$ in Corollary \eqref{Projection of Ordinary Score onto T}, we already know that the projection exists; as such, in this proof, we omit checking whether or not $\T$ is linear.  

Now, condition (A) is true by Lemma \eqref{Smooth Score Function for Pi}. Towards checking  condition (B), we first find the covariance matrix of the residual of the projection of $S_{\pi^\ast}$ on $\T$. Using the formula for this projection from  Corollary \eqref{Projection of Ordinary Score onto T}, the covariance matrix in question is given by: 
\begin{align*}
	\Var_{\Jcal^{\pi^\ast,\mathbf{p},\tau}}\Big[S_{\pi^\ast} - \big( Ds_{\pi^\ast} -  \mathbf{V}^\textup{eff}(\tau)^{-1}\psi_\tau^\textup{eff} \big)\Big] &= \Var_{\Jcal^{\pi^\ast,\mathbf{p},\tau}}\Big[  \mathbf{V}^\textup{eff}(\tau)^{-1}\psi_\tau^\textup{eff} \Big]  \\ 
	&=  \mathbf{V}^\textup{eff}(\tau)^{-1} \Var_{\Jcal^{\pi^\ast,\mathbf{p},\tau}}\Big[ \psi_\tau^\textup{eff} \Big]  \mathbf{V}^\textup{eff}(\tau)^{-1}.
\end{align*}
where the first equality follows from Lemma \eqref{Smooth Score Function for Pi}. Now, note that:
\small
\begin{align*}
	&\Var_{\Jcal^{\pi^\ast,\mathbf{p},\tau}}\Big[ \FisherInfo{\gamma^\ast(\tau)} \psi_\tau^\textup{eff} \Big] \\
	&=  \Var_{\Jcal^{\pi^\ast,\mathbf{p},\tau}}\Bigg[ \frac{d}{\tau} \big(s_{\gamma^\ast(\tau)}  -  \E_{\pi^\ast}[s_{\gamma^\ast(\tau)}]\big) - \frac{1-d}{1-\tau}  \sum_{j=0}^m \I{y=j} \frac{\pi_j^\ast}{\pi_j^\train}\big(s_{\gamma^\ast(\tau)}  -  \E_{j}[s_{\gamma^\ast(\tau)}]\big)  \Bigg] \\\\ 
	&= \E_{\Jcal^{\pi^\ast,\mathbf{p},\tau}} \Bigg[  \frac{d}{\tau^2} \big(s_{\gamma^\ast(\tau)}  -  \E_{\pi^\ast}[s_{\gamma^\ast(\tau)}]\big) \big(s_{\gamma^\ast(\tau)}  -  \E_{\pi^\ast}[s_{\gamma^\ast(\tau)}]\big)' \Bigg] \\
	&\qquad+  \E_{\Jcal^{\pi^\ast,\mathbf{p},\tau}} \Bigg[\frac{1-d}{(1-\tau)^2}  \bigg(\sum_{j=0}^m \I{y=j} \frac{\pi_j^\ast}{\pi_j^\train}\big(s_{\gamma^\ast(\tau)}  -  \E_{j}[s_{\gamma^\ast(\tau)}]\big)\bigg) \bigg(\sum_{j=0}^m \I{y=j} \frac{\pi_j^\ast}{\pi_j^\train}\big(s_{\gamma^\ast(\tau)}  -  \E_{j}[s_{\gamma^\ast(\tau)}]\big)\bigg)' \Bigg] \\ \\ 
	&= \E_{\Jcal^{\pi^\ast,\mathbf{p},\tau}} \Bigg[  \frac{d}{\tau^2} \big(s_{\gamma^\ast(\tau)}  -  \E_{\pi^\ast}[s_{\gamma^\ast(\tau)}]\big) \big(s_{\gamma^\ast(\tau)}  -  \E_{\pi^\ast}[s_{\gamma^\ast(\tau)}]\big)' \Bigg] \\
	&\qquad+  \E_{\Jcal^{\pi^\ast,\mathbf{p},\tau}} \Bigg[\frac{1-d}{(1-\tau)^2}  \sum_{j=0}^m \I{y=j}  \frac{(\pi_j^\ast)^2}{(\pi_j^\train)^2}\big(s_{\gamma^\ast(\tau)}  -  \E_{j}[s_{\gamma^\ast(\tau)}]\big)\big(s_{\gamma^\ast(\tau)}  -  \E_{j}[s_{\gamma^\ast(\tau)}]\big)' \Bigg] \\ \\ 
	&= \frac{1}{\tau}\E_{\pi^\ast} \Bigg[   \big(s_{\gamma^\ast(\tau)}  -  \E_{\pi^\ast}[s_{\gamma^\ast(\tau)}]\big) \big(s_{\gamma^\ast(\tau)}  -  \E_{\pi^\ast}[s_{\gamma^\ast(\tau)}]\big)' \Bigg] \\
	&\qquad+  \frac{1}{1-\tau} \sum_{j=0}^m\frac{(\pi_j^\ast)^2}{\pi_j^\train} \E_{j} \Bigg[  \big(s_{\gamma^\ast(\tau)}  -  \E_{j}[s_{\gamma^\ast(\tau)}]\big)\big(s_{\gamma^\ast(\tau)}  -  \E_{j}[s_{\gamma^\ast(\tau)}]\big)' \Bigg] \\ \\ 
	&= \frac{1}{\tau} \Var_{\pi^\ast}[s_{\gamma^\ast(\tau)}] +  \frac{1}{1-\tau} \sum_{j=0}^m \frac{(\pi_j^\ast)^2}{\pi_j^\train} \Var_j[s_{\gamma^\ast(\tau)}].
\end{align*}
\normalsize
Now, by a derivation nearly identical to the work done in the proof of Corollaries \eqref{First Order Constant for Any Matching Function} and \eqref{First Order Constant When Matching Function is Score Function}, the last line in the display above equals:
\small
\begin{align*}
	&\Var_{\Jcal^{\pi^\ast,\mathbf{p},\tau}}\Big[ \FisherInfo{\gamma^\ast(\tau)} \psi_\tau^\textup{eff} \Big]\\
	 &= \Bigg[ \frac{1}{\tau} + \frac{1}{1-\tau}\sum_{j=0}^m \frac{(\pi_j^\ast)^2}{\pi_j^\train} \Bigg]\Big( \FisherInfo{\gamma^\ast(\tau)} - \FisherInfo{\gamma^\ast(\tau)} \FisherInfo{\gamma^\ast(\tau);\textup{Cat}}^{-1}  \FisherInfo{\gamma^\ast(\tau)}  \Big) + \frac{1}{\tau}  \FisherInfo{\gamma^\ast(\tau)}  \FisherInfo{\pi^\ast;\textup{Cat}}^{-1}\FisherInfo{\gamma^\ast(\tau)}.
\end{align*}
\normalsize
This implies that:
\begin{align}
	\implies \Var_{\Jcal^{\pi^\ast,\mathbf{p},\tau}}\Big[\psi_\tau^\textup{eff} \Big] &=  \FisherInfo{\gamma^\ast(\tau)}^{-1} \Var_{\Jcal^{\pi^\ast,\mathbf{p},\tau}}\Big[ \FisherInfo{\gamma^\ast(\tau)} \psi_\tau^\textup{eff} \Big] \FisherInfo{\gamma^\ast(\tau)}^{-1} \nonumber \\ 
	&= \Bigg[ \frac{1}{\tau} + \frac{1}{1-\tau}\sum_{j=0}^m \frac{(\pi_j^\ast)^2}{\pi_j^\train} \Bigg]\Big( \FisherInfo{\gamma^\ast(\tau)}^{-1} -  \FisherInfo{\gamma^\ast(\tau);\textup{Cat}}^{-1}   \Big) + \frac{1}{\tau}   \FisherInfo{\pi^\ast;\textup{Cat}}^{-1} \nonumber \\ 
	&= \mathbf{V}^\textup{eff}(\tau), \label{variance matrix of efficient influence function}
\end{align}
as was claimed in the statement of the present Lemma. Hence, it follows that
\begin{align} 
	\Var_{\Jcal^{\pi^\ast,\mathbf{p},\tau}}\Big[S_{\pi^\ast} - \big( Ds_{\pi^\ast} -  \mathbf{V}^\textup{eff}(\tau)^{-1}\psi_\tau^\textup{eff} \big)\Big] &=  \mathbf{V}^\textup{eff}(\tau)^{-1} \Var_{\Jcal^{\pi^\ast,\mathbf{p},\tau}}\Big[ \psi_\tau^\textup{eff} \Big]  \mathbf{V}^\textup{eff}(\tau)^{-1} \nonumber \\ 
	&=  \mathbf{V}^\textup{eff}(\tau)^{-1},\label{variance matrix of residual}
\end{align}
which since $\mathbf{V}^\textup{eff}(\tau)$ can be shown to be nonsingular, implies the nonsingularity of $\mathbf{V}^\textup{eff}(\tau)^{-1}$ and hence of $\Var_{\Jcal^{\pi^\ast,\mathbf{p},\tau}}\Big[S_{\pi^\ast} - \big( Ds_{\pi^\ast} -  \mathbf{V}^\textup{eff}(\tau)^{-1}\psi_\tau^\textup{eff} \big)\Big]$. \\ 

Thus, both conditions (A) and (B) are met.  So, by Theorem 3.2 in \citep{Newey1990}, it follows that the efficient influence function is
$\Var_{\Jcal^{\pi^\ast,\mathbf{p},\tau}}\Big[S_{\pi^\ast} - \big( Ds_{\pi^\ast} -  \mathbf{V}^\textup{eff}(\tau)^{-1} \psi_\tau^\textup{eff} \big)\Big]^{-1} \Big( S_{\pi^\ast}(z) - \big( ds_{\pi^\ast}(x) -  \mathbf{V}^\textup{eff}(\tau)^{-1}\psi_\tau^\textup{eff}(z) \big) \Big)$. By line \eqref{variance matrix of residual} and Lemma  \eqref{Smooth Score Function for Pi}, this can be rewritten as:
\begin{align*}
	&\Var_{\Jcal^{\pi^\ast,\mathbf{p},\tau}}\Big[S_{\pi^\ast} - \big( Ds_{\pi^\ast} -  \mathbf{V}^\textup{eff}(\tau)^{-1}\psi_\tau^\textup{eff} \big)\Big]^{-1} \Big( S_{\pi^\ast}(z) - \big( ds_{\pi^\ast}(x) -  \mathbf{V}^\textup{eff}(\tau)^{-1}\psi_\tau^\textup{eff}(z) \big) \Big) \\
	&=  \mathbf{V}^\textup{eff}(\tau)\mathbf{V}^\textup{eff}(\tau)^{-1}\psi_\tau^\textup{eff}(z) \\ 
	&= \psi_\tau^\textup{eff}(z) ,
\end{align*}
so $\psi_\tau^\textup{eff}(z) $ is indeed the efficient influence function. Theorem 3.2 in \citep{Newey1990} also tells us that the efficiency bound is given by $\Var_{\Jcal^{\pi^\ast,\mathbf{p},\tau}}\Big[S_{\pi^\ast} - \big( Ds_{\pi^\ast} -  \mathbf{V}^\textup{eff}(\tau)^{-1} \psi_\tau^\textup{eff} \big)\Big]^{-1}$, which by line \eqref{variance matrix of residual}, is equal to  $\mathbf{V}^\textup{eff}(\tau)$. \\

Thus, we have proven that $\psi_\tau^\textup{eff}$ is the efficient influence function and that $ \mathbf{V}^\textup{eff}(\tau)$ is the efficiency bound!

\end{proof}
\vspace{0.4in}

\begin{proof}[\uline{Proof of Lemma \eqref{Largest CRLB}}]

First, we compute the form of $\CRLB{\mathbf{M}^{\widetilde{\textup{sub}},\tau}}$ for an arbitrary smooth parametric submodel $\mathbf{M}^{\widetilde{\textup{sub}},\tau}$. Let $\Theta^{\widetilde{\textup{sub}}}$ denote the parameter space for the submodel, and let $\theta = (\pi,\rho) \in \Theta^{\widetilde{\textup{sub}}}$ denote an arbitrary member of that space. Let $k$ denote the number of components in the vector $\theta$. The true parameter is $\theta^\ast = (\pi^\ast,\rho^\ast)$. We are interested in learning only $\pi^\ast$, not $\rho^\ast$; this is equivalent to learning $[I_{m\times m}, \ 0_{m\times k-m}] \theta^\ast$. The CRLB for learning this function of $\pi^\ast$ is:
\begin{align*}
	\CRLB{\mathbf{M}^{\widetilde{\textup{sub}},\tau}} &= [I_{m\times m}, \ 0_{m\times k-m}] \Var_{\Jcal^{\pi^\ast,\mathbf{p},\tau}}\big[ S_{\pi,\rho}^{\widetilde{\text{sub}},\tau}  \big]^{-1} [I_{m\times m}, \ 0_{m\times k-m}]',
\end{align*}
where $S_{\pi,\rho}^{\widetilde{\text{sub}},\tau} = \big[S_{\pi^\ast} , S_{\rho^\ast}^{\widetilde{\text{sub}},\tau} \big]$ is the submodel's $\mathbb{R}^{k}$-valued score function for both the parameter of interest and the nuisance parameter, evaluated at the truth. Now, define
\[
	K^{\widetilde{\textup{sub}},\tau} := 	\E_{\Jcal^{\pi^\ast,\mathbf{p},\tau}}\Big[S_{\pi^\ast} \big(S_{\rho^\ast}^{\widetilde{\text{sub}},\tau}\big)' \Big] \Var_{\Jcal^{\pi^\ast,\mathbf{p},\tau}}\Big[  S_{\rho^\ast}^{\widetilde{\text{sub}},\tau}   \Big]^{-1} \in \mathbb{R}^{m\times k}.
\] 
Then, by the partitioned inverse formula, it follows that:
\begin{align*}
	&\CRLB{\mathbf{M}^{\widetilde{\textup{sub}},\tau}} \\
	&= \bigg\{ \Var_{\Jcal^{\pi^\ast,\mathbf{p},\tau}}[S_{\pi^\ast} ]    -    K^{\widetilde{\textup{sub}},\tau}  \E_{\Jcal^{\pi^\ast,\mathbf{p},\tau}}\Big[ S_{\rho^\ast}^{\widetilde{\text{sub}},\tau}  S_{\pi^\ast}' \Big]    \bigg\}^{-1} \\ 
	&= \bigg\{ \E_{\Jcal^{\pi^\ast,\mathbf{p},\tau}}[S_{\pi^\ast}S_{\pi^\ast}' ]    -      \E_{\Jcal^{\pi^\ast,\mathbf{p},\tau}}\Big[ K^{\widetilde{\textup{sub}},\tau} S_{\rho^\ast}^{\widetilde{\text{sub}},\tau}  S_{\pi^\ast}' \Big]  - 0   \bigg\}^{-1} \\ 
	&= \bigg\{ \E_{\Jcal^{\pi^\ast,\mathbf{p},\tau}}\Big[ \big(S_{\pi^\ast}    -     K^{\widetilde{\textup{sub}},\tau} S_{\rho^\ast}^{\widetilde{\text{sub}},\tau}\big)  S_{\pi^\ast}' \Big] - \E_{\Jcal^{\pi^\ast,\mathbf{p},\tau}}\Big[ \big(S_{\pi^\ast}    -     K^{\widetilde{\textup{sub}},\tau} S_{\rho^\ast}^{\widetilde{\text{sub}},\tau}\big)  \big( K^{\widetilde{\textup{sub}},\tau} S_{\rho^\ast}^{\widetilde{\text{sub}},\tau}\big)' \Big] \bigg\}^{-1} \\ 
	&= \Var_{\Jcal^{\pi^\ast,\mathbf{p},\tau}}\Big[ S_{\pi^\ast}    -     K^{\widetilde{\textup{sub}},\tau} S_{\rho^\ast}^{\widetilde{\text{sub}},\tau} \Big]^{-1},
\end{align*}
where the second to last line is because $S_{\pi^\ast}    -     K^{\widetilde{\textup{sub}},\tau} S_{\rho^\ast}^{\widetilde{\text{sub}},\tau}$ is the residual from the mean-square projection of $S_{\pi^\ast}$ onto the linear space spanned by $ S_{\rho^\ast}^{\widetilde{\text{sub}},\tau}$, meaning that $S_{\pi^\ast}    -     K^{\widetilde{\textup{sub}},\tau} S_{\rho^\ast}^{\widetilde{\text{sub}},\tau}$ is orthogonal to $K^{\widetilde{\textup{sub}},\tau} S_{\rho^\ast}^{\widetilde{\text{sub}},\tau}$. \\ 

Next, observe that:
\begin{align*}
	&\CRLB{\mathbf{M}^{\widetilde{\textup{sub}},\tau}}^{-1}\\
	 &= \Var_{\Jcal^{\pi^\ast,\mathbf{p},\tau}}\Big[ S_{\pi^\ast}    -     K^{\widetilde{\textup{sub}},\tau} S_{\rho^\ast}^{\widetilde{\text{sub}},\tau} \Big] \\ 
	&= \Var_{\Jcal^{\pi^\ast,\mathbf{p},\tau}}\bigg[ \Big(\underbrace{S_{\pi^\ast}    -  \big( Ds_{\pi^\ast} -  \mathbf{V}^\textup{eff}(\tau)^{-1}\psi_\tau^\textup{eff} \big)}_{=:W_1}  \Big) + \Big( \underbrace{\big( Ds_{\pi^\ast} -  \mathbf{V}^\textup{eff}(\tau)^{-1}\psi_\tau^\textup{eff} \big)  -  K^{\widetilde{\textup{sub}},\tau} S_{\rho^\ast}^{\widetilde{\text{sub}},\tau}}_{=:W_2} \Big) \bigg] \\ 
	&= \Var_{\Jcal^{\pi^\ast,\mathbf{p},\tau}}[W_1 + W_2]. 
\end{align*}

From the definition of $\T$ in Lemma \eqref{Nuisance Tangent Set is a Subset of G}, we know that  $S_{\rho^\ast}^{\widetilde{\text{sub}},\tau} \in \T$; further, since $\T \subseteq \G$ by Lemma   \eqref{Nuisance Tangent Set is a Subset of G}, we know that $S_{\rho^\ast}^{\widetilde{\text{sub}},\tau} \in \G$. Now, it is easily seen that $\G$ is linear, so since $d s_{\pi^\ast} - \mathbf{V}^\textup{eff}(\tau)^{-1} \psi_\tau^\textup{eff} \in \G$ by Lemma \eqref{Projection of Ordinary Score onto G}, it follows that $W_2 = \big(d s_{\pi^\ast} - \mathbf{V}^\textup{eff}(\tau)^{-1} \psi_\tau^\textup{eff}\big) - K^{\widetilde{\textup{sub}},\tau} S_{\rho^\ast}^{\widetilde{\text{sub}},\tau} \in \G$ as well. It is also easy to see that $\G$ is a $m$-self replicating linear space (in the sense of \cite[p.~44]{Tsiatis2006}). Also, 
$W_1$ is orthogonal to $\G$ by Lemma \eqref{Projection of Ordinary Score onto G}. Thus, all the conditions of the Multivariate Pythagorean Theorem (Theorem 3.3 in \cite{Tsiatis2006}) are met, and so it holds that 
\[
	\Var_{\Jcal^{\pi^\ast,\mathbf{p},\tau}}[W_1 + W_2] = \Var_{\Jcal^{\pi^\ast,\mathbf{p},\tau}}[W_1]  + \Var_{\Jcal^{\pi^\ast,\mathbf{p},\tau}}[W_2]
\]
Consequentially:
\begin{align}
	\CRLB{\mathbf{M}^{\widetilde{\textup{sub}},\tau}}^{-1} &= \Var_{\Jcal^{\pi^\ast,\mathbf{p},\tau}}[W_1 + W_2]  \nonumber \\ 
	&\succeq \Var_{\Jcal^{\pi^\ast,\mathbf{p},\tau}}[W_1] \nonumber \\
	&=  \Var_{\Jcal^{\pi^\ast,\mathbf{p},\tau}}\Big[ S_{\pi^\ast}    -  \big( Ds_{\pi^\ast} -  \mathbf{V}^\textup{eff}(\tau)^{-1}\psi_\tau^\textup{eff} \big)\Big] \label{tough ping pong}.
\end{align}
Next, we will argue that, in fact, $\Var_{\Jcal^{\pi^\ast,\mathbf{p},\tau}}[S_{\pi^\ast}    -  \big( Ds_{\pi^\ast} -  \mathbf{V}^\textup{eff}(\tau)^{-1}\psi_\tau^\textup{eff} \big)] = \CRLB{\mathbf{M}^{\textup{sub},\tau}}^{-1}$. By our work earlier in this proof, this is equivalent to showing that 
\[
	 \Var_{\Jcal^{\pi^\ast,\mathbf{p},\tau}}[S_{\pi^\ast}    -  \big( Ds_{\pi^\ast} -  \mathbf{V}^\textup{eff}(\tau)^{-1}\psi_\tau^\textup{eff} \big)]  = \Var_{\Jcal^{\pi^\ast,\mathbf{p},\tau}}\Big[ S_{\pi^\ast}    -     K^{\textup{sub},\tau} S_{0}^{\text{sub},\tau} \Big],
\]
where $S_{0}^{\text{sub},\tau}$ is the score function for the nuisance parameter in the submodel $\mathbf{M}^{\textup{sub},\tau}$ and is equal to the last $m$ components of the random vector defined in Lemma \eqref{A Particular Smooth Parametric Submodel} evaluated at the truth $(\pi,\rho) = (\pi^\ast,0)$, and $K^{\textup{sub},\tau} \in \mathbb{R}^{m \times m}$ is such that $K^{\textup{sub},\tau} S_{0}^{\text{sub},\tau}$ is the mean-square projection of $S_{\pi^\ast}$ onto $\big\{ KS_{0}^{\text{sub},\tau} \ : \ K\in\mathbb{R}^{m\times m} \big\}$. Now, by the argument employed at the very end of the proof of Corollary \eqref{Projection of Ordinary Score onto T}, we know that $\frac{1}{1-\tau} \mathbf{V}^\textup{eff}(\tau)^{-1} \FisherInfo{\gamma^\ast(\tau)}^{-1} S_0^{\text{sub},\tau} = d s_{\pi^\ast}  - \mathbf{V}^\textup{eff}(\tau)^{-1} \psi_\tau^\textup{eff}$, and so to show that the display above is true, it suffices to show that
\[
	S_{\pi^\ast}    -  \frac{1}{1-\tau} \mathbf{V}^\textup{eff}(\tau)^{-1} \FisherInfo{\gamma^\ast(\tau)}^{-1} S_{0}^{\text{sub},\tau} =  S_{\pi^\ast}    -     K^{\textup{sub},\tau} S_{0}^{\text{sub},\tau},
\]
i.e., it suffices to show that:
\begin{equation} \label{Ksub Projection Matrix Identity}
	  \frac{1}{1-\tau} \mathbf{V}^\textup{eff}(\tau)^{-1} \FisherInfo{\gamma^\ast(\tau)}^{-1}  = K^{\textup{sub},\tau}.
\end{equation}
To verify the equality in the display above, notice that:
\begin{align*}
	&\E_{\Jcal^{\pi^\ast,\mathbf{p},\tau}}\bigg[ \Bignorm{ S_{\pi^\ast}    -  \frac{1}{1-\tau} \mathbf{V}^\textup{eff}(\tau)^{-1} \FisherInfo{\gamma^\ast(\tau)}^{-1} S_{0}^{\text{sub},\tau}  }{2}^2\bigg] \\
	&= \E_{\Jcal^{\pi^\ast,\mathbf{p},\tau}}\bigg[ \Bignorm{ S_{\pi^\ast}    - \big( Ds_{\pi^\ast} -  \mathbf{V}^\textup{eff}(\tau)^{-1}\psi_\tau^\textup{eff} \big)   }{2}^2\bigg] \\ 
	&\leq \E_{\Jcal^{\pi^\ast,\mathbf{p},\tau}}\bigg[ \Bignorm{ S_{\pi^\ast}    -  g}{2}^2\bigg] \quad \forall g \in \G,
\end{align*}
where the last line follows from Lemma \eqref{Projection of Ordinary Score onto G}. However, note that for any  $K\in\mathbb{R}^{m\times m}$, we have that $KS_{0}^{\text{sub},\tau} \in \T \subseteq\G$, and so the display above implies that
\begin{align*}
	\E_{\Jcal^{\pi^\ast,\mathbf{p},\tau}}\bigg[ \Bignorm{ S_{\pi^\ast}    -  \frac{1}{1-\tau} \mathbf{V}^\textup{eff}(\tau)^{-1} \FisherInfo{\gamma^\ast(\tau)}^{-1} S_{0}^{\text{sub},\tau}  }{2}^2\bigg]  &\leq \E_{\Jcal^{\pi^\ast,\mathbf{p},\tau}}\bigg[ \Bignorm{ S_{\pi^\ast}    -  K S_{0}^{\text{sub},\tau}}{2}^2\bigg]  \quad \forall  K \in \mathbb{R}^{m \times m},
\end{align*}
which means that the equality in display \eqref{Ksub Projection Matrix Identity} is indeed correct, because $\frac{1}{1-\tau} \mathbf{V}^\textup{eff}(\tau)^{-1} \FisherInfo{\gamma^\ast(\tau)}^{-1} S_{0}^{\text{sub},\tau} $ is the matrix $K$ for which the residual $\E_{\Jcal^{\pi^\ast,\mathbf{p},\tau}}\bigg[ \Bignorm{ S_{\pi^\ast}    -  K S_{0}^{\text{sub},\tau}}{2}^2\bigg]$ is the smallest. As was reasoned earlier, this implies that $\Var_{\Jcal^{\pi^\ast,\mathbf{p},\tau}}[S_{\pi^\ast}    -  \big( Ds_{\pi^\ast} -  \mathbf{V}^\textup{eff}(\tau)^{-1}\psi_\tau^\textup{eff} \big)] = \CRLB{\mathbf{M}^{\textup{sub},\tau}}^{-1}$. So, by line \eqref{tough ping pong}, it follows that 
\begin{align*}
	\CRLB{\mathbf{M}^{\widetilde{\textup{sub}},\tau}}^{-1} &\succeq \Var_{\Jcal^{\pi^\ast,\mathbf{p},\tau}}\Big[ S_{\pi^\ast}    -  \big( Ds_{\pi^\ast} -  \mathbf{V}^\textup{eff}(\tau)^{-1}\psi_\tau^\textup{eff} \big)\Big] \\
	&= \CRLB{\mathbf{M}^{\textup{sub},\tau}}^{-1},
\end{align*}
implying that $\CRLB{\mathbf{M}^{\widetilde{\textup{sub}},\tau}} \preceq \CRLB{\mathbf{M}^{\textup{sub},\tau}}$, as claimed by the present Lemma! Lastly, notice that, by line \eqref{variance matrix of residual}, we have that $\Var_{\Jcal^{\pi^\ast,\mathbf{p},\tau}}\Big[S_{\pi^\ast} - \big( Ds_{\pi^\ast} -  \mathbf{V}^\textup{eff}(\tau)^{-1}\psi_\tau^\textup{eff} \big)\Big] =  \mathbf{V}^\textup{eff}(\tau)^{-1}$, so $\CRLB{\mathbf{M}^{\textup{sub},\tau}}^{-1} = \mathbf{V}^\textup{eff}(\tau)^{-1}$, i.e., $\CRLB{\mathbf{M}^{\textup{sub},\tau}} = \mathbf{V}^\textup{eff}(\tau)$, as was also claimed.

\end{proof}
\vspace{0.4in}

\subsection{$\tau_n$-IID Regime}


\subsubsection{Assumptions}

We assume that the sequence $\tau_n$ satisfies Assumptions \eqref{tau n is inside open interval 0 and 1} and \eqref{tau n convergence not too fast}. We also assume Assumption \eqref{Tau n Regime: lambda and uniform convergence properties}, which concerns the complexity of the space $\H$. Finally, we assume access to an estimator $\widehat{\gamma}$ that satisfies conditions in \eqref{tau n regime, estimator of gamma bar}. All of the foregoing assumptions can be found in the Theoretical Results section of the paper.

\subsubsection{Definition of  $\curlyE$}

The family of quantifiers $\curlyE$ is defined in the Theoretical Results section of the paper (see Definitions \eqref{Definition of E estimator: it should be tau RAL for each tau}, \eqref{Definition of E estimator: second order error to zero with changing tau n} and \eqref{Definition of E estimator: moment condition on influence function with changing tau n}).


\subsubsection{Lemmas}

\begin{lemma}[\uline{Asymptotics of a Sample Proportion}] \label{Asymptotics of a Sample Proportion}
Let $\varphi_n \in (0,1)$ be a sequence with the property that $n\varphi_n \to \infty$. For each $n$, let $\zeta_{1,n},\dots,\zeta_{n,n} \simiid \textup{Bern}(\varphi_n)$ and define 
\[
	\widehat{\varphi}_n := \frac{1}{n}\sum_{i=1}^n\zeta_{i,n}.
\]
Then, the following asymptotic statements are true:
\[
	\frac{\widehat{\varphi}_n - \varphi_n}{\varphi_n} = O_{\P}\bigg( \frac{1}{\sqrt{n\varphi_n}}\bigg)
\]
\[
	\frac{\widehat{\varphi}_n - \varphi_n}{\widehat{\varphi}_n} = O_{\P}\bigg( \frac{1}{\sqrt{n\varphi_n}}\bigg)
\]
\[
	 \frac{1}{n\widehat{\varphi}_n} =  \frac{1}{n\varphi_n} +   O_{\P}\bigg( \frac{1}{(n\varphi_n)^{3/2}} \bigg)
\]
\end{lemma}
\vspace{0.4in}

\begin{lemma}[\uline{Rate of $\abs{\gamma_j^\ast - \gamma_j^\ast(\tau_n)}$}] \label{Rate of gamma star sample size convergence}
Define the random vector  $\gamma^\ast := (\gamma_1^\ast,\dots, \gamma_m^\ast)$ where 
\begin{equation}\label{definition of random gamma ast}
	\gamma_j^\ast := \frac{\frac{\pi_j^\ast}{N^\test} + \frac{1}{N^\train} \frac{(\pi_j^\ast)^2}{\widehat{\pi}_j^\train}  }{ \frac{1}{N^\test} + \frac{1}{N^\train} \sum_{k=0}^m \frac{( \pi_k^\ast)^2}{\widehat{\pi}_k^\train } } \qquad 	\forall \  j \in[m],
\end{equation}
Then, under Assumptions \eqref{tau n is inside open interval 0 and 1} and \eqref{tau n convergence not too fast}, we have for each $j\in[m]$ that:
\[
	\abs{\gamma_j^\ast - \gamma_j^\ast(\tau_n)}  = O_{\P}\bigg( \frac{1}{\sqrt{n\tau_n(1-\tau_n)} }\bigg).
\]
\end{lemma}
\vspace{0.4in}

\begin{lemma}[\uline{Error of $s_{\gamma^\ast}$ vs. $s_{\gamma^\ast(\tau_n)}$ in $\tau_n$-IID Regime}] \label{Error of gamma star difference in tau n IID Regime}
Under Assumptions \eqref{tau n is inside open interval 0 and 1} and \eqref{tau n convergence not too fast}, we have that 
\small
\begin{align*}
	\Biggnorm{ \frac{1}{\sqrt{n}} \sum_{i=1}^n \frac{D_i}{\sqrt{\tau_n}}\Big[\big( s_{\gamma^\ast}(X_i) - \E_{\pi^\ast}[s_{\gamma^\ast}] \big)  - \big( s_{\gamma^\ast(\tau_n)}(X_i) - \E_{\pi^\ast}[s_{\gamma^\ast(\tau_n)}] \big) \Big] \bigg/ \sum_{k=0}^m \absbig{ \gamma_{k}^\ast - \gamma_{k}^\ast(\tau_n) }}{2}
\end{align*}
\normalsize
is $O_{\P}(1)$ and 
\small
\begin{align*}
	\Biggnorm{ \frac{1}{\sqrt{n}} \sum_{i=1}^{n} \frac{(1-D_i)\I{Y_i=j}}{ \sqrt{(1-\tau_n)\pi_j^\train}  } \Big[\big( s_{\gamma^\ast(\tau_n)}(X_i) - \E_{j}[s_{\gamma^\ast(\tau_n)}] \big) - \big( s_{\gamma^\ast}(X_i) - \E_{j}[s_{\gamma^\ast}] \big) \Big] \bigg/ \sum_{k=0}^m \absbig{ \gamma_{k}^\ast - \gamma_{k}^\ast(\tau_n) } }{2}
\end{align*}
\normalsize
is $O_{\P}(1)$ for each $j\in[m]$.

\end{lemma}
\vspace{0.4in}

\begin{lemma}[\uline{Lower Bound on $\minEval{\FisherInfo{\gamma^\ast}}$}] \label{Lower Bound on min eval of Fisher Info with random gamma ast}
Under Assumption  \eqref{Fixed Tau Regime: Mixture FIM not too small}, we have that
\[
	\minEval{\FisherInfo{\gamma^\ast}} > \Lambda \frac{L^2}{m^2} 
\]
with probability $1$. 
\end{lemma}
\vspace{0.4in}

\begin{lemma}[\uline{$\widehat{\pi}_n$ is $\tau$-RAL w.r.t $\textbf{M}^{\text{semi},\tau}$ in Fixed $\tau$-IID Regime}] \label{pi hat is RAL}
Under Assumptions \eqref{Fixed Tau Regime: Mixture FIM not too small}, \eqref{Fixed Tau Regime: Cat minus Mixture FIM}, \eqref{Fixed Tau Regime: expo family}, \eqref{Fixed Tau Regime: H closed under scalar multiplication}, \eqref{Fixed Tau Regime: Omega Properties},  \eqref{Tau n Regime: lambda and uniform convergence properties} and \eqref{tau n regime, estimator of gamma bar}, the estimator $\widehat{\pi}_n$ is regular in the Fixed $\tau$-IID Regime, as well as asymptotically linear with influence function $\psi_\tau^\eff$ and second order error $\delta_\tau$. 
\end{lemma}
\vspace{0.4in}

\subsubsection{Proofs}

\begin{proof}[\uline{Proof of Lemma \eqref{Asymptotics of a Sample Proportion}}]

First, we will prove that $\frac{\widehat{\varphi}_n - \varphi_n}{\varphi_n} = O_{\P}\Big( \frac{1}{\sqrt{n\varphi_n}}\Big)$. Observe that:
\begin{align*}
	\Var\bigg[ \frac{\widehat{\varphi}_n - \varphi_n}{\varphi_n} \bigg] &= \frac{1}{\varphi_n^2 }\Var\big[ \widehat{\varphi}_n  \big] \\ 
	&= \frac{\varphi_n(1-\varphi_n)}{\varphi_n^2 n }  \\ 
	&\leq \frac{1}{\varphi_n n },
\end{align*}
which since $\E\Big[\frac{\widehat{\varphi}_n - \varphi_n}{\varphi_n}\Big] = 0$, implies that 
\[
	\implies \frac{\widehat{\varphi}_n - \varphi_n}{\varphi_n} = O_{\P}\bigg(  \frac{1}{ \sqrt{\varphi_n n }} \bigg),
\]
	
as desired. Second, we will prove that $\frac{\widehat{\varphi}_n - \varphi_n}{\widehat{\varphi}_n} = O_{\P}\Big( \frac{1}{\sqrt{n\varphi_n}}\Big)$. Towards that end, it will be helpful to first demonstrate that $  \frac{\sqrt{n}(\widehat{\varphi}_n - \varphi_n)}{\sqrt{\widehat{\varphi}_n}}  = O_{\P}(1)$. Note that for any $a >0$, we have that:
\begin{align*}
	\P \Bigg[ \sqrt{n}\absbigg{ \frac{\widehat{\varphi}_n - \varphi_n}{\sqrt{\widehat{\varphi}_n}} }  \geq a\Bigg] &= \P \bigg[  \sqrt{n}\absbig{\widehat{\varphi}_n - \varphi_n  }  \geq a\sqrt{\widehat{\varphi}_n}\bigg] \\ 
	&= \P \bigg[  n(\widehat{\varphi}_n - \varphi_n)^2  \geq a^2 \widehat{\varphi}_n \bigg] \\  
	&= \P \bigg[  n(\widehat{\varphi}_n - \varphi_n)^2  \geq a^2\varphi_n + a^2(\widehat{\varphi}_n-\varphi_n) \bigg] \\ 
	&\leq \P \bigg[  a^2\abs{\widehat{\varphi}_n-\varphi_n} + n(\widehat{\varphi}_n - \varphi_n)^2  \geq a^2\varphi_n \bigg] \\  
	&\leq \P \bigg[  a^2\abs{\widehat{\varphi}_n-\varphi_n}   \geq a^2\varphi_n/2 \bigg]  + \P \bigg[   n(\widehat{\varphi}_n - \varphi_n)^2   \geq a^2\varphi_n/2 \bigg] \\  
	&= \P \bigg[ \abs{\widehat{\varphi}_n-\varphi_n}   \geq \varphi_n/2 \bigg]  + \P \bigg[   \abs{\widehat{\varphi}_n - \varphi_n}   \geq a \sqrt{\frac{\varphi_n}{2n} } \bigg] \\  
	&\leq 4\frac{\Var[\widehat{\varphi}_n] }{\varphi_n^2} + \frac{2n}{\varphi_n} \frac{\Var[\widehat{\varphi}_n]}{a^2} \\ 
	&\leq 4\frac{ \frac{\varphi_n}{n} }{\varphi_n^2} + \frac{2n}{\varphi_n} \frac{ \frac{\varphi_n}{n} }{a^2} \\ 
	&= \frac{ 4 }{n\varphi_n} +  \frac{ 2}{a^2},
\end{align*}
where the second to last inequality is due to Chebychev, and the last inequality is because $0 < 1-\varphi_n < 1$. Now, since $n\varphi_n  \to \infty$, the work above shows that the LHS can be made arbitrarily small by choosing $n$ and $a$ to be sufficiently large, so indeed, $\frac{\sqrt{n}(\widehat{\varphi}_n - \varphi_n)}{\sqrt{\widehat{\varphi}_n}} = O_{\P}(1)$. Thus:
\begin{align*}
	 \absbigg{\frac{\widehat{\varphi}_n - \varphi_n}{\widehat{\varphi}_n}} &=  \frac{1}{\sqrt{\widehat{\varphi}_n}} \absbigg{\frac{\widehat{\varphi}_n - \varphi_n}{\sqrt{\widehat{\varphi}_n}} } \\
	 &=  \frac{1}{\sqrt{n\widehat{\varphi}_n}} \sqrt{n} \absbigg{\frac{\widehat{\varphi}_n - \varphi_n}{\sqrt{\widehat{\varphi}_n}} } \\
	 &=  \frac{1}{\sqrt{n\widehat{\varphi}_n}} \cdot O_{\P}(1).
\end{align*}
Now, we can also show that $\frac{1}{\sqrt{\widehat{\varphi}_n n}} = O_{\P}\Big(   \frac{1}{\sqrt{\varphi_n n}}   \Big)$. Towards that end, observe that:
\begin{align*}
	\absBigg{ \frac{1}{\sqrt{\widehat{\varphi}_n}} - \frac{1}{\sqrt{\varphi_n}} } \sqrt{n}\varphi_n &= \absBigg{  \frac{\sqrt{\varphi_n} - \sqrt{\widehat{\varphi}_n} }{\sqrt{\widehat{\varphi}_n}\sqrt{\varphi_n} }   } \sqrt{n} \varphi_n \\ 
	&= \absBigg{  \frac{\varphi_n - \widehat{\varphi}_n }{\sqrt{\widehat{\varphi}_n}\sqrt{\varphi_n} (\sqrt{\varphi_n} + \sqrt{\widehat{\varphi}_n }) }   } \sqrt{n} \varphi_n \\ 
	&= \sqrt{n}\absBigg{    \frac{\varphi_n - \widehat{\varphi}_n }{ \sqrt{\widehat{\varphi}_n} }      }\cdot   \frac{ \varphi_n }{\sqrt{\varphi_n} (\sqrt{\varphi_n} + \sqrt{\widehat{\varphi}_n }) }       \\ 
	&= \sqrt{n}\absBigg{    \frac{\varphi_n - \widehat{\varphi}_n }{ \sqrt{\widehat{\varphi}_n} }      }\cdot  \frac{ \sqrt{\varphi_n} }{ \sqrt{\varphi_n} + \sqrt{\widehat{\varphi}_n } }       \\ 
	&\leq \sqrt{n}\absBigg{    \frac{\varphi_n - \widehat{\varphi}_n }{ \sqrt{\widehat{\varphi}_n} }      }.
\end{align*}
Since we already showed that $\frac{\sqrt{n}(\widehat{\varphi}_n - \varphi_n)}{\sqrt{\widehat{\varphi}_n}} = O_{\P}(1)$, the work above implies that:
\[
	\implies \absBig{ \frac{1}{\sqrt{\widehat{\varphi}_n}} - \frac{1}{\sqrt{\varphi_n}} } \sqrt{n}\varphi_n  = O_{\P}(1)
\]
\[
	\implies \frac{1}{\sqrt{\widehat{\varphi}_n}} - \frac{1}{\sqrt{\varphi_n}} = O_{\P}\bigg( \frac{1}{\sqrt{n}\varphi_n} \bigg)
\]
\begin{equation}\label{GHIRARDELLI CHOCOLATE}
	\implies \frac{1}{\sqrt{n\widehat{\varphi}_n}}  =  \frac{1}{\sqrt{n\varphi_n}} + O_{\P}\bigg( \frac{1}{n\varphi_n} \bigg)
\end{equation}
\[
	\implies \frac{1}{\sqrt{n\widehat{\varphi}_n}}  = O_{\P}\bigg(  \frac{1}{\sqrt{n\varphi_n}}  \bigg).
\]
Thus, we have that:
\[
	\implies  \absbigg{\frac{\widehat{\varphi}_n - \varphi_n}{\widehat{\varphi}_n}} = O_{\P}\bigg(  \frac{1}{\sqrt{n\varphi_n}}  \bigg).
\]
Finally, note that, by line \eqref{GHIRARDELLI CHOCOLATE}, we have that:
\begin{align*}
	 \frac{1}{n\widehat{\varphi}_n} &=  \frac{1}{n\varphi_n} +   O_{\P}\bigg( \frac{1}{(n\varphi_n)^{3/2}} +  \frac{1}{(n\varphi_n)^{2}} \bigg) \\ 
	 &=  \frac{1}{n\varphi_n} +   O_{\P}\bigg( \frac{1}{(n\varphi_n)^{3/2}} \bigg),
\end{align*}
where the last line is because $n\varphi_n \to\infty$.

\end{proof}
\vspace{0.4in}

\begin{proof}[\uline{Proof of Lemma \eqref{Rate of gamma star sample size convergence}}]

Define the following notation for each $k\in\Y$:
\begin{itemize}
	\item $\widehat{\psi}_k := \frac{\pi_k^\ast}{N^\test} + \frac{1}{N^\train} \frac{(\pi_k^\ast)^2}{\widehat{\pi}_k^\train} $
	\item $\widehat{\Psi} := \frac{1}{N^\test} + \frac{1}{N^\train} \sum_{l=0}^m \frac{( \pi_l^\ast)^2}{\widehat{\pi}_l^\train }$
	\item $\psi_k := \frac{\pi_k^\ast}{\tau_n n} + \frac{1}{(1-\tau_n)n} \frac{(\pi_k^\ast)^2}{\pi_k^\train} $
	\item $\Psi :=  \frac{1}{\tau_n n}  + \frac{1}{(1-\tau_n)n} \sum_{l=0}^{m} \frac{(\pi_l^\ast)^2}{\pi_l^\train}$.
\end{itemize}
 Clearly, $\gamma_k^\ast = \widehat{\psi}_k / \widehat{\Psi}$ and $\gamma_k^\ast(\tau_n) = \psi_k / \Psi$. Now, let any $j\in\Y$ and $\epsilon > 0$ be given, and define $\delta := \frac{\epsilon}{\sqrt{n\tau_n(1-\tau_n)}}$. It is easy to show that:
\[
	\gamma_j^\ast - \gamma_j^\ast(\tau_n) = \frac{\widehat{\psi}_j - \psi_j }{\widehat{\Psi} } + \frac{\psi_j}{\Psi} \frac{ \Psi - \widehat{\Psi}}{\widehat{\Psi}},
\]
from which it follows that
\begin{align*}
	\implies \abs{\gamma_j^\ast - \gamma_j^\ast(\tau_n)} &\leq  \frac{ \abs{ \widehat{\psi}_j - \psi_j }  }{\widehat{\Psi} } +  \frac{ \abs{   \widehat{\Psi} - \Psi }}{\widehat{\Psi}} \\ 
	&\leq  \frac{ \abs{ \widehat{\psi}_j - \psi_j }}{\widehat{\Psi}} + \frac{\sum_{k=0}^m \abs{   \widehat{\psi}_k - \psi_k } }{\widehat{\Psi} } \\
	&\leq  \frac{ \abs{ \widehat{\psi}_j - \psi_j }}{\widehat{\psi}_j} + \sum_{k=0}^m \frac{ \abs{   \widehat{\psi}_k - \psi_k } }{\widehat{\psi}_k } \\ 
	&\leq 2 \sum_{k=0}^m \frac{ \abs{   \widehat{\psi}_k - \psi_k } }{\widehat{\psi}_k }.
\end{align*}
Consequentially: 
\begin{align*}
	&\P_{D_{1:n},Y_{1:n}} \Big[  \abs{\gamma_j^\ast - \gamma_j^\ast(\tau_n)}  \geq \delta \Big] \\
	&\leq \P_{D_{1:n},Y_{1:n}} \Bigg[  \sum_{k=0}^m \frac{ \abs{   \widehat{\psi}_k - \psi_k } }{\widehat{\psi}_k } \geq \frac{\delta}{2} \Bigg]  \\ 
	&\leq \sum_{k=0}^m  \P_{D_{1:n},Y_{1:n}} \Bigg[  \frac{ \abs{   \widehat{\psi}_k - \psi_k } }{\widehat{\psi}_k } \geq \frac{\delta}{2m} \Bigg]  \\ 
	&= \sum_{k=0}^m  \P_{D_{1:n},Y_{1:n}} \Big[   \abs{   \widehat{\psi}_k - \psi_k   }  \geq \frac{\delta}{2m} \widehat{\psi}_k  \Big]  \\ 
	&\leq \sum_{k=0}^m  \P_{D_{1:n},Y_{1:n}} \Bigg[   \absbigg{   \frac{\pi_k^\ast}{N^\test} - \frac{\pi_k^\ast}{\tau_n n}   } + \absbigg{    \frac{1}{N^\train} \frac{(\pi_k^\ast)^2}{\widehat{\pi}_k^\train} - \frac{1}{(1-\tau_n)n} \frac{(\pi_k^\ast)^2}{\pi_k^\train}     }   \geq \frac{\delta}{2m} \bigg(\frac{\pi_k^\ast}{N^\test} + \frac{1}{N^\train} \frac{(\pi_k^\ast)^2}{\widehat{\pi}_k^\train} \bigg)  \Bigg]  \\ \\
	&\leq  \sum_{k=0}^m  \P_{D_{1:n},Y_{1:n}} \Bigg[   \absbigg{   \frac{\pi_k^\ast}{N^\test} - \frac{\pi_k^\ast}{\tau_n n}   }  \geq \frac{\delta}{2m} \frac{\pi_k^\ast}{N^\test}    \Bigg] \\
	&\qquad\qquad+  \sum_{k=0}^m  \P_{D_{1:n},Y_{1:n}} \Bigg[   \absbigg{    \frac{1}{N^\train} \frac{(\pi_k^\ast)^2}{\widehat{\pi}_k^\train} - \frac{1}{(1-\tau_n)n} \frac{(\pi_k^\ast)^2}{\pi_k^\train}     }   \geq \frac{\delta}{2m}  \frac{1}{N^\train} \frac{(\pi_k^\ast)^2}{\widehat{\pi}_k^\train}   \Bigg] \\ \\ 
	&= (m+1) \P_{D_{1:n},Y_{1:n}} \Bigg[   \absbigg{   \frac{1}{N^\test} - \frac{1}{\tau_n n}   }  \geq \frac{\delta}{2m} \frac{1}{N^\test}    \Bigg] \\
	&\qquad\qquad+  \sum_{k=0}^m  \P_{D_{1:n},Y_{1:n}} \Bigg[   \absbigg{    \frac{1}{\widehat{\pi}_k^\train N^\train}  - \frac{1}{\pi_k^\train(1-\tau_n)n}      }   \geq \frac{\delta}{2m}  \frac{1}{\widehat{\pi}_k^\train N^\train}    \Bigg] \\ \\ 
	&=  (m+1)  \P_{D_{1:n},Y_{1:n}} \Bigg[      \frac{ \absbig{N^\test - \tau_n n} }{\tau_n n}      \geq \frac{\delta}{2m}     \Bigg] +  \sum_{k=0}^m  \P_{D_{1:n},Y_{1:n}} \Bigg[       \frac{ \absbig{\widehat{\pi}_k^\train N^\train - \pi_k^\train(1-\tau_n)n} }{\pi_k^\train(1-\tau_n)n}        \geq \frac{\delta}{2m}      \Bigg] \\ 
	&=  (m+1)  \P_{D_{1:n},Y_{1:n}} \Bigg[      \frac{ \absbig{N^\test/n - \tau_n } }{\tau_n }      \geq \frac{\delta}{2m}     \Bigg] +  \sum_{k=0}^m  \P_{D_{1:n},Y_{1:n}} \Bigg[       \frac{ \absbig{\widehat{\pi}_k^\train N^\train/n - \pi_k^\train(1-\tau_n)} }{\pi_k^\train(1-\tau_n)}        \geq \frac{\delta}{2m}      \Bigg].
\end{align*}

Note that:
\begin{align*}
	\P_{D_{1:n},Y_{1:n}} \Bigg[      \frac{ \absbig{N^\test/n - \tau_n } }{\tau_n }      \geq \frac{\delta}{2m}     \Bigg] &= \P_{D_{1:n},Y_{1:n}} \Bigg[      \frac{ \absbig{N^\test/n - \tau_n } }{\tau_n }      \geq \frac{1}{2m}  \frac{\epsilon}{\sqrt{n\tau_n(1-\tau_n)}}   \Bigg] \\
	&\leq \P_{D_{1:n},Y_{1:n}} \Bigg[      \frac{ \absbig{N^\test/n - \tau_n } }{\tau_n }      \geq \frac{\epsilon}{2m}  \frac{1}{\sqrt{n\tau_n}}   \Bigg] \\
	&= \P_{D_{1:n},Y_{1:n}} \Bigg[      \sqrt{n\tau_n} \frac{ \absbig{N^\test/n - \tau_n } }{\tau_n }      \geq \frac{\epsilon}{2m}     \Bigg],
\end{align*}
and likewise:
\small
\[
	 \P_{D_{1:n},Y_{1:n}} \Bigg[       \frac{ \absbig{\widehat{\pi}_k^\train N^\train/n - \pi_k^\train(1-\tau_n)} }{\pi_k^\train(1-\tau_n)}        \geq \frac{\delta}{2m}      \Bigg] \leq  \P_{D_{1:n},Y_{1:n}} \Bigg[ \sqrt{n(1-\tau_n)}      \frac{ \absbig{\widehat{\pi}_k^\train N^\train/n - \pi_k^\train(1-\tau_n)} }{\pi_k^\train(1-\tau_n)}        \geq \frac{\epsilon}{2m}      \Bigg].
\]
\normalsize
Ergo, we have that:
\begin{align*}
	&\P_{D_{1:n},Y_{1:n}} \Big[ \sqrt{n\tau_n(1-\tau_n)} \abs{\gamma_j^\ast - \gamma_j^\ast(\tau_n)}  \geq \epsilon \Big] \\
	 &= \P_{D_{1:n},Y_{1:n}} \Big[  \abs{\gamma_j^\ast - \gamma_j^\ast(\tau_n)}  \geq  \delta \Big] \\  
	&\leq (m+1) \P_{D_{1:n},Y_{1:n}} \Bigg[      \sqrt{n\tau_n} \frac{ \absbig{N^\test/n - \tau_n } }{\tau_n }      \geq \frac{\epsilon}{2m}     \Bigg]   \\
	&\qquad\qquad+    \sum_{k=0}^m  \P_{D_{1:n},Y_{1:n}} \Bigg[ \sqrt{n(1-\tau_n)}      \frac{ \absbig{\widehat{\pi}_k^\train N^\train/n - \pi_k^\train(1-\tau_n)} }{\pi_k^\train(1-\tau_n)}        \geq \frac{\epsilon}{2m}      \Bigg].
\end{align*}

Due to Assumptions \eqref{tau n is inside open interval 0 and 1} and \eqref{tau n convergence not too fast},   Lemma \eqref{Asymptotics of a Sample Proportion}  implies that $\sqrt{n\tau_n} \frac{ \abs{N^\test/n - \tau_n } }{\tau_n }  = O_{\P}(1)$, and since $\pi_k^\train$ is fixed, $\sqrt{n(1-\tau_n)}      \frac{ \abs{\widehat{\pi}_k^\train N^\train/n - \pi_k^\train(1-\tau_n)} }{\pi_k^\train(1-\tau_n)}  = O_{\P}(1)$. Thus, by the display above, it follows that 
\[
	\P_{D_{1:n},Y_{1:n}} \Big[ \sqrt{n\tau_n(1-\tau_n)} \abs{\gamma_j^\ast - \gamma_j^\ast(\tau_n)}  \geq \epsilon \Big]
\] 
can be made arbitrarily small by choosing $n$ and $\epsilon$ to be sufficiently large, so  $\sqrt{n\tau_n(1-\tau_n)} \abs{\gamma_j^\ast - \gamma_j^\ast(\tau_n)}  = O_{\P}(1)$, i.e.,
\[
	\abs{\gamma_j^\ast - \gamma_j^\ast(\tau_n)}  = O_{\P}\bigg( \frac{1}{\sqrt{n\tau_n(1-\tau_n)} }\bigg),
\]
as desired!

\end{proof}
\vspace{0.4in}

\begin{proof}[\uline{Proof of Lemma \eqref{Error of gamma star difference in tau n IID Regime}}]

The proof for the second asymptotic statement is equivalent to the proof for the first statement, so for conciseness, we only focus on proving the first statement. Towards that end, define
\begin{align*}
	\phi_i^\test &:=  \frac{D_i}{\sqrt{\tau_n}}\Big[ \big( s_{\gamma^\ast}(X_i) - \E_{\pi^\ast}[s_{\gamma^\ast}] \big) - \big( s_{\gamma^\ast(\tau_n)}(X_i) - \E_{\pi^\ast}[s_{\gamma^\ast(\tau_n)}] \big) \Big] \Bigg/ \sum_{k=0}^m \abs{\gamma_k^\ast - \gamma_k^\ast(\tau)}   \in \mathbb{R}^m.
\end{align*}
We want to prove that:
\begin{align*}
	 \Biggnorm{ \frac{1}{\sqrt{n}} \sum_{i=1}^n \phi_i^\test }{2} = O_{\P}(1).
\end{align*}
So, let any $\delta > 0$ be given. Observe that:
\begin{align*}
	&\P_{Z_{1:n}}\Bigg[ \biggnorm{ \frac{1}{\sqrt{n}}\sum_{i=1}^n\phi_i^\test }{2} \geq \delta \Bigg]\\
	 &= \E_{N^\test, \widehat{\pi}^\train }\Bigg\{ \P_{Z_{1:n} \mid N^\test, \widehat{\pi}^\train}\Bigg[ \biggnorm{ \sum_{i=1}^n\phi_i^\test }{2} \geq \sqrt{n}\delta \Bigg] \Bigg\} \\
	&\leq \E_{N^\test, \widehat{\pi}^\train }\Bigg\{ \P_{Z_{1:n} \mid N^\test, \widehat{\pi}^\train}\Bigg[ \biggnorm{ \sum_{i=1}^n\phi_i^\test }{1} \geq \sqrt{n}\delta \Bigg] \Bigg\} \\
	&\leq \sum_{j=1}^m \E_{N^\test, \widehat{\pi}^\train }\Bigg\{ \P_{Z_{1:n} \mid N^\test, \widehat{\pi}^\train}\Bigg[ \absbigg{ \sum_{i=1}^n\phi_{i,j}^\test } \geq \frac{\sqrt{n}\delta}{m} \Bigg] \Bigg\} \\ \\ 
	&= \sum_{j=1}^m \E_{N^\test, \widehat{\pi}^\train }\Bigg\{ \P_{Z_{1:n} \mid N^\test, \widehat{\pi}^\train}\Bigg[ \absbigg{ \sum_{i=1}^n\phi_{i,j}^\test } \geq \frac{\sqrt{n}\delta}{m} \Bigg]\I{ \widehat{\pi}_y^\train \in (\xi,1-\xi)\ \forall y\in\Y}  \Bigg\} \\
	&\qquad\qquad +  \sum_{j=1}^m \E_{N^\test, \widehat{\pi}^\train }\Bigg\{ \P_{Z_{1:n} \mid N^\test, \widehat{\pi}^\train}\Bigg[ \absbigg{ \sum_{i=1}^n\phi_{i,j}^\test } \geq \frac{\sqrt{n}\delta}{m} \Bigg]\I{ \exists \ y \in \Y \text{ s.t. } \widehat{\pi}_y^\train \notin (\xi,1-\xi)}  \Bigg\} \\ \\ 
	&\leq \sum_{j=1}^m \E_{N^\test, \widehat{\pi}^\train }\Bigg\{ \P_{Z_{1:n} \mid N^\test, \widehat{\pi}^\train}\Bigg[ \absbigg{ \sum_{i=1}^n\phi_{i,j}^\test } \geq \frac{\sqrt{n}\delta}{m} \Bigg]\I{ \widehat{\pi}_y^\train \in (\xi,1-\xi)\ \forall y\in\Y}  \Bigg\} \\
	&\qquad\qquad +  m\P_{N^\test, \widehat{\pi}^\train }\Big[  \exists \ y \in \Y \text{ s.t. } \widehat{\pi}_y^\train \notin (\xi,1-\xi)  \Big] \\ \\ 
	&= \sum_{j=1}^m \E_{N^\test, \widehat{\pi}^\train }\Bigg\{ \P_{Z_{1:n} \mid N^\test, \widehat{\pi}^\train}\Bigg[ \absbigg{ \sum_{i=1}^n\phi_{i,j}^\test } \geq \frac{\sqrt{n}\delta}{m} \Bigg]\I{ \widehat{\pi}_y^\train \in (\xi,1-\xi)\ \forall y\in\Y}\I{ N^\test \geq 1}  \Bigg\} \\
	&\qquad\qquad +  m\P_{N^\test, \widehat{\pi}^\train }\Big[  \exists \ y \in \Y \text{ s.t. } \widehat{\pi}_y^\train \notin (\xi,1-\xi)  \Big],
\end{align*}
where the last line is because if $N^\test = 0$, then $\sum_{i=1}^n\phi_{i,j}^\test = 0$ with probability $1$,  in which case we must have that  $\P_{Z_{1:n} \mid N^\test, \widehat{\pi}^\train}\Big[ \absbig{ \sum_{i=1}^n\phi_{i,j}^\test } \geq \frac{\sqrt{n}\delta}{m} \Big] = 0$ since $\delta > 0$. Next, let's focus on the $\P_{N^\test, \widehat{\pi}^\train }\Big[  \exists \ y \in \Y \text{ s.t. } \widehat{\pi}_y^\train \notin (\xi,1-\xi)  \Big]$ term. Note that, for each $y\in\Y$, we have that $\pi_y^\train \in (\xi,1-\xi)$, so if $\widehat{\pi}_y^\train \notin (\xi,1-\xi)$, then $\abs{\widehat{\pi}_y^\train - \pi_y^\train} \geq (\pi_y^\train - \xi) \wedge (1-\xi - \pi_y^\train)$. Thus:
\begin{align*}
	&\P_{N^\test, \widehat{\pi}^\train }\Big[  \exists \ y \in \Y \text{ s.t. } \widehat{\pi}_y^\train \notin (\xi,1-\xi)  \Big] \\
	&\leq \sum_{y = 0}^m \P_{N^\test, \widehat{\pi}^\train }\Big[  \widehat{\pi}_y^\train \notin (\xi,1-\xi)  \Big] \\ 
	&\leq \sum_{y = 0}^m \P_{N^\test, \widehat{\pi}^\train }\Big[  \abs{\widehat{\pi}_y^\train - \pi_y^\train} \geq (\pi_y^\train - \xi) \wedge (1-\xi - \pi_y^\train) \Big] \\ 
	&= \sum_{y = 0}^m \E_{N^\test}\Bigg\{ \P_{\widehat{\pi}^\train \mid N^\test}\Big[ (n-N^\test) \abs{\widehat{\pi}_y^\train - \pi_y^\train} \geq (n-N^\test)\cdot (\pi_y^\train - \xi) \wedge (1-\xi - \pi_y^\train) \Big] \Bigg\}.
\end{align*}
Note that, conditional on $N^\test$, the random variable $(n-N^\test)\widehat{\pi}_y^\train$ is the sum of $n-N^\test$ independent $\text{Bernoulli}(\pi_y^\train)$ random variables. Thus, by Hoeffding's inequality:
\begin{align*}
	&\P_{\widehat{\pi}^\train \mid N^\test}\Big[ (n-N^\test) \abs{\widehat{\pi}_y^\train - \pi_y^\train} \geq (n-N^\test)\cdot (\pi_y^\train - \xi) \wedge (1-\xi - \pi_y^\train) \Big] \\
	&\leq 2\exp\bigg(-2(n-N^\test)\Big((\pi_y^\train - \xi) \wedge (1-\xi - \pi_y^\train)  \Big)^2 \bigg) \\
	&\leq 2\exp\big(-c(n-N^\test) \big),
\end{align*}
where $c := 2 \min\limits_{0\leq y \leq m}\Big((\pi_y^\train - \xi) \wedge (1-\xi - \pi_y^\train)  \Big)^2$. Thus, we have that:
\begin{align*}
	\P_{N^\test, \widehat{\pi}^\train }\Big[  \exists \ y \in \Y \text{ s.t. } \widehat{\pi}_y^\train \notin (\xi,1-\xi)  \Big] &\leq \sum_{y = 0}^m \E_{N^\test}\Big\{  2\exp\big(-c(n-N^\test) \big) \Big\} \\ 
	&= 2(m+1) \E_{N^\test}\Big\{  \exp\big(-c(n-N^\test) \big) \Big\} \\ 
	&= 2(m+1) \Big(\big(1-[1-\tau_n]\big) + [1-\tau_n]e^{-c} \Big)^n \\ 
	&= 2(m+1) \Big(1 - [1-\tau_n](1-e^{-c}) \Big)^n \\ 
	&\leq 2(m+1) e^{-n[1-\tau_n](1-e^{-c})},
\end{align*}
where the third line uses the formula for the MGF of a Binomial random variable and the fact that $n - N^\test \sim \text{Binomial}(n,1-\tau_n)$, and the last line uses the fact that $[1-\tau_n](1-e^{-c}) \in (0,1)$ under Assumption \eqref{tau n is inside open interval 0 and 1} and $(1-u)^n \leq e^{-nu}$ for all $u\in(0,1)$. \\ 

Having obtained a bound on  $\P_{N^\test, \widehat{\pi}^\train }\Big[  \exists \ y \in \Y \text{ s.t. } \widehat{\pi}_y^\train \notin (\xi,1-\xi)  \Big]$, we now turn to bounding 
\[
	\sum_{j=1}^m \E_{N^\test, \widehat{\pi}^\train }\Bigg\{ \P_{Z_{1:n} \mid N^\test, \widehat{\pi}^\train}\Bigg[ \absbigg{ \sum_{i=1}^n\phi_{i,j}^\test } \geq \frac{\sqrt{n}\delta}{m} \Bigg]\I{ \widehat{\pi}_y^\train \in (\xi,1-\xi)\ \forall y\in\Y}\I{ N^\test \geq 1}   \Bigg\}. 
\]
Towards that end, we make the following observations about the display above. By conditioning on $N^\test, \widehat{\pi}^\train$, we have that $\gamma^\ast$ is no longer random, and $ \sum_{i=1}^n\phi_{i,j}^\test$ has the same distribution as a sum of $N^\test$ independent copies of 
\[
	   \frac{1}{\sqrt{\tau_n}}  \Big[ \big( s_{\gamma^\ast,j}(X) - \E_{\pi^\ast}[s_{\gamma^\ast,j}] \big) - \big( s_{\gamma^\ast(\tau),j}(X) - \E_{\pi^\ast}[s_{\gamma^\ast(\tau),j}] \big) \Big] \Bigg/ \sum_{k=0}^m \abs{\gamma_k^\ast - \gamma_k^\ast(\tau)} 
\]
where $X \sim p_{\pi^\ast}$. Additionally, note that if $\widehat{\pi}_y^\train \in (\xi,1-\xi)\ \forall y\in\Y$, then by Lemma \eqref{Bounded Gamma Star}, we have that $\gamma^\ast \in \Gamma$, and so by Lemma \eqref{Bounds On Score Function Differences at Different Parameters}, it follows that 
\begin{align*}
	  \frac{1}{\sqrt{\tau_n}} \absBig{ \big( s_{\gamma^\ast,j}(X) - \E_{\pi^\ast}[s_{\gamma^\ast,j}] \big) - \big( s_{\gamma^\ast(\tau),j}(X) - \E_{\pi^\ast}[s_{\gamma^\ast(\tau),j}] \big) } \Bigg/ \sum_{k=0}^m \abs{\gamma_k^\ast - \gamma_k^\ast(\tau)} &\leq \frac{4}{ \sqrt{\tau_n} L^2} ,
\end{align*}
i.e., 
\small
\begin{align*}
	  \frac{1}{\sqrt{\tau_n}}\Big[ \big( s_{\gamma^\ast,j}(X) - \E_{\pi^\ast}[s_{\gamma^\ast,j}] \big) - \big( s_{\gamma^\ast(\tau),j}(X) - \E_{\pi^\ast}[s_{\gamma^\ast(\tau),j}] \big) \Big] \Bigg/ \sum_{k=0}^m \abs{\gamma_k^\ast - \gamma_k^\ast(\tau)}  &\in \Bigg[-\frac{4}{ \sqrt{\tau_n} L^2} , \ \frac{4}{ \sqrt{\tau_n} L^2} \Bigg].
\end{align*}
\normalsize
Bringing these observations together, it follows from Hoeffding's inequality that:
\[
	\P_{Z_{1:n} \mid N^\test, \widehat{\pi}^\train}\Bigg[ \absbigg{ \sum_{i=1}^n\phi_{i,j}^\test } \geq \frac{\sqrt{n}\delta}{m} \Bigg]\I{ \widehat{\pi}_y^\train \in (\xi,1-\xi)\ \forall y\in\Y} \I{ N^\test \geq 1} 
\]
\[
	\leq  2\exp\bigg(- \frac{1}{32}\delta^2m^{-2} L^4 \cdot \frac{n\tau_n}{N^\test}   \bigg) \I{ \widehat{\pi}_y^\train \in (\xi,1-\xi)\ \forall y\in\Y} \I{ N^\test \geq 1} 
\]
\[
	\leq  2\exp\bigg(- \frac{1}{32}\delta^2m^{-2} L^4 \cdot \frac{n\tau_n}{1+N^\test}   \bigg)
\]
\[
	=  2\exp\bigg(-  \frac{h n\tau_n \delta^2}{1+N^\test}   \bigg).
\]
where 
\[
	h := \frac{1}{32}m^{-2} L^4 
\]

Ergo:
\small
\begin{align*}
	 &\sum_{j=1}^m \E_{N^\test, \widehat{\pi}^\train }\Bigg\{ \P_{Z_{1:n} \mid N^\test, \widehat{\pi}^\train}\Bigg[ \absbigg{ \sum_{i=1}^n\phi_{i,j}^\test } \geq \frac{\sqrt{n}\delta}{m} \Bigg]\I{ \widehat{\pi}_y^\train \in (\xi,1-\xi)\ \forall y\in\Y}\I{ N^\test \geq 1}   \Bigg\} \\
	 &\leq 2\sum_{j=1}^m \E_{N^\test }\Bigg\{  \exp\bigg(-  \frac{h n\tau_n \delta^2}{1+N^\test}   \bigg)  \Bigg\}.
\end{align*}
\normalsize

Now, observe that:
\begin{align*}
	 \E_{N^\test }\Bigg\{  \exp\bigg(-  \frac{h_n\delta^2}{1+N^\test}   \bigg)  \Bigg\} &=  \E_{N^\test \mid N^\test \geq h n\tau_n \delta}\Bigg\{  \exp\bigg(-  \frac{h_n\delta^2}{1+N^\test}   \bigg)  \Bigg\} \P[ N^\test \geq h n\tau_n \delta] \\
	 &\qquad\qquad + \E_{N^\test \mid N^\test < h n\tau_n  \delta}\Bigg\{  \exp\bigg(-  \frac{h n\tau_n  \delta^2}{1+N^\test}   \bigg)  \Bigg\}\P[ N^\test < h n\tau_n  \delta] \\ \\
	 &\leq \P[ N^\test \geq h n\tau_n \delta] +  \E_{N^\test \mid N^\test < h n\tau_n \delta}\Bigg\{  \exp\bigg(-  \frac{h n\tau_n \delta}{1+N^\test}\delta   \bigg)  \Bigg\}\\ 
	 &\leq \P[ N^\test \geq h n\tau_n \delta] +   \exp\bigg(-  \frac{h n\tau_n \delta}{1+h n\tau_n \delta} \delta  \bigg) \\ 
	 &\leq \frac{\E[N^\test]}{h n\tau_n \delta }  +   \exp\bigg(-  \frac{h n\tau_n  \delta}{1+h n\tau_n \delta}\delta   \bigg)\\
	 &= \frac{1}{h  \delta }  +   \exp\bigg(-  \frac{h n\tau_n  \delta}{1+h n\tau_n \delta}\delta   \bigg).
\end{align*}
Thus:
\small
\[
	\implies \sum_{j=1}^m \E_{N^\test, \widehat{\pi}^\train }\Bigg\{ \P_{Z_{1:n} \mid N^\test, \widehat{\pi}^\train}\Bigg[ \absbigg{ \sum_{i=1}^n\phi_{i,j}^\test } \geq \frac{\sqrt{n}\delta}{m} \Bigg]\I{ \widehat{\pi}_y^\train \in (\xi,1-\xi)\ \forall y\in\Y}\I{ N^\test \geq 1}   \Bigg\} 
\]
\[
	\leq  \frac{2m}{h  \delta }  +   2m\exp\bigg(-  \frac{h n\tau_n  \delta}{1+h n\tau_n \delta}\delta   \bigg)
\]
\normalsize
Thus, overall, we have that:
\begin{align*}
	\P_{Z_{1:n}}\Bigg[ \biggnorm{ \frac{1}{\sqrt{n}}\sum_{i=1}^n\phi_i^\test }{2} \geq \delta \Bigg]  &\leq  \frac{2m}{h  \delta }  +   2m\exp\bigg(-  \frac{h n\tau_n  \delta}{1+h n\tau_n \delta}\delta   \bigg) + 2m(m+1) e^{-n[1-\tau_n](1-e^{-c})},
\end{align*}
which can be made arbitrarily small by making $n$ and $\delta$ sufficiently large due to Assumption \eqref{tau n convergence not too fast}. This implies that $ \bignorm{ \frac{1}{\sqrt{n}}\sum_{i=1}^n\phi_i^\test }{2} = O_{\P}(1)$, as claimed.




\end{proof}
\vspace{0.4in}

\begin{proof}[\uline{Proof of Lemma \eqref{Lower Bound on min eval of Fisher Info with random gamma ast}}]

By inspection, both $\gamma^\ast$ and $\gamma^\ast(\tau_n)$ are members of the $m+1$ dimensional probability simplex. As such, by Lemma \eqref{Moment Identity}, we have that 
\[
	\FisherInfo{\gamma^\ast(\tau_n)} = \Cov_{\gamma^\ast(\tau_n)}[s_{\gamma^\ast(\tau_n)} , s_{\gamma^\ast(\tau_n)}] = A_{s_{\gamma^\ast(\tau_n)}}. 
\]   
However, Lemma \eqref{Moment Identity} also implies that:
\[
	A_{s_{\gamma^\ast(\tau_n)}} = \Cov_{\gamma^\ast}[s_{\gamma^\ast(\tau_n)}, s_{\gamma^\ast} ],	
\]
meaning that $\FisherInfo{\gamma^\ast(\tau_n)} = \Cov_{\gamma^\ast}[s_{\gamma^\ast(\tau_n)}, s_{\gamma^\ast}]$. Now, by Lemma \eqref{Bounded Gamma Star}, we know that $\gamma^\ast(\tau_n) \in \Gamma$, so by Assumption \eqref{Fixed Tau Regime: Mixture FIM not too small}, we have that $\sqrt{\Lambda} < \minSing{\FisherInfo{\gamma^\ast(\tau_n)}} = \minEval{\FisherInfo{\gamma^\ast(\tau_n)}}$. Bringing these observations together, we have that:
\begin{align*}
	\sqrt{\Lambda} &< \minEval{\FisherInfo{\gamma^\ast(\tau_n)}} \\
	&= \minEval{\Cov_{\gamma^\ast}[s_{\gamma^\ast(\tau_n)}, s_{\gamma^\ast}]} \\
	&= \min_{\norm{c}{2}=1} \Cov_{\gamma^\ast}[c's_{\gamma^\ast(\tau_n)}, c's_{\gamma^\ast}] \\ 
	&\leq \sqrt{\min_{\norm{c}{2}=1} \Var_{\gamma^\ast}[c's_{\gamma^\ast(\tau_n)}] \Var_{\gamma^\ast}[ c's_{\gamma^\ast}] } \\
	&\leq \sqrt{\max_{\norm{c}{2}=1} \Var_{\gamma^\ast}[c's_{\gamma^\ast(\tau_n)}]}   \sqrt{\min_{\norm{c}{2}=1}\Var_{\gamma^\ast}[ c's_{\gamma^\ast}] } \\
	&= \sqrt{\maxEval{\Var_{\gamma^\ast}[s_{\gamma^\ast(\tau_n)}]}} \sqrt{\minEval{ \FisherInfo{\gamma^\ast}  }} \\ 
	&\leq \sqrt{\sum_{i,j=1}^m \absBig{ \Cov_{\gamma^\ast}[s_{\gamma^\ast(\tau_n),i}, s_{\gamma^\ast(\tau_n),j}] }  }   \times \sqrt{\minEval{ \FisherInfo{\gamma^\ast}  }} \\ 
	&\leq \sqrt{\sum_{i,j=1}^m  \sqrt{\Var_{\gamma^\ast}[s_{\gamma^\ast(\tau_n),i}] \Var_{\gamma^\ast}[s_{\gamma^\ast(\tau_n),j}]}  }   \times \sqrt{\minEval{ \FisherInfo{\gamma^\ast}  }} \\ 
	&\leq \sqrt{  \frac{m^2}{L^2}    }   \times \sqrt{\minEval{ \FisherInfo{\gamma^\ast}  }} \\ 
	&=  \frac{m}{L}  \sqrt{\minEval{ \FisherInfo{\gamma^\ast}  }},
\end{align*}

where the second to last line is by Popoviciu's variance inequality and the fact that  $\gamma^\ast(\tau_n) \in \Gamma$ so $-\frac{1}{L} \leq s_{\gamma^\ast(\tau_n),j}  \leq \frac{1}{L}$ for each $j\in\Y$. Thus, we have that:
\[
	\Lambda \frac{L^2}{m^2} <  \minEval{ \FisherInfo{\gamma^\ast}  },
\]
as claimed.
 

\end{proof}
\vspace{0.4in}

\begin{proof}[\uline{Proof of Lemma \eqref{Behavior of First and Second Order Error Terms in tau n IID regime}}]



Under Assumption \eqref{Fixed Tau Regime: Mixture FIM not too small}, we have by Lemma \eqref{Lower Bound on min eval of Fisher Info with random gamma ast} that the smallest eigenvalue of $\FisherInfo{\gamma^\ast}$ is always greater than $0$, and so $\FisherInfo{\gamma^\ast}$ is invertible with probability $1$. Thus, $\psi^\eff$ is well-defined in Lemma \eqref{Alternative Expression for First Order Error}, and we can write:
\begin{align*}
	\sqrt{\tau_n(1-\tau_n)n}(\widehat{\pi}_n - \pi^\ast) &= \sqrt{\frac{\tau_n(1-\tau_n)}{n}} \sum_{i=1}^n \psi^\eff(Z_i) + \delta_{2,\tau_n}(Z_{1:n})
\end{align*}
where  
\[
	\delta_{2,\tau_n}(Z_{1:n}) := \sqrt{\tau_n(1-\tau_n)n}(\widehat{\pi}_n - \pi^\ast) - \sqrt{\frac{\tau_n(1-\tau_n)}{n}} \sum_{i=1}^n \psi^\eff(Z_i).
\]
Rewrite $ \sqrt{\frac{\tau_n(1-\tau_n)}{n}} \sum_{i=1}^n \psi^\eff(Z_i) $ in the following manner:

\smaller
\begin{align*}
	 &\sqrt{\frac{\tau_n(1-\tau_n)}{n}} \FisherInfo{\gamma^\ast}  \sum_{i=1}^n \psi^\eff(Z_i) \\
	 &=  \frac{\sqrt{\tau_n(1-\tau_n)}}{\widehat{\tau}_n}  \frac{1}{\sqrt{n}} \sum_{i=1}^{n} D_i\big( s_{\gamma^\ast}(X_i) - \E_{\pi^\ast}[s_{\gamma^\ast}] \big)  - \sum_{j=0}^m   \frac{\sqrt{\tau_n(1-\tau_n)}}{1-\widehat{\tau}_n}\frac{\pi_j^\ast}{ \widehat{\pi}_j^\train } \frac{1}{\sqrt{n}} \sum_{i=1}^{n} (1-D_i)\I{Y_i=j}\big( s_{\gamma^\ast}(X_i) - \E_{j}[s_{\gamma^\ast}] \big) \\ \\ 
	 &=  \frac{\sqrt{\tau_n(1-\tau_n)}}{\tau_n}  \frac{1}{\sqrt{n}} \sum_{i=1}^{n} D_i\big( s_{\gamma^\ast}(X_i) - \E_{\pi^\ast}[s_{\gamma^\ast}] \big)  -  \sum_{j=0}^m  \frac{\sqrt{\tau_n(1-\tau_n)}}{1-\tau_n}\frac{\pi_j^\ast}{ \pi_j^\train } \frac{1}{\sqrt{n}} \sum_{i=1}^{n} (1-D_i)\I{Y_i=j}\big( s_{\gamma^\ast}(X_i) - \E_{j}[s_{\gamma^\ast}] \big) \\ 
	 &\qquad\qquad+ \bigg(\frac{1}{\widehat{\tau}_n} - \frac{1}{\tau_n}\bigg)   \frac{\sqrt{\tau_n(1-\tau_n)}}{\sqrt{n}} \sum_{i=1}^{n} D_i\big( s_{\gamma^\ast}(X_i) - \E_{\pi^\ast}[s_{\gamma^\ast}] \big)  \\ 
	 &\qquad\qquad+ \sum_{j=0}^m  \bigg(\frac{1}{1-\tau_n}\frac{\pi_j^\ast}{ \pi_j^\train} - \frac{1}{1-\widehat{\tau}_n}\frac{\pi_j^\ast}{ \widehat{\pi}_j^\train}  \bigg) \frac{\sqrt{\tau_n(1-\tau_n)}}{\sqrt{n}} \sum_{i=1}^{n} (1-D_i)\I{Y_i=j}\big( s_{\gamma^\ast}(X_i) - \E_{j}[s_{\gamma^\ast}] \big) \\ \\ 
	 &=  \frac{\sqrt{\tau_n(1-\tau_n)}}{\tau_n}  \frac{1}{\sqrt{n}} \sum_{i=1}^{n} D_i\big( s_{\gamma^\ast(\tau_n)}(X_i) - \E_{\pi^\ast}[s_{\gamma^\ast(\tau_n)}] \big) -  \sum_{j=0}^m  \frac{\sqrt{\tau_n(1-\tau_n)}}{1-\tau_n}\frac{\pi_j^\ast}{ \pi_j^\train } \frac{1}{\sqrt{n}} \sum_{i=1}^{n} (1-D_i)\I{Y_i=j}\big( s_{\gamma^\ast(\tau_n)}(X_i) - \E_{j}[s_{\gamma^\ast(\tau_n)}] \big) \\
	 &\qquad\qquad+ \frac{\sqrt{\tau_n(1-\tau_n)}}{\tau_n}  \frac{1}{\sqrt{n}} \sum_{i=1}^{n} D_i\Big[\big( s_{\gamma^\ast}(X_i) - \E_{\pi^\ast}[s_{\gamma^\ast}] \big)  - \big( s_{\gamma^\ast(\tau_n)}(X_i) - \E_{\pi^\ast}[s_{\gamma^\ast(\tau_n)}] \big) \Big] \\ 
	 &\qquad\qquad+  \sum_{j=0}^m  \frac{\sqrt{\tau_n(1-\tau_n)}}{1-\tau_n}\frac{\pi_j^\ast}{ \pi_j^\train } \frac{1}{\sqrt{n}} \sum_{i=1}^{n} (1-D_i)\I{Y_i=j}\Big[\big( s_{\gamma^\ast(\tau_n)}(X_i) - \E_{j}[s_{\gamma^\ast(\tau_n)}] \big) - \big( s_{\gamma^\ast}(X_i) - \E_{j}[s_{\gamma^\ast}] \big) \Big] \\ 
	 &\qquad\qquad+ \bigg(\frac{1}{\widehat{\tau}_n} - \frac{1}{\tau_n}\bigg)   \frac{\sqrt{\tau_n(1-\tau_n)}}{\sqrt{n}} \sum_{i=1}^{n} D_i\big( s_{\gamma^\ast}(X_i) - \E_{\pi^\ast}[s_{\gamma^\ast}] \big)  \\ 
	 &\qquad\qquad+ \sum_{j=0}^m  \bigg(\frac{1}{1-\tau_n}\frac{\pi_j^\ast}{ \pi_j^\train} - \frac{1}{1-\widehat{\tau}_n}\frac{\pi_j^\ast}{ \widehat{\pi}_j^\train}  \bigg) \frac{\sqrt{\tau_n(1-\tau_n)}}{\sqrt{n}} \sum_{i=1}^{n} (1-D_i)\I{Y_i=j}\big( s_{\gamma^\ast}(X_i) - \E_{j}[s_{\gamma^\ast}] \big) \\ \\ 
	 &=  \frac{\sqrt{\tau_n(1-\tau_n)}}{\tau_n}  \frac{1}{\sqrt{n}} \sum_{i=1}^{n} D_i\big( s_{\gamma^\ast(\tau_n)}(X_i) - \E_{\pi^\ast}[s_{\gamma^\ast(\tau_n)}] \big) -  \sum_{j=0}^m  \frac{\sqrt{\tau_n(1-\tau_n)}}{1-\tau_n}\frac{\pi_j^\ast}{ \pi_j^\train } \frac{1}{\sqrt{n}} \sum_{i=1}^{n} (1-D_i)\I{Y_i=j}\big( s_{\gamma^\ast(\tau_n)}(X_i) - \E_{j}[s_{\gamma^\ast(\tau_n)}] \big) \\
	 &\qquad\qquad+ \underbrace{\frac{\sqrt{\tau_n(1-\tau_n)}}{\tau_n}  \frac{1}{\sqrt{n}} \sum_{i=1}^{n} D_i\Big[\big( s_{\gamma^\ast}(X_i) - \E_{\pi^\ast}[s_{\gamma^\ast}] \big)  - \big( s_{\gamma^\ast(\tau_n)}(X_i) - \E_{\pi^\ast}[s_{\gamma^\ast(\tau_n)}] \big) \Big]}_{=: V_{1,n}} \\ 
	 &\qquad\qquad+  \underbrace{\sum_{j=0}^m  \frac{\sqrt{\tau_n(1-\tau_n)}}{1-\tau_n}\frac{\pi_j^\ast}{ \pi_j^\train } \frac{1}{\sqrt{n}} \sum_{i=1}^{n} (1-D_i)\I{Y_i=j}\Big[\big( s_{\gamma^\ast(\tau_n)}(X_i) - \E_{j}[s_{\gamma^\ast(\tau_n)}] \big) - \big( s_{\gamma^\ast}(X_i) - \E_{j}[s_{\gamma^\ast}] \big) \Big]}_{=:V_{2,n}} \\ 
	 &\qquad\qquad+ \underbrace{\bigg(\frac{1}{\widehat{\tau}_n} - \frac{1}{\tau_n}\bigg)   \frac{\sqrt{\tau_n(1-\tau_n)}}{\sqrt{n}} \sum_{i=1}^{n} D_i\big( s_{\gamma^\ast(\tau_n)}(X_i) - \E_{\pi^\ast}[s_{\gamma^\ast(\tau_n)}] \big) }_{=:V_{3,n}} \\ 
	 &\qquad\qquad+ \underbrace{\bigg(\frac{1}{\widehat{\tau}_n} - \frac{1}{\tau_n}\bigg)   \frac{\sqrt{\tau_n(1-\tau_n)}}{\sqrt{n}} \sum_{i=1}^{n} D_i\Big[\big( s_{\gamma^\ast}(X_i) - \E_{\pi^\ast}[s_{\gamma^\ast}] \big) -\big( s_{\gamma^\ast(\tau_n)}(X_i) - \E_{\pi^\ast}[s_{\gamma^\ast(\tau_n)}] \big)\Big]}_{=:V_{4,n}}  \\ 
	 &\qquad\qquad+ \underbrace{\sum_{j=0}^m  \bigg(\frac{1}{1-\tau_n}\frac{\pi_j^\ast}{ \pi_j^\train} - \frac{1}{1-\widehat{\tau}_n}\frac{\pi_j^\ast}{ \widehat{\pi}_j^\train}  \bigg) \frac{\sqrt{\tau_n(1-\tau_n)}}{\sqrt{n}} \sum_{i=1}^{n} (1-D_i)\I{Y_i=j}\big( s_{\gamma^\ast(\tau_n)}(X_i) - \E_{j}[s_{\gamma^\ast(\tau_n)}] \big)}_{=:V_{5,n}} \\ 
	 &\qquad\qquad+ \underbrace{\sum_{j=0}^m  \bigg(\frac{1}{1-\tau_n}\frac{\pi_j^\ast}{ \pi_j^\train} - \frac{1}{1-\widehat{\tau}_n}\frac{\pi_j^\ast}{ \widehat{\pi}_j^\train}  \bigg) \frac{\sqrt{\tau_n(1-\tau_n)}}{\sqrt{n}} \sum_{i=1}^{n} (1-D_i)\I{Y_i=j}\Big[\big( s_{\gamma^\ast}(X_i) - \E_{j}[s_{\gamma^\ast}] \big) - \big( s_{\gamma^\ast(\tau_n)}(X_i) - \E_{j}[s_{\gamma^\ast(\tau_n)}] \big) \Big]}_{=:V_{6,n}}.
\end{align*}
\normalsize
Thus, making use of the definition of $\psi_{\tau_n}^\eff$ in Lemma \eqref{Projection of Ordinary Score onto G}, we have that:
\small
\[
	\implies  \sqrt{\frac{\tau_n(1-\tau_n)}{n}} \FisherInfo{\gamma^\ast} \sum_{i=1}^n \psi^\eff(Z_i) =  \sqrt{\frac{\tau_n(1-\tau_n)}{n}} \FisherInfo{\gamma^\ast(\tau_n)} \sum_{i=1}^n \psi_{\tau_n}^\eff(Z_i) + \sum_{l=1}^6 V_{l,n}
\]
\[ 
	\implies \sqrt{\frac{\tau_n(1-\tau_n)}{n}}  \sum_{i=1}^n \psi^\eff(Z_i) = \sqrt{\frac{\tau_n(1-\tau_n)}{n}}  \FisherInfo{\gamma^\ast}^{-1}\FisherInfo{\gamma^\ast(\tau_n)} \sum_{i=1}^n \psi_{\tau_n}^\eff(Z_i) + \FisherInfo{\gamma^\ast}^{-1}\sum_{l=1}^6 V_{l,n}
\]
\[
	\implies \sqrt{\frac{\tau_n(1-\tau_n)}{n}}   \sum_{i=1}^n \psi^\eff(Z_i) = \sqrt{\frac{\tau_n(1-\tau_n)}{n}}   \sum_{i=1}^n \psi_{\tau_n}^\eff(Z_i) + \delta_{1,\tau_n}(Z_{1:n}),
\]
\normalsize
where 
\[
	\delta_{1,\tau_n}(Z_{1:n}) := \big(\FisherInfo{\gamma^\ast}^{-1}\FisherInfo{\gamma^\ast(\tau_n)} - I_{m\times m}\big) \sqrt{\frac{\tau_n(1-\tau_n)}{n}}  \sum_{i=1}^n \psi_{\tau_n}^\eff(Z_i) + \FisherInfo{\gamma^\ast}^{-1}\sum_{l=1}^6 V_{l,n}.
\]
Then we have that:
\begin{align}
	\sqrt{\tau_n(1-\tau)n}(\widehat{\pi}_n - \pi^\ast) &=  \sqrt{\frac{\tau_n(1-\tau_n)}{n}}  \sum_{i=1}^n \psi^\eff(Z_i) + \delta_{2,\tau_n}(Z_{1:n})  \nonumber \\
	&= \sqrt{\frac{\tau_n(1-\tau_n)}{n}}   \sum_{i=1}^n \psi_{\tau_n}^\eff(Z_i) + \underbrace{\delta_{1,\tau_n}(Z_{1:n}) + \delta_{2,\tau_n}(Z_{1:n})}_{=: \delta_{\tau_n}(Z_{1:n}) } \nonumber \\
	&=  \sqrt{\frac{\tau_n(1-\tau_n)}{n}}   \sum_{i=1}^n \psi_{\tau_n}^\eff(Z_i) + \delta_{\tau_n}(Z_{1:n}). \label{this is the day that madoka was attacked by homura!!} \\ \nonumber
\end{align}

\uline{First, we analyze $V_{1,n}$}. Observe that:
\small
\begin{align*}
	&\norm{V_{1,n}}{2} \\
	&\leq  \biggnorm{\frac{1}{\sqrt{n}} \sum_{i=1}^{n} \frac{D_i}{\sqrt{\tau_n}}\Big[\big( s_{\gamma^\ast}(X_i) - \E_{\pi^\ast}[s_{\gamma^\ast}] \big)  - \big( s_{\gamma^\ast(\tau_n)}(X_i) - \E_{\pi^\ast}[s_{\gamma^\ast(\tau_n)}] \big) \Big]}{2} \\ 
	&= \sum_{k=0}^m \absbig{ \gamma_{k}^\ast - \gamma_{k}^\ast(\tau_n) } \times \biggnorm{ \frac{1}{\sqrt{n}} \sum_{i=1}^{n}  \frac{D_i}{\sqrt{\tau_n}}\Big[\big( s_{\gamma^\ast}(X_i) - \E_{\pi^\ast}[s_{\gamma^\ast}] \big)  - \big( s_{\gamma^\ast(\tau_n)}(X_i) - \E_{\pi^\ast}[s_{\gamma^\ast(\tau_n)}] \big) \Big] \bigg/ \sum_{k=0}^m \absbig{ \gamma_{k}^\ast - \gamma_{k}^\ast(\tau_n) }  }{2} \\ 
	&=  O_{\P}\Bigg( \frac{1}{\sqrt{n\tau_n(1-\tau_n)} }\Bigg) \times O_{\P}(1) \\ 
	&=  O_{\P}\Bigg( \frac{1}{\sqrt{n\tau_n(1-\tau_n)} }\Bigg),
\end{align*}
\normalsize
where the second to last line is due to Assumptions \eqref{tau n is inside open interval 0 and 1} and \eqref{tau n convergence not too fast}, and Lemmas  \eqref{Rate of gamma star sample size convergence} and \eqref{Error of gamma star difference in tau n IID Regime} . \\

\uline{Second, we analyze $V_{2,n}$}. Observe that:
\small
\begin{align*}
	&\norm{V_{2,n}}{2} \\
	&\leq \frac{1}{\sqrt{\xi}} \sum_{j=0}^m  \Biggnorm{ \frac{1}{\sqrt{n}} \sum_{i=1}^{n}  \frac{ (1-D_i)\I{Y_i=j} }{\sqrt{(1-\tau_n) \pi_j^\train}}   \Big[\big( s_{\gamma^\ast(\tau_n)}(X_i) - \E_{j}[s_{\gamma^\ast(\tau_n)}] \big) - \big( s_{\gamma^\ast}(X_i) - \E_{j}[s_{\gamma^\ast}] \big) \Big]}{2} \\ 
	&= \sum_{k=0}^m \absbig{ \gamma_{k}^\ast - \gamma_{k}^\ast(\tau_n) } \times \frac{1}{\sqrt{\xi}} \sum_{j=0}^m  \Biggnorm{ \frac{1}{\sqrt{n}} \sum_{i=1}^{n}  \frac{\frac{ (1-D_i)\I{Y_i=j} }{\sqrt{(1-\tau_n) \pi_j^\train}}   \Big[\big( s_{\gamma^\ast(\tau_n)}(X_i) - \E_{j}[s_{\gamma^\ast(\tau_n)}] \big) - \big( s_{\gamma^\ast}(X_i) - \E_{j}[s_{\gamma^\ast}] \big) \Big]}{ \sum_{k=0}^m \absbig{ \gamma_{k}^\ast - \gamma_{k}^\ast(\tau_n) } } }{2}  \\ 
	&=  O_{\P}\Bigg( \frac{1}{\sqrt{n\tau_n(1-\tau_n)} }\Bigg) \times O_{\P}(1) \\ 
	&=  O_{\P}\Bigg( \frac{1}{\sqrt{n\tau_n(1-\tau_n)} }\Bigg),
\end{align*}
\normalsize
where the second to last line is due to Assumptions \eqref{tau n is inside open interval 0 and 1} and \eqref{tau n convergence not too fast}, and Lemmas  \eqref{Rate of gamma star sample size convergence} and \eqref{Error of gamma star difference in tau n IID Regime} . \\ \\

\uline{Third, we analyze $V_{3,n}$}. Observe that
\small
\begin{align*}
	\norm{V_{3,n}}{2} &= \Biggnorm{\bigg(\frac{1}{\widehat{\tau}_n} - \frac{1}{\tau_n}\bigg)   \frac{\sqrt{\tau_n(1-\tau_n)}}{\sqrt{n}} \sum_{i=1}^{n} D_i\big( s_{\gamma^\ast(\tau_n)}(X_i) - \E_{\pi^\ast}[s_{\gamma^\ast(\tau_n)}] \big)}{2}  \\
	&= \Biggnorm{ \frac{\tau_n - \widehat{\tau}_n}{\widehat{\tau}_n\tau_n}  \frac{\sqrt{\tau_n(1-\tau_n)}}{\sqrt{n}} \sum_{i=1}^{n} D_i\big( s_{\gamma^\ast(\tau_n)}(X_i) - \E_{\pi^\ast}[s_{\gamma^\ast(\tau_n)}] \big) }{2} \\ 
	&= \Biggnorm{ \frac{\tau_n - \widehat{\tau}_n}{\widehat{\tau}_n\sqrt{\tau_n}}  \frac{\sqrt{1-\tau_n}}{\sqrt{n}} \sum_{i=1}^{n} D_i\big( s_{\gamma^\ast(\tau_n)}(X_i) - \E_{\pi^\ast}[s_{\gamma^\ast(\tau_n)}] \big) }{2} \\ 
	&\leq \Biggnorm{ \frac{\tau_n - \widehat{\tau}_n}{\widehat{\tau}_n}  \frac{1}{\sqrt{n}} \sum_{i=1}^{n} \frac{D_i}{\sqrt{\tau_n}} \big( s_{\gamma^\ast(\tau_n)}(X_i) - \E_{\pi^\ast}[s_{\gamma^\ast(\tau_n)}] \big) }{2} \\ 
	&=  \absbigg{\frac{\widehat{\tau}_n - \tau_n}{\widehat{\tau}_n}} \Biggnorm{  \frac{1}{\sqrt{n}} \sum_{i=1}^{n} \frac{D_i}{\sqrt{\tau_n}} \big( s_{\gamma^\ast(\tau_n)}(X_i) - \E_{\pi^\ast}[s_{\gamma^\ast(\tau_n)}] \big) }{2} \\ 
	&=  O_{\P}\bigg(\frac{1}{\sqrt{n\tau_n}}\bigg) \Biggnorm{  \frac{1}{\sqrt{n}} \sum_{i=1}^{n} \frac{D_i}{\sqrt{\tau_n}} \big( s_{\gamma^\ast(\tau_n)}(X_i) - \E_{\pi^\ast}[s_{\gamma^\ast(\tau_n)}] \big) }{2},
\end{align*}
\normalsize
where the last line is because, under Assumptions \eqref{tau n is inside open interval 0 and 1} and \eqref{tau n convergence not too fast}, we have by Lemma \eqref{Asymptotics of a Sample Proportion} that $\frac{\widehat{\tau}_n - \tau_n}{\widehat{\tau}_n} = O_{\P}\big(\frac{1}{\sqrt{n\tau_n}}\big)$. We now show that $\Bignorm{  \frac{1}{\sqrt{n}} \sum_{i=1}^{n} \frac{D_i}{\sqrt{\tau_n}} \big( s_{\gamma^\ast(\tau_n)}(X_i) - \E_{\pi^\ast}[s_{\gamma^\ast(\tau_n)}] \big) }{2} = O_{\P}(1)$. Towards that end, observe that for each $j\in [m]$:
\begin{align*}
	\Var_{\Jcal^{\pi^\ast,\mathbf{p},\tau_n}}\bigg[ \frac{D}{\sqrt{\tau_n}}\big( s_{\gamma^\ast(\tau_n),j} - \E_{\pi^\ast}[s_{\gamma^\ast(\tau_n),j}] \big) \bigg] &= \frac{1}{\tau_n}\E_{\Jcal^{\pi^\ast,\mathbf{p},\tau_n}}\Big[ D\big( s_{\gamma^\ast(\tau_n),j} - \E_{\pi^\ast}[s_{\gamma^\ast(\tau_n),j}] \big)^2 \Big] \\ 
	&=  \E_{\pi^\ast}\Big[\big( s_{\gamma^\ast(\tau_n),j} - \E_{\pi^\ast}[s_{\gamma^\ast(\tau_n),j}] \big)^2 \Big] \\ 
	&=  \Var_{\pi^\ast}\Big[s_{\gamma^\ast(\tau_n),j}\Big] \\ 
	&\leq  \frac{1}{L^2},
\end{align*}
where the last line is by Lemma \eqref{Bounded Gamma Star} and Popoviciu's variance inequality. So, since
\[
	\E_{\Jcal^{\pi^\ast,\mathbf{p},\tau_n}}\bigg[  \frac{D}{\sqrt{\tau_n}} \big( s_{\gamma^\ast(\tau_n)} - \E_{\pi^\ast}[s_{\gamma^\ast(\tau_n)}] \big)  \bigg] = 0,
\]
it follows that 
\begin{align*}
	&\Bignorm{  \frac{1}{\sqrt{n}} \sum_{i=1}^{n} \frac{D_i}{\sqrt{\tau_n}} \big( s_{\gamma^\ast(\tau_n)}(X_i) - \E_{\pi^\ast}[s_{\gamma^\ast(\tau_n)}] \big) }{2} \\
	 &= O_{\P}\Bigg(   \sqrt{\sum_{j=1}^m\Var_{\Jcal^{\pi^\ast,\mathbf{p},\tau_n}}\bigg[  \frac{D}{\sqrt{\tau_n}} \big( s_{\gamma^\ast(\tau_n),j} - \E_{\pi^\ast}[s_{\gamma^\ast(\tau_n),j}] \big)  \bigg]}  \Bigg) \\
	&= O_{\P}\Bigg(   \sqrt{\frac{m}{L^2}   }  \Bigg) \\ 
	&= O_{\P}(1).
\end{align*}
Thus, we have shown that:
\begin{align*}
	\norm{V_{3,n}}{2} &= O_{\P}\bigg(  \frac{1}{\sqrt{n\tau_n}}  \bigg). \\ \\ 
\end{align*}

\uline{Fourth, we analyze $V_{4,n}$}. Observe that:
\begin{align*}
	\norm{V_{4,n}}{2} &\leq \Biggnorm{\bigg(\frac{1}{\widehat{\tau}_n} - \frac{1}{\tau_n}\bigg)   \frac{\tau_n}{\sqrt{n}} \sum_{i=1}^{n} \frac{D_i}{\sqrt{\tau_n}} \Big[\big( s_{\gamma^\ast}(X_i) - \E_{\pi^\ast}[s_{\gamma^\ast}] \big) -\big( s_{\gamma^\ast(\tau_n)}(X_i) - \E_{\pi^\ast}[s_{\gamma^\ast(\tau_n)}] \big)\Big]}{2} \\
	&= \Biggnorm{ \frac{\tau_n - \widehat{\tau}_n}{\widehat{\tau}_n\tau_n}  \frac{\tau_n}{\sqrt{n}} \sum_{i=1}^{n} \frac{D_i}{\sqrt{\tau_n}} \Big[\big( s_{\gamma^\ast}(X_i) - \E_{\pi^\ast}[s_{\gamma^\ast}] \big) -\big( s_{\gamma^\ast(\tau_n)}(X_i) - \E_{\pi^\ast}[s_{\gamma^\ast(\tau_n)}] \big)\Big]}{2} \\ 
	&=  \frac{\abs{ \tau_n - \widehat{\tau}_n} }{\widehat{\tau}_n}   \times \Biggnorm{\frac{1}{\sqrt{n}} \sum_{i=1}^{n} \frac{D_i}{\sqrt{\tau_n}} \Big[\big( s_{\gamma^\ast}(X_i) - \E_{\pi^\ast}[s_{\gamma^\ast}] \big) -\big( s_{\gamma^\ast(\tau_n)}(X_i) - \E_{\pi^\ast}[s_{\gamma^\ast(\tau_n)}] \big)\Big]}{2} \\ 
	&=   O_{\P}\bigg( \frac{1}{\sqrt{n\tau_n}}\bigg)  \times O_{\P}\Bigg( \frac{1}{\sqrt{n\tau_n(1-\tau_n)} }\Bigg) \\
	&=   O_{\P}\Bigg( \frac{1}{\sqrt{n\tau_n} \sqrt{n\tau_n(1-\tau_n)} }\Bigg),
\end{align*}
where the second to last line follows from the work used to analyze $\norm{V_{1,n}}{2}$, as well as from  Assumptions \eqref{tau n is inside open interval 0 and 1} and \eqref{tau n convergence not too fast} and Lemma \eqref{Asymptotics of a Sample Proportion}.\\ \\

\uline{Fifth, we analyze $V_{5,n}$}. Observe that:
\small
\begin{align*}
	&V_{5,n}\\
	 &= \sum_{j=0}^m  \bigg(\frac{1}{1-\tau_n}\frac{\pi_j^\ast}{ \pi_j^\train} - \frac{1}{1-\widehat{\tau}_n}\frac{\pi_j^\ast}{ \widehat{\pi}_j^\train}  \bigg) \frac{\sqrt{\tau_n(1-\tau_n)}}{\sqrt{n}} \sum_{i=1}^{n} (1-D_i)\I{Y_i=j}\big( s_{\gamma^\ast(\tau_n)}(X_i) - \E_{j}[s_{\gamma^\ast(\tau_n)}] \big) \\ 
	&= \sum_{j=0}^m  \bigg(\frac{1}{1-\tau_n}\frac{\pi_j^\ast}{ \pi_j^\train} - \frac{1}{1-\widehat{\tau}_n}\frac{\pi_j^\ast}{ \widehat{\pi}_j^\train}  \bigg) \frac{\sqrt{\tau_n}(1-\tau_n)\sqrt{\pi_j^\train} }{\sqrt{n}} \sum_{i=1}^{n} \frac{(1-D_i)\I{Y_i=j}}{\sqrt{(1-\tau_n)\pi_j^\train} } \big( s_{\gamma^\ast(\tau_n)}(X_i) - \E_{j}[s_{\gamma^\ast(\tau_n)}] \big),
\end{align*}
\normalsize
and that, for each $j\in[m]$:
\begin{align*}
	&\absbigg{\frac{1}{1-\tau_n}\frac{\pi_j^\ast}{ \pi_j^\train} - \frac{1}{1-\widehat{\tau}_n}\frac{\pi_j^\ast}{ \widehat{\pi}_j^\train}  } \sqrt{\tau_n}(1-\tau_n)\sqrt{\pi_j^\train} \\
	&= \absbigg{\frac{1}{1-\tau_n}\frac{1}{ \pi_j^\train} - \frac{1}{1-\widehat{\tau}_n}\frac{1}{ \widehat{\pi}_j^\train}  } \pi_j^\ast \sqrt{\pi_j^\train \tau_n}(1-\tau_n) \\
	&= \absbigg{\frac{1}{ \pi_j^\train} - \frac{1-\tau_n}{1-\widehat{\tau}_n}\frac{1}{ \widehat{\pi}_j^\train}  } \pi_j^\ast \sqrt{\pi_j^\train \tau_n} \\
	&= \absbigg{\frac{1}{ \pi_j^\train} -\frac{1}{ \widehat{\pi}_j^\train}  + \frac{1}{ \widehat{\pi}_j^\train}   - \frac{1-\tau_n}{1-\widehat{\tau}_n}\frac{1}{ \widehat{\pi}_j^\train}  } \pi_j^\ast \sqrt{\pi_j^\train \tau_n} \\
	&\leq \Bigg(\absbigg{\frac{1}{ \pi_j^\train} - \frac{1}{ \widehat{\pi}_j^\train}}  + \absbigg{1   - \frac{1-\tau_n}{1-\widehat{\tau}_n}   } \frac{1}{ \widehat{\pi}_j^\train}\Bigg) \pi_j^\ast \sqrt{\pi_j^\train \tau_n} \\
	&\leq \Bigg(\absbigg{\frac{1}{ \pi_j^\train} - \frac{1}{ \widehat{\pi}_j^\train}}  + \absbigg{1   - \frac{1-\tau_n}{1-\widehat{\tau}_n}   } \bigg(\frac{1}{\pi_j^\train}  + \absbigg{ \frac{1}{ \widehat{\pi}_j^\train} - \frac{1}{\pi_j^\train}} \bigg)\Bigg) \pi_j^\ast \sqrt{\pi_j^\train \tau_n} \\
	&= \Bigg(\absbigg{\frac{\widehat{\pi}_j^\train - \pi_j^\train}{ \pi_j^\train \widehat{\pi}_j^\train }  }  + \absbigg{\frac{(1-\widehat{\tau}_n)-(1-\tau_n)}{1-\widehat{\tau}_n}   } \bigg(\frac{1}{\pi_j^\train}  + \absbigg{\frac{\widehat{\pi}_j^\train - \pi_j^\train}{ \pi_j^\train \widehat{\pi}_j^\train }  }    \bigg)\Bigg) \pi_j^\ast \sqrt{\pi_j^\train \tau_n} \\
	&= \Bigg(\absbigg{\frac{\widehat{\pi}_j^\train - \pi_j^\train}{  \widehat{\pi}_j^\train }  }  + \absbigg{\frac{(1-\widehat{\tau}_n)-(1-\tau_n)}{1-\widehat{\tau}_n}   } \bigg(1 + \absbigg{\frac{\widehat{\pi}_j^\train - \pi_j^\train}{  \widehat{\pi}_j^\train }  }    \bigg)\Bigg) \frac{\pi_j^\ast}{\pi_j^\train} \sqrt{\pi_j^\train \tau_n} \\ 
	&\leq \Bigg(\absbigg{\frac{\widehat{\pi}_j^\train - \pi_j^\train}{  \widehat{\pi}_j^\train }  }  + \absbigg{\frac{(1-\widehat{\tau}_n)-(1-\tau_n)}{1-\widehat{\tau}_n}   } \bigg(1 + \absbigg{\frac{\widehat{\pi}_j^\train - \pi_j^\train}{  \widehat{\pi}_j^\train }  }    \bigg)\Bigg) \frac{1}{\xi} \\ 
	&= \Bigg(\absbigg{\frac{\widehat{\pi}_j^\train - \pi_j^\train}{  \widehat{\pi}_j^\train }  }  + O_{\P}\bigg( \frac{1}{\sqrt{n(1-\tau_n)}} \bigg) \bigg(1 + \absbigg{\frac{\widehat{\pi}_j^\train - \pi_j^\train}{  \widehat{\pi}_j^\train }  }    \bigg)\Bigg) \frac{1}{\xi},
\end{align*}
where the last line is because,  under Assumptions \eqref{tau n is inside open interval 0 and 1} and \eqref{tau n convergence not too fast}, we have by Lemma \eqref{Asymptotics of a Sample Proportion} that $\frac{(1-\widehat{\tau}_n)-(1-\tau_n)}{1-\widehat{\tau}_n} = O_{\P}\big(\frac{1}{\sqrt{n(1-\tau_n)}}\big)$. Also note that:
\begin{align*}
	\absbigg{\frac{\widehat{\pi}_j^\train - \pi_j^\train}{  \widehat{\pi}_j^\train }  } &= \absbigg{\frac{ N_j^\train - \pi_j^\train N^\train}{  N_j^\train }  }  \\ 
	&\leq \frac{ \absbig{N_j^\train  - \pi_j^\train(1-\tau_n)n}}{N_j^\train} + \frac{\absbig{\pi_j^\train(1-\tau_n)n - \pi_j^\train N^\train}}{  N_j^\train }   \\ 
	&= \frac{ \absbig{N_j^\train/n  - \pi_j^\train(1-\tau_n)}}{N_j^\train/n} + \frac{\absbig{(1-\tau_n)n -  N^\train}}{  N_j^\train }\pi_j^\train   \\ 
	&= \frac{ \absbig{N_j^\train/n  - \pi_j^\train(1-\tau_n)}}{N_j^\train/n} + \frac{\absbig{(1-\tau_n) -  N^\train/n}}{  N^\train/n } \frac{\pi_j^\train N^\train }{N_j^\train}.
\end{align*}
Observe that:
\begin{align*}
	\frac{\pi_j^\train N^\train }{N_j^\train} &= \frac{\pi_j^\train(1-\tau_n)n  }{N_j^\train} \frac{N^\train}{(1-\tau_n)n} \\ 
	&= \absbigg{ \frac{\pi_j^\train(1-\tau_n)  }{N_j^\train / n} -1 + 1} \absbigg{\frac{N^\train/n}{1-\tau_n} -1 + 1} \\
	&= \absbigg{ \frac{\pi_j^\train(1-\tau_n) - N_j^\train / n  }{N_j^\train / n} + 1} \absbigg{\frac{N^\train/n - (1-\tau_n)}{1-\tau_n}  + 1}.
\end{align*}
Hence:
\small
\begin{align*}
	& \absbigg{\frac{\widehat{\pi}_j^\train - \pi_j^\train}{  \widehat{\pi}_j^\train }  } \\
	&\leq \frac{ \absbig{N_j^\train/n  - \pi_j^\train(1-\tau_n)}}{N_j^\train/n} + \frac{\absbig{(1-\tau_n) - (1-\widehat{\tau}_n)}}{  1-\widehat{\tau}_n} \Bigg( \absbigg{ \frac{\pi_j^\train(1-\tau_n) - N_j^\train / n  }{N_j^\train / n} + 1} \absbigg{ \frac{(1-\widehat{\tau}_n) - (1-\tau_n)}{1-\tau_n}  + 1} \Bigg).
\end{align*}
\normalsize
Under  Assumptions \eqref{tau n is inside open interval 0 and 1} and \eqref{tau n convergence not too fast}, we have by Lemma \eqref{Asymptotics of a Sample Proportion} that $\frac{ \abs{N_j^\train/n  - \pi_j^\train(1-\tau_n)}}{N_j^\train/n}$, $\frac{\abs{(1-\tau_n) - (1-\widehat{\tau}_n)}}{  1-\widehat{\tau}_n}$  and $\frac{(1-\widehat{\tau}_n) - (1-\tau_n)}{1-\tau_n}$ are all $O_{\P}\Big(  \frac{1}{\sqrt{n(1-\tau_n)}}\Big)$. Thus, it follows that $\absBig{\frac{\widehat{\pi}_j^\train - \pi_j^\train}{  \widehat{\pi}_j^\train }  }  = O_{\P}\Big(  \frac{1}{\sqrt{n(1-\tau_n)}}\Big)$, and so we may conclude that:
\begin{align}
	&\absbigg{\frac{1}{1-\tau_n}\frac{\pi_j^\ast}{ \pi_j^\train} - \frac{1}{1-\widehat{\tau}_n}\frac{\pi_j^\ast}{ \widehat{\pi}_j^\train}  } \sqrt{\tau_n}(1-\tau_n)\sqrt{\pi_j^\train} \\
	 &\leq  \Bigg(\absbigg{\frac{\widehat{\pi}_j^\train - \pi_j^\train}{  \widehat{\pi}_j^\train }  }  + O_{\P}\bigg( \frac{1}{\sqrt{n(1-\tau_n)}} \bigg) \bigg(1 + \absbigg{\frac{\widehat{\pi}_j^\train - \pi_j^\train}{  \widehat{\pi}_j^\train }  }    \bigg)\Bigg) \frac{1}{\xi} \nonumber \\
	&=    O_{\P}\bigg(  \frac{1}{\sqrt{n(1-\tau_n)}}\bigg)  + O_{\P}\bigg( \frac{1}{\sqrt{n(1-\tau_n)}} \bigg) \Bigg(1 +  O_{\P}\bigg(  \frac{1}{\sqrt{n(1-\tau_n)}}\bigg)    \Bigg)   \nonumber \\
	&=    O_{\P}\bigg(  \frac{1}{\sqrt{n(1-\tau_n)}}\bigg). \label{my foot kinda hurts}
\end{align}
This implies that $\norm{V_{5,n}}{2}$ is upper bounded by the summation of $m+1$ terms, where the $\numth{j}$ term has the form:
\[
	O_{\P}\bigg(  \frac{1}{\sqrt{n(1-\tau_n)}}\bigg) \times \biggnorm{\frac{1}{\sqrt{n}}\sum_{i=1}^{n} \frac{(1-D_i)\I{Y_i=j}}{\sqrt{(1-\tau_n)\pi_j^\train} } \big( s_{\gamma^\ast(\tau_n)}(X_i) - \E_{j}[s_{\gamma^\ast(\tau_n)}] \big)}{2}.
\]
Consequentially, to show that $\norm{V_{5,n}}{2} = O_{\P}\Big(  \frac{1}{\sqrt{n(1-\tau_n)}}\Big)$, it suffices to show that
\[
	\Biggnorm{\frac{1}{\sqrt{n}}\sum_{i=1}^{n} \frac{(1-D_i)\I{Y_i=j}}{\sqrt{(1-\tau_n)\pi_j^\train} } \big( s_{\gamma^\ast(\tau_n)}(X_i) - \E_{j}[s_{\gamma^\ast(\tau_n)}] \big) }{2} = O_{\P}(1)
\]
for each $j\in\Y$. So, towards verifying the display above, observe that, for each $k\in[m]$:
\begin{align*}
	&\Var_{\Jcal^{\pi^\ast,\mathbf{p},\tau_n}}\Bigg[ \frac{(1-D)\I{Y=j}}{\sqrt{(1-\tau_n)\pi_j^\train} } \big( s_{\gamma^\ast(\tau_n),k} - \E_{j}[s_{\gamma^\ast(\tau_n),k}] \big) \Bigg] \\
	&=  \frac{   \E_{\Jcal^{\pi^\ast,\mathbf{p},\tau_n}}\Big[ (1-D)\I{Y=j} \big( s_{\gamma^\ast(\tau_n),k} - \E_{j}[s_{\gamma^\ast(\tau_n),k}] \big)^2 \Big]   }{ (1-\tau_n)\pi_j^\train } \\
	&=  \frac{   (1-\tau_n) \pi_j^\train \E_{j}\Big[  \big( s_{\gamma^\ast(\tau_n),k} - \E_{j}[s_{\gamma^\ast(\tau_n),k}] \big)^2 \Big]   }{ (1-\tau_n)\pi_j^\train } \\
	&=  \Var_{j}\big[ s_{\gamma^\ast(\tau_n),k} \big]   \\ 
	&\leq  \frac{1}{L^2},
\end{align*}
where the last line is by Lemma \eqref{Bounded Gamma Star} and Popoviciu's variance inequality. So, since 
\[
	\E_{\Jcal^{\pi^\ast,\mathbf{p},\tau_n}}\Bigg[ \frac{(1-D)\I{Y=j}}{\sqrt{(1-\tau_n)\pi_j^\train} } \big( s_{\gamma^\ast(\tau_n),k} - \E_{j}[s_{\gamma^\ast(\tau_n),k}] \big) \Bigg] = 0,
\]
it follows that 
\small
\begin{align*}
	&\Biggnorm{\frac{1}{\sqrt{n}}\sum_{i=1}^{n} \frac{(1-D_i)\I{Y_i=j}}{\sqrt{(1-\tau_n)\pi_j^\train} } \big( s_{\gamma^\ast(\tau_n)}(X_i) - \E_{j}[s_{\gamma^\ast(\tau_n)}] \big) }{2} \\
	 &= O_{\P}\Bigg(\sqrt{ \sum_{k=1}^m \Var_{\Jcal^{\pi^\ast,\mathbf{p},\tau_n}}\Bigg[ \frac{(1-D)\I{Y=j}}{\sqrt{(1-\tau_n)\pi_j^\train} } \big( s_{\gamma^\ast(\tau_n),k} - \E_{j}[s_{\gamma^\ast(\tau_n),k}] \big) \Bigg]     }  \Bigg)\\
	&= O_{\P}\Bigg(   \sqrt{\frac{m}{L^2}   }  \Bigg) \\ 
	&= O_{\P}(1).
\end{align*}
\normalsize
Thus, as argued earlier, we may conclude that $\norm{V_{5,n}}{2} = O_{\P}\Big(  \frac{1}{\sqrt{n(1-\tau_n)}}\Big)$. \\


\uline{Sixth, we analyze $V_{6,n}$}. Observe that:
\smaller
\begin{align*}
	&\norm{V_{6,n}}{2} \\
	&\leq \sum_{j=0}^m \Bigg\{ \absbigg{\frac{1}{1-\tau_n}\frac{\pi_j^\ast}{ \pi_j^\train} - \frac{1}{1-\widehat{\tau}_n}\frac{\pi_j^\ast}{ \widehat{\pi}_j^\train} } \sqrt{\tau_n}(1-\tau_n)\sqrt{\pi_j^\train}  \\
	&\qquad\qquad\qquad\qquad\times \Biggnorm{ \frac{1 }{\sqrt{n}} \sum_{i=1}^{n} \frac{(1-D_i)\I{Y_i=j}}{\sqrt{(1-\tau_n) \pi_j^\train }} \Big[\big( s_{\gamma^\ast}(X_i) - \E_{j}[s_{\gamma^\ast}] \big) - \big( s_{\gamma^\ast(\tau_n)}(X_i) - \E_{j}[s_{\gamma^\ast(\tau_n)}] \big) \Big]}{2} \Bigg\},
\end{align*}
\normalsize
so to show that $\norm{V_{6,n}}{2} = O_{\P}\Big( \frac{1}{\sqrt{n(1-\tau_n)} \sqrt{n\tau_n(1-\tau_n)} }\Big)$, it suffices to show that the $\numth{j}$ summand above is $O_{\P}\Big( \frac{1}{\sqrt{n(1-\tau_n)} \sqrt{n\tau_n(1-\tau_n)} }\Big)$, for each $j\in\Y$. To see this, recall that
\[
	\absbigg{\frac{1}{1-\tau_n}\frac{\pi_j^\ast}{ \pi_j^\train} - \frac{1}{1-\widehat{\tau}_n}\frac{\pi_j^\ast}{ \widehat{\pi}_j^\train} } \sqrt{\tau_n}(1-\tau_n)\sqrt{\pi_j^\train}  = O_{\P}\bigg(  \frac{1}{\sqrt{n(1-\tau_n)}}\bigg)
\]
by line \eqref{my foot kinda hurts} and 
\small
\[
	 \Biggnorm{\frac{1 }{\sqrt{n}} \sum_{i=1}^{n} \frac{(1-D_i)\I{Y_i=j}}{\sqrt{(1-\tau_n) \pi_j^\train }} \Big[\big( s_{\gamma^\ast}(X_i) - \E_{j}[s_{\gamma^\ast}] \big) - \big( s_{\gamma^\ast(\tau_n)}(X_i) - \E_{j}[s_{\gamma^\ast(\tau_n)}] \big) \Big]}{2} = O_{\P}\Bigg( \frac{1}{\sqrt{n\tau_n(1-\tau_n)} }\Bigg)
\]
\normalsize
by the same work used to analyze $\norm{V_{2,n}}{2}$. This shows that the aforementioned $\numth{j}$ summand goes down like $ O_{\P}\Big( \frac{1}{\sqrt{n(1-\tau_n)} \sqrt{n\tau_n(1-\tau_n)} }\Big)$. Thus, $\norm{V_{6,n}}{2} =O_{\P}\Big( \frac{1}{\sqrt{n(1-\tau_n)} \sqrt{n\tau_n(1-\tau_n)} }\Big)$. 

Bringing together our analysis of $V_{1,n},\dots,V_{6,n}$, it follows that:
\small
\begin{align*}
	&\Biggnorm{\sum_{l=1}^6 V_{l,n}}{2} \\
	&\leq \sum_{l=1}^6 \bignorm{V_{l,n}}{2} \\ 
	&= O_{\P}\Bigg( \frac{1}{\sqrt{n\tau_n(1-\tau_n)} }\Bigg) + O_{\P}\Bigg( \frac{1}{\sqrt{n\tau_n(1-\tau_n)} }\Bigg) + O_{\P}\bigg(  \frac{1}{\sqrt{n\tau_n}}  \bigg) \\
	&\qquad\qquad + O_{\P}\Bigg( \frac{1}{\sqrt{n\tau_n} \sqrt{n\tau_n(1-\tau_n)} }\Bigg) +  O_{\P}\Bigg(  \frac{1}{\sqrt{n(1-\tau_n)}}\Bigg) + O_{\P}\Bigg( \frac{1}{\sqrt{n(1-\tau_n)} \sqrt{n\tau_n(1-\tau_n)} }\Bigg) \\ \\
	&= O_{\P}\Bigg( \frac{1}{\sqrt{n\tau_n(1-\tau_n)} }\Bigg),
\end{align*}
\normalsize
where the last line follows from Assumptions  \eqref{tau n is inside open interval 0 and 1} and \eqref{tau n convergence not too fast}. Now, under Assumption \eqref{Fixed Tau Regime: Mixture FIM not too small}, we have by Lemma \eqref{Lower Bound on min eval of Fisher Info with random gamma ast} that all the eigenvalues of $\FisherInfo{\gamma^\ast}$ are positive, which since $\FisherInfo{\gamma^\ast}$ is symmetric, means that $\FisherInfo{\gamma^\ast}$ is positive definite. Thus, the eigenvalues and singular values are equal, and so again by Lemma \eqref{Lower Bound on min eval of Fisher Info with random gamma ast}, we have that $\minSing{\FisherInfo{\gamma^\ast}} > \Lambda \frac{L^2}{m^2}$. Hence:
\begin{align*}
	\biggnorm{\FisherInfo{\gamma^\ast}^{-1}\sum_{l=1}^6 V_{l,n}}{2} &\leq \maxSing{\FisherInfo{\gamma^\ast}^{-1}} \Biggnorm{\sum_{l=1}^6 V_{l,n}}{2} \\
	&= \frac{1}{ \minSing{\FisherInfo{\gamma^\ast}} }  \Biggnorm{\sum_{l=1}^6 V_{l,n}}{2} \\
	&< \frac{m^2}{ \Lambda L^2 }  \Biggnorm{\sum_{l=1}^6 V_{l,n}}{2} \\
	&= O_{\P}\Bigg( \frac{1}{\sqrt{n\tau_n(1-\tau_n)} }\Bigg).
\end{align*}

Next, we focus on bounding $\big(\FisherInfo{\gamma^\ast}^{-1}\FisherInfo{\gamma^\ast(\tau_n)} - I_{m\times m}\big) \sqrt{\frac{\tau_n(1-\tau_n)}{n}}  \sum_{i=1}^n \psi_{\tau_n}^\eff(Z_i)$. Towards that end, observe that:
\begin{align*}
	\maxSing{\FisherInfo{\gamma^\ast}^{-1}\FisherInfo{\gamma^\ast(\tau_n)} - I_{m\times m} } &= \maxSingBig{\FisherInfo{\gamma^\ast}^{-1}\big(\FisherInfo{\gamma^\ast(\tau_n)} - \FisherInfo{\gamma^\ast}\big) } \\ 
	&\leq \maxSing{\FisherInfo{\gamma^\ast}^{-1}} \maxSing{\FisherInfo{\gamma^\ast(\tau_n)} - \FisherInfo{\gamma^\ast}} \\ 
	&< \frac{m^2}{ \Lambda L^2 }  \maxSing{\FisherInfo{\gamma^\ast(\tau_n)} - \FisherInfo{\gamma^\ast}}. 
\end{align*}
Also observe that, by Lemma \eqref{Moment Identity}
\begin{align*}
	\FisherInfo{\gamma^\ast(\tau_n)} - \FisherInfo{\gamma^\ast} = A_{s_{\gamma^\ast(\tau_n)}} - A_{s_{\gamma^\ast}} =  A_{s_{\gamma^\ast(\tau_n)} - s_{\gamma^\ast}},
\end{align*}
and so:
\begin{align*}
	\maxSing{\FisherInfo{\gamma^\ast}^{-1}\FisherInfo{\gamma^\ast(\tau_n)} - I_{m\times m} } &< \frac{m^2}{ \Lambda L^2 } \maxSing{A_{s_{\gamma^\ast(\tau_n)} - s_{\gamma^\ast}}} \\
	&\leq \frac{m^2}{ \Lambda L^2 }  \sum_{i,j=1}^m \absBig{\big[A_{s_{\gamma^\ast(\tau_n)} - s_{\gamma^\ast}}\big]_{i,j}} \\
	&= \frac{m^2}{ \Lambda L^2 }  \sum_{i,j=1}^m \absBig{ \E_j[s_{\gamma^\ast(\tau_n),i} - s_{\gamma^\ast,i} ] -  \E_0[s_{\gamma^\ast(\tau_n),i} - s_{\gamma^\ast,i} ]  }.	
\end{align*}
Now, by an argument similar to the one employed in the proof of Lemma \eqref{Error of gamma star difference in tau n IID Regime}, it is possible to show, for each $j\in\Y$ and $i\in[m]$, that 
\[
	\E_j[s_{\gamma^\ast(\tau_n),i} - s_{\gamma^\ast,i}] \Big/ \sum_{k=0}^m \absbig{ \gamma_{k}^\ast - \gamma_{k}^\ast(\tau_n) } = O_{\P}(1).
\]
As $m$ is finite, this implies that:
\begin{align*}
	\maxSing{\FisherInfo{\gamma^\ast}^{-1}\FisherInfo{\gamma^\ast(\tau_n)} - I_{m\times m} }  &= O_{\P}(1) \times \sum_{k=0}^m \absbig{ \gamma_{k}^\ast - \gamma_{k}^\ast(\tau_n) } \\ 
	&=  O_{\P}\bigg( \frac{1}{\sqrt{n\tau_n(1-\tau_n)} }\bigg).
\end{align*}
where the last line is due to Lemma \eqref{Rate of gamma star sample size convergence} and Assumptions  \eqref{tau n is inside open interval 0 and 1} and \eqref{tau n convergence not too fast}. Next, observe that:
\begin{align*}
	&\Var_{Z_{1:n}}\Bigg[ \sqrt{\frac{\tau_n(1-\tau_n)}{n}}  \sum_{i=1}^n \psi_{\tau_n}^\eff(Z_i) \Bigg] \\
	&=\tau_n(1-\tau_n) \Var_{\Jcal^{\pi^\ast,\mathbf{p},\tau_n}}\Big[\psi_{\tau_n}^\eff \Big] \\
	&=\tau_n(1-\tau_n)  \mathbf{V}^\textup{eff}(\tau_n) \\
	&=  \Bigg[ (1-\tau_n) + \tau_n \sum_{k=0}^{m} \frac{(\pi_k^\ast)^2}{\pi_k^\train} \Bigg]\Big(\FisherInfo{ \gamma^\ast(\tau_n) }^{-1} - \FisherInfo{\gamma^\ast(\tau_n);\textup{Cat}}^{-1} \Big) +  (1-\tau_n)\FisherInfo{\pi^\ast;\textup{Cat}}^{-1}, 
\end{align*}

where the second equality is due to line \eqref{variance matrix of efficient influence function}. Now, we know by Lemma \eqref{Bounded Gamma Star} that $\gamma^\ast(\tau_n)$ is bounded, and so by Lemma \eqref{Harder Problem}, we have that $\Tr\big\{ \FisherInfo{ \gamma^\ast(\tau_n) }^{-1}\big\} - \Tr\big\{ \FisherInfo{\gamma^\ast(\tau_n);\textup{Cat}}^{-1} \big\} \geq 0$. We also know that $\FisherInfo{\pi^\ast;\textup{Cat}}$ is positive definite by Lemma \eqref{Bounds on Eigenvalues of Categorical Fisher Info}, meaning that $\FisherInfo{\pi^\ast;\textup{Cat}}^{-1}$ is positive definite and hence $\Tr\big\{ \FisherInfo{\pi^\ast;\textup{Cat}}^{-1} \big\} \geq 0$. One can also use the same logic to conclude that $\Tr\big\{ \FisherInfo{\gamma^\ast(\tau_n);\textup{Cat}}^{-1} \big\}  \geq 0$.  Thus:
\small
\begin{align*}
	&\Tr\Bigg\{ \Var_{Z_{1:n}}\Bigg[ \sqrt{\frac{\tau_n(1-\tau_n)}{n}}  \sum_{i=1}^n \psi_{\tau_n}^\eff(Z_i) \Bigg]  \Bigg\} \\
	&\leq \Bigg[ 1 +  \sum_{k=0}^{m} \frac{(\pi_k^\ast)^2}{\pi_k^\train} \Bigg]\Big( \Tr\big\{ \FisherInfo{ \gamma^\ast(\tau_n) }^{-1}\big\} - \Tr\big\{ \FisherInfo{\gamma^\ast(\tau_n);\textup{Cat}}^{-1} \big\} \Big) + \Tr\big\{ \FisherInfo{\pi^\ast;\textup{Cat}}^{-1} \big\} \\ 
	&\leq \Bigg[ 1 +  \sum_{k=0}^{m} \frac{(\pi_k^\ast)^2}{\pi_k^\train} \Bigg] \Tr\big\{ \FisherInfo{ \gamma^\ast(\tau_n) }^{-1}\big\}  + \Tr\big\{ \FisherInfo{\pi^\ast;\textup{Cat}}^{-1} \big\} \\ 
	&\leq \Bigg[ 1 +  \sum_{k=0}^{m} \frac{(\pi_k^\ast)^2}{\pi_k^\train} \Bigg]  m \maxEvalBig{\FisherInfo{ \gamma^\ast(\tau_n) }^{-1}}  + m \maxEvalBig{ \FisherInfo{\pi^\ast;\textup{Cat}}^{-1} }\\ 
	&= \Bigg[ 1 +  \sum_{k=0}^{m} \frac{(\pi_k^\ast)^2}{\pi_k^\train} \Bigg]  \frac{m}{ \minSingBig{\FisherInfo{ \gamma^\ast(\tau_n) }} }  + \frac{m}{  \minEvalBig{ \FisherInfo{\pi^\ast;\textup{Cat}} }  } \\ 
	&\leq \Bigg[ 1 +  \sum_{k=0}^{m} \frac{(\pi_k^\ast)^2}{\pi_k^\train} \Bigg]  \frac{m}{ \sqrt{\Lambda} }  + \frac{m}{ 2 } \\
	&= O(1),
\end{align*}
\normalsize
where the last inequality is by Assumption \eqref{Fixed Tau Regime: Mixture FIM not too small} and Lemma \eqref{Bounds on Eigenvalues of Categorical Fisher Info}. Since $\E_{\Jcal^{\pi^\ast,\mathbf{p},\tau_n}}\big[ \psi_{\tau_n}^\eff \big]= 0$, it follows that
\begin{align*}
	\Biggnorm{\sqrt{\frac{\tau_n(1-\tau_n)}{n}}  \sum_{i=1}^n \psi_{\tau_n}^\eff(Z_i)}{2} &= O_{\P}\Bigg(   \sqrt{\Tr\Bigg\{ \Var_{Z_{1:n}}\Bigg[ \sqrt{\frac{\tau_n(1-\tau_n)}{n}}  \sum_{i=1}^n \psi_{\tau_n}^\eff(Z_i) \Bigg]  \Bigg\}}   \Bigg) \\ 
	&= O_{\P}(1),
\end{align*}
meaning that:
\begin{align*}
	&\Biggnorm{\big(\FisherInfo{\gamma^\ast}^{-1}\FisherInfo{\gamma^\ast(\tau_n)} - I_{m\times m}\big) \sqrt{\frac{\tau_n(1-\tau_n)}{n}}  \sum_{i=1}^n \psi_{\tau_n}^\eff(Z_i) }{2}\\
	 &\leq \maxSing{\FisherInfo{\gamma^\ast}^{-1}\FisherInfo{\gamma^\ast(\tau_n)} - I_{m\times m} } \Biggnorm{\sqrt{\frac{\tau_n(1-\tau_n)}{n}}  \sum_{i=1}^n \psi_{\tau_n}^\eff(Z_i)}{2}  \\ 
	&= O_{\P}\bigg( \frac{1}{\sqrt{n\tau_n(1-\tau_n)} }\bigg).
\end{align*}

Thus, overall, we have that:
\begin{align*}
	&\bignorm{\delta_{1,\tau_n}(Z_{1:n})}{2}\\
	 &\leq \biggnorm{\big(\FisherInfo{\gamma^\ast}^{-1}\FisherInfo{\gamma^\ast(\tau_n)} - I_{m\times m}\big) \sqrt{\frac{\tau_n(1-\tau_n)}{n}}  \sum_{i=1}^n \psi_{\tau_n}^\eff(Z_i) }{2} + \biggnorm{\FisherInfo{\gamma^\ast}^{-1}\sum_{l=1}^6 V_{l,n}}{2} \\ 
	&= O_{\P}\Bigg( \frac{1}{\sqrt{n\tau_n(1-\tau_n)} } \Bigg).
\end{align*}

Next, we bound $\norm{\delta_{2,\tau_n}(Z_{1:n})}{2}$. Let $\epsilon$ be as defined in Corollary \eqref{Error Rate of Full Estimator}, i.e., $\epsilon = (\widehat{\pi}_n - \pi^\ast) - \frac{1}{n} \sum_{i= 1}^{n} \psi^\eff(Z_i)$. Then:
\begin{align*}
	\norm{\delta_{2,\tau_n}(Z_{1:n})}{2} &= \biggnorm{\sqrt{\tau_n(1-\tau_n)n}(\widehat{\pi}_n - \pi^\ast) - \sqrt{\frac{\tau_n(1-\tau_n)}{n}} \sum_{i=1}^n \psi^\eff(Z_i)}{2} \\
	&= \sqrt{\tau_n(1-\tau_n)n}\biggnorm{ (\widehat{\pi}_n - \pi^\ast) - \frac{1}{n} \sum_{i=1}^n \psi^\eff(Z_i)}{2} \\
	&= \sqrt{\tau_n(1-\tau_n)n}\norm{\epsilon}{2}.
\end{align*}

Thus, for any $b,b_n > 0$ and $c^\test,c_0^\train,\dots,c_m^\train > 0$ (where $b, c^\test,c_0^\train,\dots,c_m^\train $ do not depend on $n$):
\small
\begin{align*}
	&\P_{Z_{1:n} }\Big[ \norm{\delta_{2,\tau_n}(Z_{1:n})}{2} / b_n   \geq b \Big] \\
	 &= \P_{Z_{1:n}}\Big[ \sqrt{\tau_n(1-\tau_n)n}\norm{\epsilon}{2}  /  b_n\geq b   \Big] \\ \\
	&\leq \P_{Z_{1:n}}\Big[ \sqrt{\tau_n(1-\tau_n)n}\norm{\epsilon}{2} /  b_n \geq b  \ \Big\lvert \ \abs{N^\test - \tau_n n} \leq  c^\test \sqrt{\tau_n(1-\tau_n)n},\\
	&\qquad\qquad\qquad\qquad\qquad\qquad\qquad \ \abs{N_j^\train- (1-\tau_n)\pi_j^\train n}    \leq  c_j^\train\sqrt{(1-\tau_n)\pi_j^\train [1-(1-\tau_n)\pi_j^\train]n} \ \forall j\in\Y \Big] \\
	&\qquad+ \P_{Z_{1:n}}\Big[  \abs{N^\test - \tau_n n} >  c^\test \sqrt{\tau_n(1-\tau_n)n} \Big]\\
	&\qquad+ \sum_{j=0}^m \P_{Z_{1:n}}\Big[\abs{N_j^\train- (1-\tau_n)\pi_j^\train n}    >  c_j^\train\sqrt{(1-\tau_n)\pi_j^\train [1-(1-\tau_n)\pi_j^\train]n} \Big].
\end{align*}
\normalsize
For each $n$, let $\bar{n}^\test, \bar{n}_0^\train,\dots,\bar{n}_m^\train$ be the sample sizes for which 
\[
	\P_{Z_{1:n}}\Big[ \sqrt{\tau_n(1-\tau_n)n}\norm{\epsilon}{2} /  b_n \geq b \ \Big\lvert \ N^\test  = \bar{n}^\test, \ N_j^\train =  \bar{n}_j^\train  \ \forall j\in\Y \Big] 
\]
is maximized, subject to the following constraints:
\begin{itemize}
	\item $\bar{n}^\test + \bar{n}_0^\train + \dots +\bar{n}_m^\train = n$
	\item $ \abs{\bar{n}^\test - \tau_n n} \leq  c^\test \sqrt{\tau_n(1-\tau_n)n}$
	\item $\abs{\bar{n}_j^\train- (1-\tau_n)\pi_j^\train n}    \leq  c_j^\train\sqrt{(1-\tau_n)\pi_j^\train [1-(1-\tau_n)\pi_j^\train]n}   \ \forall j\in\Y$.
\end{itemize}

Such sample sizes must exist, because the feasible set is nonempty (e.g.,  $\bar{n}^\test= \tau_n n$ and $\bar{n}_j^\train = (1-\tau_n)\pi_j^\train n$ for each $j\in\Y$  satisfies the constraints) and the feasible set is countably finite. It follows that:
\begin{align}
	&\P_{Z_{1:n} }\Big[ \norm{\delta_{2,\tau_n}(Z_{1:n})}{2}  /  b_n \geq b \Big] \nonumber\\
	&\leq \P_{Z_{1:n}}\Big[ \sqrt{\tau_n(1-\tau_n)n}\norm{\epsilon}{2}  / b_n \geq b \ \Big\lvert \  N^\test  = \bar{n}^\test  , \ N_j^\train = \bar{n}_j^\train  \ \forall j\in\Y \Big] \nonumber \\
	&\qquad+ \P_{Z_{1:n}}\Big[  \abs{N^\test - \tau_n n} >  c^\test \sqrt{\tau_n(1-\tau_n)n} \Big] \nonumber \\
	&\qquad+ \sum_{j=0}^m \P_{Z_{1:n}}\Big[\abs{N_j^\train- (1-\tau_n)\pi_j^\train n}    >  c_j^\train\sqrt{(1-\tau_n)\pi_j^\train [1-(1-\tau_n)\pi_j^\train]n} \Big]. \label{yo yo, crappy homura out to get youz!}
\end{align}

Now, due to the aforementioned constraints on $\bar{n}^\test, \bar{n}_0^\train,\dots,\bar{n}_m^\train$, it is easy to show that $\bar{n}^\test,\bar{n}^\train \to \infty$, as a result of Assumption  \eqref{tau n convergence not too fast}, where $\bar{n}^\train = n - \bar{n}^\test$. In addition, since $c^\test,c_0^\train,\dots,c_m^\train$ and $\pi^\train$ do not depend on $n$, it is not hard to show that $\frac{\bar{n}_j^\train}{\bar{n}^\train } \to \pi_j^\train$ for each $j\in\Y$ as $n\to\infty$. This latter observation implies that, for all sufficiently large $n$, $\frac{\bar{n}_j^\train}{\bar{n}^\train }  \in (\xi,1-\xi)$, since $\pi_j^\train \in (\xi,1-\xi)$. Altogether, this means that for all sufficiently large $n$, when conditioning on the event that  $N^\test  = \bar{n}^\test $ and $\ N_j^\train = \bar{n}_j^\train  \ \forall j\in\Y$, the data generation mechanism in the $\tau_n$-IID Regime is \textit{identical} to the data generation mechanism in a Fixed Sequence Regime with the following sequence of sample sizes: 
\begin{itemize}
	\item at each $n$,  $\bar{n}^\test$ samples are drawn IID from $p_{\pi^\ast}$
	\item at each $n$ and for each $j\in \Y$, $\bar{n}_j^\train = \frac{\bar{n}_j^\train}{\bar{n}^\train } \bar{n}^\train$ samples are drawn IID from $p_{j}$.
\end{itemize}
The asymptotic behavior of the true parameters in the $\tau_n$-IID Regime is also a special case of the Fixed Sequence Regime: for each $n$, the parameters $\pi^\ast$ and $\mathbf{p}$ are unchanged. One can also show that  the assumptions of the  $\tau_n$-IID Regime (when we condition on the event  that  $N^\test  = \bar{n}^\test $ and $\ N_j^\train = \bar{n}_j^\train  \ \forall j\in\Y$) are a special case of the assumptions of the Fixed Sequence Regime, whenever $n$ is sufficiently large. In particular:
\begin{itemize}
	\item Assumption \eqref{members of function class} in the Fixed Sequence Regime is satisfied by Assumption \eqref{Fixed Tau Regime: H closed under scalar multiplication} and the fact that $s_{\gamma^\ast}\in\H$ in the $\tau_n$-IID Regime, whenever $n$ is sufficiently large and  we condition on the event  that  $N^\test  = \bar{n}^\test $ and $N_j^\train = \bar{n}_j^\train  \ \forall j\in\Y$. To see that the latter is true, note that $\mathbf{p}\in \Q$, so to show that $s_{\gamma^\ast} \in \H^\text{pre}$, we only need to show that $\gamma^\ast \in \Gamma$. This would prove that $s_{\gamma^\ast}\in\H$ because $\H^\text{pre} \subseteq \H$. Now, to see that $\gamma^\ast \in \Gamma$, note that because we are conditioning on the event that  $N^\test  = \bar{n}^\test $ and $N_j^\train = \bar{n}_j^\train  \ \forall j\in\Y$, it follows that $\gamma^\ast$ is non-random, with the $\numth{y}$ component of $\gamma^\ast$ equal to:
	\begin{equation} \label{temporary definition of gamma ast for this particular sequence in proof}
		\gamma_y^\ast = \frac{  \frac{\pi_y^\ast}{\bar{n}^\test} + \frac{1}{\bar{n}^\train } \frac{(\pi_y^\ast)^2}{\bar{n}_y^\train / \bar{n}^\train  }           }{    \frac{1}{\bar{n}^\test} + \frac{1}{\bar{n}^\train } \sum_{k=0}^m \frac{(\pi_k^\ast)^2}{\bar{n}_k^\train / \bar{n}^\train  }          }.
	\end{equation}
	Since  $\pi^\ast \in \Delta$, $\sum_{y=0}^m \frac{\bar{n}_y^\train }{ \bar{n}^\train } = 1$, and  $\frac{\bar{n}_y^\train}{\bar{n}^\train }  \in (\xi,1-\xi)$ for all sufficiently large $n$ and $y\in\Y$, it follows from Lemma \eqref{Bounded Gamma Star} that $L \leq \gamma_y^\ast \leq U$ for all sufficiently large $n$ and $y\in\Y$. That is, $\gamma^\ast \in \Gamma$ for all sufficiently large $n$. \\ 

	\item Assumption \eqref{Omega Properties}  in the Fixed Sequence Regime is satisfied by Assumption \eqref{Fixed Tau Regime: Omega Properties} in the $\tau_n$-IID Regime. \\
	
	\item Assumption \eqref{lambda and uniform convergence properties}  in the Fixed Sequence Regime is satisfied by Assumptions \eqref{Fixed Tau Regime: Omega Properties}  and \eqref{Tau n Regime: lambda and uniform convergence properties} in the $\tau_n$-IID Regime, when we condition on the event  that  $N^\test  = \bar{n}^\test $ and $N_j^\train = \bar{n}_j^\train  \ \forall j\in\Y$. To demonstrate this, we begin by verifying that each of the conditions needed to invoke Assumption  \eqref{Tau n Regime: lambda and uniform convergence properties} hold. Towards that end, note that $\pi^\ast \in \Delta$ and $\mathbf{p} \in \Q$, and that by virtue of how  $(\bar{n}^\test, \bar{n}_0^\train,\dots,\bar{n}_m^\train)$ was constructed, each of the foregoing sample sizes grow towards infinity. Also, the definition of $\gamma^\ast$ in \eqref{temporary definition of gamma ast for this particular sequence in proof} equals the definition of $\bar{\gamma}$ in Assumptions section for the $\tau_n$-IID Regime. Thus, all of the conditions have been met, so by Assumption  \eqref{Tau n Regime: lambda and uniform convergence properties}, we have for each $j\in[m]$ that  
	\begin{itemize}
		\item $\lambda_j\sup_{\gamma\in\Gamma} \Omega^2\Big( \frac{q_j - q_0}{q_{\gamma}} \Big)  \to 0$
		\item $\lambda_j^2\bar{n}^\train  \to \infty$
		\item \small $\E_{\DtrainII}\Bigg\{   \sup\limits_{ \Omega(h_j) \leq \frac{1}{L\sqrt{\lambda_j} } } \absBig{\widehat{\E}_{0}^\trainII[h_j] - \E_{0}[h_j] }  + \sup\limits_{ \Omega(h_j) \leq \frac{1}{L\sqrt{\lambda_j} } } \absBig{\widehat{\E}_{j}^\trainII[h_j] - \E_{j}[h_j] } + \sup\limits_{ \Omega(h_j) \leq \frac{1}{L\sqrt{\lambda_j} } } \absBig{\widehat{\Var}_{    \gamma^\ast  }^\trainII[h_j] - \Var_{  \gamma^\ast  } [h_j]  }   \Bigg\} \to 0$, \normalsize
	\end{itemize}
	where, for each $n$, $\DtrainII$ is the analogue of $\barDtrainII$ described in the Assumptions section for the $\tau_n$-IID Regime, i.e., $\DtrainII$ contains $\frac{1}{2}\bar{n}_y^\train$ IID samples from $p_y$ for each $y\in\Y$. Using the display above, we will now verify that all the conditions listed in Assumption \eqref{lambda and uniform convergence properties} are met. Since $\lambda_j\sup_{\gamma\in\Gamma} \Omega^2(s_{\gamma,j})  = o(1)$ for each $j \in [m]$, it follows that $\lambda_j \Omega^2( s_{\gamma^\ast,j} ) = o(1)$ for each $j\in[m]$, because as was argued earlier,  $\gamma^\ast \in \Gamma$ for all sufficiently large $n$. In addition, consider any fixed $\gamma \in \Gamma$. Clearly, $\lambda_j\Omega(s_{\gamma,j}) = o(1)$, so because $s_{\gamma,j}$ is independent of $n$ and $\Omega(s_{\gamma,j})$ is finite due to Assumption  \eqref{Fixed Tau Regime: Omega Properties}, we have that $\lambda_j = o(1)$. We also have that  $\lambda_j^2\bar{n}^\train  \to \infty$. Finally, the uniform convergence in the display above is equivalent to the requirement in Assumption \eqref{lambda and uniform convergence properties}  that $\E_{\DtrainII}[\U_j(\lambda_j)] \to 0$. Thus, Assumption  \eqref{lambda and uniform convergence properties} does indeed hold. \\ 
		
	\item Assumption \eqref{FIM Difference is Bounded}  in the Fixed Sequence Regime is satisfied by Assumption \eqref{Fixed Tau Regime: Cat minus Mixture FIM} in the $\tau_n$-IID Regime, whenever $n$ is sufficiently large and we condition on the event  that  $N^\test  = \bar{n}^\test $ and $N_j^\train = \bar{n}_j^\train  \ \forall j\in\Y$. To see why, note that as argued earlier, conditioning on the aforementioned event ensures that $\gamma^\ast \in \Gamma$ for all sufficiently large $n$. Since $\mathbf{p} \in\Q$, it therefore follows from Assumption \eqref{Fixed Tau Regime: Cat minus Mixture FIM} that $\minSing{\FisherInfo{\gamma^\ast; \text{Cat}} - \FisherInfo{\gamma^\ast}} > \nu$ for all sufficiently large $n$. \\

	\item Assumption \eqref{bounded pis}  in the Fixed Sequence Regime is always true in the $\tau_n$-IID Regime, because $\pi^\ast,\pi^\train \in \Delta$. \\ 
		
	\item Assumption \eqref{min Fisher Info}  in the Fixed Sequence Regime is satisfied by Assumption \eqref{Fixed Tau Regime: Mixture FIM not too small} in the $\tau_n$-IID Regime, whenever $n$ is sufficiently large and we condition on the event  that  $N^\test  = \bar{n}^\test $ and $N_j^\train = \bar{n}_j^\train  \ \forall j\in\Y$. To see why, note that as argued earlier, conditioning on the aforementioned event ensures that $\gamma^\ast \in \Gamma$ for all sufficiently large $n$. Since $\mathbf{p} \in\Q$, it therefore follows from Assumption \eqref{Fixed Tau Regime: Mixture FIM not too small} that $\minSing{\FisherInfo{\gamma^\ast}} > \sqrt{\Lambda}$ for all sufficiently large $n$. \\ 
	
	\item Assumption \eqref{gamma hat properties} in the Fixed Sequence Regime is satisfied by Assumption \eqref{tau n regime, estimator of gamma bar} in the $\tau_n$-IID Regime, whenever $n$ is sufficiently large and we condition on the event  that  $N^\test  = \bar{n}^\test $ and $N_j^\train = \bar{n}_j^\train  \ \forall j\in\Y$. To see why, note that under Assumption \eqref{tau n regime, estimator of gamma bar}, $\widehat{\gamma} \in \Gamma$ for each $n$. Furthermore, conditional on the aforementioned event,  $\gamma^\ast$ is equal to $\bar{\gamma}$ and $\allDII$ is equal to $\barallDII$, where $\bar{\gamma}$ and $\barallDII$ are defined in the Assumptions section of the $\tau_n$-IID Regime. Thus, indeed, it holds that $\E_{ \allDII }\Big[ \norm{\widehat{\gamma} - \gamma^\ast}{2}^2 \Big] \to 0$ and, for each $a > 0$, $\P_{ \allDII}\big[ \norm{\widehat{\gamma} - \gamma^\ast }{2}\geq a\big] = o\Big(\sqrt{\frac{1}{\bar{n}^\test} + \frac{1}{\bar{n}^\train  }}\Big)$. Thus, Assumption  \eqref{gamma hat properties} holds.\\
\end{itemize}

Thus, we have shown that the data generation mechanism / assumptions of the $\tau_n$-IID Regime are a special case of the data generation mechanism / assumptions of the Fixed Sequence Regime, whenever $n$ is sufficiently large and we condition on the event that $N^\test = \bar{n}^\test$ and $N_y^\train = \bar{n}_y^\train$ for each $y\in[m]$. Thus, we may invoke Corollary \eqref{Error Rate of Full Estimator} from the Fixed Sequence Regime to make the following asymptotic statement about the distribution of $\norm{\epsilon}{2}$ in the $\tau_n$-IID Regime, conditional on the event that $N^\test = \bar{n}^\test$ and $N_y^\train = \bar{n}_y^\train$ for each $y\in[m]$ and $n$:
\smaller
\begin{align*}
	&\norm{\epsilon}{2}\\
	 &=   O_{\P}\Bigg(\sqrt{\bigg( \frac{1}{\bar{n}^\test} + \frac{1}{\bar{n}^\train} \bigg) \bigg\{ \frac{1}{\sqrt{\bar{n}^\train}} + \max_y \lambda_y \Omega(s_{\bar{\gamma},y})^2 +  \frac{1}{\sqrt{\min_y\lambda_y}}e^{-C\bar{n}^\train} +  \max_y \E_{\barDtrainII}\Big[ \bar{\U}_y(\lambda_y)\Big] + \sqrt{\E_{\barallDII}\Big[ \norm{\widehat{\gamma} - \bar{\gamma} }{2}^2 \Big]}  \bigg\}   }\Bigg) \\ 
	&\qquad + O_{\P}\Bigg( e^{-C (\min_y\lambda_y)^2 \bar{n}^\train}  + \P_{\barallDII}\Big[\bignorm{ \widehat{\gamma} - \bar{\gamma} }{2} \geq C \Big]  \Bigg),
\end{align*}
\normalsize
where $C > 0$ is some fixed, global constant. Now, notice that, due to the constraints on $\bar{n}^\test, \bar{n}_0^\train, \dots, \bar{n}_m^\train$, it holds that:
\begin{align*}
	\tau_n(1-\tau_n)n \bigg( \frac{1}{\bar{n}^\test} + \frac{1}{\bar{n}^\train} \bigg) &= (1-\tau_n)\frac{\tau_n }{\bar{n}^\test / n} + \tau_n\frac{1-\tau_n}{\bar{n}^\train/n} \to 1,
\end{align*}
so:
\small
\begin{align*}
	&\sqrt{\tau_n(1-\tau_n)n}\norm{\epsilon}{2}\\
	 &=   O_{\P}\Bigg(\sqrt{  \frac{1}{\sqrt{\bar{n}^\train}} + \max_y \lambda_y \Omega(s_{\bar{\gamma},y})^2 +  \frac{1}{\sqrt{\min_y\lambda_y}}e^{-C\bar{n}^\train} +  \max_y \E_{\barDtrainII}\Big[ \bar{\U}_y(\lambda_y)\Big] + \sqrt{\E_{\barallDII}\Big[ \norm{\widehat{\gamma} - \bar{\gamma} }{2}^2 \Big]}    }\Bigg) \\ 
	&\qquad + O_{\P}\Bigg(\sqrt{\tau_n(1-\tau_n)n} e^{-C (\min_y\lambda_y)^2 \bar{n}^\train}  + \sqrt{\tau_n(1-\tau_n)n} \P_{\barallDII}\Big[\bignorm{ \widehat{\gamma} - \bar{\gamma} }{2} \geq C \Big]  \Bigg),
\end{align*}
\normalsize
which, by our arguments from earlier, is $O_{\P}(1)$. Now, set $b_n$ equal to: 
\small
\begin{align*}
	b_n &= \sqrt{  \frac{1}{\sqrt{\bar{n}^\train}} + \max_y \lambda_y \Omega(s_{\bar{\gamma},y})^2 +  \frac{1}{\sqrt{\min_y\lambda_y}}e^{-C\bar{n}^\train} +  \max_y \E_{\barDtrainII}\Big[ \bar{\U}_y(\lambda_y)\Big] + \sqrt{\E_{\barallDII}\Big[ \norm{\widehat{\gamma} - \bar{\gamma} }{2}^2 \Big]}    } \\
	&\qquad\qquad + \sqrt{\tau_n(1-\tau_n)n} e^{-C (\min_y\lambda_y)^2 \bar{n}^\train}  + \sqrt{\tau_n(1-\tau_n)n} \P_{\barallDII}\Big[\bignorm{ \widehat{\gamma} - \bar{\gamma} }{2} \geq C \Big].
\end{align*}
\normalsize
Then, by definition of $O_{\P}(\cdot)$, it follows that 
\[
	\P_{Z_{1:n}}\Big[ \sqrt{\tau_n(1-\tau_n)n} \norm{\epsilon}{2}   / b_n \geq b \ \Big\lvert \  N^\test  = \bar{n}^\test  , \ N_j^\train = \bar{n}_j^\train  \ \forall j\in\Y \Big] 
\]
can be made arbitrarily small by choosing  $b$ and $n$ to be sufficiently large. Next, we will argue that the probabilities 
\begin{itemize}
	\item $\P_{Z_{1:n}}\Big[  \abs{N^\test - \tau_n n} >  c^\test \sqrt{\tau_n(1-\tau_n)n} \Big]$ 
	\item $\sum_{j=0}^m \P_{Z_{1:n}}\Big[\abs{N_j^\train- (1-\tau_n)\pi_j^\train n}    >  c_j^\train\sqrt{(1-\tau_n)\pi_j^\train [1-(1-\tau_n)\pi_j^\train]n} \Big]$
\end{itemize}
can be made arbitrarily small, whenever $n$ and $c^\test,c_0^\train, \dots,c_m^\train$  are taken to be sufficiently large. This is simple to prove. For each $n$,  $N^\test \sim \text{Bin}(n,\ \tau_n)$ and $N_j^\train \sim \text{Bin}(n,\ (1-\tau_n)\pi_j^\train)$. Thus, it follows that $(N^\test - \tau_n n) / \sqrt{n\tau_n(1-\tau_n)} \goesto{d} \N(0,1)$ and $(N_j^\train - (1-\tau_n)\pi_j^\train n) / \sqrt{(1-\tau_n)\pi_j^\train [1-(1-\tau_n)\pi_j^\train]n} \goesto{d} \N(0,1)$. So, by definition of convergence in distribution, it follows that 
\begin{itemize}
	\item $\P_{Z_{1:n}}\Big[  \abs{N^\test - \tau_n n} >  c^\test \sqrt{\tau_n(1-\tau_n)n} \Big] \to 2 \P[\N(0,1) \geq c^\test ]$ 
	\item $\sum_{j=0}^m \P_{Z_{1:n}}\Big[\abs{N_j^\train- (1-\tau_n)\pi_j^\train n}    >  c_j^\train\sqrt{(1-\tau_n)\pi_j^\train [1-(1-\tau_n)\pi_j^\train]n} \Big] \to 2 \sum_{j=0}^m\P[ \N(0,1) \geq c_j^\train]$.
\end{itemize}
Clearly, the limits can be made arbitrarily small by choosing $c^\test,c_0^\train, \dots,c_m^\train$ to be sufficiently large. Since the display above implies that the probabilities on the LHS can be made arbitrarily close to their limits whenever $n$ is sufficiently large, it follows that those probabilities can be made arbitrarily small by choosing $n$ and $c^\test,c_0^\train, \dots,c_m^\train$ to be sufficiently large, as claimed. 

Now, set $b = c^\test = c_0^\train = \dots = c_m^\train$. Then, bringing all of our work together, it follows from line \eqref{yo yo, crappy homura out to get youz!} that $\P_{Z_{1:n} }\big[ \norm{\delta_{2,\tau_n}(Z_{1:n})}{2}  /  b_n \geq b \big] $ can be made arbitrarily small by choosing $b$ and $n$ to be sufficiently large. This implies that:
\begin{align*}
	 &\norm{\delta_{2,\tau_n}(Z_{1:n})}{2} \\
	 &= O_{\P}(b_n) \\ 
	 &= O_{\P}\Bigg(\sqrt{  \frac{1}{\sqrt{\bar{n}^\train}} + \max_y \lambda_y \Omega(s_{\bar{\gamma},y})^2 +  \frac{1}{\sqrt{\min_y\lambda_y}}e^{-C\bar{n}^\train} +  \max_y \E_{\barDtrainII}\Big[ \bar{\U}_y(\lambda_y)\Big] + \sqrt{\E_{\barallDII}\Big[ \norm{\widehat{\gamma} - \bar{\gamma} }{2}^2 \Big]}    }\Bigg) \\ 
	&\qquad + O_{\P}\Bigg(\sqrt{\tau_n(1-\tau_n)n} e^{-C (\min_y\lambda_y)^2 \bar{n}^\train}  + \sqrt{\tau_n(1-\tau_n)n} \P_{\barallDII}\Big[\bignorm{ \widehat{\gamma} - \bar{\gamma} }{2} \geq C \Big]  \Bigg).
\end{align*}

Since we already showed that $\bignorm{\delta_{1,\tau_n}(Z_{1:n})}{2}  =  O_{\P} \Big( \frac{1}{\sqrt{n\tau_n(1-\tau_n)} } \Big)$, we may conclude from line \eqref{this is the day that madoka was attacked by homura!!} that:
\begin{align*}
	\sqrt{\tau_n(1-\tau)n}(\widehat{\pi}_n - \pi^\ast) &=  \sqrt{\frac{\tau_n(1-\tau_n)}{n}}   \sum_{i=1}^n \psi_{\tau_n}^\eff(Z_i) + \delta_{\tau_n}(Z_{1:n}) 
\end{align*}
where
\smaller
\begin{align*}
	&\bignorm{\delta_{\tau_n}(Z_{1:n})}{2} \\
	 &\leq \bignorm{ \delta_{1,\tau_n}(Z_{1:n}) }{2} + \bignorm{ \delta_{2,\tau_n}(Z_{1:n}) }{2} \\
	&= O_{\P}\Bigg( \frac{1}{\sqrt{n\tau_n(1-\tau_n)} }   +   \sqrt{  \frac{1}{\sqrt{\bar{n}^\train}} + \max_y \lambda_y \Omega(s_{\bar{\gamma},y})^2 +  \frac{1}{\sqrt{\min_y\lambda_y}}e^{-C\bar{n}^\train} +  \max_y \E_{\barDtrainII}\Big[ \bar{\U}_y(\lambda_y)\Big] + \sqrt{\E_{\barallDII}\Big[ \norm{\widehat{\gamma} - \bar{\gamma} }{2}^2 \Big]}    } \Bigg) \\
	&\qquad\qquad + O_{\P}\Bigg(\sqrt{\tau_n(1-\tau_n)n} e^{-C (\min_y\lambda_y)^2 \bar{n}^\train}  + \sqrt{\tau_n(1-\tau_n)n} \P_{\barallDII}\Big[\bignorm{ \widehat{\gamma} - \bar{\gamma} }{2} \geq C \Big]  \Bigg),
\end{align*}
\normalsize
as desired! Finally, we turn to verifying the  asymptotic properties of $\psi_{\tau_n}^\eff$ that were claimed by the present lemma. Observe that, for each $x\in\X$:
\begin{align*}
	 \bignorm{ \tau_n \psi_{\tau_n}^\eff(x,-1,1) }{2} &=  \biggnorm{ \tau_n \frac{1}{\tau_n} \FisherInfo{\gamma^\ast(\tau_n)}^{-1}\big(s_{\gamma^\ast(\tau_n)}(x)  -  \E_{\pi^\ast}[s_{\gamma^\ast(\tau_n)}]\big) }{2} \\
	&=  \bignorm{ \FisherInfo{\gamma^\ast(\tau_n)}^{-1}\big(s_{\gamma^\ast(\tau_n)}(x)  -  \E_{\pi^\ast}[s_{\gamma^\ast(\tau_n)}]\big) }{2} \\ 
	&\leq \maxSing{ \FisherInfo{\gamma^\ast(\tau_n)}^{-1} } \bignorm{s_{\gamma^\ast(\tau_n)}(x)  -  \E_{\pi^\ast}[s_{\gamma^\ast(\tau_n)}]}{2} \\
	&= \frac{1}{ \minSing{ \FisherInfo{\gamma^\ast(\tau_n)} } } \bignorm{s_{\gamma^\ast(\tau_n)}(x)  -  \E_{\pi^\ast}[s_{\gamma^\ast(\tau_n)}]}{2} \\
	&\leq \frac{1}{ \sqrt{\Lambda} } \bignorm{s_{\gamma^\ast(\tau_n)}(x)  -  \E_{\pi^\ast}[s_{\gamma^\ast(\tau_n)}]}{2} \\ 
	&= \frac{1}{ \sqrt{\Lambda} } \sqrt{   \sum_{j=1}^m  \Big(s_{\gamma^\ast(\tau_n),j}(x)  -  \E_{\pi^\ast}[s_{\gamma^\ast(\tau_n),j}] \Big)^2   } \\ 
	&\leq \frac{1}{ \sqrt{\Lambda} } \frac{2\sqrt{m}}{L}, 
\end{align*}
where the third to last line uses Assumption \eqref{Fixed Tau Regime: Omega Properties} and the fact that $\gamma^\ast(\tau_n) \in \Gamma$ by virtue of Lemma \eqref{Bounded Gamma Star}, and the last line uses Assumption \eqref{Fixed Tau Regime: Omega Properties}. This implies that 
\begin{align*}
	\frac{(1-{\tau}_n)^2}{{\tau}_n n} \E_{\pi^\ast}\Big[ \bignorm{ \tau_n \psi_{\tau_n}^\eff(X,-1,1) }{2}^4 \Big] &\leq  \frac{(1-{\tau}_n)^2}{{\tau}_n n}   \frac{16m^2}{\Lambda^2 L^4} \\
	&\leq  \frac{1}{{\tau}_n n}   \frac{16m^2}{\Lambda^2 L^4} \to 0,
\end{align*}
where the limit is because of Assumption \eqref{tau n convergence not too fast}. Similarly, observe that, for each $x\in\X$ and $y\in\Y$:
\begin{align*}
	&\bignorm{(1-{\tau}_n) \psi^\eff_{{\tau}_n}(x,y,0)}{2}\\
	 &= \Biggnorm{  (1-{\tau}_n) \frac{1}{1-\tau_n} \FisherInfo{\gamma^\ast(\tau_n)}^{-1} \sum_{j=0}^m \I{y=j} \frac{\pi_j^\ast}{\pi_j^\train}\big(s_{\gamma^\ast(\tau_n)}(x)  -  \E_{j}[s_{\gamma^\ast(\tau_n)}]\big)  }{2} \\ 
	&\leq \frac{1}{\minSing{\FisherInfo{\gamma^\ast(\tau_n)} }} \sum_{j=0}^m \I{y=j} \frac{\pi_j^\ast}{\pi_j^\train}  \bignorm{ s_{\gamma^\ast(\tau_n)}(x)  -  \E_{j}[s_{\gamma^\ast(\tau_n)}]  }{2} \\ 
	&\leq \frac{1}{ \sqrt{\Lambda} } \frac{1-\xi}{\xi}     \frac{2m^{3/2}}{L}.
\end{align*}
This implies that:
\begin{align*}
	 \frac{{\tau}_n^2}{(1-{\tau}_n)n} \sum_{y=0}^m \pi_y^\train \E_y \Big[  \norm{(1-{\tau}_n) \psi^\eff_{{\tau}_n}(X,y,0)}{2}^4  \Big] &\leq  \frac{1}{(1-{\tau}_n)n}  \frac{1}{ \Lambda^2} \frac{1-\xi}{\xi}     \frac{16m^{6}}{L^4} \to 0,
\end{align*}
where the limit is because of Assumption  \eqref{tau n convergence not too fast}. Overall, it follows that:
\[
	\frac{(1-{\tau}_n)^2}{{\tau}_n n}\E_{\pi^\ast} \Big[  \norm{{\tau}_n \psi^\eff_{{\tau}_n}(X,-1,1)}{2}^4  \Big] + \frac{{\tau}_n^2}{(1-{\tau}_n)n} \sum_{y=0}^m \pi_y^\train \E_y \Big[  \norm{(1-{\tau}_n) \psi^\eff_{{\tau}_n}(X,y,0)}{2}^4  \Big] \to 0,
\]
as desired!

\end{proof}
\vspace{0.4in}


\begin{proof}[\uline{Proof of Lemma \eqref{pi hat is RAL}}]
To prove the desideratum, we will verify the conditions of Theorem 2.2 in \citep{Newey1990}. Since in our semiparametric model each $\theta \equiv (\pi,\mathbf{q}) \in\Theta^\text{semi}$ can be partitioned into a parametric component of interest, $\pi$, and a nonparametric nuisance component, $\mathbf{q}$, it suffices to show the following statements hold:
\begin{itemize}
	\item $\widehat{\pi}_n$ is asymptotically linear with influence function $\psi_\tau^\eff$.
	\item For each regular parametric submodel $\mathbf{M}^{\text{sub},\tau}$ with true parameter $\theta^\ast$,  $\E_{\Jcal^{\theta,\tau}}\big[ \norm{ \psi_\tau^\eff }{2}^2 \big] $ exists and is continuous on a neighborhood of $\theta^\ast$.
	\item For each regular parametric submodel $\mathbf{M}^{\text{sub},\tau}$ with true parameter $\theta^\ast$,  $\E_{\Jcal^{\theta^\ast,\tau}}\big[  \psi_\tau^\eff  S_{\pi^\ast}' \big] = I_{m\times m}$, and $\E_{\Jcal^{\theta^\ast,\tau}}\big[  \psi_\tau^\eff  S_{\rho^\ast}' \big]$ is a matrix of all zeros. \\
\end{itemize}

First, we will prove that the asymptotic linearity condition holds. Recall that the Fixed $\tau$-IID Regime is a special case of the $\tau_n$-IID Regime, wherein $\tau_n = \tau$ for each $n$. Thus, under Assumptions  \eqref{Fixed Tau Regime: H closed under scalar multiplication}, \eqref{Fixed Tau Regime: Omega Properties}, \eqref{Fixed Tau Regime: Cat minus Mixture FIM}, \eqref{Fixed Tau Regime: Mixture FIM not too small}, \eqref{Tau n Regime: lambda and uniform convergence properties} and \eqref{tau n regime, estimator of gamma bar}, we can apply Lemma \eqref{Behavior of First and Second Order Error Terms in tau n IID regime} to the Fixed $\tau$-IID Regime. From this, it follows that:
\[
	\sqrt{n}(\widehat{\pi}_n - \pi^\ast) = \frac{1}{\sqrt{n}} \sum_{i=1}^{n} \psi_\tau^\eff(Z_i) + \delta_{\tau}(Z_{1:n})
\]
where it is easily seen that $\norm{\delta_\tau(Z_{1:n}) }{2} = o_{\P}(1)$. We can also verify that $\E_{\Jcal^{\pi^\ast, \mathbf{p}, \tau}}[  \psi_\tau^\eff] = 0$, as follows:
\small
\begin{align*}
	&\E_{\Jcal^{\pi^\ast, \mathbf{p}, \tau}}[  \psi_\tau^\eff] \\
	&= \E_{\Jcal^{\pi^\ast, \mathbf{p}, \tau}}\bigg[ \frac{D}{\tau} \FisherInfo{\gamma^\ast(\tau)}^{-1}\big(s_{\gamma^\ast(\tau)}(X)  -  \E_{\pi^\ast}[s_{\gamma^\ast(\tau)}]\big) - \frac{1-D}{1-\tau} \FisherInfo{\gamma^\ast(\tau)}^{-1} \sum_{j=0}^m \I{Y=j} \frac{\pi_j^\ast}{\pi_j^\train}\big(s_{\gamma^\ast(\tau)}(X)  -  \E_{j}[s_{\gamma^\ast(\tau)}]\big)  \bigg] \\ 
	&= \FisherInfo{\gamma^\ast(\tau)}^{-1} \E_{ \pi^\ast }\bigg[ s_{\gamma^\ast(\tau)}(X)  -  \E_{\pi^\ast}[s_{\gamma^\ast(\tau)}] \bigg]   -   \FisherInfo{\gamma^\ast(\tau)}^{-1} \sum_{j=0}^m \pi_j^\ast \E_j \bigg[ s_{\gamma^\ast(\tau)}(X)  -  \E_{j}[s_{\gamma^\ast(\tau)}]  \bigg]    \\ 
	&= 0.
\end{align*}
\normalsize
In addition, under Assumptions \eqref{Fixed Tau Regime: Cat minus Mixture FIM}, \eqref{Fixed Tau Regime: Mixture FIM not too small} and \eqref{Fixed Tau Regime: expo family}, we have by Lemma \eqref{Efficiency Bound} that $\Var_{\Jcal^{\pi^\ast,\mathbf{p},\tau_n}}[ \psi_\tau^\eff ] = \mathbf{V}^\eff(\tau)$, the eigenvalues of which are bounded away from $0$ and $\infty$ due to Lemma \eqref{Projection of Ordinary Score onto G}. Thus, we have verified all the conditions necessary for showing that $\widehat{\pi}_n$ is asymptotically linear with influence function $\psi_\tau^\eff$.

Second, we will prove that, if  $\mathbf{M}^{\text{sub},\tau}$ is a regular parametric submodel with nuisance parameter $\rho$, and $\theta^\ast = (\pi^\ast, \rho^\ast)$ is such that $\Jcal^{\theta^\ast,\tau} = \Jcal^{\pi^\ast, \mathbf{p}, \tau}$, then  $\E_{\Jcal^{\theta,\tau}}\big[ \norm{ \psi_\tau^\eff }{2}^2 \big] $ exists and is continuous on a neighborhood of $\theta^\ast$. For any $\theta$, the existence of $\E_{\Jcal^{\theta,\tau}}\big[ \norm{ \psi_\tau^\eff }{2}^2 \big]$ can be established by arguing that $ \norm{ \psi_\tau^\eff }{2} $ is a bounded function. To see this, note that $\gamma^\ast(\tau) \in \Gamma$ by Lemma \eqref{Bounded Gamma Star}, and so by Assumption \eqref{Fixed Tau Regime: Mixture FIM not too small} we have that $\maxSing{ \FisherInfo{\gamma^\ast(\tau) }^{-1 }} < 1/\sqrt{\Lambda} < \infty$. Further, the fact that $\gamma^\ast(\tau) \in \Gamma$ also implies that  $\norm{s_{\gamma^\ast(\tau)}}{2}$ is a bounded function. Thus, by inspecting the definition of $\psi_\tau^\eff$, it follows that $ \norm{ \psi_\tau^\eff }{2} $ is indeed a bounded function, and so  $\E_{\Jcal^{\theta,\tau}}\big[ \norm{ \psi_\tau^\eff }{2}^2 \big]$ exists. As for the continuity of $\E_{\Jcal^{\theta,\tau}}\big[ \norm{ \psi_\tau^\eff }{2}^2 \big]$ on a neighborhood of $\theta^\ast$, this follows from an application of the dominated convergence theorem and the definition of regularity in Appendix A of \citep{Newey1990}, which ensures that $\Jcal^{\theta,\tau}(z)$ is continuous for every $\theta$ and at almost all $z$.

Third, we need to prove that, if if  $\mathbf{M}^{\text{sub},\tau}$ is a regular parametric submodel with nuisance parameter $\rho$, and $\theta^\ast = (\pi^\ast, \rho^\ast)$ is such that $\Jcal^{\theta^\ast,\tau} = \Jcal^{\pi^\ast, \mathbf{p}, \tau}$, then $\E_{\Jcal^{\theta^\ast,\tau}}\big[  \psi_\tau^\eff  S_{\pi^\ast}' \big] = I_{m\times m}$, and $\E_{\Jcal^{\theta^\ast,\tau}}\big[  \psi_\tau^\eff  S_{\rho^\ast}' \big]$ is a matrix of all zeros. Towards that end, first observe that:
\begin{align*}
	\E_{\Jcal^{\theta^\ast,\tau}}\big[  \psi_\tau^\eff  S_{\pi^\ast}' \big] &= \tau \E_{ \pi^\ast }\big[  \psi_\tau^\eff(X,-1,1)  s_{\pi^\ast}(X)' \big]   \\  
	 &= \tau \E_{ \pi^\ast }\bigg[  \frac{1}{\tau} \FisherInfo{\gamma^\ast(\tau)}^{-1}\big(s_{\gamma^\ast(\tau)}(X)  -  \E_{\pi^\ast}[s_{\gamma^\ast(\tau)}]\big)   s_{\pi^\ast}(X)' \bigg]   \\  
	 &=  \FisherInfo{\gamma^\ast(\tau)}^{-1} \E_{ \pi^\ast }\bigg[  \big(s_{\gamma^\ast(\tau)}(X)  -  \E_{\pi^\ast}[s_{\gamma^\ast(\tau)}]\big)   s_{\pi^\ast}(X)' \bigg]   \\  
	 &=  \FisherInfo{\gamma^\ast(\tau)}^{-1} \Cov_{ \pi^\ast }\bigg[  s_{\gamma^\ast(\tau)}(X) ,  s_{\pi^\ast}(X) \bigg]   \\  
	 &=  \FisherInfo{\gamma^\ast(\tau)}^{-1} A_{s_{\gamma^\ast(\tau)}}   \\  
	 &=  \FisherInfo{\gamma^\ast(\tau)}^{-1}  \FisherInfo{\gamma^\ast(\tau)}  \\  
	 &= I_{m\times m},
\end{align*}
where the fifth and sixth lines are due to Lemma \eqref{Moment Identity}. Also observe that:
\begin{align*}
	&\E_{\Jcal^{\theta^\ast,\tau}}\big[  \psi_\tau^\eff  S_{\rho^\ast}' \big] \\
	&= \E_{\Jcal^{\theta^\ast,\tau}}\bigg[  \psi_\tau^\eff  S_{\rho^\ast}' \bigg] \\ 
	&=  \E_{\Jcal^{\theta^\ast,\tau}}\Bigg[ D \psi_\tau^\eff(X,-1,1)  S_{\rho^\ast}(X,-1,1)'   +   (1-D) \sum_{j=0}^m \I{Y = j} \psi_\tau^\eff(X,j,0)  S_{\rho^\ast}(X,j,0)' \Bigg].
\end{align*}
Note that:
\begin{align*}
	& \E_{\Jcal^{\theta^\ast,\tau}}\Bigg[ D \psi_\tau^\eff(X,-1,1)  S_{\rho^\ast}(X,-1,1)'  \Bigg] \\
	&= \FisherInfo{\gamma^\ast(\tau)}^{-1}\E_{\pi^\ast}\Bigg[ \big(s_{\gamma^\ast(\tau)}(X)  -  \E_{\pi^\ast}[s_{\gamma^\ast(\tau)}]\big) \sum_{j=0}^m \pi_j^\ast \frac{p_j(X)}{p_{\pi^\ast}(X) } \partialDerivative{\rho}\big\{ \log q_j^\rho(X) \big\}\big\lvert_{\rho = \rho^\ast}  '  \Bigg] \\
	 &= \FisherInfo{\gamma^\ast(\tau)}^{-1} \sum_{j=0}^m \pi_j^\ast \E_{\pi^\ast}\Bigg[  \frac{p_j(X)}{p_{\pi^\ast}(X) } \big(s_{\gamma^\ast(\tau)}(X)  -  \E_{\pi^\ast}[s_{\gamma^\ast(\tau)}]\big) \partialDerivative{\rho}\big\{ \log q_j^\rho(X) \big\}\big\lvert_{\rho = \rho^\ast}'  \Bigg] \\
	 &= \FisherInfo{\gamma^\ast(\tau)}^{-1} \sum_{j=0}^m \pi_j^\ast \E_{j}\Bigg[  \ \big(s_{\gamma^\ast(\tau)}(X)  -  \E_{\pi^\ast}[s_{\gamma^\ast(\tau)}]\big) \partialDerivative{\rho}\big\{ \log q_j^\rho(X) \big\}\big\lvert_{\rho = \rho^\ast}'  \Bigg] \\ 
	 &= \FisherInfo{\gamma^\ast(\tau)}^{-1} \sum_{j=0}^m \pi_j^\ast \E_{j}\Bigg[  s_{\gamma^\ast(\tau)}(X) \partialDerivative{\rho}\big\{ \log q_j^\rho(X) \big\}\big\lvert_{\rho = \rho^\ast}'  \Bigg] \\ 
\end{align*}
and that:
\begin{align*}
	  & \E_{\Jcal^{\theta^\ast,\tau}}\Bigg[  (1-D) \sum_{j=0}^m \I{Y = j} \psi_\tau^\eff(X,j,0)  S_{\rho^\ast}(X,j,0)' \Bigg]\\
	   &=  (1-\tau) \sum_{j=0}^m \pi_j^\train  \E_{j}\Bigg[ \psi_\tau^\eff(X,j,0)  S_{\rho^\ast}(X,j,0)' \Bigg] \\ 
	   &=   \FisherInfo{\gamma^\ast(\tau)}^{-1}  \sum_{j=0}^m \pi_j^\ast \E_{j}\Bigg[ \big(s_{\gamma^\ast(\tau)}(X)  -  \E_{j}[s_{\gamma^\ast(\tau)}]\big)  S_{\rho^\ast}(X,j,0)' \Bigg] \\ 
	   &=   \FisherInfo{\gamma^\ast(\tau)}^{-1}  \sum_{j=0}^m \pi_j^\ast \E_{j}\Bigg[ \big(s_{\gamma^\ast(\tau)}(X)  -  \E_{j}[s_{\gamma^\ast(\tau)}]\big)  \partialDerivative{\rho}\big\{ \log q_j^\rho(X) \big\}\big\lvert_{\rho = \rho^\ast}' \Bigg] \\ 
 	&=   \FisherInfo{\gamma^\ast(\tau)}^{-1}  \sum_{j=0}^m \pi_j^\ast \E_{j}\Bigg[ s_{\gamma^\ast(\tau)}(X)  \partialDerivative{\rho}\big\{ \log q_j^\rho(X) \big\}\big\lvert_{\rho = \rho^\ast}' \Bigg].
\end{align*}

Thus, we have that
\[
	 \E_{\Jcal^{\theta^\ast,\tau}}\Bigg[ D \psi_\tau^\eff(X,-1,1)  S_{\rho^\ast}(X,-1,1)'  \Bigg]  =  \E_{\Jcal^{\theta^\ast,\tau}}\Bigg[  (1-D) \sum_{j=0}^m \I{Y = j} \psi_\tau^\eff(X,j,0)  S_{\rho^\ast}(X,j,0)' \Bigg], 
\]
and so it must be that $\E_{\Jcal^{\theta^\ast,\tau}}\big[  \psi_\tau^\eff  S_{\rho^\ast}' \big]$ is a matrix of all zeros. Thus, we have shown that all the conditions in Theorem 2.2 in \citep{Newey1990} are satisfied, and so $\widehat{\pi}_n$ is indeed regular.




\end{proof}
\vspace{0.4in}

\bibliography{MyBib}

\end{document}